\documentclass{article}

\usepackage[preprint]{neurips_2026}

\usepackage[utf8]{inputenc} 
\usepackage[T1]{fontenc}    
\usepackage{hyperref}       
\usepackage{url}            
\usepackage{booktabs}       
\usepackage{amsfonts}       
\usepackage{nicefrac}       
\usepackage{microtype}      

\usepackage{microtype}
\usepackage{graphicx} 
\usepackage{subcaption}
\usepackage{booktabs} 
\usepackage{multirow}

\usepackage[table]{xcolor} 
\usepackage{amsmath,amssymb}
\usepackage{array}
\usepackage{makecell}
\usepackage{float}
\usepackage{standalone}
\usepackage{subcaption}
\usepackage{style}
\usepackage{enumitem}
\usepackage{wrapfig}
\usepackage{physics}
\newcommand{\1}{\mathbf{1}}
\usepackage{makecell}

\usepackage[most]{tcolorbox}
\tcbset{
  colback=blue!5!white,   
  colframe=blue!50!black, 
  arc=6pt,                
  boxrule=0.8pt,          
  left=6pt, right=6pt, top=6pt, bottom=6pt
}

\newtcolorbox{findingbox}{
  title=\textbf{Desideratum},
  colback=blue!5!white,
  colframe=blue!50!black,
  fonttitle=\bfseries,
  coltitle=black
}

\newtcolorbox{findingblue}{
  title=\textbf{Desideratum},
  colback=blue!5!white,
  colframe=blue!50!black,
  arc=6pt,
  boxrule=0.8pt,
  enhanced,
  sharp corners=all, 
  boxshadow=0.5pt
}

\newtcolorbox{findinggreen}{
  title=\textbf{Maximal Scale Stability Desideratum},
  colback=green!5!white,
  colframe=green!50!black,
  arc=6pt,
  boxrule=0.8pt
}

\newtcolorbox{findinggreenA}{
  colback=green!8!white,
  colframe=green!60!black,
  arc=4pt,
  boxrule=0.6pt,
  left=5pt, right=5pt, top=3pt, bottom=3pt,
  boxsep=0pt,
  before skip=6pt,
  after skip=6pt
}

\definecolor{forestbg}{RGB}{236,245,240}
\definecolor{forestframe}{RGB}{40,90,70}

\newtcolorbox{findinggreenB}{
  colback=forestbg,
  colframe=forestframe,
  arc=4pt,
  boxrule=0.6pt,
  left=5pt, right=5pt, top=3pt, bottom=3pt,
  boxsep=0pt,
  before skip=5pt,
  after skip=5pt
}

\definecolor{slatebluebg}{RGB}{238,242,248}
\definecolor{slateblueframe}{RGB}{55,80,120}

\newtcolorbox{findingblueA}{
  colback=blue!6!white,
  colframe=blue!65!black,
  arc=4pt,
  boxrule=0.6pt,
  left=5pt, right=5pt, top=3pt, bottom=3pt,
  boxsep=0pt,
  before skip=5pt,
  after skip=5pt
}

\newtcolorbox{findingblueB}{
  colback=slatebluebg,
  colframe=slateblueframe,
  arc=4pt,
  boxrule=0.6pt,
  left=5pt, right=5pt, top=3pt, bottom=3pt,
  boxsep=0pt,
  before skip=5pt,
  after skip=5pt
}

\newtcolorbox{findingblueC}{
  colback=blue!10!white,
  colframe=blue!80!black,
  arc=3pt,
  boxrule=0.7pt,
  left=4pt, right=4pt, top=2pt, bottom=2pt,
  boxsep=0pt,
  before skip=4pt,
  after skip=4pt
}

\newtcolorbox{findingredA}{
  colback=red!6!white,
  colframe=red!65!black,
  arc=4pt,
  boxrule=0.6pt,
  left=5pt, right=5pt, top=3pt, bottom=3pt,
  boxsep=0pt,
  before skip=5pt,
  after skip=5pt
}

\definecolor{winebg}{RGB}{248,240,242}
\definecolor{wineframe}{RGB}{110,50,60}

\newtcolorbox{findingredB}{
  colback=winebg,
  colframe=wineframe,
  arc=4pt,
  boxrule=0.6pt,
  left=5pt, right=5pt, top=3pt, bottom=3pt,
  boxsep=0pt,
  before skip=5pt,
  after skip=5pt
}

\newtcolorbox{findingredC}{
  colback=red!10!white,
  colframe=red!80!black,
  arc=3pt,
  boxrule=0.7pt,
  left=4pt, right=4pt, top=2pt, bottom=2pt,
  boxsep=0pt,
  before skip=4pt,
  after skip=4pt
}

\newtcolorbox{findinggreyA}{
  colback=black!6!white,
  colframe=black!65!white,
  arc=4pt,
  boxrule=0.6pt,
  left=5pt, right=5pt, top=3pt, bottom=3pt,
  boxsep=0pt,
  before skip=5pt,
  after skip=5pt
}

\definecolor{coolgreybg}{RGB}{242,244,247}
\definecolor{coolgreyframe}{RGB}{90,100,115}

\newtcolorbox{findinggreyB}{
  colback=coolgreybg,
  colframe=coolgreyframe,
  arc=4pt,
  boxrule=0.6pt,
  left=5pt, right=5pt, top=3pt, bottom=3pt,
  boxsep=0pt,
  before skip=5pt,
  after skip=5pt
}

\newtcolorbox{findinggreyC}{
  colback=black!10!white,
  colframe=black!80!white,
  arc=3pt,
  boxrule=0.7pt,
  left=4pt, right=4pt, top=2pt, bottom=2pt,
  boxsep=0pt,
  before skip=4pt,
  after skip=4pt
}

\newtcolorbox{finding}{
  colback=blue!5!white,
  colframe=blue!50!black,
  arc=4pt,              
  boxrule=0.6pt,        
  left=4pt, right=4pt, top=3pt, bottom=3pt, 
  boxsep=0pt
}

\newtcbox{\findinginline}{
  colback=blue!5!white,
  colframe=blue!50!black,
  arc=3pt,
  boxrule=0.5pt,
  boxsep=1pt,
  left=2pt, right=2pt, top=1pt, bottom=1pt
}

\renewcommand{\i}{\text{in}}
\renewcommand{\o}{\text{out}}

\usepackage{hyperref}

\usepackage[page]{appendix}

\renewcommand{\appendixtocname}{Appendix Contents.}

\makeatletter
\let\oldappendix\appendices

\renewcommand{\appendices}{%
  \clearpage
  \renewcommand{\thesection}{\Roman{section}}
  \let\tf@toc\tf@app
  \addtocontents{app}{\protect\setcounter{tocdepth}{2}}
  \immediate\write\@auxout{%
    \string\let\string\tf@toc\string\tf@app^^J
  }
  \oldappendix
}%

\newcommand{\listofappendices}{%
  \begingroup
  \renewcommand{\contentsname}{\appendixtocname}
  \let\@oldstarttoc\@starttoc
  \def\@starttoc##1{\@oldstarttoc{app}}
  \tableofcontents
  \endgroup
}

\makeatother

\newcommand{\eps}{\varepsilon}
\newcommand{\fanin}{\texttt{fan\_in}}

\usepackage{amsmath}
\usepackage{amssymb}
\usepackage{mathtools}
\usepackage{amsthm}

\usepackage{bbold}
\usepackage{physics}
\usepackage{bm}
\usepackage{braket}
\newcommand*\diff{\mathop{}\!\mathrm{d}}

\usepackage[capitalize,noabbrev]{cleveref}

\theoremstyle{plain}
\newtheorem{theorem}{Theorem}[section]
\newtheorem{proposition}[theorem]{Proposition}
\newtheorem{lemma}[theorem]{Lemma}

\theoremstyle{definition}
\newtheorem{definition}[theorem]{Definition}

\theoremstyle{remark}

\usepackage[textsize=tiny]{todonotes}

\setlist[itemize]{leftmargin=*}
\setlist[enumerate]{leftmargin=*}

\definecolor{darkforestgreen}{HTML}{15803d} 
\definecolor{mygreen}{RGB}{0,180,0}

\definecolor{correctscaling}{named}{black}
\newcommand{\w}[1]{\textcolor{darkforestgreen}{#1}}

\hypersetup{colorlinks,
    linkcolor={blue!50!black},
    citecolor={blue!50!black},
    urlcolor={blue!50!black}
}

\title{How to Scale Mixture-of-Experts: From \texorpdfstring{$\mu$}{mu}P to\\ the Maximally Scale-Stable Parameterization}

 \definecolor{myPink}{HTML}{FF1493}         

\author{
Leena Chennuru Vankadara$^{1,}$\thanks{Equal first author. $^\dagger$Equal contributor. Correspondence to: \texttt{l.vankadara@ucl.ac.uk}. \\ \phantom{Equ} This work does not relate to MH's position at Amazon.} \quad Moritz Haas$^{2,*}$ \\ 
\phantom{.}\\
\textbf{Luke Hayward}$^{1,\dagger}$ \quad \textbf{Sebastian Bordt}$^{3,\dagger}$ \quad \textbf{Alessandro Breccia}$^{1,\dagger}$ \\
\phantom{.}\\
$^1$Gatsby Computational Neuroscience Unit, University College London \\ %
\phantom{.}\\
$^2$Amazon \qquad $^3$University of Tübingen, Tübingen AI Center
}

\begin{document}
\maketitle

\begin{abstract}

Recent frontier large language models predominantly rely on Mixture-of-Experts (MoE) architectures. Despite empirical progress, there is still no principled understanding of how hyperparameters should scale with network width $N$, expert width $N_e$, number of experts $M$, sparsity $K$, and depth $L$ to ensure both stability and optimal performance at scale. We take a principled step toward resolving this gap by analyzing three different scaling regimes: (I) co-scaling $N\asymp N_e$, (II) co-scaling $N\asymp M\asymp K$, and (III) full proportional scaling of $N, N_e, M$, and $K$. For each regime, we develop a novel Dynamical Mean Field Theory (DMFT) description of the limiting training dynamics of MoEs that provides a formal foundation for our analysis. Within this framework, we derive the unique parameterization for SGD and Adam satisfying all maximal-update ($\mu$) desiderata. We then show that the resulting $\mu$P prescription does not reliably induce monotonic improvement with scale or robust learning-rate transfer. We trace these pathologies to scale-dependent observables in the aggregation dynamics, which motivates a refined set of desiderata that we term \textit{maximal scale stability}. Guided by this principle, we derive a \textit{Maximally Scale-Stable Parameterization} (MSSP) for both SGD and Adam in all three scaling regimes, and characterize the corresponding limiting dynamics - qualitatively distinct from the $\mu$P limit - through a separate DMFT analysis. Experiments verify that MSSP robustly recovers learning rate transfer and monotonic improvement with scale across regimes. Combined with existing depth-scaling theory, these results provide a complete scaling prescription for MoE architectures as a function of width, depth, expert width, and number of experts.

\end{abstract}

\section{Introduction}

Scaling the number of parameters consistently improves network quality \citep{kaplan2020scaling,hoffmann2022empirical}, yet the cost of architectural and hyperparameter tuning grows alongside it. At the frontier, where models exceed hundreds of billions of parameters \citep{brown2020language, guo2025deepseek}, even a single training run makes iterative tuning infeasible. An influential line of work proposes to design parameterizations under
which finite-size networks can be regarded as discretizations of a well-defined
infinite-size limit
\citep{yang_feature_2021, bordelon2022self, bordelon2023depthwise} - specifically,
parameterizations under which feature and prediction dynamics remain
approximately invariant across scales and finite networks
closely track the limiting object. Under such parameterizations, smaller
models serve as faithful proxies for larger ones: optimal hyperparameters, such as
learning rates, transfer across scales
\citep{tp5_2022, everett2024scaling, dey2025don}, and performance improves
predictably with scale \citep{tp5_2022, dey2025don}. The Maximal Update
Parameterization ($\mu$P) is the canonical instance for width
\citep{tp5_2022}, with analogous prescriptions developed for depth
\citep{bordelon2023depthwise, bordelon2024infinite, dey2025don}.\looseness=-1

\begin{figure}[t]
    \centering
    \begin{subfigure}[b]{0.99\textwidth}
    \centering
    \includegraphics[width=\textwidth]{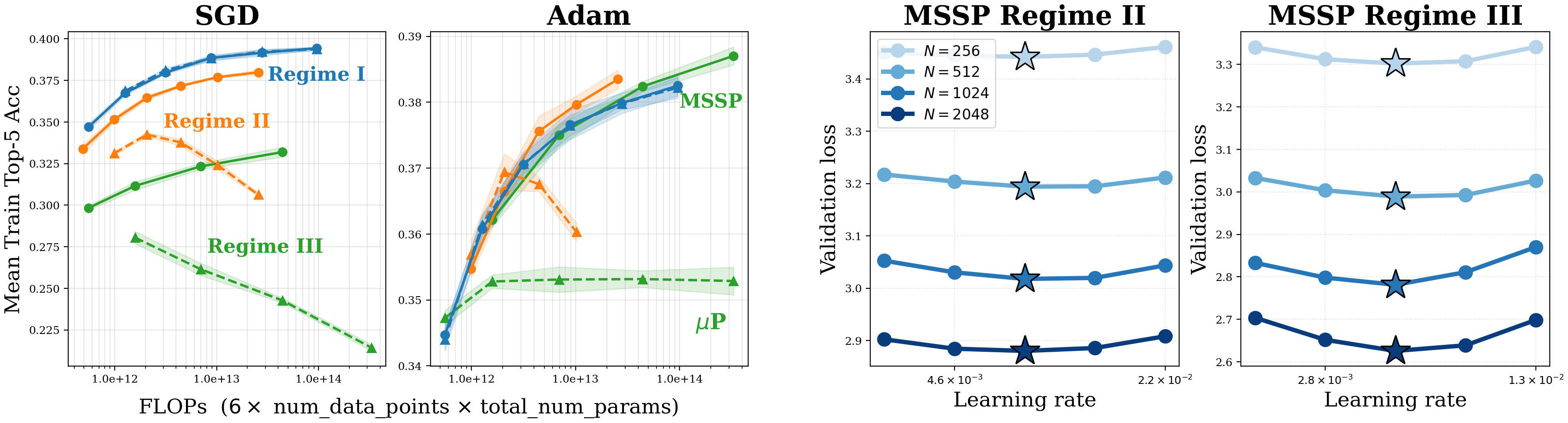}
    \end{subfigure}
    \caption{\textbf{$\mu$P does not reliably improve performance with scale in MoEs. MSSP recovers monotonic improvement and delivers LR transfer across MoE co-scaling regimes.} 
    \emph{Left: Across optimizers and regimes, MSSP (solid lines) outperforms $\mu$P (dashed lines) at large scale for MLP MoEs on TinyImageNet. Right:} LR transfer in validation loss for GPT MoEs trained with Adam in MSSP for 2.5B tokens when co-scaling width and number of (active) experts, with fixed expert width (Regime II) or co-scaled expert width (Regime III). 
    All details in App.~\ref{sec:fig_details}.}
    \label{fig:performance_transfer_main1}
\end{figure}

While scale-invariant parametrizations for dense models are well-developed, modern frontier large language models
increasingly employ Mixture-of-Experts (MoE) layers
\citep{jiang2024mixtral, guo2025deepseek}, which decouple parameter count
from computational cost via sparse routing \citep{shazeer2017}. In doing so, they introduce additional scaling axes that must be coordinated
jointly with width $N$ and depth $L$: \textit{the number of experts $M$, the
expert width $N_e$, and the routing sparsity $K/M$}. Recent engineering advances
render scaling along all of these dimensions practically tractable
\citep{nvidiaMoEMegatron2026}, and a growing body of evidence indicates that
increasingly fine-grained experts outperform configurations comprising a few
large experts
\citep{pmlr-v162-clark22a, krajewski2024scaling, dai2024deepseekmoe}. Yet, little is known about how to optimally scale 
fine-grained, sparse MoE architectures, which motivates the question that this paper addresses: \looseness-1

\begin{center}
    \textit{How should MoE architectures and training hyperparameters be
    scaled to yield scale-invariant, non-degenerate
    feature and prediction dynamics in various co-scaling regimes of
    $M, N, N_e,$ and $K$?}
\end{center}

As we will see in this work, developing scaling theories for MoEs is substantially more
involved than for dense networks. One cannot generally assume commutativity across axes ($M, N, N_e, K$); instead, joint limits governed by their relative rates must be analyzed. Deriving such limits is technically challenging since routing and aggregation in MoEs couple the dynamics
of the router, the experts, and update statistics
\citep{shazeer2017, fedus2022switch}. 
This leads to a combinatorial space of co-scaling regimes, each, as we demonstrate, exhibit qualitatively distinct behavior.

\textbf{Main Contributions.} We study three co-scaling regimes for MoEs: (Regime I) $N \asymp N_e \to \infty$ with $M, K$ fixed; (Regime II) $N \asymp M \asymp K \to \infty$ with $N_e$ fixed - the fine-grained / bottleneck regime increasingly favored in modern MoE designs; and (Regime III) joint scaling $N \asymp N_e \asymp M \asymp K \to \infty$. Across these regimes, we make the following contributions:

\begin{itemize}
    \item \textbf{$\mu$P for MoEs across regimes and optimizers.} In each scaling regime, we derive the parameterizations that satisfy the $\mu$P desiderata for SGD and Adam via signal propagation analyses which are formally justified by a novel Dynamical Mean Field Theory (DMFT) for each scaling regime.
    
    \item \textbf{Scale-dependence of $\mu$P in MoEs.} Despite formally satisfying the $\mu$P desiderata, we find that $\mu-$parameterizations do not reliably deliver learning-rate transfer or monotonic improvement with scale in MoEs (Fig.~\ref{fig:performance_transfer_main1}, \ref{fig:mlp_transfer_sgd} and \ref{fig:llm_lr_sweep_main}). We trace this to scale-dependence in the training dynamics.
    
    \item \textbf{Maximally Scale-Stable Parameterization (MSSP).} We propose \emph{maximal scale stability} as a refined principle generalizing $\mu$P desiderata, and derive the Maximally Scale-Stable Parameterization (MSSP) for SGD and Adam in each regime. The required corrections to $\mu$P are structurally distinct across regimes: zero router initialization in Regime I, amplified expert-output initialization variance ($1/N_e$ to $M/N_e$) in Regime II, and shared expert weights at initialization in Regime III.
    
    \item \textbf{DMFT under MSSP.} We derive a self-consistent DMFT for the limiting training dynamics in each regime under MSSP, including a four-level conditional mean-field hierarchy in Regime III induced by the shared expert initialization, qualitatively distinct from the limiting dynamics under $\mu$P.
    
    \item \textbf{Empirical validation.} We verify on MLP and Transformer MoEs that MSSP robustly recovers learning-rate transfer, and restores monotonic improvement with scale across all three regimes. We provide a complete scaling prescription for modern Transformer MoE architectures along width $N$, depth $L$, number of experts $M$, expert width $N_e$ and number of active experts $K$.
\end{itemize}

Overall, this paper takes important steps towards stable, predictable and optimal MoE scaling that preserves 
reliable hyperparameter transfer and monotonic improvement with increasing network size.

\textbf{Independent and concurrent work.}  \citet{jiang2026hyperparameter} derive a signSGD parameterization under Regime III via DMFT and present Transformer MoE experiments showing approximate LR transfer along individual scaling axes.
We derive $\mu$P for \emph{SGD and Adam across Regimes I, II, and III} using a signal propagation analysis complemented by a DMFT for each regime, and our experiments scale jointly along the axes each regime prescribes. {In Regime III, our $\mu$P for Adam coincides with theirs for signSGD \emph{up to $\epsilon$ scaling and a larger router initialization.}} Crucially, we identify that in Regimes II and III, $\mu$P does not reliably yield learning rate transfer or monotonic improvement with scale in MoEs and propose MSSP as a resolution. In Regime III, our DMFT for MSSP is qualitatively distinct: it exhibits a four-layer mean-field hierarchy (versus a three-layer hierarchy under $\mu$P), and we additionally derive novel DMFT limits for Regimes I and II under MSSP. See \Cref{sec:related_work_app} for extended discussion.\looseness=-1

\begin{figure}[htbp!]
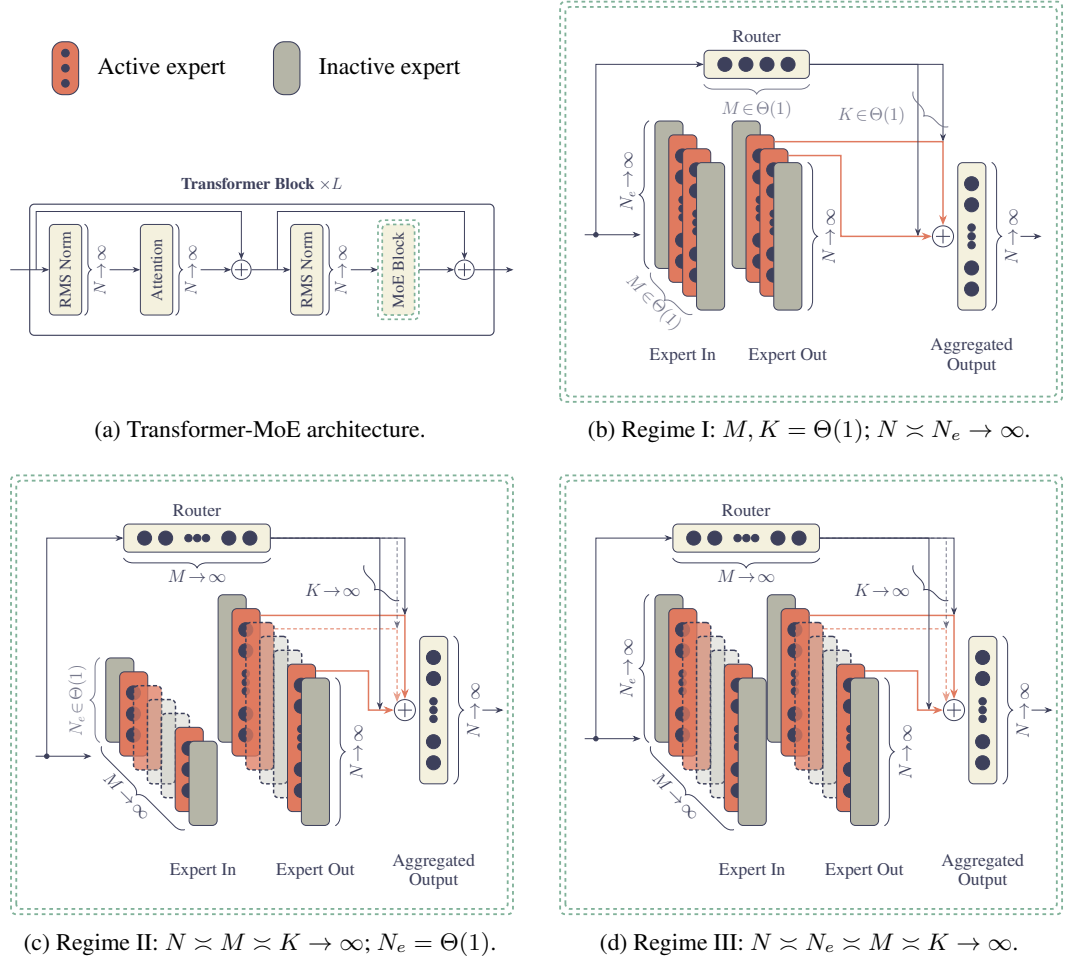

  \centering
  \vspace{-2mm}
  \begin{subfigure}[b]{0.48\textwidth}
    \centering
    \begin{tikzpicture}[
        miniexpert/.style={draw=OutputNeuronFill, rectangle, rounded corners, minimum width=0.35cm, minimum height=0.7cm, inner sep=0pt},
        minineuron/.style={circle, fill=OutputNeuronFill, draw=OutputNeuronFill, minimum size=0.1cm, inner sep=0pt}
      ]
      \node[miniexpert, fill=ExpertHighlight] (active) {};
      \foreach \y in {0.2, 0, -0.2} {
        \node[minineuron] at ($(active.center) + (0, \y)$) {};
      }
      \node[right=0.12cm of active, font=\small] (active_label) {Active expert};
      \node[miniexpert, fill=DisabledColor, right=0.5cm of active_label] (inactive) {};
      \node[right=0.12cm of inactive, font=\small] {Inactive expert};
    \end{tikzpicture}
    \\[1cm]
    \resizebox{\textwidth}{!}{
\colorlet{strokeNavy}{blockborder}

\begin{tikzpicture}[
    every path/.append style={draw=strokeNavy},
    every node/.append style={text=strokeNavy},
    block/.style={
      rectangle, draw=strokeNavy, fill=blockfill,
      rounded corners,
      minimum height=2.8cm, minimum width=1cm,
      font=\Large,
    },
    arrow/.style={->, >=Stealth, thick, draw=strokeNavy},
    add/.style={circle, draw=strokeNavy, thick, minimum size=0.6cm, font=\Huge},
    container/.style={draw=strokeNavy, line width=1pt, rounded corners}
  ]

  \coordinate (input) at (0,0);

  \node[block, right=1.2cm of input] (rms1) {\rotatebox{90}{RMS Norm}};
  \node[block, right=1.8cm of rms1] (attention) {\rotatebox{90}{Attention}};

  \node[add, right=1.8cm of attention] (add1) {};
  \draw[thick] ($(add1.center) + (-0.2, 0)$) -- ($(add1.center) + (0.2, 0)$);
  \draw[thick] ($(add1.center) + (0, -0.2)$) -- ($(add1.center) + (0, 0.2)$);

  \node[block, right=1.2cm of add1] (rms2) {\rotatebox{90}{RMS Norm}};
  \node[block, draw=containerline, dashed, line width=1.5pt, right=1.8cm of rms2] (moe) {\rotatebox{90}{MoE Block}};
  \draw[draw=containerline, dashed, line width=1.5pt, rounded corners]
    ($(moe.north west) + (-3pt, 3pt)$) rectangle ($(moe.south east) + (3pt, -3pt)$);

  \node[add, right=1.2cm of moe] (add2) {};
  \draw[thick] ($(add2.center) + (-0.2, 0)$) -- ($(add2.center) + (0.2, 0)$);
  \draw[thick] ($(add2.center) + (0, -0.2)$) -- ($(add2.center) + (0, 0.2)$);

  \coordinate[right=1.2cm of add2] (output);

  \draw[decorate, thick, decoration={brace, amplitude=6pt, raise=0.5ex}]
  (rms1.north east) -- (rms1.south east)
  node[midway, xshift={1em + 2mm}, rotate=90, anchor=center] (rms1_label) {\Large $N\!\to\!\infty$};

  \draw[decorate, thick, decoration={brace, amplitude=6pt, raise=0.5ex}]
  (attention.north east) -- (attention.south east)
  node[midway, xshift={1em + 2mm}, rotate=90, anchor=center] (attention_label) {\Large $N\!\to\!\infty$};

  \draw[decorate, thick, decoration={brace, amplitude=6pt, raise=0.5ex}]
  (rms2.north east) -- (rms2.south east)
  node[midway, xshift={1em + 2mm}, rotate=90, anchor=center] (rms2_label) {\Large $N\!\to\!\infty$};

  \draw[arrow] (input) -- (rms1);
  \draw[arrow] (rms1_label.south) -- (attention);
  \draw[arrow] (attention_label.south) -- (add1);
  \draw[arrow] (add1) -- (rms2);
  \draw[arrow] (rms2_label.south) -- ($(moe.west) + (-3pt, 0)$);
  \draw[arrow] ($(moe.east) + (3pt, 0)$) -- (add2);
  \draw[arrow] (add2) -- (output);

  \draw[arrow] ($(rms1.west) + (-0.4,0)$) -- ++(0,1.8) -| (add1.north);
  \draw[arrow] ($(rms2.west) + (-0.4,0)$) -- ++(0,1.8) -| (add2.north);

  \node[container, fit={(rms1) (attention) (add1) (rms2) (moe) (add2)}, inner sep=0.6cm,
    label={[font=\Large\bfseries, yshift=0.3cm]above:Transformer Block $\times L$}] (container) {};

\end{tikzpicture}}
    \vspace{0.5cm}
    \caption{Transformer-MoE architecture.}
  \end{subfigure}
  \hfill
  \begin{subfigure}[b]{0.48\textwidth}
    \centering
    \resizebox{\textwidth}{!}{\input{tikz_files/moE_block_regime_i_dotted.tex}}
    \caption{Regime I: $M, K = \Theta(1)$; $N \asymp N_e \to \infty$.}
  \end{subfigure}
  \\[1em]
  \begin{subfigure}[b]{0.48\textwidth}
    \centering
    \resizebox{\textwidth}{!}{\input{tikz_files/moE_block_regime_ii_dotted.tex}}
    \caption{Regime II: $N \asymp M \asymp K \to \infty$; $N_e = \Theta(1)$.}
  \end{subfigure}
  \hfill
  \begin{subfigure}[b]{0.48\textwidth}
    \centering
    \resizebox{\textwidth}{!}{\input{tikz_files/moE_block_regime_iii_dotted.tex}}
    \caption{Regime III: $N \asymp N_e \asymp M \asymp K \to \infty$.}
  \end{subfigure}
  \caption{Transformer-MoE architecture and the MoE block under the three scaling regimes.}
\end{figure}

\section{Setting: Architecture and Scaling Regimes}\label{sec:setup}

\textbf{MoE architecture.} Given input $x_t\in\mathbb{R}^D$, the residual stream $h^0_t = W^{\mathrm{in}}_t x_t$ is transformed as:
\begin{equation*}
    h^{l}_t = h^{l-1}_t + K^{-\alpha_{\mathrm{agg}}}\textstyle\sum_{i\in\text{top-K}}\phi^l_{t,i}\cdot W^{l,\mathrm{out},i}_t\,\varphi(W^{l,\mathrm{in},i}_t h^{l-1}_t),
    \qquad \phi^l_{t,i} = \sigma\!\big(\beta\,(Q^l_t h^{l-1}_t)_i\big), 
\end{equation*}
for $l\in[L]$, and is finally transformed into output logits $f_t = (W_t^{\mathrm{out}})^\top h^L_t$. Here $\varphi$ is a coordinatewise nonlinearity, $\beta$ a tunable inverse temperature, and $\sigma$ either sigmoid or softmax.
Our scaling prescriptions are \emph{local} to the MoE block and apply unchanged when stacked with attention, residual connections, and normalization (with non-MoE components parameterized via $\mu$P/mean-field), as validated by our Transformer MoE experiments in \Cref{sec:lr_sweeps_llm,sec:rcc_llm}.

\textbf{Coordinate-wise update rules.} We consider optimizers of the form $W_t = W_{t-1} - \eta_t\,\Psi_t(g_0,\ldots,g_t)$ with $g_t = \nabla_W\mathcal{L}_t$ and $\Psi_t$ acting entrywise \citep{yang_tp4b_2023}; this covers SGD and Adam.

\textbf{Scaling axes, regimes, and parameterizations}
We study joint scaling in embedding width $N$, expert width $N_e$, number of experts $M$, and active experts $K$ ($1\le K\le M$). All $\Theta,\mathcal{O},\Omega$ are taken as $n\to\infty$ along a trajectory $\mathcal{S}(n)=(N,N_e,M,K)(n)$. We organize results by three regimes: (I) \emph{fixed number of experts}: $M,K=\Theta(1)$ and $N\asymp N_e \to \infty$; (II) \emph{infinite number of experts with fixed expert width}: $N_e=\Theta(1)$ and $N\asymp M\asymp K \to \infty$; (III) \emph{joint proportional scaling}: $N\asymp N_e\asymp M\asymp K\to \infty$. Depth requires separate analytical treatment; we defer depth scaling to the end of Section \ref{sec:mssp}. Unless stated otherwise, we treat all other quantities as fixed $\Theta(1)$ quantities.

\textbf{$bcd\alpha$-parametrization.} A \emph{$bcd\alpha$-parametrization} fixes exponents $(b_W,c_W,d_W)$ for each trainable tensor $W$ and an aggregation exponent $\alpha_{\mathrm{agg}}\in [0,1]$ such that $W_0 \sim \mathcal{N}(0,n^{-2b_W})$,
\begin{equation*}
    W_t = W_{t-1} - \eta\, n^{-c_W}\,\Psi_t\!\big(n^{d_W} g_0(W),\ldots,n^{d_W} g_t(W)\big),
    \qquad
    h^l_t = n^{-\alpha_{\mathrm{agg}}}\textstyle\sum_{i=1}^{M}\phi^l_{t,i}\,h^{l,\mathrm{out}}_{t,i},
\end{equation*}
with $\Psi_t$ applied entrywise. Our goal is to identify a set of exponents $\{\alpha_{\mathrm{agg}}\}\cup\{b_W,c_W,d_W\}_{W}$ that yields stable, performant dynamics at large model scale.

\textbf{Experiment Setup.} Our experiments cover both MLP and Transformer MoEs. We utilize MLP MoEs to isolate scaling degeneracies at a manageable scale; this allows for the exhaustive hyperparameter tuning that would be computationally prohibitive with Transformers, which we reserve for validating our foundational claims. Specifically, we perform MLP MoE experiments with a single MoE block without residual connection but nonlinearity applied to $h^0_t$. We train single pass using 100 classes of TinyImageNet \citep{le2015tiny} with batch size $50$. This amounts to $1000$ steps over $50000$ images. For each scaling configuration, we separately tune a global initialization multiplier and a learning rate multiplier for each weight matrix. Due to interactions between these multipliers, we sweep all six multipliers jointly at small size $N=128$, amounting to at least $5^6=15\, 625$ runs for each combination of parameterization and scaling regime.

In our transformer MoE experiments, we train GPTs where dense layers are replaced with MoE layers \citep{muennighoff2024olmoe}. We train with warmup and cosine learning rate decay on Dolma3 \citep{olmo2025olmo3} up to width 2048 or 2.5B parameters using AdamW \citep{loshchilov2017decoupled}. We use RMS pre-norm and qk-norm and an auxiliary loss with weight 0.01 for load balancing \citep{shazeer2017}. Open source code to fully reproduce our experiments is \href{https://github.com/vankadara-lab/mssp-moe}{publicly available}. All details can be found in \Cref{sec:exp_setup}. Additional coordinate checks, learning rate sweeps and supporting evaluations are provided in \Cref{sec:add_exper}, for example showing the importance of multiplier tuning (\Cref{sec:exp_mult_tuning}) and of layerwise Adam $\epsilon$-scaling (\Cref{sec:global_eps}).

\newcommand{\dpt}[1]{\textcolor{wineframe}{#1}} 
\newcommand{\tune}[1]{\textcolor{blue}{#1}} 

\begin{table}[htbp!]
\centering
\caption{\textbf{MoE SGD and AdamW hyperparameter scaling.} Columns correspond to three MoE scaling regimes, using \textbf{expert aggregation factor $\alpha^{\text{agg}} = 1$} for sigmoid routers. Entries where $\mu$P and MSSP differ are highlighted as $\mu$P | {\color{darkforestgreen}\textbf{MSSP}}. All other weight matrices should be scaled as in $\mu$P.
In AdamW as proposed by \citet{loshchilov2017decoupled}, \textbf{weight decay should stay fixed across model scales}. In the PyTorch implementation of AdamW, the weight decay multiplier should always be the inverse of the learning rate multiplier so that the effective weight decay $\eta \cdot \text{wd}$ stays scale-independent. \textbf{Adam's $\beta_1, \beta_2$ should also be scale-independent.} 
\\ \texttt{tied}: \textbf{In Regime III, expert weights should be shared at initialization.}}
\label{tab:moe_mixtral_summary_template_main}

\small 

\setlength{\tabcolsep}{6pt}
\renewcommand{\arraystretch}{1.15}

\adjustbox{max width=\textwidth}{\begin{tabular}{>{\raggedright\arraybackslash}p{0.36\linewidth}
                >{\centering\arraybackslash}p{0.2\linewidth}
                >{\centering\arraybackslash}p{0.22\linewidth}
                >{\centering\arraybackslash}p{0.22\linewidth}}
\toprule
\textbf{Parameterization} &
\textbf{Regime I} &
\textbf{Regime II} &
\textbf{Regime III} \\
\addlinespace[2pt]
& \makecell[c]{\scriptsize $N,N_e\to\infty$\\[-1pt]\scriptsize $M,K$ fixed}
& \makecell[c]{\scriptsize $N,M,K\to\infty$\\[-1pt]\scriptsize $N_e$ fixed}
& \makecell[c]{\scriptsize $N,M,K,N_e\to\infty$} \\
\midrule

\textbf{Emb. Init. Std.} &
$d_{\rm in}^{-1/2}$ & $d_{\rm in}^{-1/2}$ & $d_{\rm in}^{-1/2}$ \\

\textbf{Emb. SGD LR} &
$N$ & $N$ & $N$ \\

\textbf{Emb. Adam LR} &
$d_{\rm in}^{-1}$ & $d_{\rm in}^{-1}$ & $d_{\rm in}^{-1}$ \\

\textbf{Emb. Adam $\epsilon$} &
$N^{-1}$ & $N^{-1}$ & $N^{-1}$ \\

\midrule

\textbf{Pre-LN Init. Std.} &
$1$ & $1$ & $1$ \\

\textbf{Pre-LN SGD LR} &
$N$ & $N$ & $N$ \\

\textbf{Pre-LN Adam LR} &
$1$ & $1$ & $1$ \\

\textbf{Pre-LN Adam $\epsilon$} &
$N^{-1}L^{-1}$ & $N^{-1}L^{-1}$ & $N^{-1}L^{-1}$ \\

\midrule

\textbf{Hidden Init. Std.} &
$N^{-1/2}$ & $N^{-1/2}$ & $N^{-1/2}$ \\

\textbf{Hidden SGD LR} &
$1$ & $1$ & $1$ \\

\textbf{Hidden Adam LR} &
$N^{-1}$ & $N^{-1}$ & $N^{-1}$ \\

\textbf{Hidden Bias SGD LR} &
$1$ & $1$ & $1$ \\

\textbf{Hidden Bias Adam LR} &
$1$ & $1$ & $1$ \\

\textbf{Hidden Adam $\epsilon$} &
$N^{-1}L^{-1}$ & $N^{-1}L^{-1}$ & $N^{-1}L^{-1}$ \\

\midrule

\multicolumn{4}{l}{\textbf{MoE routing \& experts}}\\
\addlinespace[2pt]

\textbf{Router (gating) Init. Std.} &
$N^{-1}$ | {\color{darkforestgreen}$\mathbf{0}$} & $N^{-1/2}$ & $N^{-1/2}$ \\

\textbf{Router (gating) SGD LR} &
$N^{-1}$ & $MN^{-1}$ & $1$ \\

\textbf{Router (gating) Adam LR} &
$N^{-1}$ & $N^{-1}$ & $N^{-1}$ \\

\textbf{Router Adam $\epsilon$} &
$L^{-1}$ & $M^{-1}L^{-1}$ & $M^{-1}L^{-1}$ \\

\addlinespace[2pt]

\textbf{Expert Layer 1 Init. Std.} &
$N^{-1/2}$ & $N^{-1/2}$ & $N^{-1/2}$ | {\color{darkforestgreen}$\mathbf{(N^{-1/2})^{\texttt{tied}}}$} \\

\textbf{Expert Layer 1 SGD LR} &
$1$ & $MN^{-1}$ & $M$ \\

\textbf{Expert Layer 1 Adam LR} &
$N^{-1}$ & $N^{-1}$ & $N^{-1}$ \\

\textbf{Expert Layer 1 Adam $\epsilon$} &
$N^{-1}L^{-1}$ & $M^{-1}L^{-1}$ & $N^{-1}M^{-1}L^{-1}$ \\

\addlinespace[2pt]

\textbf{Expert Layer 2 Init. Std.} &
$N_e^{-1/2}$ & $N_e^{-1/2}$ | {\color{darkforestgreen}$\mathbf{M^{1/2}N_e^{-1/2}}$} & $N_e^{-1/2}$ | {\color{darkforestgreen}$\mathbf{(N_e^{-1/2})^{\texttt{tied}}}$} \\

\textbf{Expert Layer 2 SGD LR} &
$1$ & $MN$ & $M$ \\

\textbf{Expert Layer 2 Adam LR} &
$N_e^{-1}$ & $N_e^{-1}$ & $N_e^{-1}$ \\

\textbf{Expert Layer 2 Adam $\epsilon$} &
$N^{-1}L^{-1}$ & $N^{-1}M^{-1}L^{-1}$ & $N^{-1}M^{-1}L^{-1}$ \\

\addlinespace[2pt]

\textbf{Aggregation multiplier} &
$K^{-1}$ & $K^{-1}$ & $K^{-1}$ \\

\addlinespace[2pt]

\textbf{Aux load-balancing loss multiplier} &
$1$ & $1$ & $1$ \\

\textbf{Router z-loss multiplier} &
$1$ & $1$ & $1$ \\

\midrule

\textbf{MHA Residual} &
$X^l+ L^{-1}\cdot MHA(LN(X^l))$ &
$X^l+ L^{-1}\cdot MHA(LN(X^l))$ &
$X^l+ L^{-1}\cdot MHA(LN(X^l))$ \\

\textbf{MoE FFN Residual} &
$X^l+ L^{-1}\cdot MoE(LN(X^l))$ &
$X^l+ L^{-1}\cdot MoE(LN(X^l))$ &
$X^l+ L^{-1}\cdot MoE(LN(X^l))$ \\

\midrule

\textbf{Final-LN Init. Std.} &
- & - & - \\

\textbf{Final-LN SGD LR} &
$N$ & $N$ & $N$ \\

\textbf{Final-LN Adam LR} &
$1$ & $1$ & $1$ \\

\textbf{Final-LN Adam $\epsilon$} &
$N^{-1}$ & $N^{-1}$ & $N^{-1}$ \\

\midrule

\textbf{Unemb. Init. Std.} &
$N^{-1}$ & $N^{-1}$ & $N^{-1}$ \\

\textbf{Unemb. SGD LR} &
$N^{-1}$ & $N^{-1}$ & $N^{-1}$ \\

\textbf{Unemb. Adam LR} &
$N^{-1}$ & $N^{-1}$ & $N^{-1}$ \\

\textbf{Unemb. AdamW $\epsilon$} &
1 & 1 & 1 \\

\textbf{Unemb. Fwd.} &
1 & 1 & 1 \\

\bottomrule
\end{tabular}}

\end{table}

\section{Maximal Update Desiderata for MoEs and their Shortcomings}
\label{sec:mup}

The maximal-update ($\mu$) desiderata have served as the fundamental guiding principle for deriving principled scaling rules such as $\mu$P for dense networks that deliver stable and predictable training dynamics, monotonic improvement in performance with scale, and learning-rate transfer across model sizes~\citep{yang_feature_2021, bordelon2023depthwise, vyas2024feature, dey2025don}. A natural starting point for MoE scaling is therefore to seek a parameterization that satisfies these desiderata in each of our scaling regimes. We now recapitulate the $\mu$-desiderata and then derive the unique layerwise initialization and learning-rate scaling rules satisfying them in each regime. We will then see that, despite satisfying the $\mu$-criteria by construction, $\mu$P can induce degeneracies in MoEs, motivating the refined desiderata we propose in the next section.

\textbf{Norm choice.} For any vector, matrix or tensor $T$, we measure average entry size with the RMS norm $\|T\|_{\mathrm{RMS}} := \Big(\tfrac{1}{|T|}\sum_{i\in\mathrm{entries}} T_i^2\Big)^{1/2}$, and measure spectral properties of matrices $W$ using the operator norm when equipping input and output space with $\|\cdot\|_{RMS}$: $\|W\|_{\mathrm{op}} := \sup_{x\neq 0}\frac{\|Wx\|_{\mathrm{RMS}}}{\|x\|_{\mathrm{RMS}}}.$

\textbf{Recapitulating the Maximal Update Desiderata.} The $\mu$-desiderata for non-degenerate training dynamics under model scaling are captured in the following three criteria \citep{yang_tp4b_2023}. Intuitively, they require that parameter updates induce $\Theta(1)$ feature updates that propagate forward through all linear layers without vanishing or diverging.

\begin{findinggreyB}
\textbf{Desideratum $\mu$-1: Stability and feature learning.} For every layer (including router and expert sub-layers), activations and their updates neither vanish nor diverge with model scale: $\|h_t\|_{\mathrm{RMS}}=O(1)$ for all $t\geq 0$ and $\|\Delta h_{t+1}\|_{\mathrm{RMS}}=\Theta(1)$ for some $t\geq 0$.
\end{findinggreyB}

\begin{findinggreyB}
\textbf{Desideratum $\mu$-2: Maximal effective and propagating updates.} For any trainable layer with weights $W_t$ and input activations $x_t$, write $h_t := W_t x_t$. The pre-activation update decomposes into an \emph{effective update} (from updates to the current layer's parameters) and a \emph{propagating update} (from updates to upstream layers): $\Delta h_t \;=\; {\Delta W_t\,x_t} \;+\; {W_0\,\Delta x_t}.$ Both contributions are $\Theta(1)$ at some $t \in O(1)$.
\end{findinggreyB}

\begin{findinggreyB}
\textbf{Desideratum $\mu$-3: Faithfulness.} The input to the update function $\Psi_t$ is $\Theta(1)$ at some $t>0$.
\end{findinggreyB}

For Adam, faithfulness ensures $\epsilon$ 
scales as the layerwise gradient RMS norm\footnote{Under infinite precision, gradient scaling is equivalent to scaling $\epsilon$, that is $\Psi(c\cdot g; \epsilon) = \Psi(g; c^{-1}\cdot\epsilon)$ for all $c>0$. Numerically, scaling gradients or Adam's moments $m_t, v_t$ is more stable than scaling $\epsilon$.}. $\epsilon$ acts as a crucial scale-dependent hyperparameter: values that are too large relative to gradients lead to suboptimal performance, while values that are too small are numerically unstable \citep{everett2024scaling}. For homogeneous optimizers such as SGD and signSGD, faithfulness imposes no additional constraint, since gradient rescalings can be absorbed into an effective learning-rate rescaling.

We derive the $bcd\alpha$-parameterizations that satisfy Desiderata $\mu$-1 to $\mu$-3 in each of our three 
scaling regimes. The derivation proceeds by analyzing forward and 
backward signal propagation through the network 
(App.~\ref{sec:heuristic_derivation}). For SGD the scaling is 
formally justfied by the corresponding DMFT description of the limiting training dynamics 
(App.~\ref{sec:dmft_regime1}--App.~\ref{sec:dmft-regime-iii-mup}). For Adam, the scalings are heuristically derived. For the reader's convenience, App.~\ref{sec:intuitive_scaling_explanation} provides a self-contained walkthrough of all mechanisms that underlie the scaling behaviour of MoEs in all three regimes, accessible without having to read the full technical derivation.

\begin{findinggreenB}
\textbf{Result 1 ($\mu$-parameterization for MoEs).}
For each scaling regime (I--III) and each optimizer (SGD, Adam),
there exists a unique $bcd\alpha$-parameterization under which
the MoE training dynamics satisfy Desiderata $\mu$-(1-3). The
exponents $(b_W, c_W, d_W, \alpha_{\mathrm{agg}})$ are summarized in
\Cref{tab:moe_mixtral_summary_template_main}.
\end{findinggreenB}

\begin{figure}[t]
    \centering

    \begin{subfigure}[t]{0.99\textwidth}
        \centering
        \includegraphics[width=\textwidth]{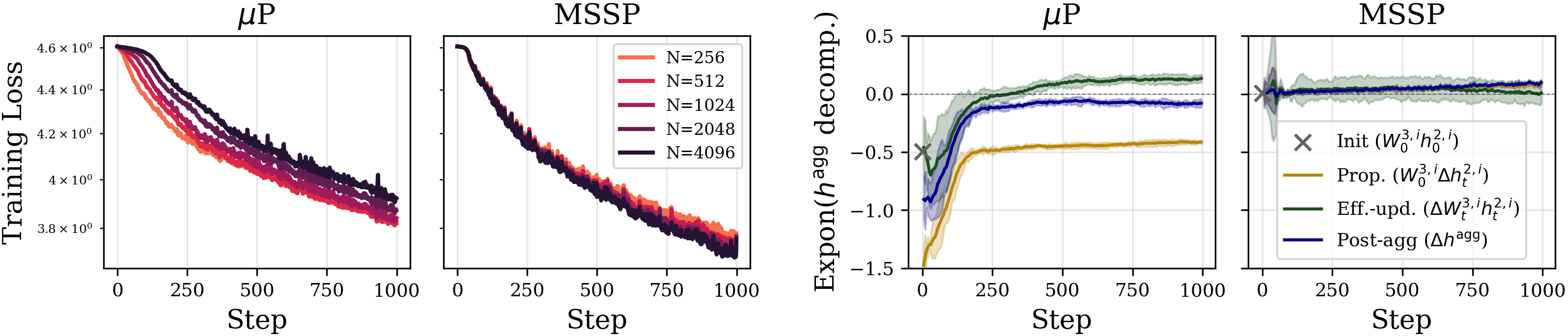} 
    \end{subfigure}

    \caption{\textbf{Delayed learning and scale dependent dynamics of $\mu$P resolved by MSSP (SGD, Regime II).} Training loss (left) is worse at large scale in $\mu$P (the darker, the wider), but monotonically improves in MSSP. Scaling exponents of sub-terms of the aggregated MoE activations $h^l_t$ (right) are approximately $0$ in all time steps in MSSP, signaling scale-independent training dynamics. In $\mu$P, initially vanishing sub-terms cause cascading time-dependent scale dependence of all terms. Even after 1000 updates, post-aggregation propagating updates still vanish in $\mu$P, as we predict.}
    \label{fig:bottleneck_sgd_mup}
\end{figure}

\textbf{Empirical degeneracies of $\mu$P in MoEs.} Having established the unique $\mu$-parameterization for each 
regime, we now examine whether it robustly delivers the scaling benefits 
expected in dense networks such as learning-rate transfer and monotonic improvement with increasing scale. 
\begin{figure}[t]
    \centering

    \begin{subfigure}[t]{0.99\textwidth}
        \centering
        \includegraphics[width=\textwidth]{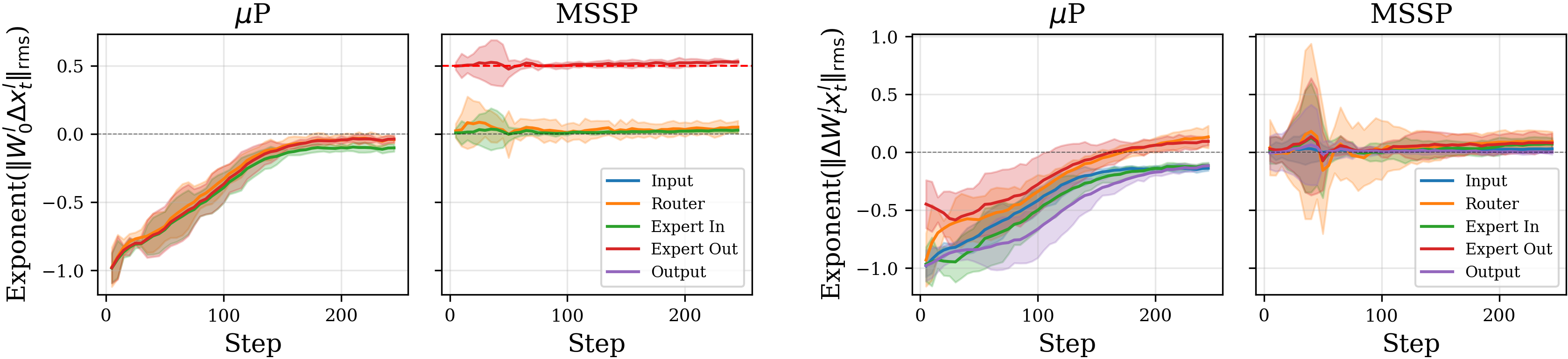}
    \end{subfigure}

    \caption{\textbf{Consistent exponents in MSSP, but not $\mu$P (SGD, Regime II).} Corresponds to \Cref{fig:bottleneck_sgd_mup}. Initially vanishing post-aggregation terms in $\mu$P induce width-dependence to cascade into all layers. Over time, exponents partially self-correct in $\mu$P and almost recover width-independent propagating updates (left) as well as effective updates (right) in all layers. By allowing diverging pre-aggregation propagating updates (red dashed line), MSSP recovers approximately width-independent effective updates at all time steps.
    }
    \label{fig:exponents_over_time}
\end{figure}

\emph{Regime I.} Under $\mu$P, training in Regime~I broadly behaves as expected: 
loss improves monotonically with scale, and learning-rate 
transfer holds across scales (\Cref{sec:lr_regime1}). 
One empirical anomaly stands out: propagating updates of the router typically vanish with scale (\Cref{fig:fixedE_base_adam_soft_ll0_0416}), even though our 
derivation predicts them to be $\Theta(1)$. We will return to this 
discrepancy in \Cref{sec:mssp}.

\begin{figure}[t]
    \centering
    \begin{subfigure}[b]{0.99\textwidth}
    \centering
    \includegraphics[width=\textwidth]{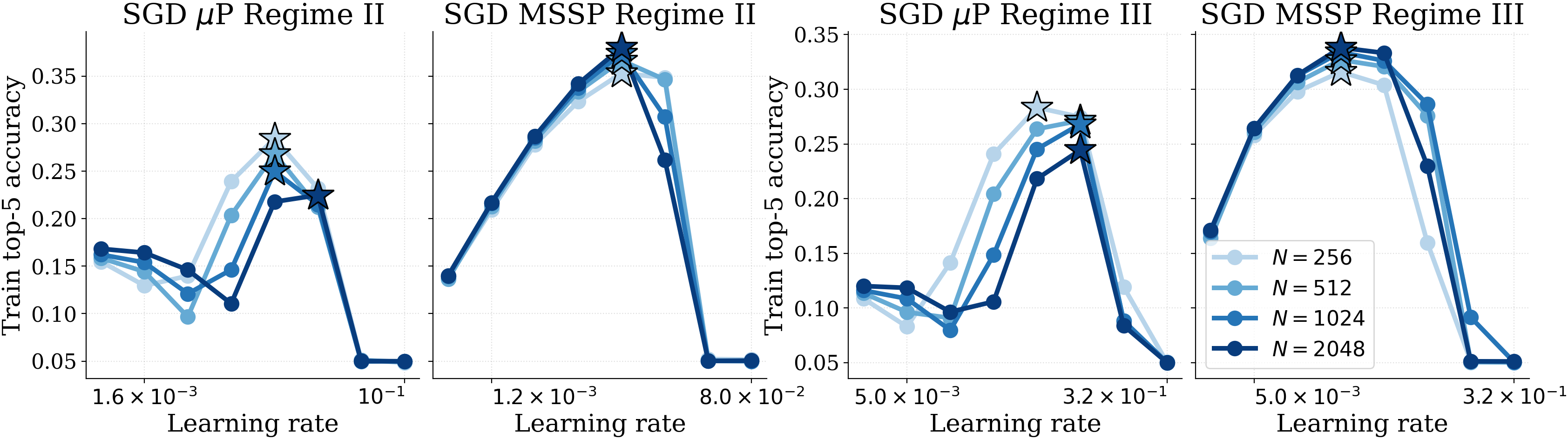} 
    \end{subfigure}
    \caption{\textbf{Robust LR transfer in MSSP, but not $\mu$P.} Top-5 training accuracy of MLP MoEs trained with SGD on TinyImageNet. The optimal learning rate often grows in $\mu$P saturating at the maximal stable learning rate with degrading performance at large scale. MSSP recovers learning rate transfer and monotonic improvements with increasing scale.}
    \label{fig:mlp_transfer_sgd}
\end{figure}

\emph{Regimes II and III.} 
Interestingly, however, we find that $\mu$P \emph{can fail} empirically in Regimes~II and III: (i) Effective and propagating updates do not remain 
$\Theta(1)$ at initialization, and can take many steps to drift toward 
their predicted asymptotic values 
(Figure~\ref{fig:exponents_over_time}).
(ii) Performance often degrades with scale
(Fig.~\ref{fig:performance_transfer_main1} and \ref{fig:bottleneck_sgd_mup} and , 
App.~\ref{sec:rcc_regime2} and \ref{sec:rcc_regime3}) and (iii) Learning rates that 
are optimal at small scales do not remain optimal at larger scales (Fig.~\ref{fig:mlp_transfer_sgd} and \ref{fig:llm_lr_sweep_main}, App.~\ref{sec:lr_regime2} and \ref{sec:lr_regime3}). \emph{In short, $\mu$P does not reliably deliver the scaling benefits {in MoEs} that motivate it in dense networks.}

\textbf{HP tuning.} These findings are robust to extensive layerwise HP tuning conducted at $N=128$ and transferred to larger scales (\Cref{sec:exp_setup}).
We resolve the underlying mechanisms in the next section.

 \textbf{Diagnosing the mechanism for failure in Regimes II and III.} To understand these empirical failures under $\mu$P, 
 examine the aggregation over experts unique to MoEs.
 The post-aggregation activations $h_t^{l+1}$ can be written as $h^l_t + h^{\text{agg}}_t$, where $h^{\text{agg}}_t$ admits the following decomposition:
\vspace{-2mm}
\begin{equation}
h^{\text{agg}}_t \;=\; \frac{1}{M}\sum_{i=1}^{M}\phi^l_i\cdot \bigg(
  \underbrace{W^{l,\text{out},i}_0\,h^{l,\text{in}}_{0,i}}_{\text{init.}}
  +\underbrace{\Delta W^{l,\text{out},i}_t\,h^{l,\text{in}}_{t,i}}_{\text{effective}}
  +\underbrace{W^{l,\text{out},i}_0\,\Delta h^{l,\text{in}}_{t,i}}_{\text{propagating}}\bigg).
  \tag{{Agg}}\label{eq:agg_decomp}
\end{equation}
A single principle organizes the scaling properties of each of these quantities. The cross-expert average exhibits LLN-like behaviour and contributes at $\Theta(1)$ whenever the per-expert summands share a \emph{coherent direction across experts}; when the per-expert summands have independent random directions, the corresponding aggregate exhibits \emph{Central Limit Theorem (CLT)-like} behaviour which suppresses a factor of $\sqrt{M}$. The underlying mechanism that produces or destroys this coherence varies across these aggregate terms and across regimes, so they can occupy distinct orders in $N$, and the order of $h^{l+1}_t$ at any finite scale is set by whichever sub-term dominates. We now provide intuition for how each of these aggregates behaves. We recommend non-technical readers to skip to `\emph{Empirical consequences}'.

\emph{Init aggregate.} Each per-expert summand $\phi^l_i\cdot W^{l,\text{out},i}_0\,h^{l,\text{in}}_{0,i}$ is a chain $W^{l,\text{out},i}_0\,W^{l,\text{in},i}_0$ of two independent Gaussian matrices acting on the shared input $x^l_0$. By rotation invariance, the cross-expert dependence between summands is mediated by its scaled norm, $\|x^l_0\|^2/M$, which as $M \to \infty$ converges to a deterministic scalar. Since summands are asymptotically i.i.d., the init aggregate  exhibits CLT-like behaviour and is $\Theta(1/\sqrt M)$ in both Regimes II and III.

\emph{Propagating aggregate.} The propagating aggregate admits a further decomposition. At leading order, $\Delta h^{l,\text{in},i}_t \approx W^{l,\text{in},i}_0\,\Delta x^l_t + \Delta W^{l,\text{in},i}_t\,x^l_t$, splitting the propagating aggregate into two contributions: $\frac{1}{M}\sum_i \phi^l_i\,W^{l,\text{out},i}_0\,W^{l,\text{in},i}_0\,\Delta x^l_t$ and $\frac{1}{M}\sum_i \phi^l_i\,W^{l,\text{out},i}_0\,\Delta W^{l,\text{in},i}_t\,x^l_t$.
The former is governed by the same CLT mechanism as the init aggregate, with one caveat: $\Delta x^l_t$ depends on each $W^{l,\text{in},i}_0$ through backpropagation. However, a single vector in $\mathbb{R}^N$ cannot align nontrivially with $M\to\infty$ independent random Gaussian chains simultaneously. The per-expert summands therefore remain approximately independent across experts, and the same CLT argument applies. This contribution is therefore also of order $\Theta(1/\sqrt M)$ in both Regimes II and III.
The second contribution behaves differently in different scaling regimes. Substituting the rank-one gradient $\Delta W^{l,\text{in},i}_t \propto (W^{l,\text{out},i}_0)^{\!\top} \delta^{l+1}_t (x^l_t)^{\!\top}$ (where $\delta^{l+1}_t = \nabla_{h_t^{l+1}} \mathcal{L}$ represents the backpropagated error signal from the subsequent layer) yields the form $\frac{1}{M}\sum_{i=1}^{M} \phi^{l}_{i}\, G_i\, u$, where $G_i := W^{l,\text{out},i}_0(W^{l,\text{out},i}_0)^{\!\top}$ and $u \propto \delta^{l+1}_t$ is shared across experts. 

\begin{enumerate}
    \item[(a)] \emph{Regime II.} When $N_e$ is held fixed, $G_i$ has rank $N_e = \Theta(1)$ and after rescaling, it is the orthogonal projector onto a $O(1)$ dimensional subspace of $\mathbb{R}^N$ that is Haar-distributed on the Grassmannian $\mathrm{Gr}(N_e, N)$ (i.e., uniformly random among $N_e$-dimensional subspaces), and these subspaces are independent across experts. Therefore, the per-expert vectors $\{G_i u\}_{i=1}^{M}$ are approximately i.i.d.\ across experts with no coherent direction, and the cross-expert average exhibits (CLT-like behaviour). Both contributions to the propagating aggregate are therefore $\Theta(1/\sqrt M)$ under $\mu$P.
    \item[(b)] \emph{Regime III.} When $N_e \asymp N$, $W^{l,\text{out},i}_0$ is generically full rank and the empirical spectrum of $G_i = W^{l,\text{out},i}_0(W^{l,\text{out},i}_0)^\top$ concentrates around its mean \citep{marchenko1967distribution}. Consequently $G_i \approx c\, I_N$ for a deterministic scalar $c = \Theta(1)$, so $G_i u \approx c\, u$ for every $i$. The per-expert summands therefore share a coherent direction $u$, and this contribution exhibits a \emph{Law of Large Numbers(LLN)-like} behaviour at $\Theta(1)$ scale. Together with the $\Theta(1/\sqrt M)$ first contribution, the propagating aggregate contains a strictly subleading contribution and exhibits a milder version of the same CLT/LLN imbalance. Moreover, the init term in both regimes exhibits CLT behaviour so remains subleading.
\end{enumerate}

\emph{Effective aggregate.}
Substituting the rank-one gradient $\Delta W^{l,\text{out},i}_t \propto \delta^{l+1}_t \bigl(h^{l,\text{in},i}_{t-1}\bigr)^{\!\top}$ reveals that the summands of the effective aggregate share a coherent direction $\delta^{l+1}_t$. The effective aggregate therefore exhibits LLN-like behaviour at $\Theta(1)$ along $\delta^{l+1}_t$ in both Regimes II and III.

\emph{Empirical consequences.} In summary, \emph{$h^{\mathrm{agg}}_t$ is composed of unbalanced contributions}: the init term is of order $\Theta(1/\sqrt M)$ in both regimes; the effective term is of order $\Theta(1)$ in both; the propagating term decomposes into a first contribution of $\Theta(1/\sqrt M)$ in both regimes, and a second contribution of $\Theta(1/\sqrt M)$ in Regime II but $\Theta(1)$ in Regime III. The training-induced update $\Delta h^{\mathrm{agg}}_t$, formed from the effective and propagating terms, is therefore unbalanced in both regimes, with the imbalance more pronounced in Regime II. This is confirmed empirically in Regime II under $\mu$P (Fig.~\ref{fig:bottleneck_sgd_mup}): at finite scale $\Delta h^{\mathrm{agg}}_t$ is dominated by the scale-suppressed init and propagating terms. Feature learning is therefore vanishing early in training and takes many optimization steps to become $\Theta(1)$ and performance consequently degrades with increasing scale (Fig.~\ref{fig:performance_transfer_main1} and \ref{fig:bottleneck_sgd_mup}).

\section{A More Fundamental Desideratum Beyond Maximal Updates: Scaling MoEs Requires Maximal Scale Stability}
\label{sec:mssp}
In \Cref{sec:mup}, we traced the failure modes of $\mu$P to scale-dependent contributions arising in the training dynamics. This motivates a more general and fundamental desideratum, which we call \emph{maximal scale stability}: \emph{every primitive interaction obtained by decomposing each weight as $W = W_0 + \Delta W$ and propagating this split through the network should have a $\Theta(1)$ impact on the forward and (appropriately normalized) backward dynamics.} For most settings of interest, this reduces to a compact set of operational conditions; we instantiate the desideratum for MoEs as follows:

\begin{findinggreyB}
\textbf{Maximal Scale Stability Desiderata for MoEs.}
\begin{enumerate}
    \item \emph{Forward.} The maximal update Desiderata~$\mu$-1, $\mu$-2, $\mu$-3 hold.
    \item \emph{Backward.} Along any linear map $h^\ell_t = W^\ell_t\,x^{\ell-1}_t$, analogous to activations, gradients admit the transpose recursion $\bar\delta^{\ell-1}_t = (W^\ell_t)^\top\,\delta^\ell_t$ and consequently the analogous effective/propagating decomposition $\bar\delta^{\ell-1}_t \;=\; {(W^\ell_0)^\top\,\delta^\ell_t} \;+\; {(\Delta W^\ell_t)^\top\,\delta^\ell_t},$ where $\bar\delta^\ell_t := \nabla_{x^\ell_t}\mathcal{L}$ and $\delta^\ell_t := \nabla_{h^\ell_t}\mathcal{L}$. Both contributions are balanced and remain $\Theta(1)$ under appropriate normalization.
    \item \emph{Forward and Backward Aggregation.} Each of the init, propagating, and effective contributions of the cross-expert aggregation \eqref{eq:agg_decomp} is $\Theta(1)$. The analogous requirement applies to the expert-aggregated gradient at the shared input $h^l$ (see Eq.~\eqref{eq:bwd_agg_decomp}).
\end{enumerate}
\end{findinggreyB}

We refer to a parameterization that satisfies these conditions as the \emph{Maximally Scale-Stable Parameterization} (MSSP). For linear layers with diverging input and output dimensions (\texttt{fan-in}~$\asymp$~\texttt{fan-out}), forward and backward desiderata coincide by the symmetry of the Gaussian initialization. MSSP desiderata reduce to $\mu$-desiderata in dense architectures such as MLPs and Transformers. But in general, forward scale stability does not imply backward scale stability as we will see below.
 \begin{figure}[t]
    \centering
    \begin{subfigure}[b]{0.99\textwidth}
    \centering
    \includegraphics[width=\textwidth]{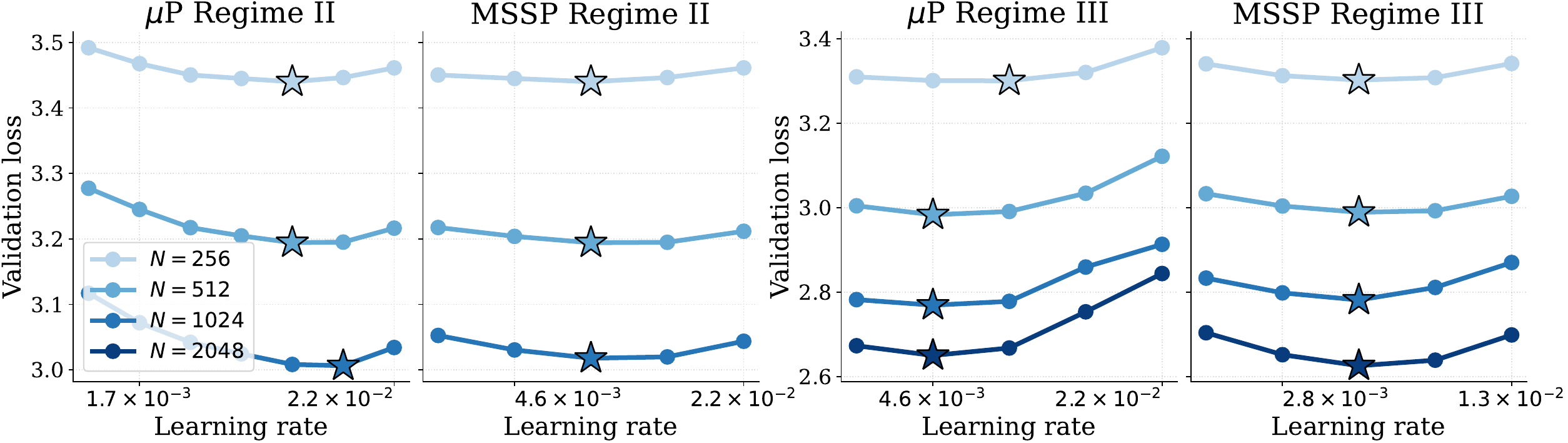}
    \end{subfigure}
    \caption{\textbf{Learning rate transfer in Transformers.} Validation loss for GPT MoEs trained with Adam in $\mu$P and MSSP for 2.5B tokens in Regime II ($N, M, K \to \infty, N_e \in \Theta(1)$ left) and Regime III ($N, N_e, M, K \to \infty$ right). Observe LR transfer and monotonic improvement with scale in MSSP.}
    \label{fig:llm_lr_sweep_main}
\end{figure}

\textbf{MSSP for MoEs.} We derive parameterizations that satisfy the above desiderata in each scaling regime. The derivation proceeds analogously by analysing forward and backward signal propagation through the network (App.~\ref{sec:heuristic_derivation}), and is formally justified by the corresponding DMFT description of the limiting training dynamics (App.~\ref{sec:dmft_regime2} and App.~\ref{sec:dmft_regime3}).
 
\begin{findinggreenB}
\textbf{Result 2 (MSSP for MoEs).}
The parameterization in \Cref{tab:moe_mixtral_summary_template_main} satisfies all MSSP desiderata in Regimes~I and~III. In Regime~II, it satisfies Desiderata~1--3 except that the propagating update of $h^{3,i}$ is of order $\Theta(\sqrt{M})$; however, its impact on the aggregate dynamics remains $\Theta(1)$.
\end{findinggreenB}

\emph{Regime~II $(N, M, K \to \infty, N_e \in \Theta(1))$.} Under $\mu$P in Regime II, gradients at \emph{every layer} except $h^{l,\textrm{out},i}$ have scale-dependent contributions. In addition, we saw in Sec.~\ref{sec:mup} that both the forward and backward aggregations receive scale-dependent contributions. MSSP rectifies all these imbalances with a single intervention: \textbf{the initialization variance of the expert output is amplified from $\mathbf{1/N_e}$ to $\mathbf{M/N_e}$}. Although failure modes in $\mu$P arise through structurally distinct mechanisms, 
this variance amplification simultaneously rebalances every forward and backward cross-expert sub-aggregate, every previously sub-leading contribution to per-expert gradients as well as to the gradient flowing through the router (cf. tables in App.~\ref{ssec:mup_r2_summary} for $\mu$P versus App.~\ref{ssec:ssp_r2_summary} for MSSP).

The only quantity that does not remain $\Theta(1)$ under MSSP is the per-expert propagating update $W^{l,\text{out},i}_0\,h^{l,\text{in},i}_t = \Theta(\sqrt M)$, which is impossible to avoid. We argue that this divergence is benign since its impact on the training dynamics is $\Theta(1)$, and no training observable inherits this divergence.
The dynamics admits a well-defined DMFT limit under MSSP (App.~\ref{sec:dmft_regime2}). We empirically verify that all relevant terms remain scale-stable over time (Figure~\ref{fig:exponents_over_time}, App.~\ref{sec:rcc_regime2}). 
Consequently, MSSP-Regime-II restores both monotonic performance improvement with scale and learning rate transfer 
from small to large scale for both SGD and Adam (Fig.~\ref{fig:performance_transfer_main1}, \ref{fig:mlp_transfer_sgd} and \ref{fig:llm_lr_sweep_main}, App.~\ref{sec:lr_regime2}). 

\emph{Regime~III $(N, N_e, M, K, \to \infty)$.} When $N_e \asymp N$, amplifying the expert output initialization variance 
induces divergence in the backward pass. The viable fix in Regime~III is instead structural: \textbf{share the expert initialization, $\mathbf{W^{l,\text{out},i}_0 = W^{l,\text{out}}_0}$ and $\mathbf{W^{l,\text{in},i}_0 = W^{l,\text{in}}_0}$ for all $i$} and the allow the routing mechanism to diversify experts over time. Under this intervention, complete scale stability is restored (see theory App.~\ref{ssec:ssp_r3_summary} and experiment App.~\ref{sec:rcc_regime3}). 
\Cref{fig:performance_transfer_main1} shows that this intervention can have impactful practical consequences: performance of both SGD and Adam on TinyImageNet reliably improves with scale only under MSSP but not $\mu$P,  
all relevant quantities remain $\Theta(1)$ through training (see 
App.~\ref{sec:rcc_regime3}), and learning rate reliably transfers across scales (Fig. \ref{fig:mlp_transfer_sgd} and \ref{fig:llm_lr_sweep_main}).

\emph{Hence, both regimes call for structurally distinct interventions: increased expert output initialization 
in Regime~II, and shared initialization in Regime~III. Neither solution transfers to the other regime.} A more detailed but intuitive mechanism-by-mechanism account of how the MSSP prescription rectifies each $\mu$P failure is provided in App.~\ref{sec:intuitive_scaling_explanation}. 

\emph{Regime I $(M, K \in \Theta(1))$.} $\mu$P satisfies our operational MSSP desiderata for MoEs in Regime I. 
However, splitting $\Delta h^{l-1}$ into the contributions of the router-gradient and the expert-gradient pathways reveals that the former is correlated with $Q_0$ and contributes coherently at $\Theta(1)$, while the latter is weakly correlated with $Q_0$ and is CLT-suppressed at $\Theta(1/\sqrt N)$. 
To recover residual balance in the router's propagating update $Q^l_0 \Delta h^{l-1}$, \textbf{MSSP initializes the router to zero}. The empirical impacts of this adjustment are marginal (\Cref{fig:performance_transfer_main1}, App.~\ref{sec:rcc_regime1}).

\textbf{Scaling depth $L$.} In residual MoEs, including Transformer MoEs, 
depth composes naturally with our analysis: once the per-layer MoE block
is parameterized according to MSSP, with $\Theta(1/L)$ scaling of the residual MoE blocks \citep{dey2025don},
the analysis of \citet{bordelon2024infinite} extends with minimal
modification to give a well-defined infinite-depth limit. \Cref{tab:moe_mixtral_summary_template_main} gives
a complete scaling prescription for modern transformer MoE architectures as a function of width $N$, depth $L$, expert width $N_e$, and number of (active) experts $K, M$.

\section{Self-consistent DMFT for MoE training dynamics}
\label{sec:insights_from_dmft}

For each regime, we derive self-consistent equations describing the limiting MoE training dynamics using a path-integral formulation based on the Martin–Siggia–Rose–De Dominicis–Janssen (MSRDJ) framework \cite{martin1973statistical, dedominicis1976techniques, janssen1976lagrangean}. This framework underlies pioneering DMFT analyses of training dynamics in infinite-width and infinite-depth deep neural networks \cite{bordelon2022self, bordelon2023depthwise, bordelon2024infinite}. We derive DMFT limits for MoE architectures in each of the three scaling regimes, under both $\mu$P and MSSP; each admits a well-defined scaling limit. In each regime, the limiting MoE dynamics under MSSP enjoy several analytical properties described below. The full set of DMFT equations is provided in App.\ref{sec:dmft_regime1}-\ref{sec:dmft_regime3}. We defer the comparison to $\mu$P limits to App.~\ref{sec:dmft-regime-iii-mup}.

\textbf{Regime I $(M, K \in \Theta(1))$.} The DMFT has a finite-expert mean-field factorization. Residual-stream neurons and their backward signals are i.i.d.\ draws from a global single-site distribution; within each expert, hidden-layer neurons are i.i.d.\ draws from an expert-local single-site distribution. Both distributions are characterized self-consistently by a finite set of macroscopic order parameters. Because the number of experts is finite, the expert index is not replaced by a population mean field: the experts and router variables remain explicitly represented. Averages over the global distribution define the global feature and gradient kernels; Together with router variables, these order parameters close the limiting dynamics. Under soft routing, the stochastic router field vanishes in this limit, inducing deterministic routing. Without an additional symmetry-breaking mechanism (Top-$K$, routing noise, or random router biases), the router remains uniform and experts stay identical throughout training. This is empirically verified in \Cref{fig:training_dynamics_fixed_E_soft}.

\textbf{Regime II.} The simultaneous $N, M, K \asymp n \to \infty$ limit with $N_e \in \Theta(1)$ reveals a {finite expert width mean-field factorization} in MSSP. Residual-stream neurons and their backward signals are i.i.d.\ draws from an effective global single-site distribution; experts together with their router variables are i.i.d.\ draws from an effective expert/router distribution. Both are self-consistently characterized by a finite set of order parameters. Since per-expert hidden width is fixed, the hidden layer of a typical expert is a finite-dimensional component of the expert/router process. Averages over the global distribution define the global feature and gradient kernels; averages over the expert/router distribution define the routed expert-level kernels. These order parameters close the limiting dynamics.

\textbf{Regime III.} The simultaneous $N, N_e, M, K \asymp n \to \infty$ limit reveals a \emph{four-level, conditional, mean-field hierarchy} in MSSP. Schematically,
\[
\mathcal X_{\mathrm{glob}} \sim P_{\mathrm{glob}}(\cdot;\mathcal K),
\qquad
\mathcal F_{\mathrm{sh}} \sim P_{\mathrm{sh}}(\cdot;\mathcal K),
\]
\[
\mathcal E \mid \mathcal F_{\mathrm{sh}}
\sim P_{\mathrm{ex/r}}(\cdot \mid \mathcal F_{\mathrm{sh}};\mathcal K),
\qquad
\mathcal U \mid (\mathcal F_{\mathrm{sh}},\mathcal E)
\sim P_{\mathrm{loc}}(\cdot \mid \mathcal F_{\mathrm{sh}},\mathcal E;\mathcal K),
\]
where \(\mathcal X_{\mathrm{glob}}\) denotes the global single-site process, \(\mathcal F_{\mathrm{sh}}\) the shared expert-hidden single-site process, \(\mathcal E\) the expert/router single-site process, and \(\mathcal U\) the within-expert hidden-neuron single-site process.

\emph{Global single-site process.} Residual-stream coordinates and their backward signals are i.i.d.\ draws from an effective distribution characterized self-consistently by a finite set of order parameters; averages over this process define the global feature and gradient kernels.

\emph{Shared expert-hidden single-site process.} Coordinates of \emph{expert-averaged} hidden-layer fields, arising from expert-averaged hidden activations and gradients, are i.i.d.\ draws from an effective shared distribution. Averages define the shared expert-hidden kernels, which summarize the correlation structure of these fields induced by the shared initial expert weights.

\emph{Expert/router single-site process.} Conditional on the shared expert-hidden fields, expert/router sites are i.i.d.\ draws from an effective expert/router distribution; averages define expert-level kernels.

\emph{Within-expert hidden-neuron single-site process.} Conditional on the shared expert-hidden fields and the corresponding router state, neurons in the expert hidden layers are i.i.d.\ draws from an effective distribution; averages define the expert-local forward and backward kernels.

The four processes form a closed self-consistent system: kernels defined at each level enter as parameters of the single-site distributions at the others, and the limiting dynamics are determined by solving all four levels simultaneously.

\section{Discussion and Future Work}\label{sec:discussion}

Scaling laws and hyperparameter transfer have become the backbone of
modern model development. Yet, for MoE architectures there remained a critical gap of how to scale the optimization hyperparameters jointly with MoE model dimensions $M, N, N_e, K, L$. The Maximally Scale-Stable Parameterization (MSSP) introduced
here closes this gap. We believe generalizing the $\mu$-desiderata to
maximally scale-stable desiderata provides a framework for scale-invariant
dynamics across a broader set of architectures, optimizers, and scaling
dimensions, with practical benefits including predictable performance
gains and hyperparameter transfer.

MSSP enables the first principled compute-optimal comparison of the three
co-scaling regimes. Earlier work on optimal expert granularity
\citep{pmlr-v162-clark22a,krajewski2024scaling,dai2024deepseekmoe} was
conducted under suboptimal model scaling and merits reevaluation. More fundamentally, why hyperparameters transfer under
scale-invariant parameterizations, in MoEs, dense networks, or beyond,
remains an open theoretical problem.

The DMFT limits are of independent interest, opening routes to studying
finite-size corrections, expert specialization, and the role of routing. They also provide a framework for analyzing auxiliary
load-balancing and $z$-losses, which we verify empirically do not alter
the scaling exponents but may shed light on router collapse through their
interaction with the limiting dynamics.

\paragraph{Limitations.} In Regime II, MSSP causes the per-expert
pre-aggregation $W^{l,\text{out},i}_0\,h^{l,\text{in},i}_t$ to scale as
$\Theta(\sqrt{M})$. Though its dynamical impact remains $\Theta(1)$, at
very large $M$ this may raise numerical-precision concerns, which a fused
kernel computing the cross-expert aggregate without materializing the
per-expert quantity could address. Our DMFT derivations are at the level
of physical rigor; fully rigorous proofs remain open. Of practical interest would also be an analysis that captures increasing sparsity $M/K\to\infty$.

\section*{Acknowledgments}
This work has been supported by the Gatsby Charitable Foundation (GAT4058). SB has been supported by the German Research Foundation through
the Cluster of Excellence “Machine Learning - New Perspectives for
Science" (EXC 2064/1 number 390727645).

\bibliography{neurips}

@article{ghosh2025understanding,
  title={Understanding the Mechanisms of Fast Hyperparameter Transfer},
  author={Ghosh, Nikhil and Wu, Denny and Bietti, Alberto},
  journal={arXiv preprint arXiv:2512.22768},
  year={2025}
}

@article{le2015tiny,
  title={Tiny imagenet visual recognition challenge},
  author={Le, Yann and Yang, Xuan and others},
  journal={CS 231N},
  volume={7},
  number={7},
  pages={3},
  year={2015}
}

@misc{mayaki2026generalization,
  title={Generalization and Scaling Laws for Mixture-of-Experts Transformers},
  author={Mayaki, Mansour Zoubeirou},
  year={2026},
  note={Manuscript}
}

@article{he2024mixture,
  title={Mixture of a million experts},
  author={He, Xu Owen},
  journal={arXiv preprint arXiv:2407.04153},
  year={2024}
}

@article{krajewski2024scaling,
  title={Scaling laws for fine-grained mixture of experts},
  author={Krajewski, Jakub and Ludziejewski, Jan and Adamczewski, Kamil and Pi{\'o}ro, Maciej and Krutul, Micha{\l} and Antoniak, Szymon and Ciebiera, Kamil and Kr{\'o}l, Krystian and Odrzyg{\'o}{\'z}d{\'z}, Tomasz and Sankowski, Piotr and others},
  journal={arXiv preprint arXiv:2402.07871},
  year={2024}
}

@article{boix2025power,
  title={The power of fine-grained experts: Granularity boosts expressivity in Mixture of Experts},
  author={Boix-Adsera, Enric and Rigollet, Philippe},
  journal={arXiv preprint arXiv:2505.06839},
  year={2025}
}

@article{chaintron2026resnets,
  title={Resnets of all shapes and sizes: Convergence of training dynamics in the large-scale limit},
  author={Chaintron, Louis-Pierre and Chizat, L{\'e}na{\"\i}c and Maas, Javier},
  journal={arXiv preprint arXiv:2603.18168},
  year={2026}
}

@article{nvidiaMoEMegatron2026,
  title={Scalable Training of Mixture-of-Experts Models with Megatron Core},
  author={NVIDIA},
  journal={arXiv preprint arXiv:2603.07685},
  year={2026}
}

@article{jiang2026hyperparameter,
  title={Hyperparameter Transfer with Mixture-of-Expert Layers},
  author={Jiang, Tianze and Bordelon, Blake and Pehlevan, Cengiz and Hanin, Boris},
  journal={arXiv preprint arXiv:2601.20205},
  year={2026}
}

@article{dai2024deepseekmoe,
  title={Deepseekmoe: Towards ultimate expert specialization in mixture-of-experts language models},
  author={Dai, Damai and Deng, Chengqi and Zhao, Chenggang and Xu, RX and Gao, Huazuo and Chen, Deli and Li, Jiashi and Zeng, Wangding and Yu, Xingkai and Wu, Yu and others},
  journal={arXiv preprint arXiv:2401.06066},
  year={2024}
}

@article{malasnicki2025mu,
  title={mu-Parametrization for Mixture of Experts},
  author={Ma{\l}a{\'s}nicki, Jan and Ciebiera, Kamil and Boru{\'n}, Mateusz and Pi{\'o}ro, Maciej and Ludziejewski, Jan and Stefaniak, Maciej and Krutul, Micha{\l} and Jaszczur, Sebastian and Cygan, Marek and Adamczewski, Kamil and others},
  journal={arXiv preprint arXiv:2508.09752},
  year={2025}
}

@article{muennighoff2024olmoe,
  title={Olmoe: Open mixture-of-experts language models},
  author={Muennighoff, Niklas and Soldaini, Luca and Groeneveld, Dirk and Lo, Kyle and Morrison, Jacob and Min, Sewon and Shi, Weijia and Walsh, Pete and Tafjord, Oyvind and Lambert, Nathan and others},
  journal={arXiv preprint arXiv:2409.02060},
  year={2024}
}

@article{zoph2022st,
  title={St-moe: Designing stable and transferable sparse expert models},
  author={Zoph, Barret and Bello, Irwan and Kumar, Sameer and Du, Nan and Huang, Yanping and Dean, Jeff and Shazeer, Noam and Fedus, William},
  journal={arXiv preprint arXiv:2202.08906},
  year={2022}
}

@article{guo2025deepseek,
  title={Deepseek-r1: Incentivizing reasoning capability in llms via reinforcement learning},
  author={Guo, Daya and Yang, Dejian and Zhang, Haowei and Song, Junxiao and Zhang, Ruoyu and Xu, Runxin and Zhu, Qihao and Ma, Shirong and Wang, Peiyi and Bi, Xiao and others},
  journal={arXiv preprint arXiv:2501.12948},
  year={2025}
}

@article{eigen2013learning,
  title={Learning factored representations in a deep mixture of experts},
  author={Eigen, David and Ranzato, Marc'Aurelio and Sutskever, Ilya},
  journal={arXiv preprint arXiv:1312.4314},
  year={2013}
}

@article{fedus2022switch,
  title={Switch transformers: Scaling to trillion parameter models with simple and efficient sparsity},
  author={Fedus, William and Zoph, Barret and Shazeer, Noam},
  journal={Journal of Machine Learning Research},
  volume={23},
  number={120},
  pages={1--39},
  year={2022}
}

@article{chen2022towards,
  title={Towards understanding mixture of experts in deep learning},
  author={Chen, Zixiang and Deng, Yihe and Wu, Yue and Gu, Quanquan and Li, Yuanzhi},
  journal={arXiv preprint arXiv:2208.02813},
  year={2022}
}

@article{jacobs1991adaptive,
  title={Adaptive mixtures of local experts},
  author={Jacobs, Robert A and Jordan, Michael I and Nowlan, Steven J and Hinton, Geoffrey E},
  journal={Neural computation},
  volume={3},
  number={1},
  pages={79--87},
  year={1991},
  publisher={MIT Press}
}

@article{chizat2018global,
  title={On the global convergence of gradient descent for over-parameterized models using optimal transport},
  author={Chizat, Lenaic and Bach, Francis},
  journal={Advances in neural information processing systems},
  volume={31},
  year={2018}
}

@article{dey2025don,
  title={Don't be lazy: CompleteP enables compute-efficient deep transformers},
  author={Dey, Nolan and Zhang, Bin Claire and Noci, Lorenzo and Li, Mufan and Bordelon, Blake and Bergsma, Shane and Pehlevan, Cengiz and Hanin, Boris and Hestness, Joel},
  journal={arXiv:2505.01618},
  year={2025}
}

@article{hoffmann2022empirical,
  title={An empirical analysis of compute-optimal large language model training},
  author={Hoffmann, Jordan and Borgeaud, Sebastian and Mensch, Arthur and Buchatskaya, Elena and Cai, Trevor and Rutherford, Eliza and de Las Casas, Diego and Hendricks, Lisa Anne and Welbl, Johannes and Clark, Aidan and others},
  journal={Advances in Neural Information Processing Systems (NeurIPS)},
  volume={35},
  pages={30016--30030},
  year={2022}
}

@inproceedings{noci2024super,
  title={Super Consistency of Neural Network Landscapes and Learning Rate Transfer},
  author={Noci, Lorenzo and Meterez, Alexandru and Hofmann, Thomas and Orvieto, Antonio},
  booktitle={The Thirty-eighth Annual Conference on Neural Information Processing Systems (NeurIPS)},
  year={2024}
}

@article{brown2020language,
  title={Language models are few-shot learners},
  author={Brown, Tom and Mann, Benjamin and Ryder, Nick and Subbiah, Melanie and Kaplan, Jared D and Dhariwal, Prafulla and Neelakantan, Arvind and Shyam, Pranav and Sastry, Girish and Askell, Amanda and others},
  journal={Advances in Neural Information Processing Systems (NeurIPS)},
  volume={33},
  pages={1877--1901},
  year={2020}
}

@inproceedings{vankadara24ssms,
 author = {Leena Chennuru Vankadara and Jin Xu and Moritz Haas and Volkan Cevher},
 booktitle = {The Thirty-eighth Annual Conference on Neural Information Processing Systems (NeurIPS)},
 title = {On Feature Learning in Structured State Space Models},
 url = {https://openreview.net/forum?id=aQv5AbN1wF},
 year = {2024}
}

@article{everett2024scaling,
  title={Scaling exponents across parameterizations and optimizers},
  author={Everett, Katie and Xiao, Lechao and Wortsman, Mitchell and Alemi, Alexander A and Novak, Roman and Liu, Peter J and Gur, Izzeddin and Sohl-Dickstein, Jascha and Kaelbling, Leslie Pack and Lee, Jaehoon and others},
  journal={arXiv:2407.05872},
  year={2024}
}

@inproceedings{
blake2024u,
title={u-$\mu$P: The Unit-Scaled Maximal Update Parametrization},
author={Charlie Blake and Constantin Eichenberg and Josef Dean and Lukas Balles and Luke Yuri Prince and Bj{\"o}rn Deiseroth and Andres Felipe Cruz-Salinas and Carlo Luschi and Samuel Weinbach and Douglas Orr},
booktitle={The Thirteenth International Conference on Learning Representations (ICLR)},
year={2025},
url={https://openreview.net/forum?id=P7KRIiLM8T}
}

@inproceedings{
wortsman2023small,
title={Small-scale proxies for large-scale Transformer training instabilities},
author={Mitchell Wortsman and Peter J Liu and Lechao Xiao and Katie E Everett and Alexander A Alemi and Ben Adlam and John D Co-Reyes and Izzeddin Gur and Abhishek Kumar and Roman Novak and Jeffrey Pennington and Jascha Sohl-Dickstein and Kelvin Xu and Jaehoon Lee and Justin Gilmer and Simon Kornblith},
booktitle={The Twelfth International Conference on Learning Representations (ICLR)},
year={2024},
url={https://openreview.net/forum?id=d8w0pmvXbZ}
}

@inproceedings{
bordelon2024infinite,
title={Infinite Limits of Multi-head Transformer Dynamics},
author={Blake Bordelon and Hamza Tahir Chaudhry and Cengiz Pehlevan},
booktitle={The Thirty-eighth Annual Conference on Neural Information Processing Systems (NeurIPS)},
year={2024},
url={https://openreview.net/forum?id=p0BBKhD5aI}
}

@article{kaplan2020scaling,
  title={Scaling laws for neural language models},
  author={Kaplan, Jared and McCandlish, Sam and Henighan, Tom and Brown, Tom B and Chess, Benjamin and Child, Rewon and Gray, Scott and Radford, Alec and Wu, Jeffrey and Amodei, Dario},
  journal={arXiv:2001.08361},
  year={2020}
}

@inproceedings{
bordelon2023depthwise,
title={Depthwise Hyperparameter Transfer in Residual Networks: Dynamics and Scaling Limit},
author={Blake Bordelon and Lorenzo Noci and Mufan Bill Li and Boris Hanin and Cengiz Pehlevan},
booktitle={The Twelfth International Conference on Learning Representations (ICLR)},
year={2024},
url={https://openreview.net/forum?id=KZJehvRKGD}
}

@article{noci2024shaped,
  title={The shaped transformer: Attention models in the infinite depth-and-width limit},
  author={Noci, Lorenzo and Li, Chuning and Li, Mufan and He, Bobby and Hofmann, Thomas and Maddison, Chris J and Roy, Dan},
  journal={Advances in Neural Information Processing Systems (NeurIPS)},
  volume={36},
  year={2024}
}

@inproceedings{hayou2023width,
  title={Width and depth limits commute in residual networks},
  author={Hayou, Soufiane and Yang, Greg},
  booktitle={International Conference on Machine Learning (ICML)},
  pages={12700--12723},
  year={2023},
  organization={PMLR}
}

@article{mei2018mean,
  title={A mean field view of the landscape of two-layer neural networks},
  author={Mei, Song and Montanari, Andrea and Nguyen, Phan-Minh},
  journal={Proceedings of the National Academy of Sciences},
  volume={115},
  number={33},
  pages={E7665--E7671},
  year={2018},
  publisher={National Acad Sciences}
}

@article{bordelon2022self,
  title={Self-consistent dynamical field theory of kernel evolution in wide neural networks},
  author={Bordelon, Blake and Pehlevan, Cengiz},
  journal={Advances in Neural Information Processing Systems (NeurIPS)},
  volume={35},
  pages={32240--32256},
  year={2022}
}

@article{vyas2024feature,
  title={Feature-learning networks are consistent across widths at realistic scales},
  author={Vyas, Nikhil and Atanasov, Alexander and Bordelon, Blake and Morwani, Depen and Sainathan, Sabarish and Pehlevan, Cengiz},
  journal={Advances in Neural Information Processing Systems (NeurIPS)},
  volume={36},
  year={2024}
}

@techreport{cifar10,
  title={Learning multiple layers of features from tiny images},
  author={Krizhevsky, Alex and Hinton, Geoffrey and others},
  year={2009},
  institution={University of Toronto}
}

@article{yang_tp1_2019,
  title={Wide feedforward or recurrent neural networks of any architecture are gaussian processes},
  author={Yang, Greg},
  journal={Advances in Neural Information Processing Systems (NeurIPS)},
  volume={32},
  year={2019}
}

@article{tp5_2022,
  title={Tensor programs v: Tuning large neural networks via zero-shot hyperparameter transfer},
  author={Yang, Greg and Hu, Edward J and Babuschkin, Igor and Sidor, Szymon and Liu, Xiaodong and Farhi, David and Ryder, Nick and Pachocki, Jakub and Chen, Weizhu and Gao, Jianfeng},
  journal={arXiv:2203.03466},
  year={2022}
}

@article{yang_tp4b_2023,
      title={Tensor Programs IVb: Adaptive Optimization in the Infinite-Width Limit}, 
      author={Greg Yang and Etai Littwin},
      year={2023},
      journal={arXiv:2308.01814}
}

@InProceedings{yang_feature_2021,
  title = 	 {Tensor Programs IV: Feature Learning in Infinite-Width Neural Networks},
  author =       {Yang, Greg and Hu, Edward J.},
  booktitle = 	 {International Conference on Machine Learning (ICML)},
  year = 	 {2021},
}

@inproceedings{pytorch,
title = {PyTorch: An Imperative Style, High-Performance Deep Learning Library},
author = {Paszke, Adam and Gross, Sam and Massa, Francisco and Lerer, Adam and Bradbury, James and Chanan, Gregory and Killeen, Trevor and Lin, Zeming and Gimelshein, Natalia and Antiga, Luca and Desmaison, Alban and Kopf, Andreas and Yang, Edward and DeVito, Zachary and Raison, Martin and Tejani, Alykhan and Chilamkurthy, Sasank and Steiner, Benoit and Fang, Lu and Bai, Junjie and Chintala, Soumith},
booktitle = {Advances in Neural Information Processing Systems (NeurIPS)},
year = {2019},
}

@incollection{neal_priors_1996,
  author    = {Neal, Radford M.},
  title     = {Priors for Infinite Networks},
  booktitle = {Bayesian Learning for Neural Networks},
  year      = {1996},
  publisher = {Springer New York},
  pages     = {29--53}
}

@inproceedings{jacot_neural_2018,
	title = {Neural {Tangent} {Kernel}: {Convergence} and generalization in neural networks},
	shorttitle = {Neural tangent kernel},
	booktitle = {Advances in Neural Information Processing Systems (NeurIPS)},
	author = {Jacot, Arthur and Gabriel, Franck and Hongler, Clément},
	year = {2018},
	pages = {8571--8580},
}

@inproceedings{he_delving_2015,
	title = {Delving deep into rectifiers: {Surpassing} human-level performance on imagenet classification},
	shorttitle = {Delving deep into rectifiers},
	booktitle = {{IEEE} international conference on computer vision (ICCV)},
	author = {He, Kaiming and Zhang, Xiangyu and Ren, Shaoqing and Sun, Jian},
	year = {2015},
	pages = {1026--1034},
}

@inproceedings{matthews_gaussian_2018,
	title = {Gaussian process behaviour in wide deep neural networks},
	booktitle = {International Conference on Learning Representations (ICLR)},
	author = {Matthews, Alexander G. de G. and Hron, Jiri and Rowland, Mark and Turner, Richard E. and Ghahramani, Zoubin},
	year = {2018},
}

@book{vershynin2018high,
  title={High-dimensional probability: An introduction with applications in data science},
  author={Vershynin, Roman},
  volume={47},
  year={2018},
  publisher={Cambridge University Press}
}

@inproceedings{loshchilov2017decoupled,
  title={Decoupled weight decay regularization},
  author={Loshchilov, Ilya and Hutter, Frank},
  booktitle={International Conference on Learning Representations (ICLR)},
  year={2019}
}

@inproceedings{
haas2025largelrs,
title={On the Surprising Effectiveness of Large Learning Rates under Standard Width Scaling},
author={Moritz Haas and Sebastian Bordt and Ulrike von Luxburg and Leena Chennuru Vankadara},
booktitle={The Thirty-ninth Annual Conference on Neural Information Processing Systems},
year={2025},
url={https://openreview.net/forum?id=hTxnm6H93P}
}

@article{sun2025curse,
  title={The curse of depth in large language models},
  author={Sun, Wenfang and Song, Xinyuan and Li, Pengxiang and Yin, Lu and Zheng, Yefeng and Liu, Shiwei},
  journal={arXiv preprint arXiv:2502.05795},
  year={2025}
}

@article{jordan1994hierarchical,
  title={Hierarchical mixtures of experts and the EM algorithm},
  author={Jordan, Michael I and Jacobs, Robert A},
  journal={Neural computation},
  volume={6},
  number={2},
  pages={181--214},
  year={1994},
  publisher={MIT Press}
}

@article{shahbaba2009nonlinear,
  title={Nonlinear models using Dirichlet process mixtures},
  author={Shahbaba, Babak and Neal, Radford},
  journal={Journal of Machine Learning Research},
  volume={10},
  number={8},
  year={2009}
}

@article{tresp2000mixtures,
  title={Mixtures of Gaussian processes},
  author={Tresp, Volker},
  journal={Advances in neural information processing systems},
  volume={13},
  year={2000}
}

@article{yang2023spectral,
  title={A spectral condition for feature learning},
  author={Yang, Greg and Simon, James B and Bernstein, Jeremy},
  journal={arXiv preprint arXiv:2310.17813},
  year={2023}
}

@inproceedings{glorot2010understanding,
  title={Understanding the difficulty of training deep feedforward neural networks},
  author={Glorot, Xavier and Bengio, Yoshua},
  booktitle={Proceedings of the thirteenth international conference on artificial intelligence and statistics},
  year={2010},
  organization={JMLR Workshop and Conference Proceedings}
}

@inproceedings{
shazeer2017,
title={ Outrageously Large Neural Networks: The Sparsely-Gated Mixture-of-Experts Layer},
author={Noam Shazeer and Azalia Mirhoseini and Krzysztof Maziarz and Andy Davis and Quoc Le and Geoffrey Hinton and Jeff Dean},
booktitle={International Conference on Learning Representations (ICLR)},
year={2017},
}

@article{jiang2024mixtral,
  title={Mixtral of experts},
  author={Jiang, Albert Q and Sablayrolles, Alexandre and Roux, Antoine and Mensch, Arthur and Savary, Blanche and Bamford, Chris and Chaplot, Devendra Singh and Casas, Diego de las and Hanna, Emma Bou and Bressand, Florian and others},
  journal={arXiv preprint arXiv:2401.04088},
  year={2024}
}

@InProceedings{pmlr-v162-clark22a,
  title = 	 {Unified Scaling Laws for Routed Language Models},
  author =       {Clark, Aidan and De Las Casas, Diego and Guy, Aurelia and Mensch, Arthur and Paganini, Michela and Hoffmann, Jordan and Damoc, Bogdan and Hechtman, Blake and Cai, Trevor and Borgeaud, Sebastian and Van Den Driessche, George Bm and Rutherford, Eliza and Hennigan, Tom and Johnson, Matthew J and Cassirer, Albin and Jones, Chris and Buchatskaya, Elena and Budden, David and Sifre, Laurent and Osindero, Simon and Vinyals, Oriol and Ranzato, Marc'Aurelio and Rae, Jack and Elsen, Erich and Kavukcuoglu, Koray and Simonyan, Karen},
  booktitle = 	 {Proceedings of the 39th International Conference on Machine Learning},
  year = 	 {2022},
  volume = 	 {162},
  series = 	 {Proceedings of Machine Learning Research},
  publisher =    {PMLR},
}

@misc{olmo2025olmo3,
title={Olmo 3},
author={Team Olmo and Allyson Ettinger and Amanda Bertsch and Bailey Kuehl and David Graham and David Heineman and Dirk Groeneveld and Faeze Brahman and Finbarr Timbers and Hamish Ivison and Jacob Morrison and Jake Poznanski and Kyle Lo and Luca Soldaini and Matt Jordan and Mayee Chen and Michael Noukhovitch and Nathan Lambert and Pete Walsh and Pradeep Dasigi and Robert Berry and Saumya Malik and Saurabh Shah and Scott Geng and Shane Arora and Shashank Gupta and Taira Anderson and Teng Xiao and Tyler Murray and Tyler Romero and Victoria Graf and Akari Asai and Akshita Bhagia and Alexander Wettig and Alisa Liu and Aman Rangapur and Chloe Anastasiades and Costa Huang and Dustin Schwenk and Harsh Trivedi and Ian Magnusson and Jaron Lochner and Jiacheng Liu and Lester James V. Miranda and Maarten Sap and Malia Morgan and Michael Schmitz and Michal Guerquin and Michael Wilson and Regan Huff and Ronan Le Bras and Rui Xin and Rulin Shao and Sam Skjonsberg and Shannon Zejiang Shen and Shuyue Stella Li and Tucker Wilde and Valentina Pyatkin and Will Merrill and Yapei Chang and Yuling Gu and Zhiyuan Zeng and Ashish Sabharwal and Luke Zettlemoyer and Pang Wei Koh and Ali Farhadi and Noah A. Smith and Hannaneh Hajishirzi},
year={2025},
eprint={2512.13961},
archivePrefix={arXiv},
primaryClass={cs.CL},
url={https://arxiv.org/abs/2512.13961},
}

@article{marchenko1967distribution,
  author  = {Marchenko, V. A. and Pastur, L. A.},
  title   = {Distribution of eigenvalues for some sets of random matrices},
  journal = {Mat. Sbornik (N.S.)},
  volume  = {72(114)},
  number  = {4},
  pages   = {507--536},
  year    = {1967}
}

@book{bai2010spectral,
  author    = {Bai, Zhidong and Silverstein, Jack W.},
  title     = {Spectral Analysis of Large Dimensional Random Matrices},
  publisher = {Springer},
  edition   = {2},
  series    = {Springer Series in Statistics},
  year      = {2010}
}

@book{anderson2010introduction,
  author    = {Anderson, Greg W. and Guionnet, Alice and Zeitouni, Ofer},
  title     = {An Introduction to Random Matrices},
  publisher = {Cambridge University Press},
  series    = {Cambridge Studies in Advanced Mathematics},
  volume    = {118},
  year      = {2010}
}

@article{Hubbard,
  author  = {Hubbard, J.},
  title   = {Calculation of Partition Functions},
  journal = {Physical Review Letters},
  volume  = {3},
  number  = {2},
  pages   = {77--78},
  year    = {1959}
}

@article{Stein,
  author  = {Stein, Charles M.},
  title   = {Estimation of the Mean of a Multivariate Normal Distribution},
  journal = {The Annals of Statistics},
  volume  = {9},
  number  = {6},
  pages   = {1135--1151},
  year    = {1981}
}

@article{martin1973statistical,
  author  = {Martin, Paul Cecil and Siggia, E. D. and Rose, H. A.},
  title   = {Statistical Dynamics of Classical Systems},
  journal = {Physical Review A},
  volume  = {8},
  number  = {1},
  pages   = {423},
  year    = {1973}
}

@article{dedominicis1976techniques,
  author  = {De Dominicis, C.},
  title   = {Techniques de renormalisation de la th{\'e}orie des champs et dynamique des ph{\'e}nom{\`e}nes critiques},
  journal = {Journal de Physique Colloques},
  volume  = {37},
  number  = {C1},
  pages   = {C1-247--C1-253},
  year    = {1976},
  doi     = {10.1051/jphyscol:1976138}
}

@article{janssen1976lagrangean,
  title={On a Lagrangean for classical field dynamics and renormalization group calculations of dynamical critical properties},
  author={Janssen, Hans-Karl},
  journal={Zeitschrift f{\"u}r Physik B Condensed Matter},
  volume={23},
  number={4},
  pages={377--380},
  year={1976},
  publisher={Springer}
}
\bibliographystyle{plainnat_clean.bst}

\newpage

\begin{appendices}
\listofappendices

\counterwithin{figure}{section}
\counterwithin{table}{section}
\counterwithin{equation}{section}

\crefalias{section}{appendix}
\crefalias{subsection}{appendix}

\newpage

\part{Quick Reference}

This part serves as a concise entrypoint to the appendix that enables the reader to orient themselves quickly and to navigate the subsequent parts in a structured way. \Cref{sec:roadmap} provides a roadmap and suggested reading paths depending on the reader's background and interests.

For practitioners, \Cref{sec:complete_parameterization_mixtral} concisely provides our \textit{complete MSSP scaling prescription for modern transformer MoE architectures}.

\section{Appendix Roadmap and Suggested Reading Paths}
\label{sec:roadmap}

The remaining parts of this appendix are self-contained and can be summarized follows:
\begin{itemize}
\item \textbf{Part~II (Background and Extended Related Work)} provides background on the established infinite width literature, in particular the most common $abc$-parameterizations for standard dense networks, necessary to understand our corrected parameterizations for MoEs (\Cref{sec:background}) and a detailed account of related work (\Cref{sec:related_work_app}).
\item \textbf{Part~III (Theory)} provides three complementary levels of formalism: an intuitive scaling explanation (\Cref{sec:intuitive_scaling_explanation}), followed by a full forward and backward signal propagation analysis for deriving the $\mu$P and MSSP hyperparameter scaling rules in each scaling regime (\Cref{sec:heuristic_derivation}) followed by a rigorous self-consistent DMFT for $\mu$P and MSSP in each of the three scaling regimes (Appendices~\ref{sec:dmft_regime1} to \ref{sec:dmft-regime-iii-mup}). Note that Regimes II and III provide differing challenges that require their own resolutions, both in terms of scaling and analysis technique. Whenever the DMFT equations for a differing parameterization our routing mechanism follows from an analogous derivation, we omit the duplication and just provide the final set of self-consistent DMFT equations.
\item \textbf{Part~IV (Experiments)} contains the experimental setup for our MLP MoE and Transformer MoE experiments (\Cref{sec:exp_setup}), followed by extensive empirical evidence for our claims about learning rate transfer and scaling properties of $\mu$P and MSSP for both SGD and Adam in all 3 scaling regimes (\Cref{sec:add_exper}). We close with further empirical scaling insights that practitioners should be aware of in Sections~\ref{sec:exp_softmax_collapse} to \ref{sec:global_eps}. 
\end{itemize}

\textbf{If you only have 30 minutes:} read (i) \Cref{sec:complete_parameterization_mixtral} for our complete Mixtral-style MSSP recipe, and (ii)~\Cref{ssec:summary} for the regime-by-regime summary of how $\mu$P breaks and how MSSP fixes it.

\textbf{For conceptual understanding}: We recommend familiarizing oneself with the background on width scaling (\Cref{sec:background}), then reading \Cref{sec:intuitive_scaling_explanation} for a comprehensive introduction into the scaling arguments necessary for understanding the signal propagation pathologies of MoEs and how to fix them. These arguments will provide the reader with tools that are more broadly applicable to similar scaling derivations for other non-standard architectures and optimizers. We then recommend reading the complete MSSP recipe for modern MoEs (\Cref{sec:complete_parameterization_mixtral}), and ending with a quick pass over the empirical results in \Cref{sec:add_exper}. In particular, we recommend skipping Appendices~\ref{sec:heuristic_derivation} to \ref{sec:exp_setup} during this first read.

\textbf{Deep dive:} For the ambitious reader, we recommend starting with the above path for generating sufficient conceptual understanding, before diving into the detailed signal propagation analyses in \Cref{sec:heuristic_derivation}, and the DMFT analyses in Appendices~\ref{sec:dmft_regime1} to \ref{sec:dmft-regime-iii-mup}. The signal propagation analysis decomposes all terms in the training dynamics and determines their scaling with arguments from random matrix theory. This makes the order of contributions of all observables explicit. The DMFT provides complementary value by clarifying how different variables interact via evolution of their kernels. 

\textbf{Practitioners:} For readers that are primarily interested in the practical takeaways, we recommend starting with our complete Mixtral-style MSSP recipe in \Cref{sec:complete_parameterization_mixtral}. Our Transformer setup is introduced in \Cref{sec:transformer_setup} and our empirical findings are provided in \Cref{sec:add_exper}.

\textbf{Notation mapping.} The notation in the appendix differs slightly from the main paper. For readability, we write out the theory for a single block, but, as is common in the related DMFT and $\mu$P literature \citep{tp5_2022,jiang2026hyperparameter}, every derivation follows in the same way for the architecture that stacks multiple blocks that is considered in the main paper. A detailed explanation is provided in \Cref{ssec:ssp_r2_depth}.

Recall the following main paper notation.

\textbf{MoE architecture.} Given input $x_t\in\mathbb{R}^D$, the residual stream $h^0_t = W^{\mathrm{in}}_t x_t$ is transformed as:
\begin{equation*}
    h^{l}_t = h^{l-1}_t + K^{-\alpha_{\mathrm{agg}}}\textstyle\sum_{i\in\text{top-K}}\phi^l_{t,i}\cdot W^{l,\mathrm{out},i}_t\,\varphi(W^{l,\mathrm{in},i}_t h^{l-1}_t),
    \qquad \phi^l_{t,i} = \sigma\!\big(\beta\,(Q^l_t h^{l-1}_t)_i\big), 
\end{equation*}
for $l\in[L]$, and is finally transformed into output logits $f_t = (W_t^{\mathrm{out}})^\top h^L_t$. Here $\varphi$ is a coordinatewise nonlinearity, $\beta$ a tunable inverse temperature, and $\sigma$ either sigmoid or softmax.

For theoretical purposes, it suffices to analyze the signal propagation through the following single MoE block without residual connection.

\textbf{Minimal MoE architecture.}
For readability, we work with the minimal MoE architecture
\begin{align}
    \label{eq:architecture_appendix}
    h^1      &= W^1 x,                                                      \\
    \psi     &= Q h^1,                                                      \\
    \phi     &= \sigma(\psi),                                               \\
    h^{2,i}  &= W^{2,i} h^1,                   \qquad i = 1,\dots,M,        \\
    h^{3,i}  &= W^{3,i} h^{2,i},                                            \\
    h^3      &= \frac{1}{M}\sum_{i=1}^M \phi_i\, h^{3,i},                   \\
    f        &= W^4 h^3,
\end{align}
where $x \in \mathbb{R}^D$ is the input, $f \in \mathbb{R}$ the scalar output, and the trainable parameters are
\[
\theta \;=\; \bigl(W^1,\; Q,\; \{W^{2,i},\, W^{3,i}\}_{i=1}^M,\; W^4\bigr)
\]
with shapes $W^1 \in \mathbb{R}^{N\times D}$, $Q \in \mathbb{R}^{M\times N}$, $W^{2,i} \in \mathbb{R}^{N_e\times N}$, $W^{3,i} \in \mathbb{R}^{N\times N_e}$, $W^4 \in \mathbb{R}^{1\times N}$. The same scaling exponents hold under sigmoid gating with aggregation factor $\alpha_{\text{agg}}=1$ and under softmax gating without an explicit aggregation multiplier.

In particular, here we write $h^1$ instead of $h^l$, and $h^3 = \frac{1}{M}\sum_{i=1}^M \phi_i\, h^{3,i}$ instead of $h^{l+1}=h^l+h^{\text{agg}}$.

\newpage
\section{Complete Maximally Scale-Stable Parameterization for Mixtral-Style MoEs}
\label{sec:complete_parameterization_mixtral}

\begin{table}[htbp!]
\centering
\caption{\textbf{MoE AdamW hyperparameter scaling.} Columns correspond to three MoE scaling regimes.
\w{Width} and \dpt{depth} control terms are highlighted. For example in the bottleneck Regime II, width $N$, number of experts $M$ and top-$K$ are all scaling proportionally, thus could be replaced by the width multiplier ${m_N}=N/N_0$ in relation to some base width $N_0$. 
In AdamW as proposed by \citet{loshchilov2017decoupled}, weight decay should stay fixed across model scales. In the PyTorch implementation of AdamW, the weight decay multiplier should always be the inverse of the learning rate multiplier so that the effective weight decay $\eta \cdot \text{wd}$ stays scale-independent. The minimal fine-graining of multiplier tuning at small model size recommended includes a global initialization multiplier, 
and learning rate multipliers based on layer type (embedding, readout, gain, router, expert in, expert out, other).\\ $\ast$: Router zero initialization requires initial randomness in the routing mechanism.\\ \texttt{tied}: \textbf{In Regime III, expert weights should be shared at initialization.}}
\label{tab:moe_mixtral_summary_template}

\small 

\setlength{\tabcolsep}{6pt}
\renewcommand{\arraystretch}{1.15}

\adjustbox{max width=\textwidth}{\begin{tabular}{>{\raggedright\arraybackslash}p{0.36\linewidth}
                >{\centering\arraybackslash}p{0.2\linewidth}
                >{\centering\arraybackslash}p{0.22\linewidth}
                >{\centering\arraybackslash}p{0.22\linewidth}}
\toprule
\textbf{Parameterization} &
\textbf{Regime I} &
\textbf{Regime II} &
\textbf{Regime III} \\
\addlinespace[2pt]
& \makecell[c]{\scriptsize $\w{N,N_e\to\infty}$\\[-1pt]\scriptsize $M,K$ fixed}
& \makecell[c]{\scriptsize $\w{N,M,K\to\infty}$\\[-1pt]\scriptsize $N_e$ fixed}
& \makecell[c]{\scriptsize $\w{N,M,K,N_e\to\infty}$} \\
\midrule

\textbf{Emb. Init. Std.} &
$d_{\rm in}^{-1/2}$ & $d_{\rm in}^{-1/2}$ & $d_{\rm in}^{-1/2}$ \\

\textbf{Emb. Adam LR} &
$d_{\rm in}^{-1}$ & $d_{\rm in}^{-1}$ & $d_{\rm in}^{-1}$ \\

\textbf{Emb. Adam $\epsilon$} &
$\w{N^{-1}}$ & $\w{N^{-1}}$ & $\w{N^{-1}}$ \\

\midrule

\textbf{Pre-LN Init. Std.} &
$1$ & $1$ & $1$ \\

\textbf{Pre-LN Adam LR} &
$1$ & $1$ & $1$ \\

\textbf{Pre-LN Adam $\epsilon$} &
$\w{N^{-1}}\dpt{L^{-1}}$ & $\w{N^{-1}}\dpt{L^{-1}}$ & $\w{N^{-1}}\dpt{L^{-1}}$ \\

\midrule

\textbf{Hidden Init. Std.} &
$\w{N^{-1/2}}$ & $\w{N^{-1/2}}$ & $\w{N^{-1/2}}$ \\

\textbf{Hidden Adam LR} &
$\w{N^{-1}}$ & $\w{N^{-1}}$ & $\w{N^{-1}}$ \\

\textbf{Hidden Bias Adam LR} &
$1$ & $1$ & $1$ \\

\textbf{Hidden Adam $\epsilon$} &
$\w{N^{-1}}\dpt{L^{-1}}$ & $\w{N^{-1}}\dpt{L^{-1}}$ & $\w{N^{-1}}\dpt{L^{-1}}$ \\

\midrule

\multicolumn{4}{l}{\textbf{MoE routing \& experts}}\\
\addlinespace[2pt]

\textbf{Router (gating) Init. Std.} &
$0^\ast$ & $\w{N^{-1/2}}$ & $\w{N^{-1/2}}$ \\

\textbf{Router (gating) Adam LR} &
$\w{N^{-1}}$ & $\w{N^{-1}}$ & $\w{N^{-1}}$ \\

\textbf{Router Adam $\epsilon$} &
$\dpt{L^{-1}}$ & $\w{M^{-1}}\dpt{L^{-1}}$ & $\w{M^{-1}}\dpt{L^{-1}}$ \\

\addlinespace[2pt]

\textbf{Expert Layer 1 Init. Std.} &
$\w{N^{-1/2}}$ & $\w{N^{-1/2}}$ & $\w{(N^{-1/2})^{\texttt{tied}}}$ \\

\textbf{Expert Layer 1 Adam LR} &
$\w{N^{-1}}$ & $\w{N^{-1}}$ & $\w{N^{-1}}$ \\

\textbf{Expert Layer 1 Adam $\epsilon$} &
$\w{N^{-1}}\dpt{L^{-1}}$ & $\w{M^{-1}}\dpt{L^{-1}}$ & $\w{N^{-1}M^{-1}}\dpt{L^{-1}}$ \\

\addlinespace[2pt]

\textbf{Expert Layer 2 Init. Std.} &
$\w{N_e^{-1/2}}$ & $\w{M^{1/2}}N_e^{-1/2}$ & $\w{(N_e^{-1/2})^{\texttt{tied}}}$ \\

\textbf{Expert Layer 2 Adam LR} &
$\w{N_e^{-1}}$ & ${N_e^{-1}}$ & $\w{N_e^{-1}}$ \\

\textbf{Expert Layer 2 Adam $\epsilon$} &
$\w{N^{-1}}\dpt{L^{-1}}$ & $\w{N^{-1}M^{-1}}\dpt{L^{-1}}$ & $\w{N^{-1}M^{-1}}\dpt{L^{-1}}$ \\

\addlinespace[2pt]

\textbf{Aggregation multiplier} &
$K^{-1}$ & $\w{K^{-1}}$ & $\w{K^{-1}}$ \\

\addlinespace[2pt]

\textbf{Aux load-balancing loss multiplier} &
$1$ & $1$ & $1$ \\

\textbf{Router z-loss multiplier} &
$1$ & $1$ & $1$ \\

\midrule

\textbf{MHA Residual} &
$X^l+ \dpt{L^{-1}}\cdot MHA(LN(X^l))$ &
$X^l+ \dpt{L^{-1}}\cdot MHA(LN(X^l))$ &
$X^l+ \dpt{L^{-1}}\cdot MHA(LN(X^l))$ \\

\textbf{MoE FFN Residual} &
$X^l+ \dpt{L^{-1}}\cdot MoE(LN(X^l))$ &
$X^l+ \dpt{L^{-1}}\cdot MoE(LN(X^l))$ &
$X^l+ \dpt{L^{-1}}\cdot MoE(LN(X^l))$ \\

\midrule

\textbf{Final-LN Init. Std.} &
- & - & - \\

\textbf{Final-LN Adam LR} &
$1$ & $1$ & $1$ \\

\textbf{Final-LN Adam $\epsilon$} &
$\w{N^{-1}}$ & $\w{N^{-1}}$ & $\w{N^{-1}}$ \\

\midrule

\textbf{Unemb. Init. Std.} &
$\w{N^{-1}}$ & $\w{N^{-1}}$ & $\w{N^{-1}}$ \\

\textbf{Unemb. Adam LR} &
$\w{N^{-1}}$ & $\w{N^{-1}}$ & $\w{N^{-1}}$ \\

\textbf{Unemb. AdamW $\epsilon$} &
1 & 1 & 1 \\

\textbf{Unemb. Fwd.} &
1 & 1 & 1 \\

\bottomrule
\end{tabular}}

\end{table}

\newpage

\part{Background and Extended Related Work}

This part collects background on existing width-scaling parameterizations and a detailed account of related-work.

\vspace{-2mm}
\section{Background on Width-scaling Parameterizations}\label{sec:background}

\vspace{-2mm}
\Cref{tab:abc} gives an overview over the most common $abc$-parameterizations for training dense networks with SGD. \citet{yang_feature_2021} use width-dependent weight multipliers $n^{-a}\cdot W$ to scale gradients and avoid layerwise learning rates. This extra degree of freedom introduces an equivalence class of valid $\mu$P $abc$-parameterizations. Mean field parameterization \citep{mei2018mean,chizat2018global,bordelon2022self} is equivalent to $\mu$P, using the multipliers \texttt{fan-in}$^{-1/2}$ except $N^{-1}$ in the output layer. For practical readability, we remove this extra degree of freedom in the main paper by using naive weight multipliers $a=0$ for all trainable weights.

\begin{table}[H]
    \centering
    \adjustbox{max width=\textwidth}{\begin{tabular}{lcccccccc}
        \toprule
        & & & \multicolumn{3}{c}{\textbf{Weight-multiplier version}} & \multicolumn{3}{c}{\textbf{Weight-multiplier-free version}}\\
        & & & Input-like & Hidden-like & Output-like & Input-like & Hidden-like & Output-like \\
        \hline\hline
        \multirow{3}{*}{SP} & $\alpha_l\cdot W^l,$ & $\alpha_l\propto$ &  & \multirow{3}{*}{-} & & $1$ & $1$ & $1$\\
                             & $\mathcal{N}(0,\sigma^2_l),$ & $\sigma_l\propto$ & & & & $1$ & $N^{-1/2}$ & $N^{-1/2}$\\
                             & $\eta_l\cdot \nabla_{W^l}\mathcal{L},$ & $\eta_l\propto$ & & & & $N^{-c}$ & $N^{-c}$ & $N^{-c}$\\
        \hline
        \multirow{3}{*}{NTP} & $\alpha_l\cdot W^l,$ & $\alpha_l\propto$ & $1$ & $N^{-1/2}$ & $N^{-1/2}$ & $1$ & $1$ & $1$\\
                             & $\mathcal{N}(0,\sigma^2_l),$ & $\sigma_l\propto$ & $1$ & $1$ & $1$ & $1$ & $N^{-1/2}$ & $N^{-1/2}$\\
                             & $\eta_l\cdot \nabla_{W^l}\mathcal{L},$ & $\eta_l\propto$ & $1$ & $1$ & $1$ & $1$ & $N^{-1}$ & $N^{-1}$\\
        \hline
        \multirow{3}{*}{$\mu$P} & $\alpha_l\cdot W^l,$ & $\alpha_l\propto$ & $N^{1/2}$ & $1$ & $N^{-1/2}$ & $1$ & $1$ & $1$\\
                             & $\mathcal{N}(0,\sigma^2_l),$ & $\sigma_l\propto$ & $N^{-1/2}$ & $N^{-1/2}$ & $N^{-1/2}$ & $1$ & $N^{-1/2}$ & $N^{-1}$\\
                             & $\eta_l\cdot \nabla_{W^l}\mathcal{L},$ & $\eta_l\propto$ & $1$ & $1$ & $1$ & $n$ & $1$ & $N^{-1}$\\
    \bottomrule
    \end{tabular}}
    \caption{\textbf{(Common $abc$-parameterizations)} Standard parameterization (SP), neural tangent parameterization (NTP) and the maximal update parameterization ($\mu$P) for MLPs trained with SGD in their multiplier version which purely adapts the architecture and allows width-independent global learning rates (\textit{left}) and in their weight multiplier-free version (\textit{right}). Parameterizations differ in their layerwise choice of width-dependent weight multipliers $\alpha_l$, initialization variances $\sigma_l$ and learning rates $\eta_l$. Weight multiplier-free representatives of an $abc$-equivalence class purely adapt the optimization algorithm highlighting the fact that parameterizations effectively only induce layerwise learning rates. Knowing that $\mu$P correctly scales the updates in all layers, observe that the input- and hidden-layer learning rates in NTP induce vanishing updates. The same holds in SP when choosing $c\geq 1$ as is necessary for avoiding logit blowup in the infinite-width limit. SP with large learning rates $c=1/2$ recovers stable hidden-layer feature learning despite logit divergence \citep{haas2025largelrs}.
    \vspace{-2mm}}
    \label{tab:abc}
\end{table}

\begin{wraptable}[12]{r}{0.5\textwidth}
\centering
\vspace{-5mm}
\caption{\textbf{$\mu$P-heuristic} $(b,c_{\text{SGD}}, c_{\text{Adam}},d_{\text{Adam}})$ by layer type. Both SGD and Adam share the same initialization $\mathcal{N}(0,N^{-b})$. SGD requires learning rates $\eta_N=N^{-c_{\text{SGD}}}$, and Adam requires learning rates $\eta_N=N^{-c_{\text{Adam}}}$ and\\ gradient scaling $N^{-d_{\text{Adam}}}\cdot \nabla_{W}\mathcal{L}$.}
\label{tab:mup_heuristic}
\begin{tabular}{l c c c c}
\toprule
 & \textbf{$b$} & \textbf{$c_{\text{SGD}}$} & \textbf{$c_{\text{Adam}}$} & \textbf{$d_{\text{Adam}}$} \\
\midrule
Input-like   & 0   & -1 & 0 & 1 \\
Matrix-like  & 0.5 & 0 & 1 & 1 \\
Output-like  & 1   & 1 & 1 & 0 \\
\bottomrule
\end{tabular}
\end{wraptable}
\textbf{$\mu$P for Adam in standard architectures.} Adam requires the same layerwise initialization variances as SGD. For clarity, consider the weight-multiplier-free version of $\mu$P. Since Adam normalizes its update, but the update direction remains correlated with the incoming activations, Adam's layerwise learning rate scaling simplifies to $\eta(W)=1/\texttt{fan-in}(W)$ for any linear weight matrix $W: \mathbb{R}^{\texttt{fan-in}}\to\mathbb{R}^{\texttt{fan-out}}$. For faithfulness, Adam $\eps(W)$ should always scale as the gradient RMS norm of $W$, that is $\Theta(1)$ for the readout layer and $\Theta(N^{-1})$ otherwise.

\textbf{Baseline Parameterizations: SP and the \texorpdfstring{$\mu$}{mu}P heuristic.}
Standard parametrization (SP) uses Xavier/Glorot-type variance scaling \citep{glorot2010understanding,he_delving_2015}, i.e.\ typical dense layers take $b_W=\tfrac12$ (with common exceptions for input embeddings), and uses the same optimizer hyperparameters such as learning rate and Adam $\eps$ for all layers.

In contrast, $\mu$P is designed so that maximal-update behavior (and, for adaptive methods, faithfulness) hold in the infinite-width limit for dense networks \citep{yang_feature_2021,yang_tp4b_2023}.
Practitioners often use the following \emph{$\mu$P heuristic}: classify a weight tensor by whether it maps
fixed$\to n$ (\emph{input-like}), $n\to n$ (\emph{matrix-like}), or $n\to$fixed (\emph{output-like}), and assign $(b,c,d)$ accordingly as summarized in \Cref{tab:mup_heuristic}. 
For MoEs in Regime I, this heuristic treats expert MLP weights matrix-like, while router projections are output-like due to the fixed number of experts $M$. In Regime I, $\mu$P heuristic indeed satisfies all $\mu$ desiderata, but it does not suffice in Regimes II and III, when co-scaling the number of experts $M\to\infty$.

\section{Detailed Related Work}\label{sec:related_work_app}

\vspace{-2mm}
\textbf{Mixture of Experts models.} Mixture of experts models have a long history in machine learning \citep{jacobs1991adaptive}, with instantiations such as hierarchical models \citep{jordan1994hierarchical}, Gaussian processes \citep{tresp2000mixtures} or Dirichlet processes \citep{shahbaba2009nonlinear}. After the first application in deep networks by \citet{eigen2013learning}, \citet{shazeer2017} sparked the use of MoEs in modern architectures, using the top-$k$ operation for sparsification and trainable noise for load balancing. \citet{fedus2022switch} lowered communication and computational cost for improved scalability, using Kaiming initialization and simplifying the load balancing auxiliary loss. Most MoE architectures still use a width-independent initialization of $0.006$ \citep{dai2024deepseekmoe}. The trend in recent frontier models goes toward increased fine-graining, from 2 out of 8 in Mixtral 8x7B and 8x22B \citep{jiang2024mixtral} to 8 out of 256 in Deepseek-R1 \citep{guo2025deepseek}. \citet{zoph2022st} introduce the router z-loss for router logit regularization; \citet{haas2025largelrs} suggest that miss-scaling causes logit divergence, hence the additional regularization might be unnecessary in MSSP. Insightful empirical investigations can be found in the fully open OlMoE report \citep{muennighoff2024olmoe}. They identify removing the need for the load balancing loss as an important future direction, since it significantly constrains the model flexibility and may prevent experts from sufficiently specializing over the course of training. To date, the theory of MoEs is very limited. \citet{chen2022towards} show that MoEs are able to learn data with cluster structure using two-layer nonlinear convolutional neural networks as experts.

\textbf{Scaling theory.} Infinite-width theory dates back to \citet{neal_priors_1996,matthews_gaussian_2018,jacot_neural_2018}. The Tensor Program series \citep{yang_tp1_2019,yang_feature_2021,yang_tp4b_2023} has served as a crucial tool for developing flexible and general infinite-width theory allowing to study non-vanishing feature learning limits \citep{everett2024scaling,vankadara24ssms,haas2025largelrs}. This theory gave rise to the maximal update parameterization ($\mu$P) \citep{tp5_2022} under which the optimal learning rate and further dynamical properties \citep{noci2024super} have been observed to transfer to larger model width. The mean-field parameterization is equivalent to $\mu$P. DMFT  has proven useful as it allows to simulate from the limit \citep{chizat2018global,mei2018mean,bordelon2022self,bordelon2024infinite}. Similar techniques allow infinite depth limits \citep{bordelon2023depthwise,hayou2023width,noci2024shaped,chaintron2026resnets}. These works apply depth-dependent scaling factors like $L^{-1/2}$ or $L^{-1}$ \citep{dey2025don} to the residual stream; it remains open whether these interventions suffice. \citet{dey2025don} show that $L^{-1}$ outperforms $L^{-1/2}$ in Transformer training, but \citet{sun2025curse} argue that scaling should depend on the layer index.

\textbf{Scaling mixture of experts models.} \citet{pmlr-v162-clark22a,krajewski2024scaling} and \citet[Figure 3]{dai2024deepseekmoe} all suggest that increasingly fine-grained experts outperform fewer larger ones in terms of learning performance, with saturating gains. \citet{boix2025power} prove an exponentially improved expressivity due to fine graining. \citet{mayaki2026generalization} provides theory showing that performance can improve either by scaling active capacity or by increasing the number of experts, depending on the dominant bottleneck. In practice, eventually the increased communication cost of routing dominates compute-optimality considerations, but increasingly efficient implementations are being developed. \citet{nvidiaMoEMegatron2026} provides a highly optimized distributed training framework across scaling regimes, that also allows efficient training with many fine-grained experts. \citet{he2024mixture} proposes PEER layers that take fine graining to the extreme. However, all existing works use suboptimal model scaling, hence it will be crucial to evaluate the impact of granularity on scaling in MSSP.

\textbf{Independent and concurrent work.} \citet{jiang2026hyperparameter} derive a parameterization for \emph{Sign SGD} in Regime III via a DMFT analysis, and present Transformer MoE experiments showing approximate learning rate transfer when scaling one axis at a time with all others fixed. We derive $\mu$P for \emph{SGD and Adam across all three Regimes I, II, and III} using a signal propagation analysis complemented by a DMFT for each regime, and our experiments scale jointly along the axes each regime prescribes. Our $\mu$P in Regime III coincides with theirs up to a larger router initialization. More importantly, we identify that in Regimes II and III, $\mu$P principles do not suffice to robustly yield learning rate transfer or monotonic improvement with scale in MoEs, due to scale-dependent dynamics, and we propose MSSP as a resolution. In Regime III, our DMFT for MSSP is qualitatively distinct: it exhibits a four-layer mean-field hierarchy (versus a three-layer hierarchy under $\mu$P), and we additionally derive novel DMFT limits for Regimes I and II under MSSP. We provide a detailed account of further related work in \Cref{sec:related_work_app}.\looseness=-1

\textbf{Under Regimes II and III ($M,K\to \infty$), $\mu$P differs from shape-based $\mu$P-heuristic.} For dense networks, it is common to scale the layerwise initialization variance and learning rate of a weight matrix purely based on its shape \citep{tp5_2022,yang2023spectral}, distinguishing input-like (fixed $\to\infty$), hidden-like ($\infty\to\infty$) and output-like ($\infty\to$ fixed) layers. \citet{malasnicki2025mu} apply this heuristic to MoEs, treating the router output-like in Regime I ($M,K\in\Theta(1)$). \Cref{tab:moe_mixtral_summary_template_main} shows that $\mu$P-heuristic coincides with $\mu$P in Regime I, but fails to induce all $\mu$-desiderata in Regimes II and III.

\newpage
\part{Theory}

\maketitle

This part of the appendix presents our scaling analysis at three levels of formalism, each progressively more precise. We start with an intuitive explanation in \Cref{sec:intuitive_scaling_explanation}, then provide the detailed signal propagation analysis in \Cref{sec:heuristic_derivation}, and finally provide a full DMFT analysis for each scaling regime in \Cref{sec:dmft_regime1,sec:dmft_regime2,sec:dmft_regime2_mup,sec:dmft_regime3,sec:dmft-regime-iii-mup}.

\section{Intuitive Explanations of the Shortcomings of \texorpdfstring{$\mu$}{mu}P for MoEs and How MSSP Fixes Them}
\label{sec:intuitive_scaling_explanation}

To set the stage for our formal results, we provide intuition for how to derive scaling rules for MoE blocks that satisfy Desiderata~1 and~2. In each scaling regime, these desiderata determine a unique $(b,c,d,\alpha)$-parameterization (up to the rescaling invariance of homogeneous optimizers discussed in Section~\ref{sec:setup}). Table~\ref{tab:moe_mixtral_summary_template_main} reports a canonical parameterization of this equivalence class.
The SGD scaling in each regime is formally justified by the corresponding DMFT derivation. In addition, Adam scaling is heuristically derived but is directly guided by the same DMFT analysis. The conditions under which these parameterizations satisfy Desideratum~3 are discussed in Section~\ref{sec:heuristic_derivation}. Here and throughout, asymptotic notation (e.g., $\Theta(\cdot)$) is understood with respect to the joint scaling trajectory $\mathcal S(n)$ as $n\to\infty$, specialized to the scaling regime (I--III) under consideration. For the intuitive explanation, we focus mainly on Regimes~II and~III. Regime~I is more straightforward and is omitted for brevity. 

\subsection{Overview}
Let's start by recalling the Maximal update desiderata\footnote{We omit an explicit discussion of the optimizer faithfulness desideratum here for brevity but we will briefly discuss $\epsilon$ scaling.}.
\subsubsection{Maximal Update Desiderata}
\label{sec:appendix_desiderata}
The Maximal Update Desiderata require that in each layer, the effective and propagating updates remain $\Theta(1)$ in RMS at some $t>0$:
\begin{equation}
\|\mathcal{T}^\ell_{\mathrm{eff}}(t)\|_{\mathrm{RMS}} = \|\Delta W^\ell_t\,x^{\ell-1}_t\|_{\mathrm{RMS}} =\Theta(1),
\qquad
\|\mathcal{T}^\ell_{\mathrm{prop}}(t)\|_{\mathrm{RMS}} = \|W^\ell_0\,\Delta x^{\ell-1}_t\|_{\mathrm{RMS}} =\Theta(1).
\label{eq:theta1_updates_appendix}
\end{equation}
For the MoE aggregation layer, we additionally require that the updates to the aggregated representation remains non-vanishing and non-diverging:
\begin{equation}
    \|\Delta h^{\mathrm{agg}}_t(t)\|_{\mathrm{RMS}} = \Theta(1).
    \label{eq:theta1_agg}
\end{equation}

\subsubsection{Effective and propagating updates}
\label{ssec:Intuitive_explanation_of_the_scaling_of_effective_and_propagating_updates}
\textbf{Alignment exponents for effective and propagating updates.} In all regimes, the asymptotic scales of forward and backward effective and propagating terms are governed by the training-induced correlation structure between the two factors appearing in each product. Following \citet{yang2023spectral, everett2024scaling}, we summarise this effect via \emph{alignment exponents} $p_\ell$ and $q_\ell$ ($p^\nabla_\ell$ and $q^\nabla_\ell$ analogously for backward), defined by
\begin{equation}
    \begin{aligned}
    \|\Delta W^\ell_t x^{\ell-1}_t\|_{\mathrm{RMS}} &= \Theta\!\left((n_{\mathrm{in}}^\ell)^{p_\ell}\,\|\Delta W^\ell_t\|_{\mathrm{RMS}}\;\|x^{\ell-1}_t\|_{\mathrm{RMS}}\right), \\
    \|W^\ell_0 \Delta x^{\ell-1}_t\|_{\mathrm{RMS}} &= \Theta\!\left((n_{\mathrm{in}}^\ell)^{q_\ell}\,\|W^\ell_0\|_{\mathrm{RMS}}\;\|\Delta x^{\ell-1}_t\|_{\mathrm{RMS}}\right),
    \end{aligned}
    \label{eq:alignment-ratios}
\end{equation}
where $n_{\mathrm{in}}^\ell := \dim(x^{\ell-1})$ denotes the fan-in of layer $\ell$. Informally, $p_\ell$ quantifies the degree of alignment between $\Delta W^\ell_t$ and the current features $x^{\ell-1}_t$, while $q_\ell$ quantifies the degree of alignment between the initial weights $W^\ell_0$ and the upstream feature change $\Delta x^{\ell-1}_t$.

\textbf{From $(p_\ell,q_\ell)$ to scaling rules for optimizers.}
Fix a layer $\ell$ and suppose the alignment exponents $(p_\ell,q_\ell)$ are known. Assume inductively that at some time $t>0$ the previous-layer quantities satisfy $\|x^{\ell-1}_t\|_{\mathrm{RMS}}=\Theta(1)$ and $\|\Delta x^{\ell-1}_t\|_{\mathrm{RMS}}=\Theta(1)$. Enforcing Desideratum~1 at layer $\ell$ then reduces, via \eqref{eq:alignment-ratios}, to choosing scales for $\|W^\ell_0\|_{\mathrm{RMS}}$ and $\|\Delta W^\ell_t\|_{\mathrm{RMS}}$ so that the effective and propagating updates remain $\Theta(1)$. These marginal scales are directly controlled by optimizer hyperparameters: initialization variances set the scaling of $\|W^\ell_0\|_{\mathrm{RMS}}$, while learning-rate scaling sets the scaling of $\|\Delta W^\ell_t\|_{\mathrm{RMS}}$ once the scale of the layerwise gradients is known. For Adam, optimizer faithfulness further requires $\epsilon$ to remain $\Theta(1)$ relative to the typical gradient scale; accordingly, we choose $\epsilon$ to scale with the layerwise gradient RMS norm. {The same reduction applies to the backward pass, with backward alignment exponents $(p^\nabla_\ell, q^\nabla_\ell)$ playing the role of $(p_\ell, q_\ell)$ in setting the leading-order scales of $(W^\ell_0)^\top\delta^\ell_t$ and $(\Delta W^\ell_t)^\top\delta^\ell_t$, where $\delta^\ell_t := \nabla_{h^\ell_t}\mathcal{L}$ denotes the pre-activation gradient passed backward.}

\subsubsection{DMFT predictions.}
Our DMFT analysis provides the regime-specific asymptotic inputs needed to instantiate the above program. It predicts the forward alignment behaviour $(p_\ell,q_\ell)$, the backward alignment behaviour $(p^\nabla_\ell,q^\nabla_\ell)$, and the leading-order scaling of expert-indexed aggregates that enter both $\Delta h^{\mathrm{agg}}(t)$ and the gradients at shared representations when $M$ and $K$ scale. Together, these predictions determine the regime-appropriate scaling of initialization variances, learning rates, and (for Adam) $\epsilon$, yielding the canonical parameterizations reported in Table~\ref{tab:moe_mixtral_summary_template_main}.

\vspace{1em}
{The remainder of this section follows the programme described above in two stages. Section~\ref{ssec:per-layer} treats single linear maps $W \cdot x$ and the alignment exponents $(p_\ell, q_\ell)$ that govern them. Section~\ref{ssec:cross-expert-mechanisms} treats the MoE-specific cross-expert aggregations, organized as a small taxonomy of structural mechanisms that determine whether each aggregation term survives at leading scale or is suppressed by the cross-expert CLT effect. The same taxonomy makes the imbalances under $\mu$P in Regimes~II and~III transparent, and identifies, for each mechanism, the structural change that MSSP introduces to restore balance.}

\subsection{Per-layer alignment: scaling of linear (forward) maps \texorpdfstring{$W\cdot x$}{W.x}}
\label{ssec:per-layer}

Each of the products entering the alignment-exponent decomposition~\eqref{eq:alignment-ratios} is a single linear map $W\,x$, and the prefactors in $\|Wx\|_{\mathrm{RMS}}$ are governed by random-matrix and rank-1-alignment arguments below. We focus mainly on the forward pass; the backward arguments are analogous.

\subsubsection{Random-matrix preliminaries}
\label{ssec:random-matrix-preliminaries}

Let $A \in \mathbb{R}^{n \times n}$ have i.i.d.\ entries $A_{ij} \sim \mathcal{N}(0,\sigma^2)$. Then $A$ acts as a near-isometry rescaled by $\sigma\sqrt{n}$ along almost every direction in $\mathbb{R}^n$: its singular values are $\Theta(\sigma\sqrt{n})$ throughout the Marchenko--Pastur bulk \citep{marchenko1967distribution,bai2010spectral}, and the corresponding singular vectors are uniformly distributed on $S^{n-1}$ \citep[][Ch.~2]{anderson2010introduction}.

The directions along which $A$ fails to act as such an isometry, namely those aligned with its smallest singular values, span only a vanishing fraction of $\mathbb{R}^n$ and form an exponentially thin set on the sphere \citep[e.g.][\S 3.1]{vershynin2018high}. Therefore, for generic $x \in \mathbb{R}^n$ (even when $x$ depends on $A$, provided it is not adversarially chosen to lie in this subspace),
\[
\|Ax\|_2 \;\asymp\; \sigma\sqrt{n}\,\|x\|_2.
\]
Equivalently, in RMS norm,
\[
\|Ax\|_{\mathrm{RMS}} \;\asymp\; \sqrt{n}\,\|A\|_{\mathrm{RMS}}\,\|x\|_{\mathrm{RMS}}.
\]

We refer to this as \emph{CLT-like scaling}: each entry of $Ax$ is a sum of $n$ uncorrelated mean-zero terms, so its typical magnitude grows only as $\sqrt{n}$ relative to the per-term scale, in analogy with the Central Limit Theorem. Later, we will contrast this with \emph{LLN-like scaling}, where the prefactor becomes $n$ rather than $\sqrt{n}$.

\subsubsection{Propagating updates}
\label{sssec:propagating}

\paragraph{Standard MLP hidden layer: CLT-like scaling.}
Applying the arguments above to a standard MLP hidden layer, with initial weights ($W_0$) an i.i.d.\ Gaussian matrix whose fan-in and fan-out both diverge at rate $n$, a propagating perturbation $\Delta x$ inherits this CLT-like scaling:
\[
\|W_0\,\Delta x\|_{\mathrm{RMS}} \;\asymp\; \sqrt{n}\,\|W_0\|_{\mathrm{RMS}}\,\|\Delta x\|_{\mathrm{RMS}}.
\]
Non-trivial feature learning requires the previous layer's updates to satisfy $\|\Delta x\|_{\mathrm{RMS}} = \Theta(1)$, and for the propagating update to remain $\Theta(1)$ after passing through the weight matrix we therefore need
\[
\|W_0\|_{\mathrm{RMS}} \;\asymp\; \tfrac{1}{\sqrt{n}}, \qquad \text{equivalently,}\qquad \sigma \;\asymp\; \tfrac{1}{\sqrt{n}}.
\]

The CLT-like scaling above hinges on $A$ mapping between two diverging dimensions: for an $n \times n$ i.i.d.\ Gaussian, the operator norm is $\Theta(\sigma\sqrt{n})$, which caps the prefactor at $\sqrt{n}$ regardless of how $x$ depends on $A$. No alignment, however structured, can lift the scaling to $n$ in the $\infty\!\to\!\infty$ case. When at least one dimension of $A$ is finite or $A$ has an extremely skewed aspect ratio, this argument no longer applies and alignment-induced LLN scaling becomes possible, although whether it actually occurs depends on the surrounding network dynamics; we shall give some examples below to illustrate this. In the $\infty\!\to\!\infty$ case, $x$ has too few degrees of freedom to align coherently with all $n$ rows of $A$ at once: maximal alignment with any single row boosts only the corresponding entry of $Ax$ to LLN order, while the remaining $n-1$ entries stay at CLT order and dominate the norm. There is no way for one $x$ to align with $n$ independent rows simultaneously.

\vspace{1em}
\textbf{Remark.} In all three regimes, propagating updates (both forward and backward) of all the layers that have diverging input and output dimensions admit CLT-like scaling. For instance, in Regime~III all the layers (including the experts and the router), except the input and the output layer, fall into this category.

\vspace{1em}
\paragraph{Readout layer: the canonical LLN-like scaling.} For the readout layer of a standard network, $W_0 \in \mathbb{R}^{k \times n}$ maps the diverging width $n$ to a fixed output dimension $k$. The perturbation $\Delta x \in \mathbb{R}^n$ entering the layer has been shaped by backpropagation through $W_0^\top$, so its entries acquire systematic alignment with the columns of $W_0$. The inner products $\langle a^{(j)}, \Delta x \rangle$ no longer cancel, and the LLN-like prefactor is realized in the diverging dimension being summed over:
\[
\|W_0\,\Delta x\|_{\mathrm{RMS}} \;\asymp\; n\,\|W_0\|_{\mathrm{RMS}}\,\|\Delta x\|_{\mathrm{RMS}}.
\]

\paragraph{MoE first expert layer (Regime II): alignment broken by expert independence.} The MoE bottleneck regime (Regime~II) is an instructive intermediate case. The first expert layer $W_0^{(i)} \in \mathbb{R}^{N_e \times n}$ maps the diverging width $n$ to the finite expert hidden width $N_e$, geometrically the same situation as a readout layer, so alignment-induced LLN scaling is in principle permitted. Yet the alignment mechanism is broken here: the $M$ expert first-layers $\{W_0^{(i)}\}_{i=1}^{M}$ are independent at initialization, and a single shared input perturbation $\Delta x$ cannot simultaneously align with the column structure of every $W_0^{(i)}$. CLT-like scaling therefore survives,
\[
\|W_0^{(i)}\,\Delta x\|_{\mathrm{RMS}} \;\asymp\; \sqrt{n}\,\|W_0^{(i)}\|_{\mathrm{RMS}}\,\|\Delta x\|_{\mathrm{RMS}}.
\]

\paragraph{MoE second expert layer (Regime II): alignment is restored at finite dimension.} The second expert layer $W_0^{(i,2)} \in \mathbb{R}^{n \times N_e}$ maps the finite hidden width $N_e$ back to the diverging width $n$. Here the relevant sum runs over only $N_e$ terms, so the LLN prefactor cannot exceed $N_e$ regardless of alignment. The perturbation entering this layer is private to expert $i$ and is propagated by training through its own $W_0^{(i,2)\top}$, so the alignment is realized:
\[
\|W_0^{(i,2)}\,\Delta x\|_{\mathrm{RMS}} \;\asymp\; N_e\,\|W_0^{(i,2)}\|_{\mathrm{RMS}}\,\|\Delta x\|_{\mathrm{RMS}} \;\asymp\; \|W_0^{(i,2)}\|_{\mathrm{RMS}}\,\|\Delta x\|_{\mathrm{RMS}} \quad \text{since $N_e \in \Theta(1)$}.
\]
LLN-like in the fixed dimension being summed over.

\subsubsection{Effective updates: rank-1 alignment}
\label{sssec:effective}

The SGD update $\Delta W = -\eta\,(\nabla_{x'} L)\,x^\top$ is rank-1, so applying it to $x$ saturates Cauchy-Schwarz: $\|\Delta W \cdot x\|_2 = \|\Delta W\|_F\,\|x\|_2$, equivalently
\[
    \|\Delta W \cdot x\|_{\mathrm{RMS}} \;\asymp\; n\,\|\Delta W\|_{\mathrm{RMS}}\,\|x\|_{\mathrm{RMS}},
\]
where $n$ is the layer's fan-in. The prefactor $n$ (vs.\ the $\sqrt n$ of CLT) reflects the rank-1 alignment between $\Delta W$ and $x$ due to gradient descent, which precludes any mean-zero cancellation. When the fan-in is fixed (as for the second expert layer in Regime~II, with $N_e = \Theta(1)$), the prefactor is $\Theta(1)$ rather than diverging. By transpose, the backward effective piece $(\Delta W)^\top\delta$ satisfies the same alignment relation with fan-out replacing fan-in. Together with the gradient scale at the layer, the rank-1 relation places a constraint on the SGD learning rate at every layer in any regime; the gradient scale is the only missing input to instantiate it.

In contrast, for Adam, entrywise normalization sets $\Delta W$ entries to $\Theta(\eta_{\mathrm{Adam}})$ regardless of the gradient scale, so the alignment argument alone constrains $\eta_{\mathrm{Adam}}$: the input-side LLN factor gives $\eta_{\mathrm{Adam}} = 1/n$ when the input dimension diverges and $\eta_{\mathrm{Adam}} = 1$ when it is fixed. Setting Adam's $\varepsilon$, however, still requires the gradient scale: $\varepsilon$ must remain $\Theta$ of the typical per-coordinate gradient magnitude (else either $\varepsilon$ dominates and the update becomes trivial, or the gradients dominate and the normalization becomes vacuous), so it scales with the gradient at the layer.

Together with the propagating analysis above, this yields per-layer constraints relating $\sigma$, $\eta$ (and $\varepsilon$ for Adam) to the layerwise gradient scale, which is in turn determined by the cross-expert aggregations of Section~\ref{ssec:cross-expert-mechanisms}. Our goal is a parameterization that simultaneously satisfies these per-layer constraints together with the cross-expert balance constraints below; trade-offs between competing constraints can in principle require expanding the hyperparameter space (e.g., layer-specific scalings) to admit a feasible solution.

\vspace{1em}
{Crucially, the $\mu$P desideratum (that the layerwise effective and propagating terms in the \emph{forward} pass remain $\Theta(1)$) already pins down a unique parameterization: the per-layer initialization variances, learning rates, and (for Adam) $\varepsilon$. We refer to this parameterization as the \emph{$\mu$P baseline}. The alignment-exponent framework~\eqref{eq:alignment-ratios} furnishes the analytical machinery for solving the corresponding $\Theta(1)$ conditions. The $\mu$P desideratum, however, imposes no explicit constraints on the backward pass or on the cross-expert aggregations~\eqref{eq:agg_decomp_appendix},~\eqref{eq:bwd_agg_decomp}; as discussed in §\ref{sec:appendix_desiderata}, balance across the terms of these decompositions constitutes the additional desiderata required of an MoE parameterization. In what follows, we proceed case by case through the $\mu$P baseline, identifying where it produces imbalances under our extended desiderata and showing how MSSP resolves each.}

\subsection{MSSP Desiderata}
For MoE blocks, the analogous requirement applies to each term in the effective/propagating decomposition of the aggregated representation. Writing $W^{3,i}(t)=W^{3,i}(0)+\Delta W^{3,i}(t)$ and $h^{2,i}(t)=h^{2,i}(0)+\Delta h^{2,i}(t)$, the aggregation $h^{\mathrm{agg}}(t)=\sum\nolimits_{i=1}^{M}\phi_i(t)\,W^{3,i}(t)\,h^{2,i}(t)$ decomposes as
\begin{equation}
h^{\mathrm{agg}}(t) =
  \underbrace{\sum\nolimits_{i=1}^{M}\phi_i\,W^{3,i}(0)\,h^{2,i}(0)}_{\text{A1: init.}}
  +\underbrace{\sum\nolimits_{i=1}^{M}\phi_i\,W^{3,i}(0)\,\Delta h^{2,i}(t)}_{\text{A2: propagating}}
  +\underbrace{\sum\nolimits_{i=1}^{M}\phi_i\,\Delta W^{3,i}(t)\,h^{2,i}(t)}_{\text{A3: effective}}.
  \label{eq:agg_decomp_appendix}
\end{equation}
{The MoE-specific extension of Desideratum~1 requires each of these three contributions (init., propagating, effective) to remain $\Theta(1)$ in RMS at some $t>0$. Expanding the propagating term one step further (substituting $\Delta h^{2,i}(t) \approx W^{2,i}(0)\,\Delta h^1 + \Delta W^{2,i}(t)\,h^1(0)$) yields the four-term decomposition $h^{\mathrm{agg}}(t) = A_1 + A_{2,1} + A_{2,2}' + A_3$ which will be used in the case studies below.}

{\textbf{Backward desiderata.} Recall that the same structure carries over to the backward pass. Along any linear map $h^\ell_t = W^\ell_t\,x^{\ell-1}_t$, writing $\bar\delta^\ell_t := \nabla_{x^\ell_t}\mathcal{L}$ and $\delta^\ell_t := \nabla_{h^\ell_t}\mathcal{L}$, the transpose recursion $\bar\delta^{\ell-1}_t = (W^\ell_t)^\top \delta^\ell_t$ admits the analogous effective/propagating split}
\begin{equation}
\bar\delta^{\ell-1}_t \;=\; \underbrace{(W^\ell_0)^\top\,\delta^\ell_t}_{\text{propagating}} \;+\; \underbrace{(\Delta W^\ell_t)^\top\,\delta^\ell_t}_{\text{effective}}.
\label{eq:bwd_layer_decomp}
\end{equation}
{When backpropagation crosses a representation that fans out into $M$ expert branches, the gradient at the shared node aggregates expert contributions in the same form as the forward aggregation. Writing $g^{2,i}(t) := \nabla_{h^{2,i}(t)}\mathcal{L}$ for the per-expert gradient at the bottleneck representation, the expert-pathway contribution to the gradient at the input shared across experts is $\bar\delta^{1,\mathrm{exp}}(t) = \sum\nolimits_{i=1}^{M}\phi_i(t)\,(W^{2,i}(t))^\top\,g^{2,i}(t)$, and expanding $W^{2,i}(t) = W^{2,i}(0) + \Delta W^{2,i}(t)$ and $g^{2,i}(t) = g^{2,i}(0) + \Delta g^{2,i}(t)$ gives}
\begin{equation}
\adjustbox{max width=\textwidth}{$
\bar\delta^{1,\mathrm{exp}}(t) \;=\;
  \underbrace{\sum\nolimits_{i=1}^{M}\,(W^{2,i}(0))^\top\,g^{2,i}(0)}_{\text{A4: init.}}
  +\underbrace{\sum\nolimits_{i=1}^{M}\,(W^{2,i}(0))^\top\,\Delta g^{2,i}(t)}_{\text{A5: propagating}}
+\underbrace{\sum\nolimits_{i=1}^{M}\,(\Delta W^{2,i}(t))^\top\,g^{2,i}(t)}_{\text{A6: effective}}
$}
  \label{eq:bwd_agg_decomp}
\end{equation}
{Desideratum~1 extends naturally to the backward pass: each piece of the backward decomposition (the layerwise effective/propagating split \eqref{eq:bwd_layer_decomp} and each term in the expert-aggregation decomposition \eqref{eq:bwd_agg_decomp}) is required to be \emph{balanced}, i.e.\ to share a common scale within each decomposition. (Under an appropriate per-layer normalization that places backward signals on a scale comparable to forward signals, the common scale is $\Theta(1)$; since this normalization depends on architecture, algorithm, and layer type, we work with the normalization-free \emph{balance} statement throughout, and omit the explicit normalization.)}

\subsection{Backward scaling} 
The forward arguments above transpose directly to the backward pass at every layer with diverging input and output dimensions: $(W^\ell_0)^\top \delta$ inherits the same CLT-like scaling as $W^\ell_0\,\Delta x$, and $(\Delta W^\ell)^\top \delta$ inherits the same rank-1 alignment as $\Delta W^\ell \cdot x$ (with fan-out replacing fan-in). We do not repeat these case studies.

\subsection{Aggregation layers} In MoEs, aggregations across experts arise in both the forward pass (Eq.~\eqref{eq:agg_decomp_appendix}) and the backward pass (Eq.~\eqref{eq:bwd_agg_decomp}). The aggregation operator introduces sums over \textit{expert-indexed} contributions (weighted by routing), and the leading-order scale of these sums depends on the evolving cross-expert correlation structure. In particular, expert sums can exhibit CLT-type scaling (weak cross-expert correlations), LLN-type scaling (strongly correlated contributions), or mixed behaviour across different contributions; our analysis makes these distinctions regime-by-regime.

\subsection{Width dependence in \texorpdfstring{$\mu$}{mu}P and how MSSP fixes it}

Below, we provide some case studies on the key imbalances in MoE dynamics under $\mu$P and follow this by how MSSP fixes them. 

\subsubsection{Expert hidden layer gradient $\partial f/\partial h^{2,i}$}
\label{sssec:bwd-bottleneck}

Consider the four-piece decomposition
\[
    \partial f/\partial h^{2,i} \;\propto\; (W_0^{3,i} + \Delta W^{3,i})^\top (W_0^4 + \Delta W^4)^\top.
\]

In Regime~III, $(W_0^{3,i})^\top(\Delta W^4)^\top$ admits CLT-like scaling, since $W_0^{3,i}$ has diverging input and output dimensions. In Regime~II, $W_0^{3,i}$ is a low-rank matrix and $\Delta W^{4}$ is shaped by $W_0^{3,i}$ through backpropagation, so LLN-like scaling is in principle possible. However, $\Delta W^{4}$ is built from sums over all $M \to \infty$ experts' $W_0^{3,i}$, and hence cannot simultaneously align with any single $W_0^{3,i}$; the term therefore also admits CLT-like scaling.

\[
\|(W_0^{3,i})^\top(\Delta W^4)^\top\|_{\mathrm{RMS}} = \sqrt{N}\,\|(W_0^{3,i})^\top\|_{\mathrm{RMS}}\,\|(\Delta W^4)^\top\|_{\mathrm{RMS}} \asymp \sqrt{N} \cdot 1/\sqrt{N_e} \cdot 1/{N} \quad \text{under $\mu$P}.
\]

Now consider the term $(\Delta W^{3,i})^\top(W_0^4)^\top$. As discussed in the effective-updates section, since the updates to the second expert-layer weights are rank-1 in structure, the rank-1 alignment argument applies, yielding LLN-like scaling:
\[
\|(\Delta W^{3,i})^\top(W_0^4)^\top\|_{\mathrm{RMS}} = N\,\|(\Delta W^{3,i})^\top\|_{\mathrm{RMS}}\,\|(W_0^4)^\top\|_{\mathrm{RMS}} \asymp N \cdot 1/{N_e} \cdot 1/{N} \quad \text{under $\mu$P}.
\]

The remaining two contractions in the four-piece expansion follow the same scaling laws: $(W_0^{3,i})^\top(W_0^4)^\top$ behaves identically to the CLT-like term analysed above, and $(\Delta W^{3,i})^\top(\Delta W^4)^\top$ behaves identically to the rank-1 term.

In Regime~III we have $N_e \asymp N$, so both terms are of order $1/N$ and are therefore balanced (and, under the appropriate normalization, both terms are of order $1$). In Regime~II, $N_e \in \Theta(1)$, and the two terms have scales $1/\sqrt{N}$ and $1$ respectively.

In summary, in Regime~III, $|W_0^{3,i}| = \Theta(1/\sqrt N)$ and $|\Delta W^{3,i}| = \Theta(1/N)$, leading to an entry-size gap of $\sqrt N$. This $\sqrt N$ exactly compensates the CLT-vs.-rank-1 alignment-type gap, so both contractions land at $\Theta(1/N)$: \emph{balanced}. In Regime~II, $|W_0^{3,i}| = \Theta(1)$ and $|\Delta W^{3,i}| = \Theta(1)$, so the entry sizes are equal (since $\sigma_3^2 = 1/N_e = \Theta(1)$). The CLT-vs.-rank-1 alignment-type gap is no longer cancelled, so the two pieces split by $\sqrt N$: \emph{imbalanced}.

{\paragraph{MSSP correction in Regime II.} The MSSP-Regime-II boost $\sigma_3^2 = M/N_e$ increases the scale of $|W_0^{3,i}|$ from $\Theta(1)$ to $\Theta(\sqrt M) = \Theta(\sqrt N)$, while the update $|\Delta W^{3,i}| = \Theta(1)$ is unchanged (the SGD learning rate $\eta_3 = \eta MN$ is identical to its $\mu$P-Regime-II value). The entry-size ratio $|\Delta W^{3,i}|/|W_0^{3,i}|$ thus moves from $\Theta(1)$ under $\mu$P to $\Theta(1/\sqrt N)$ under MSSP, restoring the buffer to the same $\sqrt N$ value as in Regime~III. The CLT-vs-rank-1 alignment-type gap is then cancelled exactly, and the four pieces of $\partial f/\partial h^{2,i}$ all sit at the leading $\Theta(1/N)$ scale.}

\subsection{Cross-expert aggregation: a taxonomy of mechanisms}
\label{ssec:cross-expert-mechanisms}

{We now turn to the MoE-specific cross-expert aggregations. Two such aggregations arise: the forward aggregation $h^{\mathrm{agg}}(t)$ of~\eqref{eq:agg_decomp_appendix}, with expanded four-term form $h^{\mathrm{agg}}(t) = A_1 + A_{2,1} + A_{2,2}' + A_3$; and the backward aggregation $\bar\delta^{1,\mathrm{exp}}(t)$ of~\eqref{eq:bwd_agg_decomp}, which collects the expert-pathway contributions to the gradient with respect to the input $h^1$ shared across experts, with an analogous decomposition. Each summand depends on the per-expert weights $W_0^{2,i}, W_0^{3,i}$ and their updates; the leading-order scale of the aggregate is determined by the cross-$i$ correlation structure of these summands. The scaling of these aggregation terms is determined by the crude CLT--LLN dichotomy: summands sharing a coherent direction aggregate by LLN (no $1/\sqrt M$ suppression), whereas summands with i.i.d.\ random directions aggregate by CLT ($1/\sqrt M$ suppression). The structurally interesting question, however, is: \emph{What generates a coherent direction in the first place, or prevents one from emerging?} The four mechanisms identified below (A--D) classify the case studies that follow according to the algebraic feature that controls this question, and we examine each in turn under the $\mu$P baseline and under MSSP.} {For each mechanism we (i) state the algebraic feature and case-study term it governs, (ii) state its scaling under the $\mu$P baseline, and (iii) close with a paragraph describing the corresponding behaviour under MSSP.}

\subsubsection{Mechanism A: cross-expert CLT under independent or weakly correlated summands}
\label{sssec:mech-A}

Mechanism A governs aggregates $A := \frac{1}{M}\sum_i \phi_i\,W^{3,i}_0\,W^{2,i}_0\,v$ with $v$ shared across experts. Terms $A_1$ and $A_{2,1}$ in the forward aggregation are of this form, with $v = h^1$ and $v = \Delta h^1$ respectively. In both cases a cross-expert central limit theorem applies, suppressing the aggregate by $1/\sqrt{M}$ relative to the per-summand scale.

\paragraph{Independent case ($A_1$).}
Take $v\in\mathbb{R}^N$ independent of all expert weights. With $W^{2,i}_0\in\mathbb{R}^{N_e\times N}$ Gaussian (entries $\mathcal{N}(0,\sigma_2^2)$),
\[
    W^{2,i}_0\,v \;\sim\; \mathcal{N}\!\bigl(0,\,\sigma_2^2\,\|v\|^2\,I_{N_e}\bigr),
\]
isotropic on $\mathbb{R}^{N_e}$ and depending on $v$ only through $\|v\|^2$. Conditional on $W^{2,i}_0\,v$,
\[
    u_i := W^{3,i}_0\,W^{2,i}_0\,v \;\big|\; W^{2,i}_0\,v \;\sim\; \mathcal{N}\!\bigl(0,\,\sigma_3^2\,\|W^{2,i}_0\,v\|^2\,I_N\bigr).
\]
Marginalising over $W^{2,i}_0$ preserves rotation invariance: the distribution of $u_i$ depends on $v$ only through $\|v\|^2$. Across experts, $(W^{2,i}_0, W^{3,i}_0)_{i=1}^M$ are i.i.d., so the only source of cross-expert dependence among $\{u_i\}$ is the shared $\|v\|^2$.

The scalar $\|v\|^2 = \Theta(N)$ does not itself converge. Combined with the $1/M$ aggregation prefactor and $M\asymp N$, it appears in the aggregate variance $\mathbb{E}\|A\|^2 \asymp M^{-1}\,\mathbb{E}\|u_i\|^2$ as $\|v\|^2/M \asymp \|v\|^2/N = \|v\|_{\mathrm{RMS}}^2$, which converges deterministically under standard initialisation. The summands are asymptotically independent at the aggregate level, and a multivariate CLT applies.

\paragraph{Weakly dependent case ($A_{2,1}$).}
$\Delta h^1$ is shaped by gradient backflow through every expert and depends on each $W^{2,i}_0$, so strict independence fails. A single vector in $\mathbb{R}^N$ cannot, however, align nontrivially with $M\to\infty$ independent Gaussian matrices simultaneously: the dependence of $\Delta h^1$ on any one $W^{2,i}_0$ enters through a single summand of the cross-expert loss gradient and is of relative order $1/M$. Decompose $\Delta h^1 = \Delta h^{1,(\neq i)} + \Delta h^{1,(i)}$ with $\Delta h^{1,(\neq i)}$ independent of $W^{2,i}_0$ and $\|\Delta h^{1,(i)}\|/\|\Delta h^1\| = O(1/M)$; the dominant term $W^{3,i}_0\,W^{2,i}_0\,\Delta h^{1,(\neq i)}$ is governed by the previous paragraph, the residual by a $1/M$-suppressed correction.

\paragraph{Per-summand and aggregate magnitudes.}
\[
    \mathbb{E}\|u_i\|^2 \;=\; N\sigma_3^2 \cdot N_e\sigma_2^2\,\|v\|^2 \;=\; \sigma_2^2\,\sigma_3^2\,N\,N_e\,\|v\|^2,
\]
\[
    \|u_i\|_{\mathrm{RMS}} \;\asymp\; (\sigma_2\sqrt{N})(\sigma_3\sqrt{N_e})\,\|v\|_{\mathrm{RMS}}, \qquad
    \|A\|_{\mathrm{RMS}} \;\asymp\; \frac{\sigma_2\,\sigma_3\,\sqrt{N\,N_e}}{\sqrt{M}}\,\|v\|_{\mathrm{RMS}}.
\]
This holds in both Regime~II ($N_e$ fixed) and Regime~III ($N_e\asymp N$). Under $\mu$P, $\sigma_2 = 1/\sqrt{N}$ and $\sigma_3 = 1/\sqrt{N_e}$, so $\|A_1\|_{\mathrm{RMS}}, \|A_{2,1}\|_{\mathrm{RMS}}$ are of order $1/\sqrt{M}$, strictly subleading to $A_{2,2}'$ and $A_3$ at $\Theta(1)$.

{\paragraph{MSSP correction.} The two MSSP fixes resolve this in regime-specific ways. In Regime~III, MSSP shares expert weights at initialization ($W_0^{2,i} = W_0^2$ and $W_0^{3,i} = W_0^3$ for all $i$); the per-expert summands then become strongly correlated across $i$ rather than independent, the cross-expert CLT collapses to an LLN, and both terms are lifted to $\Theta(1)$ along the shared direction $W_0^{3} W_0^{2}\,h^1$ (or its $\Delta h^1$ analogue). In Regime~II, MSSP boosts $\sigma_3$ from $1/\sqrt{N_e}$ to $\sqrt{M/N_e}$; the per-summand RMS scale is amplified by $\sqrt{M}$, exactly compensating the $1/\sqrt M$ cross-expert CLT suppression and lifting both terms to $\Theta(1)$ in expert-specific directions.}

\subsubsection{Mechanism B: Gram operators on a shared vector}
\label{sssec:mech-B}

{We will use the quantity $A_{2,2}' \;=\; \frac{1}{M}\sum_i \phi_{i,2}\,W^{3,i}_0\,\Delta_t W^{2,i}\,h^1_0$ to illustrate this mechanism. Each summand has the form $G_i u$, where $G_i = W_0^{3,i}(W_0^{3,i})^\top$ acts on a vector $u$ that is shared across experts (no $i$ index). The cross-expert direction of $G_i u$ is then governed by whether $G_i$ is full-rank (so $G_i \approx N_e\sigma_3^2 \cdot I$ and $G_i u$ inherits the shared direction $u$) or rank-deficient (so $G_i$ is, after rescaling, a projection onto a random $N_e$-dimensional subspace of $\mathbb R^N$, distinct across experts). The same rank threshold separates Regime~III ($N_e \asymp N$, isotropic, LLN) from Regime~II ($N_e = \Theta(1)$, anisotropic, CLT).}

\paragraph{Intuitive scaling of $A_{2,2}'$ across regimes.}
Substituting the rank-1 update $\Delta_t W^{2,i} \propto (W^{3,i}_0)^\top(W^4_0)^\top\,(h^1_0)^\top$ and using $\Delta_t W^{2,i}\,h^1_0 \propto \|h^1_0\|^2\,(W^{3,i}_0)^\top(W^4_0)^\top$, each summand of $A_{2,2}'$ reduces to
\[
    W^{3,i}_0\,\Delta_t W^{2,i}\,h^1_0 \;\propto\; \|h^1_0\|^2\,\underbrace{W^{3,i}_0(W^{3,i}_0)^\top}_{=:\,G_i}\,(W^4_0)^\top.
\]
Writing $u := (W^4_0)^\top \in \mathbb{R}^N$, the analysis splits into a within-expert step (the action of $G_i$ on the fixed vector $u$) and a cross-expert step (the aggregation $(1/M)\sum_i$).

\vspace{1em}

\textit{Within-expert structure (fix $i$).} The matrix $W^{3,i}_0 \in \mathbb{R}^{N\times N_e}$ has rank $\min(N, N_e)$ generically, and so does $G_i$. The image of $G_i$ is the column space of $W^{3,i}_0$ in $\mathbb{R}^N$; let $\Pi_i$ denote the orthogonal projection onto it. By rotational invariance of the entries of $W^{3,i}_0$, the random subspace $\mathrm{col}(W^{3,i}_0)$ is Haar-distributed on the Grassmannian $\mathrm{Gr}(\min(N,N_e),\,N)$.

The non-zero eigenvalues of $G_i$ are the squared singular values of $W^{3,i}_0$, which concentrate around $N\sigma_3^2$ by Marchenko--Pastur \citep{marchenko1967distribution}. Consequently $G_i$ admits the operator-level approximation
\[
    G_i \;\approx\; (N\sigma_3^2)\,\Pi_i,
\]
giving $G_i u \approx (N\sigma_3^2)\,\Pi_i u$. Since $u$ is independent of $W^{3,i}_0$\footnote{The same argument holds under weak correlations} and $\mathrm{col}(W^{3,i}_0)$ is Haar-distributed, each projection retains
\[
    \|\Pi_i u\|_2^2 \;\asymp\; \frac{\min(N,N_e)}{N}\,\|u\|_2^2.
\]

\textit{Cross-expert structure.} The matrices $\{W^{3,i}_0\}_{i=1}^M$ are i.i.d., so the projections $\{\Pi_i u\}_{i=1}^M$ are independent across experts, each marginally Haar-distributed in direction within $\mathbb{R}^N$.

\bigskip

\textit{Regime~III ($N_e \asymp N$): isotropic Gram matrix preserves LLN scaling.} For each fixed $i$, since $W^{3,i}_0$ is generically full rank at leading order, so $\Pi_i \approx I_N$ and
\[
    G_i\,u \;\approx\; (N_e\sigma_3^2)\,u \;=\; \Theta(1)\cdot u
\]
(taking $\sigma_3^2 = 1/N_e$). The Gram acts as a near-scalar multiple of the identity, and the result lies along $u = (W^4_0)^\top$ (shared direction across all experts). Cross-expert aggregation admits a law of large numbers like scaling:
\[
    \|A_{2,2}'\|_{\mathrm{RMS}} \;\asymp\; \|h^1_0\|^2 \cdot N_e\sigma_3^2 \cdot \|(W^4_0)^\top\|_{\mathrm{RMS}} \;=\; \Theta(1) \text{ along }(W^4_0)^\top \quad \text{under $\mu$P}
\]

\vspace{1em}
\textit{Regime~II ($N_e$ fixed, $N\to\infty$): anisotropic Gram yields CLT scaling.} For each fixed $i$, the column span $\mathrm{col}(W^{3,i}_0)$ is a fixed-dimensional random subspace within a diverging ambient space, on which $G_i$ has eigenvalue $N\sigma_3^2 = N/N_e$ under $\mu$P. By rotational invariance, the Haar distribution of this subspace implies $\|\Pi_i u\|_2^2 \asymp (N_e/N)\,\|u\|_2^2$: each orthonormal basis vector of $\mathrm{col}(W^{3,i}_0)$ is uniform on $S^{N-1}$ by rotation invariance, contributing $\|u\|^2/N$ in expectation, and there are $N_e$ such basis vectors. Therefore,
\[
    \|G_i u\|_2 \;\asymp\; (N\sigma_3^2)\sqrt{N_e/N}\,\|u\|_2 \;=\; \sigma_3^2\sqrt{N N_e}\,\|u\|_2,
\]
with $G_i u$ lying entirely in the expert-specific subspace $\mathrm{col}(W^{3,i}_0)$. Across experts, these subspaces are independent random elements of the Grassmannian, so the directions $\{G_i u / \|G_i u\|_2\}_{i=1}^M$ are independent and there is no shared coherent direction along which $\{G_i u\}_{i}$ can accumulate. Cross-expert aggregation therefore proceeds by the central limit theorem rather than the law of large numbers: $A_{2,2}'$ is suppressed by an additional $1/\sqrt M$ relative to the per-summand scale, and is not aligned with $(W^4_0)^\top$.

{\paragraph{MSSP correction.} The Regime~II imbalance is removed under MSSP-Regime-II by the amplification $\sigma_3^2 = M/N_e$, which raises the eigenvalue of $G_i$ on its rank-$N_e$ column span from $N\sigma_3^2 = N/N_e$ under $\mu$P to $N\sigma_3^2 = NM/N_e$ under MSSP. The per-summand scale $\|G_i u\|_2 \asymp \sigma_3^2\sqrt{NN_e}\,\|u\|_2$ is thereby amplified by exactly $\sqrt M$, and the cross-expert $1/\sqrt M$ CLT cancellation is exactly absorbed; $A_{2,2}'$ then sits at $\Theta(1)$, in expert-specific directions rather than along $(W_0^4)^\top$. In Regime~III, $A_{2,2}'$ is already $\Theta(1)$ under $\mu$P (the Gram is isotropic and aligned along $u$); MSSP-Regime-III's shared experts collapse the marginal randomness of $\{\Pi_i\}$ entirely, but do not change the $\Theta$-class.}

\subsubsection{Mechanism C: Expert sums of effective contributions}
\label{sssec:mech-C}

{Let us use the term $A_3$ of the forward aggregation to illustrate. Each summand has the form $\Delta W^{3,i}\,x_i$ with $\Delta W^{3,i}$ rank-1 of the form $u_{\mathrm{shared}}\,v_i^\top$, where $u_{\mathrm{shared}}$ is a backpropagated factor shared across experts and $v_i$ is expert-specific. The product $\Delta W^{3,i}\,x_i = \langle v_i, x_i\rangle\,u_{\mathrm{shared}}$ then writes the summand as a (per-expert) scalar times a shared direction. Cross-expert aggregation reduces to a scalar mean, and the LLN survives whenever that scalar mean is dominated by a strictly positive contribution. Unlike Mechanism~B, the shared direction is forced by the rank-1 structure of the gradient itself rather than recovered from a Gram acting on a fixed vector, so this mechanism is regime-insensitive.}

\paragraph{Scaling $A_3$.} Recall that
\[
    A_3 \;=\; \frac{1}{M}\sum_i \phi_{i,t}\,\Delta_t W^{3,i}\,h^{2,i}_t,
\]
where
\[
    \Delta_t W^{3,i}\,h^{2,i}_t \;=\; -\frac{\eta_3\,\chi\,\phi_{i,t-1}}{M}\,\langle h^{2,i}_{t-1},\,h^{2,i}_t\rangle\,(W^4_{t-1})^\top.
\]
 Therefore, the per-summand is a single inner product times the shared readout direction. Decomposing the inner product,
\[
    \langle h^{2,i}_{t-1},\,h^{2,i}_t\rangle \;=\; \|h^{2,i}_{t-1}\|_2^2 \;+\; \langle h^{2,i}_{t-1},\,\Delta_t h^{2,i}\rangle.
\]
The first term, $\|h^{2,i}_{t-1}\|_2^2$, is strictly positive for every expert (non-zero mean across experts). The second term is a smaller fluctuation around it: at $t=1$ it is random across $i$, while at $t=2$ it becomes coherent via the alignment chain. In either case the sum is dominated by the strictly positive $\|h^{2,i}_{t-1}\|_2^2$, so the cross-expert scalar average is LLN-preserved:
\[
    \frac{1}{M}\sum_i \phi_{i,t}\,\phi_{i,t-1}\,\langle h^{2,i}_{t-1},\,h^{2,i}_t\rangle \;\asymp\; \|h^{2,i}\|_2^2.
\]
Substituting back,
\[
    A_3 \;\asymp\; -\frac{\eta_3\,\chi}{M}\,\|h^{2,i}\|_2^2\,(W^4_{t-1})^\top
\]
along the shared readout direction and in $\Theta(1)$ in Regime~II under both $\mu$P and MSSP. In Regime~III, the same argument shows that these terms are $\Theta(1)$ under $\mu$P. Under MSSP (which uses shared experts), similar arguments show that these terms are $\Theta(1)$ as well.

\subsubsection{Mechanism D: Scalar self-pairing structural invariance}
\label{sssec:mech-D}

{Certain quantities in the aggregation are built from summands that comprise scalar self-pairings $u^\top G_i u$ rather than a vector $G_i u$. The expectation $\mathbb{E}[u^\top G_i u] = \|u\|_2^2$ depends only on the diagonal mean $\mathbb{E}[(G_i)_{aa}] = N_e\sigma_3^2$, which is $\Theta(1)$ in both Regimes~II and~III under $\mu$P (and is therefore regime-symmetric). The per-layer parameterization, however, is calibrated against the (regime-asymmetric) vector form, so when the structure happens to be scalar the calibration finds nothing to act on and passes through unattenuated, leading to width dependence.}

\paragraph{Mechanism: The structural mechanism underlying the imbalance of $A_6$.}

The term $A_6$ arises in the expansion of the backward expert-pathway gradient $\bar\delta^{1,\mathrm{exp}}(t)$ of~\eqref{eq:bwd_agg_decomp}. Specifically, $A_6 = (1/M)\sum_i \phi_{i,t}\,(\Delta W^{2,i})^\top\,(W_0^{3,i})^\top\,(W^4_t)^\top$, the contribution in which $\Delta W^{2,i}$ is the perturbed expert factor while $W^{3,i}$ remains at initialization. Substituting $\Delta W^{2,i}$ into $A_6$ produces the scalar self-pairing
\[
    S_i \;:=\; u^\top G_i\,u \;=\; \big\|(W_0^{3,i})^\top u\big\|_2^2,
    \qquad u := (W_0^4)^\top,\;\; G_i := W_0^{3,i}(W_0^{3,i})^\top.
\]
Since $(G_i)_{ab} = \sum_{k=1}^{N_e}(W_0^{3,i})_{ak}(W_0^{3,i})_{bk}$ with entries i.i.d.\ $\mathcal N(0,\sigma_3^2)$,
\[
    \mathbb{E}[(G_i)_{ab}] \;=\; N_e\sigma_3^2\,\delta_{ab} \;=\; \delta_{ab}
    \qquad (\sigma_3^2 = 1/N_e),
\]
hence
\[
    \mathbb{E}[S_i] \;=\; \|u\|_2^2 \;=\; \Theta(1/N) \quad\text{in both regimes.}
\]
The remaining terms of the decomposition of the aggregated gradient depend on $W_0^{3,i}$ only through linear, vector-form quantities, which are regime-asymmetric by a factor of $\sqrt N$ in per-coordinate magnitude and which are accounted for by the parameterization. Due to the regime symmetry of $A_6$, the parameterization therefore passes through it unattenuated, yielding \emph{imbalanced} scaling in Regime~II.
\[
    A_6 \;\in\; \Theta(1/N^2)\;\text{ in Regime II},
    \qquad A_6 \;\in\; \Theta(1/N)\;\text{ in Regime III.}
\]

{\paragraph{MSSP correction.} The MSSP-Regime-II boost $\sigma_3^2 = M/N_e$ promotes the diagonal mean to $\mathbb{E}[(G_i)_{aa}] = N_e\sigma_3^2 = M = \Theta(N)$, so the scalar self-pairing satisfies $\mathbb{E}[S_i] = M\|u\|_2^2 = \Theta(N)\cdot\Theta(1/N) = \Theta(1)$. The $\Theta(N)$ amplification of the scalar exactly fills the $\sqrt N \cdot \sqrt N$ deficit that $\mu$P-Regime-II's vector-form-calibrated parameterization had been unable to address, and $A_6$ moves from $\Theta(1/N^2)$ to $\Theta(1/N)$, the leading scale in Regime~II. The structural symmetry of $\mathbb{E}[G_i] \propto I$ that made $A_6$ inert under $\mu$P-Regime-II is exploited in reverse by MSSP: the boost is precisely tuned so that the regime-symmetric part of $G_i$ becomes regime-asymmetric in the way the parameterization expects. In Regime~III, $A_6$ is already at the leading scale under $\mu$P, and MSSP-Regime-III's shared experts do not change its $\Theta$-class.}

\subsection{Summary: regime imbalances under \texorpdfstring{$\mu$}{mu}P and how MSSP resolves them}
\label{ssec:summary}

\textbf{Regime II.} In Regime~II, MSSP introduces a single change to the $\mu$P baseline: the initialization variance of the second expert layer is increased to $\sigma_3^2 = M/N_e$. Although the change operates through a single parameter, it restores width independence in all four sources of imbalance present under $\mu$P in Regime~II. The cross-expert CLT under independent or weakly correlated summands (Mechanism~A) is corrected because the increased variance amplifies each per-summand by $\sqrt M$, exactly offsetting the $1/\sqrt M$ suppression of the cross-expert sum. The Gram operators on a shared vector (Mechanism~B) and the scalar self-pairing structural invariance (Mechanism~D) are corrected by the corresponding amplification of the Gram's eigenvalues and diagonal entries. The gradient at the expert hidden layer (§\ref{sssec:bwd-bottleneck}) is corrected because the increased $\sigma_3$ raises $\|W_0^{3,i}\|$ to a scale that allows the four contractions of $\partial f/\partial h^{2,i}$ to balance. The same change to the parameterization therefore corrects all four structurally distinct sources of width dependence in Regime~II simultaneously.

\textbf{Regime III.} In Regime~III, MSSP introduces a different single change: expert weights are shared at initialization, so that all experts begin from a common pair $(W_0^2, W_0^3)$. Under $\mu$P, the only mechanism that produces a width dependence in Regime~III is the cross-expert CLT under independent or weakly correlated summands (Mechanism~A). With shared experts, the per-expert summands are identical across $i$ and accumulate by an LLN rather than cancelling by a CLT, and the affected terms are lifted to the leading scale. The remaining mechanisms (the Gram operators on a shared vector, the expert sums of effective contributions, the scalar self-pairing structural invariance, and the gradient at the expert hidden layer) are already balanced under $\mu$P in Regime~III, and shared experts leave their leading behaviour unchanged.

\textbf{The two fixes cannot be interchanged.} The two fixes are matched to the two distinct sources of width dependence in their respective regimes, and cannot be interchanged. Applying the Regime-II fix in Regime~III would amplify the second expert layer's initial weights too aggressively, causing the backward gradient at the expert hidden layer to scale as $\sqrt N$ above its target; this is the very imbalance that the Regime-II fix was introduced to repair in Regime~II. Conversely, sharing experts in Regime~II would correct only Mechanism~A: the Gram operators on a shared vector, the scalar self-pairing structural invariance, and the gradient at the expert hidden layer would all remain imbalanced, since shared experts do not affect the $\sigma_3$-dependent scales through which those mechanisms operate. Regime~II requires the boosted variance; Regime~III requires shared experts; each is the appropriate fix for its regime and would cause width dependence in the other.

\subsection{Justification of MSSP in Regime II}
\label{ssec:r2_ssp_justification}

We close the Regime~II analysis by addressing a structural feature of the MSSP parameterization (Definition~\ref{def:ssp_regime_ii}) that distinguishes it from conventional feature-learning parameterizations: the per-expert hidden state $h^{3,i}$ (specifically the initial and propagating contributions $W_0^{3,i} h_0^{2,i}$ and $W_0^{3,i}\Delta h^{2,i}$) has coordinate scale $\Theta(\sqrt N)$ vs the standard $\Theta(1)$ scale. We argue in this subsection that this is admissible because $h^{3,i}$ never functions as a primitive variable in the dynamics, and that MSSP-Regime-II is, up to structurally equivalent reformulations, the unique parameterization that achieves the correct $\Theta(1)$ scale for the decomposition of the aggregated activation $h^3$ in Regime~II while confining the resulting amplification to a quantity that the dynamics avoid as a primitive.

\paragraph{The dynamics never realize $h^{3,i}$ as a primitive variable.}
We claim that, throughout both forward and backward passes, every place at which $h^{3,i}$ would arise in a natural decomposition of the dynamics admits a contraction-order rewriting that bypasses it. The forward aggregate is computable directly:
\begin{equation}\label{eq:ssp_r2_h3_forward}
    h^3 \;=\; \frac{1}{M}\sum_{i=1}^M \phi_i\,W^{3,i}\,h^{2,i},
\end{equation}
without forming any $h^{3,i}$ explicitly. The backward gradients factor through small-dimensional contractions:
\begin{align}
    \frac{\partial f}{\partial h^{2,i}} \;&=\; \frac{\phi_i}{M}\,(W^{3,i})^\top(W^4)^\top, \label{eq:ssp_r2_grad_h2i}\\
    \frac{\partial L}{\partial W^{3,i}} \;&=\; \frac{\chi\,\phi_i}{M}\,(W^4)^\top(h^{2,i})^\top, \label{eq:ssp_r2_grad_W3i}\\
    \frac{\partial f}{\partial \phi_i} \;&=\; \frac{1}{M}\,(W^4\,W^{3,i})\,h^{2,i}. \label{eq:ssp_r2_grad_phi}
\end{align}
The gating gradient \eqref{eq:ssp_r2_grad_phi} is the only quantity in which $h^{3,i}$ would appear under the natural decomposition $\partial f/\partial\phi_i = (1/M)\,(h^{3,i})^\top(W^4)^\top$. The contraction reordering on the right-hand side replaces the formation of $h^{3,i}\in\mathbb{R}^N$ with the inner-product chain $W^4\,W^{3,i}\in\mathbb{R}^{1\times N_e}$, an $N_e$-dimensional row vector. Because $N_e=\Theta(1)$, this object is small-dimensional and may be formed at numerical magnitude $\Theta(\sqrt N)$ without storage or precision concerns.

The full forward and backward dynamics therefore close on the variable set
\begin{equation}\label{eq:ssp_r2_variables}
    \mathcal V \;=\; \bigl\{\,h^1,\; \psi,\; \phi,\; \{h^{2,i}\}_{i=1}^M,\; h^3,\; f,\; \{W^{2,i}\}_{i=1}^M,\; \{W^{3,i}\}_{i=1}^M,\; W^4,\; Q\,\bigr\},
\end{equation}
none of whose elements simultaneously carry both large coordinate magnitude ($\Theta(\sqrt N)$) and large dimension ($\Theta(N)$). The amplified per-expert hidden state $h^{3,i}$ — the unique object in the architecture that combines both — is structurally absent from $\mathcal V$ as a primitive variable.

\paragraph{Mean-field interpretation.}
The structure above is the signature of a mean-field theory: per-particle quantities are amplified, the aggregate is the natural order parameter, and the closed dynamics involve only aggregated moments of per-particle quantities together with low-dimensional inputs. The DMFT effective fields associated with MSSP-Regime-II can be expressed entirely in terms of aggregated weight-product moments such as
\[
    \bar G_t \;=\; \frac{1}{M}\sum_{i=1}^M \phi_{i,t}\,W_t^{3,i}\,W_t^{2,i}, \qquad \bar K_t \;=\; \frac{1}{M}\sum_{i=1}^M \phi_{i,t}^2\,(W_t^{2,i})^\top\,W_t^{2,i},
\]
together with the per-expert layer-1 activations $\{h^{2,i}\}$, all of which sit at the standard $\Theta(1)$ coordinate scale. The per-expert hidden state $h^{3,i}$ does not appear in the effective theory.

\subsection{Composability with depth: stacked MoE blocks under MSSP-Regime-II}
\label{ssec:ssp_r2_depth}

The boost-and-cancel mechanism that defines MSSP-Regime-II is structurally local to a single MoE block: it operates within that block's cross-$i$ CLT on the aggregated activation, and its scaling justification (Definition~\ref{def:ssp_regime_ii}) does not refer to any architectural element outside the block. We formalize this observation by showing that stacking $K$ MoE blocks under MSSP-Regime-II preserves the leading scales of every primitive variable in the forward pass and every backward gradient at every block boundary. The parameterization composes with depth without per-block modification.

\paragraph{Stacked architecture.}
Let $K\geq 2$. We consider the composition
\[
    x \;\xrightarrow{W^1}\; h^1 \;\xrightarrow{\text{block }1}\; h^{3,(1)} \;\xrightarrow{\text{block }2}\; h^{3,(2)} \;\xrightarrow{}\; \cdots \;\xrightarrow{\text{block }K}\; h^{3,(K)} \;\xrightarrow{W^4}\; f,
\]
where each block $k\in[K]$ has its own gating $Q^{(k)}$, expert weights $\{W^{2,i,(k)}, W^{3,i,(k)}\}_{i\in[M]}$, and aggregation
\begin{align*}
    \psi^{(k)} &= Q^{(k)} h^{3,(k-1)}, \qquad \phi_i^{(k)} = \sigma(\psi_i^{(k)}),\\
    h^{2,i,(k)} &= W^{2,i,(k)} h^{3,(k-1)},\\
    h^{3,i,(k)} &= W^{3,i,(k)} h^{2,i,(k)},\\
    h^{3,(k)} &= \frac{1}{M} \sum_{i=1}^M \phi_i^{(k)}\, h^{3,i,(k)},
\end{align*}
with the convention $h^{3,(0)} := h^1$. All weights are drawn independently across blocks at initialization. Each block adopts the MSSP-Regime-II parameterization (Definition~\ref{def:ssp_regime_ii}): per-block init variances $\sigma_Q^2 = 1/N$, $\sigma_2^2 = 1/N$, $\sigma_3^2 = M/N_e$; output $W_0^4 = 0$; and SGD learning rates $\eta_Q = \eta_2 = \eta\,M/N$, $\eta_3 = \eta MN$, $\eta_4 = \eta/N$ within each block.

\paragraph{Forward composability.}

\begin{proposition}[Forward depth composition]
\label{prop:ssp_r2_depth_forward}
Under the stacked MSSP-Regime-II architecture above, for every $k\in[K]$ at initialization:
\begin{enumerate}[label=(\roman*),leftmargin=2em]
    \item $h_0^{2,i,(k)}\in\Theta(1)$ entry-wise.
    \item $h_0^{3,i,(k)}\in\Theta(\sqrt N)$ entry-wise.
    \item The aggregate $h_0^{3,(k)}\in\Theta(1)$ entry-wise via cross-$i$ CLT.
\end{enumerate}
\end{proposition}

\begin{proof}[Sketch]
Induction on $k$. The base case $k=1$ is the single-block analysis (\S\ref{ssec:ssp_r2_fwd1}) with input $h^1\in\Theta(1)$. For the inductive step, suppose $h_0^{3,(k-1)}\in\Theta(1)$ entry-wise. Then $h_0^{2,i,(k)} = W_0^{2,i,(k)} h_0^{3,(k-1)}$ has variance per entry $\sigma_2^2\,\|h_0^{3,(k-1)}\|^2 = (1/N)\cdot\Theta(N) = \Theta(1)$, giving (i). Next, $h_0^{3,i,(k)} = W_0^{3,i,(k)} h_0^{2,i,(k)}$ has variance per entry $N_e\,\sigma_3^2\,\Theta(1) = N_e\cdot(M/N_e) = M = \Theta(N)$, giving (ii). Finally, the aggregate variance is $(1/M^2)\sum_i\phi_{i,0}^2\,N_e\,\sigma_3^2\,\Theta(1) = \Theta(\sigma_3^2/M) = \Theta(1)$ via cross-$i$ CLT, giving (iii). The inductive hypothesis enters only through the $\Theta(1)$ entry scale of the input $h_0^{3,(k-1)}$ — exactly the property required to start the per-block boost-and-cancel.
\end{proof}

\paragraph{Backward composability.}

\begin{proposition}[Backward depth composition]
\label{prop:ssp_r2_depth_backward}
At $t=1$ — when only $W^4$ has updated, with $W_1^4 = -\eta_4\chi_0\,(h_0^{3,(K)})^\top$ having entries $\Theta(1/N)$ — the propagated gradient at every inter-block aggregate satisfies
\[
    \partial f_1/\partial h_0^{3,(k)} \in \Theta(1/N) \quad\text{entry-wise}, \qquad k\in\{0,1,\dots,K-1\},
\]
where $h_0^{3,(0)} := h^1$.
\end{proposition}

\begin{proof}[Sketch]
Reverse induction on $k$. Base case $k = K-1$: the expert pathway through block $K$ is
\[
    \partial f_1/\partial h_0^{3,(K-1)}\big|_\text{exp} \;=\; \frac{1}{M}\sum_{i=1}^M \phi_{i,0}^{(K)}\,(W_0^{2,i,(K)})^\top(W_0^{3,i,(K)})^\top(W_1^4)^\top.
\]
The intermediate $(W_0^{3,i,(K)})^\top(W_1^4)^\top\in\mathbb{R}^{N_e}$ has entries $\Theta(1)$ — the boost in $\sigma_3^2 = M/N_e$ produces $(W_0^{3,i,(K)})^\top h_0^{3,(K)} \in \Theta(N)$, multiplied by $-\eta_4\chi_0 = -1/N$ to give $\Theta(1)$. Then $(W_0^{2,i,(K)})^\top$ acting on this $\Theta(1)$ vector via the rank-deficient initialization Gram (\S\ref{ssec:mup_r2_lemmas}) produces per-summand entries $\Theta(1/\sqrt N)$ in mostly-random directions; cross-$i$ CLT over $M$ approximately-independent summands cancels this to $\Theta(1/\sqrt{MN}) = \Theta(1/N)$. The router pathway through block $K$ contributes $\Theta(1/N)$ similarly.

Inductive step: assume $\partial f_1/\partial h_0^{3,(k+1)}\in\Theta(1/N)$ entries. The expert pathway through block $k+1$ is
\[
    \partial f_1/\partial h_0^{3,(k)}\big|_\text{exp} \;=\; \frac{1}{M}\sum_{i=1}^M \phi_{i,0}^{(k+1)}\,(W_0^{2,i,(k+1)})^\top(W_0^{3,i,(k+1)})^\top\,\partial f_1/\partial h_0^{3,(k+1)}.
\]
The intermediate $(W_0^{3,i,(k+1)})^\top\,\partial f_1/\partial h_0^{3,(k+1)}\in\mathbb{R}^{N_e}$ has variance per entry $N\cdot\sigma_3^2\cdot\Theta(1/N^2) = N\cdot N\cdot(1/N^2) = 1$, hence coordinate scale $\Theta(1)$ — the boost in $\sigma_3^2$ exactly compensates the inductive $\Theta(1/N)$ scale of the incoming gradient. The remainder of the calculation reduces to the base case: $(W_0^{2,i,(k+1)})^\top$ acting on a $\Theta(1)$ vector produces per-summand $\Theta(1/\sqrt N)$, cross-$i$ CLT yields aggregate $\Theta(1/N)$. Router pathway analogous.
\end{proof}

\begin{theorem}[Depth composability of MSSP-Regime-II]
\label{thm:ssp_r2_depth}
Under the stacked MSSP-Regime-II architecture above, for every $K\geq 1$ and every block $k\in[K]$:
\begin{enumerate}[label=(\roman*),leftmargin=2em]
    \item Every primitive variable at initialization in block $k$ has the same coordinate scale as in the single-block ($K=1$) MSSP-Regime-II analysis (\S\ref{ssec:ssp_r2_fwd1}).
    \item Every backward gradient at every primitive variable at $t=1$ in block $k$ has the same coordinate scale as in the single-block analysis (\S\ref{ssec:ssp_r2_bwd2}), and consequently the SGD updates $\Delta W^{2,i,(k)}, \Delta W^{3,i,(k)}, \Delta Q^{(k)}$ within block $k$ retain their single-block scales.
    \item Every per-expert hidden state $h^{3,i,(k)}$ remains non-primitive: the contraction-reordering identity
    \[
        (h^{3,i,(k)})^\top v \;=\; (v^\top W^{3,i,(k)})\, h^{2,i,(k)}
    \]
    bypasses formation of $h^{3,i,(k)}$ for any vector $v\in\mathbb{R}^N$ at coordinate scale $\Theta(1/N)$, with the small-dimensional intermediate $v^\top W^{3,i,(k)}\in\mathbb{R}^{N_e}$ at coordinate scale $\Theta(1)$.
\end{enumerate}
Consequently, MSSP-Regime-II composes with arbitrary depth $K$ without any per-block parameter modification.
\end{theorem}

\begin{proof}[Sketch]
(i) is Proposition~\ref{prop:ssp_r2_depth_forward}. (ii) follows from Proposition~\ref{prop:ssp_r2_depth_backward} together with the single-block backward analysis: once the gradient at block $k$'s output aggregate $\partial f_1/\partial h_0^{3,(k)}\in\Theta(1/N)$ is established, all backward computations within block $k$ proceed exactly as in \S\ref{ssec:ssp_r2_bwd2}, since they depend only on the input gradient's coordinate scale and the in-block weight structure. (iii) is the algebraic identity together with the entry-scale calculation $v^\top W^{3,i,(k)}\in\Theta(1)$ for $v\in\Theta(1/N)$ entries, which is precisely the intermediate computation in the proof of Proposition~\ref{prop:ssp_r2_depth_backward}.
\end{proof}

\subsection{Approaches We Tried That Do Not Restore Scale Stability}
\label{sec:failed_treatments}

Before settling on the parameterization-level prescription that defines MSSP, we explored a number of architectural and parameterization-level interventions intended to remove the delayed-learning phenomenon and the underlying width-dependent subterms in MoE training under $\mu$P. None of the interventions enumerated below, applied on its own, restored clean refined-coordinate-check exponents while avoiding delayed learning at scale. We document them here because each is a natural intervention that practitioners may consider, and the reasons they fail clarify why the underlying problem is structural and why a parameterization-level fix is necessary.

\subsubsection{RMSNorm (or LayerNorm) after the aggregation}
\label{ssec:try_postnorm}

A natural first response to the observation that $\|h^{\mathrm{agg}}\|_{\mathrm{RMS}}$ vanishes with width under $\mu$P-Regime-II is to insert a normalization layer after the aggregation, rescaling $h^{\mathrm{agg}}$ back to $\Theta(1)$. We tested this intervention with both RMSNorm and post-aggregation sigmoid and observed that neither restores width-independent dynamics. The reasons fall into four distinct points.

\paragraph{Norm layers preserve ratios, not balance.} The post-aggregation activation $h^{\mathrm{agg}}$ decomposes into multiple subterms with distinct width scalings under $\mu$P-Regime-II~--- most notably the LLN-aligned effective contribution $A_3 \in \Theta(1)$ and the CLT-suppressed propagating contribution $A_{2,1} \in \Theta(1/\sqrt N)$ (\S\ref{ssec:mup_r2_summary}). RMSNorm divides the entire aggregate by its own RMS norm, rescaling every subterm by the same factor. The relative magnitudes of the subterms are preserved: after normalization, the post-norm activation is dominated by the effective contribution at the leading scale, while the propagating contribution sits at $\Theta(1/\sqrt N)$ of the signal~--- a vanishing fraction. The propagating channel that should carry information from upstream-layer updates into the current block has been effectively erased.

\paragraph{The norm's $\epsilon$ has to be width-scaled, and the right scaling is exactly the imbalanced-subterm structure we are trying to avoid having to track.} Practical RMSNorm computes $\tilde h = h/\sqrt{\|h\|^2 + \epsilon^2}$, where $\epsilon$ prevents over- and under-flow. With $\|h^{\mathrm{agg}}_0\|^2 \in \Theta(1/N)$ in $\mu$P-Regime-II, there is a width-dependent crossover between two regimes: when $\|h\|^2 \ll \epsilon^2$ the norm acts as a constant rescaling $h/\epsilon$ (no normalization in the limit), while when $\|h\|^2 \gg \epsilon^2$ the norm acts as a full RMS normalization. A fixed $\epsilon$ therefore behaves qualitatively differently at small versus large widths, and the threshold between the two behaviors moves with $N$. To recover width-independent behavior, $\epsilon$ would need to be scaled in proportion to $\|h^{\mathrm{agg}}\|$; but $\|h^{\mathrm{agg}}\|$ in $\mu$P-Regime-II is itself a sum of subterms with mixed scaling exponents, and the dominant subterm changes during training. There is no single power-of-$N$ schedule for $\epsilon$ that works across widths without first having solved the underlying scaling problem. The norm therefore moves the scaling burden into a different hyperparameter rather than eliminating it.

\paragraph{The backward pass through the norm introduces its own width-dependent rescaling.} The Jacobian of RMSNorm scales as $1/\|h^{\mathrm{agg}}\|$, so gradients flowing backward through the norm layer are inflated by that factor. At initialization in $\mu$P-Regime-II this factor is $\sqrt N$, and gradients upstream of the norm are amplified by a width-dependent multiplier whose magnitude also evolves over training as $\|h^{\mathrm{agg}}\|$ climbs. The effective layerwise learning rate on every parameter upstream of the norm therefore becomes a function of both width and step number~--- the time-varying width-dependence that scale stability is designed to remove. So even granting the loss of the propagating channel discussed above and the hyperparameter burden of scaling $\epsilon$, the backward pass through a normalization layer is itself a fresh source of instability rather than a fix for the existing one.

\paragraph{A post-aggregation norm cannot reach imbalances that live elsewhere in the dynamics.} The imbalanced subterms in $\mu$P-Regime-II are not localized to $h^{\mathrm{agg}}$. The four-piece expert-hidden-gradient buffer $\partial f/\partial h^{2,i}$ (\S\ref{sssec:bwd-bottleneck}) and the scalar self-pairing imbalance in $A_6$ (Mechanism~D, \S\ref{sssec:mech-D}) sit on the backward pathway and never pass through any post-$h^{\mathrm{agg}}$ operation. Even granting that a post-aggregation norm somehow fixed the forward aggregation, three of the four mechanisms identified in \S\ref{ssec:cross-expert-mechanisms} would remain broken. A single normalization layer placed at one point in the architecture cannot address imbalances that arise across multiple structurally distinct points in the dynamics; this is one of the empirical reasons we converged on a parameterization-level fix rather than an architectural patch.

\newpage
\section{Signal Propagation Analysis}\label{sec:heuristic_derivation}

Here we provide a self-consistent forward and backward signal propagation analysis through an MoE block for the two more interesting Regimes II and III.

This analysis complements the rigorous DMFT treatment in the subsequent sections. We use the minimal MoE architecture defined below, which places the MoE block between an input and an output layer so that its incoming and outgoing signals carry the same asymptotic scaling as in a standard dense architecture. The arguments therefore extend to MoE blocks embedded in larger networks, such as a Transformer MoEs, whenever additional modules preserve this scaling. This is precisely what $\mu$P's layer-wise parametrization guarantees~\citep{yang_feature_2021,tp5_2022}. This compositionality is verified by our Transformer MoE coordinate checks in \Cref{sec:rcc_llm}. For brevity, we omit the analysis for Regime I, which follows the known arguments for standard architectures.

\subsection{Preliminaries}
\label{sec:preliminaries}

\textbf{Minimal MoE architecture.}
We work with the minimal MoE architecture
\begin{align}
    \label{eq:architecture}
    h^1      &= W^1 x,                                                      \\
    \psi     &= Q h^1,                                                      \\
    \phi     &= \sigma(\psi),                                               \\
    h^{2,i}  &= W^{2,i} h^1,                   \qquad i = 1,\dots,M,        \\
    h^{3,i}  &= W^{3,i} h^{2,i},                                            \\
    h^3      &= \frac{1}{M}\sum_{i=1}^M \phi_i\, h^{3,i},                   \\
    f        &= W^4 h^3,
\end{align}
where $x \in \mathbb{R}^D$ is the input, $f \in \mathbb{R}$ the scalar output, and the trainable parameters are
\[
\theta \;=\; \bigl(W^1,\; Q,\; \{W^{2,i},\, W^{3,i}\}_{i=1}^M,\; W^4\bigr)
\]
with shapes $W^1 \in \mathbb{R}^{N\times D}$, $Q \in \mathbb{R}^{M\times N}$, $W^{2,i} \in \mathbb{R}^{N_e\times N}$, $W^{3,i} \in \mathbb{R}^{N\times N_e}$, $W^4 \in \mathbb{R}^{1\times N}$. Here $N$ is the hidden width, $N_e$ the expert hidden width, $M$ the number of experts, and $D$ the input dimension. The derivations below are written for \emph{sigmoid} gating, $\sigma(\psi)_i = (1+e^{-\psi_i})^{-1}$, paired with the explicit $1/M$ aggregation factor in~\eqref{eq:architecture}; the same scaling exponents hold under softmax gating without an explicit aggregation multiplier.

\textbf{Scaling regimes.}
As described in the main paper, we analyze three asymptotic regimes, distinguished by which of the width parameters $N, N_e, M$ diverge: 
\begin{itemize}
    \item \textbf{Regime I:} ($N, N_e \asymp n \to \infty$ with $M, K = \Theta(1)$),
    \item \textbf{Regime II:} ($N, M, K \asymp n \to \infty$ with $N_e = \Theta(1)$),
    \item \textbf{Regime III:} ($N, N_e, M, K \asymp n \to \infty$).
\end{itemize}
The heuristic derivation for Regimes II and III is provided in Sections \ref{sec:heuristic_scaling_derivation_regime_ii}, and \ref{sec:heuristic_scaling_derivation_regime_iii} respectively.

In each of these regimes, we want to understand the scale of all quantities that arise in the forward and backward computations in the network, both at initialization and during training. While the initial random parameters are independent of the data distribution, correlations develop over time since updates carry information about the data; this is the case for both SGD and Adam~\citep{tp5_2022,yang_tp4b_2023, everett2024scaling}. To derive the correct scaling it is crucial to consider the correlations that arise during training and how these correlations propagate forward and backward through the network. In general, after one step of SGD the full set of post-update correlations is already present\citep{yang_feature_2021, tp5_2022, bordelon2022self, everett2024scaling}, so analysing the first two forward and backward passes suffices to determine the asymptotic width scaling of every quantity arising in the computation\footnote{When $W_{L+1}^{(0)} = 0$, only the readout updates in the first SGD step, since the gradient into every non-readout parameter carries a factor of $W_{L+1}^\top$ that vanishes at initialization. A third forward and backward pass at $\theta^{(2)}$ is then needed, since the non-readout parameters first move at $t=2$.}.

\subsection{Standing assumptions and notation}
\label{ssec:standing}

We track leading-order coordinate-scale exponents only, ignoring $\Theta(1)$ prefactors.

\paragraph{Standing assumptions.}
\begin{description}[font=\bfseries,leftmargin=2em,labelindent=0pt]
   \item[\textbf{(B1)}] $\chi_t = \partial L/\partial f_t \in \Theta(1)$ at every step $t$; absorbed into $\Theta(1)$ scalar prefactors throughout.
    \item[\textbf{(B2)}] Nonlinearities preserve leading scaling exponents. For sigmoid gating with $1/M$ aggregation, $\phi_{i,t},\,\dot\phi_{i,t}\in\Theta(1)$.
\end{description}

\paragraph{Notation.} $X\in\Theta(\alpha)$ denotes entry-wise (coordinate) scale. Cumulative SGD updates: $W_t^\ell = W_0^\ell + \Delta_t W^\ell$, $\Delta_t X = X_t - X_{t-1}$, with init values subscripted by $0$.

\paragraph{Standard tools.} We invoke the following standard probabilistic facts as needed, without per-instance citation:
\begin{itemize}[leftmargin=1cm]
    \item Central limit theorem (CLT) for sums of approximately independent terms;
    \item Law of large numbers (LLN) for empirical averages;
    \item Operator-norm concentration for iid (sub-)Gaussian matrices: for $W\in\mathbb{R}^{n\times m}$ with iid $\mathcal{N}(0,\sigma^2)$ entries, $\|W\|_{op}=\sigma(\sqrt n+\sqrt m)\,(1+o(1))$ with high probability~\citep{vershynin2018high}.
\end{itemize}

\paragraph{Feature update decomposition.} For $h^\ell = W^\ell h^{\ell-1}$,
\[
    \Delta h^\ell \;=\; \underbrace{W_0^\ell\,\Delta h^{\ell-1}}_{\text{propagating}} \;+\; \underbrace{\Delta W^\ell\,h_0^{\ell-1}}_{\text{effective}} \;+\; \underbrace{\Delta W^\ell\,\Delta h^{\ell-1}}_{\text{cross term}}.
\]

\subsection{Scaling derivation for Regime II}
\label{sec:heuristic_scaling_derivation_regime_ii}

In this section, we provide the heuristic scaling derivation for Regime II.

\begin{definition}[$\mu$P baseline, Regime II]
    \label{def:mup_baseline_regime_ii}
    \leavevmode
    \begin{itemize}
        \item \textbf{Initialization.} All parameters are drawn \textit{independently} according to:
        \[
            W_0^{1} \sim \mathcal{N}(0, D^{-1}), \; 
            Q_0 \sim \mathcal{N}(0, N^{-1}), \; 
            W_0^{2,i} \sim \mathcal{N}(0, N^{-1}), 
        \]
        \[
            W_0^{3,i} \sim \mathcal{N}(0, N_e^{-1}), \; 
            W_0^{4} \sim \mathcal{N}(0, N^{-2}), \; \text{for } i \in [M].
        \]
       
        \item \textbf{SGD learning rates.} $\eta_1 = \eta\, N$, $\;\eta_Q = \eta (M/N)$, $\;\eta_2 = (M/N) \eta$, $\;\eta_3 = (M N) \eta$, $\;\eta_4 = \eta\, N^{-1}$.
        \item \textbf{Adam learning rates.} $\eta_1 = \eta$, $\;\eta_Q = \eta_2 = \eta_4 = \eta\, N^{-1}$, $\;\eta_3 = 1/N_e \eta$.
        \item \textbf{Adam epsilon.} $\epsilon_1 = \epsilon\, N^{-1}$, $\;\epsilon_Q = \epsilon\, M^{-1}$, $\;\epsilon_2 = \epsilon\, M^{-1}$, $\;\epsilon_3 = \epsilon\, N^{-1} M^{-1}$, $\;\epsilon_4 = \epsilon$.
    \end{itemize}
\end{definition}

\begin{definition}[SSP, Regime II]
    \label{def:ssp_regime_ii}
    \leavevmode
    \begin{itemize}
        \item \textbf{Initialization.} All parameters are drawn \textit{independently} according to:
        \[
            W_0^{1} \sim \mathcal{N}(0, D^{-1}), \; 
            Q_0 \sim \mathcal{N}(0, N^{-1}), \; 
            W_0^{2,i} \sim \mathcal{N}(0, N^{-1}), 
        \]
        \[
            W_0^{3,i} \sim \mathcal{N}(0, M N_e^{-1}), \; 
            W_0^{4} \sim \mathcal{N}(0, N^{-2}), \; \text{for } i \in [M].
        \]
       
        \item \textbf{SGD learning rates.} $\eta_1 = \eta\, N$, $\;\eta_Q = \eta (M/N)$, $\;\eta_2 = (M/N) \eta$, $\;\eta_3 = (M N) \eta$, $\;\eta_4 = \eta\, N^{-1}$.
        \item \textbf{Adam learning rates.} $\eta_1 = \eta$, $\;\eta_Q = \eta_2 = \eta_4 = \eta\, N^{-1}$, $\;\eta_3 = 1/N_e \eta$.
        \item \textbf{Adam epsilon.} $\epsilon_1 = \epsilon\, N^{-1}$, $\;\epsilon_Q = \epsilon\, M^{-1}$, $\;\epsilon_2 = \epsilon\, M^{-1}$, $\;\epsilon_3 = \epsilon\, N^{-1} M^{-1}$, $\;\epsilon_4 = \epsilon$.
    \end{itemize}
\end{definition}

\subsubsection{Deriving $\mu$P in Regime II}
\label{ssec:mup_r2_consolidated}

This subsection presents a self-contained derivation for the $\mu$P baseline (Definition~\ref{def:mup_baseline_regime_ii}) in Regime II.

\paragraph{Setup and standing assumptions}
\label{ssec:mup_r2_setup}

\paragraph{Architecture.} As in §\ref{sec:preliminaries}: minimal MoE with sigmoid gating and explicit $1/M$ aggregation $h^3 = (1/M)\sum_i\phi_i\,h^{3,i}$.

\paragraph{Initialization ($\mu$P baseline).} Per Definition~\ref{def:mup_baseline_regime_ii}, with all parameters drawn independently:
\begin{align*}
    (W_0^1)_{ab}&\sim\mathcal{N}(0,1/D), &
    (Q_0)_{ia}&\sim\mathcal{N}(0,1/N),\\
    (W_0^{2,i})_{ab}&\sim\mathcal{N}(0,1/N), &
    (W_0^{3,i})_{ab}&\sim\mathcal{N}(0,1/N_e),\\
    (W_0^4)_a&\sim\mathcal{N}(0,1/N^2).
\end{align*}
Resulting entry scales: $W_0^4$ entries $\Theta(1/N)$, $W_0^{2,i}$ entries $\Theta(1/\sqrt N)$, $W_0^{3,i}$ entries $\Theta(1)$ (since $\sigma_3^2=1/N_e=\Theta(1)$).

\paragraph{Learning rates (SGD).} $\eta_1=\eta N$, $\eta_Q=\eta M/N$, $\eta_2=\eta M/N$, $\eta_3=\eta MN$, $\eta_4=\eta/N$, $\eta\in\Theta(1)$.

\paragraph{Gating.} Sigmoid: $\phi_{i,t}, \dot\phi_{i,t}\in\Theta(1)$.

\paragraph{Auxiliary scaling lemmas}
\label{ssec:mup_r2_lemmas}

We reuse Lemma~\ref{lem:ssp_gram} from §\ref{ssec:ssp_consol_lemmas}.

\textbf{Specialization of Lemma~\ref{lem:ssp_gram} to the rank-deficient initialization Gram matrices in Regime~II.} For the random initialization $W_0^{3,i}\in\mathbb{R}^{N\times N_e}$ with $\sigma_W^2=1/N_e$ and $N_e=\Theta(1)$, the Gram matrix $W_0^{3,i}(W_0^{3,i})^\top\in\mathbb{R}^{N\times N}$ has rank at most $N_e=\Theta(1)$: its diagonal mean is $1$ and its off-diagonal entries have coordinate scale $\Theta(1)$ (variance $1/N_e=\Theta(1)$). Consequently, acting on a vector $v=(W_0^4)^\top$ with entries of coordinate scale $\Theta(1/N)$, the product $W_0^{3,i}(W_0^{3,i})^\top\,v$ has entries of coordinate scale $\Theta(1/\sqrt N)$ in mostly-random directions, not aligned along $v$.

Analogously, for the random initialization $W_0^{2,i}\in\mathbb{R}^{N_e\times N}$ with $\sigma_W^2=1/N$, the Gram matrix $(W_0^{2,i})^\top W_0^{2,i}\in\mathbb{R}^{N\times N}$ has rank at most $N_e=\Theta(1)$ and entries of coordinate scale $\Theta(1/N)$. Acting on $h_0^1$, it produces entries of coordinate scale $\Theta(1/\sqrt N)$ in mostly-random directions, not aligned along $h_0^1$.

We refer to $W_0^{3,i}(W_0^{3,i})^\top$ and $(W_0^{2,i})^\top W_0^{2,i}$ collectively as the \emph{rank-deficient initialization Gram matrices} of Regime~II.

\paragraph{First forward pass}
\label{ssec:mup_r2_fwd1}

\[
    h_0^1 = W_0^1\,x \in \Theta(1). \quad (\text{CLT};\ \sigma_1^2=1/D,\ \|x\|^2\in\Theta(D)) \tag{R2$\mu$-F1.1}
\]
\[
    \psi_0 = Q_0\,h_0^1 \in \Theta(1). \quad (\text{CLT};\ \sigma_Q^2=1/N,\ \|h_0^1\|^2\in\Theta(N)) \tag{R2$\mu$-F1.2}
\]
\[
    \phi_{i,0} = \sigma(\psi_{i,0}) \in \Theta(1). \quad (\text{sigmoid bounded; gating assumption}) \tag{R2$\mu$-F1.3}
\]
\begin{align*}
    h_0^{2,i} &= W_0^{2,i}\,h_0^1 \in \Theta(1),\quad \text{independent across }i\\
    & \quad (\text{CLT};\ \sigma_2^2=1/N,\ \|h_0^1\|^2\in\Theta(N);\\
    & \qquad \|h_0^{2,i}\|^2=N_e\cdot\Theta(1)=\Theta(1)\text{ since }N_e=\Theta(1),\text{ not }\Theta(N_e)\to\infty). \tag{R2$\mu$-F1.4}
\end{align*}
\begin{align*}
    h_0^{3,i} &= W_0^{3,i}\,h_0^{2,i} \in \Theta(1)\\
    & \quad (\text{variance per entry }\sigma_3^2\|h_0^{2,i}\|^2=(1/N_e)\cdot\Theta(1)=\Theta(1);\\
    & \qquad \text{direct variance, not asymptotic CLT, since }N_e=O(1)). \tag{R2$\mu$-F1.5}
\end{align*}
\begin{align*}
    h_0^3 &= (1/M)\textstyle\sum_i\phi_{i,0}\,h_0^{3,i} \in \Theta(1/\sqrt M)=\Theta(1/\sqrt N)\\
    & \quad (\text{cross-$i$ CLT on entries of coordinate scale }\Theta(1);\\
    & \qquad \{h_0^{3,i}\}\text{ independent across }i). \tag{R2$\mu$-F1.6}
\end{align*}
\begin{align*}
    f_0 &= W_0^4\,h_0^3 = \langle W_0^4, h_0^3\rangle \in \Theta(1/N)\\
    & \quad (\text{variance }\textstyle\sum_a\Theta(1/N^2)\cdot\Theta(1/N)=\Theta(1/N^2);\\
    & \qquad h_0^3\text{ entry scale }\Theta(1/\sqrt N)\text{ via (R2$\mu$-F1.6)}). \tag{R2$\mu$-F1.7}
\end{align*}

\paragraph{First backward pass and step-1 updates}
\label{ssec:mup_r2_bwd1}

\paragraph{New intermediate scaling.} The intermediate quantity $(W_0^{3,i})^\top(W_0^4)^\top$ has entries of variance $\sigma_3^2\,\|W_0^4\|^2 = \Theta(1)\cdot\Theta(1/N) = \Theta(1/N)$, hence coordinate scale $\Theta(1/\sqrt N)$.

\paragraph{Per-layer gradients (with $W_0^4/\Delta W^4$ split).}

\[
    \partial f_0/\partial h_0^3 = (W_0^4)^\top \in \Theta(1/N). \quad (W_0^4\text{ entries }\Theta(1/N);\ \Delta_t W^4=0\text{ at }t=0) \tag{R2$\mu$-B1.1}
\]
\[
    \partial f_0/\partial h_0^{3,i} = (\phi_{i,0}/M)(W_0^4)^\top \in \Theta(1/(MN))=\Theta(1/N^2). \quad (\phi_{i,0}\in\Theta(1)) \tag{R2$\mu$-B1.2}
\]
\begin{align*}
    \partial f_0/\partial h_0^{2,i} &= (\phi_{i,0}/M)(W_0^{3,i})^\top(W_0^4)^\top \in \Theta(1/(M\sqrt N))=\Theta(1/N^{3/2})\\
    & \quad ((W_0^{3,i})^\top(W_0^4)^\top\in\Theta(1/\sqrt N)\text{ from new intermediate scaling above}). \tag{R2$\mu$-B1.3}
\end{align*}
\begin{align*}
    \partial f_0/\partial \phi_{i,0} &= (1/M)\langle h_0^{3,i}, W_0^4\rangle \in \Theta(1/N^{3/2})\\
    & \quad (\langle h_0^{3,i}, W_0^4\rangle\text{ variance }\textstyle\sum_a\Theta(1)\cdot\Theta(1/N^2)=\Theta(1/N),\text{ coord }\Theta(1/\sqrt N);\\
    & \qquad h_0^{3,i}\text{ entry scale }\Theta(1)\text{ via (R2$\mu$-F1.5)}). \tag{R2$\mu$-B1.4}
\end{align*}
\begin{align*}
    (\partial f_0/\partial h_0^1)_\mathrm{exp} &= (1/M)\textstyle\sum_i\phi_{i,0}(W_0^{2,i})^\top(W_0^{3,i})^\top(W_0^4)^\top \in \Theta(1/N^{3/2})\\
    & \quad (\text{at }t=0\text{ only the all-init }A_{4.1}\text{ piece is non-zero;}\\
    & \qquad \text{per summand }(W_0^{2,i})^\top\text{ acts on }(W_0^{3,i})^\top(W_0^4)^\top\in\Theta(1/\sqrt N)\\
    & \qquad \text{with }\|v\|^2=N_e\,\Theta(1/N)=\Theta(1/N);\\
    & \qquad \text{variance }(1/N)\,\Theta(1/N)=\Theta(1/N^2),\text{ per-}i\text{ coord }\Theta(1/N);\\
    & \qquad \text{cross-$i$ CLT}). \tag{R2$\mu$-B1.5}
\end{align*}
\begin{align*}
    (\partial f_0/\partial h_0^1)_\mathrm{router} &= (1/M)Q_0^\top v,\quad v_i:=\dot\phi_{i,0}\langle h_0^{3,i}, W_0^4\rangle\\
    &\in \Theta(1/N^{3/2})\\
    & \quad (v_i\in\Theta(1/\sqrt N)\text{ via (R2$\mu$-B1.4)};\ \|v\|^2=M\,\Theta(1/N)=\Theta(1);\\
    & \qquad \text{cross-layer CLT giving }Q_0^\top v\text{ coord }\Theta(1/\sqrt N);\\
    & \qquad \text{W}^{3,i}\text{-split }v_i = v^I_i + v^U_i\text{ where}\\
    & \qquad v^I_i := \dot\phi_{i,0}\langle (W_0^{3,i})h_0^{2,i}, W_0^4\rangle\\
    & \qquad v^U_i := \dot\phi_{i,0}\langle (\Delta_0 W^{3,i})h_0^{2,i}, W_0^4\rangle = 0\text{ since }\Delta_0 W^{3,i}=0;\\
    & \qquad \text{so }(1/M)Q_0^\top v^I = \Theta(1/N^{3/2})\text{ and }(1/M)Q_0^\top v^U = 0). \tag{R2$\mu$-B1.6}
\end{align*}
\[
    \partial f_0/\partial h_0^1 = (\partial f_0/\partial h_0^1)_\mathrm{exp} + (\partial f_0/\partial h_0^1)_\mathrm{router} \in \Theta(1/N^{3/2}). \tag{R2$\mu$-B1.7}
\]

\paragraph{Step-1 parameter updates.}

\begin{align*}
    \Delta_1 W^4 &= -(\eta/N)\chi_0(h_0^3)^\top \in \Theta(1/N^{3/2})\\
    & \quad (h_0^3\text{ entry scale }\Theta(1/\sqrt N)\text{ via (R2$\mu$-F1.6)};\ \text{sub-leading vs }W_0^4\in\Theta(1/N)). \tag{R2$\mu$-U1.4}
\end{align*}
\begin{align*}
    \Delta_1 W^{3,i} &= -\eta_3\chi_0(\phi_{i,0}/M)(W_0^4)^\top(h_0^{2,i})^\top = -\eta N\chi_0\phi_{i,0}\,(W_0^4)^\top(h_0^{2,i})^\top\\
    &\in \Theta(1)\text{ rank-1 along }(W_0^4)^\top\otimes h_0^{2,i}\\
    & \quad (\eta_3=\eta MN;\ h_0^{2,i}\in\Theta(1)\text{ via (R2$\mu$-F1.4)};\\
    & \qquad \text{same scale as }W_0^{3,i}\in\Theta(1)). \tag{R2$\mu$-U1.3}
\end{align*}
\begin{align*}
    \Delta_1 W^{2,i} &= -(\eta\chi_0\phi_{i,0}/N)(W_0^{3,i})^\top(W_0^4)^\top(h_0^1)^\top \in \Theta(1/N^{3/2})\text{ rank-1}\\
    & \quad (\eta_2=\eta M/N;\ (W_0^{3,i})^\top(W_0^4)^\top\in\Theta(1/\sqrt N)\text{ from new intermediate scaling};\\
    & \qquad \text{sub-leading vs }W_0^{2,i}\in\Theta(1/\sqrt N)). \tag{R2$\mu$-U1.2}
\end{align*}
\begin{align*}
    \Delta_1 W^1 &= -\eta_1\chi_0(\partial f_0/\partial h_0^1)\,x^\top \in \Theta(1/\sqrt N)\\
    & \quad (\eta_1=\eta N;\ \partial f_0/\partial h_0^1\in\Theta(1/N^{3/2})\text{ via (R2$\mu$-B1.7)}). \tag{R2$\mu$-U1.1a}
\end{align*}
\[
    \Delta_1 h^1 = \Delta_1 W^1\,x \in \Theta(1/\sqrt N) \text{ aligned along } \partial f_0/\partial h_0^1. \tag{R2$\mu$-U1.1b}
\]
\[
    \Delta_1 Q = -\eta_Q\chi_0(\partial f_0/\partial\phi)(h_0^1)^\top \in \Theta(1/N^{3/2}). \quad (\partial f_0/\partial\phi\in\Theta(1/N^{3/2})\text{ via (R2$\mu$-B1.4)}) \tag{R2$\mu$-U1.Q}
\]

\paragraph{Second forward pass}
\label{ssec:mup_r2_fwd2}

\[
    h_1^1 = h_0^1 + \Delta_1 h^1 \in \Theta(1). \quad (h_0^1\in\Theta(1)\text{ dominates }\Delta_1 h^1\in\Theta(1/\sqrt N)\text{ via (R2$\mu$-U1.1b)}) \tag{R2$\mu$-F2.1}
\]
\begin{align*}
    \psi_1 &= \psi_0 + \Delta_1\psi \in \Theta(1),\quad \phi_1\in\Theta(1) \tag{R2$\mu$-F2.2--R2$\mu$-F2.3}\\
    & \quad (\Delta_1\psi\in\Theta(1/\sqrt N)\text{ from }Q_0\Delta_1 h^1\text{ and }\Delta_1 Q\,h_0^1;\\
    & \qquad \text{scales by (R2$\mu$-U1.1b), (R2$\mu$-U1.Q)};\\
    & \qquad \text{cross }\Delta_1 Q\,\Delta_1 h^1\in\Theta(1/N^{3/2})\text{ sub-leading via }\partial f_0/\partial h_0^1\text{ via (R2$\mu$-B1.7)})
\end{align*}

\noindent (R2$\mu$-F2.4) $h_1^{2,i} = W_1^{2,i}\,h_1^1$, four-piece decomposition:
\[
    h_1^{2,i} = h_0^{2,i} + W_0^{2,i}\Delta_1 h^1 + \Delta_1 W^{2,i}\,h_0^1 + \Delta_1 W^{2,i}\,\Delta_1 h^1. \tag{R2$\mu$-F2.4}
\]
\[
    \text{init: } W_0^{2,i}h_0^1 = h_0^{2,i} \in \Theta(1). \quad (\text{R2$\mu$-F1.4}) \tag{R2$\mu$-F2.4a}
\]

\begin{align*}
    \text{prop: } W_0^{2,i}\Delta_1 h^1 &\in \Theta(1/\sqrt N) \tag{R2$\mu$-F2.4b}\\
    & \quad (\sigma_2^2\|\Delta_1 h^1\|^2 = (1/N)\,\Theta(1) = \Theta(1/N);\\
    & \qquad \|\Delta_1 h^1\|^2\in\Theta(1)\text{ via (R2$\mu$-U1.1b)})
\end{align*}
\begin{align*}
    \text{eff: } \Delta_1 W^{2,i}\,h_0^1 &= -(\eta\chi_0\phi_{i,0}/N)\,\|h_0^1\|^2\,(W_0^{3,i})^\top(W_0^4)^\top \in \Theta(1/\sqrt N) \tag{R2$\mu$-F2.4c}\\
    & \quad (\text{(R2$\mu$-U1.2)substitution};\\
    & \qquad (W_0^{3,i})^\top(W_0^4)^\top\in\Theta(1/\sqrt N)\text{ from new intermediate scaling};\\
    & \qquad \|h_0^1\|^2\in\Theta(N))
\end{align*}

\begin{align*}
    \text{cross: } \Delta_1 W^{2,i}\,\Delta_1 h^1 &= -(\eta\chi_0\phi_{i,0}/N)\,\langle h_0^1,\Delta_1 h^1\rangle\,(W_0^{3,i})^\top(W_0^4)^\top \\
    &\in \Theta(1/N^{3/2}) \tag{R2$\mu$-F2.4d}\\
    & \quad ((W_0^{3,i})^\top(W_0^4)^\top\in\Theta(1/\sqrt N)\text{ from new intermediate scaling};\\
    & \qquad \langle h_0^1,\Delta_1 h^1\rangle\in\Theta(1)\text{ random, derived via} \\
    & \qquad\quad \partial f_0/\partial h_0^1\in\Theta(1/N^{3/2})\text{ via (R2$\mu$-B1.7)};\\
    & \qquad \text{sub-leading vs propagating and effective})
\end{align*}

\noindent (R2$\mu$-F2.5) $h_1^{3,i} = W_1^{3,i}\,h_1^{2,i}$, four-piece decomposition:
\[
    h_1^{3,i} = h_0^{3,i} + W_0^{3,i}\Delta_1 h^{2,i} + \Delta_1 W^{3,i}\,h_0^{2,i} + \Delta_1 W^{3,i}\,\Delta_1 h^{2,i}. \tag{R2$\mu$-F2.5}
\]
\[
    \text{init: } h_0^{3,i} \in \Theta(1). \quad (\text{R2$\mu$-F1.5}) \tag{R2$\mu$-F2.5a}
\]
\begin{align*}
    \text{prop: } W_0^{3,i}\Delta_1 h^{2,i} &\in \Theta(1/\sqrt N) \tag{R2$\mu$-F2.5b}\\
    & \quad (\sigma_3^2\|\Delta_1 h^{2,i}\|^2 = \Theta(1)\cdot\Theta(1/N) = \Theta(1/N);\\
    & \qquad \|\Delta_1 h^{2,i}\|^2\in\Theta(1)\text{ via (R2$\mu$-F2.4)})
\end{align*}
\begin{align*}
    \text{eff: } \Delta_1 W^{3,i}\,h_0^{2,i} &= -\eta N\chi_0\phi_{i,0}\,\|h_0^{2,i}\|^2\,(W_0^4)^\top \in \Theta(1) \text{ along }(W_0^4)^\top \tag{R2$\mu$-F2.5c}\\
    & \quad (\text{(R2$\mu$-U1.3)substitution};\ \|h_0^{2,i}\|^2\in\Theta(1)\text{ via (R2$\mu$-F1.4)};\\
    & \qquad (W_0^4)^\top\in\Theta(1/N))
\end{align*}
\begin{align*}
    \text{cross: } \Delta_1 W^{3,i}\,\Delta_1 h^{2,i} &= -\eta N\chi_0\phi_{i,0}\,\langle h_0^{2,i},\Delta_1 h^{2,i}\rangle\,(W_0^4)^\top \in \Theta(1/\sqrt N) \text{ along }(W_0^4)^\top \tag{R2$\mu$-F2.5d}\\
    & \quad (\text{(R2$\mu$-U1.3)substitution};\\
    & \qquad \text{dominant piece of }\Delta_1 h^{2,i}\propto(W_0^{3,i})^\top(W_0^4)^\top\text{ via (R2$\mu$-F2.4c)},\\
    & \qquad \text{so }\langle h_0^{2,i},\Delta_1 h^{2,i}\rangle\sim -(\eta\chi_0\phi/N)\|h_0^1\|^2\langle h_0^{3,i},W_0^4\rangle\in\Theta(1/\sqrt N)\\
    & \qquad \text{random across }i,\text{ via (R2$\mu$-B1.4)};\\
    & \qquad \text{random sign across }i,\text{ sub-leading vs effective})
\end{align*}
\textbf{Effective dominates:} $\Delta_1 h^{3,i}\in\Theta(1)$ aligned along $(W_0^4)^\top$ — the engine of feature learning at $t=1$.

\noindent (R2$\mu$-F2.6) $h_1^3 = A_1 + A_{2,1} + A_{2,2} + A_3 + D$, where
\begin{align*}
    A_1 &:= \tfrac{1}{M}\textstyle\sum_i\phi_{i,1}\,h_0^{3,i}, &
    A_{2,1} &:= \tfrac{1}{M}\textstyle\sum_i\phi_{i,1}\,W_0^{3,i}W_0^{2,i}\,\Delta_1 h^1,\\
    A_{2,2} &:= \tfrac{1}{M}\textstyle\sum_i\phi_{i,1}\,W_0^{3,i}\,\Delta_1 W^{2,i}\,h_0^1, &
    A_3 &:= \tfrac{1}{M}\textstyle\sum_i\phi_{i,1}\,\Delta_1 W^{3,i}\,h_0^{2,i},\\
    D &:= \tfrac{1}{M}\textstyle\sum_i\phi_{i,1}\,\Delta_1 W^{3,i}\,\Delta_1 h^{2,i}.
\end{align*}
\[
    A_1 \in \Theta(1/\sqrt N). \quad (\text{cross-$i$ CLT on }\Theta(1)\text{-entry independent vectors}) \tag{R2$\mu$-F2.6a}
\]
\begin{align*}
    A_{2,1} &\in \Theta(1/(\sqrt M\sqrt N)) = \Theta(1/N) \tag{R2$\mu$-F2.6b}\\
    & \quad (\text{per summand chain CLT entries }\Theta(1/\sqrt N)\text{ random};\\
    & \qquad \|\Delta_1 h^1\|^2\in\Theta(1)\text{ via (R2$\mu$-U1.1b)};\ \text{cross-$i$ CLT})
\end{align*}
\begin{align*}
    A_{2,2} &= -(\eta\chi_0/N)\,\|h_0^1\|^2\,\tfrac{1}{M}\textstyle\sum_i\phi_{i,1}\phi_{i,0}\,W_0^{3,i}(W_0^{3,i})^\top(W_0^4)^\top\\
    &\in \Theta(1/N)\text{ in mostly-random directions} \tag{R2$\mu$-F2.6c}\\
    & \quad (\text{(R2$\mu$-U1.2)substitution};\\
    & \qquad W_0^{3,i}(W_0^{3,i})^\top(W_0^4)^\top\in\Theta(1/\sqrt N)\text{ in mostly-random directions}\\
    & \qquad \text{by rank-deficient initialization Gram (\S\ref{ssec:mup_r2_lemmas})};\\
    & \qquad \|h_0^1\|^2\in\Theta(N);\ \text{per summand entries }\Theta(1/\sqrt N),\text{ cross-$i$ CLT})
\end{align*}
\begin{align*}
    A_3 &= -\eta N\chi_0\,(W_0^4)^\top\,\Bigl(\tfrac{1}{M}\textstyle\sum_i\phi_{i,1}\phi_{i,0}\,\|h_0^{2,i}\|^2\Bigr) \in \Theta(1) \text{ along } (W_0^4)^\top \tag{R2$\mu$-F2.6d}\\
    & \quad (\text{(R2$\mu$-U1.3)substitution};\ \|h_0^{2,i}\|^2\in\Theta(1)\text{ via (R2$\mu$-F1.4)};\ \text{LLN};\\
    & \qquad (W_0^4)^\top\in\Theta(1/N);\ \text{entry }\eta N\cdot(1/N)\cdot\Theta(1)=\Theta(1))
\end{align*}
\begin{align*}
    D &= -\eta N\chi_0\,(W_0^4)^\top\,\tfrac{1}{M}\textstyle\sum_i\phi_{i,1}\phi_{i,0}\,\langle h_0^{2,i},\Delta_1 h^{2,i}\rangle\\
    &\in \Theta(1/N)\text{ along }(W_0^4)^\top \tag{R2$\mu$-F2.6e}\\
    & \quad (\text{(R2$\mu$-U1.3)substitution};\\
    & \qquad \langle h_0^{2,i},\Delta_1 h^{2,i}\rangle\in\Theta(1/\sqrt N)\text{ random across }i\\
    & \qquad \text{via }\langle h_0^{3,i},W_0^4\rangle\in\Theta(1/\sqrt N)\text{ via (R2$\mu$-B1.4)};\\
    & \qquad \text{cross-$i$ CLT in the random factor};\\
    & \qquad \text{sub-leading vs }A_3\in\Theta(1))
\end{align*}
\[
    h_1^3 \in \Theta(1) \text{ along } (W_0^4)^\top, \text{ from } A_3 \text{ alone}. \tag{R2$\mu$-F2.6}
\]
\begin{align*}
    f_1 &= W_1^4\,h_1^3 \approx \langle W_0^4, A_3\rangle = -\eta N\chi_0\,\|W_0^4\|^2\,\Bigl(\tfrac{1}{M}\textstyle\sum_i\phi_{i,1}\phi_{i,0}\,\|h_0^{2,i}\|^2\Bigr) \in \Theta(1) \tag{R2$\mu$-F2.7}\\
    & \quad (A_3\text{ aligned along }(W_0^4)^\top\text{ via (R2$\mu$-F2.6d)};\\
    & \qquad \|W_0^4\|^2 = N\cdot(1/N^2) = 1/N;\ \|h_0^{2,i}\|^2\in\Theta(1)\text{ via (R2$\mu$-F1.4)};\\
    & \qquad \text{net }\eta N\cdot\Theta(1)\cdot(1/N) = \Theta(1))
\end{align*}

\paragraph{Second backward pass and step-2 updates}
\label{ssec:mup_r2_bwd2}

\paragraph{New intermediate at $t=1$.}
\begin{align*}
    (\Delta_1 W^{3,i})^\top(W_1^4)^\top &\approx -\eta N\chi_0\phi_{i,0}\,\|W_0^4\|^2\,h_0^{2,i} \in \Theta(1) \text{ aligned } \parallel h_0^{2,i}\\
    & \quad (\text{(R2$\mu$-U1.3)rank-1 substitution};\ \|W_0^4\|^2 = 1/N;\\
    & \qquad h_0^{2,i}\in\Theta(1)\text{ via (R2$\mu$-F1.4)};\ \text{net }\eta N\cdot(1/N) = \Theta(1))
\end{align*}
This $\Theta(1)$ aligned-along-$h_0^{2,i}$ contribution from the rank-1 update is what drives $\partial f_1/\partial h_1^{2,i}$ to climb from $\Theta(1/N^{3/2})$ at $t=0$ to $\Theta(1/N)$ at $t=1$.

\paragraph{Per-layer gradients (with $W_0^4/\Delta W^4$ binary split).}

\begin{align*}
    \partial f_1/\partial h_1^3 &= (W_0^4)^\top + (\Delta_1 W^4)^\top \in \Theta(1/N) \tag{R2$\mu$-B2.1}\\
    & \quad (W_0^4\in\Theta(1/N)\text{ dominates }\Delta_1 W^4\in\Theta(1/N^{3/2})\text{ via (R2$\mu$-U1.4)})
\end{align*}
\begin{align*}
    \partial f_1/\partial h_1^{3,i} &= (\phi_{i,1}/M)(W_1^4)^\top \in \Theta(1/N^2) \tag{R2$\mu$-B2.2}\\
    & \quad (\text{init piece }\Theta(1/N^2);\ \Delta_1 W^4\text{ piece }\Theta(1/N^{5/2})\text{ via (R2$\mu$-U1.4)})
\end{align*}

\noindent (R2$\mu$-B2.3) $\partial f_1/\partial h_1^{2,i} = (\phi_{i,1}/M)(W_1^{3,i})^\top(W_1^4)^\top$, four-piece decomposition:
\begin{align*}
    (\phi_{i,1}/M)(W_0^{3,i})^\top(W_0^4)^\top &\in \Theta(1/N^{3/2}) \tag{R2$\mu$-B2.3a}\\
    & \quad ((W_0^{3,i})^\top(W_0^4)^\top\in\Theta(1/\sqrt N)\text{ from new intermediate scaling at }t=0)
\end{align*}
\begin{align*}
    (\phi_{i,1}/M)(W_0^{3,i})^\top(\Delta_1 W^4)^\top &= -(\eta\chi_0\phi_{i,0}/(MN))(W_0^{3,i})^\top h_0^3 \\
    &\in \Theta(1/N^2) \tag{R2$\mu$-B2.3b}\\
    & \quad (\text{(R2$\mu$-U1.4)substitution};\\
    & \qquad (W_0^{3,i})^\top h_0^3\in\Theta(1)\\
    & \qquad\qquad \text{from }j=i\text{ coherent contribution to the average})
\end{align*}
\begin{align*}
    (\phi_{i,1}/M)(\Delta_1 W^{3,i})^\top(W_0^4)^\top &\in \Theta(1/N) \text{ along } h_0^{2,i} \tag{R2$\mu$-B2.3c}\\
    & \quad ((\Delta_1 W^{3,i})^\top(W_0^4)^\top\in\Theta(1)\text{ along }h_0^{2,i}\\
    & \qquad \text{from new intermediate at }t=1\text{ above};\ \textbf{dominant})
\end{align*}
\begin{align*}
    (\phi_{i,1}/M)(\Delta_1 W^{3,i})^\top(\Delta_1 W^4)^\top &\in \Theta(1/N^2) \tag{R2$\mu$-B2.3d}\\
    & \quad (\text{same alignment as (R2$\mu$-B2.3c)};\\
    & \qquad \Delta_1 W^4\text{ smaller by }1/\sqrt N\text{ via (R2$\mu$-U1.4)})
\end{align*}
\[
    \partial f_1/\partial h_1^{2,i} \in \Theta(1/N) \text{ along } h_0^{2,i}, \text{ from (R2$\mu$-B2.3c)}. \tag{R2$\mu$-B2.3}
\]
\begin{align*}
    \partial f_1/\partial \phi_{i,1} &= (1/M)\langle h_1^{3,i}, W_1^4\rangle \in \Theta(1/N) \tag{R2$\mu$-B2.4}\\
    & \quad (\text{Decompose }h_1^{3,i} = (W_0^{3,i})\,h_1^{2,i} + (\Delta_1 W^{3,i})\,h_1^{2,i}\text{ and }W_1^4 = W_0^4 + \Delta_1 W^4;\\
    & \qquad \text{four-piece grid:}\\
    & \qquad (\text{init}\cdot\text{init})\ \langle (W_0^{3,i})\,h_1^{2,i}, W_0^4\rangle\in\Theta(1/\sqrt N)\\
    & \qquad\quad \text{via random IP }\langle h_0^{3,i}, W_0^4\rangle\in\Theta(1/\sqrt N)\text{ (R2$\mu$-B1.4)};\\
    & \qquad (\text{init}\cdot\text{update})\ \langle (W_0^{3,i})\,h_1^{2,i}, \Delta_1 W^4\rangle = -(\chi_0/N)\langle h_0^{3,i}, h_0^3\rangle\in\Theta(1/N)\\
    & \qquad\quad \text{via }\langle h_0^{3,i}, h_0^3\rangle\in\Theta(1)\text{ (}j=i\text{ diagonal }(\phi_{i,0}/M)\|h_0^{3,i}\|^2=\Theta(1)\text{ in }\mu\text{P)};\\
    & \qquad (\text{update}\cdot\text{init})\ \langle (\Delta_1 W^{3,i})\,h_1^{2,i}, W_0^4\rangle = -\eta N\chi_0\phi_{i,0}\|h_0^{2,i}\|^2\|W_0^4\|^2\in\Theta(1)\\
    & \qquad\quad \text{via rank-1 alignment of }\Delta_1 W^{3,i}h_0^{2,i}\text{ with }(W_0^4)^\top;\ \eta N\cdot\Theta(1)\cdot\Theta(1/N)=\Theta(1);\\
    & \qquad (\text{update}\cdot\text{update})\ \langle (\Delta_1 W^{3,i})\,h_1^{2,i}, \Delta_1 W^4\rangle = \kappa\langle W_0^4, \Delta_1 W^4\rangle\in\Theta(1/N)\\
    & \qquad\quad \text{via }\kappa\in\Theta(N)\text{ (rank-1 prefactor) and }\langle W_0^4, \Delta_1 W^4\rangle\in\Theta(1/N^2);\\
    & \qquad /M\text{: }(\Theta(1/N^{3/2}),\Theta(1/N^2),\Theta(1/N),\Theta(1/N^2))\text{ respectively};\\
    & \qquad (\text{update}\cdot\text{init})\text{ dominates; total }\Theta(1/N))
\end{align*}

\paragraph{Expert pathway: $(\partial f_1/\partial h_1^1)_\mathrm{exp} = A_4+A_5+A_6+E$, full 8-piece expansion.}

Each of $A_4, A_5, A_6, E$ splits into a $.1$ piece (using $W_0^4$) and a $.2$ piece (using $\Delta_1 W^4$):
\begin{align*}
    A_{4.1} &\in \Theta(1/N^{3/2}). \quad (\text{R2$\mu$-B1.5}) \tag{R2$\mu$-B2.5a.1}\\
    A_{4.2} &\in \Theta(1/N^2). \quad (\Delta_1 W^4\text{ smaller by }1/\sqrt N\text{ via (R2$\mu$-U1.4)}) \tag{R2$\mu$-B2.5a.2}
\end{align*}
\begin{align*}
    A_{5.1} &= \tfrac{1}{M}\textstyle\sum_i\phi_{i,1}(W_0^{2,i})^\top(\Delta_1 W^{3,i})^\top(W_0^4)^\top\\
    &\in \Theta(1/N) \text{ in mostly-random directions} \tag{R2$\mu$-B2.5b.1}\\
    & \quad ((\Delta_1 W^{3,i})^\top(W_0^4)^\top\in\Theta(1)\text{ along }h_0^{2,i}\text{ from new intermediate at }t=1;\\
    & \qquad (W_0^{2,i})^\top h_0^{2,i}=(W_0^{2,i})^\top W_0^{2,i}h_0^1\in\Theta(1/\sqrt N)\text{ in mostly-random directions}\\
    & \qquad \text{by rank-deficient initialization Gram (\S\ref{ssec:mup_r2_lemmas})};\\
    & \qquad \text{coherent }(N_e/N)\,h_0^1\text{ piece is }\Theta(1/N)\text{ sub-leading};\\
    & \qquad \text{cross-$i$ CLT: }\Theta(1/\sqrt N)/\sqrt M = \Theta(1/N))
\end{align*}
\[
    A_{5.2} \in \Theta(1/N^2). \quad (\text{same structure as (R2$\mu$-B2.5b.1)};\ \Delta_1 W^4\text{ smaller by }1/\sqrt N\text{ via (R2$\mu$-U1.4)}) \tag{R2$\mu$-B2.5b.2}
\]
\begin{align*}
    A_{6.1} &= \tfrac{1}{M}\textstyle\sum_i\phi_{i,1}(\Delta_1 W^{2,i})^\top(W_0^{3,i})^\top(W_0^4)^\top\\
    &= -(\eta\chi_0/N)\,h_0^1\,\tfrac{1}{M}\textstyle\sum_i\phi_{i,1}\phi_{i,0}\,S_i \in \Theta(1/N^2) \text{ along } h_0^1 \tag{R2$\mu$-B2.5c.1}\\
    & \quad (\text{(R2$\mu$-U1.2)substitution};\ S_i := W_0^4 W_0^{3,i}(W_0^{3,i})^\top(W_0^4)^\top\in\Theta(1/N)\\
    & \qquad \text{quadratic form in }W_0^4\text{ across rank-}N_e\text{ Gram, leading piece }\|W_0^4\|^2;\\
    & \qquad \text{per summand entries }(1/N)\cdot 1\cdot(1/N) = \Theta(1/N^2)\text{ aligned along }h_0^1)
\end{align*}
\[
    A_{6.2} \in \Theta(1/N^3). \quad (\text{same structure as (R2$\mu$-B2.5c.1)};\ \Delta_1 W^4\text{ smaller by }1/\sqrt N\text{ via (R2$\mu$-U1.4)}) \tag{R2$\mu$-B2.5c.2}
\]
\begin{align*}
    E_1 &= \tfrac{1}{M}\textstyle\sum_i\phi_{i,1}(\Delta_1 W^{2,i})^\top(\Delta_1 W^{3,i})^\top(W_0^4)^\top \in \Theta(1/N^2) \tag{R2$\mu$-B2.5d.1}\\
    & \quad (\text{combining (R2$\mu$-U1.2), (R2$\mu$-U1.3) substitutions};\\
    & \qquad \text{per summand involves }\langle W_0^4, h_0^{3,i}\rangle\in\Theta(1/\sqrt N)\text{ random via (R2$\mu$-B1.4)};\\
    & \qquad \text{per summand entries }\Theta(1/N^{3/2});\ \text{cross-$i$ CLT})
\end{align*}
\[
    E_2 \in \Theta(1/N^{5/2}). \quad (\text{same as (R2$\mu$-B2.5d.1)};\ \Delta_1 W^4\text{ smaller by }1/\sqrt N\text{ via (R2$\mu$-U1.4)}) \tag{R2$\mu$-B2.5d.2}
\]
\[
    (\partial f_1/\partial h_1^1)_\mathrm{exp} \in \Theta(1/N) \text{ in mostly-random directions, from } A_{5.1} \text{ alone}. \tag{R2$\mu$-B2.5}
\]

\paragraph{Router pathway.} Decomposing $\mathbf{v}=\mathbf{v}^{(0)}+\mathbf{v}^{(\Delta)}$ where $\mathbf{v}^{(0)}_i=\dot\phi_{i,1}\langle h_1^{3,i}, W_0^4\rangle\in\Theta(1)$ (coherent via the alignment in $h_1^{3,i}$ from (R2$\mu$-F2.5c) / (R2$\mu$-B2.4)) and $\mathbf{v}^{(\Delta)}_i\in\Theta(1/N)$ (via (R2$\mu$-U1.4)):
\begin{align*}
    (1/M)Q_0^\top\mathbf{v}^{(0)} &\in \Theta(1/N) \text{ random direction} \tag{R2$\mu$-B2.6a}\\
    & \quad (Q_0\text{ indep.\ of }\mathbf{v}^{(0)};\ \sigma_Q^2=1/N;\ \|\mathbf{v}^{(0)}\|^2\in\Theta(M))
\end{align*}
\[
    (1/M)Q_0^\top\mathbf{v}^{(\Delta)} \in \Theta(1/N^2). \tag{R2$\mu$-B2.6b}
\]
\begin{align*}
    (1/M)\Delta_1 Q^\top\mathbf{v}^{(0)} &\in \Theta(1/N^2) \text{ along } h_0^1 \tag{R2$\mu$-B2.6c}\\
    & \quad (\Delta_1 Q\text{ rank-1 along }(h_0^1)^\top\text{ via (R2$\mu$-U1.Q)};\\
    & \qquad \langle w_0, \mathbf{v}^{(0)}\rangle = (\dot\phi^2/M)\sum_i[r_i c_i + r_i^2]\in\Theta(1/N)\\
    & \qquad \text{(both the cross-i CLT piece }\sum r_i c_i\sim 1\text{ and the LLN self-pairing}\\
    & \qquad \sum r_i^2\sim M/N\text{ contribute }\Theta(1)\text{ before the }(1/M)\text{ prefactor)};\\
    & \qquad \Delta_1 Q^\top \mathbf{v}^{(0)} = h_0^1\eta_Q\chi_0\langle w_0, \mathbf{v}^{(0)}\rangle\sim 1/N\text{ entries};\\
    & \qquad \text{after }(1/M)\text{: }\Theta(1/N^2)\text{ — sub-leading vs (R2$\mu$-B2.6a)})
\end{align*}
\[
    (1/M)\Delta_1 Q^\top\mathbf{v}^{(\Delta)} \in \Theta(1/N^3). \tag{R2$\mu$-B2.6d}
\]
\[
    (\partial f_1/\partial h_1^1)_\mathrm{router} \in \Theta(1/N), \text{ from (R2$\mu$-B2.6a) alone (other three pieces sub-leading)}. \tag{R2$\mu$-B2.6}
\]
\textit{W$^{3,i}$-imbalance propagates into the router pathway.} The W$^{3,i}$-split of (R2$\mu$-B2.4), $\mathbf{v}=\mathbf{v}^I+\mathbf{v}^U$ with $\mathbf{v}^I_i=\dot\phi_{i,1}\langle (W_0^{3,i})h_1^{2,i}, W_1^4\rangle$ and $\mathbf{v}^U_i=\dot\phi_{i,1}\langle (\Delta_1 W^{3,i})h_1^{2,i}, W_1^4\rangle$, transmits as: $\mathbf{v}^I_i\in\Theta(1/\sqrt N)$ random across $i$ (from $\langle h_0^{3,i}, W_0^4\rangle$), giving $\|\mathbf{v}^I\|^2\in\Theta(1)$ and $(1/M)Q_t^\top\mathbf{v}^I\in\Theta(1/N^{3/2})$ via cross-layer CLT; whereas $\mathbf{v}^U_i\in\Theta(1)$ coherent across $i$ (rank-1 alignment with $(W_0^4)^\top$), giving $\|\mathbf{v}^U\|^2\in\Theta(M)$ and $(1/M)Q_t^\top\mathbf{v}^U\in\Theta(1/N)$. The same $\sqrt N$ deficit between $W_0^{3,i}$- and $\Delta W^{3,i}$-pieces is transmitted into $\partial f_1/\partial h_1^1$ via the router pathway.

\[
    \partial f_1/\partial h_1^1 = (\partial f_1/\partial h_1^1)_\mathrm{exp} + (\partial f_1/\partial h_1^1)_\mathrm{router} \in \Theta(1/N). \tag{R2$\mu$-B2.7}
\]

\paragraph{Step-2 parameter updates.}
\[
    \Delta_2 W^4 = -(\chi_1/N)(h_1^3)^\top \in \Theta(1/N). \quad (h_1^3\in\Theta(1)\text{ via (R2$\mu$-F2.6)};\ \text{now comparable to }W_0^4) \tag{R2$\mu$-U2.4}
\]
\begin{align*}
    \Delta_2 W^{3,i} &= -\eta N\chi_1\phi_{i,1}(W_1^4)^\top(h_1^{2,i})^\top \in \Theta(1) \text{ rank-1 along } (W_1^4)^\top\otimes h_1^{2,i} \tag{R2$\mu$-U2.3}\\
    & \quad (\text{same structure as (R2$\mu$-U1.3)};\\
    & \qquad (W_1^4)^\top\in\Theta(1/N);\ h_1^{2,i}\in\Theta(1)\text{ via (R2$\mu$-F2.4)})
\end{align*}
\begin{align*}
    \Delta_2 W^{2,i} &= -(\eta\chi_1\phi_{i,1}/N)(W_1^{3,i})^\top(W_1^4)^\top(h_1^1)^\top \in \Theta(1/N) \text{ rank-1 along } h_0^{2,i}\otimes h_1^1 \tag{R2$\mu$-U2.2}\\
    & \quad ((W_1^{3,i})^\top(W_1^4)^\top\in\Theta(1)\text{ along }h_0^{2,i}\text{ via new intermediate at }t=1;\\
    & \qquad \sqrt N\text{ larger than }\Delta_1 W^{2,i}\in\Theta(1/N^{3/2}))
\end{align*}
\begin{align*}
    \Delta_2 W^1 &= -\eta_1\chi_1(\partial f_1/\partial h_1^1)\,x^\top \in \Theta(1) \tag{R2$\mu$-U2.1a}\\
    & \quad (\eta_1=\eta N;\ \partial f_1/\partial h_1^1\in\Theta(1/N)\text{ via (R2$\mu$-B2.7)})
\end{align*}
\[
    \Delta_2 h^1 = \Delta_2 W^1\,x \in \Theta(1) \text{ aligned along } h_0^1. \quad (\text{embedding feature-learns at }t=2) \tag{R2$\mu$-U2.1b}
\]
\[
    \Delta_2 Q \in \Theta(1/N). \quad (\partial f_1/\partial\phi\in\Theta(1/N)\text{ via (R2$\mu$-B2.4)}) \tag{R2$\mu$-U2.Q}
\]

\paragraph{Third forward pass}
\label{ssec:mup_r2_fwd3}

We compute activations at $\theta^{(2)}$. The cumulative embedding change $\Delta h^1:=h_2^1-h_0^1 = \Delta_1 h^1 + \Delta_2 h^1$, with $\Delta_1 h^1\in\Theta(1/\sqrt N)$ via (R2$\mu$-U1.1b) and $\Delta_2 h^1\in\Theta(1)$ aligned along $h_0^1$ via (R2$\mu$-U2.1b); the latter dominates.

\[
    h_2^1 = h_0^1 + \Delta h^1 \in \Theta(1). \tag{R2$\mu$-F3.1}
\]

\noindent (R2$\mu$-F3.2) $h_2^{2,i} = W_2^{2,i}h_2^1$, four-piece decomposition with cumulative $\Delta W^{2,i}$:
\[
    \text{init: } W_0^{2,i}h_0^1 = h_0^{2,i} \in \Theta(1). \quad (\text{R2$\mu$-F1.4}) \tag{R2$\mu$-F3.2a}
\]
\begin{align*}
    \text{prop: } W_0^{2,i}\Delta h^1 &\in \Theta(1) \tag{R2$\mu$-F3.2b}\\
    & \quad (\sigma_2^2\|\Delta h^1\|^2 = (1/N)\,\Theta(N) = \Theta(1);\\
    & \qquad \|\Delta h^1\|^2\in\Theta(N)\text{ via }\Delta_2 h^1\in\Theta(1)\text{ via (R2$\mu$-U2.1b)})
\end{align*}
\begin{align*}
    \text{eff: } \Delta_2 W^{2,i}\,h_0^1 &= -(\eta\chi_1\phi_{i,1}/N)(W_1^{3,i})^\top(W_1^4)^\top\,\langle h_1^1, h_0^1\rangle \in \Theta(1) \text{ along } h_0^{2,i} \tag{R2$\mu$-F3.2c}\\
    & \quad (\text{(R2$\mu$-U2.2)substitution};\\
    & \qquad (W_1^{3,i})^\top(W_1^4)^\top\in\Theta(1)\text{ along }h_0^{2,i}\text{ via new intermediate at }t=1;\\
    & \qquad \langle h_1^1, h_0^1\rangle\in\Theta(N)\text{ coherent})
\end{align*}
\begin{align*}
    \text{cross: } \Delta_2 W^{2,i}\,\Delta h^1 &= -(\eta\chi_1\phi_{i,1}/N)(W_1^{3,i})^\top(W_1^4)^\top\,\langle h_1^1,\Delta h^1\rangle \in \Theta(1) \text{ along } h_0^{2,i} \tag{R2$\mu$-F3.2d}\\
    & \quad (\text{(R2$\mu$-U2.2)substitution};\\
    & \qquad (W_1^{3,i})^\top(W_1^4)^\top\in\Theta(1)\text{ along }h_0^{2,i}\text{ via new intermediate at }t=1;\\
    & \qquad \langle h_1^1,\Delta h^1\rangle\in\Theta(N)\text{ coherent via }\Delta h^1\parallel h_0^1\text{ from (R2$\mu$-U2.1b)};\\
    & \qquad \text{leading-scale contribution})
\end{align*}
$\Delta h^{2,i}\in\Theta(1)$ entry-wise, with the coherent $\Theta(1)$-along-$h_0^{2,i}$ piece emerging from (R2$\mu$-F3.2c) + (R2$\mu$-F3.2d).

\noindent (R2$\mu$-F3.3) $h_2^{3,i} = W_2^{3,i}h_2^{2,i}$, four-piece decomposition with cumulative $\Delta W^{3,i}$:
\[
    \text{init: } h_0^{3,i} \in \Theta(1). \quad (\text{R2$\mu$-F1.5}) \tag{R2$\mu$-F3.3a}
\]
\begin{align*}
    \text{prop: } W_0^{3,i}\Delta h^{2,i} &\in \Theta(1) \tag{R2$\mu$-F3.3b}\\
    & \quad (\sigma_3^2\|\Delta h^{2,i}\|^2 = \Theta(1)\cdot\Theta(1) = \Theta(1);\\
    & \qquad \|\Delta h^{2,i}\|^2\in\Theta(1)\text{ via (R2$\mu$-F3.2)};\ N_e=\Theta(1))
\end{align*}
\begin{align*}
    \text{eff: } \Delta_2 W^{3,i}\,h_0^{2,i} &= -\eta N\chi_1\phi_{i,1}\,\|h_0^{2,i}\|^2\,(W_1^4)^\top \in \Theta(1) \text{ along }(W_0^4)^\top \tag{R2$\mu$-F3.3c}\\
    & \quad (\text{(R2$\mu$-U2.3)substitution};\ \|h_0^{2,i}\|^2\in\Theta(1)\text{ via (R2$\mu$-F1.4)};\\
    & \qquad (W_1^4)^\top\in\Theta(1/N)\text{ dominated by }(W_0^4)^\top)
\end{align*}
\[
    \text{cross: } \Delta_2 W^{3,i}\,\Delta h^{2,i} \text{ tracked as part of } D' \text{ in (R2$\mu$-F3.4)}. \tag{R2$\mu$-F3.3d}
\]
$\Delta h^{3,i}\in\Theta(1)$ aligned along $(W_0^4)^\top$.

\noindent (R2$\mu$-F3.4) $h_2^3 = A_1' + A_{2,1}' + A_{2,2}' + A_3' + D'$, where
\begin{align*}
    A_1' &:= \tfrac{1}{M}\textstyle\sum_i\phi_{i,2}\,h_0^{3,i}, &
    A_{2,1}' &:= \tfrac{1}{M}\textstyle\sum_i\phi_{i,2}\,W_0^{3,i}W_0^{2,i}\,\Delta h^1,\\
    A_{2,2}' &:= \tfrac{1}{M}\textstyle\sum_i\phi_{i,2}\,W_0^{3,i}\,\Delta_2 W^{2,i}\,h_0^1, &
    A_3' &:= \tfrac{1}{M}\textstyle\sum_i\phi_{i,2}\,\Delta_2 W^{3,i}\,h_0^{2,i},\\
    D' &:= \tfrac{1}{M}\textstyle\sum_i\phi_{i,2}\,\Delta_2 W^{3,i}\,\Delta h^{2,i}.
\end{align*}
\[
    A_1' \in \Theta(1/\sqrt N). \quad (\text{cross-$i$ CLT on }\Theta(1)\text{-entry independent vectors;\ same as (R2$\mu$-F2.6a)}) \tag{R2$\mu$-F3.4a}
\]
\begin{align*}
    A_{2,1}' &\in \Theta(1/\sqrt N) \tag{R2$\mu$-F3.4b}\\
    & \quad (\text{climbs from }\Theta(1/N)\text{ at }t=1\text{ because }\|\Delta h^1\|^2\\
    & \qquad \text{grows from }\Theta(1)\text{ to }\Theta(N)\text{ via (R2$\mu$-U2.1b)})
\end{align*}
\begin{align*}
    A_{2,2}' &\in \Theta(1/\sqrt N) \text{ in mostly-random directions} \tag{R2$\mu$-F3.4c}\\
    & \quad (\text{same mechanism as }A_{2,1}';\ \|\Delta h^1\|^2\in\Theta(N)\text{ via (R2$\mu$-U2.1b)};\\
    & \qquad \text{rank-deficient initialization Gram (\S\ref{ssec:mup_r2_lemmas})}\\
    & \qquad \text{still does not collapse, so sub-leading})
\end{align*}
\begin{align*}
    A_3' &= -\eta N\chi_1\,(W_0^4)^\top\,\Bigl(\tfrac{1}{M}\textstyle\sum_i\phi_{i,2}\phi_{i,1}\,\|h_0^{2,i}\|^2\Bigr) \in \Theta(1) \text{ along }(W_0^4)^\top \tag{R2$\mu$-F3.4d}\\
    & \quad (\text{same structure as (R2$\mu$-F2.6d) with cumulative }\Delta_2 W^{3,i})
\end{align*}
\begin{align*}
    D' &= -\eta N\chi_1\,(W_0^4)^\top\,\tfrac{1}{M}\textstyle\sum_i\phi_{i,2}\phi_{i,1}\,\langle h_0^{2,i},\Delta h^{2,i}\rangle\\
    &\in \Theta(1) \text{ along }(W_0^4)^\top \tag{R2$\mu$-F3.4e}\\
    & \quad (\textbf{climbs from }\Theta(1/N)\text{ at }t=1\text{ to }\Theta(1)\text{ at }t=2;\\
    & \qquad \langle h_0^{2,i},\Delta h^{2,i}\rangle\in\Theta(1)\text{ coherent across }i\text{ via (R2$\mu$-F3.2c)+(R2$\mu$-F3.2d)},\\
    & \qquad \text{replacing the }t=1\text{ random factor }\langle h_0^{3,i},W_0^4\rangle\in\Theta(1/\sqrt N)\text{ via (R2$\mu$-B1.4)};\\
    & \qquad \text{LLN gives empirical average }\Theta(1)\text{ rather than }\Theta(1/\sqrt M))
\end{align*}
\[
    h_2^3 \in \Theta(1) \text{ along }(W_0^4)^\top, \text{ from } A_3' + D'. \tag{R2$\mu$-F3.4}
\]
\begin{align*}
    f_2 &= W_2^4\,h_2^3 \in \Theta(1). \tag{R2$\mu$-F3.5}\\
    &\quad (A_3'+D'\text{ aligned along }(W_0^4)^\top;\ \|W_0^4\|^2=1/N;\ \text{net }\eta N\cdot\Theta(1)\cdot(1/N)=\Theta(1))
\end{align*}

\paragraph{Third backward pass}
\label{ssec:mup_r2_bwd3}

The structure parallels (R2$\mu$-B2.1)--(R2$\mu$-B2.7) with cumulative $\Delta W$'s. The key change at $t=2$ vs $t=1$:
\begin{align*}
    (\Delta W^4)^\top &\in \Theta(1/N),\quad \text{climbs from }\Theta(1/N^{3/2})\text{ at }t=1 \tag{R2$\mu$-B3.1}\\
    & \quad (h_1^3\in\Theta(1)\text{ via (R2$\mu$-F2.6)};\ (\Delta_2 W^4)^\top\in\Theta(1/N)\text{ via (R2$\mu$-U2.4)};\\
    & \qquad \text{now comparable to }(W_0^4)^\top\in\Theta(1/N))
\end{align*}

Combined with the alignment mechanism (driven by $\langle W_1^4, h_1^{3,i}\rangle\in\Theta(1)$ coherent across $i$ via (R2$\mu$-B2.4)), several pieces climb at $t=2$:

\paragraph{Router pathway $\partial f_2/\partial \phi_{i,2}$.}
\begin{align*}
    \partial f_2/\partial \phi_{i,2} &= (1/M)\langle h_2^{3,i}, W_2^4\rangle \in \Theta(1/N), \tag{R2$\mu$-B3.4}\\
    & \quad \text{four-piece grid (same decomposition as (R2$\mu$-B2.4) at }t=2):\\
    & \quad (\text{init}\cdot\text{init})\ \langle (W_0^{3,i})\,h_2^{2,i}, W_0^4\rangle\in\Theta(1/\sqrt N)\text{ random, unchanged from }t=1;\\
    & \quad (\text{init}\cdot\text{update})\ \langle (W_0^{3,i})\,h_2^{2,i}, \Delta_2 W^4\rangle\in\Theta(1/\sqrt N),\ \text{climbs from }\Theta(1/N)\text{ at }t=1\\
    & \qquad \text{via }\langle h_0^{3,i}, A_3\rangle\in\Theta(\sqrt N)\text{ (}A_3\parallel(W_0^4)^\top\text{ from R2$\mu$-F2.6d)}\\
    & \qquad \text{followed by the }1/N\text{ prefactor of }\Delta_2 W^4;\\
    & \quad (\text{update}\cdot\text{init})\ \langle (\Delta_2 W^{3,i})\,h_2^{2,i}, W_0^4\rangle\in\Theta(1)\text{ via rank-1 alignment, unchanged};\\
    & \quad (\text{update}\cdot\text{update})\ \langle (\Delta_2 W^{3,i})\,h_2^{2,i}, \Delta_2 W^4\rangle\in\Theta(1),\ \text{climbs from }\Theta(1/N)\text{ at }t=1\\
    & \qquad \text{via }\langle W_0^4, h_1^3\rangle\in\Theta(1)\text{ from }A_3\\
    & \qquad \text{(}\kappa\in\Theta(N)\text{ rank-1 prefactor times }\langle W_0^4, \Delta_2 W^4\rangle\in\Theta(1/N));\\
    & \quad /M\text{: }(\Theta(1/N^{3/2}),\Theta(1/N^{3/2}),\Theta(1/N),\Theta(1/N))\text{ respectively};\\
    & \quad \text{both update}\cdot*\text{ pieces dominate; total }\Theta(1/N);\\
    & \quad \text{the entire }W_0^{3,i}\text{ row stays at }\Theta(1/N^{3/2})\text{: }\sqrt N\text{ deficit preserved}.
\end{align*}

\paragraph{Climbing pieces in expert pathway $(\partial f_2/\partial h_2^1)_\mathrm{exp}$.}
\begin{align*}
    A_{4.2} &\in \Theta(1/N^{3/2}),\quad \text{climbs from }\Theta(1/N^2)\text{ at }t=1 \tag{R2$\mu$-B3.5a.2}\\
    & \quad (\sqrt N\text{ lift via (R2$\mu$-B3.1)};\ \text{still sub-leading vs }A_{4.1})
\end{align*}
\begin{align*}
    A_{5.2} &\in \Theta(1/N),\quad \text{climbs from }\Theta(1/N^2)\text{ at }t=1 \tag{R2$\mu$-B3.5b.2}\\
    & \quad (\text{combined }\sqrt N\text{ lift via (R2$\mu$-B3.1) and alignment-driven mechanism};\\
    & \qquad \langle W_1^4, h_1^{3,i}\rangle\in\Theta(1)\text{ coherent via alignment of }h_1^{3,i}\text{ along }(W_0^4)^\top\\
    & \qquad \text{from (R2$\mu$-F2.5c)};\ \text{joins }A_{5.1}\text{ at leading scale})
\end{align*}
\begin{align*}
    A_{6.1} &\in \Theta(1/N^2)\text{ at every step};\quad A_{6.2}\in\Theta(1/N^3)\text{ at }t=1,\ \Theta(1/N^2)\text{ at }t=2 \tag{R2$\mu$-B3.5c}\\
    & \quad (A_{6.1}: S_i := W_0^4\cdot M_W\cdot(W_0^4)^\top\text{ has trace-induced coherent piece }\\
    & \qquad \sigma_4^2\cdot\operatorname{tr}(M_W)=\Theta(1/N)\text{ at every step, LLN preserves }\Theta(1/N^2);\\
    & \qquad A_{6.2}\text{ at }t=1: S_i\text{ random (}\Delta_1 W^4\text{ indep of }W_0^4\text{), cross-$i$ CLT gives }1/\sqrt N\\
    & \qquad \text{combined with }\Delta_1 W^4\text{ smaller by }1/\sqrt N\text{ via (R2$\mu$-U1.4)}\Rightarrow\Theta(1/N^3);\\
    & \qquad A_{6.2}\text{ at }t=2: \Delta W^4\text{ acquires alignment with }W_0^4\text{ via }h_1^3\\
    & \qquad \text{(FL piece in (R2$\mu$-F2.5c)), restoring trace-induced coherent piece in }S_i;\\
    & \qquad \text{combined with }\Delta W^4\text{ climb to }\Theta(1/N)\text{ via (R2$\mu$-B3.1) gives }\Theta(1/N^2);\\
    & \qquad \text{both stay sub-leading vs }A_{5.x}, E_x\in\Theta(1/N))
\end{align*}
\begin{align*}
    E_1 &\in \Theta(1/N),\quad \text{climbs from }\Theta(1/N^2)\text{ at }t=1 \tag{R2$\mu$-B3.5d.1}\\
    & \quad (\text{alignment-driven climb};\\
    & \qquad \text{inner product structure inside }E\text{ no longer CLT-cancels but LLN-accumulates}\\
    & \qquad \text{once }\langle W_1^4, h_1^{3,i}\rangle\in\Theta(1)\text{ is coherent via (R2$\mu$-B2.4)};\\
    & \qquad \text{joins leading scale})
\end{align*}
\begin{align*}
    E_2 &\in \Theta(1/N),\quad \text{climbs from }\Theta(1/N^{5/2})\text{ at }t=1 \tag{R2$\mu$-B3.5d.2}\\
    & \quad (\text{combined alignment-driven climb and }\sqrt N\text{ lift via (R2$\mu$-B3.1)};\\
    & \qquad \text{joins leading scale})
\end{align*}
\[
    (\partial f_2/\partial h_2^1)_\mathrm{exp} \in \Theta(1/N), \text{ from }A_{5.1}, A_{5.2}, E_1, E_2. \tag{R2$\mu$-B3.5}
\]

\paragraph{Router pathway.}
\[
    (\partial f_2/\partial h_2^1)_\mathrm{router} \in \Theta(1/N), \text{ all four pieces at leading scale}. \tag{R2$\mu$-B3.6}
\]
The $\mathbf{v}^{(\Delta)}$-pieces lift via (R2$\mu$-B3.1); the $\Delta_t Q^\top\mathbf{v}^{(0)}$ piece remains coherent via the rank-1 alignment of $\Delta_t Q$ with $\mathbf{v}^{(0)}$ (extending (R2$\mu$-B2.6c) to cumulative $\Delta_t Q$).

\textit{W$^{3,i}$-imbalance preserved at $t=2$.} Define $v_i := \dot\phi_{i,2}\langle h_2^{3,i}, W_2^4\rangle$ and split $v_i = v^I_i + v^U_i$ where
\begin{align*}
    v^I_i &:= \dot\phi_{i,2}\langle (W_0^{3,i})h_2^{2,i}, W_2^4\rangle,\\
    v^U_i &:= \dot\phi_{i,2}\langle (\Delta_2 W^{3,i})h_2^{2,i}, W_2^4\rangle.
\end{align*}
$v^I_i\in\Theta(1/\sqrt N)$ random across $i$ (both $(W_0^{3,i}, W_0^4)$ and $(W_0^{3,i}, \Delta_2 W^4)$ entries are $\Theta(1/\sqrt N)$, the latter via $\langle h_0^{3,i}, A_3\rangle\in\Theta(\sqrt N)$ followed by the $1/N$ prefactor of $\Delta_2 W^4$). $v^U_i\in\Theta(1)$ coherent. Cross-layer CLT then gives $(1/M)Q_2^\top v^I\in\Theta(1/N^{3/2})$ (sub-leading) and $(1/M)Q_2^\top v^U\in\Theta(1/N)$ (leading). The entire $W_0^{3,i}$-row of the router pathway sits a factor $\sqrt N$ below the $\Delta_2 W^{3,i}$-row.

\[
    \partial f_2/\partial h_2^1 \in \Theta(1/N), \tag{R2$\mu$-B3.7}
\]
with leading-scale contributions from $A_{5.1}, A_{5.2}, E_1, E_2$ in the expert pathway and all four pieces in the router pathway.

\subsubsection{Summary tables of signal propagation for $\mu$P in Regime II}
\label{ssec:mup_r2_summary}

Notation: $\Delta_t W^\ell$ denotes the cumulative update $W_t^\ell - W_0^\ell$. Green = at the proper feature-learning scale at $t=2$; red = sub-leading throughout.

\paragraph{Forward.}
\begin{center}
\renewcommand{\arraystretch}{1.2}
\begin{tabular}{l|l|l|l}
\textbf{Quantity} & $t=0$ & $t=1$ & $t=2$\\\hline
$h_t^1 = h_0^1 + \Delta_t h^1$                                                            & $\Theta(1)$         & $\Theta(1)$           & \textcolor{green!50!black}{$\Theta(1)$}\\
$\quad$ init: $h_0^1 = W_0^1 x$                                                           & $\Theta(1)$         & $\Theta(1)$           & \textcolor{green!50!black}{$\Theta(1)$}\\
$\quad$ effective: $\Delta_t h^1 = \Delta_t W^1\,x$                                       & $0$                 & $\Theta(1/\sqrt N)$   & \textcolor{green!50!black}{$\Theta(1)$}\\
$\psi_t,\,\phi_t$                                                                         & $\Theta(1)$         & $\Theta(1)$           & \textcolor{green!50!black}{$\Theta(1)$}\\
$h_t^{2,i}$                                                                               & $\Theta(1)$         & $\Theta(1)$           & \textcolor{green!50!black}{$\Theta(1)$}\\
$\quad$ init: $h_0^{2,i} = W_0^{2,i}h_0^1$                                                & $\Theta(1)$         & $\Theta(1)$           & \textcolor{green!50!black}{$\Theta(1)$}\\
$\quad$ propagating: $W_0^{2,i}\Delta_t h^1$                                              & $0$                 & $\Theta(1/\sqrt N)$   & \textcolor{green!50!black}{$\Theta(1)$}\\
$\quad$ effective: $\Delta_t W^{2,i}\,h_0^1$                                              & $0$                 & $\Theta(1/\sqrt N)$   & \textcolor{green!50!black}{$\Theta(1)$}\\
$\quad$ cross: $\Delta_t W^{2,i}\,\Delta_t h^1$                                           & $0$                 & $\Theta(1/N^{3/2})$   & \textcolor{green!50!black}{$\Theta(1)$}\\
$h_t^{3,i}$                                                                               & $\Theta(1)$         & $\Theta(1)$           & \textcolor{green!50!black}{$\Theta(1)$}\\
$\quad$ init: $h_0^{3,i} = W_0^{3,i}h_0^{2,i}$                                            & $\Theta(1)$         & $\Theta(1)$           & \textcolor{green!50!black}{$\Theta(1)$}\\
$\quad$ propagating: $W_0^{3,i}\Delta_t h^{2,i}$                                          & $0$                 & $\Theta(1/\sqrt N)$   & \textcolor{green!50!black}{$\Theta(1)$}\\
$\quad$ effective: $\Delta_t W^{3,i}\,h_0^{2,i}$                                          & $0$                 & $\Theta(1)$         & \textcolor{green!50!black}{$\Theta(1)$}\\
$\quad$ cross: $\Delta_t W^{3,i}\,\Delta_t h^{2,i}$                                       & $0$                 & $\Theta(1/\sqrt N)$ & \textcolor{green!50!black}{$\Theta(1)$}\\
$h_t^3 = A_1+A_{2,1}+A_{2,2}+A_3+D$                                                       & $\Theta(1/\sqrt N)$ & $\Theta(1)$           & \textcolor{green!50!black}{$\Theta(1)$}\\
$\quad A_1 = (1/M)\sum_i\phi_{i,t}\,h_0^{3,i}$                                            & $\Theta(1/\sqrt N)$ & $\Theta(1/\sqrt N)$   & \textcolor{red!70!black}{$\Theta(1/\sqrt N)$}\\
$\quad A_{2,1} = (1/M)\sum_i\phi_{i,t}\,W_0^{3,i}W_0^{2,i}\,\Delta_t h^1$                 & $0$                 & $\Theta(1/N)$         & \textcolor{red!70!black}{$\Theta(1/\sqrt N)$}\\
$\quad A_{2,2} = (1/M)\sum_i\phi_{i,t}\,W_0^{3,i}\,\Delta_t W^{2,i}\,h_0^1$               & $0$                 & $\Theta(1/N)$         & \textcolor{red!70!black}{$\Theta(1/\sqrt N)$}\\
$\quad A_3 = (1/M)\sum_i\phi_{i,t}\,\Delta_t W^{3,i}\,h_0^{2,i}$                          & $0$                 & $\Theta(1)$         & \textcolor{green!50!black}{$\Theta(1)$}\\
$\quad D = (1/M)\sum_i\phi_{i,t}\,\Delta_t W^{3,i}\,\Delta_t h^{2,i}$                     & $0$                 & $\Theta(1/N)$         & \textcolor{green!50!black}{$\Theta(1)$}\\
$f_t = W_t^4\,h_t^3$                                                                      & $\Theta(1/N)$       & $\Theta(1)$           & \textcolor{green!50!black}{$\Theta(1)$}\\
\end{tabular}
\end{center}

\paragraph{Backward.}
\begin{center}
\renewcommand{\arraystretch}{1.2}
\adjustbox{max width=\textwidth}{\begin{tabular}{l|l|l|l}
\textbf{Quantity} & $t=0$ & $t=1$ & $t=2$\\\hline
$\partial f_t/\partial h_t^3 = (W_0^4)^\top + (\Delta_t W^4)^\top$                                                              & $\Theta(1/N)$       & $\Theta(1/N)$           & \textcolor{green!50!black}{$\Theta(1/N)$}\\
$\quad$ init: $(W_0^4)^\top$                                                                                                    & $\Theta(1/N)$       & $\Theta(1/N)$           & \textcolor{green!50!black}{$\Theta(1/N)$}\\
$\quad$ update: $(\Delta_t W^4)^\top$                                                                                           & $0$                 & $\Theta(1/N^{3/2})$     & \textcolor{green!50!black}{$\Theta(1/N)$}\\
$\partial f_t/\partial h_t^{3,i} = (\phi_{i,t}/M)(W_t^4)^\top$                                                                  & $\Theta(1/N^2)$     & $\Theta(1/N^2)$         & \textcolor{green!50!black}{$\Theta(1/N^2)$}\\
$\quad$ init: $(\phi_{i,t}/M)(W_0^4)^\top$                                                                                      & $\Theta(1/N^2)$     & $\Theta(1/N^2)$         & \textcolor{green!50!black}{$\Theta(1/N^2)$}\\
$\quad$ update: $(\phi_{i,t}/M)(\Delta_t W^4)^\top$                                                                             & $0$                 & $\Theta(1/N^{5/2})$     & \textcolor{green!50!black}{$\Theta(1/N^2)$}\\
$\partial f_t/\partial h_t^{2,i} = (\phi_{i,t}/M)(W_t^{3,i})^\top(W_t^4)^\top$                                                  & $\Theta(1/N^{3/2})$ & $\Theta(1/N)$ & \textcolor{green!50!black}{$\Theta(1/N)$}\\
$\quad$ init$\cdot$init: $(\phi_{i,t}/M)(W_0^{3,i})^\top(W_0^4)^\top$                                                           & $\Theta(1/N^{3/2})$ & $\Theta(1/N^{3/2})$     & \textcolor{red!70!black}{$\Theta(1/N^{3/2})$}\\
$\quad$ init$\cdot$update: $(\phi_{i,t}/M)(W_0^{3,i})^\top(\Delta_t W^4)^\top$                                                  & $0$                 & $\Theta(1/N^2)$         & \textcolor{red!70!black}{$\Theta(1/N^{3/2})$}\\
$\quad$ update$\cdot$init: $(\phi_{i,t}/M)(\Delta_t W^{3,i})^\top(W_0^4)^\top$                                                  & $0$                 & $\Theta(1/N)$ & \textcolor{green!50!black}{$\Theta(1/N)$}\\
$\quad$ update$\cdot$update: $(\phi_{i,t}/M)(\Delta_t W^{3,i})^\top(\Delta_t W^4)^\top$                                         & $0$                 & $\Theta(1/N^2)$         & \textcolor{green!50!black}{$\Theta(1/N)$}\\
$\partial f_t/\partial \phi_{i,t} = (1/M)\langle h_t^{3,i}, W_t^4\rangle$                                                       & $\Theta(1/N^{3/2})$ & $\Theta(1/N)$           & \textcolor{green!50!black}{$\Theta(1/N)$}\\
$\quad$ init$\cdot$init: $(1/M)\langle (W_0^{3,i})\,h_t^{2,i}, W_0^4\rangle$                                                    & $\Theta(1/N^{3/2})$ & $\Theta(1/N^{3/2})$     & \textcolor{red!70!black}{$\Theta(1/N^{3/2})$}\\
$\quad$ init$\cdot$update: $(1/M)\langle (W_0^{3,i})\,h_t^{2,i}, \Delta_t W^4\rangle$                                           & $0$                 & $\Theta(1/N^2)$         & \textcolor{red!70!black}{$\Theta(1/N^{3/2})$}\\
$\quad$ update$\cdot$init: $(1/M)\langle (\Delta_t W^{3,i})\,h_t^{2,i}, W_0^4\rangle$                                           & $0$                 & $\Theta(1/N)$           & \textcolor{green!50!black}{$\Theta(1/N)$}\\
$\quad$ update$\cdot$update: $(1/M)\langle (\Delta_t W^{3,i})\,h_t^{2,i}, \Delta_t W^4\rangle$                                  & $0$                 & $\Theta(1/N^2)$         & \textcolor{green!50!black}{$\Theta(1/N)$}\\
$(\partial f_t/\partial h_t^1)_\mathrm{exp}$                                                                                    & $\Theta(1/N^{3/2})$ & $\Theta(1/N)$           & \textcolor{green!50!black}{$\Theta(1/N)$}\\
$\quad A_{4.1} = (1/M)\sum_i\phi_{i,t}\,(W_0^{2,i})^\top(W_0^{3,i})^\top(W_0^4)^\top$                                           & $\Theta(1/N^{3/2})$ & $\Theta(1/N^{3/2})$     & \textcolor{red!70!black}{$\Theta(1/N^{3/2})$}\\
$\quad A_{4.2} = (1/M)\sum_i\phi_{i,t}\,(W_0^{2,i})^\top(W_0^{3,i})^\top(\Delta_t W^4)^\top$                                    & $0$                 & $\Theta(1/N^2)$         & \textcolor{red!70!black}{$\Theta(1/N^{3/2})$}\\
$\quad A_{5.1} = (1/M)\sum_i\phi_{i,t}\,(W_0^{2,i})^\top(\Delta_t W^{3,i})^\top(W_0^4)^\top$                                    & $0$                 & $\Theta(1/N)$           & \textcolor{green!50!black}{$\Theta(1/N)$}\\
$\quad A_{5.2} = (1/M)\sum_i\phi_{i,t}\,(W_0^{2,i})^\top(\Delta_t W^{3,i})^\top(\Delta_t W^4)^\top$                             & $0$                 & $\Theta(1/N^2)$         & \textcolor{green!50!black}{$\Theta(1/N)$}\\
$\quad A_{6.1} = (1/M)\sum_i\phi_{i,t}\,(\Delta_t W^{2,i})^\top(W_0^{3,i})^\top(W_0^4)^\top$                                    & $0$                 & $\Theta(1/N^2)$         & \textcolor{red!70!black}{$\Theta(1/N^2)$}\\
$\quad A_{6.2} = (1/M)\sum_i\phi_{i,t}\,(\Delta_t W^{2,i})^\top(W_0^{3,i})^\top(\Delta_t W^4)^\top$                             & $0$                 & $\Theta(1/N^3)$         & \textcolor{red!70!black}{$\Theta(1/N^2)$}\\
$\quad E_1 = (1/M)\sum_i\phi_{i,t}\,(\Delta_t W^{2,i})^\top(\Delta_t W^{3,i})^\top(W_0^4)^\top$                                 & $0$                 & $\Theta(1/N^2)$         & \textcolor{green!50!black}{$\Theta(1/N)$}\\
$\quad E_2 = (1/M)\sum_i\phi_{i,t}\,(\Delta_t W^{2,i})^\top(\Delta_t W^{3,i})^\top(\Delta_t W^4)^\top$                          & $0$                 & $\Theta(1/N^{5/2})$     & \textcolor{green!50!black}{$\Theta(1/N)$}\\
$(\partial f_t/\partial h_t^1)_\mathrm{router} = (1/M)Q_t^\top[\dot\phi_{i,t}\langle h_t^{3,i}, W_t^4\rangle]_i$                  & $\Theta(1/N^{3/2})$ & $\Theta(1/N)$           & \textcolor{green!50!black}{$\Theta(1/N)$}\\
$\quad (1/M)Q_0^\top[\dot\phi_{i,t}\langle (W_0^{3,i})h_t^{2,i}, W_0^4\rangle]_i$                                               & $\Theta(1/N^{3/2})$ & $\Theta(1/N^{3/2})$     & \textcolor{red!70!black}{$\Theta(1/N^{3/2})$}\\
$\quad (1/M)Q_0^\top[\dot\phi_{i,t}\langle (W_0^{3,i})h_t^{2,i}, \Delta_t W^4\rangle]_i$                                        & $0$                 & $\Theta(1/N^2)$         & \textcolor{red!70!black}{$\Theta(1/N^{3/2})$}\\
$\quad (1/M)Q_0^\top[\dot\phi_{i,t}\langle (\Delta_t W^{3,i})h_t^{2,i}, W_0^4\rangle]_i$                                        & $0$                 & $\Theta(1/N)$           & \textcolor{green!50!black}{$\Theta(1/N)$}\\
$\quad (1/M)Q_0^\top[\dot\phi_{i,t}\langle (\Delta_t W^{3,i})h_t^{2,i}, \Delta_t W^4\rangle]_i$                                 & $0$                 & $\Theta(1/N^2)$         & \textcolor{green!50!black}{$\Theta(1/N)$}\\
$\quad (1/M)\Delta_t Q^\top[\dot\phi_{i,t}\langle (W_0^{3,i})h_t^{2,i}, W_0^4\rangle]_i$                                        & $0$                 & $\Theta(1/N^2)$         & \textcolor{red!70!black}{$\Theta(1/N^2)$}\\
$\quad (1/M)\Delta_t Q^\top[\dot\phi_{i,t}\langle (W_0^{3,i})h_t^{2,i}, \Delta_t W^4\rangle]_i$                                 & $0$                 & $\Theta(1/N^3)$         & \textcolor{red!70!black}{$\Theta(1/N^2)$}\\
$\quad (1/M)\Delta_t Q^\top[\dot\phi_{i,t}\langle (\Delta_t W^{3,i})h_t^{2,i}, W_0^4\rangle]_i$                                 & $0$                 & $\Theta(1/N^2)$         & \textcolor{green!50!black}{$\Theta(1/N)$}\\
$\quad (1/M)\Delta_t Q^\top[\dot\phi_{i,t}\langle (\Delta_t W^{3,i})h_t^{2,i}, \Delta_t W^4\rangle]_i$                          & $0$                 & $\Theta(1/N^3)$         & \textcolor{green!50!black}{$\Theta(1/N)$}\\
$\partial f_t/\partial h_t^1 = (\partial f_t/\partial h_t^1)_\mathrm{exp} + (\partial f_t/\partial h_t^1)_\mathrm{router}$      & $\Theta(1/N^{3/2})$ & $\Theta(1/N)$           & \textcolor{green!50!black}{$\Theta(1/N)$}\\
\end{tabular}}
\end{center}

\subsubsection{Deriving MSSP in Regime II}
\label{ssec:ssp_r2_consolidated}

This subsection presents the MSSP-Regime-II derivation per Definition~\ref{def:ssp_regime_ii}, with the user-specified modification $W_0^4 = 0$. MSSP-Regime-II differs from $\mu$P-Regime-II (Definition~\ref{def:mup_baseline_regime_ii}) in exactly two places: the variance of $W_0^{3,i}$ is $M/N_e$ (entries $\Theta(\sqrt N)$) rather than $1/N_e$ (entries $\Theta(1)$); and $W_0^4 = 0$ rather than variance $1/N^2$. All other init variances, learning rates, gating, and standing assumptions are identical to $\mu$P-Regime-II.

Two structural consequences propagate through the derivation:

\begin{enumerate}
    \item The boosted $W_0^{3,i}$ variance gives $h_0^{3,i}$ entries of scale $\Theta(\sqrt N)$ (vs $\Theta(1)$ in $\mu$P), which after $1/M$-aggregation with the cross-$i$ CLT gives $h_0^3$ entries $\Theta(1)$. So the aggregated activation is at the leading scale at init, even though the prediction $f_0 = 0$ vanishes.
    \item Since $W_0^4 = 0$, every gradient at $t=0$ that carries a $W_0^4$ factor is zero, so only $W^4$ updates at $t=1$. Hidden weights first move at $t=2$. Three forward + three backward passes are needed to capture all post-update correlations.
\end{enumerate}

Compared to $\mu$P-Regime-II: MSSP-Regime-II's larger $W^{3,i}$ init means the bottleneck-layer expert weights start big and the per-expert hidden state $h^{3,i}$ inherits that scale; updates to $W^{3,i}$ are then sub-leading relative to its init (vs comparable in $\mu$P-Regime-II). The aggregated $h^3$ still feature-learns at $\Theta(1)$, just through a different mechanism: the prediction reaches $f_1 \in\Theta(1)$ at $t=1$ purely from the $W^4 \cdot h_0^3$ contraction (since $h_0^3 \in\Theta(1)$ entries, $\|h_0^3\|^2\in\Theta(N)$), with no hidden updates required.

\paragraph{Setup and standing assumptions}
\label{ssec:ssp_r2_setup}

\paragraph{Architecture.} Same as in §\ref{ssec:mup_r2_setup}.

\paragraph{Initialization.} Per Definition~\ref{def:ssp_regime_ii} with $W_0^4 = 0$:
\begin{align*}
    (W_0^1)_{ab}&\sim\mathcal{N}(0,1/D), &
    (Q_0)_{ia}&\sim\mathcal{N}(0,1/N),\\
    (W_0^{2,i})_{ab}&\sim\mathcal{N}(0,1/N), &
    (W_0^{3,i})_{ab}&\sim\mathcal{N}(0, M/N_e),\\
    W_0^4 &= 0. &
\end{align*}
All parameters drawn independently across $i$. Resulting entry scales: $W_0^4 = 0$, $W_0^{2,i}\in\Theta(1/\sqrt N)$, $W_0^{3,i}\in\Theta(\sqrt M)=\Theta(\sqrt N)$, $W_0^1\in\Theta(1)$, $Q_0\in\Theta(1/\sqrt N)$.

\paragraph{Learning rates (SGD).} Same as $\mu$P-Regime-II: $\eta_1=\eta N$, $\eta_Q=\eta M/N$, $\eta_2=\eta M/N$, $\eta_3=\eta MN$, $\eta_4=\eta/N$.

\paragraph{Gating.} Sigmoid + $1/M$ aggregation: $\phi_{i,t}, \dot\phi_{i,t}\in\Theta(1)$.

\paragraph{Auxiliary scaling lemmas}
\label{ssec:ssp_r2_lemmas}

We reuse CLT, LLN, and Lemma~\ref{lem:ssp_gram}.

\textbf{Specialization of Lemma~\ref{lem:ssp_gram} to MSSP-Regime-II.} For $W=W_0^{3,i}\in\mathbb{R}^{N\times N_e}$ with $\sigma_W^2 = M/N_e$ and $N_e=\Theta(1)$:
\begin{itemize}
    \item $W_0^{3,i}(W_0^{3,i})^\top\in\mathbb{R}^{N\times N}$ has rank $\le N_e=\Theta(1)$. Diagonal mean $n\sigma^2 = M = \Theta(N)$, off-diagonal entries variance $n\sigma^4 = M^2/N_e = \Theta(N^2)$, coordinate scale $\Theta(N)$. \textbf{Every entry is $\Theta(N)$} — much larger than $\mu$P-Regime-II's $\Theta(1)$ entries.
    \item $(W_0^{3,i})^\top W_0^{3,i}\in\mathbb{R}^{N_e\times N_e}$, rank-$N_e$ full. Diagonal mean $m\sigma^2 = NM/N_e = \Theta(N^2)$, off-diagonal coord scale $\Theta(\sigma^2\sqrt m)=\Theta(N^{3/2})$. So acting on a $\Theta(1)$-coordinate $N_e$-vector $h_0^{2,i}$ gives entries of coordinate scale $\Theta(N)\cdot\Theta(1) + \Theta(N^{3/2})\cdot\Theta(1) = \Theta(N^2)$ at leading.
\end{itemize}
The other Gram products ($W_0^{2,i}$ versions) are unchanged from $\mu$P-Regime-II.

\paragraph{First forward pass}
\label{ssec:ssp_r2_fwd1}

\[
    h_0^1 = W_0^1\,x \in \Theta(1). \quad (\text{CLT};\ \sigma_1^2=1/D,\ \|x\|^2\in\Theta(D)) \tag{R2$s$-F1.1}
\]
\[
    \psi_0 = Q_0\,h_0^1 \in \Theta(1). \quad (\text{CLT};\ \sigma_Q^2=1/N,\ \|h_0^1\|^2\in\Theta(N)) \tag{R2$s$-F1.2}
\]
\[
    \phi_{i,0} = \sigma(\psi_{i,0}) \in \Theta(1). \quad (\text{sigmoid bounded; gating assumption}) \tag{R2$s$-F1.3}
\]
\begin{align*}
    h_0^{2,i} &= W_0^{2,i}\,h_0^1 \in \Theta(1),\quad \text{independent across }i\\
    & \quad (\text{CLT};\ \sigma_2^2=1/N,\ \|h_0^1\|^2\in\Theta(N);\\
    & \qquad \|h_0^{2,i}\|^2=N_e\cdot\Theta(1)=\Theta(1)\text{ since }N_e=\Theta(1)). \tag{R2$s$-F1.4}
\end{align*}
\begin{align*}
    h_0^{3,i} &= W_0^{3,i}\,h_0^{2,i} \in \Theta(\sqrt N),\quad \|h_0^{3,i}\|^2\in\Theta(N^2) \quad (\textbf{boosted by MSSP variance}) \tag{R2$s$-F1.5}\\
    & \quad (\text{CLT};\ \sigma_3^2=M/N_e=\Theta(N);\ \|h_0^{2,i}\|^2\in\Theta(1)\text{ via (R2$s$-F1.4)};\\
    & \qquad \text{variance per entry }\Theta(N)\cdot\Theta(1)=\Theta(N))
\end{align*}
\begin{align*}
    h_0^3 &= (1/M)\textstyle\sum_i\phi_{i,0}\,h_0^{3,i} \in \Theta(1) \tag{R2$s$-F1.6}\\
    & \quad (\text{cross-$i$ CLT on }h_0^{3,i}\text{ entries }\Theta(\sqrt N)\text{ via (R2$s$-F1.5)};\\
    & \qquad \{h_0^{3,i}\}\text{ independent across }i;\\
    & \qquad \text{variance per entry }(1/M^2)\textstyle\sum_i\phi_{i,0}^2\,\Theta(N) = \Theta(N/M) = \Theta(1);\\
    & \qquad \sqrt M\text{ cross-}i\text{ CLT cancellation balances the }\sqrt M\text{ amplification of }h_0^{3,i})
\end{align*}
\[
    f_0 = W_0^4\,h_0^3 = 0. \quad (W_0^4 = 0\text{ by MSSP-Regime-II standing assumption}) \tag{R2$s$-F1.7}
\]

\paragraph{First backward pass and step-1 updates}
\label{ssec:ssp_r2_bwd1}

By the chain rule, every hidden gradient at $t=0$ ($\partial f_0/\partial h_0^3$, $\partial f_0/\partial h_0^{3,i}$, $\partial f_0/\partial h_0^{2,i}$, $\partial f_0/\partial \phi_{i,0}$, and both expert and router contributions to $\partial f_0/\partial h_0^1$) carries a $W_0^4 = 0$ factor and is therefore zero. Hence $W_1^\ell = W_0^\ell$ for $\ell\in\{1,Q,(2,i),(3,i)\}$; only $W^4$ is updated at $t=1$.

\paragraph{Step-1 parameter update.}
\[
    \Delta_1 W^4 = -(\eta\chi_0/N)\,(h_0^3)^\top \in \Theta(1/N). \quad (h_0^3\in\Theta(1)\text{ via (R2$s$-F1.6)}) \tag{R2$s$-U1.4}
\]

\paragraph{Second forward pass}
\label{ssec:ssp_r2_fwd2}

Since the non-readout weights are unchanged at $t=1$, every hidden activation is unchanged: $h_1^\ell=h_0^\ell$ for $\ell\in\{1,(2,i),(3,i),3\}$, $\psi_1=\psi_0$, $\phi_1=\phi_0$.

\[
    f_1 = W_1^4\,h_0^3 = -\tfrac{\eta\chi_0}{N}\,\|h_0^3\|^2 \in \Theta(1). \quad (\|h_0^3\|^2\in\Theta(N)\text{ via (R2$s$-F1.6)}) \tag{R2$s$-F2.7}
\]

\paragraph{Second backward pass and step-2 updates}
\label{ssec:ssp_r2_bwd2}

\paragraph{New intermediate at $t=1$.}
\begin{align*}
    (W_0^{3,i})^\top h_0^3 &\in \Theta(N) \text{ entry-wise}\\
    & \quad (\text{using (R2$s$-F1.6)}\text{ to expand }h_0^3=(1/M)\sum_j\phi_{j,0}h_0^{3,j};\\
    & \qquad j=i\text{ piece }(\phi_{i,0}/M)(W_0^{3,i})^\top W_0^{3,i}h_0^{2,i}\text{ with}\\
    & \qquad (W_0^{3,i})^\top W_0^{3,i}h_0^{2,i}\in\Theta(N^2)\text{ from Lemma~\ref{lem:ssp_gram} specialization},\\
    & \qquad \text{giving }(\phi_{i,0}/M)\cdot\Theta(N^2)=\Theta(N);\\
    & \qquad j\neq i\text{ pieces give cross-$j$ noise of comparable }\Theta(N)\text{ scale})\\[2pt]
    \Rightarrow\ (W_0^{3,i})^\top(W_1^4)^\top &= -(\eta\chi_0/N)(W_0^{3,i})^\top h_0^3 \in \Theta(1) \text{ entry-wise}\\
    & \quad (W_1^4\text{ via (R2$s$-U1.4)})
\end{align*}

\paragraph{Per-layer gradients.}

\[
    \partial f_1/\partial h_1^3 = (W_1^4)^\top \in \Theta(1/N). \quad (\text{via (R2$s$-U1.4)}) \tag{R2$s$-B2.1}
\]
\[
    \partial f_1/\partial h_1^{3,i} = (\phi_{i,0}/M)(W_1^4)^\top \in \Theta(1/N^2). \quad (\phi_{i,0}\in\Theta(1);\ M=\Theta(N)) \tag{R2$s$-B2.2}
\]
\begin{align*}
    \partial f_1/\partial h_1^{2,i} &= (\phi_{i,0}/M)(W_0^{3,i})^\top(W_1^4)^\top \in \Theta(1/N) \tag{R2$s$-B2.3}\\
    & \quad ((W_0^{3,i})^\top(W_1^4)^\top\in\Theta(1)\text{ via the new intermediate above})
\end{align*}
\begin{align*}
    \partial f_1/\partial \phi_{i,0} &= (1/M)\langle h_0^{3,i}, W_1^4\rangle = -(\eta\chi_0/(MN))\langle h_0^{3,i}, h_0^3\rangle \in \Theta(1/N) \tag{R2$s$-B2.4}\\
    & \quad (\text{Four-piece grid: only (init}\cdot\text{update) is non-zero at }t=1,\\
    & \qquad \text{since }W_0^4=0\text{ kills the *}\cdot\text{init column}\\
    & \qquad \text{and }\Delta_1 W^{3,i}=0\text{ (hidden weights frozen) kills the update}\cdot\text{* row;}\\
    & \qquad (\text{init}\cdot\text{update})\ \langle (W_0^{3,1})\,h_0^{2,i}, \Delta_1 W^4\rangle\in\Theta(1):\\
    & \qquad\quad \text{expanding }h_0^3=(1/M)\sum_j\phi_{j,0}h_0^{3,j}\text{ gives}\\
    & \qquad\quad \langle h_0^{3,i}, h_0^3\rangle = (\phi_{i,0}/M)\|h_0^{3,i}\|^2 + (1/M)\textstyle\sum_{j\ne i}\phi_{j,0}\langle h_0^{3,i}, h_0^{3,j}\rangle;\\
    & \qquad\quad j=i\text{ diagonal term: }(\phi_{i,0}/M)\|h_0^{3,i}\|^2=(1/M)\Theta(N^2)=\Theta(N)\\
    & \qquad\quad \text{using }\|h_0^{3,i}\|^2\in\Theta(N^2)\text{ via (R2$s$-F1.5)};\\
    & \qquad\quad j\neq i\text{ off-diagonal terms give comparable }\Theta(N)\text{ via cross-}j\text{ CLT};\\
    & \qquad\quad \text{so }\langle h_0^{3,i}, h_0^3\rangle\in\Theta(N);\\
    & \qquad \text{substituting: }\partial f_1/\partial\phi_{i,0} = -(\eta\chi_0/(MN))\cdot\Theta(N) = \Theta(1/M) = \Theta(1/N);\\
    & \qquad \text{lifted from }\mu\text{P-R2's (init}\cdot\text{update)}=\Theta(1/N^2)\text{ entry}\\
    & \qquad \text{by the boost }\sigma_3^2=M/N_e)
\end{align*}
\begin{align*}
    (\partial f_1/\partial h_1^1)_\mathrm{exp} &= (1/M)\textstyle\sum_i\phi_{i,0}(W_0^{2,i})^\top(W_0^{3,i})^\top(W_1^4)^\top \in \Theta(1/N) \tag{R2$s$-B2.5}\\
    & \quad ((W_0^{3,i})^\top(W_1^4)^\top\in\Theta(1)\text{ via the new intermediate above};\\
    & \qquad (W_0^{2,i})^\top\text{ acts on a }\Theta(1)\text{ vector with }\|v\|^2=N_e\Theta(1)=\Theta(1);\\
    & \qquad \text{variance per entry }(1/N)\Theta(1)=\Theta(1/N),\text{ coord }\Theta(1/\sqrt N)\text{ random per }i;\\
    & \qquad \text{cross-$i$ CLT: }\Theta(1/N))
\end{align*}
\begin{align*}
    (\partial f_1/\partial h_1^1)_\mathrm{router} &= (1/M)Q_0^\top v,\quad v_i:=\dot\phi_{i,0}\langle h_0^{3,i}, W_1^4\rangle \in \Theta(1/N) \tag{R2$s$-B2.6}\\
    & \quad (v_i\in\Theta(1)\text{ via (R2$s$-B2.4)};\ \|v\|^2=M\Theta(1)=\Theta(M);\\
    & \qquad Q_0^\top v\text{ entries variance }(1/N)\Theta(M)=\Theta(1)\text{ via cross-layer CLT};\\
    & \qquad (1/M)\Theta(1)=\Theta(1/N);\\
    & \qquad \text{W}^{3,i}\text{-split }v_i = v^I_i + v^U_i\text{ where}\\
    & \qquad v^I_i := \dot\phi_{i,0}\langle (W_0^{3,1})h_0^{2,i}, W_1^4\rangle\\
    & \qquad v^U_i := \dot\phi_{i,0}\langle (\Delta_1 W^{3,i})h_0^{2,i}, W_1^4\rangle = 0\text{ since }\Delta_1 W^{3,i}=0;\\
    & \qquad v^I_i\in\Theta(1)\text{ coherent via boost (cf.\ R2$s$-B2.4)};\\
    & \qquad (1/M)Q_0^\top v^I = \Theta(1/N)\text{ and }(1/M)Q_0^\top v^U = 0)
\end{align*}
\[
    \partial f_1/\partial h_1^1 = (\partial f_1/\partial h_1^1)_\mathrm{exp} + (\partial f_1/\partial h_1^1)_\mathrm{router} \in \Theta(1/N). \tag{R2$s$-B2.7}
\]

\paragraph{Step-2 parameter updates.}
\[
    \Delta_2 W^4 = -(\eta\chi_1/N)(h_0^3)^\top \in \Theta(1/N). \quad (h_0^3\in\Theta(1)\text{ via (R2$s$-F1.6)}) \tag{R2$s$-U2.4}
\]
\begin{align*}
    \Delta_2 W^{3,i} &= -\eta N\chi_1\phi_{i,0}\,(W_1^4)^\top(h_0^{2,i})^\top \in \Theta(1) \text{ rank-1} \tag{R2$s$-U2.3}\\
    & \quad ((W_1^4)^\top\in\Theta(1/N)\text{ via (R2$s$-U1.4)};\ h_0^{2,i}\in\Theta(1)\text{ via (R2$s$-F1.4)};\\
    & \qquad \text{sub-leading vs }W_0^{3,i}\in\Theta(\sqrt N)\text{ via (R2$s$-F1.5)})
\end{align*}
\begin{align*}
    \Delta_2 W^{2,i} &= -(\eta\chi_1\phi_{i,0}/N)(W_0^{3,i})^\top(W_1^4)^\top(h_0^1)^\top \in \Theta(1/N) \text{ rank-1} \tag{R2$s$-U2.2}\\
    & \quad ((W_0^{3,i})^\top(W_1^4)^\top\in\Theta(1)\text{ via the new intermediate};\\
    & \qquad \text{sub-leading vs }W_0^{2,i}\in\Theta(1/\sqrt N))
\end{align*}
\begin{align*}
    \Delta_2 W^1 &= -\eta_1\chi_1(\partial f_1/\partial h_1^1)\,x^\top \in \Theta(1) \tag{R2$s$-U2.1a}\\
    & \quad (\eta_1=\eta N;\ \partial f_1/\partial h_1^1\in\Theta(1/N)\text{ via (R2$s$-B2.7)})
\end{align*}
\[
    \Delta_2 h^1 = \Delta_2 W^1\,x \in \Theta(1), \text{ aligned along } h_0^1. \tag{R2$s$-U2.1b}
\]
\[
    \Delta_2 Q \in \Theta(1/N). \quad (\partial f_1/\partial\phi\in\Theta(1/N)\text{ via (R2$s$-B2.4)}) \tag{R2$s$-U2.Q}
\]

\paragraph{Third forward pass}
\label{ssec:ssp_r2_fwd3}

\[
    h_2^1 = h_0^1 + \Delta_2 h^1 \in \Theta(1). \quad (\Delta_2 h^1\in\Theta(1)\text{ via (R2$s$-U2.1b)};\ h_0^1\in\Theta(1)) \tag{R2$s$-F3.1}
\]

\noindent (R2$s$-F3.2) $h_2^{2,i} = W_2^{2,i}h_2^1$, four-piece decomposition:
\[
    h_2^{2,i} = h_0^{2,i} + W_0^{2,i}\Delta_2 h^1 + \Delta_2 W^{2,i}h_0^1 + \Delta_2 W^{2,i}\Delta_2 h^1. \tag{R2$s$-F3.2}
\]
\[
    \text{init: } h_0^{2,i} \in \Theta(1). \quad (\text{R2$s$-F1.4}) \tag{R2$s$-F3.2a}
\]
\begin{align*}
    \text{prop: } W_0^{2,i}\Delta_2 h^1 &\in \Theta(1)\\
    & \quad (\sigma_2^2\|\Delta_2 h^1\|^2 = (1/N)\Theta(N) = \Theta(1);\\
    & \qquad \|\Delta_2 h^1\|^2\in\Theta(N)\text{ via (R2$s$-U2.1b)}). \tag{R2$s$-F3.2b}
\end{align*}
\begin{align*}
    \text{eff: } \Delta_2 W^{2,i}h_0^1 &= -(\eta\chi_1\phi_{i,1}/N)\|h_0^1\|^2(W_0^{3,i})^\top(W_1^4)^\top \in \Theta(1)\\
    & \quad (\text{(R2$s$-U2.2)substitution};\\
    & \qquad (W_0^{3,i})^\top(W_1^4)^\top\in\Theta(1)\text{ via the new intermediate at }t=1;\\
    & \qquad \|h_0^1\|^2\in\Theta(N)). \tag{R2$s$-F3.2c}
\end{align*}
\begin{align*}
    \text{cross: } \Delta_2 W^{2,i}\Delta_2 h^1 &= -(\eta\chi_1\phi_{i,1}/N)\langle h_0^1,\Delta_2 h^1\rangle(W_0^{3,i})^\top(W_1^4)^\top \in \Theta(1)\\
    & \quad ((W_0^{3,i})^\top(W_1^4)^\top\in\Theta(1)\text{ via the new intermediate};\\
    & \qquad \langle h_0^1,\Delta_2 h^1\rangle\in\Theta(N)\text{ coherent via (R2$s$-U2.1b)}). \tag{R2$s$-F3.2d}
\end{align*}
$\Delta_2 h^{2,i}\in\Theta(1)$ entry-wise.

\noindent (R2$s$-F3.3) $h_2^{3,i} = W_2^{3,i}h_2^{2,i}$, four-piece decomposition:
\[
    h_2^{3,i} = h_0^{3,i} + W_0^{3,i}\Delta_2 h^{2,i} + \Delta_2 W^{3,i}h_0^{2,i} + \Delta_2 W^{3,i}\Delta_2 h^{2,i}. \tag{R2$s$-F3.3}
\]
\[
    \text{init: } h_0^{3,i} \in \Theta(\sqrt N). \quad (\text{R2$s$-F1.5}) \tag{R2$s$-F3.3a}
\]
\begin{align*}
    \text{prop: } W_0^{3,i}\Delta_2 h^{2,i} &\in \Theta(\sqrt N)\\
    & \quad (\sigma_3^2\|\Delta_2 h^{2,i}\|^2 = \Theta(N)\cdot\Theta(1) = \Theta(N);\\
    & \qquad \|\Delta_2 h^{2,i}\|^2 = N_e\Theta(1) = \Theta(1)\text{ via (R2$s$-F3.2)};\ N_e=\Theta(1)). \tag{R2$s$-F3.3b}
\end{align*}
\begin{align*}
    \text{eff: } \Delta_2 W^{3,i}h_0^{2,i} &= -\eta N\chi_1\phi_{i,1}\|h_0^{2,i}\|^2(W_1^4)^\top \in \Theta(1)\\
    & \quad (\text{(R2$s$-U2.3)substitution};\\
    & \qquad \|h_0^{2,i}\|^2\in\Theta(1)\text{ via (R2$s$-F1.4)};\\
    & \qquad (W_1^4)^\top\in\Theta(1/N)\text{ via (R2$s$-U1.4)}). \tag{R2$s$-F3.3c}
\end{align*}
\[
    \text{cross: } \Delta_2 W^{3,i}\Delta_2 h^{2,i} \text{ tracked as part of }D\text{ in (R2$s$-F3.4)}. \tag{R2$s$-F3.3d}
\]
\textbf{Propagating dominates}: $\Delta_2 h^{3,i}\in\Theta(\sqrt N)$ — same scale as $h_0^{3,i}$ at init. The effective piece $\Theta(1)$ is sub-leading per-expert but contributes coherently to the aggregation in (R2$s$-F3.4d).

\noindent (R2$s$-F3.4) $h_2^3 = A_1 + A_{2,1} + A_{2,2} + A_3 + D$, where
\begin{align*}
    A_1 &:= \tfrac{1}{M}\textstyle\sum_i\phi_{i,2}h_0^{3,i}, &
    A_{2,1} &:= \tfrac{1}{M}\textstyle\sum_i\phi_{i,2}W_0^{3,i}W_0^{2,i}\Delta_2 h^1,\\
    A_{2,2} &:= \tfrac{1}{M}\textstyle\sum_i\phi_{i,2}W_0^{3,i}\Delta_2 W^{2,i}h_0^1, &
    A_3 &:= \tfrac{1}{M}\textstyle\sum_i\phi_{i,2}\Delta_2 W^{3,i}h_0^{2,i},\\
    D &:= \tfrac{1}{M}\textstyle\sum_i\phi_{i,2}\Delta_2 W^{3,i}\Delta_2 h^{2,i}.
\end{align*}
\[
    A_1 \in \Theta(1). \quad (\text{cross-$i$ CLT, same calculation as (R2$s$-F1.6)}) \tag{R2$s$-F3.4a}
\]
\begin{align*}
    A_{2,1} &\in \Theta(1)\\
    & \quad (\text{per summand }W_0^{3,i}W_0^{2,i}\Delta_2 h^1\in\Theta(\sqrt N)\text{ random};\\
    & \qquad \|\Delta_2 h^1\|^2\in\Theta(N)\text{ via (R2$s$-U2.1b)};\\
    & \qquad \text{cross-$i$ CLT: }\Theta(\sqrt N/\sqrt M)=\Theta(1)). \tag{R2$s$-F3.4b}
\end{align*}
\begin{align*}
    A_{2,2} &\in \Theta(1)\\
    & \quad (\text{per summand }W_0^{3,i}[\Delta_2 W^{2,i}h_0^1]\in\Theta(\sqrt N)\\
    & \qquad \text{since }\Delta_2 W^{2,i}h_0^1\in\Theta(1)\text{ via (R2$s$-F3.2c)};\\
    & \qquad \text{cross-$i$ CLT: }\Theta(\sqrt N/\sqrt M)=\Theta(1)). \tag{R2$s$-F3.4c}
\end{align*}
\begin{align*}
    A_3 &= -\eta N\chi_1(W_1^4)^\top\Bigl(\tfrac{1}{M}\textstyle\sum_i\phi_{i,2}\phi_{i,1}\|h_0^{2,i}\|^2\Bigr) \in \Theta(1)\\
    & \quad (\text{(R2$s$-U2.3)substitution};\\
    & \qquad \|h_0^{2,i}\|^2\in\Theta(1)\text{ via (R2$s$-F1.4)};\ \text{LLN};\\
    & \qquad (W_1^4)^\top\in\Theta(1/N)). \tag{R2$s$-F3.4d}
\end{align*}
\begin{align*}
    D &= -\eta N\chi_1(W_1^4)^\top\,\tfrac{1}{M}\textstyle\sum_i\phi_{i,2}\phi_{i,1}\langle h_0^{2,i},\Delta_2 h^{2,i}\rangle \in \Theta(1)\\
    & \quad (\text{(R2$s$-U2.3)substitution};\\
    & \qquad \langle h_0^{2,i},\Delta_2 h^{2,i}\rangle\in\Theta(1)\text{ coherent across }i,\\
    & \qquad \text{from two pieces: }\Delta_2 h^1\parallel h_0^1\text{ via (R2$s$-U2.1b)},\\
    & \qquad \text{and the }j=i\text{ Gram self-overlap in the effective piece (R2$s$-F3.2c)};\\
    & \qquad \text{LLN gives empirical average }\Theta(1)). \tag{R2$s$-F3.4e}
\end{align*}
\[
    h_2^3 \in \Theta(1) \text{ entry-wise}. \tag{R2$s$-F3.4}
\]
\begin{align*}
    f_2 &= W_2^4\,h_2^3 = -(\eta(\chi_0+\chi_1)/N)\langle h_0^3, h_2^3\rangle \in \Theta(1)\\
    & \quad (W_2^4 = W_1^4 + \Delta_2 W^4\text{ via (R2$s$-U2.4)};\\
    & \qquad \langle h_0^3, h_2^3\rangle\in\Theta(N)\text{ coherent, dominated by }A_3+D\text{ via (R2$s$-F3.4)};\\
    & \qquad \text{net }\eta\cdot(1/N)\cdot\Theta(N) = \Theta(1)). \tag{R2$s$-F3.5}
\end{align*}

\paragraph{Third backward pass}
\label{ssec:ssp_r2_bwd3}

The structure parallels (R2$s$-B2.1)--(R2$s$-B2.7) with cumulative weights at step 2. Since $W_0^4=0$ in MSSP, the four ``$.1$'' pieces $A_{4.1}, A_{5.1}, A_{6.1}, E_1$ — using $W_0^4$ rather than $\Delta W^4$ — are identically zero at every step; only the four ``$.2$'' pieces contribute. With $W_2^4 = \Delta W^4$ now non-zero (via (R2$s$-U2.4)), all gradients are non-zero.

\paragraph{Per-layer gradients.}
\[
    \partial f_2/\partial h_2^3 = (W_2^4)^\top \in \Theta(1/N). \quad (\text{(R2$s$-U2.4)}) \tag{R2$s$-B3.1}
\]
\[
    \partial f_2/\partial h_2^{3,i} = (\phi_{i,2}/M)(W_2^4)^\top \in \Theta(1/N^2). \quad (\phi_{i,2}\in\Theta(1);\ M=\Theta(N)) \tag{R2$s$-B3.2}
\]
\begin{align*}
    \partial f_2/\partial h_2^{2,i} &= (\phi_{i,2}/M)(W_2^{3,i})^\top(W_2^4)^\top \in \Theta(1/N)\\
    & \quad ((W_0^{3,i})^\top(W_2^4)^\top\in\Theta(1)\text{ via the cumulative analog of the new intermediate at }t=1;\\
    & \qquad (\Delta_2 W^{3,i})^\top(W_2^4)^\top\in\Theta(1)\text{ via (R2$s$-U2.3)}). \tag{R2$s$-B3.3}
\end{align*}
\paragraph{Router pathway $\partial f_2/\partial \phi_{i,2}$.}
\begin{align*}
    \partial f_2/\partial \phi_{i,2} &= (1/M)\langle h_2^{3,i}, W_2^4\rangle \in \Theta(1/N), \tag{R2$s$-B3.4}\\
    & \quad \text{Four-piece grid: }*\cdot\text{init column is identically zero }(W_0^4=0);\\
    & \quad (\text{init}\cdot\text{update})\ \langle (W_0^{3,1})\,h_2^{2,i}, \Delta_2 W^4\rangle\in\Theta(1)\text{: }\\
    & \qquad \langle h_0^{3,i}, W_2^4\rangle = -\tfrac{\chi_0+\chi_1}{N}\langle h_0^{3,i}, h_0^3\rangle,\\
    & \qquad \text{expanding }h_0^3=(1/M)\textstyle\sum_j\phi_{j,0}h_0^{3,j}\text{ gives the diagonal }j=i\text{ term}\\
    & \qquad (\phi_{i,0}/M)\|h_0^{3,i}\|^2 = (1/M)\Theta(N^2)=\Theta(N)\text{ via (R2$s$-F1.5)};\\
    & \qquad j\neq i\text{ off-diagonal terms give comparable }\Theta(N);\\
    & \qquad \text{so }\langle h_0^{3,i}, W_2^4\rangle\in\Theta(1);\\
    & \quad (\text{update}\cdot\text{update})\ \langle (\Delta_2 W^{3,i})\,h_2^{2,i}, \Delta_2 W^4\rangle\in\Theta(1)\text{: }\\
    & \qquad \langle (\Delta_2 W^{3,i})\,h_2^{2,i}, W_2^4\rangle = \kappa\langle (W_1^4)^\top, W_2^4\rangle\\
    & \qquad \text{with }\kappa = -\eta N\chi_1\phi_{i,1}\|h^{2,i}\|^2\in\Theta(N)\\
    & \qquad \text{and }\langle W_1^4, W_2^4\rangle = (\chi_0(\chi_0+\chi_1)/N^2)\|h_0^3\|^2\in\Theta(1/N);\\
    & \qquad \text{net }\Theta(N)\cdot\Theta(1/N)=\Theta(1);\\
    & \quad \text{substituting: }\partial f_2/\partial\phi_{i,2} = (1/M)\bigl(\Theta(1) + \Theta(1)\bigr) = \Theta(1/M) = \Theta(1/N);\\
    & \quad \text{lifted from }\mu\text{P-R2's grid }(\Theta(1/N^{3/2}),\Theta(1/N^{3/2}),\Theta(1/N),\Theta(1/N))\\
    & \quad \text{to MSSP-R2's }(0,\Theta(1/N),0,\Theta(1/N))\text{ via }W_0^4=0\text{ and }\sigma_3^2=M/N_e.
\end{align*}

\paragraph{Expert pathway: $(\partial f_2/\partial h_2^1)_\mathrm{exp} = A_{4.2} + A_{5.2} + A_{6.2} + E_2$.}
\begin{align*}
    A_{4.2} &= \tfrac{1}{M}\textstyle\sum_i\phi_{i,2}(W_0^{2,i})^\top(W_0^{3,i})^\top(W_2^4)^\top \in \Theta(1/N)\\
    & \quad ((W_0^{3,i})^\top(W_2^4)^\top\in\Theta(1)\text{ via the cumulative analog of the new intermediate};\\
    & \qquad (W_0^{2,i})^\top\text{ acts on }\Theta(1)\text{ vector with }\|v\|^2=\Theta(1);\\
    & \qquad \text{coord }\Theta(1/\sqrt N)\text{ random per }i,\text{ cross-$i$ CLT: }\Theta(1/N)). \tag{R2$s$-B3.5a}
\end{align*}
\begin{align*}
    A_{5.2} &= \tfrac{1}{M}\textstyle\sum_i\phi_{i,2}(W_0^{2,i})^\top(\Delta_2 W^{3,i})^\top(W_2^4)^\top \in \Theta(1/N)\\
    & \quad (\text{(R2$s$-U2.3)substitution gives }(\Delta_2 W^{3,i})^\top(W_2^4)^\top\in\Theta(1)\text{ along }h_0^{2,i};\\
    & \qquad (W_0^{2,i})^\top h_0^{2,i}\in\Theta(1/\sqrt N)\text{ random per }i;\\
    & \qquad \text{cross-$i$ CLT: }\Theta(1/N)). \tag{R2$s$-B3.5b}
\end{align*}
\begin{align*}
    A_{6.2} &= \tfrac{1}{M}\textstyle\sum_i\phi_{i,2}(\Delta_2 W^{2,i})^\top(W_0^{3,i})^\top(W_2^4)^\top \in \Theta(1/N)\\
    & \quad (\text{(R2$s$-U2.2)substitution};\\
    & \qquad \text{per summand }-(\eta\chi_1\phi_{i,1}/N)h_0^1\cdot S_i\text{ where}\\
    & \qquad S_i := W_1^4\,W_0^{3,i}(W_0^{3,i})^\top(W_2^4)^\top\in\Theta(1)\\
    & \qquad \text{(Lemma~\ref{lem:ssp_gram} specialization: }W_0^{3,i}(W_0^{3,i})^\top\text{ entries }\Theta(N),\\
    & \qquad \text{acting on }(W_2^4)^\top\in\Theta(1/N)\text{ gives }\Theta(\sqrt N);\\
    & \qquad W_1^4\text{ contraction yields }\Theta(1));\\
    & \qquad \text{per summand entries }\Theta(1/N),\text{ cross-$i$ avg: }\Theta(1/N)). \tag{R2$s$-B3.5c}
\end{align*}
\begin{align*}
    E_2 &= \tfrac{1}{M}\textstyle\sum_i\phi_{i,2}(\Delta_2 W^{2,i})^\top(\Delta_2 W^{3,i})^\top(W_2^4)^\top \in \Theta(1/N)\\
    & \quad (\text{combining (R2$s$-U2.2), (R2$s$-U2.3) substitutions};\\
    & \qquad (\Delta_2 W^{3,i})^\top(W_2^4)^\top\in\Theta(1)\text{ along }h_0^{2,i};\\
    & \qquad \langle W_1^4, h_0^{3,i}\rangle = -(\eta\chi_0/N)\langle h_0^3, h_0^{3,i}\rangle\in\Theta(1)\\
    & \qquad \text{(coherent }j=i\text{ piece via }\|h_0^{3,i}\|^2\in\Theta(N^2)\text{ from (R2$s$-F1.5)};\\
    & \qquad \text{plus }j\neq i\text{ random of comparable scale)};\\
    & \qquad \text{per summand entries }\Theta(1/N);\\
    & \qquad \text{cross-$i$ LLN on coherent piece: }\Theta(1/N)). \tag{R2$s$-B3.5d}
\end{align*}
\[
    (\partial f_2/\partial h_2^1)_\mathrm{exp} \in \Theta(1/N), \text{ all four non-zero pieces at leading scale}. \tag{R2$s$-B3.5}
\]

\paragraph{Router pathway.}
\begin{align*}
    (\partial f_2/\partial h_2^1)_\mathrm{router} &= (1/M)Q_2^\top v,\quad v_i:=\dot\phi_{i,2}\langle h_2^{3,i}, W_2^4\rangle \in \Theta(1/N)\\
    & \quad (v_i\in\Theta(1)\text{ via (R2$s$-B3.4)};\\
    & \qquad \text{by the analysis of (R2$s$-B2.6) extended to cumulative }Q_2 = Q_0 + \Delta_2 Q;\\
    & \qquad Q_0\text{ piece via cross-layer CLT};\\
    & \qquad \Delta_2 Q\text{ piece coherent via (R2$s$-U2.Q)};\\
    & \qquad \text{both contribute at }\Theta(1/N)). \tag{R2$s$-B3.6}
\end{align*}
\textit{W$^{3,i}$-balance transmits through the router pathway.} The W$^{3,i}$-split of (R2$s$-B3.4), $\mathbf{v}=\mathbf{v}^I+\mathbf{v}^U$ with $\mathbf{v}^I_i=\dot\phi_{i,2}\langle (W_0^{3,1})h_2^{2,i}, \Delta_2 W^4\rangle$ and $\mathbf{v}^U_i=\dot\phi_{i,2}\langle (\Delta_2 W^{3,i})h_2^{2,i}, \Delta_2 W^4\rangle$, transmits as: both $\mathbf{v}^I_i\in\Theta(1)$ and $\mathbf{v}^U_i\in\Theta(1)$ coherent across $i$ — the boost $\sigma_3^2 = M/N_e$ promotes the $W_0^{3,1}$-piece's diagonal $j=i$ contribution to $\Theta(1)$ (cf.\ R2$s$-B3.4), matching the rank-1-aligned $\Delta W^{3,i}$-piece. So $(1/M)Q_t^\top\mathbf{v}^I\in\Theta(1/N)$ and $(1/M)Q_t^\top\mathbf{v}^U\in\Theta(1/N)$: balanced, in contrast to the $\sqrt N$ deficit under $\mu$P-R2.
\[
    \partial f_2/\partial h_2^1 = (\partial f_2/\partial h_2^1)_\mathrm{exp} + (\partial f_2/\partial h_2^1)_\mathrm{router} \in \Theta(1/N). \tag{R2$s$-B3.7}
\]

\subsubsection{Summary tables of signal propagation for MSSP in Regime II}
\label{ssec:ssp_r2_summary}

\paragraph{Forward features.}

\begin{center}
\renewcommand{\arraystretch}{1.15}
\begin{tabular}{l|l|l|l}
\textbf{Quantity} & $t=0$ & $t=1$ & $t=2$\\\hline
$h_t^1$ (init) & $\Theta(1)$ & $\Theta(1)$ & \textcolor{green!50!black}{$\Theta(1)$}\\
$\Delta_t h^1$ & — & $0$ (no update) & \textcolor{green!50!black}{$\Theta(1)$}\\
$\quad$ effective $\Delta W^1\,x$ & — & $0$ & \textcolor{green!50!black}{$\Theta(1)$}\\
$\psi_t$ & $\Theta(1)$ & $\Theta(1)$ & \textcolor{green!50!black}{$\Theta(1)$}\\
$\Delta_t \psi$ & — & $0$ & \textcolor{green!50!black}{$\Theta(1)$}\\
$\quad$ effective $\Delta Q\,h_0^1$ & — & $0$ & \textcolor{green!50!black}{$\Theta(1)$}\\
$\quad$ propagating $Q_0\,\Delta h^1$ & — & $0$ & \textcolor{green!50!black}{$\Theta(1)$}\\
$\phi_t$ & $\Theta(1)$ & $\Theta(1)$ & \textcolor{green!50!black}{$\Theta(1)$}\\
$\Delta_t \phi$ & — & $0$ & \textcolor{green!50!black}{$\Theta(1)$}\\
$h_t^{2,i}$ & $\Theta(1)$ & $\Theta(1)$ & \textcolor{green!50!black}{$\Theta(1)$}\\
$\Delta_t h^{2,i}$ & — & $0$ & \textcolor{green!50!black}{$\Theta(1)$}\\
$\quad$ effective $\Delta W^{2,i}h_0^1$ & — & $0$ & \textcolor{green!50!black}{$\Theta(1)$}\\
$\quad$ propagating $W_0^{2,i}\Delta h^1$ & — & $0$ & \textcolor{green!50!black}{$\Theta(1)$}\\
$h_t^{3,i}$ & $\Theta(\sqrt N)$ & $\Theta(\sqrt N)$ & $\Theta(\sqrt N)$\\
$\Delta_t h^{3,i}$ & — & $0$ & {$\Theta(\sqrt N)$}\\
$\quad$ effective $\Delta W^{3,i}h_0^{2,i}$ & — & $0$ & \textcolor{green!50!black}{$\Theta(1)$}\\
$\quad$ propagating $W_0^{3,i}\Delta h^{2,i}$ & — & $0$ & $\Theta(\sqrt N)$\\
$h_t^3$ & $\Theta(1)$ & $\Theta(1)$ & \textcolor{green!50!black}{$\Theta(1)$}\\
$f_t$ & $0$ & $\Theta(1)$ & \textcolor{green!50!black}{$\Theta(1)$}\\
\end{tabular}
\end{center}

\paragraph{Forward aggregation: $h_t^3 = A_1 + A_{2,1} + A_{2,2} + A_3 + D$.}

\begin{center}
\renewcommand{\arraystretch}{1.15}
\begin{tabular}{l|l|l|l}
\textbf{Piece} & $t=0$ & $t=1$ & $t=2$\\\hline
$A_1=(1/M)\sum\phi h_0^{3,i}$ & $\Theta(1)$ & $\Theta(1)$ & \textcolor{green!50!black}{$\Theta(1)$}\\
$A_{2,1}=(1/M)\sum\phi W_0^{3,i}W_0^{2,i}\Delta h^1$ & $0$ & $0$ & \textcolor{green!50!black}{$\Theta(1)$}\\
$A_{2,2}=(1/M)\sum\phi W_0^{3,i}\Delta W^{2,i}h_0^1$ & $0$ & $0$ & \textcolor{green!50!black}{$\Theta(1)$}\\
$A_3=(1/M)\sum\phi\Delta W^{3,i}h_0^{2,i}$ & $0$ & $0$ & \textcolor{green!50!black}{$\Theta(1)$}\\
$D=(1/M)\sum\phi\Delta W^{3,i}\Delta h^{2,i}$ & $0$ & $0$ & \textcolor{green!50!black}{$\Theta(1)$}\\
\textbf{Total $h_t^3$} & $\Theta(1)$ & $\Theta(1)$ & \textcolor{green!50!black}{$\Theta(1)$}\\
\end{tabular}
\end{center}

\paragraph{Backward gradients (per-layer with $W_0^4/\Delta W^4$ split).}

\begin{center}
\renewcommand{\arraystretch}{1.15}
\begin{tabular}{l|l|l|l}
\textbf{Piece} & $t=0$ & $t=1$ & $t=2$\\\hline
$(W_0^4)^\top$ & $0$ & $0$ & $0$\\
$(\Delta W^4)^\top$ & $0$ & $\Theta(1/N)$ & \textcolor{green!50!black}{$\Theta(1/N)$}\\
$\partial f_t/\partial h_t^3$ \textbf{total} & $0$ & $\Theta(1/N)$ & \textcolor{green!50!black}{$\Theta(1/N)$}\\\hline
$\partial f_t/\partial h_t^{3,i}$ \textbf{total} & $0$ & $\Theta(1/N^2)$ & \textcolor{green!50!black}{$\Theta(1/N^2)$}\\\hline
$(\phi/M)(W_0^{3,i})^\top(W_0^4)^\top$ & $0$ & $0$ & $0$\\
$(\phi/M)(W_0^{3,i})^\top(\Delta W^4)^\top$ & $0$ & $\Theta(1/N)$ & \textcolor{green!50!black}{$\Theta(1/N)$}\\
$(\phi/M)(\Delta W^{3,i})^\top(W_0^4)^\top$ & $0$ & $0$ & $0$\\
$(\phi/M)(\Delta W^{3,i})^\top(\Delta W^4)^\top$ & $0$ & $0$ & \textcolor{green!50!black}{$\Theta(1/N)$}\\
$\partial f_t/\partial h_t^{2,i}$ \textbf{total} & $0$ & $\Theta(1/N)$ & \textcolor{green!50!black}{$\Theta(1/N)$}\\\hline
init$\cdot$init: $(1/M)\langle (W_0^{3,i})\,h_t^{2,i}, W_0^4\rangle$ & $0$ & $0$ & $0$\\
init$\cdot$update: $(1/M)\langle (W_0^{3,i})\,h_t^{2,i}, \Delta_t W^4\rangle$ & $0$ & $\Theta(1/N)$ & \textcolor{green!50!black}{$\Theta(1/N)$}\\
update$\cdot$init: $(1/M)\langle (\Delta_t W^{3,i})\,h_t^{2,i}, W_0^4\rangle$ & $0$ & $0$ & $0$\\
update$\cdot$update: $(1/M)\langle (\Delta_t W^{3,i})\,h_t^{2,i}, \Delta_t W^4\rangle$ & $0$ & $0$ & \textcolor{green!50!black}{$\Theta(1/N)$}\\
$\partial f_t/\partial \phi_{i,t}$ \textbf{total} & $0$ & $\Theta(1/N)$ & \textcolor{green!50!black}{$\Theta(1/N)$}\\
\end{tabular}
\end{center}

\paragraph{Expert pathway: 8-piece decomposition.}

\begin{center}
\renewcommand{\arraystretch}{1.15}
\begin{tabular}{l|c|l|l|l}
\textbf{Piece} & \textbf{$\Delta$ factors} & $t=0$ & $t=1$ & $t=2$\\\hline
$A_{4.1}$ (no $\Delta W^4$) & none & $0$ & $0$ & $0$\\
$A_{4.2}$ (with $\Delta W^4$) & $W^4$ & $0$ & $\Theta(1/N)$ & \textcolor{green!50!black}{$\Theta(1/N)$}\\
$A_{5.1}$ (no $\Delta W^4$) & $W^{3,i}$ & $0$ & $0$ & $0$\\
$A_{5.2}$ (with $\Delta W^4$) & $W^{3,i},W^4$ & $0$ & $0$ & \textcolor{green!50!black}{$\Theta(1/N)$}\\
$A_{6.1}$ (no $\Delta W^4$) & $W^{2,i}$ & $0$ & $0$ & $0$\\
$A_{6.2}$ (with $\Delta W^4$) & $W^{2,i},W^4$ & $0$ & $0$ & \textcolor{green!50!black}{$\Theta(1/N)$}\\
$E_1$ (no $\Delta W^4$) & $W^{2,i},W^{3,i}$ & $0$ & $0$ & $0$\\
$E_2$ (with $\Delta W^4$) & all three & $0$ & $0$ & \textcolor{green!50!black}{$\Theta(1/N)$}\\
\textbf{Expert total} & — & $0$ & $\Theta(1/N)$ & \textcolor{green!50!black}{$\Theta(1/N)$}\\
\end{tabular}
\end{center}

\paragraph{Router pathway.}

\begin{center}
\renewcommand{\arraystretch}{1.15}
\begin{tabular}{l|l|l|l}
\textbf{Piece} & $t=0$ & $t=1$ & $t=2$\\\hline
$(1/M)Q_t^\top[\dot\phi_{i,t}\langle h_t^{3,i}, \Delta_t W^4\rangle]_i$ \textbf{(total)} & $0$ & $\Theta(1/N)$ & \textcolor{green!50!black}{$\Theta(1/N)$}\\
$\quad (1/M)Q_0^\top[\dot\phi_{i,t}\langle (W_0^{3,1})h_t^{2,i}, \Delta_t W^4\rangle]_i$ & $0$ & $\Theta(1/N)$ & \textcolor{green!50!black}{$\Theta(1/N)$}\\
$\quad (1/M)Q_0^\top[\dot\phi_{i,t}\langle (\Delta_t W^{3,i})h_t^{2,i}, \Delta_t W^4\rangle]_i$ & $0$ & $0$ & \textcolor{green!50!black}{$\Theta(1/N)$}\\
$\quad (1/M)\Delta Q^\top[\dot\phi_{i,t}\langle (W_0^{3,1})h_t^{2,i}, \Delta_t W^4\rangle]_i$ & $0$ & $0$ & \textcolor{green!50!black}{$\Theta(1/N)$}\\
$\quad (1/M)\Delta Q^\top[\dot\phi_{i,t}\langle (\Delta_t W^{3,i})h_t^{2,i}, \Delta_t W^4\rangle]_i$ & $0$ & $0$ & \textcolor{green!50!black}{$\Theta(1/N)$}\\
\end{tabular}
\end{center}

\paragraph{Total backward.}

\begin{center}
\begin{tabular}{l|l|l|l}
\textbf{Quantity} & $t=0$ & $t=1$ & $t=2$\\\hline
$\partial f_t/\partial h_t^1$ & $0$ & $\Theta(1/N)$ & \textcolor{green!50!black}{$\Theta(1/N)$}\\
\end{tabular}
\end{center}

\subsection{Scaling derivation for Regime III}
\label{sec:heuristic_scaling_derivation_regime_iii}

In this section, we provide the heuristic scaling derivation for Regime III. Throughout we work under the proportional limit $N \asymp N_e \asymp M \asymp K \to \infty$.

\begin{definition}[$\mu$P baseline, Regime III]
    \label{def:mup_baseline_regime_iii}
    \leavevmode
    \begin{itemize}
        \item \textbf{Initialization.} All parameters are drawn \textit{independently} according to:
        \[
            W_0^{1} \sim \mathcal{N}(0, D^{-1}), \;
            Q_0 \sim \mathcal{N}(0, N^{-1}), \;
            W_0^{2,i} \sim \mathcal{N}(0, N^{-1}), \;
        \]
        \[
            W_0^{3,i} \sim \mathcal{N}(0, N_e^{-1}), \;
            W_0^{4} \sim \mathcal{N}(0, N^{-2}), \; \text{for } i \in [M].
        \]
        \item \textbf{SGD learning rates.} $\eta_1 = \eta\, N$, $\;\eta_Q = \eta$, $\;\eta_2 = \eta_3 = \eta\, M$, $\;\eta_4 = \eta\, N^{-1}$.
        \item \textbf{Adam learning rates.} $\eta_1 = \eta$, $\;\eta_Q = \eta_2 = \eta_3 = \eta_4 = \eta\, N^{-1}$.
        \item \textbf{Adam epsilon.} $\epsilon_1 = \epsilon\, N^{-1}$, $\;\epsilon_Q = \epsilon\, M^{-1}$, $\;\epsilon_2 = \epsilon_3 = \epsilon\, N^{-1} M^{-1}$, $\;\epsilon_4 = \epsilon$.
    \end{itemize}
\end{definition}

\begin{definition}[Maximally Scale-Stable Parameterization (MSSP), Regime III]
    \label{def:ssp_regime_iii}
    MSSP adopts the same per-layer initialization variances and learning rate scalings as the $\mu$P baseline (Definition~\ref{def:mup_baseline_regime_iii}), with the sole exception that expert weights are \textbf{shared across experts at initialization}: a single pair $(W_0^{2}, W_0^{3})$ is sampled and assigned to every expert, such that
    \[
        W_0^{2,i} = W_0^{2} \quad \text{and} \quad W_0^{3,i} = W_0^{3} \quad \text{for all } i = 1, \dots, M.
    \]
\end{definition}

\subsubsection{Compound aggregate lemmas in Regime~III}
\label{ssec:r3_lemmas}

The analyses in this section invoke three compound aggregate-level results stating how sums across $M$ independent experts of products and Gram products of the per-expert weight matrices behave under proportional scaling. All three are exact moment computations under Gaussian initialization, with independence across experts and from the input vector $v$ as the only inputs.

\begin{lemma}[Cross-layer product sum across experts]
\label{lem:r3_cross_layer}
Let $\{W^{2,i}\}_{i=1}^M$ have i.i.d.\ centered Gaussian entries of variance $\sigma_2^2 = 1/N$ in $\mathbb{R}^{N_e\times N}$, and let $\{W^{3,i}\}_{i=1}^M$ have i.i.d.\ centered Gaussian entries of variance $\sigma_3^2 = 1/N_e$ in $\mathbb{R}^{N\times N_e}$, with all matrices mutually independent across $i$ and across the two layers. Let $c_i\in\Theta(1)$ be approximately independent of these weights, and let $v\in\mathbb{R}^N$ have entries of coordinate scale $\Theta(\alpha)$, independent of all $W^{2,i},W^{3,i}$. Then
\[
    G \;:=\; \frac{1}{M}\sum_{i=1}^M c_i\,(W^{2,i})^\top (W^{3,i})^\top \;\in\;\mathbb{R}^{N\times N}
\]
has zero mean, entries of coordinate scale $\Theta(1/\sqrt{NM})$, and $Gv$ has entries of coordinate scale $\Theta(\alpha/\sqrt M) = \Theta(\alpha/\sqrt N)$.
\end{lemma}

\begin{proof}
$\mathbb{E}[G_{a,b}] = (1/M)\sum_i c_i\,\mathbb{E}[\sum_k W^{2,i}_{k,a}W^{3,i}_{b,k}] = 0$ by independence across the two layers. Variance:
\[
    \mathrm{Var}(G_{a,b}) = \frac{1}{M^2}\sum_i c_i^2\,\mathrm{Var}\!\Bigl(\sum_k W^{2,i}_{k,a}W^{3,i}_{b,k}\Bigr) = \frac{1}{M^2}\sum_i c_i^2\cdot N_e\sigma_2^2\sigma_3^2 = \Theta\!\Bigl(\tfrac{N_e\sigma_2^2\sigma_3^2}{M}\Bigr) = \Theta\!\Bigl(\tfrac{1}{MN}\Bigr).
\]
Cross-entry covariances $\mathbb{E}[G_{a,b_1}G_{a,b_2}]$ for $b_1\neq b_2$ vanish by independence, so
$\mathrm{Var}((Gv)_a) = \sum_b\mathrm{Var}(G_{a,b})\,v_b^2 = N\cdot\Theta(1/(MN))\cdot\alpha^2 = \Theta(\alpha^2/M)$.
\end{proof}

\begin{lemma}[Gram concentration across experts]
\label{lem:r3_gram}
Let $\{W^{a,i}\}_{i=1}^M$ have i.i.d.\ centered Gaussian entries of variance $\sigma^2$ in $\mathbb{R}^{m\times n}$, independent across $i$. Let $c_i\in\Theta(1)$ be approximately independent of the weights, with empirical mean $\bar c$. Let $G^{a,i}$ denote either Gram $(W^{a,i})^\top W^{a,i}\in\mathbb{R}^{n\times n}$ or $W^{a,i}(W^{a,i})^\top\in\mathbb{R}^{m\times m}$, and let $v$ be a vector of compatible dimension at coordinate scale $\Theta(\alpha)$, independent of all $W^{a,i}$. Then
\[
    \frac{1}{M}\sum_{i=1}^M c_i\,G^{a,i}\,v = \bar c\,(\mathrm{tr\text{-}dim}\cdot\sigma^2)\,v + r,\qquad r\text{ at coordinate scale }\Theta(\sigma^2\sqrt{mn/M}\,\alpha),
\]
where $\mathrm{tr\text{-}dim}$ is $m$ for the inner Gram $((W^{a,i})^\top W^{a,i})$ and $n$ for the outer Gram $(W^{a,i}(W^{a,i})^\top)$.

For the specific instances used in this section:
\begin{itemize}
    \item Inner Gram: $W^{2,i}\in\mathbb{R}^{N_e\times N}$ with $\sigma_2^2=1/N$ gives leading constant $m\sigma^2 = N_e/N$; $W^{3,i}\in\mathbb{R}^{N\times N_e}$ with $\sigma_3^2=1/N_e$ gives leading constant $m\sigma^2 = N/N_e$.
    \item Outer Gram: both layers give $n\sigma^2 = 1$.
\end{itemize}
All leading constants are $\Theta(1)$ under proportional scaling, with random correction $\Theta(\alpha/\sqrt N)$.
\end{lemma}

\begin{proof}
Inner Gram case: $\mathbb{E}[G_{j,k}] = (1/M)\sum_i c_i\,\mathbb{E}[\sum_a W^{a,i}_{a,j}W^{a,i}_{a,k}] = \bar c\,m\sigma^2\,\delta_{j,k}$, with off-diagonal variance $\Theta(m\sigma^4/M)$ giving coordinate scale $\Theta(\sigma^2\sqrt{m/M})$ and off-diagonal contribution to $(Gv)_j - \bar c m\sigma^2 v_j$ at coordinate scale $\Theta(\sigma^2\sqrt{mn/M}\,\alpha)$. Outer Gram case is the same statement with the roles of $m$ and $n$ swapped, equivalently obtained by applying the inner-Gram result to $W^\top$ (whose entries are i.i.d. Gaussian of the same variance with the dimension pair swapped).
\end{proof}

\subsubsection{Deriving $\mu$P in Regime III}
\label{sec:mup_r3_consolidated}

This subsection provides a self-contained derivation of the heuristic width scaling for the $\mu$P baseline (Definition~\ref{def:mup_baseline_regime_iii}) in Regime III. We invoke the compound aggregate lemmas of §\ref{ssec:r3_lemmas} for sums across independent experts of weight-product structures, and use CLT/LLN inline for single matrix-vector products and gating averages.

The principal structural consequence is the following. Although hidden weights move at $t=1$ (since $W_0^4\neq 0$ in $\mu$P), the embedding update $\Delta_1 h^1$ comes out at $\Theta(1/\sqrt N)$, not $\Theta(1)$, because both the expert and router contributions to $\partial f_0/\partial h_0^1$ cancel across independent experts via CLT. Full $\Theta(1)$ feature learning at the embedding emerges at $t=2$, when the rank-one updates $\Delta_1 W^{2,i}, \Delta_1 W^{3,i}$ inject coherent contributions ($A_5,A_6$) into the expert pathway gradient that align across experts along $(W_0^4)^\top$ and so escape the cross-$i$ cancellation. We therefore analyse three forward and three backward passes.

\paragraph{First forward pass}
\label{ssec:mup_r3_fwd1}

\[
    h_0^1=W_0^1\,x \in \Theta(1). \quad (\text{CLT};\ \sigma_1^2=1/D,\ \|x\|^2\in\Theta(D)) \tag{R3$\mu$-F1.1}
\]

\[
    \psi_0 = Q_0\,h_0^1 \in \Theta(1). \quad (\text{CLT};\ \sigma_Q^2=1/N,\ \|h_0^1\|^2\in\Theta(N)) \tag{R3$\mu$-F1.2}
\]
\[
    \phi_0 = \sigma(\psi_0) \in \Theta(1). \quad (\text{sigmoid bounded; gating assumption}) \tag{R3$\mu$-F1.3}
\]
\[
    h_0^{2,i} = W_0^{2,i}\,h_0^1 \in \Theta(1), \text{ indep.\ across } i. \quad (\text{CLT};\ \sigma_2^2=1/N,\ \|h_0^1\|^2\in\Theta(N)) \tag{R3$\mu$-F1.4}
\]
\[
    h_0^{3,i} = W_0^{3,i}\,h_0^{2,i} \in \Theta(1), \text{ indep.\ across } i. \quad (\text{CLT};\ \sigma_3^2=1/N_e,\ \|h_0^{2,i}\|^2\in\Theta(N_e)) \tag{R3$\mu$-F1.5}
\]
\[
    h_0^3 = (1/M)\sum_i\phi_{i,0}\,h_0^{3,i} \in \Theta(1/\sqrt M) = \Theta(1/\sqrt N). \quad (\text{cross-}i\text{ CLT}) \tag{R3$\mu$-F1.6}
\]
\[
    f_0 = W_0^4\,h_0^3 = \langle W_0^4, h_0^3\rangle \in \Theta(1/N). \quad (\text{CLT};\ W_0^4\text{ entries }\Theta(1/N)\text{ indep.\ of } h_0^3) \tag{R3$\mu$-F1.7}
\]

\paragraph{First backward pass and step-1 updates}
\label{ssec:mup_r3_bwd1}

\[
    \partial f_0/\partial h_0^3 = (W_0^4)^\top \in \Theta(1/N). \tag{R3$\mu$-B1.1}
\]
\[
    \partial f_0/\partial h_0^{3,i} = (\phi_{i,0}/M)(W_0^4)^\top \in \Theta(1/(MN)). \tag{R3$\mu$-B1.2}
\]
\begin{align*}
    \partial f_0/\partial h_0^{2,i} &= (\phi_{i,0}/M)(W_0^{3,i})^\top(W_0^4)^\top \in \Theta(1/(MN)). \tag{R3$\mu$-B1.3}\\
    & \quad ((W_0^{3,i})^\top(W_0^4)^\top\in\Theta(1/N)\text{ by CLT};\ \sigma_3^2=1/N_e)
\end{align*}
\begin{align*}
    \partial f_0/\partial\phi_{i,0} &= (1/M)\langle h_0^{3,i}, W_0^4\rangle \in \Theta(1/(M\sqrt N)) = \Theta(1/N^{3/2}). \tag{R3$\mu$-B1.4}\\
    & \quad (\langle h_0^{3,i},W_0^4\rangle\in\Theta(1/\sqrt N)\text{ by CLT})
\end{align*}
\begin{align*}
    \Bigl(\tfrac{\partial f_0}{\partial h_0^1}\Bigr)_{\!\!\mathrm{exp}} &= \tfrac{1}{M}\textstyle\sum_i\phi_{i,0}\,(W_0^{2,i})^\top(W_0^{3,i})^\top(W_0^4)^\top\\
    &\in \Theta\bigl(\sigma_2\sigma_3\sqrt{N_e}\,\|(W_0^4)^\top\|\,/\sqrt M\bigr) \text{ random direction} \quad (\text{Lem.~\ref{lem:r3_cross_layer}; cross-$i$ CLT})\\
    &= \Theta(1/\sqrt N\cdot 1/\sqrt{N_e}\cdot \sqrt{N_e}\cdot 1/\sqrt N\cdot 1/\sqrt M) = \Theta(1/N^{3/2}). \tag{R3$\mu$-B1.5}
\end{align*}
\begin{align*}
    \Bigl(\tfrac{\partial f_0}{\partial h_0^1}\Bigr)_{\!\!\mathrm{router}} &= \tfrac{1}{M}Q_0^\top v, \qquad v_i := \dot\phi_{i,0}\langle W_0^4,h_0^{3,i}\rangle\\
    &\in \Theta\bigl(\sigma_Q\|v\|/M\bigr) \text{ random direction} \quad (Q_0\text{ indep.\ of }v;\ \text{cross-layer CLT})\\
    &= \Theta(1/\sqrt N\cdot 1\cdot 1/N) = \Theta(1/N^{3/2}) \tag{R3$\mu$-B1.6}\\
    & \quad (\sigma_Q^2=1/N;\ v_i\in\Theta(1/\sqrt N)\text{ random across }i\Rightarrow\|v\|^2\in\Theta(1)).
\end{align*}
\[
    \partial f_0/\partial h_0^1 = (\partial f_0/\partial h_0^1)_\mathrm{exp} + (\partial f_0/\partial h_0^1)_\mathrm{router} \in \Theta(1/N^{3/2}). \tag{R3$\mu$-B1.7}
\]

\paragraph{Step-1 parameter updates.}

\[
    \Delta_1 W^4 = -\tfrac{\chi_0}{N}(h_0^3)^\top \in \Theta(1/N^{3/2}). \quad (h_0^3\in\Theta(1/\sqrt N)\text{ by (R3$\mu$-F1.6)}) \tag{R3$\mu$-U1.4}
\]
\[
    \Delta_1 W^{3,i} = -\eta\chi_0\phi_{i,0}\,(W_0^4)^\top(h_0^{2,i})^\top \in \Theta(1/N). \quad (\eta_3=\eta M) \tag{R3$\mu$-U1.3}
\]
\begin{align*}
    \Delta_1 W^{2,i} &= -\eta\chi_0\phi_{i,0}\,(W_0^{3,i})^\top(W_0^4)^\top(h_0^1)^\top \in \Theta(1/N). \tag{R3$\mu$-U1.2}\\
    & \quad ((W_0^{3,i})^\top(W_0^4)^\top\in\Theta(1/N)\text{ by (R3$\mu$-B1.3)})
\end{align*}
\begin{align*}
    \Delta_1 W^1 &= -\eta N\chi_0\,(\partial f_0/\partial h_0^1)\,x^\top \in \Theta(1/\sqrt N). \tag{R3$\mu$-U1.1a}\\
    & \quad (\eta_1=\eta N,\ \partial f_0/\partial h_0^1\in\Theta(1/N^{3/2})\text{ by (R3$\mu$-B1.7)})
\end{align*}
\[
    \Delta_1 h^1 = \Delta_1 W^1\,x \in \Theta(1/\sqrt N), \text{ sub-leading vs } h_0^1\in\Theta(1). \tag{R3$\mu$-U1.1b}
\]
\begin{align*}
    \Delta_1 Q &= -\eta_Q\chi_0\,\mathrm{diag}(\dot\phi_0)\,(\partial f_0/\partial\phi)\,(h_0^1)^\top \in \Theta(1/N^{3/2}). \tag{R3$\mu$-U1.Q}\\
    & \quad (\partial f_0/\partial\phi\in\Theta(1/N^{3/2})\text{ by (R3$\mu$-B1.4)};\ \eta_Q=\eta)
\end{align*}

\paragraph{Second forward pass}
\label{ssec:mup_r3_fwd2}

\[
    h_1^1 = h_0^1 + \Delta_1 h^1 \in \Theta(1). \quad (h_0^1\in\Theta(1)\text{ dominates }\Delta_1 h^1\in\Theta(1/\sqrt N)\text{ by (R3$\mu$-U1.1b)}) \tag{R3$\mu$-F2.1}
\]
\begin{align*}
    \psi_1 &= \psi_0 + \Delta_1\psi \in \Theta(1),\quad \phi_1\in\Theta(1). \tag{R3$\mu$-F2.2--R3$\mu$-F2.3}\\
    & \quad (\Delta_1\psi\in\Theta(1/\sqrt N)\text{ from }Q_0\Delta_1 h^1\text{ and }\Delta_1 Q\,h_0^1\text{; scales by (R3$\mu$-U1.1b), (R3$\mu$-U1.Q)})
\end{align*}

\noindent (R3$\mu$-F2.4) $h_1^{2,i} = W_1^{2,i}\,h_1^1$, four-piece decomposition:
\[
    h_1^{2,i} = h_0^{2,i} + W_0^{2,i}\Delta_1 h^1 + \Delta_1 W^{2,i}\,h_0^1 + \Delta_1 W^{2,i}\,\Delta_1 h^1. \tag{R3$\mu$-F2.4}
\]
\[
    \text{init: } W_0^{2,i}h_0^1 = h_0^{2,i} \in \Theta(1). \quad (\text{R3$\mu$-F1.4}) \tag{R3$\mu$-F2.4a}
\]
Op-norm of iid Gaussian $W_0^{2,i}\; \Delta_1 h^1\in\Theta(1/\sqrt N)$ by (R3$\mu$-U1.1b):
\[
    \text{prop: } W_0^{2,i}\Delta_1 h^1 \in \Theta(1/\sqrt N). \tag{R3$\mu$-F2.4b}
\]
\begin{align*}
    \text{eff: } \Delta_1 W^{2,i}\,h_0^1 &= -\eta\chi_0\phi_{i,0}(W_0^{3,i})^\top(W_0^4)^\top\,\|h_0^1\|^2 \in \Theta(1). \tag{R3$\mu$-F2.4c}\\
    & \quad ((W_0^{3,i})^\top(W_0^4)^\top\in\Theta(1/N)\text{ by (R3$\mu$-B1.3)};\ \|h_0^1\|^2\in\Theta(N))
\end{align*}
\begin{align*}
    \text{cross: } \Delta_1 W^{2,i}\,\Delta_1 h^1 &= -\eta\chi_0\phi_{i,0}(W_0^{3,i})^\top(W_0^4)^\top\langle h_0^1,\Delta_1 h^1\rangle \in \Theta(1/N). \tag{R3$\mu$-F2.4d}\\
    & \quad ((W_0^{3,i})^\top(W_0^4)^\top\in\Theta(1/N)\text{ by (R3$\mu$-B1.3)};\ \langle h_0^1,\Delta_1 h^1\rangle\in\Theta(1)\text{ by CLT})
\end{align*}

\noindent (R3$\mu$-F2.5) $h_1^{3,i} = W_1^{3,i}\,h_1^{2,i}$, four-piece decomposition:
\[
    h_1^{3,i} = h_0^{3,i} + W_0^{3,i}\Delta_1 h^{2,i} + \Delta_1 W^{3,i}\,h_0^{2,i} + \Delta_1 W^{3,i}\,\Delta_1 h^{2,i}. \tag{R3$\mu$-F2.5}
\]
\[
    \text{init: } h_0^{3,i} \in \Theta(1). \quad (\text{R3$\mu$-F1.5}) \tag{R3$\mu$-F2.5a}
\]
\[
    \text{prop: } W_0^{3,i}\Delta_1 h^{2,i} \in \Theta(1). \quad (\text{op-norm of iid Gaussian } W_0^{3,i};\ \Delta_1 h^{2,i}\in\Theta(1)) \tag{R3$\mu$-F2.5b}
\]
\begin{align*}
    \text{eff: } \Delta_1 W^{3,i}\,h_0^{2,i} &= -\eta\chi_0\phi_{i,0}(W_0^4)^\top\,\|h_0^{2,i}\|^2 \in \Theta(1). \tag{R3$\mu$-F2.5c}\\
    & \quad (\|h_0^{2,i}\|^2\in\Theta(N_e);\ (W_0^4)^\top\in\Theta(1/N))
\end{align*}
\[
    \text{cross: } \Delta_1 W^{3,i}\,\Delta_1 h^{2,i} \text{ tracked as part of } D \text{ in (R3$\mu$-F2.6)}. \tag{R3$\mu$-F2.5d}
\]

\noindent (R3$\mu$-F2.6) $h_1^3 = A_1 + A_{2,1} + A_{2,2} + A_3 + D$, where
\begin{align*}
    A_1 &:= \tfrac{1}{M}\textstyle\sum_i\phi_{i,1}\,h_0^{3,i}, &
    A_{2,1} &:= \tfrac{1}{M}\textstyle\sum_i\phi_{i,1}\,W_0^{3,i}W_0^{2,i}\,\Delta_1 h^1,\\
    A_{2,2} &:= \tfrac{1}{M}\textstyle\sum_i\phi_{i,1}\,W_0^{3,i}\,\Delta_1 W^{2,i}\,h_0^1, &
    A_3 &:= \tfrac{1}{M}\textstyle\sum_i\phi_{i,1}\,\Delta_1 W^{3,i}\,h_0^{2,i},\\
    D &:= \tfrac{1}{M}\textstyle\sum_i\phi_{i,1}\,\Delta_1 W^{3,i}\,\Delta_1 h^{2,i}.
\end{align*}
\[
    A_1 \in \Theta(1/\sqrt N). \quad (\text{cross-}i\text{ CLT on }\Theta(1)\text{-entry independent vectors}) \tag{R3$\mu$-F2.6a}
\]
\begin{align*}
    A_{2,1} &\in \Theta(1/(\sqrt N\,\sqrt M)) = \Theta(1/N). \tag{R3$\mu$-F2.6b}\\
    & \quad (\text{Lemma~\ref{lem:r3_cross_layer}};\ \Delta_1 h^1\text{ entries }\Theta(1/\sqrt N)\text{ by (R3$\mu$-U1.1b)})
\end{align*}
\begin{align*}
    A_{2,2} &= -\eta\chi_0\,\|h_0^1\|^2\,\tfrac{1}{M}\textstyle\sum_i\phi_{i,1}\phi_{i,0}\,W_0^{3,i}(W_0^{3,i})^\top(W_0^4)^\top\\
    &\in \Theta\bigl(\|h_0^1\|^2\,\tfrac{1}{M}\textstyle\sum_i\phi_{i,1}\phi_{i,0}\bigr)\cdot(W_0^4)^\top \text{ along }(W_0^4)^\top \quad (\text{Lem.~\ref{lem:r3_gram}})\\
    &= \Theta(N\cdot 1\cdot 1/N) = \Theta(1) \text{ along }(W_0^4)^\top \tag{R3$\mu$-F2.6c}\\
    & \quad (\|h_0^1\|^2\in\Theta(N);\ \text{LLN: }\tfrac{1}{M}\textstyle\sum\phi\phi\in\Theta(1);\ (W_0^4)^\top\in\Theta(1/N))
\end{align*}
\begin{align*}
    A_3 &= -\eta\chi_0\,(W_0^4)^\top\,\Bigl(\tfrac{1}{M}\textstyle\sum_i\phi_{i,1}\phi_{i,0}\,\|h_0^{2,i}\|^2\Bigr) \in \Theta(1) \text{ along } (W_0^4)^\top. \tag{R3$\mu$-F2.6d}\\
    & \quad (\|h_0^{2,i}\|^2\in\Theta(N_e);\ \text{LLN})
\end{align*}
\begin{align*}
    D &= \tfrac{1}{M}\textstyle\sum_i \phi_{i,1}\,\Delta_1 W^{3,i}\,\Delta_1 h^{2,i} = -\eta\chi_0\,(W_0^4)^\top\,\tfrac{1}{M}\textstyle\sum_i\phi_{i,1}\phi_{i,0}\,\langle h_0^{2,i},\Delta_1 h^{2,i}\rangle\\
    &\in \Theta\bigl(\tfrac{1}{M}\textstyle\sum_i\phi_{i,1}\phi_{i,0}\,\langle h_0^{2,i},\Delta_1 h^{2,i}\rangle\bigr)\cdot(W_0^4)^\top \text{ along }(W_0^4)^\top \\
    & \quad (\text{cross-$i$ CLT: }\langle\cdot,\cdot\rangle\text{ random across }i,\text{ mean }0)\\
    &= \Theta(\sqrt N/\sqrt M\cdot 1/N) = \Theta(1/N) \text{ along }(W_0^4)^\top \tag{R3$\mu$-F2.6e}\\
    & \quad (\langle h_0^{2,i},\Delta_1 h^{2,i}\rangle\in\Theta(\sqrt N);\ (W_0^4)^\top\in\Theta(1/N);\ M\in\Theta(N))
\end{align*}
\[
    h_1^3 \in \Theta(1) \text{ along } (W_0^4)^\top, \text{ from } A_{2,2}+A_3. \tag{R3$\mu$-F2.6}
\]

\begin{align*}
    W_0^4\,h_1^3 &= \langle W_0^4, A_{2,2}+A_3\rangle + \cdots \in \Theta(1). \tag{R3$\mu$-F2.7a}\\
    & \quad (A_{2,2}, A_3\text{ aligned along }(W_0^4)^\top;\ \langle W_0^4, A_3\rangle = -\eta\chi_0\|W_0^4\|^2\bar c\,N_e\in\Theta(1))
\end{align*}
\[
    \Delta_1 W^4\,h_1^3 = -(\chi_0/N)\langle h_0^3, h_1^3\rangle \in \Theta(1/N). \tag{R3$\mu$-F2.7b}
\]
\[
    f_1 = W_1^4\,h_1^3 \in \Theta(1). \tag{R3$\mu$-F2.7}
\]

\paragraph{Second backward pass and step-2 updates}
\label{ssec:mup_r3_bwd2}

\begin{align*}
    \partial f_1/\partial h_1^3 &= (W_1^4)^\top = (W_0^4)^\top + (\Delta_1 W^4)^\top \in \Theta(1/N). \tag{R3$\mu$-B2.1}\\
    & \quad (W_0^4\in\Theta(1/N)\text{ dominates }\Delta_1 W^4\in\Theta(1/N^{3/2})\text{ by (R3$\mu$-U1.4)})
\end{align*}
\[
    \partial f_1/\partial h_1^{3,i} = (\phi_{i,1}/M)(W_1^4)^\top \in \Theta(1/(MN)). \tag{R3$\mu$-B2.2}
\]
\begin{align*}
    \partial f_1/\partial h_1^{2,i} &= (\phi_{i,1}/M)(W_1^{3,i})^\top(W_1^4)^\top \in \Theta(1/(MN)). \tag{R3$\mu$-B2.3}\\
    & \quad ((W_0^{3,i})^\top(W_0^4)^\top\in\Theta(1/N)\text{ by (R3$\mu$-B1.3)};\ (\Delta_1 W^{3,i})^\top W_0^4\in\Theta(1/N);\\
    & \qquad\quad \phi_{i,1}\in\Theta(1);\ M\in\Theta(N))
\end{align*}
\begin{align*}
    \partial f_1/\partial\phi_{i,1} &= (1/M)\langle h_1^{3,i}, W_1^4\rangle \in \Theta(1/M). \tag{R3$\mu$-B2.4}\\
    & \quad (\langle h_1^{3,i},W_0^4\rangle = -\eta\chi_0\phi_{i,0}\|h_0^{2,i}\|^2\|W_0^4\|^2\in\Theta(1)\text{ via eff.\ term of (R3$\mu$-F2.5)};\\
    & \qquad\quad \|h_0^{2,i}\|^2\in\Theta(N_e);\ \|W_0^4\|^2\in\Theta(1/N))
\end{align*}

\noindent (R3$\mu$-B2.5) $(\partial f_1/\partial h_1^1)_\mathrm{exp} = A_4+A_5+A_6+E$, where
\begin{align*}
    A_4 &:= \tfrac{1}{M}\textstyle\sum_i\phi_{i,1}(W_0^{2,i})^\top(W_0^{3,i})^\top(W_1^4)^\top,\\
    A_5 &:= \tfrac{1}{M}\textstyle\sum_i\phi_{i,1}(W_0^{2,i})^\top(\Delta_1 W^{3,i})^\top(W_1^4)^\top,\\
    A_6 &:= \tfrac{1}{M}\textstyle\sum_i\phi_{i,1}(\Delta_1 W^{2,i})^\top(W_0^{3,i})^\top(W_1^4)^\top,\\
    E &:= \tfrac{1}{M}\textstyle\sum_i\phi_{i,1}(\Delta_1 W^{2,i})^\top(\Delta_1 W^{3,i})^\top(W_1^4)^\top.
\end{align*}
\[
    A_4 \in \Theta(1/N^{3/2}). \quad (\text{Lemma~\ref{lem:r3_cross_layer}};\ v=(W_1^4)^\top\in\Theta(1/N)) \tag{R3$\mu$-B2.5a}
\]
\begin{align*}
    A_5 &= -\eta\chi_0\,\|W_0^4\|^2\,\tfrac{1}{M}\textstyle\sum_i\phi_{i,1}\phi_{i,0}\,(W_0^{2,i})^\top W_0^{2,i}\,h_0^1\\
    &\in \Theta\bigl(\|W_0^4\|^2\,\tfrac{1}{M}\textstyle\sum_i\phi_{i,1}\phi_{i,0}\bigr)\cdot h_0^1 \text{ along }h_0^1 \quad (\text{Lem.~\ref{lem:r3_gram}})\\
    &= \Theta(1/N\cdot 1\cdot 1) = \Theta(1/N) \text{ along }h_0^1 \tag{R3$\mu$-B2.5b}\\
    & \quad (\|W_0^4\|^2\in\Theta(1/N);\ \text{LLN: }\tfrac{1}{M}\textstyle\sum\phi\phi\in\Theta(1);\ h_0^1\in\Theta(1))
\end{align*}
\begin{align*}
    A_6 &= -\eta\chi_0\,h_0^1\,\tfrac{1}{M}\textstyle\sum_i\phi_{i,1}\phi_{i,0}\,W_0^4 W_0^{3,i}(W_0^{3,i})^\top(W_1^4)^\top\\
    &\in \Theta\bigl(\tfrac{1}{M}\textstyle\sum_i\phi_{i,1}\phi_{i,0}\,\|W_0^4\|^2\bigr)\cdot h_0^1 \text{ along }h_0^1 \\
    & \quad (\text{Lem.~\ref{lem:r3_gram}: }W_0^{3,i}(W_0^{3,i})^\top(W_1^4)^\top\to(W_1^4)^\top;\ \langle W_0^4,(W_1^4)^\top\rangle\to\|W_0^4\|^2)\\
    &= \Theta(1\cdot 1/N) = \Theta(1/N) \text{ along }h_0^1 \tag{R3$\mu$-B2.5c}\\
    & \quad (\|W_0^4\|^2\in\Theta(1/N);\ \text{LLN};\ h_0^1\in\Theta(1))
\end{align*}
\begin{align*}
    E &= \tfrac{1}{M}\textstyle\sum_i\phi_{i,1}(\Delta_1 W^{2,i})^\top(\Delta_1 W^{3,i})^\top(W_1^4)^\top\\
    &= \eta^2\chi_0^2\,h_0^1\,\tfrac{1}{M}\textstyle\sum_i\phi_{i,1}\phi_{i,0}^2\,\langle W_0^4,h_0^{3,i}\rangle\,(h_0^3)^\top(W_1^4)^\top \quad (\text{substituting (R3$\mu$-U1.2),(R3$\mu$-U1.3)})\\
    &\in \Theta\bigl(\tfrac{1}{M}\textstyle\sum_i\langle W_0^4,h_0^{3,i}\rangle\bigr)\cdot h_0^1 \text{ along }h_0^1 \quad (\text{cross-$i$ CLT: }\langle W_0^4,h_0^{3,i}\rangle\text{ random across }i,\text{ mean }0)\\
    &= \Theta(1/\sqrt N\cdot 1/\sqrt M\cdot \|h_0^3\|^2/N) = \Theta(1/N^2) \text{ along }h_0^1 \tag{R3$\mu$-B2.5d}\\
    & \quad (\langle W_0^4,h_0^{3,i}\rangle\in\Theta(1/\sqrt N)\text{ by (R3$\mu$-B1.4)};\ \|h_0^3\|^2\in\Theta(1);\ M\in\Theta(N))
\end{align*}
\[
    (\partial f_1/\partial h_1^1)_\mathrm{exp} \in \Theta(1/N) \text{ along } h_0^1. \tag{R3$\mu$-B2.5}
\]
\begin{align*}
    (\partial f_1/\partial h_1^1)_\mathrm{router} &= \tfrac{1}{M}Q_0^\top v, \qquad v_i := \dot\phi_{i,1}\langle h_1^{3,i},W_1^4\rangle\\
    &\in \Theta\bigl(\sigma_Q\|v\|/M\bigr) \text{ random direction} \\
    & \quad (Q_0\text{ indep.\ of }v;\ \text{cross-layer CLT};\ \Delta_1 Q\text{ sub-leading by (R3$\mu$-U1.Q)})\\
    &= \Theta(1/\sqrt N\cdot \sqrt M\cdot 1/M) = \Theta(1/\sqrt{MN}) = \Theta(1/N) \tag{R3$\mu$-B2.6}\\
    & \quad (\sigma_Q^2=1/N;\ v_i\in\Theta(1)\text{ via (R3$\mu$-B2.4)}\Rightarrow\|v\|^2\in\Theta(M);\ M\in\Theta(N))
\end{align*}
\[
    \partial f_1/\partial h_1^1 = (\partial f_1/\partial h_1^1)_\mathrm{exp} + (\partial f_1/\partial h_1^1)_\mathrm{router} \in \Theta(1/N), \tag{R3$\mu$-B2.7}
\]
the proper feature-learning rate, an order of magnitude larger than $\partial f_0/\partial h_0^1$.

\paragraph{Step-2 parameter updates.}
\[
    \Delta_2 W^4 = -\tfrac{\chi_1}{N}(h_1^3)^\top \in \Theta(1/N). \quad (h_1^3\in\Theta(1)\text{ by (R3$\mu$-F2.6)}) \tag{R3$\mu$-U2.4}
\]
\begin{align*}
    \Delta_2 W^{3,i},\ \Delta_2 W^{2,i} &\in \Theta(1/N). \tag{R3$\mu$-U2.3, R3$\mu$-U2.2}\\
    & \quad (\text{same structure as (R3$\mu$-U1.3), (R3$\mu$-U1.2) with cumulative weights})
\end{align*}
\begin{align*}
    \Delta_2 W^1 &= -\eta_1\chi_1(\partial f_1/\partial h_1^1)\,x^\top \in \Theta(1). \tag{R3$\mu$-U2.1a}\\
    & \quad (\eta_1=N,\ \partial f_1/\partial h_1^1\in\Theta(1/N)\text{ by (R3$\mu$-B2.7)})
\end{align*}
\[
    \Delta_2 h^1 = \Delta_2 W^1\,x \in \Theta(1) \text{ aligned along } h_0^1. \quad (\text{embedding feature-learns at }t=2) \tag{R3$\mu$-U2.1b}
\]
\[
    \Delta_2 Q \in \Theta(1/N). \quad (\partial f_1/\partial\phi\in\Theta(1/N)\text{ by (R3$\mu$-B2.4)}) \tag{R3$\mu$-U2.Q}
\]

\paragraph{Third forward pass}
\label{ssec:mup_r3_fwd3}

We compute activations at $\theta^{(2)}$. The cumulative embedding change $\Delta h^1:=h_2^1-h_0^1 = \Delta_1 h^1 + \Delta_2 h^1$, with $\Delta_1 h^1\in\Theta(1/\sqrt N)$ by (R3$\mu$-U1.1b) and $\Delta_2 h^1\in\Theta(1)$ aligned along $h_0^1$ by (R3$\mu$-U2.1b).

\[
    h_2^1 = h_0^1 + \Delta h^1 \in \Theta(1). \tag{R3$\mu$-F3.1}
\]

\noindent (R3$\mu$-F3.2) $h_2^{2,i} = W_2^{2,i}h_2^1$, four-piece decomposition with cumulative $\Delta W^{2,i}$:
\[
    \text{init: } W_0^{2,i}h_0^1 = h_0^{2,i} \in \Theta(1). \quad (\text{R3$\mu$-F1.4}) \tag{R3$\mu$-F3.2a}
\]
\[
    \text{prop: } W_0^{2,i}\Delta h^1 \in \Theta(1). \quad (\text{op-norm of iid Gaussian } W_0^{2,i};\ \Delta h^1\in\Theta(1)\text{ by (R3$\mu$-U2.1b)}) \tag{R3$\mu$-F3.2b}
\]
\begin{align*}
    \text{eff: } \Delta_t W^{2,i}h_0^1 &= -\eta\chi\phi(W_{t-1}^{3,i})^\top(W_{t-1}^4)^\top\langle h_{t-1}^1, h_0^1\rangle \in \Theta(1). \tag{R3$\mu$-F3.2c}\\
    & \quad ((W_{t-1}^{3,i})^\top(W_{t-1}^4)^\top\in\Theta(1/N)\text{ by (R3$\mu$-B1.3)/(R3$\mu$-B2.3)};\\
    & \qquad\quad \langle h_{t-1}^1,h_0^1\rangle\in\Theta(N)\text{ coherent via (R3$\mu$-U2.1b)})
\end{align*}
\begin{align*}
    \text{cross: } \Delta_t W^{2,i}\Delta h^1 &= -\eta\chi\phi(W_{t-1}^{3,i})^\top(W_{t-1}^4)^\top\langle h_{t-1}^1, \Delta h^1\rangle \in \Theta(1). \tag{R3$\mu$-F3.2d}\\
    & \quad ((W_{t-1}^{3,i})^\top(W_{t-1}^4)^\top\in\Theta(1/N)\text{ by (R3$\mu$-B1.3)/(R3$\mu$-B2.3)};\\
    & \qquad\quad \langle h_{t-1}^1,\Delta h^1\rangle\in\Theta(N)\text{ coherent via (R3$\mu$-U2.1b)})
\end{align*}
\[
    h_2^{2,i} \in \Theta(1). \tag{R3$\mu$-F3.2}
\]

\noindent (R3$\mu$-F3.3) $h_2^{3,i} = W_2^{3,i}h_2^{2,i}$, four-piece decomposition:
\[
    \text{init: } W_0^{3,i}h_0^{2,i} = h_0^{3,i} \in \Theta(1). \quad (\text{R3$\mu$-F1.5}) \tag{R3$\mu$-F3.3a}
\]
\[
    \text{prop: } W_0^{3,i}\Delta h^{2,i} \in \Theta(1). \quad (\text{op-norm of iid Gaussian } W_0^{3,i};\ \Delta h^{2,i}\in\Theta(1)) \tag{R3$\mu$-F3.3b}
\]
\begin{align*}
    \text{eff: } \Delta_t W^{3,i}h_0^{2,i} &= -\eta\chi\phi(W_{t-1}^4)^\top\langle h_{t-1}^{2,i}, h_0^{2,i}\rangle \in \Theta(1). \tag{R3$\mu$-F3.3c}\\
    & \quad ((W_{t-1}^4)^\top\in\Theta(1/N);\ \langle h_{t-1}^{2,i},h_0^{2,i}\rangle\in\Theta(N_e)\text{ coherent})
\end{align*}
\[
    \text{cross: } \Delta W^{3,i}\Delta h^{2,i} \text{ tracked as part of } D' \text{ in (R3$\mu$-F3.4)}. \tag{R3$\mu$-F3.3d}
\]
\[
    h_2^{3,i} \in \Theta(1). \tag{R3$\mu$-F3.3}
\]

\noindent (R3$\mu$-F3.4) $h_2^3 = A_1' + A_{2,1}' + A_{2,2}' + A_3' + D'$:
\[
    A_1' = (1/M)\textstyle\sum_i\phi_{i,2}h_0^{3,i} \in \Theta(1/\sqrt N). \quad (\text{cross-}i\text{ CLT}) \tag{R3$\mu$-F3.4a}
\]
\[
    A_{2,1}' \in \Theta(1/\sqrt M) = \Theta(1/\sqrt N). \quad (\text{Lemma~\ref{lem:r3_cross_layer}};\ \Delta h^1\in\Theta(1)\text{ by (R3$\mu$-U2.1b)}) \tag{R3$\mu$-F3.4b}
\]
\begin{align*}
    A_{2,2}' &= -\eta\chi_1\,\langle h_1^1, h_0^1\rangle\,\tfrac{1}{M}\textstyle\sum_i\phi_{i,2}\phi_{i,1}\,W_0^{3,i}(W_1^{3,i})^\top(W_1^4)^\top\\
    &\in \Theta\bigl(\langle h_1^1,h_0^1\rangle\,\tfrac{1}{M}\textstyle\sum_i\phi_{i,2}\phi_{i,1}\bigr)\cdot(W_0^4)^\top \text{ along }(W_0^4)^\top \quad (\text{Lem.~\ref{lem:r3_gram}})\\
    &= \Theta(N\cdot 1\cdot 1/N) = \Theta(1) \text{ along }(W_0^4)^\top \\
    &\quad (\langle h_1^1,h_0^1\rangle\in\Theta(N)\text{ coherent};\ \text{LLN};\ (W_0^4)^\top\in\Theta(1/N)). \tag{R3$\mu$-F3.4c}
\end{align*}
\[
    A_3' \in \Theta(1) \text{ along } (W_0^4)^\top. \quad (\Delta_2 W^{3,i}h_0^{2,i}\in\Theta(1)\text{ along }(W_1^4)^\top;\ \text{LLN}) \tag{R3$\mu$-F3.4d}
\]
\begin{align*}
    D' &\in \Theta(1) \text{ along } (W_0^4)^\top. \tag{R3$\mu$-F3.4e}\\
    & \quad (\langle h_1^{2,i},\Delta h^{2,i}\rangle\in\Theta(N_e)\text{ coherent via alignment chain through (R3$\mu$-F3.2)};\ \text{LLN})
\end{align*}
\[
    h_2^3 \in \Theta(1) \text{ along } (W_0^4)^\top. \tag{R3$\mu$-F3.4}
\]

\noindent (R3$\mu$-F3.5) $f_2 = W_2^4 h_2^3$, with $W_2^4 = W_0^4 + \Delta_1 W^4 + \Delta_2 W^4$:
\[
    W_0^4\,h_2^3 \in \Theta(1). \quad (A_{2,2}'+A_3'+D' = \kappa(W_0^4)^\top\text{ with }\kappa\in\Theta(N);\ \kappa\|W_0^4\|^2\in\Theta(1)) \tag{R3$\mu$-F3.5a}
\]
\begin{align*}
    \Delta_2 W^4\,h_2^3 &= -(\chi_1/N)\langle h_1^3, h_2^3\rangle \in \Theta(1). \tag{R3$\mu$-F3.5b}\\
    & \quad (\langle h_1^3, h_2^3\rangle\in\Theta(N)\text{ via aligned components along }(W_0^4)^\top;\\
    & \qquad\quad h_1^3\in\Theta(1)\text{ along }(W_0^4)^\top\text{ by (R3$\mu$-F2.6)})
\end{align*}
\[
    \Delta_1 W^4\,h_2^3 \in \Theta(1/\sqrt N). \quad (\Delta_1 W^4\in\Theta(1/N^{3/2})\text{ sub-leading by (R3$\mu$-U1.4)}) \tag{R3$\mu$-F3.5c}
\]
\[
    f_2 \in \Theta(1). \tag{R3$\mu$-F3.5}
\]

\paragraph{Third backward pass}
\label{ssec:mup_r3_bwd3}

We compute gradients at $\theta^{(2)}$, with $W_2^4 = W_0^4 + \Delta_1 W^4 + \Delta_2 W^4$ where $W_0^4\in\Theta(1/N)$, $\Delta_1 W^4\in\Theta(1/N^{3/2})$ by (R3$\mu$-U1.4), and $\Delta_2 W^4 = -(\chi_1/N)(h_1^3)^\top\in\Theta(1/N)$ aligned along $(h_1^3)^\top$ by (R3$\mu$-U2.4).

\begin{align*}
    \partial f_2/\partial h_2^3 &= (W_2^4)^\top \in \Theta(1/N). \tag{R3$\mu$-B3.1}\\
    & \quad (W_0^4\in\Theta(1/N)\text{ and }\Delta_2 W^4\in\Theta(1/N)\text{ by (R3$\mu$-U2.4) both contribute};\\
    & \qquad\quad \Delta_1 W^4\in\Theta(1/N^{3/2})\text{ by (R3$\mu$-U1.4) sub-leading})
\end{align*}
\[
    \partial f_2/\partial h_2^{3,i} = (\phi_{i,2}/M)(W_2^4)^\top \in \Theta(1/(MN)). \quad (\phi_{i,2}\in\Theta(1)) \tag{R3$\mu$-B3.2}
\]
\begin{align*}
    \partial f_2/\partial h_2^{2,i} &= (\phi_{i,2}/M)(W_2^{3,i})^\top(W_2^4)^\top \in \Theta(1/(MN)). \tag{R3$\mu$-B3.3}\\
    & \quad ((W_0^{3,i})^\top(W_0^4)^\top\in\Theta(1/N)\text{ via (R3$\mu$-B1.3)};\\
    & \qquad\quad (\Delta W^{3,i})^\top(W_2^4)^\top\in\Theta(1/N)\text{ via Lem.~\ref{lem:r3_gram} on aligned components})
\end{align*}
\begin{align*}
    \partial f_2/\partial \phi_{i,2} &= (1/M)\langle h_2^{3,i}, W_2^4\rangle \in \Theta(1/M). \tag{R3$\mu$-B3.4}\\
    & \quad (\langle h_2^{3,i},W_2^4\rangle\in\Theta(1)\text{ via the }(W_0^4)^\top\text{-aligned effective}\\
    & \qquad\quad \text{and propagating components of (R3$\mu$-F3.3)})
\end{align*}

\noindent (R3$\mu$-B3.5) Expert contribution. The chain through the experts gives
\[
    \Bigl(\frac{\partial f_2}{\partial h_2^1}\Bigr)_{\!\!\mathrm{exp}} = \tfrac{1}{M}\textstyle\sum_i\phi_{i,2}(W_2^{2,i})^\top(W_2^{3,i})^\top(W_2^4)^\top.
\]
Expand $W_2^{a,i}=W_0^{a,i}+\Delta W^{a,i}$ with cumulative update $\Delta W^{a,i}=\Delta_1 W^{a,i}+\Delta_2 W^{a,i}$. The four-term decomposition is:
\begin{align*}
    A_4'' &:= \tfrac{1}{M}\textstyle\sum_i\phi_{i,2}(W_0^{2,i})^\top(W_0^{3,i})^\top(W_2^4)^\top,\\
    A_5'' &:= \tfrac{1}{M}\textstyle\sum_i\phi_{i,2}(W_0^{2,i})^\top(\Delta W^{3,i})^\top(W_2^4)^\top,\\
    A_6'' &:= \tfrac{1}{M}\textstyle\sum_i\phi_{i,2}(\Delta W^{2,i})^\top(W_0^{3,i})^\top(W_2^4)^\top,\\
    E''   &:= \tfrac{1}{M}\textstyle\sum_i\phi_{i,2}(\Delta W^{2,i})^\top(\Delta W^{3,i})^\top(W_2^4)^\top.
\end{align*}

\[
    A_4'' \in \Theta(1/N^{3/2}). \quad (\text{Lemma~\ref{lem:r3_cross_layer}};\ v=(W_2^4)^\top\in\Theta(1/N)) \tag{R3$\mu$-B3.5a}
\]
\begin{align*}
    A_5'' &= -\eta\chi\,\|W_0^4\|^2\,\tfrac{1}{M}\textstyle\sum_i\phi_{i,2}\phi\,(W_0^{2,i})^\top W_0^{2,i}\,h_0^1 + \cdots\\
    &\in \Theta\bigl(\|W_0^4\|^2\,\tfrac{1}{M}\textstyle\sum_i\phi_{i,2}\phi\bigr)\cdot h_0^1 \text{ along }h_0^1 \\
    & \quad (\text{Lem.~\ref{lem:r3_gram}};\ \Delta_1\text{ and }\Delta_2\text{ contributions share leading direction})\\
    &= \Theta(1/N\cdot 1\cdot 1) = \Theta(1/N) \text{ along }h_0^1 \tag{R3$\mu$-B3.5b}\\
    & \quad (\|W_0^4\|^2\in\Theta(1/N);\ \text{LLN};\ h_0^1\in\Theta(1))
\end{align*}
\begin{align*}
    A_6'' &\in \Theta(1/N) \text{ along } h_0^1. \tag{R3$\mu$-B3.5c}\\
    & \quad (\text{Lemma~\ref{lem:r3_gram}};\ \text{LLN};\ (\Delta W^{2,i})^\top\text{ rank-1 with leading direction }h_0^1\otimes(W_0^4 W_0^{3,i}))
\end{align*}
\begin{align*}
    E'' &= \eta^2\chi^2\,h_0^1\,\tfrac{1}{M}\textstyle\sum_i\phi_{i,2}\phi^2\,\langle W_t^4, h_t^{3,i}\rangle\,(h_t^3)^\top(W_2^4)^\top + \cdots\\
    &\in \Theta\bigl(\tfrac{1}{M}\textstyle\sum_i\langle W_t^4, h_t^{3,i}\rangle\cdot\langle h_t^3, W_2^4\rangle\bigr)\cdot h_0^1 \text{ along }h_0^1 \\
    & \quad (\text{at }t=2\text{: }\langle W_t^4, h_t^{3,i}\rangle\in\Theta(1)\text{ coherent via eff.\ term of (R3$\mu$-F2.5)};\ \text{LLN})\\
    &= \Theta(1\cdot 1) = \Theta(1/N) \text{ along }h_0^1 \tag{R3$\mu$-B3.5d}\\
    & \quad (\text{after multiplying by }\langle h_t^3, W_2^4\rangle\in\Theta(1)\text{ by (R3$\mu$-F3.5)};\ \text{prefactor }\eta^2\|W_0^4\|^2\in\Theta(1/N))
\end{align*}
\[
    (\partial f_2/\partial h_2^1)_\mathrm{exp} = A_4'' + A_5'' + A_6'' + E'' \in \Theta(1/N) \text{ along } h_0^1. \tag{R3$\mu$-B3.5}
\]
\begin{align*}
    (\partial f_2/\partial h_2^1)_\mathrm{router} &= \tfrac{1}{M}Q_0^\top v + \tfrac{1}{M}(\Delta_2 Q)^\top v, \qquad v_i := \dot\phi_{i,2}\langle h_2^{3,i},W_2^4\rangle\\
    &\in \Theta\bigl(\sigma_Q\|v\|/M\bigr) \text{ random direction} + \Theta\bigl(\langle\partial f_1/\partial\phi,v\rangle/M\bigr)\,h_1^1 \text{ along }h_1^1 \\
    & \quad (Q_0\text{ piece via cross-layer CLT};\ \Delta_2 Q\text{ piece coherent})\\
    &= \Theta(1/\sqrt N\cdot\sqrt M\cdot 1/M) = \Theta(1/\sqrt{MN}) = \Theta(1/N) \tag{R3$\mu$-B3.6}\\
    & \quad (\sigma_Q^2=1/N;\ v_i\in\Theta(1)\text{ via (R3$\mu$-B2.4)}\Rightarrow\|v\|^2\in\Theta(M);\ M\in\Theta(N))
\end{align*}
\[
    \partial f_2/\partial h_2^1 = (\partial f_2/\partial h_2^1)_\mathrm{exp} + (\partial f_2/\partial h_2^1)_\mathrm{router} \in \Theta(1/N), \tag{R3$\mu$-B3.7}
\]
dominated by $A_5''+A_6''+E''$ along $h_0^1$; same scale as $\partial f_1/\partial h_1^1$ at $t=1$, with $E''$ now matching $A_5'',A_6''$ via the alignment-coherence shift.

\subsubsection{Summary table of signal propagation for $\mu$P in Regime III}
\label{ssec:mup_consol_summary}

Notation: $\Delta_t W^\ell$ denotes the cumulative update $W_t^\ell - W_0^\ell$.

\newpage
\paragraph{Forward.}
\begin{center}
\renewcommand{\arraystretch}{1.2}
\begin{tabular}{l|l|l|l}
\textbf{Quantity} & $t=0$ & $t=1$ & $t=2$\\\hline
$h_t^1 = h_0^1 + \Delta_t h^1$                                                            & $\Theta(1)$         & $\Theta(1)$         & \textcolor{green!50!black}{$\Theta(1)$}\\
$\quad$ init: $h_0^1 = W_0^1 x$                                                           & $\Theta(1)$         & $\Theta(1)$         & \textcolor{green!50!black}{$\Theta(1)$}\\
$\quad$ effective: $\Delta_t h^1 = \Delta_t W^1 x$                                        & $0$                 & $\Theta(1/\sqrt N)$ & \textcolor{green!50!black}{$\Theta(1)$}\\
$\psi_t,\,\phi_t$                                                                         & $\Theta(1)$         & $\Theta(1)$         & \textcolor{green!50!black}{$\Theta(1)$}\\
$h_t^{2,i}$                                                                               & $\Theta(1)$         & $\Theta(1)$         & \textcolor{green!50!black}{$\Theta(1)$}\\
$\quad$ init: $h_0^{2,i} = W_0^{2,i}h_0^1$                                                & $\Theta(1)$         & $\Theta(1)$         & \textcolor{green!50!black}{$\Theta(1)$}\\
$\quad$ propagating: $W_0^{2,i}\Delta_t h^1$                                              & $0$                 & $\Theta(1/\sqrt N)$ & \textcolor{green!50!black}{$\Theta(1)$}\\
$\quad$ effective: $\Delta_t W^{2,i}\,h_0^1$                                              & $0$                 & $\Theta(1)$         & \textcolor{green!50!black}{$\Theta(1)$}\\
$\quad$ cross: $\Delta_t W^{2,i}\,\Delta_t h^1$                                           & $0$                 & $\Theta(1/N)$       & \textcolor{green!50!black}{$\Theta(1)$}\\
$h_t^{3,i}$                                                                               & $\Theta(1)$         & $\Theta(1)$         & \textcolor{green!50!black}{$\Theta(1)$}\\
$\quad$ init: $h_0^{3,i} = W_0^{3,i}h_0^{2,i}$                                            & $\Theta(1)$         & $\Theta(1)$         & \textcolor{green!50!black}{$\Theta(1)$}\\
$\quad$ propagating: $W_0^{3,i}\Delta_t h^{2,i}$                                          & $0$                 & $\Theta(1)$         & \textcolor{green!50!black}{$\Theta(1)$}\\
$\quad$ effective: $\Delta_t W^{3,i}\,h_0^{2,i}$                                          & $0$                 & $\Theta(1)$         & \textcolor{green!50!black}{$\Theta(1)$}\\
$\quad$ cross: $\Delta_t W^{3,i}\,\Delta_t h^{2,i}$                                       & $0$                 & $\Theta(1/N^{3/2})$ & \textcolor{green!50!black}{$\Theta(1)$}\\
$h_t^3 = A_1+A_{2,1}+A_{2,2}+A_3+D$                                                       & $\Theta(1/\sqrt N)$ & $\Theta(1)$         & \textcolor{green!50!black}{$\Theta(1)$}\\
$\quad A_1 = (1/M)\sum_i\phi_{i,t}\,h_0^{3,i}$                                            & $\Theta(1/\sqrt N)$ & $\Theta(1/\sqrt N)$ & \textcolor{red!70!black}{$\Theta(1/\sqrt N)$}\\
$\quad A_{2,1} = (1/M)\sum_i\phi_{i,t}\,W_0^{3,i}W_0^{2,i}\,\Delta_t h^1$                 & $0$                 & $\Theta(1/N)$       & \textcolor{red!70!black}{$\Theta(1/\sqrt N)$}\\
$\quad A_{2,2} = (1/M)\sum_i\phi_{i,t}\,W_0^{3,i}\,\Delta_t W^{2,i}\,h_0^1$               & $0$                 & $\Theta(1)$         & \textcolor{green!50!black}{$\Theta(1)$}\\
$\quad A_3 = (1/M)\sum_i\phi_{i,t}\,\Delta_t W^{3,i}\,h_0^{2,i}$                          & $0$                 & $\Theta(1)$         & \textcolor{green!50!black}{$\Theta(1)$}\\
$\quad D = (1/M)\sum_i\phi_{i,t}\,\Delta_t W^{3,i}\,\Delta_t h^{2,i}$                     & $0$                 & $\Theta(1/N)$       & \textcolor{green!50!black}{$\Theta(1)$}\\
$f_t = W_t^4 h_t^3$                                                                       & $\Theta(1/N)$       & $\Theta(1)$         & \textcolor{green!50!black}{$\Theta(1)$}\\
$\quad$ init: $W_0^4\,h_0^3$                                                              & $\Theta(1/N)$       & $\Theta(1/N)$       & \textcolor{red!70!black}{$\Theta(1/N)$}\\
$\quad$ propagating: $W_0^4\,\Delta_t h^3$                                                & $0$                 & $\Theta(1)$         & \textcolor{green!50!black}{$\Theta(1)$}\\
$\quad$ effective: $\Delta_t W^4\,h_0^3$                                                  & $0$                 & $\Theta(1/N)$       & \textcolor{red!70!black}{$\Theta(1/N)$}\\
$\quad$ cross: $\Delta_t W^4\,\Delta_t h^3$                                               & $0$                 & $\Theta(1/N)$       & \textcolor{green!50!black}{$\Theta(1)$}\\
\end{tabular}
\end{center}

\newpage
\paragraph{Backward.}
\begin{center}
\renewcommand{\arraystretch}{1.2}
\adjustbox{max width=\textwidth}{\begin{tabular}{l|l|l|l}
\textbf{Quantity} & $t=0$ & $t=1$ & $t=2$\\\hline
$\partial f_t/\partial h_t^3 = (W_0^4)^\top + (\Delta_t W^4)^\top$                                                              & $\Theta(1/N)$         & $\Theta(1/N)$         & \textcolor{green!50!black}{$\Theta(1/N)$}\\
$\quad$ init: $(W_0^4)^\top$                                                                                                    & $\Theta(1/N)$         & $\Theta(1/N)$         & \textcolor{green!50!black}{$\Theta(1/N)$}\\
$\quad$ update: $(\Delta_t W^4)^\top$                                                                                           & $0$                   & $\Theta(1/N^{3/2})$   & \textcolor{green!50!black}{$\Theta(1/N)$}\\
$\partial f_t/\partial h_t^{3,i} = (\phi_{i,t}/M)(W_t^4)^\top$                                                                  & $\Theta(1/(MN))$      & $\Theta(1/(MN))$      & \textcolor{green!50!black}{$\Theta(1/(MN))$}\\
$\quad$ init: $(\phi_{i,t}/M)(W_0^4)^\top$                                                                                      & $\Theta(1/(MN))$      & $\Theta(1/(MN))$      & \textcolor{green!50!black}{$\Theta(1/(MN))$}\\
$\quad$ update: $(\phi_{i,t}/M)(\Delta_t W^4)^\top$                                                                             & $0$                   & $\Theta(1/(MN^{3/2}))$ & \textcolor{green!50!black}{$\Theta(1/(MN))$}\\
$\partial f_t/\partial h_t^{2,i} = (\phi_{i,t}/M)(W_t^{3,i})^\top(W_t^4)^\top$                                                  & $\Theta(1/(MN))$      & $\Theta(1/(MN))$      & \textcolor{green!50!black}{$\Theta(1/(MN))$}\\
$\quad$ init$\cdot$init: $(\phi_{i,t}/M)(W_0^{3,i})^\top(W_0^4)^\top$                                                           & $\Theta(1/(MN))$      & $\Theta(1/(MN))$      & \textcolor{green!50!black}{$\Theta(1/(MN))$}\\
$\quad$ init$\cdot$update: $(\phi_{i,t}/M)(W_0^{3,i})^\top(\Delta_t W^4)^\top$                                                  & $0$                   & $\Theta(1/(MN^{3/2}))$ & \textcolor{green!50!black}{$\Theta(1/(MN))$}\\
$\quad$ update$\cdot$init: $(\phi_{i,t}/M)(\Delta_t W^{3,i})^\top(W_0^4)^\top$                                                  & $0$                   & $\Theta(1/(MN))$      & \textcolor{green!50!black}{$\Theta(1/(MN))$}\\
$\quad$ update$\cdot$update: $(\phi_{i,t}/M)(\Delta_t W^{3,i})^\top(\Delta_t W^4)^\top$                                         & $0$                   & $\Theta(1/(MN^2))$    & \textcolor{green!50!black}{$\Theta(1/(MN))$}\\
$\partial f_t/\partial \phi_{i,t} = (1/M)\langle h_t^{3,i}, W_t^4\rangle$                                                       & $\Theta(1/N^{3/2})$   & $\Theta(1/M)$         & \textcolor{green!50!black}{$\Theta(1/M)$}\\
$\quad$ init: $(1/M)\langle h_t^{3,i}, W_0^4\rangle$                                                                            & $\Theta(1/N^{3/2})$   & $\Theta(1/M)$         & \textcolor{green!50!black}{$\Theta(1/M)$}\\
$\quad$ update: $(1/M)\langle h_t^{3,i}, \Delta_t W^4\rangle$                                                                   & $0$                   & $\Theta(1/N^{3/2})$   & \textcolor{green!50!black}{$\Theta(1/M)$}\\
$(\partial f_t/\partial h_t^1)_\mathrm{exp}$                                                                                    & $\Theta(1/N^{3/2})$   & $\Theta(1/N)$         & \textcolor{green!50!black}{$\Theta(1/N)$}\\
$\quad A_{4,1} = (1/M)\sum_i\phi_{i,t}\,(W_0^{2,i})^\top(W_0^{3,i})^\top(W_0^4)^\top$                                           & $\Theta(1/N^{3/2})$   & $\Theta(1/N^{3/2})$   & \textcolor{red!70!black}{$\Theta(1/N^{3/2})$}\\
$\quad A_{4,2} = (1/M)\sum_i\phi_{i,t}\,(W_0^{2,i})^\top(W_0^{3,i})^\top(\Delta_t W^4)^\top$                                    & $0$                   & $\Theta(1/N^2)$       & \textcolor{red!70!black}{$\Theta(1/N^{3/2})$}\\
$\quad A_{5,1} = (1/M)\sum_i\phi_{i,t}\,(W_0^{2,i})^\top(\Delta_t W^{3,i})^\top(W_0^4)^\top$                                    & $0$                   & $\Theta(1/N)$         & \textcolor{green!50!black}{$\Theta(1/N)$}\\
$\quad A_{5,2} = (1/M)\sum_i\phi_{i,t}\,(W_0^{2,i})^\top(\Delta_t W^{3,i})^\top(\Delta_t W^4)^\top$                             & $0$                   & $\Theta(1/N^2)$       & \textcolor{green!50!black}{$\Theta(1/N)$}\\
$\quad A_{6,1} = (1/M)\sum_i\phi_{i,t}\,(\Delta_t W^{2,i})^\top(W_0^{3,i})^\top(W_0^4)^\top$                                    & $0$                   & $\Theta(1/N)$         & \textcolor{green!50!black}{$\Theta(1/N)$}\\
$\quad A_{6,2} = (1/M)\sum_i\phi_{i,t}\,(\Delta_t W^{2,i})^\top(W_0^{3,i})^\top(\Delta_t W^4)^\top$                             & $0$                   & $\Theta(1/N^2)$       & \textcolor{green!50!black}{$\Theta(1/N)$}\\
$\quad E_1 = (1/M)\sum_i\phi_{i,t}\,(\Delta_t W^{2,i})^\top(\Delta_t W^{3,i})^\top(W_0^4)^\top$                                 & $0$                   & $\Theta(1/N^2)$       & \textcolor{green!50!black}{$\Theta(1/N)$}\\
$\quad E_2 = (1/M)\sum_i\phi_{i,t}\,(\Delta_t W^{2,i})^\top(\Delta_t W^{3,i})^\top(\Delta_t W^4)^\top$                          & $0$                   & $\Theta(1/N^2)$       & \textcolor{green!50!black}{$\Theta(1/N)$}\\
$(\partial f_t/\partial h_t^1)_\mathrm{router} = (1/M)Q_t^\top v$                                                               & $\Theta(1/N^{3/2})$   & $\Theta(1/\sqrt{MN})$ & \textcolor{green!50!black}{$\Theta(1/\sqrt{MN})$}\\
$\quad$ init: $(1/M)Q_0^\top v$                                                                                                 & $\Theta(1/N^{3/2})$   & $\Theta(1/\sqrt{MN})$ & \textcolor{green!50!black}{$\Theta(1/\sqrt{MN})$}\\
$\quad$ update: $(1/M)(\Delta_t Q)^\top v$                                                                                      & $0$                   & $\Theta(1/(MN))$      & \textcolor{green!50!black}{$\Theta(1/\sqrt{MN})$}\\
\end{tabular}}
\end{center}

\subsubsection{Deriving MSSP in Regime III}
\label{sec:ssp_consolidated}

This subsection provides a self-contained, heuristic scaling analysis for the Maximally Scale-Stable Parameterization (MSSP, Definition~\ref{def:ssp_regime_iii}) in Regime III.

\paragraph{Auxiliary scaling lemma}
\label{ssec:ssp_consol_lemmas}

\begin{lemma}[Gram concentration]\label{lem:ssp_gram}
Let $W\in\mathbb{R}^{m\times n}$ have i.i.d.\ centered entries of variance $\sigma_W^2$. Then
\[
    \mathbb{E}\bigl[(W^\top W)_{jk}\bigr]=m\sigma_W^2\,\delta_{jk},\qquad
    \mathbb{E}\bigl[(WW^\top)_{ij}\bigr]=n\sigma_W^2\,\delta_{ij},
\]
with off-diagonal fluctuations of order $\sigma_W^2\sqrt m$ (resp.\ $\sigma_W^2\sqrt n$). For any fixed $v\in\mathbb{R}^n$ with $\|v\|^2\in\Theta(n)$ that is independent of $W$, $W^\top W\,v = m\sigma_W^2\,v + r$ with $r$ of coordinate scale $\Theta(\sigma_W^2\sqrt{mn})$. The analogous statement holds for $WW^\top$ acting on $v'\in\mathbb{R}^m$ with $\|v'\|^2\in\Theta(m)$.
\end{lemma}
\begin{proof}[Sketch]
$(W^\top W)_{jk}=\sum_a W_{aj}W_{ak}$ is a sum of $m$ centered i.i.d.\ products with mean $\sigma_W^2\delta_{jk}$ and variance $\sigma_W^4$ off-diagonal. The residual $r=(W^\top W-m\sigma_W^2 I)v$ has entry-wise variance $\sum_k m\sigma_W^4\,v_k^2=m\sigma_W^4\,\|v\|^2$, hence coordinate scale $\sigma_W^2\sqrt{mn}$.
\end{proof}

\textbf{Specializations of Lemma~\ref{lem:ssp_gram} under proportional scaling.} Both Gram matrices of $W_0^{2,1}$ and of $W_0^{3,1}$ produce $\Theta(1)$-coordinate output when applied to $\Theta(1)$-coordinate input vectors (independent of or weakly correlated with the corresponding $W$):
\begin{align*}
    (W_0^{2,1})^\top W_0^{2,1}\,v &= (N_e/N)\,v + r,
        & W_0^{2,1}(W_0^{2,1})^\top\,v' &= v' + r',\\
    (W_0^{3,1})^\top W_0^{3,1}\,u &= (N/N_e)\,u + r'',
        & W_0^{3,1}(W_0^{3,1})^\top\,u' &= u' + r''',
\end{align*}
where each residual is of the same $\Theta(1)$ entry-wise order as the leading deterministic part. The Gram approximation is order-of-magnitude correct, with comparable random fluctuations.

\paragraph{First forward pass}
\label{ssec:ssp_consol_fwd1}

\[
    h_0^1 = W_0^1\,x \in \Theta(1). \quad (\text{CLT};\ \sigma_1^2=1/D,\ \|x\|^2\in\Theta(D)) \tag{R3$s$-F1.1}
\]
\[
    \psi_0 = Q_0\,h_0^1 \in \Theta(1). \quad (\text{CLT};\ \sigma_Q^2=1/N,\ \|h_0^1\|^2\in\Theta(N)) \tag{R3$s$-F1.2}
\]
\[
    \phi_0 = \sigma(\psi_0) \in \Theta(1). \quad (\text{sigmoid bounded; gating assumption}) \tag{R3$s$-F1.3}
\]
\begin{align*}
    h_0^{2,i} &= h_0^{2,1} = W_0^{2,1}\,h_0^1 \in \Theta(1)\\
    & \quad (\text{CLT};\ \sigma_2^2=1/N,\ \|h_0^1\|^2\in\Theta(N);\ h_0^{2,i}=h_0^{2,1}\text{ by MSSP}). \tag{R3$s$-F1.4}
\end{align*}
\begin{align*}
    h_0^{3,i} &= h_0^{3,1} = W_0^{3,1}\,h_0^{2,1} \in \Theta(1)\\
    & \quad (\text{CLT};\ \sigma_3^2=1/N_e,\ \|h_0^{2,1}\|^2\in\Theta(N_e);\ h_0^{3,i}=h_0^{3,1}\text{ by MSSP}). \tag{R3$s$-F1.5}
\end{align*}
\begin{align*}
    h_0^3 &= (1/M)\textstyle\sum_i\phi_{i,0}\,h_0^{3,1} = (\textstyle\sum_k\phi_{k,0}/M)\,h_0^{3,1} \in \Theta(1)\\
    & \quad ((\textstyle\sum_k\phi_{k,0}/M)\in\Theta(1)\text{ by LLN}). \tag{R3$s$-F1.6}
\end{align*}
\[
    f_0 = W_0^4\,h_0^3 = 0. \quad (\text{standing assumption } W_0^4=0) \tag{R3$s$-F1.7}
\]

\paragraph{First backward pass and step-1 updates}
\label{ssec:ssp_consol_bwd1}

By the chain rule, every hidden gradient ($\partial f_0/\partial h_0^3$, $\partial f_0/\partial h_0^{3,i}$, $\partial f_0/\partial h_0^{2,i}$, $\partial f_0/\partial \phi_{i,0}$, $\partial f_0/\partial h_0^1$ via either the expert or router path) carries a $W_0^4=0$ factor and is therefore zero. Hence every non-readout weight is unchanged at $t=1$: $W_1^\ell = W_0^\ell$ for $\ell\in\{1,Q,2,3\}$.

\paragraph{Step-1 parameter updates.}
\[
    \Delta_1 W^4 = -\tfrac{\chi_0}{N}\,(h_0^3)^\top \in \Theta(1/N). \quad (\text{aligned along } (h_0^3)^\top) \tag{R3$s$-U1.4}
\]

\paragraph{Second forward pass}
\label{ssec:ssp_consol_fwd2}

Since the non-readout weights are unchanged at $t=1$, every hidden activation is unchanged: $h_1^\ell=h_0^\ell$ for $\ell\in\{1,(2,i),(3,i),3\}$, $\psi_1=\psi_0$, $\phi_1=\phi_0$.

\[
    f_1 = W_1^4\,h_0^3 = -\tfrac{\chi_0}{N}\|h_0^3\|^2 \in \Theta(1). \quad (\|h_0^3\|^2\in\Theta(N)\text{ via (R3$s$-F1.6)}) \tag{R3$s$-F2.7}
\]

\paragraph{Second backward pass and step-2 updates}
\label{ssec:ssp_consol_bwd2}

\[
    \partial f_1/\partial h_1^3 = (W_1^4)^\top = (\Delta_1 W^4)^\top \in \Theta(1/N). \quad (\text{(R3$s$-U1.4)};\ W_0^4=0) \tag{R3$s$-B2.1}
\]
\[
    \partial f_1/\partial h_1^{3,i} = (\phi_{i,0}/M)(W_1^4)^\top \in \Theta(1/(MN)). \quad (\phi_{i,0}\in\Theta(1)) \tag{R3$s$-B2.2}
\]
\begin{align*}
    \partial f_1/\partial h_1^{2,i} &= (\phi_{i,0}/M)(W_0^{3,1})^\top(W_1^4)^\top \in \Theta(1/(MN))\\
    & \quad ((W_0^{3,1})^\top(W_1^4)^\top = -\tfrac{\chi_0}{MN}(\textstyle\sum\phi)(W_0^{3,1})^\top W_0^{3,1}h_0^{2,1}\in\Theta(1/N)\text{ by Lem.~\ref{lem:ssp_gram}};\\
    & \qquad \phi_{i,0}\in\Theta(1)). \tag{R3$s$-B2.3}
\end{align*}
\begin{align*}
    \partial f_1/\partial \phi_{i,0} &= (1/M)(h_0^{3,1})^\top(W_1^4)^\top \in \Theta(1/M)\\
    & \quad ((h_0^{3,1})^\top(W_1^4)^\top = -\tfrac{\chi_0}{M N}(\textstyle\sum\phi)\|h_0^{3,1}\|^2\in\Theta(1)\text{ by LLN};\\
    & \qquad \|h_0^{3,1}\|^2\in\Theta(N_e)\text{ via (R3$s$-F1.5)};\ N_e/N\in\Theta(1)\text{ in Regime III}). \tag{R3$s$-B2.4}
\end{align*}
\begin{align*}
    \Bigl(\tfrac{\partial f_1}{\partial h_1^1}\Bigr)_{\!\!\mathrm{exp}} &= \bigl(\tfrac{1}{M}\textstyle\sum\phi\bigr)(W_0^{2,1})^\top(W_0^{3,1})^\top(W_1^4)^\top\\
    &\in \Theta\bigl(\tfrac{1}{M}\textstyle\sum\phi\bigr)\cdot(W_0^{2,1})^\top h_0^{2,1}\,/N \text{ along }(W_0^{2,1})^\top h_0^{2,1}\\
    & \quad (\text{Lem.~\ref{lem:ssp_gram} on }(W_0^{3,1})^\top h_0^3)\\
    &= \Theta(1\cdot 1/N) = \Theta(1/N)\\
    & \quad (\text{LLN};\ (W_0^{2,1})^\top h_0^{2,1}\in\Theta(1)\text{ by Lem.~\ref{lem:ssp_gram}}). \tag{R3$s$-B2.5}
\end{align*}
\begin{align*}
    \Bigl(\tfrac{\partial f_1}{\partial h_1^1}\Bigr)_{\!\!\mathrm{router}} &= \tfrac{1}{M}Q_0^\top v, \qquad v_i := \dot\phi_{i,0}(h_0^{3,1})^\top(W_1^4)^\top\\
    &\in \Theta\bigl(\sigma_Q\|v\|/M\bigr) \text{ random direction} \quad (Q_0\text{ indep.\ of }v;\ \text{cross-layer CLT})\\
    &= \Theta(1/\sqrt N\cdot\sqrt M\cdot 1/M) = \Theta(1/\sqrt{MN})\\
    & \quad (\sigma_Q^2=1/N;\ v_i\in\Theta(1)\text{ via (R3$s$-B2.4)}\Rightarrow\|v\|^2\in\Theta(M)). \tag{R3$s$-B2.6}
\end{align*}
\[
    \partial f_1/\partial h_1^1 = (\partial f_1/\partial h_1^1)_\mathrm{exp} + (\partial f_1/\partial h_1^1)_\mathrm{router} \in \Theta(1/N). \tag{R3$s$-B2.7}
\]

\paragraph{Step-2 parameter updates.}
\begin{align*}
    \Delta_2 W^4 &= -\tfrac{\chi_1}{N}(h_1^3)^\top \in \Theta(1/N), \quad W_2^4 = -\tfrac{\chi_0+\chi_1}{N}(h_0^3)^\top\\
    & \quad (h_1^3=h_0^3). \tag{R3$s$-U2.4}
\end{align*}
\[
    \Delta_2 W^{3,i} = -\tfrac{\chi_1\phi_{i,0}}{N}\,h_0^3\,(h_0^{2,1})^\top \in \Theta(1/N). \quad (\eta_3=M;\ \text{(R3$s$-B2.2)}) \tag{R3$s$-U2.3}
\]
\[
    \Delta_2 W^{2,i} = -\tfrac{\chi_1\phi_{i,0}}{N}\,(W_0^{3,1})^\top h_0^3\,(h_0^1)^\top \in \Theta(1/N). \quad ((W_0^{3,1})^\top h_0^3\in\Theta(1)\text{ via (R3$s$-B2.3)}) \tag{R3$s$-U2.2}
\]
\begin{align*}
    \Delta_2 W^1 &= -\eta_1\chi_1(\partial f_1/\partial h_1^1)\,x^\top \in \Theta(1)\\
    & \quad (\eta_1=N;\ \partial f_1/\partial h_1^1\in\Theta(1/N)\text{ via (R3$s$-B2.7)}). \tag{R3$s$-U2.1a}
\end{align*}
\[
    \Delta_2 h^1 = \Delta_2 W^1\,x \in \Theta(1) \text{ aligned along } \partial f_1/\partial h_1^1. \tag{R3$s$-U2.1b}
\]
\begin{align*}
    \Delta_2 Q &= -\eta_Q\chi_1\,(\partial f_1/\partial\phi)\,(h_1^1)^\top \in \Theta(1/M)=\Theta(1/N)\\
    & \quad (\eta_Q=1;\ \partial f_1/\partial\phi\in\Theta(1/M)\text{ via (R3$s$-B2.4)}). \tag{R3$s$-U2.Q}
\end{align*}

\paragraph{Third forward pass}
\label{ssec:ssp_consol_fwd3}

We compute activations at $\theta^{(2)}$, expanding each updated weight as $W_2^\ell = W_0^\ell+\Delta_2 W^\ell$.

\[
    h_2^1 = h_0^1 + \Delta_2 h^1 \in \Theta(1). \quad (\Delta_2 h^1\in\Theta(1)\text{ via (R3$s$-U2.1b)}) \tag{R3$s$-F3.1}
\]

\noindent (R3$s$-F3.2) $h_2^{2,i} = W_2^{2,i}\,h_2^1$, four-piece decomposition:
\[
    h_2^{2,i} = h_0^{2,1} + W_0^{2,1}\,\Delta_2 h^1 + \Delta_2 W^{2,i}\,h_0^1 + \Delta_2 W^{2,i}\,\Delta_2 h^1. \tag{R3$s$-F3.2}
\]
\[
    \text{init: } W_0^{2,1}h_0^1 = h_0^{2,1} \in \Theta(1). \quad (\text{R3$s$-F1.4}) \tag{R3$s$-F3.2a}
\]
\begin{align*}
    \text{prop: } W_0^{2,1}\,\Delta_2 h^1 &\in \Theta(1)\\
    & \quad (\text{op-norm of iid Gaussian } W_0^{2,1};\ \Delta_2 h^1\in\Theta(1)\text{ via (R3$s$-U2.1b)}). \tag{R3$s$-F3.2b}
\end{align*}
\begin{align*}
    \text{eff: } \Delta_2 W^{2,i}\,h_0^1 &= -\tfrac{\chi_1\phi_{i,0}}{N}(W_0^{3,1})^\top h_0^3\,\|h_0^1\|^2 \in \Theta(1)\\
    & \quad ((W_0^{3,1})^\top h_0^3\in\Theta(1)\text{ via (R3$s$-B2.3)};\ \|h_0^1\|^2\in\Theta(N)). \tag{R3$s$-F3.2c}
\end{align*}
\begin{align*}
    \text{cross: } \Delta_2 W^{2,i}\,\Delta_2 h^1 &= -\tfrac{\chi_1\phi_{i,0}}{N}(W_0^{3,1})^\top h_0^3\,\langle h_0^1,\Delta_2 h^1\rangle \in \Theta(1)\\
    & \quad ((W_0^{3,1})^\top h_0^3\in\Theta(1)\text{ via (R3$s$-B2.3)};\\
    & \qquad \langle h_0^1,\Delta_2 h^1\rangle\in\Theta(N)\text{ coherent via (R3$s$-U2.1b)}). \tag{R3$s$-F3.2d}
\end{align*}

\noindent (R3$s$-F3.3) $h_2^{3,i} = W_2^{3,i}\,h_2^{2,i}$, four-piece decomposition:
\[
    h_2^{3,i} = h_0^{3,1} + W_0^{3,1}\,\Delta_2 h^{2,i} + \Delta_2 W^{3,i}\,h_0^{2,1} + \Delta_2 W^{3,i}\,\Delta_2 h^{2,i}. \tag{R3$s$-F3.3}
\]
\[
    \text{init: } h_0^{3,1} \in \Theta(1). \quad (\text{R3$s$-F1.5}) \tag{R3$s$-F3.3a}
\]
\begin{align*}
    \text{prop: } W_0^{3,1}\,\Delta_2 h^{2,i} &\in \Theta(1)\\
    & \quad (\text{op-norm of iid Gaussian } W_0^{3,1};\ \Delta_2 h^{2,i}\in\Theta(1)\text{ via (R3$s$-F3.2)}). \tag{R3$s$-F3.3b}
\end{align*}
\begin{align*}
    \text{eff: } \Delta_2 W^{3,i}\,h_0^{2,1} &= -\tfrac{\chi_1\phi_{i,0}}{N}\,h_0^3\,\|h_0^{2,1}\|^2 \in \Theta(1)\\
    & \quad (\|h_0^{2,1}\|^2\in\Theta(N_e);\ h_0^3\in\Theta(1)\text{ via (R3$s$-F1.6)}). \tag{R3$s$-F3.3c}
\end{align*}
\[
    \text{cross: } \Delta_2 W^{3,i}\,\Delta_2 h^{2,i} \text{ tracked as part of } D \text{ in (R3$s$-F3.4)}. \tag{R3$s$-F3.3d}
\]

\noindent (R3$s$-F3.4) $h_2^3 = A_1 + A_2 + A_3 + D$, where
\begin{align*}
    A_1 &:= \tfrac{1}{M}\textstyle\sum_i\phi_{i,2}\,W_0^{3,1}h_0^{2,1}, &
    A_2 &:= \tfrac{1}{M}\textstyle\sum_i\phi_{i,2}\,\Delta_2 W^{3,i}\,h_0^{2,1},\\
    A_3 &:= \tfrac{1}{M}\textstyle\sum_i\phi_{i,2}\,W_0^{3,1}\,\Delta_2 h^{2,i}, &
    D   &:= \tfrac{1}{M}\textstyle\sum_i\phi_{i,2}\,\Delta_2 W^{3,i}\,\Delta_2 h^{2,i}.
\end{align*}
\[
    A_1 = \bigl(\textstyle\sum\phi_{\cdot,2}/M\bigr)\,h_0^{3,1} \in \Theta(1). \quad (i\text{-independent matrix; LLN}) \tag{R3$s$-F3.4a}
\]
\begin{align*}
    A_2 &= -\tfrac{\chi_1\,\|h_0^{2,1}\|^2}{N}\Bigl(\tfrac{1}{M}\textstyle\sum_i\phi_{i,2}\phi_{i,0}\Bigr)\,h_0^3 \in \Theta(1) \text{ along } h_0^3\\
    & \quad (\|h_0^{2,1}\|^2/N\in\Theta(1);\ \text{LLN}). \tag{R3$s$-F3.4b}
\end{align*}
\begin{align*}
    A_3 &\in \Theta(1)\\
    & \quad (\text{three-piece expansion of }\Delta_2 h^{2,i}\text{ from (R3$s$-F3.2)};\\
    & \qquad \text{Lemma~\ref{lem:ssp_gram} on }h_0^3, h_0^1;\ \langle h_0^1,\Delta_2 h^1\rangle\in\Theta(N);\ \text{LLN}). \tag{R3$s$-F3.4c}
\end{align*}
\begin{align*}
    D &= -\tfrac{\chi_1}{N}\,h_0^3\,\tfrac{1}{M}\textstyle\sum_i\phi_{i,2}\phi_{i,0}\,\langle h_0^{2,1},\Delta_2 h^{2,i}\rangle\\
    &\in \Theta\bigl(\tfrac{1}{M}\textstyle\sum_i\phi_{i,2}\phi_{i,0}\,\langle h_0^{2,1},\Delta_2 h^{2,i}\rangle/N\bigr)\cdot h_0^3 \text{ along }h_0^3\\
    & \quad (\text{Lem.~\ref{lem:ssp_gram};\ LLN})\\
    &= \Theta(N/N) = \Theta(1) \text{ along }h_0^3\\
    & \quad (\langle h_0^{2,1},\Delta_2 h^{2,i}\rangle\in\Theta(N)\text{ coherent via (R3$s$-U2.1b)};\ h_0^3\in\Theta(1)). \tag{R3$s$-F3.4d}
\end{align*}
\[
    h_2^3 = A_1+A_2+A_3+D \in \Theta(1). \tag{R3$s$-F3.4}
\]
\begin{align*}
    f_2 &= W_2^4\,h_2^3 = -\tfrac{\chi_0+\chi_1}{N}\langle h_0^3, h_2^3\rangle \in \Theta(1)\\
    & \quad (\langle h_0^3,A_1\rangle = (\textstyle\sum\phi_{\cdot,2}/M)\langle h_0^3,h_0^{3,1}\rangle\in\Theta(N)\text{ coherent via (R3$s$-F3.4a)};\\
    & \qquad \langle h_0^3,A_2+A_3+D\rangle\in\Theta(N)\text{ via aligned components (R3$s$-F3.4b)--(R3$s$-F3.4d)}). \tag{R3$s$-F3.5}
\end{align*}

\paragraph{Third backward pass}
\label{ssec:ssp_consol_bwd3}

\[
    \partial f_2/\partial h_2^3 = (W_2^4)^\top = -\tfrac{\chi_0+\chi_1}{N}\,h_0^3 \in \Theta(1/N). \quad (\text{(R3$s$-U2.4)};\ h_0^3\in\Theta(1)) \tag{R3$s$-B3.1}
\]
\[
    \partial f_2/\partial h_2^{3,i} = (\phi_{i,2}/M)(W_2^4)^\top \in \Theta(1/(MN)). \quad (\phi_{i,2}\in\Theta(1)) \tag{R3$s$-B3.2}
\]
\begin{align*}
    \partial f_2/\partial h_2^{2,i} &= (\phi_{i,2}/M)(W_2^{3,i})^\top(W_2^4)^\top \in \Theta(1/(MN))\\
    & \quad ((W_0^{3,1})^\top(W_2^4)^\top\in\Theta(1/N)\text{ via (R3$s$-B2.3)};\\
    & \qquad (\Delta_2 W^{3,i})^\top(W_2^4)^\top\in\Theta(1/N)\text{ along }h_0^{2,1}\text{ using }\|h_0^3\|^2\in\Theta(N)\text{ via (R3$s$-F1.6)};\\
    & \qquad \phi_{i,2}\in\Theta(1)). \tag{R3$s$-B3.3}
\end{align*}
\begin{align*}
    \partial f_2/\partial \phi_{i,2} &= (1/M)(h_2^{3,i})^\top(W_2^4)^\top \in \Theta(1/M)\\
    & \quad ((h_2^{3,i})^\top(W_2^4)^\top = -\tfrac{\chi_0+\chi_1}{N}\langle h_2^{3,i},h_0^3\rangle\in\Theta(1);\\
    & \qquad \langle h_2^{3,i},h_0^3\rangle\in\Theta(N)\text{ coherent via (R3$s$-F3.3)}). \tag{R3$s$-B3.4}
\end{align*}

\noindent (R3$s$-B3.5) $(\partial f_2/\partial h_2^1)_\mathrm{exp} = \widetilde A_4 + \widetilde A_5 + \widetilde A_6 + \widetilde E$, where
\begin{align*}
    \widetilde A_4 &:= \tfrac{1}{M}\textstyle\sum_i\phi_{i,2}\,(W_0^{2,1})^\top(W_0^{3,1})^\top(W_2^4)^\top,\\
    \widetilde A_5 &:= \tfrac{1}{M}\textstyle\sum_i\phi_{i,2}\,(W_0^{2,1})^\top(\Delta_2 W^{3,i})^\top(W_2^4)^\top,\\
    \widetilde A_6 &:= \tfrac{1}{M}\textstyle\sum_i\phi_{i,2}\,(\Delta_2 W^{2,i})^\top(W_0^{3,1})^\top(W_2^4)^\top,\\
    \widetilde E   &:= \tfrac{1}{M}\textstyle\sum_i\phi_{i,2}\,(\Delta_2 W^{2,i})^\top(\Delta_2 W^{3,i})^\top(W_2^4)^\top.
\end{align*}
\begin{align*}
    \widetilde A_4 &= -\tfrac{\chi_0+\chi_1}{N}\bigl(\textstyle\sum\phi_{\cdot,2}/M\bigr)(W_0^{2,1})^\top(W_0^{3,1})^\top h_0^3\\
    &\in \Theta\bigl((W_0^{2,1})^\top(W_0^{3,1})^\top h_0^3/N\bigr)\\
    & \quad (i\text{-indep.\ matrix under MSSP};\ \text{Lem.~\ref{lem:ssp_gram}};\ \text{LLN})\\
    &= \Theta(1\cdot 1/N) = \Theta(1/N)\\
    & \quad ((W_0^{2,1})^\top(W_0^{3,1})^\top h_0^3\in\Theta(1)\text{ via (R3$s$-B2.3)}). \tag{R3$s$-B3.5a}
\end{align*}
\begin{align*}
    \widetilde A_5 &= \tfrac{\chi_1(\chi_0+\chi_1)\,\|h_0^3\|^2}{N^2}\,\tfrac{1}{M}\textstyle\sum_i\phi_{i,2}\phi_{i,0}\,(W_0^{2,1})^\top h_0^{2,1}\\
    &\in \Theta\bigl(\|h_0^3\|^2\,\tfrac{1}{M}\textstyle\sum_i\phi_{i,2}\phi_{i,0}/N^2\bigr)\cdot h_0^1 \text{ along }h_0^1\\
    & \quad (\text{Lem.~\ref{lem:ssp_gram}: }(W_0^{2,1})^\top h_0^{2,1}\to h_0^1)\\
    &= \Theta(N\cdot 1\cdot 1/N^2) = \Theta(1/N) \text{ along }h_0^1\\
    & \quad (\|h_0^3\|^2\in\Theta(N)\text{ via (R3$s$-F1.6)};\ \text{LLN};\ h_0^1\in\Theta(1)). \tag{R3$s$-B3.5b}
\end{align*}
\begin{align*}
    \widetilde A_6 &= -\tfrac{\chi_1}{N}\,h_0^1\,\tfrac{1}{M}\textstyle\sum_i\phi_{i,0}\,S_i, \qquad S_i := (h_0^3)^\top W_0^{3,1}(W_0^{3,1})^\top(W_2^4)^\top\\
    &\in \Theta\bigl(\tfrac{1}{M}\textstyle\sum_i\phi_{i,0}\,S_i/N\bigr)\cdot h_0^1 \text{ along }h_0^1\\
    & \quad (\text{Lem.~\ref{lem:ssp_gram}: }S_i\in\Theta(1);\ \text{LLN})\\
    &= \Theta(1\cdot 1/N) = \Theta(1/N) \text{ along }h_0^1\\
    & \quad (S_i\in\Theta(1);\ h_0^1\in\Theta(1)). \tag{R3$s$-B3.5c}
\end{align*}
\begin{align*}
    \widetilde E &= -\tfrac{\chi_1^2(\chi_0+\chi_1)\,\|h_0^3\|^2}{N^3}\,h_0^1\,\tfrac{1}{M}\textstyle\sum_i\phi_{i,2}\phi_{i,0}^2\,\langle h_0^3,h_0^{3,1}\rangle\\
    &\in \Theta\bigl(\langle h_0^3,h_0^{3,1}\rangle\,\|h_0^3\|^2/N^3\bigr)\cdot h_0^1 \text{ along }h_0^1\\
    & \quad (\text{LLN; }\langle h_0^3,h_0^{3,1}\rangle\text{ coherent under MSSP})\\
    &= \Theta(N\cdot N/N^3) = \Theta(1/N) \text{ along }h_0^1\\
    & \quad (\langle h_0^3,h_0^{3,1}\rangle\in\Theta(N)\text{ coherent via (R3$s$-F1.6)};\ \|h_0^3\|^2\in\Theta(N)\text{ via (R3$s$-F1.6)}). \tag{R3$s$-B3.5d}
\end{align*}
\[
    (\partial f_2/\partial h_2^1)_\mathrm{exp} \in \Theta(1/N) \text{ along } h_0^1. \tag{R3$s$-B3.5}
\]
\begin{align*}
    (\partial f_2/\partial h_2^1)_\mathrm{router} &= \tfrac{1}{M}Q_0^\top v + \tfrac{1}{M}(\Delta_2 Q)^\top v, \qquad v_i := \dot\phi_{i,2}(h_2^{3,i})^\top(W_2^4)^\top\\
    &\in \Theta\bigl(\sigma_Q\|v\|/M\bigr) \text{ random direction} + \Theta\bigl(\langle\partial f_1/\partial\phi,v\rangle/M\bigr)\,h_1^1 \text{ along }h_1^1\\
    & \quad (Q_0\text{ piece via cross-layer CLT};\ \Delta_2 Q\text{ piece coherent under MSSP})\\
    &= \Theta(1/\sqrt N\cdot\sqrt M\cdot 1/M) = \Theta(1/\sqrt{MN})\\
    & \quad (\sigma_Q^2=1/N;\ \|v\|^2\in\Theta(M)\text{ via }v_i\in\Theta(1)\text{ from (R3$s$-B3.4)};\\
    & \qquad \langle\partial f_1/\partial\phi,v\rangle\in\Theta(1)\text{ via MSSP coherence}). \tag{R3$s$-B3.6}
\end{align*}
\[
    \partial f_2/\partial h_2^1 = (\partial f_2/\partial h_2^1)_\mathrm{exp} + (\partial f_2/\partial h_2^1)_\mathrm{router} \in \Theta(1/N). \tag{R3$s$-B3.7}
\]

\subsubsection{Summary table of signal propagation for MSSP in Regime III}
\label{ssec:ssp_r3_summary}

Notation: $\Delta_t W^\ell$ denotes the cumulative update $W_t^\ell - W_0^\ell$.

\paragraph{Forward.}
\begin{center}
\renewcommand{\arraystretch}{1.2}
\begin{tabular}{l|l|l|l}
\textbf{Quantity} & $t=0$ & $t=1$ & $t=2$\\\hline
$h_t^1 = h_0^1 + \Delta_t h^1$                                                     & $\Theta(1)$ & $\Theta(1)$ & \textcolor{green!50!black}{$\Theta(1)$}\\
$\quad$ init: $h_0^1 = W_0^1 x$                                                    & $\Theta(1)$ & $\Theta(1)$ & \textcolor{green!50!black}{$\Theta(1)$}\\
$\quad$ effective: $\Delta_t h^1 = \Delta_t W^1 x$                                 & $0$         & $0$         & \textcolor{green!50!black}{$\Theta(1)$}\\
$\psi_t,\,\phi_t$                                                                  & $\Theta(1)$ & $\Theta(1)$ & \textcolor{green!50!black}{$\Theta(1)$}\\
$h_t^{2,i}$                                                                        & $\Theta(1)$ & $\Theta(1)$ & \textcolor{green!50!black}{$\Theta(1)$}\\
$\quad$ init: $W_0^{2,1}h_0^1$                                                     & $\Theta(1)$ & $\Theta(1)$ & \textcolor{green!50!black}{$\Theta(1)$}\\
$\quad$ propagating: $W_0^{2,1}\Delta_t h^1$                                       & $0$         & $0$         & \textcolor{green!50!black}{$\Theta(1)$}\\
$\quad$ effective: $\Delta_t W^{2,i}\,h_0^1$                                       & $0$         & $0$         & \textcolor{green!50!black}{$\Theta(1)$}\\
$\quad$ cross: $\Delta_t W^{2,i}\,\Delta_t h^1$                                    & $0$         & $0$         & \textcolor{green!50!black}{$\Theta(1)$}\\
$h_t^{3,i}$                                                                        & $\Theta(1)$ & $\Theta(1)$ & \textcolor{green!50!black}{$\Theta(1)$}\\
$\quad$ init: $W_0^{3,1}h_0^{2,1}$                                                 & $\Theta(1)$ & $\Theta(1)$ & \textcolor{green!50!black}{$\Theta(1)$}\\
$\quad$ propagating: $W_0^{3,1}\Delta_t h^{2,i}$                                   & $0$         & $0$         & \textcolor{green!50!black}{$\Theta(1)$}\\
$\quad$ effective: $\Delta_t W^{3,i}\,h_0^{2,1}$                                   & $0$         & $0$         & \textcolor{green!50!black}{$\Theta(1)$}\\
$\quad$ cross: $\Delta_t W^{3,i}\,\Delta_t h^{2,i}$                                & $0$         & $0$         & \textcolor{green!50!black}{$\Theta(1)$}\\
$h_t^3 = A_1+A_2+A_3+D$                                                            & $\Theta(1)$ & $\Theta(1)$ & \textcolor{green!50!black}{$\Theta(1)$}\\
$\quad A_1 = (1/M)\sum_i\phi_{i,t}\,W_0^{3,1}h_0^{2,1}$                            & $\Theta(1)$ & $\Theta(1)$ & \textcolor{green!50!black}{$\Theta(1)$}\\
$\quad A_2 = (1/M)\sum_i\phi_{i,t}\,\Delta_t W^{3,i}\,h_0^{2,1}$                   & $0$         & $0$         & \textcolor{green!50!black}{$\Theta(1)$}\\
$\quad A_3 = (1/M)\sum_i\phi_{i,t}\,W_0^{3,1}\,\Delta_t h^{2,i}$                   & $0$         & $0$         & \textcolor{green!50!black}{$\Theta(1)$}\\
$\quad D = (1/M)\sum_i\phi_{i,t}\,\Delta_t W^{3,i}\,\Delta_t h^{2,i}$              & $0$         & $0$         & \textcolor{green!50!black}{$\Theta(1)$}\\
$f_t = W_t^4 h_t^3$                                                                & $0$         & $\Theta(1)$ & \textcolor{green!50!black}{$\Theta(1)$}\\
$\quad$ effective: $\Delta_t W^4\,h_0^3$                                           & $0$         & $\Theta(1)$ & \textcolor{green!50!black}{$\Theta(1)$}\\
$\quad$ cross: $\Delta_t W^4\,\Delta_t h^3$                                        & $0$         & $0$         & \textcolor{green!50!black}{$\Theta(1)$}\\
\end{tabular}
\end{center}

\paragraph{Backward.}
\begin{center}
\renewcommand{\arraystretch}{1.2}
\begin{tabular}{l|l|l|l}
\textbf{Quantity} & $t=0$ & $t=1$ & $t=2$\\\hline
$\partial f_t/\partial h_t^3 = (\Delta_t W^4)^\top$                                                                                    & $0$ & $\Theta(1/N)$         & \textcolor{green!50!black}{$\Theta(1/N)$}\\
$\partial f_t/\partial h_t^{3,i} = (\phi_{i,t}/M)(\Delta_t W^4)^\top$                                                                  & $0$ & $\Theta(1/(MN))$      & \textcolor{green!50!black}{$\Theta(1/(MN))$}\\
$\partial f_t/\partial h_t^{2,i} = (\phi_{i,t}/M)(W_t^{3,i})^\top(\Delta_t W^4)^\top$                                                  & $0$ & $\Theta(1/(MN))$      & \textcolor{green!50!black}{$\Theta(1/(MN))$}\\
$\quad$ init: $(\phi_{i,t}/M)(W_0^{3,1})^\top(\Delta_t W^4)^\top$                                                                      & $0$ & $\Theta(1/(MN))$      & \textcolor{green!50!black}{$\Theta(1/(MN))$}\\
$\quad$ update: $(\phi_{i,t}/M)(\Delta_t W^{3,i})^\top(\Delta_t W^4)^\top$                                                             & $0$ & $0$                   & \textcolor{green!50!black}{$\Theta(1/(MN))$}\\
$\partial f_t/\partial \phi_{i,t} = (1/M)\langle h_t^{3,i}, \Delta_t W^4\rangle$                                                       & $0$ & $\Theta(1/M)$         & \textcolor{green!50!black}{$\Theta(1/M)$}\\
$(\partial f_t/\partial h_t^1)_\mathrm{exp}$                                                                                           & $0$ & $\Theta(1/N)$         & \textcolor{green!50!black}{$\Theta(1/N)$}\\
$\quad A_4 = (1/M)\sum_i\phi_{i,t}\,(W_0^{2,1})^\top(W_0^{3,1})^\top(\Delta_t W^4)^\top$                                               & $0$ & $\Theta(1/N)$         & \textcolor{green!50!black}{$\Theta(1/N)$}\\
$\quad A_5 = (1/M)\sum_i\phi_{i,t}\,(W_0^{2,1})^\top(\Delta_t W^{3,i})^\top(\Delta_t W^4)^\top$                                        & $0$ & $0$                   & \textcolor{green!50!black}{$\Theta(1/N)$}\\
$\quad A_6 = (1/M)\sum_i\phi_{i,t}\,(\Delta_t W^{2,i})^\top(W_0^{3,1})^\top(\Delta_t W^4)^\top$                                        & $0$ & $0$                   & \textcolor{green!50!black}{$\Theta(1/N)$}\\
$\quad E   = (1/M)\sum_i\phi_{i,t}\,(\Delta_t W^{2,i})^\top(\Delta_t W^{3,i})^\top(\Delta_t W^4)^\top$                                 & $0$ & $0$                   & \textcolor{green!50!black}{$\Theta(1/N)$}\\
$(\partial f_t/\partial h_t^1)_\mathrm{router} = (1/M)Q_t^\top v$                                                                      & $0$ & $\Theta(1/\sqrt{MN})$ & \textcolor{green!50!black}{$\Theta(1/\sqrt{MN})$}\\
$\quad$ init: $(1/M)Q_0^\top v$                                                                                                        & $0$ & $\Theta(1/\sqrt{MN})$ & \textcolor{green!50!black}{$\Theta(1/\sqrt{MN})$}\\
$\quad$ update: $(1/M)(\Delta_t Q)^\top v$                                                                                             & $0$ & $0$                   & \textcolor{green!50!black}{$\Theta(1/\sqrt{MN})$}\\
\end{tabular}
\end{center}

\newpage

\newpage
\section{DMFT Analysis for Regime I}\label{sec:dmft_regime1}

\subsection{Setup}

We use the following architecture:

\begin{equation}
\label{deep_experts_architecture}
    \begin{aligned}
        h_\mu^1 &=W^1x_\mu \\
        \psi_\mu &=\frac{1}{N} Q\sigma(h^1_\mu) \\
        \phi_\mu&=\text{softmax}(\psi_\mu)\\
        h^{2}_{\mu,i}&=\frac{1}{\sqrt{N}}W^{2}_i\sigma(h_\mu^{1}),\\
        h^{\ell+1}_{\mu,i}&=\frac{1}{\sqrt{N}}W^{\ell+1}_i\sigma(h_{\mu,i}^{\ell}), \quad \ell\in\{2,3,...,L-1\}\\
        h^L_\mu&=\sum_{i=1}^M\phi_\mu^i\cdot h_{\mu,i}^{L}\\
        h^{L+1}_\mu&=\frac{1}{\sqrt{N}}w^{L+1\mathsf{T}}h^L_\mu\\
        f_\mu&=\frac{1}{\gamma}h_\mu^{L+1}\\
        w_\alpha^{L+1}(0)&\sim\mathcal N(0,1)\\
        W_{\alpha\beta,i}^{\ell}(0)&\sim\mathcal N(0,1) \text{ for } \ell\in\{2,3,...,L\},i\in\{1,2,...,M\}\\
        W_{\alpha\beta}^{1}(0)&\sim\mathcal N(0,1)\\
        Q_{\alpha\beta}(0)&\sim\mathcal N(0,1)\\
        \eta&=\eta_0\gamma^2\\
        \gamma&=\gamma_0\sqrt N\\
        \eta_0,\gamma_0&\sim O(1)
    \end{aligned}
\end{equation}

$x_\mu\in\mathbb R^D$ is the embedding vector of input $\mu\in\{1,2,...,P\}$, $\psi_\mu\in\mathbb R^M$ gives the router's preferences for each of the $M$ experts, $\phi_\mu^i$ denotes the $i^{th} $ component of $\phi_\mu$, for $i\in\{1,2,...,M\}$, $\sigma$ is an element-wise nonlinearity, $Q\in\mathbb R^{M\times N}$, $w^4\in\mathbb R^N$, $W^{2}_i,W^{3}_i\in\mathbb R^{N\times N}$, $W^1\in\mathbb R^{N\times D}$, the preactivations $h$ are in $\mathbb R^N$, except for $h_\mu^4$, which is a scalar. We define $\text{softmax}:\mathbb R^p\to(0,1)^p,\, p>1$ as the function which takes a tuple $\bm z = (z_1,z_2,...,z_p)\in\mathbb R^p$, and computes each component of the vector $\text{softmax}(\bm z)\in(0,1)^p$ with $[\text{softmax}(\bm z)]_i=\frac{e^{\beta z_i}}{\sum_{j=1}^pe^{\beta z_j}}$, where $T=\frac1\beta$ is the temperature of the softmax. We will be interested in the low temperature limit $\beta\to\infty$, where softmax implements Top-1.

We denote for brevity $\bm\theta=\text{Vec}\{W^1,W^{\ell}_i,w^{L+1}, Q\}_{i,\ell}$, for $i\in\{1,2,...,M\}$ and $\ell\in\{2,3,...,L\}$, and define the gradients

\begin{equation}
    \label{deep_gradients_defns}
    \begin{aligned}
    &g_\mu^l=\sqrt N
    \frac{\partial h_\mu^{L+1}}{\partial h_\mu^l}, \text{   }l\in\text{Vec}\{1,(2,i),(3,i),...,(L,i),L,L+1\}_i\\
&g_\mu^\phi = \frac{1}{\sqrt N}\frac{\partial h_\mu^{L+1}}{\partial\phi_\mu}  
    \end{aligned}
\end{equation}

$g^1$ has two terms, as we must track the gradient flowing both through the router, and through the experts. Using \ref{deep_gradients}, the components of $g^\phi_\mu$ are 

\begin{equation}
    (g^\phi_\mu)_i = \frac{1}{\sqrt N} \frac{\partial h_\mu^{L+1}}{\partial\phi_\mu^i} = \frac{1}{\sqrt N} \frac{\partial h_\mu^{L+1}}{\partial \bm h^L_\mu}\cdot\frac{\partial \bm h_\mu^{L}}{\partial\phi_\mu^i} = \frac{1}{N}\bm g^L\cdot\bm h_{\mu,i}^{L} = O(1).
\end{equation}

In addition, we define the following \textit{pre-gradient fields} which are defined as the gradients with respect to the activations instead of the pre-activations. For linear layers, they clearly reduce to the gradient fields.
\begin{equation}
\label{deep_gradients}
    \begin{aligned}
        g^L_\mu&=z^L_\mu=w^{L+1}\\
        g^{L,i}_\mu&=z_{\mu,i}^{L}=g^L_\mu\phi^i_\mu\\
        g_{\mu,i}^{\ell}&=\dot\sigma(h_{\mu,i}^{\ell})\odot z_{\mu,i}^{\ell}, \quad z_{\mu,i}^{\ell}=\frac{1}{\sqrt N}W_i^{\ell+1\mathsf{T}}g_{\mu,i}^{\ell+1},\quad \ell\in\{2,3,4,...,L-1\}, i\in\{1,2,...,M\}\\
        (g_\mu^\phi)_i &= (z_\mu^\phi)_i = \frac{1}{N} g^L_\mu \cdot h_{\mu,i}^{L},\quad i\in\{1,2,...,M\}\\
        g_\mu^1&=\dot\sigma(h^1_\mu)\odot z_\mu^1, \quad z_\mu^1=\sum_{i=1}^M \frac{1}{\sqrt N} W^{2\mathsf{T}}_i g_{\mu,i}^{2} + Q^\top g^\phi_\mu
    \end{aligned}
\end{equation}
For convenience, we define $\tilde{z}_\mu^\phi := Q^\top g^\phi_\mu$ and $\tilde{z}_{\mu,i}^{1} := \frac{1}{\sqrt N} W^{2\mathsf{T}}_i g_{\mu,i}^{2}$. Then we have $z_\mu^1 = \sum_{i=1}^M \tilde{z}_{\mu,i}^{1}$ and $z_\mu^\phi = \tilde{z}_\mu^\phi$.

We define also the following feature and gradient kernels (The gradient kernels may be considered to be defined in terms of the z-objects via \ref{deep_gradients}) :

\begin{equation}
    \label{deep_kernels}
\begin{aligned}
\Phi_{\mu\nu}^{0} &:= x_\mu \cdot x_\nu, \\
\Phi_{\mu\nu}^{1}(s,t) &:= \frac{1}{N}\, \sigma(h_\mu^{1}(s))\cdot \sigma(h_\nu^{1}(t)), \\
\Phi_{\mu\nu,i}^{\ell}(s,t) &:= \frac{1}{N}\, \sigma\!\big(h_{\mu,i}^{\ell}(s)\big) \cdot \sigma\!\big(h_{\nu,i}^{\ell}(t)\big)\qquad\ell\in\{2,3,...,L-1\}, \\
\Phi_{\mu\nu}^{L}(s,t) &:= \frac{1}{N}\, h_\mu^{L}(s) \cdot h_\nu^{L}(t)\\
G_{\mu\nu}^{1}(s,t) &:= \frac1Ng^1_\mu(s) \cdot g^1_\nu(t), \\
G_{\mu\nu,i}^{\ell}(s,t) &:= \frac1Ng_{\mu,i}^{\ell}(s) \cdot g_{\nu,i}^{\ell}(t)\qquad \ell\in\{2,3,...,L\}\\
G_{\mu\nu}^\phi(s,t)&:= g^\phi_\mu(s)g^\phi_\nu(t)
\end{aligned}
\end{equation}

All of these kernels are normalised to be order unity in $N$.

\subsection{Dynamics}
\label{sec:dynamics}

We train using gradient flow with learning rate $\eta$ on the loss function

\begin{equation}
    \label{deep_loss}
    \mathcal L=\frac 1 P \sum_{\mu=1}^P\ell(f_\mu,y_\mu)
\end{equation}

This induces the following dynamics:

\begin{equation}
    \label{deep_learning_dynamics_sec1}
    \begin{aligned}
    &\frac{d\theta}{dt}
= \frac{\eta_0 \gamma}{P}
  \sum_{\mu} \Delta_{\mu}
  \frac{\partial h_{\mu}^{L+1}}{\partial \theta}\\
  &\Delta_\mu=-\frac{\partial\mathcal L}{\partial f_\mu}
    \end{aligned}
\end{equation}

In particular, for a MSE loss $\mathcal L=\frac{1}{P}\sum_{\nu}(y_\nu-f_\nu)^2$, we have $\Delta_\nu=2(y_\nu-f_\nu)$.

The logits update as 

\begin{equation}
\label{logit_updates}
\frac{d f_{\mu}(t)}{dt}
= \frac{\partial f_{\mu}(t)}{\partial \theta}\cdot \frac{d\theta}{dt}
= \frac{\eta}{P}
  \sum_{\alpha} \Delta_{\alpha} \,
  K^{\mathrm{NTK}}_{\mu \alpha}(t,t)
\end{equation}

\begin{equation}
\label{ntk_kernel}
\text{where} \quad
K^{\mathrm{NTK}}_{\mu \alpha}(t,s)
\equiv
\frac{\partial f_{\mu}(t)}{\partial \theta}
\cdot
\frac{\partial f_{\alpha}(s)}{\partial \theta}
\quad
\text{is the neural tangent kernel.}
\end{equation}

We can derive from this definition and the repeated use of chain rule that

\begin{equation}
    \gamma^2K^{NTK}_{\mu\nu}(t,t)=G_{\mu\nu}^1\Phi^0_{\mu\nu}+\sum_{i=1}^M[G_{\mu\nu,i}^{2}\Phi^1_{\mu\nu}+\sum_{\ell=3}^L G_{\mu\nu,i}^{\ell}\Phi^{\ell-1}_{\mu\nu,i}]+\Phi^{L}_{\mu\nu} + G^\phi_{\mu\nu}\Phi^1_{\mu\nu}
\end{equation}

Hence

\begin{equation}
\begin{aligned}
    \frac{\diff f_\mu(t)}{\diff t}&=\frac{\eta_0}{P}\sum_{\nu=1}^P\Big[G_{\mu\nu}^1(t,t)\Phi^0_{\mu\nu}+\sum_{i=1}^M[G_{\mu\nu,i}^{2}(t,t)\Phi^1_{\mu\nu}(t,t)+\sum_{\ell=3}^L G_{\mu\nu,i}^{\ell}(t,t)\Phi^{\ell-1}_{\mu\nu,i}(t,t)]+\Phi^{L}_{\mu\nu}(t,t)\\
    &+ G^\phi_{\mu\nu}(t,t)\Phi^1_{\mu\nu}(t,t)\Big]\Delta_\nu(t)
\end{aligned}
\end{equation}

Using the learning dynamics \ref{deep_learning_dynamics_sec1}, we find for the weight matrices

\begin{equation}
    \frac{d W^{1}(t)}{dt}
= \frac{\eta_0 \gamma_0}{P}
  \sum_{\mu} \Delta_{\mu} \, g_{\mu}^{1}(t)  x_{\mu}^\top
\end{equation}

\begin{equation}
\label{deep_dW2idt}
    \frac{d\,W^{2}_{i}(t)}{dt}
= \frac{\eta_0 \gamma_0}{\sqrt{N}\,P}
  \sum_{\mu} \Delta_{\mu}\, g_{\mu,i}^{2}  \sigma\!\big(h_{\mu}^{1}\big)^\top
\end{equation}

\begin{equation}
\label{deep_dWlidt}
    \frac{d\,W^{\ell}_{i}(t)}{dt}
= \frac{\eta_0 \gamma_0}{\sqrt{N}\,P}
  \sum_{\mu} \Delta_{\mu}\, g_{\mu,i}^{\ell} \sigma\big(h_{\mu,i}^{\ell-1}\big)^\top,\quad\ell\in\{3,4,...,L\}
\end{equation}

\begin{equation}
    \frac{d\,w^{L+1}(t)}{dt}
= \frac{\eta_0 \gamma_0 \sqrt{N}}{P}
  \sum_{\mu} \Delta_{\mu} \frac{\partial h_{\mu}^{L+1}}{\partial w^{L+1}}
= \frac{\eta_0 \gamma_0}{P} \sqrt{N}
  \sum_{\mu} \Delta_{\mu} \frac{1}{\sqrt{N}} h_{\mu}^L
= \frac{\gamma_0 \eta_0}{P}
  \sum_{\mu} \Delta_{\mu} h_{\mu}^L
\end{equation}

\begin{equation}
    \begin{aligned}
        \underbrace{\frac{dQ(t)}{dt}}_{\in\mathbb R^{M\times N}}
        &=\frac{\eta_0\gamma_0}{P}\sum_{\mu=1}^P\Delta_\mu(t) \underbrace{g^\phi_\mu(t)}_{\in\mathbb R^{M\times 1}}\underbrace{[\sigma(h^1_\mu(t))]^\top}_{\in\mathbb R^{1\times N}}
    \end{aligned}
\end{equation}

We can now integrate these expressions and use the definitions of the pre-activations from \ref{deep_experts_architecture} to obtain:

\begin{equation}
\label{deep_preactivations_integrated}
\begin{aligned}
    W^1(t)
&= W^1(0)
  + \int_0^t \! ds \, \frac{\eta_0 \gamma_0}{P}
    \sum_{\mu} \Delta_{\mu} \, g_{\mu}^1(s) \cdot x_{\mu}\\
\implies h_{\mu}^1(t)
&= W^1(0) x_{\mu}
  + \frac{\eta_0 \gamma_0}{P}
    \int_0^t \! ds
    \sum_{\nu} \Delta_{\nu} \, g_{\nu}^1(s) \, \Phi_{\mu\nu}^0\\
 W^{2}_{i}(t)
&= W^{2}_{i}(0)
  + \int_0^t \! ds \, \frac{\eta_0 \gamma_0}{\sqrt{N} P}
    \sum_{\mu} \Delta_{\mu} \, g_{\mu,i}^{2}(s) \sigma(h_{\mu}^{1}(s))^\top\\
\implies {h^{2}_{\mu, i}}(t)
&= \frac{1}{\sqrt{N}} W^{2}_{i}(0) \sigma(h_{\mu}^{1}(t))
  + \frac{\eta_0 \gamma_0}{P}
    \int_0^t \! ds
    \sum_{\nu} \Delta_{\nu} \, g_{\nu,i}^{2}(s)
    \, \Phi_{\mu \nu}^{1}(s,t)\\
W^{\ell}_{i}(t)
&= W^{\ell}_{i}(0)
  + \frac{\eta_0 \gamma_0}{\sqrt{N} P}
    \int_0^t \! ds
    \sum_{\mu} \Delta_{\mu} \, g_{\mu,i}^{\ell}(s) \, \sigma\!\big(h_{\mu,i}^{\ell-1}\big)^{\!T}\\
\implies h_{\mu,i}^{\ell}(t)&= \frac{1}{\sqrt{N}} W^{\ell}_{i}(0) \, \sigma\!\big(h_{\mu,i}^{\ell-1}(t)\big)
  + \frac{\eta_0 \gamma_0}{P}
    \int_0^t \! ds
    \sum_{\nu} \Delta_{\nu} \, g_{\nu,i}^{\ell}(s)
    \, \Phi_{\mu \nu,i}^{\ell-1}(s,t), \quad \ell\in\{3,4,...,L\}\\
Q(t)&= Q(0)
+ \frac{\eta_0 \gamma_0}{P}
  \int_{0}^{t} \! ds \,
  \sum_{\mu=1}^{P}
  \Delta_\mu(s)\,
  g^\phi_\mu(s)\,
  \bigl(h_\mu^{1}(s)\bigr)^{\top} \\
\implies \psi_\mu(t)
&= \frac{1}{N}\, Q(0)\, \sigma(h_\mu^{1}(t))
   + \frac{\eta_0 \gamma_0}{P}
     \int_{0}^{t} \! ds \,
     \sum_{\nu=1}^{P}
     \Delta_\nu(s)\,
     g^\phi_\nu(s)\,
     \Phi^1_{\mu\nu}(s,t)
    \end{aligned}
\end{equation}

This is useful, because we hope to average over the weights at initialization to obtain expressions in terms of only features. 
These expressions \ref{deep_preactivations_integrated} motivate the definition of the following fields:

\begin{equation}
\label{deep_chis}
    \begin{aligned}
        &\chi_\mu^1=W^1(0)x_\mu\in\mathbb R^N\\
        &\chi_{\mu,i}^{2}(t)=\frac{1}{\sqrt{N}} W^{2}_{i}(0) \sigma(h_{\mu}^{1}(t))\in\mathbb R^N\\
        &\chi_{\mu,i}^{\ell}(t)=\frac{1}{\sqrt{N}} W^{\ell}_{i}(0) \sigma(h_{\mu,i}^{\ell-1}(t))\in\mathbb R^N\quad\ell\in\{3,4,...,L\}\\
        &\chi^\phi_\mu(t)=\frac1N Q(0)\sigma(h_\mu^1(t))\in\mathbb R^M
    \end{aligned}
\end{equation}

We repeat the analysis of \ref{deep_preactivations_integrated} for the gradient fields which are defined as $z_{\mu}^{\ell}
= \frac{1}{\sqrt{N}}\, W^{\ell+1}(t)^{\top} \, g_{\mu}^{\ell+1}(t)$, where $\ell$ indexes the layer, and may involve an expert index too, and $z^\phi_\mu(t) =  Q(t)^\top g^\phi_\mu(t)$: In particular, we have $\forall i\in\{1,2,...,M\}$

\begin{equation}
\label{deep_gradients_fields_evolution}
    \begin{aligned}
        z_\mu^\phi(t) &=\xi^\phi_\mu(t) + \frac{\eta_0\gamma_0}{P}\int_0^tds\sum_{\nu=1}^P\Delta_\nu(s)\sigma(h^1_\nu(s))G_{\mu\nu}^\phi(t,s)  \\
        z_{\mu,i}^{1}(t)&=\xi_{\mu}^{1,i}(t)+ \frac{\eta_0 \gamma_0}{P}\int_0^t \! ds\sum_{\nu}\Delta_{\nu} \, \sigma(h^1_{\nu}(s)) \,G_{\mu \nu,i}^{2}(s,t)\\
        z_{\mu,i}^{\ell}(t)&=\xi_{\mu,i}^{\ell}(t)+ \frac{\eta_0 \gamma_0}{P}\int_0^t \! ds\sum_{\nu}\Delta_{\nu} \, \sigma(h^{\ell}_{\nu,i}(s)) \,G_{\mu \nu,i}^{\ell+1}(s,t)\quad\ell\in\{2,3,...,L-1\}\\
        z_{\mu,i}^{L}&=\phi_\mu^iz^L_\mu\\
        z_\mu^{L}(t)&=w_\mu^{L+1}(t)=\xi_\mu^{L}(t)+\frac{\gamma_0\eta_0}{P}\int_0^tds\sum_\nu\Delta_\nu h_\nu^L,
    \end{aligned}
\end{equation}

where we define the following fields:

\begin{equation}
\label{deep_xis}
    \begin{aligned}
        &\xi_\mu^\phi(t) = Q(0)^\top g_\mu^\phi(t)\\
        &\xi_{\mu,i}^{\ell}(t)=\frac{1}{\sqrt{N}} W^{\ell+1}_{i}(0)^\mathsf{T}  g_{\mu,i}^{\ell+1}(t)\quad\ell\in\{1,2,...,L-1\}\\
        &\xi_\mu^{L}=w^{L+1}(0)\sim\mathcal{N}(0,1)
    \end{aligned}
\end{equation}
Note that $\xi^{L}_\mu$ is in fact time-independent, and equal to the initialization of $w^{L+1}$. We know that this object has entries distributed as $\mathcal{N}(0,1)$, and we will also recover this from the DMFT for completeness.

Note also that the final gradient is $g^4_\mu = \sqrt N$, so we can think of the final equation in \ref{deep_gradients_fields_evolution} as involving the identity kernel.

Finally, we write

\begin{equation}
\begin{aligned}
\big[ g^{\phi} \big]_i
&=
\frac{1}{N}
\sum_{n=1}^{N}
g_n^{L}\,
\big[ h_i^{L} \big]_n
=
\frac{1}{N^{3/2}}
\sum_{n,m=1}^{N}
g_n^{L}
\big[ W_i^{L} \big]_{nm}\,
\sigma\!\big( \big[ h_i^{L-1} \big]_m \big)
\\[8pt]
\big[ g^{\phi}_{\mu}(t) \big]_i
&=
\xi^{g^{\phi}}_{\mu i}(t)\\
&
+
\frac{1}{N^{2}}
\frac{\eta_0 \gamma_0}{P}
\int_{0}^{t} ds
\sum_{\nu}
\Delta_{\nu}(s)
\sum_{n,m=1}^{N}
g_{\mu n}^{L}(t)
g_{\nu n}^{L}(s)
\phi_{\nu}^{i}(s)
\sigma\!\big( \big[ h_{\mu i}^{L-1}(t) \big]_m \big)\,
\sigma\!\big( \big[ h_{\nu i}^{L-1}(s) \big]_m \big)
\\[8pt]
&=
\xi^{g^{\phi}}_{\mu i}(t)
+
\frac{\eta_0 \gamma_0}{P}
\int_{0}^{t} ds
\sum_{\nu}
\Delta_{\nu}(s)\,
G_{\mu\nu}^{L}(t,s)\,
{\Phi}_{\mu\nu}^{L-1}(t,s)\,\phi^i_\nu(s)
\end{aligned}
\end{equation}

Where we define the field

\begin{equation}
\begin{aligned}
\xi^{g^{\phi}}_{\mu i}(t)
&=
\frac{1}{N^{3/2}}\,
g^{L\,\top}_\mu(t)\,
W_i^{L}(0)\,
\sigma\!\big( h_{\mu i}^{L-1} (t)\big).
\end{aligned}
\end{equation}

\subsection{Mean field theory}

If we can show that the kernels defined above concentrate, then our DMFT will be determined by the stochastic fields $\mathcal{F} = \{\chi_\mu^1,\chi_{\mu,i}^{2},\chi_{\mu,i}^{3},...,\chi_{\mu,i}^{L}\chi^\phi_\mu,\xi_{\mu,i}^{1},\xi_{\mu,i}^{2},...,\xi_{\mu,i}^{L-1}\xi_\mu^{L},\xi^\phi_\mu\}_{i\in\{1,2,...,M\}}$. Therefore to construct the DMFT, we must compute the moment generating functional of these stochastic processes.

For brevity, we define, alongside the set $\mathcal F$ of fields, the set $\mathcal S$ of sources, which contains a corresponding source $j$ (or $v$) for every $\chi$ (or $\xi$) in $\mathcal F$. So $\mathcal{S} = \{j_\mu^1,j_{\mu,i}^{2},j_{\mu,i}^{3},...,j_{\mu,i}^{L}j^\phi_\mu,v_{\mu,i}^{1},v_{\mu,i}^{2},...,v_{\mu,i}^{L-1}v_\mu^{L},v^\phi_\mu\}_{i\in\{1,2,...,M\}}$. This notation is constructed so that we can write $\mathcal F \cdot \mathcal S$ to denote $j_\mu^1(t)\cdot\chi_\mu^1(t)+j_{\mu,i}^{2}(t)\cdot\chi_{\mu,i}^{2}(t)+...+v_\mu^\phi(t)\cdot\xi_\mu^\phi(t)$. Note that $v^\phi_\mu\in\mathbb R^M$, but all the other dummy variables are in $\mathbb R^N$. We also define $\theta_0 = \mathrm{Vec}\!\left\{
  W^1(0), \, W^{2}_{i}(0),...,W^{L}_{i}(0) \,, w^{L+1}(0), \, Q(0)
\right\}_{i\in\{1,...,M\}}.$

The MGF with which we are concerned is therefore

\begin{equation}
    \label{deep_mgf}
    \begin{aligned}
    Z\!\left[\{ j,v\}\right]\Bigg\langle
  \exp\!\Bigg(
    \sum_{\mu}
    \int_{0}^{\infty} \! dt \,
    \mathcal F \cdot \mathcal S
  \Bigg)\Bigg\rangle_{\theta_0}
\end{aligned}
\end{equation}

$v^\phi_\mu\in\mathbb R^M$, but all the other dummy variables are in $\mathbb R^N$.

In order to integrate, we enforce definitions of our random fields using resolutions of the identity to obtain:

\begin{equation}
\begin{aligned}
Z=&\Bigg\langle\frac{1}{(2\pi)^{8N}}
\prod_{\mu=1}^P\prod_{i=1}^M\int_0^\infty 
dt  \;\Bigg\{\exp\!\Bigg(
    \sum_{\mu}
    \int_{0}^{\infty} \! dt \,
    \mathcal F \cdot \mathcal S
  \Bigg) + \\
  &\int d\mathcal F
\exp\Bigg[
i\Bigg(
\hat{\chi}_\mu^{1} \!\cdot\! 
\big( \chi_\mu^{1} - W^{1}(0) x_\mu \big)
+ \hat{\chi}_{\mu,i}^{2} \!\cdot\! 
\Big( \chi_{\mu,i}^{2} - \frac{1}{\sqrt{N}} W^{2}_{i}(0) \sigma(h_\mu^{1}(t)) \Big) \\
&\qquad
+ \sum_{\ell=3}^L\hat{\chi}_{\mu,i}^{\ell} \!\cdot\! 
\Big( \chi_{\mu,i}^{\ell} - \frac{1}{\sqrt{N}} W^{\ell}_{i}(0) 
\sigma(h_{\mu,i}^{\ell-1}(t)) \Big)
+ \sum_{\ell=1}^{L-1}\hat{\xi}_{\mu,i}^{\ell} \!\cdot\! 
\Big( \xi_{\mu,i}^{\ell} - \frac{1}{\sqrt{N}} 
W^{\ell+1}_{i}(0)^{\!\top} g_{\mu,i}^{\ell+1}(t) \Big)\\
&\qquad
+ \hat{\xi}_\mu^{L} \!\cdot\! 
\Big(\xi_\mu^{L}-w^{L+1}(0)^{\!\top})
+\hat\chi^\phi_\mu\cdot\bigg(\chi^\phi_\mu-\frac1NQ(0)\sigma(h^1_\mu(t))\bigg)
+\hat\xi^\phi_\mu\cdot\bigg(\xi^\phi_\mu-Q(0)^\top g^\phi_\mu(t)
\bigg)
\\
&\qquad
+ \hat\xi^{g^{\phi}}\,\cdot\,\left(\xi^{g^\phi}_\mu
-
\frac1{N^{3/2}}g^{L\top}_\mu W^L_i(0)\sigma(h_{\mu,i}^{L-1})
\right)
\Bigg)
\Bigg]\Bigg\}\Bigg\rangle_{\theta_0}
\end{aligned}
\end{equation}

We then integrate using the fact that for a Gaussian variable $z\sim\mathcal N(0,\sigma^2)$, $\langle e^{-ia\cdot z}\rangle_z\propto e^{-\frac12\sigma^2|a|^2}$ to obtain

\begin{equation}
\begin{aligned}
\label{integrated_Z}
Z
&\propto
   \prod_{\mu=1}^{P}\prod_{\nu=1}^{P}\prod_{i=1}^{M}
   \int_{0}^{\infty} dt
   \int
dt \int d\mathcal F\\[2pt]
&\quad\times
\exp\Bigg\{
-\frac{1}{2}\sum_{\mu,\nu=1}^{P}\int_0^\infty dt\int_0^\infty ds\bigg[
  \hat{\chi}_\mu^{1}(t)\!\cdot\!\hat{\chi}_\nu^{1}(s)\,\Phi_{\mu\nu}^{0}(t,s) \\
& \quad 
+ (\hat{\chi}_{\mu,i}^{2}(t)\!\cdot\!\hat{\chi}_{\nu,i}^{2}(s)+\frac1N\hat\chi^\phi_\mu(t)\cdot\hat\chi^\phi_\nu(s))\,\Phi_{\mu\nu}^{1}(t,s)+ \sum_{\ell=3}^L\hat{\chi}_{\mu,i}^{\ell}(s)\!\cdot\!\hat{\chi}_{\nu,i}^{\ell}(t)\,\Phi_{\mu\nu,i}^{\ell-1}(s,t)\\
&\quad
+\sum_{\ell=2}^L\hat{\xi}_{\mu,i}^{\ell}(s)\!\cdot\!\hat{\xi}_{\nu,i}^{\ell}(t)\,G_{\mu\nu,i}^{\ell}(s,t)
+ \hat{\xi}_\mu^{L+1}(s)\!\cdot\!\hat{\xi}_\nu^{L+1}(t) + \hat{\xi}_\mu^\phi(s)\!\cdot\!\hat{\xi}_\nu^{\phi}(t) G^\phi_{\mu\nu}(t,s)
\bigg]
\\
&\quad-i\sum_{\mu,\nu=1}^P\int_0^\infty dt\int_0^\infty ds\,\,\left[\sum_{\ell=2}^L\hat\chi_{\mu,i}^{\ell}(t)\cdot g_{\nu,i}^{\ell}(t)A_{\mu\nu,i}^{\ell}(t,t)+\hat\chi_\mu^{\phi}(t)\cdot g_\nu^{\phi}(t)A_{\mu\nu}^{\phi}(t,t)\right]
\\
&\quad
    +\sum_{\mu = 1}^{P}
    \int_{0}^{\infty} dt\int_0^\infty ds\,
    \Bigg[
    \sum_{\ell=2}^{L-1}\Big((j_{\mu,i}^{\ell}(t)+i\hat\chi_{\mu,i}^{\ell}(t))\cdot \chi_{\mu,i}^{\ell}(t)
    + (v_{\mu,i}^{\ell}(t)+i\hat\xi_{\mu,i}^{\ell})\cdot\xi_{\mu,i}^{\ell}(t)\Big) \\
    &
    +(j_\mu^1(t)+i\hat\chi_\mu^1(t))\cdot\chi_\mu^1(t)\quad+(v_\mu^{L}+i\hat\xi^{L}_\mu(t))\cdot\xi^{L}_\mu(t)\\
    &
+(j_{\mu,i}^{L}(t)+i\hat\chi_{\mu,i}^{L}(t))\cdot\chi_{\mu,i}^{L}(t) +(v_{\mu,i}^{1}(t)+i\hat\xi_{\mu,i}^{1}(t))\cdot\xi_{\mu,i}^{1}(t) +(v_\mu^{\phi}+i\hat\xi^{\phi}_\mu(t))\cdot\xi^{\phi}_\mu(t)\\
    &\qquad +(j_\mu^{\phi}+i\hat\chi^{\phi}_\mu(t))\cdot\chi^{\phi}_\mu(t)
    +(v_\mu^{g^\phi}+i\hat\xi^{g^\phi}_\mu(t))\cdot\xi^{g^\phi}_\mu(t)
    \Bigg]
\Bigg\}.
\end{aligned}
\end{equation}

Where we have defined the kernels

\begin{equation}
    \label{deep_A_kernels}
    A_{\mu\nu,i}^{\ell}(s,t)=\begin{cases}
        -\frac iN \hat\xi_{\mu,i}^{1}(s)\cdot\sigma(h^1_\nu(t))\quad \ell=2\\
        -\frac iN \hat\xi_{\mu,i}^{\ell-1}(s)\cdot\sigma(h^{\ell-1}_{\nu,i}(t))\quad \ell\in\{3,4,...,L\}
    \end{cases}
\end{equation}

\begin{equation}
    \label{deep_A_kernels_phi}
    A_{\mu\nu}^{\phi}(s,t)=
        -\frac iN \hat\xi_\mu^{\phi}(s)\cdot \sigma(h^1_\nu(t))
\end{equation}

There are also three terms arising from the coupling of $\xi^{g^\phi}$ to itself and to the forwards and backwards fields at layer L. These vanish as $\frac1N$, so have been excluded from \ref{integrated_Z} for brevity.

The $\Phi$ and $G$ kernels are required for computing the evolution of the function output, via the NTK. The kernels $A$ are not involved in the NTK, but rather arise from the coupling of the fields across a single layer's initial weight matrix. We now enforce the definitions of all three types of kernels using integral representations of Dirac delta-functions. In particular, for each pair $\mu,\nu$ of samples, each expert $i$ and each pair $t,s$ of times, we multiply by

\begin{equation}
\label{first_id}
1
= \int
  \frac{
    d\Phi_{\mu\nu}^{0}(t,s)\,
    d\hat{\Phi}_{\mu\nu}^{0}(t,s)
  }{2\pi i}
  \exp\!\left[
    \hat{\Phi}_{\mu\nu}^{0}(t,s)
    \left(
      \Phi_{\mu\nu}^{0}(t,s)
      -x_\mu
        \cdot
        x_\nu
    \right)
  \right]
\end{equation}
\begin{equation}
1
= \int
  \frac{
    d\Phi_{\mu\nu}^{1}(t,s)\,
    d\hat{\Phi}_{\mu\nu}^{1}(t,s)
  }{2\pi i\, N^{-1}}
  \exp\!\left[
    \hat{\Phi}_{\mu\nu}^{1}(t,s)
    \left(N
      \Phi_{\mu\nu}^{1}(t,s)
      -\sigma(h_{\mu}^{1}(t))
        \cdot
        \sigma\!\big(h_{\nu}^{1}(s)\big)
    \right)
  \right]
\end{equation}
\begin{equation}
\begin{aligned}
1
= \int
  \frac{
    d\Phi_{\mu\nu,i}^{\ell}(t,s)\,
    d\hat{\Phi}_{\mu\nu,i}^{\ell}(t,s)
  }{2\pi i\, N^{-1}}
  \exp\!\Big[
    \hat{\Phi}_{\mu\nu,i}^{\ell}(t,s)
    \big(N
      \Phi_{\mu\nu,i}^{\ell}(t,s)
      -& \sigma\!\big(h_{\mu,i}^{\ell}(t)\big)
        \cdot
        \sigma\!\big(h_{\nu,i}^{\ell}(s)\big)
    \big)
  \Big]\\
  &
  \quad\ell\in\{2,3,...,L-1\}
\end{aligned}
\end{equation}
\begin{equation}
1
= \int
  \frac{
    d\Phi_{\mu\nu}^{L}(t,s)\,
    d\hat{\Phi}_{\mu\nu}^{L}(t,s)
  }{2\pi i\, N^{-1}}
  \exp\!\left[
    \hat{\Phi}_{\mu\nu}^{L}(t,s)
    \left(N
      \Phi_{\mu\nu}^{L}(t,s)
      -h_{\mu}^{L}(t)
        \cdot
        h_{\nu}^{L}(s)
    \right)
  \right]
\end{equation}
\begin{equation}
1
= \int
  \frac{
    dG_{\mu\nu}^{1}(t,s)\,
    d\hat{G}_{\mu\nu}^{1}(t,s)
  }{2\pi i\, N^{-1}}
  \exp\!\left[
    \hat{G}_{\mu\nu}^{1}(t,s)
    \left(N
      G_{\mu\nu}^{1}(t,s)
      -g_{\mu}^{1}(t)
        \cdot
        g_{\nu}^{1}(s)
    \right)
  \right]
\end{equation}
\begin{equation}
\begin{aligned}
1
= \int
  \frac{
    dG_{\mu\nu,i}^{\ell}(t,s)\,
    d\hat{G}_{\mu\nu,i}^{\ell}(t,s)
  }{2\pi i\, N^{-1}}
  \exp\!\Big[
    \hat{G}_{\mu\nu,i}^{\ell}(t,s)
    \big(N
      G_{\mu\nu,i}^{\ell}(t,s)
      -&g_{\mu,i}^{\ell}(t)
        \cdot
        g_{\nu,i}^{\ell}(s)
    \big)
  \Big]\\
  &\ell\in\{2,3,...,L\}
\end{aligned}
\end{equation}
\begin{equation}
1
= \int
  \frac{
    dA_{\mu\nu,i}^{2}(t,s)\,
    dB_{\mu\nu,i}^{2}(t,s)
  }{2\pi i\, N^{-1}}
  \exp\!\left[
    -B_{\mu\nu,i}^{2}(t,s)
    \left(N
      A_{\mu\nu,i}^{2}(t,s)
      +i\hat\xi_{\,i}^{1}(t)
        \cdot
        \sigma(h_{\nu}^{1}(s))
    \right)
  \right]
\end{equation}

\begin{equation}
1
= \int
  \frac{
    dA_{\mu\nu}^{\phi}(t,s)\,
    dB_{\mu\nu}^{\phi}(t,s)
  }{2\pi i\, N^{-1}}
  \exp\!\left[
    -B_{\mu\nu}^{\phi}(t,s)
    \left(N
      A_{\mu\nu}^{\phi}(t,s)
      +i\hat\xi_{\mu}^{\phi}(t)
        \cdot
        \sigma(h_{\nu}^{1}(s))
    \right)
  \right]
\end{equation}

\begin{equation}
\begin{aligned}
1
= \int
  \frac{
    dA_{\mu\nu,i}^{\ell}(t,s)\,
    dB_{\mu\nu,i}^{\ell}(t,s)
  }{2\pi i\, N^{-1}}
  \exp\!\Big[
    -B_{\mu\nu,i}^{\ell}(t,s)
    \big(N
      A_{\mu\nu,i}^{\ell}(t,s)
      +i\hat\xi_{\mu,i}^{\ell-1}(t)&
        \cdot
        \sigma(h_{\nu,i}^{\ell-1}(s))
    \big)
  \Big]\\
  &\ell\in\{3,4,...,L\}
\end{aligned}
\end{equation}
\begin{equation}
\label{last_id}
1
= \int
  \frac{
    d G_{\mu\nu}^{\phi}(t,s)\,
    d\hat G_{\mu\nu}^{\phi}(t,s)
  }{2\pi i}
  \exp\!\left[
    \hat G_{\mu\nu}^{\phi}(t,s)
    \left(
      G_{\mu\nu}^{\phi}(t,s)
      +g_{\mu}^{\phi}(t)
        \cdot
        g_{\nu}^{\phi}(t)
    \right)
  \right]
\end{equation}

We call the conjugate kernel of $A$ $B$, rather than $\hat A$. This is because it will turn out to be equal to $\hat\chi\cdot g$, which appears in the gradient stream update equation in the final DMFT.

Multiplying in all these factors of unity, we arrive at a partition function which factorises over each of the N sites in each hidden layer. There are two things to note here:

\begin{enumerate}
    \item We want to treat g,h as scalars, so each of the resolutions of the identity \ref{first_id} to \ref{last_id} above will be duplicated N times leading to an extensive factor N, which pulls through to the front of the exponential, since all the neurons in a given layer are indistinguishable in the limit, so have the same $\Phi$ or $G$. That is, we can write for instance 
    \begin{equation}
        1
        = \int
          \frac{
            dG_{\mu\nu}^{1}(t,s)\,
            d\hat{G}_{\mu\nu}^{1}(t,s)
          }{2\pi i\, N^{-1}}
          \exp\!\left[
            N\hat{G}_{\mu\nu}^{1}(t,s)
            \left(
              G_{\mu\nu}^{1}(t,s)
              -g_{\mu}^{1}(t)g_{\nu}^{1}(s)\right)\right]
    \end{equation}

    Where $g^1_\mu\in\mathbb R$ are the components of the former vector of the same name, which are considered statistically indistinguishable in the limit.
    The exception to this however is \ref{last_id}, since $g^\phi_\mu\in\mathbb R^M$. M is not being taken to the infinite limit, so we cannot consider the M components of $g^\phi_\mu$ to be indistinguishable. Rather we must sum over them:
    \begin{equation}
    \label{last_id_modified}
    1
    = \int
    \frac{
    d G_{\mu\nu}^{\phi}(t,s)\,
    d\hat G_{\mu\nu}^{\phi}(t,s)
    }{2\pi i\, }
    \exp\!\left[
    \sum_{\beta=1}^M
\hat G_{\mu\nu}^{\phi}(t,s)
    \left(
    G_{\mu\nu}^{\phi}(t,s)
    +g_{\mu,\beta}^{\phi}(t)g_{\nu,\beta}^{\phi}(t)\right)\right].
    \end{equation}
    \item Each term in the argument of the exponential in \ref{integrated_Z} is a dot product of two vectors in $\mathbb R^N$, so contains N terms which are identical in the limit. We can therefore bring out a factor N here too. Once again, the exception is the term $(j_\mu^\phi(t)+i\hat\chi^\phi_\mu(t))\chi_\mu^\phi(t)$, which is repeated M times. Therefore when we bring a factor N out front of the action, we must include a prefactor $\frac M N$ with this term.
\end{enumerate}

Define the set $\mathcal K$ of all order parameters. This includes all the kernels defined above
.

\begin{equation}
    \label{deep_Z}
    \begin{aligned}
    Z\propto\int &\prod_{\mu,\nu,t,s,i}d\mathcal K\,
\exp({NS[\{\hat\Phi,\Phi,\hat G,G,A,B\}]})
\end{aligned}
\end{equation}

Where the DMFT action $S$is $O(1)$, and has the form

\begin{equation}
\label{action}
\adjustbox{max width=\textwidth}{$
\begin{aligned}
    S[\{\hat\Phi,\Phi,\hat G,G,A,B\}]&=
    \sum_{\mu,\nu}
  \int_{0}^{\infty} \! dt
  \int_{0}^{\infty} \! ds
  \Big[\hat{G}_{\mu\nu}^{1}(t,s) G_{\mu\nu}^{1}(t,s)
  + \sum_{\ell=2}^L\hat{G}_{\mu\nu,i}^{\ell}(t,s) G_{\mu\nu,i}^{\ell}(t,s)
  \\
  &\quad+\hat{\Phi}_{\mu\nu}^{1}(t,s) \Phi_{\mu\nu}^{1}(t,s)
  +\sum_{\ell=2}^{L-1}\hat{\Phi}_{\mu\nu,i}^{2}(t,s) \Phi_{\mu\nu,i}^{2}(t,s)
  +\hat{\Phi}_{\mu\nu}^{L}(t,s) \Phi_{\mu\nu}^{L}(t,s)\\
  &\quad-\sum_{L=2}^LB_{\mu\nu,i}^{\ell}(t,s) A_{\mu\nu,i}^{\ell}(t,s)
  +\frac1N\hat G_{\mu\nu}^\phi(t,s)G_{\mu\nu}^\phi(t,s)-B^\phi_{\mu\nu}(t,s)A^\phi_{\mu\nu}(t,s)
  \Big]
\\
&\quad+\frac1N\sum_{\alpha=1}^N\ln \mathcal{Z}_N[\Phi,\hat\Phi,G,\hat G,A,B,j_\alpha,v_\alpha]
+\frac1N\sum_{\beta=1}^M\ln \mathcal{Z}_M[\Phi^1,\hat G^\phi ,j^\phi_\beta]
\end{aligned}
$}
\end{equation}

such that it consists of inner products of the order parameters $\{\Phi,G,A\}$ and their duals $\{\hat\Phi,\hat G,B\}$, in addition to the single-site MGFs

\begin{equation}
\label{deep_single_site_Z}
\begin{aligned}
    \mathcal Z_N&[\{\Phi,\hat\Phi,G,\hat G,A,B,j,v\}]\\&=
    \prod_{\mu,\nu,i,t}
   \int_{0}^{\infty} dt
   \int
d\mathcal F \,\\
    &\times\exp\Bigg(+\sum_{\mu = 1}^{P}
    \int_{0}^{\infty} dt\int_0^\infty ds\,
    \Bigg[
    \sum_{\ell=2}^L\Big((j_{\mu,i}^{\ell}(t)+i\hat\chi_{\mu,i}^{\ell}(t))\cdot \chi_{\mu,i}^{\ell}(t)
    + (v_{\mu,i}^{\ell}(t)+i\hat\xi_{\mu,i}^{\ell})\cdot\xi_{\mu,i}^{\ell}(t)\Big) \\
    &
    +(j_\mu^1(t)+i\hat\chi_\mu^1(t))\cdot\chi_\mu^1(t)+(v_\mu^{L+1}+i\hat\xi^{L+1}_\mu(t))\cdot\xi^{L+1}_\mu(t)
+(j^\phi_\mu(t)+i\hat\chi^\phi_\mu(t))\cdot\chi^\phi_\mu(t)\\
&
+(v^\phi_\mu(t)+i\hat\xi^\phi_\mu(t))\cdot\xi^\phi_\mu(t)
    \Bigg]\Bigg)\\
    &\times
\exp\Bigg(
-\frac{1}{2}\sum_{\mu,\nu,i}\int_0^\infty dt\int_0^\infty ds\bigg[
  \hat{\chi}_\mu^{1}(t)\!\cdot\!\hat{\chi}_\nu^{1}(s)\,\Phi_{\mu\nu}^{0}(t,s)
+ (\hat{\chi}_{\mu,i}^{2}(t)\!\cdot\!\hat{\chi}_{\nu,i}^{2}(s)\\
&
+
\frac1N \hat\chi^\phi_\mu(t)\cdot\hat\chi_\mu^\phi(s))\Phi^1_{\mu\nu}(t,s) + \sum_{\ell=3}^L\hat{\chi}_{\mu,i}^{\ell}(s)\!\cdot\!\hat{\chi}_{\nu,i}^{\ell}(t)\,\Phi_{\mu\nu,i}^{\ell-1}(s,t)
+\sum_{\ell=2}^L\hat{\xi}_{\mu,i}^{\ell}(s)\!\cdot\!\hat{\xi}_{\nu,i}^{\ell}(t)\,G_{\mu\nu,i}^{\ell}(s,t) \\
&
+ \hat{\xi}_\mu^{L+1}(s)\!\cdot\!\hat{\xi}_\nu^{L+1}(t) + \hat\xi_\mu^\phi(t)\hat\xi_\nu^\phi(s)G_{\mu\nu}^\phi(t,s)
\bigg]
\\
&\quad-i\sum_{\mu,\nu=1}^P\int_0^\infty dt\int_0^\infty ds\sum_{\ell=2}^L\hat\chi_{\mu,i}^{\ell}(t)\cdot g_{\nu,i}^{\ell}(t)A_{\mu\nu,i}^{\ell}(t,t) \\
&+\,i\,\sum_\mu \int_0^\infty dt \left[
\hat h^3_\mu(t) h^3_\mu(t)-\sum_{i=1}^M\phi^i_\mu(t)\hat h^3_\mu(t)h^3_\mu(t)\right]
\\
&\times\exp\Bigg(-\sum_{\mu,\nu,i}\int_0^\infty dt\int_0^\infty dt\bigg[x_\mu x_\nu\hat\Phi^0_{\mu\nu}+\sigma(h^1_\mu(t))\sigma(h^1_\nu(s))\hat\Phi^1_{\mu\nu}(t,s)\\
&+\sum_{\ell=2}^{L-1}\sigma(h_{\mu,i}^{\ell}(t))\sigma(h_{\nu,i}^{\ell}(s))\hat\Phi_{\mu\nu,i}^{\ell}(t,s)+h^L_\mu(t)h^L_\nu(s)\hat\Phi^L_{\mu\nu}(t,s)
+g^1_\mu(t)g^1_\nu(s)\hat G^1_{\mu\nu}(t,s)
\\
&+\sum_{\ell=2}^{L}g_{\mu,i}^{\ell}(t)g_{\nu,i}^{\ell}(s)\hat G_{\mu\nu,i}^{\ell}(t,s)
+\hat\xi_{\mu,i}^{2}(t)\sigma(h^{1}_\nu(s))B^{2}_{\mu\nu,i}(t,s)\\
&+i\sum_{\ell=3}^{L}\hat\xi_{\mu,i}^{\ell}(t)\sigma(h^{\ell-1}_{\nu,i}(s))B_{\mu\nu,i}^{\ell}(t,s)+i\hat\xi^{\phi}_\mu(t)\sigma(h^{1}_\nu(s))B^{\phi}_{\mu\nu}(t,s)
\bigg]\Bigg)
\end{aligned}
\end{equation}

And

\begin{equation}
\label{ZM}
    \begin{aligned}
    \mathcal Z_M&[\{\Phi^1,\hat G^\phi,j^\phi_\beta\}]\\&=\prod_{\mu,\nu,t}\int_{0}^{\infty} dt\int d\hat\chi^\phi_{\mu,\beta}(t)\,d\chi_{\mu,\beta}^\phi(t) \,\text{exp}\Big((j_{\mu,\beta}^\phi(t)+i\hat\chi_{\mu,\beta}(t))\chi^\phi_\mu(t)
    +(v_\mu^{g^\phi}+i\hat\xi^{g^\phi}_\mu(t))\cdot\xi^{g^\phi}_\mu(t)
    \\
    &\qquad
    -\frac12\frac1N\hat\chi_{\mu,\beta}^\phi(t)\hat\chi_{\nu,\beta}^\phi(s)\Phi^1_{\mu\nu}(t,s)
    -g_{\mu,\beta}^\phi g_{\nu,\beta}^\phi\hat G^\phi_{\mu\nu}(t,s)-i\hat\chi_{\mu,\beta}^{\phi}(t)g_{\nu,\beta}^{\phi}(t)A_{\mu\nu}^{\phi}(t,t)\Big)
    \end{aligned}
\end{equation}
The sum over $\beta$ is an average over sites (rows of vectors in $\mathbb R^M$). We regard $h_\mu(t) $ and $g_\mu(t)$ as functions of $\chi$ and $\xi$.

It is manifest in the form of \ref{deep_Z} that the actual state of the system in the limit $N\to\infty$ will be a saddle point of the action. This is the point where $\delta S=0$ for \textit{any} variation of our many order parameters. That is, we impose the following $\forall t,s,\mu,\nu,i$:

\begin{equation}
\label{deep_saddle_equations}
\begin{aligned}
\frac{\delta \mathcal{S}}{\delta G_{\mu\nu}^{1}(t,s)} &= \hat{G}_{\mu\nu}^{1}(t,s)-\frac{1}{\mathcal Z}\frac{\delta\mathcal Z}{\delta G^1_{\mu\nu}(t,s)}=\hat{G}_{\mu\nu}^{1}(t,s)-0= 0\\
\frac{\delta \mathcal{S}}{\delta G_{\mu\nu,i}^{\ell}(t,s)} &= \hat{G}_{\mu\nu,i}^{\ell}(t,s)- \frac{1}{\mathcal Z}\frac{\delta\mathcal Z}{\delta G_{\mu\nu,i}^{\ell}(t,s)} = \hat{G}_{\mu\nu,i}^{\ell}(t,s)-\frac12\frac1N\sum_{\alpha=1}^N\langle\hat\xi_{\mu,i}^{\ell}(t)\hat\xi_{\nu,i}^{\ell}(s)\rangle_\alpha=0\\
\frac{\delta \mathcal{S}}{\delta \Phi_{\mu\nu}^{1}(t,s)} &= \hat{\Phi}_{\mu\nu}^{1}(t,s)-\frac{1}{\mathcal Z}\frac{\delta\mathcal Z}{\delta \Phi^{1}_{\mu\nu}(t,s)} =\hat{\Phi}_{\mu\nu}^{1}(t,s)-\frac1{2}(\frac1N\sum_{\alpha=1}^N\langle\hat\chi_{\mu,i}^{2}(t)\hat\chi_{\nu,i}^{2}(s)\rangle_\alpha\\
&\qquad\qquad\qquad\qquad\qquad\qquad\qquad+\frac1N\frac1N\sum_{\beta=1}^M\langle\hat\chi^\phi_\mu(t)\hat\chi^\phi_\nu(s)\rangle_\beta)= 0\\
\frac{\delta \mathcal{S}}{\delta \Phi_{\mu\nu,i}^{\ell}(t,s)} &= \hat{\Phi}_{\mu\nu,i}^{\ell}(t,s)-\frac{1}{\mathcal Z}\frac{\delta\mathcal Z}{\delta\Phi_{\mu\nu,i}^{\ell}(t,s)} =\hat{\Phi}_{\mu\nu,i}^{\ell}(t,s)-\frac12\frac1N\sum_{\alpha=1}^N\langle\hat\chi_{\mu,i}^{\ell}(t)\hat\chi_{\nu,i}^{\ell}(s)\rangle_\alpha =0\\
\frac{\delta \mathcal{S}}{\delta \Phi_{\mu\nu}^{L}(t,s)} &= \hat{\Phi}_{\mu\nu}^{L}(t,s)-\frac{1}{\mathcal Z}\frac{\delta\mathcal Z}{\delta\Phi^{L}_{\mu\nu}(t,s)} =\hat{\Phi}_{\mu\nu}^{L}(t,s)-0 =0\\
\frac{\delta \mathcal{S}}{\delta A_{\mu\nu,i}^{\ell}(t,s)} &= -B_{\mu\nu,i}^{\ell}(t,s)-\frac{1}{\mathcal Z}\frac{\delta\mathcal Z}{\delta A_{\mu\nu,i}^{\ell}(t,s)} =-B_{\mu\nu,i}^{\ell}(t,s)-i\frac1N\sum_{\alpha=1}^N\langle\hat\chi_{\mu,i}^{\ell}(t)g_{\nu,i}^{\ell}(s)\rangle_\alpha =0\quad\\
\frac{\delta \mathcal{S}}{\delta A_{\mu\nu}^{\phi}(t,s)} &= -B_{\mu\nu}^{\phi}(t,s)-\frac{1}{\mathcal Z}\frac{\delta\mathcal Z}{\delta A^{\phi}_{\mu\nu}(t,s)} =-B_{\mu\nu}^{\phi}(t,s)-i\frac1N\sum_{\beta=1}^M\langle\hat\chi_\mu^{\phi}(t)g_\nu^{\phi}(s)\rangle_\beta =0\\
\frac{\delta \mathcal{S}}{\delta G^\phi_{\mu\nu}(t,s)} &= \frac1N\hat{G}_{\mu\nu}^{\phi}(t,s)-\frac{1}{\mathcal Z}\frac{\delta\mathcal Z}{\delta G^{\phi}_{\mu\nu}(t,s)} =\frac1N\hat{G}_{\mu}^{\phi}(t,s)- \frac12\frac1N\sum_{\alpha=1}^N\langle\hat\xi_\mu^{\phi}(t)\hat\xi_\nu^{\phi}(s)\rangle_\alpha=0\\
\frac{\delta \mathcal{S}}{\delta \hat{\Phi}_{\mu\nu}^{1}(t,s)} &= \Phi_{\mu\nu}^{1}(t,s)
- \frac{1}{N} \sum_{\alpha=1}^{N} \langle \sigma(h_\mu^{1}(t) )\sigma(h_\nu^{1}(s)) \rangle_{\alpha} = 0\\
\frac{\delta \mathcal{S}}{\delta \hat{\Phi}_{\mu\nu,i}^{\ell}(t,s)}
&= \Phi_{\mu\nu,i}^{\ell}(t,s)
- \frac{1}{N} \sum_{\alpha=1}^{N} \langle \sigma(h_{\mu,i}^{\ell}(t))\, \sigma(h_{\nu,i}^{\ell}(s)) \rangle_{\alpha} = 0\quad\\
\frac{\delta \mathcal{S}}{\delta \hat{\Phi}_{\mu\nu}^{L}(t,s)}
&= \Phi_{\mu\nu}^{L}(t,s)
- \frac{1}{N} \sum_{\alpha=1}^{N} \langle h_\mu^{L}(t) h_\nu^{L}(s) \rangle_{\alpha} = 0
\end{aligned}
\end{equation}

\begin{equation}
\begin{aligned}
\frac{\delta \mathcal{S}}{\delta \hat{G}_{\mu\nu}^{1}(t,s)}
&= G_{\mu\nu}^{1}(t,s)
- \frac{1}{N} \sum_{\alpha=1}^{N} \langle g_\mu^{1}(t) g_\nu^{1}(s) \rangle_{\alpha} = 0\\
\frac{\delta \mathcal{S}}{\delta \hat{G}_{\mu\nu,i}^{\ell}(t,s)}
&= G_{\mu\nu,i}^{\ell}(t,s)
- \frac{1}{N} \sum_{\alpha=1}^{N} \langle g_{\mu,i}^{\ell}(t) g_{\nu,i}^{\ell}(s) \rangle_{\alpha} = 0\quad\\
\frac{\delta \mathcal{S}}{\delta B_{\mu\nu,i}^{2}(t,s)}
&= -A_{\mu\nu,i}^{2}(t,s)-i\frac{1}{N} \sum_{\alpha=1}^{N} \langle \hat\xi_{\mu,i}^{2}(t) \sigma(h_\nu^{1}(s)) \rangle_{\alpha} = 0\\
\frac{\delta \mathcal{S}}{\delta B_{\mu\nu,i}^{\ell}(t,s)}
&= -A_{\mu\nu,i}^{\ell}(t,s)-i\frac{1}{N} \sum_{\alpha=1}^{N} \langle \hat\xi_{\mu,i}^{\ell}(t) \sigma(h_{\nu,i}^{\ell-1}(s)) \rangle_{\alpha} = 0\\
\frac{\delta \mathcal{S}}{\delta B_{\mu\nu}^{\phi}(t,s)}
&= -A_{\mu\nu}^{\phi}(t,s)-i\frac{1}{N} \sum_{\alpha=1}^{N} \langle \hat\xi_\mu^{\phi}(t) \sigma(h_\nu^{1}(s)) \rangle_{\alpha} = 0\\
\frac{\delta \mathcal{S}}{\delta \hat{G}_{\mu\nu}^{\phi}(t,s)}
&= \frac1NG_{\mu\nu}^{\phi}(t,s)
- \frac1N\sum_{\beta=1}^{M} \langle g_\mu^{\phi}(t) g_\nu^{\phi}(s) \rangle_{\beta} = 0\\
\end{aligned}
\end{equation}

Here the average $\langle\rangle_\alpha$ denotes the $\alpha^{\text{th}}$ single-site average of an observable $\mathcal O(\{\chi,\xi, u\})$, defined as:

\begin{equation}
\label{defn_average}
    \langle\mathcal O(\{\chi,\xi\})\rangle_\alpha = \frac{1}{\mathcal{Z}[j_\alpha, v_\alpha]}\int\prod_\mu\prod_{\text{layers}}d\chi_\mu^\ell d\xi_\mu^\ell \exp(-\mathcal H[\{\chi,\xi\},\{j,v\}]) \mathcal O(\{\chi,\xi\})
\end{equation}

Where $\mathcal H$ is the logarithm of the integrand in \ref{deep_single_site_Z} or \ref{ZM}.

At zero source $j,v\to0$, all single-site averages over N terms are equivalent, and we can write (for instance) $\Phi^1_{\mu\nu}=\langle \sigma(h^1_\mu)\,\sigma(h^1_\nu)\,\rangle$, where $\langle\cdot\rangle$ is the average over single-site distributions for $j,v\to0$. The sum $\frac1N\frac1N\sum_{\beta=1}^M\langle\hat\chi^\phi_\mu(t)\hat\chi^\phi_\nu(s)\rangle_\beta\to0$ in the limit, but notably, $\frac{1}{N^2} \sum_{\beta=1}^{M} \langle g_\mu^{\phi}(t) g_\nu^{\phi}(s) \rangle_{\beta}$ remains $O(1)$ but doesn't concentrate. We have
\begin{equation}
\label{deep_saddle_equations_vanishing_dummy}
\begin{aligned}
\frac{\delta \mathcal{S}}{\delta G_{\mu\nu}^{1}(t,s)} &= \hat{G}_{\mu\nu}^{1}(t,s)-\frac{1}{\mathcal Z}\frac{\delta\mathcal Z}{\delta G^1_{\mu\nu}(t,s)}=\hat{G}_{\mu\nu}^{1}(t,s)-0= 0\\
\frac{\delta \mathcal{S}}{\delta G_{\mu\nu,i}^{\ell}(t,s)} &= \hat{G}_{\mu\nu,i}^{\ell}(t,s)- \frac{1}{\mathcal Z}\frac{\delta\mathcal Z}{\delta G_{\mu\nu,i}^{\ell}(t,s)} = \hat{G}_{\mu\nu,i}^{\ell}(t,s)-\frac12\langle\hat\xi_{\mu,i}^{\ell}(t)\hat\xi_{\nu,i}^{\ell}(s)\rangle=0\\\
\frac{\delta \mathcal{S}}{\delta \Phi_{\mu\nu}^{0}} &= \hat{\Phi}_{\mu\nu}^{0}-\frac{1}{\mathcal Z}\frac{\delta\mathcal Z}{\delta\Phi^0_{\mu\nu}} =\hat{\Phi}_{\mu\nu}^{0}-\frac1{2N}\langle\hat\chi_\mu^1(t)\hat\chi_\nu^1(s)\rangle= 0\\
\frac{\delta \mathcal{S}}{\delta \Phi_{\mu\nu}^{1}(t,s)} &= \hat{\Phi}_{\mu\nu}^{1}(t,s)-\frac{1}{\mathcal Z}\frac{\delta\mathcal Z}{\delta \Phi^{1}_{\mu\nu}(t,s)} =\hat{\Phi}_{\mu\nu}^{1}(t,s)-\frac1{2}(\langle\hat\chi_{\mu,i}^{2}(t)\hat\chi_{\nu,i}^{2}(s)\rangle= 0\\
\frac{\delta \mathcal{S}}{\delta \Phi_{\mu\nu,i}^{\ell}(t,s)} &= \hat{\Phi}_{\mu\nu,i}^{\ell}(t,s)-\frac{1}{\mathcal Z}\frac{\delta\mathcal Z}{\delta\Phi_{\mu\nu,i}^{\ell}(t,s)} =\hat{\Phi}_{\mu\nu,i}^{\ell}(t,s)-\frac12 \langle\hat\chi_{\mu,i}^{\ell}(t)\hat\chi_{\nu,i}^{\ell}(s)\rangle =0\\\
\frac{\delta \mathcal{S}}{\delta \Phi_{\mu\nu}^{L}(t,s)} &= \hat{\Phi}_{\mu\nu}^{L}(t,s)-\frac{1}{\mathcal Z}\frac{\delta\mathcal Z}{\delta\Phi^{L}_{\mu\nu}(t,s)} =\hat{\Phi}_{\mu\nu}^{L}(t,s)-0 =0\\
\frac{\delta \mathcal{S}}{\delta A_{\mu\nu,i}^{\ell}(t,s)} &= -B_{\mu\nu,i}^{\ell}(t,s)-\frac{1}{\mathcal Z}\frac{\delta\mathcal Z}{\delta A_{\mu\nu,i}^{\ell}(t,s)} =-B_{\mu\nu,i}^{\ell}(t,s)-i\langle\hat\chi_{\mu,i}^{\ell}(t)g_{\nu,i}^{\ell}(s)\rangle =0\\
\frac{\delta \mathcal{S}}{\delta A_{\mu\nu}^{\phi}(t,s)} &= -B_{\mu\nu}^{\phi}(t,s)-\frac{1}{\mathcal Z}\frac{\delta\mathcal Z}{\delta A^{\phi}_{\mu\nu}(t,s)} =-B_{\mu\nu}^{\phi}(t,s)=0\\
\frac{\delta \mathcal{S}}{\delta G^\phi_{\mu\nu}(t,s)} &= \frac1N\hat{G}_{\mu\nu}^{\phi}(t,s)-\frac{1}{\mathcal Z}\frac{\delta\mathcal Z}{\delta G^{\phi}_{\mu\nu}(t,s)} =\frac1N\hat{G}_{\mu}^{\phi}(t,s)- \frac12\langle\hat\xi_\mu^{\phi}(t)\hat\xi_\nu^{\phi}(s)\rangle=0\\
\frac{\delta \mathcal{S}}{\delta \hat{\Phi}_{\mu\nu}^{1}(t,s)} &= \Phi_{\mu\nu}^{1}(t,s)
- \langle \sigma(h_\mu^{1}(t) )\sigma(h_\nu^{1}(s)) \rangle = 0\\
\frac{\delta \mathcal{S}}{\delta \hat{\Phi}_{\mu\nu,i}^{\ell}(t,s)}
&= \Phi_{\mu\nu,i}^{\ell}(t,s)
-\langle \sigma(h_{\mu,i}^{\ell}(t))\, \sigma(h_{\nu,i}^{\ell}(s)) \rangle = 0\\
\frac{\delta \mathcal{S}}{\delta \hat{\Phi}_{\mu\nu}^{L}(t,s)}
&= \Phi_{\mu\nu}^{L}(t,s)
-\langle h_\mu^{L}(t) h_\nu^{L}(s) \rangle = 0\\
\frac{\delta \mathcal{S}}{\delta \hat{G}_{\mu\nu}^{1}(t,s)}
&= G_{\mu\nu}^{1}(t,s)
- \langle g_\mu^{1}(t) g_\nu^{1}(s) \rangle = 0\\
\frac{\delta \mathcal{S}}{\delta \hat{G}_{\mu\nu,i}^{\ell}(t,s)}
&= G_{\mu\nu,i}^{\ell}(t,s)
-\langle g_{\mu,i}^{\ell}(t) g_{\nu,i}^{\ell}(s) \rangle = 0\\
\frac{\delta \mathcal{S}}{\delta B_{\mu\nu,i}^{2}(t,s)}
&= -A_{\mu\nu,i}^{2}(t,s)-i \langle \hat\xi_{\mu,i}^{2}(t) \sigma(h_\nu^{1}(s)) \rangle = 0\\
\frac{\delta \mathcal{S}}{\delta B_{\mu\nu,i}^{\ell}(t,s)}
&= -A_{\mu\nu,i}^{\ell}(t,s)-i \langle \hat\xi_{\mu,i}^{\ell}(t) \sigma(h_{\nu,i}^{\ell-1}(s)) \rangle = 0\\
\frac{\delta \mathcal{S}}{\delta B_{\mu\nu}^{\phi}(t,s)}
&= -A_{\mu\nu}^{\phi}(t,s)-i \langle \hat\xi_\mu^{\phi}(t) \sigma(h_\nu^{1}(s)) \rangle = 0\\
\frac{\delta \mathcal{S}}{\delta \hat{G}_{\mu\nu}^{\phi}(t,s)}
&= G_{\mu\nu}^{\phi}(t,s) - \sum_{\beta=1}^{M} \langle g_\mu^{\phi}(t) g_\nu^{\phi}(s) \rangle_{\beta}= 0\\
\end{aligned}
\end{equation}

\subsection{Averages of dual variables}
\label{subsec:dualaverages}
We see that the dual variables $\{\hat\Phi,\hat G,B\}$ are given in terms of such correlators as $\langle\hat\chi\hat\chi\rangle$. We will now show that these vanish, finding along the way expressions for the fields $\{\chi,\xi\}$, as well as the kernels $\{A,B\}$.

It will serve brevity to work with a vectorised notation. To wit: let $\bm\chi^{\ell}_{i}=\text{Vec}\{\chi_{\mu,i}^{\ell}(t)\}_{\mu\in\{1,...,P\},t\in\mathbb R^+}$ denote the vectorization of the stochastic field over different samples and times, with analogous objects defined for the other fields. We denote the dot product between such objects $\bm a\cdot\bm b=\sum_{\mu=1}^P\int_0^\infty a_\mu(t)b_\mu(t)$. A similar procedure is applied to matrices by defining $\bm\Phi=\text{Mat}\{\Phi_{\mu\nu}\}_{\mu\nu\in\{1,...,P\},ts\in\mathbb R^+}$, with the appropriate matrix product
defined as $[\bm A\bm b]_{\mu,t}=\int_0^tds\frac1P\sum_{\nu=1}^P A_{\mu\nu}(t,s)b_\nu(s)$.

With this notation in place, we can write $\langle\hat{\bm\chi}^{\ell}_{i}\hat{\bm\chi}^{\ell}_{i}\rangle$ for $\ell\in\{3,...,L-1\}$ in terms of the moment generating function for $\hat{\bm\chi}^{\ell}_{i}$, and the dummy field $\bm w$

\begin{equation}
\label{chili_correlator}
\adjustbox{max width=\textwidth}{$
    \begin{aligned}
        \langle\hat{\bm\chi}^{\ell}_{i}\hat{\bm\chi}^{\ell}_{i}\rangle
        &=-\,\frac{\partial^2}{\partial \boldsymbol{w}\,\partial \boldsymbol{w}^\top}\left\langle \exp\!\big(i\,\boldsymbol{w}\cdot\hat{\boldsymbol{\chi}}^{\,\ell}_{i}\big) \right\rangle \Big|_{\boldsymbol{w}=0}\\
        &=-\frac{\partial^{2}}{\partial \boldsymbol{w}\,\partial\boldsymbol{w}^{\top}}
\Bigg[\frac{1}{\mathcal Z}\!\int d\hat{\boldsymbol{\chi}}^{\,\ell}_{i}\,d\boldsymbol{\chi}^{\,\ell}_{i}\;
  \exp\!\big(i\,\boldsymbol{w}\!\cdot\!\hat{\boldsymbol{\chi}}^{\,\ell}_{i}\big)\,
  \exp\!\big(-\mathcal H\big)\Bigg]\Bigg|_{\boldsymbol{w}=0}
\\
&= -\frac{1}{\mathcal Z}\frac{\partial^{2}}{\partial \boldsymbol{w}\,\partial \boldsymbol{w}^{\top}}
\int d\hat{\boldsymbol{\chi}}^{\,\ell}_{i}\,d\boldsymbol{\chi}^{\,\ell}_{i}\;
\exp\!\left(
-\tfrac12\,\hat{\boldsymbol{\chi}}^{\,\ell\top}_{i}
\boldsymbol{\Phi}^{\,\ell-1}_{i}
\hat{\boldsymbol{\chi}}^{\,\ell}_{i}
+ i\big(-\boldsymbol{\chi}^{\,\ell}_{i}\!+\!\boldsymbol{w}
  + \mathbf{A}^{\ell}_{i}\mathbf{g}^{\,\ell}_{i}\big)\!\cdot\!
  \hat{\boldsymbol{\chi}}^{\,\ell}_{i}
\right)\Bigg|_{\boldsymbol{w}=0}
\\
&= -\frac{1}{\mathcal Z}\frac{\partial^{2}}{\partial \boldsymbol{w}\,\partial \boldsymbol{w}^{\top}}
\int d\boldsymbol{\chi}^{\,\ell}_{i}\;
\exp\!\left(
-\tfrac12\,
\big(-\boldsymbol{\chi}^{\,\ell}_{i}+\boldsymbol{w}+\mathbf{A}^{\ell}_{i}\mathbf{g}^{\,\ell}_{i}\big)^{\!\top}
[\boldsymbol{\Phi}^{\,\ell-1}_{i}]^{-1}
\big(-\boldsymbol{\chi}^{\,\ell}_{i}+\boldsymbol{w}+\mathbf{A}^{\ell}_{i}\mathbf{g}^{\,\ell}_{i}\big)
\right)\Bigg|_{\boldsymbol{w}=0}
\\
&=\frac{1}{\mathcal Z}\!\int d\boldsymbol{\chi}^{\,\ell}_{i}\;
\exp\!\left(
-\tfrac12\,
\big(-\boldsymbol{\chi}^{\,\ell}_{i}+\boldsymbol{w}+\mathbf{A}^{\ell}_{i}\mathbf{g}^{\,\ell}_{i}\big)^{\!\top}
[\boldsymbol{\Phi}^{\,\ell-1}_{i}]^{-1}
\big(-\boldsymbol{\chi}^{\,\ell}_{i}+\boldsymbol{w}+\mathbf{A}^{\ell}_{i}\mathbf{g}^{\,\ell}_{i}\big)
\right)
\\[-1mm]
&\qquad\times\left(
[\boldsymbol{\Phi}^{\,\ell-1}_{i}]^{-1}
-
[\boldsymbol{\Phi}^{\,\ell-1}_{i}]^{-1}
\big(-\boldsymbol{\chi}^{\,\ell}_{i}+\boldsymbol{w}+\mathbf{A}^{\ell}_{i}\mathbf{g}^{\,\ell}_{i}\big)
\big(-\boldsymbol{\chi}^{\,\ell}_{i}+\boldsymbol{w}+\mathbf{A}^{\ell}_{i}\mathbf{g}^{\,\ell}_{i}\big)^{\!\top}
[\boldsymbol{\Phi}^{\,\ell-1}_{i}]^{-1}
\right)\Bigg|_{\boldsymbol{w}=0}
\\
&= [\boldsymbol{\Phi}^{\,\ell-1}_{i}]^{-1}
-
[\boldsymbol{\Phi}^{\,\ell-1}_{i}]^{-1}
\Big\langle
\big(\boldsymbol{\chi}^{\,\ell}_{i}-\mathbf{A}^{\ell}_{i}\mathbf{g}^{\,\ell}_{i}\big)
\big(\boldsymbol{\chi}^{\,\ell}_{i}-\mathbf{A}^{\ell}_{i}\mathbf{g}^{\,\ell}_{i}\big)^{\!\top}
\Big\rangle
[\boldsymbol{\Phi}^{\,\ell-1}_{i}]^{-1}
    \end{aligned}
$}
\end{equation}

Where we use the definition \ref{defn_average} of the average, and retain only the relevant parts of the integral. Through similar derivations, we obtain also the following:

\begin{equation}
    \label{chi2_correlator}
    \langle\hat{\bm\chi}^{2}_{i}\hat{\bm\chi}^{2}_{i}\rangle= [\boldsymbol{\Phi}^{1}]^{-1}-[\boldsymbol{\Phi}^{1}]^{-1}
\Big\langle
\big(\boldsymbol{\chi}^{2}_{i}-\mathbf{A}^{2}_{i}\mathbf{g}^{2}_{i}\big)
\big(\boldsymbol{\chi}^{2}_{i}-\mathbf{A}^{2}_{i}\mathbf{g}^{2}_{i}\big)^{\!\top}
\Big\rangle
[\boldsymbol{\Phi}^{1}]^{-1}
\end{equation}

\begin{equation}
    \label{chi1_correlator}
    \langle\hat{\bm\chi}^{1}\hat{\bm\chi}^{1}\rangle= \left[\boldsymbol{\Phi}^{0}\right]^{-1}-\left[\boldsymbol{\Phi}^{0}\right]^{-1}
\Big\langle
\boldsymbol{\chi}^{1}
\boldsymbol{\chi}^{1\top}
\Big\rangle
\left[\boldsymbol{\Phi}^{0}\right]^{-1}
\end{equation}

\begin{equation}
    \label{xili_correlator}
    \langle\hat{\bm\xi}^{\ell}_{i}\hat{\bm\xi}^{\ell}_{i}\rangle= [\boldsymbol{G}^{\,\ell}_{i}]^{-1}-[\boldsymbol{G}^{\,\ell}_{i}]^{-1}
\Big\langle
\big(\boldsymbol{\xi}^{\,\ell}_{i}-\mathbf{B}^{\ell}_{i}\mathbf{g}^{\,\ell}_{i}\big)
\big(\boldsymbol{\xi}^{\,\ell}_{i}-\mathbf{B}^{\ell}_{i}\mathbf{g}^{\,\ell}_{i}\big)^{\!\top}
\Big\rangle
[\boldsymbol{G}^{\ell}_{i}]^{-1}\quad\ell\in\{2,3,...,L\}
\end{equation}

\begin{equation}
    \label{xili_correlator_phi}
    \langle\hat{\bm\xi}^{\phi}\hat{\bm\xi}^{\phi}\rangle= [\boldsymbol{G}^{\,\phi}]^{-1}-[\boldsymbol{G}^{\,\phi}]^{-1}
\Big\langle
\big(\boldsymbol{\xi}^{\,\phi}-\mathbf{B}^{\phi}\mathbf{g}^{\,\phi}\big)
\big(\boldsymbol{\xi}^{\,\phi}-\mathbf{B}^{\phi}\mathbf{g}^{\,\phi}\big)^{\!\top}
\Big\rangle
[\boldsymbol{G}^{\phi}]^{-1}
\end{equation}

\begin{equation}
\adjustbox{max width=\textwidth}{$
\begin{aligned}
-\,i\,\big\langle \hat{\boldsymbol{\chi}}^{\,\ell}_{i}\,\mathbf{g}^{\,\ell}_{i} \big\rangle
&= \frac{\partial}{\partial \mathbf{w}}\,
   \Big\langle \exp\!\big(-i\,\mathbf{w}\!\cdot\!\hat{\boldsymbol{\chi}}^{\,\ell}_{i}\big)\;
   \mathbf{g}^{\,\ell\top}_{i} \Big\rangle\Big|_{\mathbf{w}=0}
\\
&= \frac{1}{\mathcal Z}\,\frac{\partial}{\partial \mathbf{w}}
   \int d\hat{\boldsymbol{\chi}}^{\,\ell}_{i}\,d\boldsymbol{\chi}^{\,\ell}_{i}\;
   \exp\!\left(
   -\tfrac12\,\hat{\boldsymbol{\chi}}^{\,\ell\top}_{i}\boldsymbol{\Phi}^{\,\ell-1}_{i}\hat{\boldsymbol{\chi}}^{\,\ell}_{i}
   + i\big(\boldsymbol{\chi}^{\,\ell}_{i}\!+\!\mathbf{w}
   - \mathbf{A}^{\,\ell}_{i}\mathbf{g}^{\,\ell}_{i}\big)\!\cdot\!\hat{\boldsymbol{\chi}}^{\,\ell}_{i}
   \right)
   \mathbf{g}^{\,\ell\top}_{i}\Big|_{\mathbf{w}=0}
\\
&= \frac{1}{\mathcal Z}\,\frac{\partial}{\partial \mathbf{w}}
   \int d\boldsymbol{\chi}^{\,\ell}_{i}\;
   \exp\!\left(
   -\tfrac12\,
   \big(\boldsymbol{\chi}^{\,\ell}_{i}+\mathbf{w}-\mathbf{A}^{\,\ell}_{i}\mathbf{g}^{\,\ell}_{i}\big)^{\!\top}
   [\boldsymbol{\Phi}^{\,\ell-1}_{i}]^{-1}
   \big(\boldsymbol{\chi}^{\,\ell}_{i}+\mathbf{w}-\mathbf{A}^{\,\ell}_{i}\mathbf{g}^{\,\ell}_{i}\big)
   \right)
   \mathbf{g}^{\,\ell\top}_{i}\Big|_{\mathbf{w}=0}
\\
&= -\frac{1}{\mathcal Z}
   \int d\boldsymbol{\chi}^{\,\ell}_{i}\;
   \exp\!\left(
   -\tfrac12\,
   \big(\boldsymbol{\chi}^{\,\ell}_{i}+\mathbf{w}-\mathbf{A}^{\,\ell}_{i}\mathbf{g}^{\,\ell}_{i}\big)^{\!\top}
   [\boldsymbol{\Phi}^{\,\ell-1}_{i}]^{-1}
   \big(\boldsymbol{\chi}^{\,\ell}_{i}+\mathbf{w}-\mathbf{A}^{\,\ell}_{i}\mathbf{g}^{\,\ell}_{i}\big)
   \right)
\\[-1mm]
&\qquad\qquad\qquad\qquad\times
   [\boldsymbol{\Phi}^{\,\ell-1}_{i}]^{-1}
   \big(\boldsymbol{\chi}^{\,\ell}_{i}+\mathbf{w}-\mathbf{A}^{\,\ell}_{i}\mathbf{g}^{\,\ell}_{i}\big)
   \mathbf{g}^{\,\ell\top}_{i}
   \Big|_{\mathbf{w}=0}
\\
&= [\boldsymbol{\Phi}^{\,\ell-1}_{i}]^{-1}
   \Big\langle
   \big(\boldsymbol{\chi}^{\,\ell}_{i}-\mathbf{A}^{\,\ell}_{i}\mathbf{g}^{\,\ell}_{i}\big)\,
   \mathbf{g}^{\,\ell\top}_{i}
   \Big\rangle \quad\ell\in\{2,3,...,L\}
\end{aligned}
$}
\end{equation}

\begin{equation}
\label{xi_sigma_correlator}
-\,i\,\Big\langle \hat{\boldsymbol{\xi}}^{\,\ell}_{i}\,
   \boldsymbol{\sigma}\!\left(\mathbf{h}^{\,\ell-1}_{i}\right) \Big\rangle
=
\big[\mathbf{G}^{\,\ell}_{i}\big]^{-1}
\Big\langle
\big(\boldsymbol{\xi}^{\,\ell}_{i} - \mathbf{B}^{\,\ell}_{i}
   \boldsymbol{\sigma}\!\left(\mathbf{h}^{\,\ell-1}_{i}\right)\big)\,
   \boldsymbol{\sigma}\!\left(\mathbf{h}^{\,\ell-1}_{i}\right)
\Big\rangle\quad\ell\in\{3,4,...,L\}
\end{equation}

\begin{equation}
-\,i\,\Big\langle \hat{\boldsymbol{\xi}}^{\phi}\,
   \sigma(\mathbf{h}^{1}) \Big\rangle
=
\big[\mathbf{G}^{\phi}\big]^{-1}
\Big\langle
\big(\boldsymbol{\xi}^{\phi} - \mathbf{B}^{\phi}
   \sigma(\mathbf{h}^{1})\big)\,
   \sigma(\mathbf{h}^{1})
\Big\rangle
\end{equation}

\begin{equation}
\label{final_correlator}
-\,i\,\Big\langle \hat{\boldsymbol{\xi}}^{\,2,i}\,
   \sigma\!\left(\mathbf{h}^{1}\right) \Big\rangle
=
\big[\mathbf{G}^{\,2,i}\big]^{-1}
\Big\langle
\big(\boldsymbol{\xi}^{\,2,i} - \mathbf{B}^{2}_{i}
   \sigma\!\left(\mathbf{h}^{1}\right)\big)\,
   \sigma\!\left(\mathbf{h}^{1}\right)
\Big\rangle
\end{equation}

Now the Hubbard Trick \cite{Hubbard} states that 

\begin{equation}
\label{hubbard_trick}
\adjustbox{max width=\textwidth}{$
    \exp\!\left(-\frac{1}{2}\,\mathbf{x}^{\top} A\,\mathbf{x}\right)
= \int_{\mathbb{R}^{d}}
\frac{d\mathbf{u}}{(2\pi)^{d/2} \sqrt{\det A}}\,
\exp\!\left(-\frac{1}{2}\,\mathbf{u}^{\top}A^{-1}\mathbf{u}
- i\,\mathbf{u}\!\cdot\!\mathbf{x}\right)
= \big\langle \exp(-i\,\mathbf{u}\!\cdot\!\mathbf{x}) \big\rangle_{\mathbf{u} \sim \mathcal{N}(0,A)}.
$}
\end{equation}

This allows us to rewrite the quadratic terms in our single-site MGF as follows:

\begin{equation}
\begin{aligned}
\exp\!&\left(
-\frac12 
\sum_{\mu\nu}
\int_{0}^{\infty}\! dt
\int_{0}^{\infty}\! ds\;
\hat{\chi}^{\,1}_{\mu}(t)\cdot
\hat{\chi}^{\,1}_{\nu}(s)\,
\,
\Phi^{\,0}_{\mu\nu}(t,s)
\right)
= \\
&
=
\left\langle
\exp\!\left(
-i
\sum_{\mu}
\int_{0}^{\infty}\! dt\;
\alpha^{\,1}_{\mu}(t)\,
\hat{\chi}^{\,1}_{\mu}(t)
\right)
\right\rangle_{\alpha^1
\sim \mathcal{GP}\,\left(0,\Phi^{\,0}\right)}
\end{aligned}
\end{equation}

\begin{equation}
\begin{aligned}
\exp&\!\left(
-\frac12 
\sum_{\mu\nu i}
\int_{0}^{\infty}\! dt
\int_{0}^{\infty}\! ds\;
\hat{\chi}_{\mu,i}^{2}(t)\cdot
\hat{\chi}_{\nu,i}^{2}(s)\,
\Phi^{1}_{\mu\nu}(t,s)
\right)=\\
&
=
\left\langle
\exp\!\left(
-i
\sum_{\mu i}
\int_{0}^{\infty}\! dt\;
\alpha_{\mu,i}^{2}(t)\,
\hat{\chi}_{\mu,i}^{2}(t)
\right)
\right\rangle_{\{\alpha^{2}_{i}\}
\sim \mathcal{GP}\,\left(0,\Phi^{\,1}\right)}
\end{aligned}
\end{equation}

\begin{equation}
\begin{aligned}
\exp&\!\Bigg(
-\frac12 
\sum_{\mu\nu i}
\int_{0}^{\infty}\! dt
\int_{0}^{\infty}\! ds\;
\hat{\chi}^{\ell}_{\mu,i}(t)\cdot
\hat{\chi}^{\ell}_{\nu,i}(s)\,
\Phi^{\ell-1}_{\mu\nu,i}(t,s)
\Bigg) =\\
&
=
\left\langle
\exp\!\left(
-i
\sum_{\mu i}
\int_{0}^{\infty}\! dt\;
\alpha^{\ell}_{\mu,i}(t)\,
\hat{\chi}^{\ell}_{\mu,i}(t)
\right)
\right\rangle_{\{\alpha^{\ell}_{i}\}
\sim \mathcal{GP}\,\left(0,\Phi^{\ell-1}_i\right)}\ell\in\{3,4,...,L\}
\end{aligned}
\end{equation}

\begin{equation}
\begin{aligned}
\exp&\!\Bigg(
-\frac12 
\sum_{\mu\nu i}
\int_{0}^{\infty}\! dt
\int_{0}^{\infty}\! ds\;
\hat{\xi}^{\ell}_{\mu,i}(t)\cdot
\hat{\xi}^{\ell}_{\nu,i}(s)\,
G_{\mu\nu,i}^{\ell}(t,s)
\Bigg)=\\
&
=
\left\langle
\exp\!\left(
-i
\sum_{\mu i}
\int_{0}^{\infty}\! dt\;
\beta^{\ell}_{\mu,i}(t)\,
\hat{\xi}^{\ell}_{\mu,i}(t)
\right)
\right\rangle_{\{\beta^{\ell}_{i}\}
\sim \mathcal{GP}\,\left(0,G^{\ell}_{i}\right)}\ell\in\{2,3,4,...,L\}
\end{aligned}
\end{equation}

\begin{equation}
\begin{aligned}
\exp&\!\Bigg(
-\frac12 
\sum_{\mu\nu i}
\int_{0}^{\infty}\! dt
\int_{0}^{\infty}\! ds\;
\hat{\xi}^{L+1}_{\mu}(t)\cdot
\hat{\xi}^{L+1}_{\nu}(s)
\Bigg)=\\
&
=
\left\langle
\exp\!\left(
-i
\sum_{\mu i}
\int_{0}^{\infty}\! dt\;
\beta^{L+1}_{\mu}(t)\,
\hat{\xi}^{L+1}_{\mu}(t)
\right)
\right\rangle_{\{\beta^{L+1}\}
\sim \mathcal{GP}\,\left(0,1\right)}
\end{aligned}
\end{equation}

\begin{equation}
\exp\!\Bigg(
-\frac12 
\sum_{\mu\nu i}
\int_{0}^{\infty}\! dt
\int_{0}^{\infty}\! ds\;
\hat{\xi}^{\phi}_{\mu}(t)\cdot
\hat{\xi}^{\phi}_{\nu}(s)
\Bigg)
=
\left\langle
\exp\!\left(
-i
\sum_{\mu i}
\int_{0}^{\infty}\! dt\;
\beta^{\phi}_{\mu}(t)\,
\hat{\xi}^{\phi}_{\mu}(t)
\right)
\right\rangle_{\{\beta^{\phi}\}
\sim \mathcal{GP}\,\left(0,G^\phi\right)}
\end{equation}

We can now easily integrate over all the $\{\hat\chi,\hat\xi\}$ variables, since the argument of the exponential in $\mathcal Z$ has been linearised with respect to them all. Doing so yields the following delta functions:

\begin{equation}
    \int \prod_{\mu,t} \frac{d\hat{\chi}_{\mu}^1(t)}{2\pi}\,
\exp\!\left(
 i\,\hat{\boldsymbol{\chi}}^1 \!\cdot\!
 \big(\boldsymbol{\chi}^1 - \boldsymbol{\alpha}^1\big)
\right)
\;=\;
\delta\!\left(\boldsymbol{\chi}^1 - \boldsymbol{\alpha}^1\right)
\end{equation}

\begin{equation}
    \int \prod_{\mu,t} \frac{d\hat{\chi}_{\mu,i}^{\,\ell}(t)}{2\pi}\;
\exp\!\left[
  i\,\hat{\boldsymbol{\chi}}^{\,\ell}_{i}
  \!\cdot\!
  \left(
     \boldsymbol{\chi}^{\,\ell}_{i}
     - \boldsymbol{\alpha}^{\,\ell}_{i}
     - \mathbf{A}^{\,\ell}_{i}\,\mathbf{g}^{\,\ell}_{i}
  \right)
\right]
=
\delta\!\left(
   \boldsymbol{\chi}^{\,\ell}_{i}
   - \boldsymbol{\alpha}^{\,\ell}_{i}
   - \mathbf{A}^{\,\ell}_{i}\,\mathbf{g}^{\,\ell}_{i}
\right)\quad\ell\in\{2,3,...,L\}
\end{equation}

\begin{equation}
\label{delta_xii2i}
\int \prod_{\mu,t}
\frac{d\hat{\xi}_{\mu}^{\,2,i}(t)}{2\pi}\;
\exp\!\left(
 i\,\hat{\boldsymbol{\xi}}^{\,\ell}_{i}
 \cdot
 \Big(
   \boldsymbol{\xi}^{2}_{i}
   - \boldsymbol{\beta}^{2}_{i}
   - \mathbf{B}^{2\top}_{i}\,
     \sigma\!\left(\mathbf{h}^{1}\right)
 \Big)
\right)
=
\delta\!\left(
 \boldsymbol{\xi}^{2}_{i}
 - \boldsymbol{\beta}^{2}_{i}
 - \mathbf{B}^{2\top}_{i}\,
   \sigma\!\left(\mathbf{h}^{1}\right)
\right)
\end{equation}

\begin{equation}
\begin{aligned}
\label{delta_xielli}
\int \prod_{\mu,t}
\frac{d\hat{\xi}_{\mu,i}^{\,\ell}(t)}{2\pi}\;
\exp\!\left(
 i\,\hat{\boldsymbol{\xi}}^{\,\ell}_{i}
 \cdot
 \Big(
   \boldsymbol{\xi}^{\,\ell}_{i}
   - \boldsymbol{\beta}^{\,\ell}_{i}
   - \mathbf{B}^{\,\ell\top}_{i}\,
     \sigma\!\left(\mathbf{h}^{\,\ell-1}_{i}\right)
 \Big)
\right)
=
\delta\!\left(
 \boldsymbol{\xi}^{\,\ell}_{i}
 - \boldsymbol{\beta}^{\,\ell}_{i}
 - \mathbf{B}^{\,\ell\top}_{i}\,
   \sigma\!\left(\mathbf{h}^{\,\ell-1}_{i}\right)
\right)\\
\quad\ell\in\{3,4,...,L\}
\end{aligned}
\end{equation}

\begin{equation}
\label{delta_xiphi}
\int \prod_{\mu,t}
\frac{d\hat{\xi}_{\mu}^{\phi}(t)}{2\pi}\;
\exp\!\left(
 i\,\hat{\boldsymbol{\xi}}^{\phi}
 \cdot
 \Big(
   \boldsymbol{\xi}^{\phi}
   - \boldsymbol{\beta}^{\phi}
   - \mathbf{B}^{\phi\top}\,
     \sigma(\mathbf{h}^{1})
 \Big)
\right)
=
\delta\!\left(
 \boldsymbol{\xi}^{\phi}
 - \boldsymbol{\beta}^{\phi}
 - \mathbf{B}^{\phi\top}\,
   \sigma(\mathbf{h}^{1})
\right)
\end{equation}

\begin{equation}
    \int \prod_{\mu,t}
\frac{d\hat{\xi}_{\mu}^{L+1}(t)}{2\pi}\;
\exp\!\left(
 i\,\hat{\boldsymbol{\xi}}^{L+1}
 \cdot
 \Big(
   \boldsymbol{\xi}^{L+1}
   - \boldsymbol{\beta}^{L+1}
 \Big)
\right)
=
\delta\!\left(
 \boldsymbol{\xi}^{L+1}
 - \boldsymbol{\beta}^{L+1}
\right)
\end{equation}

Using these in the expressions \ref{chi2_correlator} to \ref{final_correlator}, we note that all four classes of correlators vanish, since the average term collapses to the covariance matrix of the respective Gaussian process $\{\bm\alpha,\bm\beta\}$, which cancels with its inverse, giving zero overall. We summarise this as follows (where the expression is implied to hold for all allowed arguments, superscripts and subscripts):

\begin{equation}
    \hat G=\hat\Phi=\langle\hat\chi\hat\chi\rangle=\langle\hat\xi\hat\xi\rangle=0
\end{equation}

\subsection{The coupling kernels A and B}
What of $A$ and its dual $B$? Well inserting \ref{delta_xii2i} into \ref{xi_sigma_correlator}, and inserting this in turn into the relevant saddle point equation, we obtain for $\ell\in\{3,4,...,L\}$

\begin{equation}
\label{Aellli_avg}
    \begin{aligned}
A^{\,\ell}_{\mu\nu,i}(t,s)
&= -\,i\,
\big\langle
\hat{\xi}^{\,\ell}_{\nu,i}(t)\,
\sigma\!\big(h^{\,\ell-1}_{\mu,i}(s)\big)
\big\rangle
\\
&=
\big[G^{\,\ell}_{i}\big]^{-1}
\Big\langle
\big(
\xi^{\,\ell}_{i}
-
B^{\,\ell}_{i}\,
\sigma\!\big(h^{\,\ell-1}_{i}\big)
\big)\,
\sigma\!\big(h^{\,\ell-1}_{i}\big)
\Big\rangle
\\
&=
\left\langle\frac{\partial\sigma(h_{\mu,i}^{\ell-1}(t))}{\partial\beta_{\nu,i}^{\ell\top}(s)}\right\rangle
\end{aligned}
\end{equation}

In the final line, we use Stein's lemma \cite{Stein} which states that for a normally distributed random variable $X$ with expectation $\mu$ and variance $\sigma^2$, and differentiable function $g$, $\mathbb E[g(X)(X-\mu)]=\sigma^2\mathbb E[g'(X)]$. 

Similarly:

\begin{equation}
\label{A2i_avg}
A^{2}_{\mu\nu,i}(t,s)=\left\langle\frac{\partial\sigma(h_\mu^{1}(t))}{\partial\beta_{\nu,i}^{2\top}(s)}\right\rangle
\end{equation}

\begin{equation}
\label{Aphi_avg}
A^{\phi}_{\mu\nu}(t,s)=\left\langle\frac{\partial \sigma(h_\mu^{1}(t))}{\partial\beta_\nu^{\phi\top}(s)}\right\rangle
\end{equation}

\begin{equation}
\label{B_avg}
B_{\mu\nu,i}^{\ell}(t,s)=\left\langle\frac{\partial g_{\mu,i}^{\ell}(t)}{\partial\alpha_{\nu,i}^{\ell\top}(s)}\right\rangle
\end{equation}

Note that the saddle equations tell us that $B^\phi_{\mu\nu}(t,s) = 0$.

\subsection{Final DMFT}

To remedy the asymmetry manifest in equations \ref{delta_xii2i} and \ref{delta_xielli}, we redefine $B$ as its transpose. We also rescale $A$ to $\gamma_0\eta_0A$, and B to $\gamma_0\eta_0B$. This makes it clear that the non-Gaussian corrections to $h$ and $z$ are $O(\gamma_0\eta_0)$. We also replace instances of $g$ with their corresponding $z$ using \ref{deep_gradients}. 

We then have the following complete self-consistent DMFT equations:

\begin{equation}
\label{deep_dmft}
\begin{aligned}
\alpha^1(t)&\sim\mathcal{GP}(0,\Phi^0),\qquad
\alpha^{\ell}_{i}(t)\sim\begin{cases}
    \mathcal{GP}(0,\Phi^1(t,s))&\ell=2\\
    \mathcal{GP}(0,\Phi^{\ell-1}_i(t,s))&\ell\in\{3,4,...,L\}
\end{cases}\\
\beta^{\ell}_{i}(t)&\sim\mathcal{GP}(0,G^{\ell+1}_{i}(t,s))\quad\ell\in\{1,2,...,L-1\},\qquad
\beta^{L}(t)\sim\mathcal{N}(0,1),\qquad \\
\beta^\phi(t)&\sim\mathcal{GP}(0,G^\phi(t,s))
\qquad\Phi_{\mu\nu}^{0} = \langle x_\mu x_\nu \rangle,\qquad
\Phi_{\mu\nu}^{1}(t,s) = \langle \sigma(h_\mu^{1}(t))\sigma(h_\nu^{1}(s)) \rangle, \\
\Phi_{\mu\nu,i}^{\ell}(t,s) &= \langle \sigma(h_{\mu,i}^{\ell}(t))\, \sigma(h_{\nu,i}^{\ell}(s)) \rangle\quad\ell\in\{2,3,...,L-1\}, \\
\Phi_{\mu\nu}^{L}(t,s) &= \left\langle \sum_{i,j=1}^M \phi^i_{\mu}(t)\,\phi^j_\nu(s)\, h_{\mu,i}^{L}(t)\, h_{\nu,i}^{L}(s)\, \right\rangle, \\
G_{\mu\nu}^{1}(t,s) &= \langle [\dot\sigma(h_\mu^{1}(t))\odot z_\mu^{1}(t)] [\dot\sigma(h_\nu^{1}(s))\odot z_\nu^{1}(s)] \rangle\\
G_{\mu\nu,i}^{\ell}(t,s) &= \langle [\dot\sigma(h_{\mu,i}^{\ell}(t))\odot z_{\mu,i}^{\ell}(t)] [\dot\sigma(h_{\nu,i}^{\ell}(s))\odot z_{\nu,i}^{\ell}(s)] \rangle\quad\ell\in\{2,3,...,L-1\},\\
G_{\mu\nu,i}^{L}(t,s) &= \langle z_{\mu,i}^{L}(t)z_{\nu,i}^{L}(s)\rangle\\
G_{\mu\nu}^{\phi}(t,s) &= \sum_{i=1}^M\langle h_{\mu,i}^{L}(t) z_\mu^{L}(t)\rangle_i\langle h_{\nu,i}^{L}(s)z_\nu^{L}(s) \rangle_i\\
A^{2}_{\mu\nu,i}(t,s)&=\left\langle\frac{\partial\sigma(h_\mu^{1}(t))}{\partial\beta_{\nu,i}^{2\top}(s)}\right\rangle,\qquad
A^{\,\ell}_{\mu\nu,i}(t,s)=\left\langle\frac{\partial\sigma(h_{\mu,i}^{\ell-1}(t))}{\partial\beta_{\nu,i}^{\ell\top}(s)}\right\rangle\quad\ell\in\{3,4,...,L\},\qquad \\
A^{\phi}_{\mu\nu}(t,s)&=\left\langle\frac{\partial \sigma(h_\mu^{1}(t))}{\partial\beta_\nu^{\phi\top}(s)}\right\rangle\\
B_{\mu\nu,i}^{\ell}(t,s)&=\left\langle\frac{\partial(\dot\sigma(h_{\mu,i}^{\ell}(t))z_{\mu,i}^{\ell}(t))}{\partial\alpha_{\nu,i}^{\ell\top}(s)}\right\rangle\quad\ell\in\{2,3,...,L-1\},\qquad B^{L}_{\mu\nu,i}(t,s)=\left\langle\frac{\partial z_{\mu,i}^{L}(t)}{\partial\alpha_{\nu,i}^{L\top}(s)}\right\rangle\
\end{aligned}
\end{equation}
\begin{equation}
\begin{aligned}
\tilde z_{\mu,i}^{1}(t)&=\beta_{\mu,i}^{1}(t)+ \frac{\eta_0 \gamma_0}{P}\int_0^t \! ds\sum_{\nu}[B^{2}_{\mu\nu,i}(s,t)+\Delta_{\nu}(s)G_{\mu \nu,i}^{2}(s,t)] \, \sigma(h^1_{\nu}(s))\\
\tilde z_\mu^{\phi}(t)&=\beta_{\mu}^{\phi}(t)+ \frac{\eta_0 \gamma_0}{P}\int_0^t \! ds\sum_{\nu}[\Delta_{\nu}(s)G_{\mu \nu}^{\phi}(s,t)] \, \sigma(h^1_{\nu}(s)),\\
z^1_\mu(t) &= \sum_{i=1}^M \tilde z^{1}_{\mu,i}(t) + \tilde z^\phi_\mu\\
z_{\mu,i}^{\ell}(t)&=\beta_{\mu,i}^{\ell}(t)+ \frac{\eta_0 \gamma_0}{P}\int_0^t \! ds\sum_{\nu}[B^{\ell+1}_{\mu\nu,i}(s,t)+\Delta_{\nu}(s)G_{\mu \nu,i}^{\ell+1}(s,t)] \, \sigma(h^{\ell}_{\nu,i}(s))\\
& \qquad\qquad\qquad\qquad\qquad\qquad\qquad\qquad\qquad\quad\ell\in\{2,3,...,L-1\}\\
z_\mu^{L}(t)&=\beta_{\mu}^{L}(t)+ \frac{\eta_0 \gamma_0}{P}\int_0^t \! ds\sum_{\nu}\Delta_{\nu}(s) \, h^{L}_{\nu}(s), \qquad\qquad\qquad\,\,\, z_{\mu,i}^{L}(t)=\phi^i(t)z_\mu^L(t)\\
\end{aligned}
\end{equation}

\begin{equation}
\begin{aligned}
h_{\mu}^1(t)&=\alpha_{\mu}^1(t)+\frac{\eta_0 \gamma_0}{P}
    \int_0^t \! ds
    \sum_{\nu} \Delta_{\nu} \,\Phi_{\mu\nu}^0\, (\dot\sigma(h^1_\mu) \odot z^1_\mu(s)) , \\
{h^{2}_{\mu, i}}(t)&= \alpha_{\mu,i}^{2}(t)
  + \frac{\eta_0 \gamma_0}{P}
    \int_0^t \! ds
    \sum_{\nu} [A_{\mu\nu,i}^{2}(s,t)+\Delta_{\nu}(s)\Phi_{\mu \nu}^{1}(s,t)] \, (\dot\sigma(h_{\nu,i}^{2}(s))\odot z_{\nu,i}^{2}(s)) , \\
h_{\mu,i}^{\ell}(t)&= \alpha_{\mu,i}^{\ell}(t)
  + \frac{\eta_0 \gamma_0}{P}
    \int_0^t \! ds
    \sum_{\nu} [A_{\mu\nu,i}^{\ell}(s,t)+\Delta_{\nu}(s)\Phi_{\mu \nu,i}^{\ell-1}(s,t)] \, (\dot\sigma(h_{\nu,i}^{\ell}(s))\odot z_{\nu,i}^{\ell}(s))\\
    &    \qquad\qquad\qquad\qquad\qquad\qquad\qquad\qquad\qquad\qquad\qquad\qquad\ell\in\{3,4,...,L-1\} , \\
h_{\mu,i}^{L}(t)&= \alpha_{\mu,i}^{L}(t)
  + \frac{\eta_0 \gamma_0}{P}
    \int_0^t \! ds
    \sum_{\nu} [A_{\mu\nu,i}^{L}(s,t)+\Delta_{\nu}(s)\Phi_{\mu \nu,i}^{L-1}(s,t)] \, z_{\nu,i}^{L}(s)
    \, , \\
h^L_\mu&=\sum_{i=1}^M\phi_\mu^i\cdot h_{\mu,i}^{L},\qquad\qquad\phi_\mu(t)=\text{softmax}(\psi_\mu(t))\\
\frac{\diff f_\mu(t)}{\diff t}&=\frac{\eta_0}{P}\sum_{\nu=1}^P\Big[G_{\mu\nu}^1(t,t)\Phi^0_{\mu\nu}+\sum_{i=1}^M[G_{\mu\nu,i}^{2}(t,t)\Phi^1_{\mu\nu}(t,t)+\sum_{\ell=3}^L G_{\mu\nu,i}^{\ell}(t,t)\Phi^{\ell-1}_{\mu\nu,i}(t,t)] \\
&
\quad+\Phi^{L}_{\mu\nu}(t,t)+ G^\phi_{\mu\nu}(t,t)\Phi^1_{\mu\nu}(t,t)\Big]\Delta_\nu(t)\\
\psi^i_\mu(t)&=\frac{\gamma_0\eta_0}P\int_0^tds\sum_{\nu=1}^P[A^\phi_{\nu\mu}(s,t)+\Delta_\nu(s) \Phi^1_{\nu\mu}(s,t)]\left[g^\phi_\nu(s)\right]_i,
\quad \\
\big[ g^{\phi}_{\mu}(t) \big]_i
&=
\frac{\eta_0 \gamma_0}{P}
\int_{0}^{t} ds
\sum_{\nu}
\Delta_{\nu}(s)\,
G_{\mu\nu}^{L}(t,s)\,
{\Phi}_{\mu\nu}^{L-1}(t,s)\,\phi^i_\nu(s)
    \end{aligned}
\end{equation}

\newpage

\section{DMFT Analysis for Regime II (MSSP)}\label{sec:dmft_regime2}

\subsection{Architectural definitions}
Here we distinguish the external width $N$ and the internal width $N_e$, maintaining a standard continuous MoE architecture, but now with $Q\in\mathbb R^{M \times N}$ and $W^{2,\i}_i\in\mathbb R^{N_e\times N}$. All other architectural objects live in the standard spaces and the total number of experts is $M$.

We are concerned with the thermodynamic limit where $M, N \to \infty$ at a fixed ratio. Specifically, we define $\kappa = \frac{M}{N}$, where $\kappa$ is a constant of order unity in $N$. The internal expert width $N_e$ will also remain strictly order unity in this setting. The forward pass, following the \textit{MSSP}, is defined as:

\begin{equation}
\label{infinite_experts_architecture_r2}
    \begin{aligned}
\mathbb R^N \ni h_\mu^1 
&= \frac{1}{\sqrt D} W^1 x_\mu 
&\qquad
\mathbb R^M \ni \psi_\mu 
&= \frac{1}{\sqrt N} Q \sigma(h^1_\mu) \\
\mathbb R^{N_e} \ni h^{2,\text{in}}_{\mu,i} 
&= \frac{1}{\sqrt N} W^{2,\text{i}}_i \sigma(h_\mu^1) 
&\qquad
\mathbb R^M \ni \phi_\mu 
&= \text{softmax}(\psi_\mu) \\
\mathbb R^N \ni h^{2,\text{o}}_{\mu,i} 
&= \sqrt{\frac{\kappa N}{N_e}} W^{2,\text{o}}_i \sigma(h^{2,\text{in}}_{\mu,i}) 
&\qquad
\mathbb R^N \ni h^3_\mu 
&= \sum_{i=1}^M \phi_\mu^i \, h^{2,\text{o}}_{\mu,i} \\
\mathbb R \ni h^4_\mu 
&= \frac{1}{\sqrt N} w^{4\mathsf{T}} h^3_\mu 
&\qquad
f_\mu 
&= \frac{1}{\gamma} h^4_\mu
\end{aligned}
\end{equation}

\subsection{Learning rates and initialization}
To ensure that the network evolves dynamically in the infinite-width limit without gradients vanishing or exploding, we apply specific learning rate scalings:
\begin{equation}
    \begin{aligned}
        \eta&=\eta_0\gamma^2 \qquad
        \eta_Q = \eta_0\gamma^2 \kappa \qquad
        \gamma =\gamma_0\sqrt N \qquad
        \eta_0,\gamma_0\sim O(1)\\
    \end{aligned}
\end{equation}

The network weights are initialized from standard Gaussian distributions:
\begin{equation}
    w_\alpha^{4}(0),\left[W_{i}^{2,\o}(0)\right]_{\alpha\beta},\left[W_{i}^{2,\i}(0)\right]_{\alpha\beta},W_{\alpha\beta}^{1}(0),Q_{\alpha\beta}(0)\sim\mathcal N(0,1)
\end{equation}

For brevity in our derivations, we denote the collection of all network parameters as $\bm\theta = \text{Vec}\{W^1,W_i^{2,\i},W_i^{2,\o},w^4,Q\}$. 

Due to the normalization constraint of the softmax operator over $M$ experts, the routing probabilities scale as $\phi^i\sim O(\frac1{\kappa N})$. It is mathematically tidier to formulate the DMFT using variables which remain $O(1)$ in the limit. We therefore define and track the rescaled routing variables:
\begin{equation}
    \tilde\phi^i_\mu := \kappa N \phi^i_\mu
\end{equation}

\subsection{Gradient definitions}
We mathematically \textit{define} the pre-activation gradients, ensuring they contain the correct scaling factors to remain finite in the $N \to \infty$ limit:

\begin{equation}
\begin{aligned}
g^1_\mu 
&:= \sqrt N \frac{\partial h^4_\mu}{\partial h^1_\mu}
&\qquad
g^{2,\text{in}}_{\mu,i} 
&:= \kappa\sqrt N \frac{\partial h^4_\mu}{\partial h^{2,\text{in}}_{\mu,i}} \\
g^{2,\text{out}}_{\mu,i} 
&:= \kappa N^{\frac32}\frac{\partial h^4_\mu}{\partial h^{2,\text{o}}_{\mu,i}}
&\qquad
g^3_\mu 
&:= \sqrt N \frac{\partial h^4_\mu}{\partial h^3_\mu} = w^4 = z^3 \\
g^\phi_\mu 
&:= \frac{1}{\sqrt N}\frac{\partial h^4_\mu}{\partial \phi_\mu}
&\qquad
z^{2,\text{out}}_{\mu,i} 
&:= \kappa N \phi^i_\mu g^3_\mu = \tilde\phi^i_\mu g^3_\mu \\
z^{2,\text{in}}_{\mu,i} 
&:= \sqrt{\frac{\kappa}{N_e N}} \, W^{2,\text{out}\,\top}_i g^{2,\text{out}}_{\mu,i}
&\qquad
\tilde z^\phi_\mu 
&:= \frac{1}{\kappa \sqrt N} Q^\top g^{1,\phi} \\
\tilde z^1_{\mu,i} 
&:= \frac{1}{\kappa \sqrt N} W^{2,\text{in}\,\top}_i g^{2,\text{in}}_{\mu,i}
&\qquad
z^1 
&:= \frac{1}{\kappa \sqrt N} Q^\top g^{1,\phi}
   + \frac{1}{\kappa \sqrt N} \sum_{i=1}^M W^{2,\text{in}\,\top}_i g^{2,\text{in}}_{\mu,i} \\
&&&
= \frac{1}{\kappa \sqrt N} \sum_{i=1}^M \tilde z^1_{\mu,i} + \tilde z^{1,\phi}_\mu
\end{aligned}
\end{equation}

Note that the components of the router gradient $g^\phi_\mu$ naturally scale as $O(1)$:
\begin{equation}
    (g^\phi_\mu)_i =\frac1{\sqrt N}\frac{\partial h_\mu^{4}}{\partial\phi_\mu^i} = \frac1{\sqrt N}\frac{\partial h_\mu^{4}}{\partial \bm h^3_\mu}\cdot\frac{\partial \bm h_\mu^{3}}{\partial\phi_\mu^i} = \frac{1}{ N}\bm g^3\cdot\bm h_{\mu,i}^{2,\o} = O(1).
\end{equation}

The backward pass through the router dictates that the gradient at layer $1$ involves the quantity $g^{1,\phi}$, defined component-wise to handle the Jacobian of the softmax:
\begin{equation}
    g^{1,\phi}_k := \kappa N \sum_{i=1}^{\kappa N}g_i^\phi\phi^i(\delta_{ik}-\phi^k)
    = \sum_{i=1}^{\kappa N}g_i^\phi\tilde\phi^i(\delta_{ik}-\frac{\tilde\phi^k}{\kappa N})
\end{equation}

\subsection{DMFT kernels and order parameters}
We define the following macroscopic \textit{kernels}, all of which are $O_N(1)$ and will serve as the fundamental order parameters in the dynamics:

\begin{equation}
\label{inf_exp_kernels_r2}
\begin{aligned}
\begin{aligned}
\Phi^{0}_{\mu\nu}
&= \frac{1}{D}\, x_\mu \cdot x_\nu
&\qquad
\Phi^{1}_{\mu\nu}(s,t)
&= \frac{1}{N}\, \sigma\!\big(h^{1}_{\mu}(s)\big)\cdot \sigma\!\big(h^{1}_{\nu}(t)\big) \\
\Phi^{2,\text{in}}_{\mu\nu,i}(s,t)
&= \frac{1}{N_e}\, \sigma\!\big(h^{2,\text{in}}_{\mu,i}(s)\big)\cdot \sigma\!\big(h^{2,\text{in}}_{\nu,i}(t)\big)
&\qquad
\Phi^{2,\text{o}}_{\mu\nu,i}(s,t)
&= \frac{1}{N}\, \sigma\!\big(h^{2,\text{out}}_{\mu,i}(s)\big)\cdot \sigma\!\big(h^{2,\text{out}}_{\nu,i}(t)\big) \\
\Phi^{3}_{\mu\nu}(s,t)
&= \frac{1}{N}\, h^{3}_{\mu}(s)\cdot h^{3}_{\nu}(t)
&\qquad
G^{1}_{\mu\nu}(s,t)
&= \frac{1}{N}\, g^{1}_{\mu}(s)\cdot g^{1}_{\nu}(t) \\
G^{2,\text{in}}_{\mu\nu,i}(s,t)
&= \frac{1}{N_e}\, g^{2,\text{in}}_{\mu,i}(s)\cdot g^{2,\text{in}}_{\nu,i}(t)
&\qquad
G^{2,\text{out}}_{\mu\nu,i}(s,t)
&= \frac{1}{N}\, g^{2,\text{out}}_{\mu,i}(s)\cdot g^{2,\text{out}}_{\nu,i}(t) \\
G^{1,\phi}_{\mu\nu}(s,t)
&= \frac{1}{\kappa N}\, g^{1,\phi}_{\mu}(s)\cdot g^{1,\phi}_{\nu}(t)
&\qquad
G^{3}_{\mu\nu}(s,t)
&= \frac{1}{N}\, g^{3}_{\mu}(s)\cdot g^{3}_{\nu}(t)
\end{aligned}
\end{aligned}
\end{equation}

With the exception of the input data Gram matrix $\Phi^0$, all of these kernels are dynamically evolving macroscopic variables.

\subsection{Learning Dynamics and the Neural Tangent Kernel}

We train the network using continuous-time gradient flow. The variable learning rate scheme defined previously is necessary so that the activation updates within the experts and for the router scores do not vanish or explode as $N, M \to \infty$. 

Taking a standard empirical risk minimization loss of the form:
\begin{equation}
    \label{deep_loss_r2}
    \mathcal L=\frac 1 P \sum_{\mu=1}^P\ell(f_\mu,y_\mu)
\end{equation}

The network parameters evolve according to:
\begin{equation}
    \label{deep_learning_dynamics_r2}
    \begin{aligned}
    &\frac{d\theta}{dt}
= \frac{\eta}{P\gamma}
  \sum_{\mu} \Delta_{\mu}
  \frac{\partial h_{\mu}^{4}}{\partial \theta}\\
  &\Delta_\mu=-\frac{\partial\mathcal L}{\partial f_\mu}
    \end{aligned}
\end{equation}

For a Mean Squared Error (MSE) loss where $\mathcal L=\frac{1}{P}\sum_{\nu}(y_\nu-f_\nu)^2$, the error signal (or residual) simplifies to $\Delta_\nu=2(y_\nu-f_\nu)$.

The logits update dynamically via the chain rule, driven by the Neural Tangent Kernel (NTK):
\begin{equation}
\label{logit_updates_r2}
\frac{d f_{\mu}(t)}{dt}
= \frac{\partial f_{\mu}(t)}{\partial \theta}\cdot \frac{d\theta}{dt}
= \frac{\partial f_{\mu}(t)}{\partial \theta}\cdot \frac{\eta_\theta}{P}\sum_\alpha^P\Delta_\alpha\frac{\partial f_\alpha(t)}{\partial \theta}
= \frac{\eta}{P}
  \sum_{\alpha} \Delta_{\alpha} \,
  K^{\mathrm{NTK}}_{\mu \alpha}(t,t)
\end{equation}

Where the infinite-width NTK is defined as:
\begin{equation}
\label{ntk_kernel_r2}
K^{\mathrm{NTK}}_{\mu \alpha}(t,s)
\equiv
\frac{\partial f_{\mu}(t)}{\partial \theta}
\cdot
\frac{\partial f_{\alpha}(s)}{\partial \theta}
\end{equation}

\subsection{Decomposition of the MoE NTK}
To explicitly compute the NTK, we require that the magnitude $\eta_\theta \frac{\partial f_{\mu}(t)}{\partial \theta_i} \cdot \frac{\partial f_{\alpha}(s)}{\partial \theta_i}$ remains order 1 for each parameter block $\theta_i$. We evaluate this constraint block by block.

For the expert input layer $W^{2,\i}$:
\begin{equation}
    \begin{aligned}
        &\eta_0 \sum_{i=1}^{\kappa N}\sum_{l,m,j=1}^{N_e}\sum_{k=1}^N
        \frac{\partial h^4}{\partial \left[{h^{2,\text{in}}_i}\right]_l}
        \frac{\partial \left[{h^{2,\text{in}}_i}\right]_l}{\partial W^{2}_{jk,i}}
        \frac{\partial h^4}{\partial \left[{h^{2,\text{in}}_i}\right]_m}
        \frac{\partial \left[{h^{2,\text{in}}_i}\right]_m}{\partial W^{2}_{jk,i}}= \\
        &=
        \eta_0 N \sum_{i=1}^{\kappa N}\sum_{l,m,j=1}^{N_e}\sum_{k=1}^N
        \frac{\left[g^{2,\i}_i\right]_l}{\kappa \sqrt N}\frac{\left[g^{2,\i}_i\right]_m}{\kappa\sqrt N}\frac{\sigma(h^1_k)}{\sqrt N}\frac{\sigma(h^1_k)}{\sqrt N}\delta_{mj}\delta_{lj}\\
        &=\eta_0 \frac{N_e}\kappa\frac{1}{\kappa N}\sum_{i=1}^{\kappa N}\left[\frac1{N_e}\sum_{j=1}^{N_e} \left[g^{2,\i}_i\right]_j\left[g^{2,\i}_i\right]_j\right]\left[\frac1N\sum_{k=1}^N\sigma(h^1_k)^2\right]\\
        &=\eta_0 \frac{N_e}\kappa\frac{1}{\kappa N}\sum_{i=1}^{\kappa N}G^{2,\i}_i\Phi^1 = \eta_0\frac{N_e}\kappa\bar G^2\Phi^1
    \end{aligned}
\end{equation}

For the expert output layer $W^{2,\o}$:
\begin{equation}
    \begin{aligned}
        &\eta_0\sum_{i=1}^{\kappa N}\sum_{l,m,k=1}^{N_e}\sum_{j=1}^N
        \frac{\partial h^4}{\partial \left[h^{2,\o}_i\right]_l}
        \frac{\partial \left[h^{2,\o}_i\right]_l}{\partial W^{2,\o}_{jk}}
        \frac{\partial h^4}{\partial \left[h^{2,\o}_i\right]_m}
        \frac{\partial \left[h^{2,\o}_i\right]_m}{\partial W^{2,\o}_{jk}}\\
        &=
        \eta_0  \sum_{i=1}^{\kappa N}\sum_{l,m,k=1}^{N_e}\sum_{j=1}^N
        \frac{\left[g^{2,\o}_i\right]_l}{\kappa N^{3/2}}\frac{\left[g^{2,\o}_i\right]_m}{\kappa N^{3/2}}\frac{\sqrt{\kappa N}}{\sqrt{N_e}}\sigma(\left[{h^{2,\text{in}}_i}\right]_k)\frac{\sqrt{\kappa N}}{\sqrt{N_e}}\sigma(\left[{h^{2,\text{in}}_i}\right]_k)\delta_{mj}\delta_{lj}\\
        &=\eta_0 \frac{1}{\kappa N}\sum_{i=1}^{\kappa N}G^{2,\o}_i\Phi^{2,\i}_i        
    \end{aligned}
\end{equation}

For the shared base layer $W^1$:
\begin{equation}
\begin{aligned}
\eta_0
\sum_{k=1}^{D}
\sum_{j,m=1}^{N}
\frac{\partial h^{4}}{\partial h^{1}_{n}}
\frac{\partial h^{1}_{n}}{\partial W^{1}_{jk}}
\frac{\partial h^{4}}{\partial h^{1}_{m}}
\frac{\partial h^{1}_m}{\partial W^{1}_{jk}}
&=
\eta_0
\sum_{k=1}^{D}
\sum_{j,m,n=1}^{N}
\frac{g^{1}_{n}}{\sqrt{N}}
\frac{g^{1}_{m}}{\sqrt{N}}
\,\delta_{nj}\,\delta_{mj}\,
\frac{x_k}{\sqrt{D}}
\frac{x_k}{\sqrt{D}}
=
\eta_0\,G^{1}\,\Phi^{0}
\end{aligned}
\end{equation}

The router weight matrix $Q$ involves the complex Jacobian of the softmax operator. Integrating this out, we find:
\begin{equation}
\begin{aligned}
\eta_0\kappa
\sum_{j=1}^{\kappa N}
\sum_{k=1}^{N}
\frac{\partial h^{4}}{\partial Q_{jk}}
\frac{\partial h^{4}}{\partial Q_{jk}}
&=
\eta_0\kappa N
\sum_{j=1}^{\kappa N}
\sum_{i,i'=1}^{\kappa N}
\Phi^{1}
\,g^{\phi}_{i}\,g^{\phi}_{i'}
\,\phi^{i}\,\phi^{i'}
(\delta_{ij}-\phi^{j})
(\delta_{i'j}-\phi^{j})
=
{\eta_0}\Phi^1G^{1,\phi}
\end{aligned}
\end{equation}

Aggregating these contributions yields the full macroscopic evolution equation:
\begin{equation}
\begin{aligned}
    \frac{d f_{\mu}(t)}{dt}
&= \frac{\eta_0}{P}\sum_\nu\Delta_\nu
\Bigg[
G^{1}_{\mu\nu}(t,t)\,\Phi^{0}_{\mu\nu}(t,t)
+
\frac{N_e}\kappa\frac{1}{\kappa N}\sum_{i=1}^{\kappa N}G_{\mu\nu,i}^{2,\i}(t,t)\Phi^1_{\mu\nu}(t,t)\\
&
+
\frac{1}{\kappa N}\sum_{i=1}^{\kappa N}G^{2,\o}_{\mu\nu,i}(t,t)\Phi^{2,\i}_{\mu\nu,i}(t,t)\ +
\Phi^{3}_{\mu\nu}(t,t)
+ \Phi^1_{\mu\nu}(t,t) G^{1,\phi}_{\mu\nu}(t,t)
\Bigg]
\end{aligned}
\end{equation}

\subsection{Evolution of weights, preactivations, and pregradients}
The updates to the router weights are:
\begin{equation}
\label{appendixDdQdt_r2}
\begin{aligned}
\frac{d Q_{jk}}{dt}
&=\frac{\eta}{P}\sum_\mu\Delta_\mu \frac{\partial f_\mu}{\partial Q_{jk}}
=\frac{\eta_0\gamma\kappa\sqrt N}{P}\sum_\mu\Delta_\mu \frac{\partial h^4_\mu}{\partial Q_{jk}}
= \frac{\eta_0\gamma_0\kappa N}{P}\sum_\mu\Delta_\mu \frac{\partial h^4_\mu}{\partial Q_{jk}} \\
&=
\frac{\eta_0 \gamma_0 }{P \sqrt N}
\sum_{\mu} \Delta_\mu
\left[\kappa N
\sum_{a=1}^{M}
g_a^{\phi}\,
\phi_a
(\delta_{aj}-\phi_j)\right]
\, \sigma(h_k^{1}) \\[6pt]
&= 
\frac{\eta_0 \gamma_0 }{P \sqrt N}
\sum_{\mu} \Delta_\mu
g_j^{1,\phi}
\sigma(h_k^{1}) \\[6pt]
\end{aligned}
\end{equation}
We obtain thus:
\begin{equation}
\begin{aligned}
\psi_\mu(t)
&= \frac{1}{\sqrt N}\, Q(t)\, \sigma\!\big(h_{\mu}^{1}(t)\big) \\
&= \chi^{\phi}_\mu(t)
+ \frac{\eta_0 \gamma_0}{P}\,
\int_{0}^{t} ds
\sum_{\nu} \Delta_{\nu}(s)\,
\Phi^{1}_{\mu\nu}(s,t)
g^{1,\phi}_\nu(t) \\
\tilde z^{1,\phi}_\mu &= \tilde\xi^{1,\phi}_\mu +\frac{\eta_0\gamma_0}{P}\int_0^\infty dt \sum_\nu \Delta_\nu G^{1,\phi}_{\nu\mu}\,\sigma(h^1_{\nu j})
\end{aligned}
\end{equation}
The other preactivation and pregradient updates follow similarly:

\begin{equation}
\begin{aligned}
\frac{d W^{1}(t)}{dt}
&=
\frac{\eta_0 \gamma_0 \sqrt{N}}{P}
\sum_{\mu}
\Delta_{\mu}(t)\,
\frac{g_{\mu}^{1}(t)}{\sqrt{N}}\,
\frac{1}{\sqrt{D}}\,
x_{\mu}
\\[3pt]
\Rightarrow\quad
h^{1}_{\mu}(t)
&=
\frac{1}{\sqrt{D}}\,W^{1}(0)\,x_{\mu}
+
\int_{0}^{t} ds\,
\frac{\eta_0 \gamma_0}{P}
\sum_{\nu}
\Delta_{\nu}(s)\,
g_{\nu}^{1}(s)\,
\Phi^{0}_{\mu\nu}(t,s)
\\[6pt]
\frac{d W^{2,\i}_i(t)}{dt}
&=
\frac{\eta_0 \gamma_0 \sqrt{N} }{P}
\sum_{\mu}
\Delta_{\mu}(t)\,
\frac{g_{\mu,i}^{2,\i}(t)}{\kappa \sqrt{N}}\,
\frac{1}{\sqrt N}\,
\sigma\!\big(h^{1}_{\mu}(t)\big)
\\[3pt]
\Rightarrow\quad
h^{2,\text{in}}_{\mu, i}(t)
&=
\frac1{\sqrt N} W^{2,\i}_i(0)\,\sigma(h^1_\mu(t))
+
\frac{\eta_0 \gamma_0}{P}
\int_{0}^{t} ds\,
\sum_{\nu}
\Delta_{\nu}(s)\,
g_{\nu_i}^{2,\i}(s)\,
\Phi^{1}_{\mu\nu}(t,s)
\\[6pt]
\frac{d W^{2,\o}_i(t)}{dt}
&=
\frac{\eta_0 \gamma_0 \sqrt{N} }{P}
\sum_{\mu}
\Delta_{\mu}(t)\,
\frac{g_{\mu,i}^{2,\o}(t)}{N^{3/2}\kappa}\,
\frac{\sqrt{\kappa N}}{\sqrt{{N_{e}}}}\,
\sigma\!\big(h^{2,\text{in}}_{\mu, i}(t)\big)
\\
&=
\frac{\eta_0\gamma_0}{P\sqrt\kappa}\frac1{\sqrt{NN_e}}\sum_\mu\Delta_\mu(t) g_i^{2,\o}(t)\sigma(h_i^{2,\i}(t))
\\[3pt]
\Rightarrow\quad
h^{2,\o}_{\mu,i}(t)
&=
\frac{\sqrt{\kappa N}}{\sqrt{{N_{e}}}}\,
W^{2,\o}_i(0)\,
\sigma\!\big(h^{2,\text{in}}_{\mu, i}(t)\big)
+
\frac{\eta_0 \gamma_0}{P}
\int_{0}^{t} ds\,
\sum_{\nu}
\Delta_{\nu}(s)\,
g_{\nu,i}^{2,\o}(s)\,
\Phi^{2,\i}_{\mu\nu,i}(t,s)
\\[6pt]
\frac{d w^{4}(t)}{dt}
&=
\frac{\eta_0 \gamma_0 \sqrt{N}}{P}
\sum_{\mu}
\Delta_{\mu}(t)\,
\frac{1}{\sqrt{N}}\,
h^{3}_{\mu}(t)
=
\frac{\eta_0 \gamma_0}{P}
\sum_{\mu}
\Delta_{\mu}(t)\,
h^{3}_{\mu}(t)
\end{aligned}
\end{equation}

Using the same expressions for the evolution of the weights to derive expressions for the pregradients:

\begin{equation}
\begin{aligned}
z^{3}_\mu(t)
&=
g^{3}_\mu(t)
=
w^{4}_\mu(t)
=
\xi^{3}_\mu(0)
+
\frac{\eta_0 \gamma_0}{P}
\int_{0}^{t} ds
\sum_{\nu}
\Delta_{\nu}(s)\,
h^{3}_{\nu}(s)
\\[6pt]
z^{2,\o}_{\mu,i}(t)
&= \kappa g^{2,\o}_{\mu,i}(t)
=
\tilde\phi^{i}_{\mu}(t)\,g^{3}_{\nu}(t)
=
\frac{\eta_0 \gamma_0 }{P}
\tilde\phi^i_\mu(t)
\int_{0}^{t} ds
\sum_{\nu}
\Delta_{\nu}(s)\,
h^{3}_{\nu}(s)
\\[6pt]
z^{2,\i}_{\mu,i}(t)
&=
\sqrt\frac{\kappa}{N N_e}\,W^{2,\o\,\top}_i(t)\,g^{2,\o}_{\mu,i}(t)
\\
&=
\xi^{2,\i}_{\mu,i}(t)
+
\frac{\eta_0 \gamma_0}{P N_e}
\int_{0}^{t} ds
\sum_{\nu}
\Delta_{\nu}(s)\,
G^{2,\o}_{\nu\mu,i}(s,t)\,
\sigma\!\big(h^{2,\i}_{\nu,i}(s)\big)
\\[6pt]
\tilde z^{1}_{\mu,i}(t)
&=
W^{2,\i\,\top}_i(t)\,g^{2,\i}_i(t)
\\
&=
\tilde\xi^{1}_{\mu,i}(t)
+
\frac{\eta_0 \gamma_0N_e}{\sqrt N P}
\int_{0}^{t} ds
\sum_{\nu}
\Delta_{\nu}(s)\, G^{2,\i}_{\nu\mu,i}(s,t)\,
\sigma\!\big(h^{1}_{\nu}(s)\big)
\end{aligned}
\end{equation}
Finally for the router gradient:
\begin{equation}
    \begin{aligned}
        g^{1,\phi}_i
        &=
        \frac1{\kappa N}
        \sum_{m=1}^{\kappa N}g_m^\phi\tilde\phi^m(\kappa N \delta_{mi}-\tilde\phi^i)\\
        &=
        \xi^{g^{1,\phi}}_{i}
        +
        \frac{\eta_0\gamma_0}{P}\int_0^tds \sum_\nu \Delta_\nu G^3 \Phi^{2,in}_i \tilde\phi^i \tilde\phi^i
        -
        \frac{\eta_0\gamma_0}{P}\int_0^tds \sum_\nu \Delta_\nu G^3 \bar \Phi^{2,in} \tilde\phi^i
        .
    \end{aligned}
\end{equation}

\subsection{Stochastic initial fields}
To isolate the purely deterministic part of the trajectory from the random initialization, we define a set of initial stochastic fields $\mathcal{F}$. These fields encapsulate the randomness of the initial weights:

\begin{equation}
\mathcal{F} =\{\chi_\mu^1(t),\chi_{\mu,i}^{2,\i}(t),\chi_{\mu,i}^{2,\o}(t),\chi^\phi_\mu(t),\tilde\xi_{\mu,i}^{1}(t),\xi_{\mu,i}^{2,\i}(t),\xi_\mu^{3}(t),\xi^\phi_\mu(t),\xi^{g^\phi}_{\mu i}(t) \}_{i \in\{1,...,M\}, \mu\in\{1,...,P\}}
\end{equation}

\begin{equation}
\begin{aligned}
\chi^{1}_{\mu} &= \frac{1}{\sqrt{D}}\, W^{1}(0)\, x_{\mu} \\
\chi^{2,\i}_{\mu,i}(t) &= \frac{1}{\sqrt N}\, W^{2,\i}_i(0)\, \sigma\!\big(h^{1}_{\mu}(t)\big) \\
\chi^{2,\o}_{\mu,i}(t) &= \sqrt\frac{\kappa N}{{N_{e}}}W^{2,\o}_i(0)\, \sigma\!\big(h^{2,\text{in}}_{\mu, i}(t)\big) \\
\chi^{\phi}_{\mu}(t) &= \frac{1}{\sqrt N}\, Q(0)\, \sigma(h^1_{\mu}(t)) \\
\tilde\xi^{1}_{\mu,i}(t) &= \big(W^{2,\i}_i(0)\big)^{\top}\, g^{2,\i}_{\mu,i}(t) \\
\xi^{2,\i}_{\mu,i}(t) &= \sqrt\frac{\kappa }{NN_e}\, \big(W^{2,\o}_i(0)\big)^{\top}\, g^{2,\o}_{\mu,i}(t) \\
{\xi^{1,\phi}_\mu(t)} &= \frac{1}{\kappa \sqrt N} Q(0)^\top {g^{1,\phi}_\mu(t)} \\
\xi^{g^\phi}_{\mu i}(t) &= \frac1{\sqrt{{N_{e}} }N} g^{3\top}_\mu(t)\,W^{2,\o}_i(0)\,\sigma({h^{2,\text{in}}_{\mu, i}}(t))
\end{aligned}
\end{equation}
A critical step in rendering the MoE partition function computationally tractable is distinguishing between expert-local fields (which are specific to an expert $i$) and global fields (which impact the shared base layer or the final aggregated output). This distinction allows the massive partition function to factorize over the experts.

We define the router-averaged stochastic fields to capture the macroscopic effect of the local processes:
\begin{equation}
    \bar\chi^3_\mu(t) = \frac{1}{\kappa N}\sum_{i=1}^{\kappa N}\tilde\phi^i_\mu(t)\chi^{2,\o}_{\mu,i}(t)
\end{equation}
\begin{equation}
    \bar\xi^1_\mu(t) = \frac1{\kappa\sqrt N}\sum_{i=1}^{\kappa N}\tilde\xi^{1}_{\mu,i}(t)
\end{equation}

Substituting these, the global aggregated output $h^3_\mu$ and the gradient arriving at the base layer $z^1_\mu$ can be expressed entirely in terms of order parameters that are smooth over the ensemble of experts:
\begin{equation}
    h^3_\mu(t) = \bar\chi^3_\mu(t) + \frac{\eta_0\gamma_0}{ P}\int_0^t ds \sum_\nu \Delta_\nu(s) \bar\Phi^{2,\i}_{\mu\nu}(s,t) z^3_\nu(s)
\end{equation}

\begin{equation}
z^1_\mu(t) = \bar\xi^{1}_\mu(t) + \xi^{1,\phi}_\mu(t) + \frac{\eta_0 \gamma_0}{P}\int_0^t \! ds\sum_{\nu}\Delta_{\nu}(s) \left[\frac1{\sqrt N} \bar G_{\mu \nu}^{2,\i}(s,t) + G^{1,\phi}_{\mu\nu}(s,t) \right] \sigma(h^{1}_{\nu}(s))
\end{equation}

\subsection{Deriving the DMFT Action}
We formulate the DMFT by writing the moment-generating function for the system's trajectories and performing a disorder average over the initial weights $\theta_0$.
\begin{equation}
\adjustbox{max width=\textwidth}{$
\begin{aligned}
Z\propto&\Bigg\langle
\int d\mathcal{F}
\exp\Bigg(
\sum_{\mu}
    \int_{0}^{\infty} \! dt \,i\,\Bigg[
\hat{\chi}_\mu^{1} \!\cdot\! 
\big( \chi_\mu^{1} - \frac{1}{\sqrt{D}} W^{1}(0) x_\mu \big)\\
&
+ \sum_{i=1}^{\kappa N}\hat{\chi}_{\mu,i}^{2,\i} \!\cdot\! 
\Big( \chi_{\mu,i}^{2,\i} - \frac{1}{\sqrt N} W^{2,\i}_i(0) \sigma(h_\mu^{1}(t)) \Big) \\
&
+ \hat{\bar\chi}_\mu^{3} \!\cdot\! 
\Big( \bar\chi_\mu^{3} - \frac{1}{\kappa N}\sum_{i=1}^{\kappa N}\tilde\phi^i_\mu(t)\sqrt\frac{\kappa N}{{N_{e}}}W^{2,\o}_i(0) 
\sigma({h^{2,\text{in}}_{\mu, i}}(t)) \Big) \\
&
+ \hat{\bar\xi}_\mu^{1} \!\cdot\! 
\Big( \bar\xi_\mu^{1} - \frac{1}{\kappa \sqrt N}\sum_{i=1}^{\kappa N} 
W^{2,\i}_i(0)^{\!\top} g_{\mu,i}^{2,\i}(t) \Big)\\
&
+ \sum_{i=1}^{\kappa N}\hat{\xi}_{\mu,i}^{2,\i} \!\cdot\! 
\Big( \xi_{\mu,i}^{2,\i} - \sqrt\frac{\kappa}{N N_e} 
W^{2,\o}_i(0)^{\!\top} g_{\mu,i}^{2,\o}(t) \Big)
+ \hat{\xi}_\mu^{3} \!\cdot\! 
\Big(\xi_\mu^{3}-w^{4}(0)^{\!\top}\Big)
\\
&+\hat\chi^\phi_\mu\cdot\bigg(\chi^\phi_\mu-\frac{1}{\sqrt N}Q(0)\sigma(h^1_\mu(t))\bigg)
+
\hat\xi^{1,\phi}_\mu\cdot\bigg(\xi^{1,\phi}_\mu-\frac{1}{\kappa \sqrt N}Q(0)^\top g^{1,\phi}_\mu(t)
\bigg)\\
&
+ \sum_{m=1}^{\kappa N}
\hat{\xi}^{g^{1,\phi}}_{\mu, m}(t)
\Big(
\xi^{g^{1,\phi}}_{\mu, m}(t)
-
\frac{1}{\kappa N}
\sum_{i=1}^{\kappa N}
\frac{1}{\sqrt{N_e}\,N}
\,
g^{3\,T}_\mu(t)
W^{2,\text{out}}_{i}
\sigma\!\big(h^{2,\text{in}}_{\mu,i}(t)\big)
\tilde{\phi}_\mu^{\,i}(t)
\left(
\kappa N\,\delta_{im}
-
\tilde{\phi}^{\,m}_\mu(t)
\right)
\Big)
\Bigg]\Bigg)\\
&\times\exp\!\Bigg(
    \sum_{\mu}
    \int_{0}^{\infty} \! dt \,
    \Bigg[
    j_\mu^1(t)\cdot\chi_\mu^1(t)+\sum_{i=1}^Mj_{\mu,i}^{2,\i}(t)\cdot \chi_{\mu,i}^{2,\i}(t)
    + \bar j_{\mu}^{3}(t)\cdot \bar\chi_{\mu}^{3}(t)
    + \sum_{i=1}^Mv_{\mu,i}^{2,\i}(t)\cdot\xi_{\mu,i}^{2,\i}(t)   \\
&
    +\bar v_{\mu}^{1}(t)\cdot\bar\xi_{\mu}^{1}(t) + v_\mu^{3}\cdot\xi^{3}_\mu(t)
    +j^\phi_\mu(t)\cdot\chi^\phi_\mu(t) + 
    v_\mu^{1,\phi}(t)\cdot\xi_\mu^{1,\phi}(t)
    + 
    v_\mu^{g^{1,\phi}}(t)\cdot\xi_\mu^{g^{1,\phi}}(t)
    \Bigg]
  \Bigg)\Bigg\rangle_{\theta_0}.
\end{aligned}
$}
\end{equation}

The resulting  partition function obtained after the integration over the initial weights $\theta_0$ is:

\begin{equation}
\begin{aligned}
&Z  \propto 
\int d\mathcal{F}
\exp\Bigg\{-\frac12 \sum_{\mu\nu}\int_0^\infty \diff t \int_0^\infty \diff s\bigg[
\hat{\chi}_\mu^{1}(t)\!\cdot\!\hat{\chi}_\nu^{1}(s)\,\Phi_{\mu\nu}^{0}(t,s)
+(\sum_{i=1}^M\hat{\chi}_{\mu,i}^{2,\i}(t)\!\cdot\!\hat{\chi}_{\nu,i}^{2,\i}(s)\\
&+ \hat\chi^\phi_\mu(t)\cdot\hat\chi^\phi_\nu(s))\,\Phi_{\mu\nu}^{1}(t,s)+
\frac{N_{e}}{\kappa }\hat{\bar\xi}_\mu^{1}(s)\!\cdot\!\hat{\bar\xi}_\nu^{1}(t)\,\bar G_{\mu\nu}^{2,\i}(s,t)
\\
&+
\frac\kappa{N_e}\sum_{i=1}^M\hat{\xi}_{\mu,i}^{2,\i}(s)\!\cdot\!\hat{\xi}_{\nu,i}^{2,\i}(t)\,G_{\mu\nu}^{3}(s,t)
\tilde\phi^i_\mu(s)
\tilde\phi^i_\nu(t)\\
&
+
\hat{\xi}_\mu^{3}(s)\!\cdot\!\hat{\xi}_\nu^{3}(t)+ 
\frac1{\kappa }\hat{\xi}_\mu^{1,\phi}(s)\!\cdot\!\hat{\xi}_\nu^{1,\phi}(t) G^{1,\phi}_{\mu\nu}(t,s)
\\
&
+
\kappa
\sum_{i=1}^{\kappa N}
\hat{\xi}^{g^{1,\phi}}_{\mu,i}(s)\,
\hat{\xi}^{g^{1,\phi}}_{\nu,i}(t)\,
\Phi^{2,\text{in}}_{\mu\nu,i}(s,t)\,
G^{3}_{\mu\nu}(s,t)\,
\tilde{\phi}_{\mu}^{\,i}(s)\,
\tilde{\phi}_{\nu}^{\,i}(t)\\
&
+
\bar{\Phi}^{2,\mathrm{in}}_{\mu\nu}(s,t)\,
\left(
\hat{\bar\chi}^{3}_\mu(s)\cdot\hat{\bar\chi}^{3}_\nu(t)
\right)
\bigg]
-i\sum_{\mu,\nu}\int_0^\infty dt\int_0^\infty ds\,\,\Big[
\bar A_{\mu\nu}^{1}(s,t) \frac1{\kappa } \sum_{i=1}^{\kappa N}\hat\chi^{2,\i}_{\mu,i}(s)\cdot g^{2,\i}_{\nu,i}(t) \\
&
+
\frac1{\kappa}\hat\chi_\mu^{\phi}(s)\cdot g_\nu^{1,\phi}(t)A_{\mu\nu}^{\phi}(s,t)
+
\frac{1}{N_e}
\sum_{i=1}^{\kappa N}
\big(
\hat\xi^{2,\mathrm{in}}_{\mu,i}(s)\,\cdot\,
\sigma(h^{2,\mathrm{in}}_{\nu,i}(t))
\big)
\tilde{\phi}^{\,i}_\mu(s)\tilde{\phi}^{\,i}_\nu(t)\,
\bar{B}^{2,\mathrm{in}}_{\mu\nu}(s,t)\, \\
&
+
\sum_{i=1}^{\kappa N}\hat\xi^{g^{1,\phi}}_{\mu,\,i\,}(t)\tilde\phi_{\nu}^i(s)\,\tilde A^{\tilde\phi}_{\mu\nu,\,i\,}(t,s)
\Big]
    +\sum_{\mu}
    \int_{0}^{\infty} \! dt \,
    \Bigg[
    j_\mu^1(t)\cdot\chi_\mu^1(t)+\sum_{i=1}^Mj_{\mu,i}^{2,\i}(t)\cdot \chi_{\mu,i}^{2,\i}(t)
    + \bar j_{\mu}^{3}(t)\cdot \bar\chi_{\mu}^{3}(t)
     \\
&+\sum_{i=1}^Mv_{\mu,i}^{2,\i}(t)\cdot\xi_{\mu,i}^{2,\i}(t)
    +\bar v_{\mu}^{1}(t)\cdot\bar\xi_{\mu}^{1}(t)
    +v_\mu^{3}\cdot\xi^{3}_\mu(t)
    +j^\phi_\mu(t)\cdot\chi^\phi_\mu(t) + 
    v_\mu^{1,\phi}(t)\cdot\xi_\mu^{1,\phi}(t)
    \Bigg]
\Bigg\}
\end{aligned}
\end{equation}
Where we have introduced the $O(1)$ kernels (also called response functions)
\begin{equation}
\begin{aligned}
\bar B^{2,\i}_{\mu\nu}(s,t)
&=
-\frac i{ N}\hat{\bar\chi}^{3}_\mu(s)\cdot g^{3}_\nu(t)
\\
\bar A^1_{\mu\nu}(s,t)
&=
-\frac i{ N} \hat{\bar\xi}^{1}_\mu(s)\cdot \sigma(h^{1}_\nu(t))
\\
A^{1,\phi}_{\mu\nu}(s,t)
&= 
-\,\frac{i}{N}\,
\hat{\xi}^{1,\phi}_\mu(s)\cdot
\sigma(h^{1}_{\nu}(t)) \\
\tilde A^{\tilde\phi}_{\mu\nu,\,i\,}(t,s)
    &=
    \kappa \Phi^{2,\i}_{\mu\nu,\,i\,}(t,s)\,G_{\mu\nu}^3(t,s)\,\tilde\phi^i_\mu(t)\,A^{g^{1,\phi}}_{\mu\nu}(t,s)
    +
    \kappa A_{\mu\nu,\,i\,}^{2,\i}(t,s)\,G^3_{\mu\nu}(t,s)\,\tilde\phi^i_{\mu}(t)\\
    &
    +
    \Phi^{2,\i}_{\mu\nu,\,i\,}(t,s)\,\tilde\phi^i_\mu(t)\,\bar B^{2,\i}_{\mu\nu}(t,s)
    \quad
    +
    \frac12\kappa\bar\Phi^{2,\i}_{\mu\nu}(t,s)\,G^3_{\mu\nu}(t,s)\,A^{g^{1,\phi}}_{\mu\nu}(t,s)\\
    &
    +
    \bar\Phi^{2,\i}_{\mu\nu}(t,s)\,\bar B^{2,\i}_{\mu\nu}(t,s)
    +
    \kappa \bar A^{2,\i}_{\mu\nu}(t,s)\,G^3_{\mu\nu}(t,s)
\end{aligned}
\end{equation}

We must constrain the solutions to be physical, so for each $i,\mu,\nu,s,t$ we multiply in the following resolutions of the identity to enforce the definitions of the order parameters.

\begin{equation}
    1 = \int\frac{\diff \Phi^1_{\mu\nu}(s,t)\diff\hat\Phi^1_{\mu\nu}(s,t)}{2\pi i N^{-1}}\exp\left[\hat\Phi^1_{\mu\nu}(s,t)\Big(N\Phi^1_{\mu\nu}(s,t)-\sigma(h_\mu^1(s))\cdot \sigma(h^1_\nu(t))\Big)\right]
\end{equation}

\begin{equation}
\label{Phi2i_r2}
    1 = \int\frac{\diff \Phi^{2,\i}_{\mu\nu,i}(s,t)\diff\hat\Phi^{2,\i}_{\mu\nu,i}(s,t)}{2\pi i {N_{e}}^{-1}}\exp\left[\hat\Phi^{2,\i}_{\mu\nu,i}(s,t)\Big({N_{e}}\Phi^{2,\i}_{\mu\nu,i}(s,t)-\sigma({h^{2,\text{in}}_{\mu, i}}(s))\cdot \sigma(h^{2,\text{in}}_{\nu,i}(t))\Big)\right]
\end{equation}

\begin{equation}
    1 = \int\frac{\diff G^1_{\mu\nu}(s,t)\diff\hat G^1_{\mu\nu}(s,t)}{2\pi i N^{-1}}\exp\left[\hat G^1_{\mu\nu}(s,t)\Big(NG^1_{\mu\nu}(s,t)-g_\mu^1(s)\cdot g^1_\nu(t)\Big)\right]
\end{equation}

\begin{equation}
\label{G2i_r2}
    1 = \int\frac{\diff G^{2,\i}_{\mu\nu,i}(s,t)\diff\hat G^{2,\i}_{\mu\nu,i}(s,t)}{2\pi i {N_{e}}^{-1}}\exp\left[\hat G^{2,\i}_{\mu\nu,i}(s,t)\Big({N_{e}} G^{2,\i}_{\mu\nu,i}(s,t)-g_{\mu,i}^{2}(s)\cdot g_{\nu,i}^{2}(t)\Big)\right]
\end{equation}

\begin{equation}
    1 = \int\frac{\diff G^3_{\mu\nu}(s,t)\diff\hat G^3_{\mu\nu}(s,t)}{2\pi i N^{-1}}\exp\left[\hat G^3_{\mu\nu}(s,t)\Big(NG^3_{\mu\nu}(s,t)-g_\mu^3(s)\cdot g^3_\nu(t)\Big)\right]
\end{equation}

\begin{equation}
    1 = \int\frac{\diff G^{1,\phi}_{\mu\nu}(s,t)\diff\hat G^{1,\phi}_{\mu\nu}(s,t)}{2\pi i {\kappa N}^{-1}}\exp\left[\hat G^{1,\phi}_{\mu\nu}(s,t)\Big(\kappa N G^{1,\phi}_{\mu\nu}(s,t)-g_\mu^{1,\phi}(s)\cdot g^{1,\phi}_\nu(t)\Big)\right]
\end{equation}

\begin{equation}
    1 = \int\frac{\diff\bar B^{2,\i}_{\mu\nu}(s,t)\diff \bar A^{2,\i}_{\mu\nu}(s,t)}{2\pi i N^{-1}}\exp\left[-\bar A^{2,\i}_{\mu\nu}(s,t)\Big(N\bar B^{2,\i}_{\mu\nu}(s,t)+i\hat{\bar\chi}_\mu^{3}(s)\cdot g^{3}_\nu(t)\Big)\right]
\end{equation}

\begin{equation}
    1 = \int\frac{\diff \bar A^{1}_{\mu\nu}(s,t)\diff \bar B^{1}_{\mu\nu}(s,t)}{2\pi i N^{-1}}\exp\left[-\bar B^{1}_{\mu\nu}(s,t)\Big(N\bar A^{1}_{\mu\nu}(s,t)+i\hat{\bar\xi}_\mu^{1}(s)\cdot \sigma(h^1_\nu(t))\Big)\right]
\end{equation}

\begin{equation}
    1 = \int\frac{\diff A^{1,\phi}_{\mu\nu}(s,t)\diff B^{1,\phi}_{\mu\nu}(s,t)}{2\pi i N^{-1}}\exp\left[-B^{1,\phi}_{\mu\nu}(s,t)\Big(NA^{1,\phi}_{\mu\nu}(s,t)+i\hat\xi_\mu^{1,\phi}(s)\cdot \sigma(h^1_\nu(t))\Big)\right]
\end{equation}

\begin{equation}
    1 = \int\frac{\diff A^{g^{1,\phi}}_{\mu\nu}(s,t)\diff B^{g^{1,\phi}}_{\mu\nu}(s,t)}{2\pi i (\kappa N)^{-1}}\exp\left[-B^{g^{1,\phi}}_{\mu\nu}(s,t)\Big(\kappa N A^{g^{1,\phi}}_{\mu\nu}(s,t)+i\hat\xi_\mu^{g^{1,\phi}}(s)\cdot \tilde\phi_\nu(t)\Big)\right]
\end{equation}

\begin{equation}
    1 = \int\frac{\diff A^{2,\i}_{\mu\nu,i}(s,t)\diff  B^{2,\i}_{\mu\nu,i}(s,t)}{2\pi i N_e^{-1}}\exp\left[ B^{2,\i}_{\mu\nu,i}(s,t)\Big(N_e A^{2,\i}_{\mu\nu,i}(s,t)+i\hat{\xi}_{\mu,i}^{2,\i}(s)\cdot \sigma(h^{2,\i}_{\nu,i}(t))\Big)\right]
\end{equation}

\begin{equation}
1=\int\frac{d\Phi_{\mu\nu}^{3}(t,s)\,d\hat{\Phi}_{\mu\nu}^{3}(t,s)}{2\pi i\, N^{-1}}\exp\!\left[\hat{\Phi}_{\mu\nu}^{3}(t,s)\left(N\Phi_{\mu\nu}^{3}(t,s)-h_{\mu}^{3}(t)\cdot h_{\nu}^{3}(s)\right)\right]
\end{equation}

\begin{equation}
1=\int\frac{d\bar\Phi_{\mu\nu}^{2,\i}(t,s)\,d\hat{\bar\Phi}_{\mu\nu}^{2,\i}(t,s)}{2\pi i\, }\exp\!\left[\hat{\bar\Phi}_{\mu\nu}^{2,\i}(t,s)\left(\kappa N\bar\Phi_{\mu\nu}^{2,\i}(t,s)-\sum_{i=1}^{\kappa N}\tilde\phi^i_\mu(s)\tilde\phi^i_\nu(t)\Phi^{2,\i}_{\mu\nu,i}(s,t)\right)\right]
\end{equation}

\begin{equation}
1=\int\frac{d\bar G_{\mu\nu}^{2,\i}(t,s)\,d\hat{\bar G}_{\mu\nu}^{2,\i}(t,s)}{2\pi i\, (\kappa N)^{-1}}\exp\!\left[\hat{\bar G}_{\mu\nu}^{2,\i}(t,s)\left(\kappa N\bar G_{\mu\nu}^{2,\i}(t,s)-\sum_{i=1}^{\kappa N}G^{2,\i}_{\mu\nu,i}(s,t)\right)\right]
\end{equation}

\subsection{Softmax}
We need to encode the behaviour of softmax in the DMFT. Recall that $\phi^i = \text{softmax}(\psi^i)$, so $\tilde\phi^i =\kappa N \phi^i = \kappa N\, \text{softmax}(\psi^i)$. That is,

\begin{equation}
    \tilde\phi^i_\mu(t) = \kappa N \phi^i_\mu(t) = \frac{e^{\psi^i_\mu(t)}}{\frac{1}{\kappa N}\sum_{j=1}^{\kappa N}e^{\psi^j_\mu(t)}}
\end{equation}

This invites the definition for each input $\mu$ of the \textit{global} order parameter

\begin{equation}
    \mathcal S_\mu(t):=\frac{1}{\kappa N}\sum_{j=1}^{\kappa N}e^{\psi^j_\mu(t)}
    ,
\end{equation}

which we enforce via the Fourier-transformed delta function

\begin{equation}
1=\int\frac{d \mathcal S_{\mu}(t)\,d\hat{ \mathcal S}_{\mu}(t)}{2\pi i\, (\kappa N)^{-1}}\exp\!\left[\hat{\mathcal S}_{\mu}(t)\left(\kappa N \mathcal S_{\mu}^{}(t)-\sum_{i=1}^{\kappa N} e^{\psi^i_\mu(t)}\right)\right]
\end{equation}

\subsection{Partition function}
Define the set of all kernels and conjugates (indexed for time and feature, although this is omitted for brevity in \ref{kernels_set_r2}):

\begin{equation}
\label{kernels_set_r2}
\begin{aligned}
    \mathcal K = \{&\Phi^1,\hat\Phi^1,\Phi^{2,\i}_i,\hat\Phi^{2,\i}_i,\bar\Phi^{2,\i},\hat{\bar\Phi}^{2,\i},\Phi^3,\hat\Phi^3,G^1,
    \hat G^1,\bar G^{2,\i},\hat{\bar G}^{2,\i},G^{2,\i}_i,\hat G^{2,\i}_i,
    \\&G^{3},\hat G^{3},G^{1,\phi},\hat G^{1,\phi},\bar A^{1},\bar B^{1},\bar B^{2,\i},\bar A^{2,\i},A^{1,\phi},B^{1,\phi}
    ,A^{g^{1,\phi}},B^{g^{1,\phi}}
    ,B^{2,\i}_i,A^{2,\i}_i
    \}
\end{aligned}
\end{equation}

We can partition these order parameters into the set $\mathcal K_{global}$ of global order parameters, and the set $\mathcal K_{exp-loc}$ of expert-local order parameters.

\begin{equation}
\label{kernels_set_global}
\begin{aligned}
    \mathcal K_{global} = \{&
    \Phi^1,\hat\Phi^1,\bar\Phi^{2,\i},\hat{\bar\Phi}^{2,\i},\Phi^3,\hat\Phi^3,G^1,
    \hat G^1,\bar G^{2,\i},\hat{\bar G}^{2,\i},G^{1,\phi},\hat G^{1,\phi},G^{3},\hat G^{3},\\
    & A^{1},\bar B^{1},\bar B^{2,in},\bar A^{2,in},A^{1,\phi},B^{1,\phi},A^{g^{1,\phi}},B^{g^{1,\phi}}\}
\end{aligned}
\end{equation}

\begin{equation}
\label{kernels_set_expert_local}
\begin{aligned}
    \mathcal K_{exp-loc} = \{&\Phi^{2,\i}_i,\hat\Phi^{2,\i}_i,\Phi^{2,\o}_i,\hat\Phi^{2,\o}_i,G^{2,\i}_i,
    \hat G^{2,\i}_i,
    B^{2,\i}_i,A^{2,\i}_i\}
\end{aligned}
\end{equation}

\begin{equation}
Z \;\propto\; \int \Big(\prod_{\mu,\nu}\prod_{t,s} d\mathcal K_{global}\Big)\;
\exp\!\Big( N\,S[\mathcal K]\Big), \quad \textrm{where}
\label{eq:part_func_reg_ii}
\end{equation}

\begin{equation}
\begin{aligned}
S[\mathcal K]
&=
\sum_{\mu,\nu}\int dt\int ds\Bigg[
\hat\Phi^1_{\mu\nu}(t,s)\,\Phi^1_{\mu\nu}(t,s)
+\,\hat G^1_{\mu\nu}(t,s)\,G^1_{\mu\nu}(t,s)
+\kappa\,\hat G^{1,\phi}_{\mu\nu}(t,s)\,G^{1,\phi}_{\mu\nu}(t,s)\\
&
- \bar A^{2,\i}_{\mu\nu}(t,s)\,\bar B^{2,\i}_{\mu\nu}(t,s)
- \bar B^{1}_{\mu\nu}(t,s)\,\bar A^{1}_{\mu\nu}(t,s)
-\,B^{1,\phi}_{\mu\nu}(t,s)\,A^{1,\phi}_{\mu\nu}(t,s)
-
\kappa B^{g^{1,\phi}}_{\mu\nu}(t,s)
\,A^{g^{1,\phi}}_{\mu\nu}(t,s)\\
&
+\,\hat\Phi^3_{\mu\nu}(t,s)\,\Phi^3_{\mu\nu}(t,s)
+\kappa \hat{\bar\Phi}^{2,\i}_{\mu\nu}(t,s)\bar\Phi^{2,\i}_{\mu\nu}(t,s)
+\kappa \hat{\bar G}^{2,\i}_{\mu\nu}(t,s)\bar G^{2,\i}_{\mu\nu}(t,s)\\
&
+
\kappa \hat{\mathcal S}_\mu(t)\,\mathcal S_\nu(s)
+\,\hat G^3_{\mu\nu}(t,s)\,G^3_{\mu\nu}(t,s)
\Bigg]
\\
&\qquad
+
\frac1N\sum_{n=1}^N\ln\mathcal Z_N^{global}[j^1_n,\bar j^3_n,\bar v^1_n, v^{1,\phi}_n,v^3_n]
+
\frac{\kappa}{\kappa N}\sum_{i=1}^{\kappa N}\ln Z^{local}[j^{2,\i}_i,v^{2,\i}_i,j^{3,\o}_i,j^\phi_i]
\end{aligned}
\end{equation}

\begin{equation}
\begin{aligned}
\ln \mathcal Z_{N}^{global}&[j^1_n,\bar j^3_n,\bar v^1_n, v^{1,\phi}_n,v^3_n]
=
-\frac12\sum_{\mu,\nu}\int dt\int ds\Big[
\hat{\chi}_\mu^{1}(t)\hat{\chi}_\nu^{1}(s)\,\Phi_{\mu\nu}^{0}(t,s)
+ \hat{\xi}_\mu^{3}(t)\hat{\xi}_\nu^{3}(s)\\
&
+ \frac1{\kappa}\hat{\xi}_\mu^{1,\phi}(t)\hat{\xi}_\nu^{1,\phi}(s)\, G^{1,\phi}_{\mu\nu}(t,s)
+
\frac{N_{e}}{\kappa } \hat{\bar\xi}^1_\mu(t) \hat{\bar\xi}^1_\nu(s)\bar G^{2,\i}_{\mu\nu}(t,s)
+\,\hat{\bar\chi}^3_\mu(t)\,\hat{\bar\chi}^3_\mu(t) \,\bar\Phi^{2,\i}
\Big]\\
&
-\sum_{\mu,\nu}\int dt\int ds\Big[
\hat\Phi^1_{\mu\nu}(t,s)\,\sigma(h_\mu^1(t))\,\sigma(h^1_\nu(s))
\\
&
+ \hat G^1_{\mu\nu}(t,s)\,g_\mu^1(t)\,g^1_\nu(s)
+ \hat\Phi^3_{\mu\nu}(t,s)\,h^3_\mu(t) h^3_\nu(s)
+ \hat G^3_{\mu\nu}(t,s)\,g_\mu^3(t)\,g^3_\nu(s)
\Big]\\
&
-i \sum_{\mu,\nu}\int dt\int ds
\Big[
\; B^{1,\phi}_{\mu\nu}(t,s)\hat\xi^{1,\phi}_\mu(t)\sigma(h_\nu^1(s))
+ \bar B^1_{\mu\nu}(t,s)\hat{\bar\xi}^1_\mu(t)\sigma(h^1_\nu(s))
\\
&
+\bar A^{2,\i}_{\mu\nu}(t,s) \hat{\bar\chi}^3_\mu(t)g^3_\nu(s)
\Big]
\\
&
+\sum_{\mu}\int dt\Big[
(v_{\mu,n}^{3}+i\hat\xi^{3}_{\mu,n}(t))\xi^{3}_{\mu,n}(t)
+(v_{\mu,n}^{1,\phi}(t)+i\hat\xi_{\mu,n}^{1,\phi}(t))\xi_{\mu,n}^{1,\phi}(t)+(j_{\mu,n}^1(t)
\\
&
+i\hat\chi^1_{\mu,n}(t))\chi_{\mu,n}^1(t)
+ (\bar j^3_{\mu,n}(t) + i\hat{\bar\chi}^3_{\mu,n}(t))\bar\chi^3_{\mu,n}(t)
+ (\bar v^1_{\mu,n}(t) + i\hat{\bar\xi}^1_{\mu,n}(t))\bar\xi^1_{\mu,n}(t)
\Big]
\end{aligned}
\end{equation}

\begin{equation}
\adjustbox{max width=\textwidth}{$
\begin{aligned}
    \mathcal Z^{local}&[j^{2}_{i},v^{2}_{i},j^{3,i},j^\phi_i]
    =\int\Big(\prod_{\mu\nu}\prod_{t,s}d\mathcal K_{exp-loc}\Big) \times
    \exp\Bigg(
    \sum_{\mu\nu}\int_0^\infty dt \int_0^\infty ds\;
    \Big[
    {N_{e}} \,
    \hat\Phi^{2,\i}_{\mu\nu,i}(t,s)\,\Phi^{2,\i}_{\mu\nu,i}(t,s) \\
    &
+{N_{e}}\, \hat G^{2,\i}_{\mu\nu,i}(t,s)\,G^{2,\i}_{\mu\nu,i}(t,s)
+N_e\,  B^{2,\i}_{\mu\nu,i}(t,s)\, A^{2,\i}_{\mu\nu,i}(t,s)
\Big]\Bigg)\\
&
\times
    \frac1N \prod_{j=1}^{N_{e}}\mathcal Z_{{N_{e}}}\left[\left[j_i^{2,\i}\right]_j,\left[v_i^{2,\i}\right]_j\right]
    \,\times\,
    \mathcal Z_M\left[j_i^{\phi},v_i^{g^{1,\phi}}\right]
\end{aligned}
$}
\end{equation}

\begin{equation}
\begin{aligned}
\mathcal Z_{{N_{e}}\,j}&\left[\left[j_i^{2,\i}\right]_j,\left[v_i^{2,\i}\right]_j\right]
:= \int d\mathcal F\;
\exp\Bigg\{
-\frac12\sum_{\mu,\nu}\int dt\int ds\Big[
\,\left[\hat\chi^{2,\i}_{\mu,i}(t)\right]_j\,\left[\hat\chi^{2,\i}_{\nu,i}(s)\right]_j\,\Phi^1_{\mu\nu}(t,s)\\
&\quad
+\frac{\kappa}{N_e}\,\left[\hat\xi^{2,\i}_{\mu,i}(t)\right]_j\,\left[\hat\xi^{2,\i}_{\nu,i}(s)\right]_j\,G_{\mu\nu}^{3}(s,t)
\tilde\phi^i_\mu(s)
\tilde\phi^i_\nu(t)
\\
&\quad
+\frac1{N_e}
G^3_{\mu\nu}(s,t)
\sigma(\left[h^{2,\i}_{\mu,i}(s)\right]_j)\sigma(\left[h^{2,\i}_{\nu,i}(t)\right]_j)
\hat{\xi}^{g^{\phi}}_{\mu,i}(t)\hat{\xi}^{g^{\phi}}_{\nu,i}(s)
\Big]\\
&\quad
-i\sum_{\mu,\nu}\int dt\int ds\Bigg[
\frac{1}{\kappa }\,\left[\hat\chi^{2,\i}_{\mu,i}(t)\right]_j\,\left[g^{2,\i}_{\nu,i}(s)\right]_j\,\bar A^{1}_{\mu\nu}(t,s)
\\
&\quad 
-B^{2,\i}_{\mu\nu,i}(t,s) \left[\hat\xi^{2,\i}_{\mu,i}(t)\right]
\sigma(\left[h^{2,\i}_{\nu,i}(s)\right])
\Bigg]
\\
&\quad
-\sum_{\mu,\nu}\int dt\int ds\Big[
\hat\Phi^{2,\i}_{\mu\nu,i}(t,s)\,\sigma({\left[h^{2,\text{in}}_{\mu, i}(t)\right]_{j}})\,\sigma({\left[h^{2,\text{in}}_{\nu, i}(s)\right]_{j}})\\
& \quad
+\hat G^{2,\i}_{\mu\nu,i}(t,s)\,\left[g^{2,\i}_{\mu,i}(t)\right]_j\,\left[g^{2,\i}_{\nu,i}(s)\right]_j
\Big]\\
&\quad
+\sum_{\mu}\int dt\Big[
\big(\left[j^{2,\i}_{\mu,i}(t)\right]_j+i\left[\hat\chi^{2,\i}_{\mu,i}(t)\right]_j\big)\,\left[\chi^{2,\i}_{\mu,i}(t)\right]_j\\
&\quad
+\big(\left[v^{2,\i}_{\mu,i}(t)\right]_j+i\left[\hat\xi^{2,\i}_{\mu,i}(t)\right]_j\big)\,\left[\xi^{2,\i}_{\mu,i}(t)\right]_j
\Big]
\Bigg\}.
\end{aligned}
\end{equation}

\begin{equation}
\adjustbox{max width=\textwidth}{$
\begin{aligned}
    \mathcal Z_M&[j_m^\phi,v^{g^{1,\phi}}_m]
    :=
    \int d\mathcal F\;\exp\Bigg\{
    -\frac12\sum_{\mu,\nu}\int dt\int ds
    \Big[
    \hat\chi^\phi_{\mu m}(t)\hat\chi^\phi_{\nu m}(s)\,\Phi_{\mu\nu}^{1}(t,s) \\
    &
    +
    \kappa
\hat{\xi}^{g^{1,\phi}}_{\mu,m }(s)\,
\hat{\xi}^{g^{1,\phi}}_{\nu,m }(t)\,
\Phi^{2,\text{in}}_{\mu\nu,m }(s,t)\,
G^{3}_{\mu\nu}(s,t)\,
\tilde{\phi}_{\mu}^{\,m }(s)\,
\tilde{\phi}_{\nu}^{\,m }(t)
    \Big]
    \\
    &
    -i \sum_{\mu,\nu}\int dt\int ds \,\Big[\frac1\kappa\hat\chi_{\mu m}^{\phi}(t)g_\nu^{1,\phi}(s)A_{\mu\nu}^{1,\phi}(t,s)
+
\hat\xi^{g^{1,\phi}}_{\mu m}(t)\tilde\phi_\nu^m(s) 
\left[ 
B^{g^{1,\phi}}_{\mu\nu}(t,s)
+
\tilde A^{\tilde\phi}_{\mu\nu,\,m\,}(t,s)
\right]
    \Big]
\\
&
+
\sum_{\mu,\nu}\int_0^\infty dt\int_0^\infty ds\,\,\Big[
A^{2,\i}_{\mu\nu,m}(s,t)
\tilde{\phi}^{\,m}_\mu(s)\tilde{\phi}^{\,m}_\nu(t)\,
\bar{B}^{2,\mathrm{in}}_{\mu\nu}(s,t)\,
\Big]
    \\
    &
    -\sum_{\mu,\nu}\int dt\int ds
    \Big[
    \hat
    G^{1,\phi}_{\mu\nu}(t,s)\,g_{\mu m}^{1,\phi}(t)\,g^{1,\phi}_{\nu m}(s)
+
\hat{\bar\Phi}^{2,\i}_{\mu\nu}(t,s)\,\tilde\phi^{m}_{\mu}(t)\,\tilde\phi^{m}_{\nu}(s)\Phi^{2,\i}_{\mu\nu,m }(t,s)\\
&
+\hat{\bar G}^{2,\i}_{\mu\nu}(t,s)\,G^{2,\i}_{\mu\nu,m }(t,s)
\Big]
    \\
    &
-\sum_{\mu}\int dt\,\hat {\mathcal S}_\mu(t)\,e^{\psi^m_\mu(t)}
+\sum_{\mu}\int dt
    (j^\phi_{\mu m}(t)+i\hat\chi^\phi_{\mu m}(t))\,\chi^\phi_{\mu m}(t)
    +
    (v^{g^\phi}_{\mu m}(t)+i\hat\xi^{g^{1,\phi}}_{\mu m}(t))\,\xi^{g^{1,\phi}}_{\mu m}(t)
    \Bigg\}.
\end{aligned}
$}
\end{equation}

\subsection{Saddle point approximation}

To write down the saddle point equations, we define first the single-site distributions. 

We can write 

\begin{equation}
\adjustbox{max width=\textwidth}{$
    \mathcal Z_{N}^{global}[j^1_n,\bar j^3_n,\bar v^1_n, v^{1,\phi}_n,v^3_n]
    =
    \int d\mathcal F\;
    \exp\bigg\{
    -\mathcal H^{global}_{N} [\{
    \chi^{1}_{\mu n},\bar\chi^3_{\mu n},\bar\xi^1_{\mu n},\xi^{1,\phi}_{\mu n},\xi^3_{\mu n},j^1_n,\bar j^3_n,\bar v^1_n, v^{1,\phi}_n,v^3_n
    \}_\mu]
    \bigg\}
$}
\end{equation}

\begin{equation}
    \mathcal Z_{M }[j^{\phi}_m] 
    =
    \int d\mathcal F\;
    \exp\bigg\{
    -\mathcal H_{M} [\{\chi^{\phi}_{\mu,m},j^{\phi}_{\mu,m}\}_\mu]
    \bigg\}
\end{equation}

Where $\mathcal H$ in each case is the logarithm of the integrand of the corresponding $\mathcal Z$. We can then define for each $\mathcal Z\in\{\mathcal Z_{N}^{global},\mathcal Z_{{N_{e}}\,j },\mathcal Z_{M}\}$ and the corresponding $\mathcal H$ the average

\begin{equation}
\label{defn_average_single_site_r2}
    \langle\mathcal O(\{\chi,\xi\})\rangle_{\mathcal Z} 
    = \frac{1}{\mathcal Z}\int\prod_\mu d\mathcal F \exp(-\mathcal H[\{\chi,\xi\},\{j,v\}]) \mathcal O(\{\chi,\xi\})
\end{equation}

With this apparatus in place, we can take saddle equations. We treat expert-local kernels as microvariables which are implicitly defined in terms of $\chi,\xi$, and so do not take saddle equations of them at this point.

\subsection{Saddle-Point Equations}
Since the integrand in eqn.\ref{eq:part_func_reg_ii} has the form $e^{NS[\mathcal{K}]}$, in the $\lim_N \rightarrow \infty$ we can use the saddle point approximation, which consist in finding the set of equations that lead to a stationary action $S[\mathcal{K}].$

We note the following:
\begin{itemize}
    \item
At zero source, all single site averages $\langle\rangle_{\mathcal Z_{N}^{global}[j^1_n,\bar j^3_n,\bar v^1_n, v^{1,\phi}_n,v^3_n]}$ are equivalent, so we may write $\langle\rangle_{\mathcal Z_{N}^{global}}$ for the average over the single-site distributions when $\bm j^1,\bm{\bar j}^3,\bm{\bar v}^1, \bm v^{1,\phi},\bm v^3\to 0$.
\item
\textit{Conditional on the set $\mathcal K_{global}$} of global kernels, $\mathcal Z^{local}$ factorises over experts, and so we can write $\langle\rangle_{\mathcal Z_{N_e j}| \mathcal K_{global}}$ for the conditional average over the distribution defined by $\mathcal Z_{N_e\,j}\left[\left[j_i^{2,\i}\right]_j,\left[v_i^{2,\i}\right]_j\right]$ as $\bm j^{2,\i},\bm v^{2,\i}\to0$, and $\langle\rangle_{\mathcal Z_{M}| \mathcal K_{global}}$ for the conditional average over the distribution defined by $\mathcal Z_{M}[j_m^\phi,v^{g^\phi}_m]$ as $\bm j^{\phi},\bm v^{g^{1,\phi}}\to0$.
\item 
Expert-local variables follow single-site processes, so we can drop expert indices
\item
By a derivation similar to that in sec.\ref{sec:dmft_regime1}, we can prove that conjugate kernels defined as covariances between conjugate fields vanish, since they have no physical meaning and can not influence the dynamics in addition to imposing constraint when introducing kernel definitions.
\end{itemize}

The global kernels are given by their expectation values over the single-site global distribution $\mathcal Z^{\text{global}}_{N}$:

\begin{equation}
\label{global_saddle_v2_r2}
\begin{aligned}
\Phi^{1}_{\mu\nu}(s,t) &= \big\langle \sigma(h_{\mu}^{1}(s))\,\sigma(h_{\nu}^{1}(t)) \big\rangle_{\mathcal Z^{\text{global}}_{N}} \\[3pt]
\Phi^{3}_{\mu\nu}(s,t) &= \big\langle h_{\mu }^{3}(s)\,h_{\nu }^{3}(t) \big\rangle_{\mathcal Z^{\text{global}}_{N}} \\[3pt]
G^{1}_{\mu\nu}(s,t) &= \big\langle g_{\mu }^{1}(s)\,g_{\nu }^{1}(t) \big\rangle_{\mathcal Z^{\text{global}}_{N}} \\[3pt]
G^{3}_{\mu\nu}(s,t) &= \big\langle g_{\mu }^{3}(s)\,g_{\nu }^{3}(t) \big\rangle_{\mathcal Z^{global}_{N}} \\[3pt]
A^{1,\phi}_{\mu\nu}(s,t) &= -i\big\langle \hat{\xi}_{\mu}^{1,\phi}(s)\, \sigma(h^1_\nu(t))(t) \big\rangle_{\mathcal Z_{N}^{global}} \\[3pt]
\bar A^{1}_{\mu\nu}(s,t) &= - i\, \big\langle \hat{\bar\xi}^1_{\mu }(t)\sigma(h^1_{\nu }(s)) \big\rangle_{\mathcal Z_{N}^{global}} \\[3pt]
\bar B^{2,\i}_{\mu\nu}(s,t) &= - i\, \big\langle \hat{\bar\chi}^3_{\mu }(t)g^3_{\nu }(s) \big\rangle_{\mathcal Z_{N}^{global}}
\end{aligned}
\end{equation}

The router kernels require expectations over the expert ensemble distribution $\mathcal Z_{M}$, conditioned on the global fields:
\begin{equation}
\begin{aligned}
\bar\Phi^{2,\i}_{\mu\nu}(s,t) &= \big\langle \tilde\phi_{\mu}(s)\,\tilde\phi_\nu(t) \Phi^{2,\i}_{\mu\nu}(s,t) \big\rangle_{\mathcal Z_{M}|\mathcal K_{global}} \\[3pt]
\bar G^{2,\i}_{\mu\nu}(s,t) &= \big\langle G^{2,\i}_{\mu\nu,}(s,t) \big\rangle_{\mathcal Z_{M}|\mathcal K_{global}} \\[3pt]
G^{1,\phi}_{\mu\nu}(s,t) &= \big\langle g_{\mu}^{1,\phi}(s)\, g_{\nu}^{1,\phi}(t) \big\rangle_{\mathcal Z_{M}|\mathcal K_{global}} \\[3pt]
{\mathcal S}_{\mu}(t) &= \big\langle e^{\psi_\mu(t)} \big\rangle_{\mathcal Z_{M}|\mathcal K_{global}}
\end{aligned}
\end{equation}

\subsection{Expert-local order parameters}
The local kernels are determined by independent processes within each expert's internal dimensions:
\begin{equation}
\label{local_saddle_equations_v1_r2}
    \begin{aligned}
\Phi^{2,\i}_{\mu\nu}(s,t) &= \frac1{N_e} \sum_{j=1}^{N_e}  \sigma({\left[h^{2,\text{in}}_{\mu}(t)\right]_{j}})\,\sigma({\left[h^{2,\text{in}}_{\nu}(s)\right]_{j}})  \\[3pt]
G^{2,\i}_{\mu\nu}(s,t) &= \frac1{N_e} \sum_{j=1}^{N_e} {\left[g^{2,\text{in}}_{\mu}(t)\right]_{j}}\,{\left[g^{2,\text{in}}_{\nu}(s)\right]_{j}}  \\[3pt]
A^{2,\i}_{\mu\nu}(s,t) &= i\,\kappa\, \sum_{j=1}^{N_e} \left[\hat\xi^{2,\i}_{\mu}(s)\right]_j\sigma(\left[h_{\nu}^{2,\i}\right]_j) \\[3pt]
B^{2,\i}_{\mu\nu}(s,t) &= - \frac1{N_e} \langle \tilde{\phi}^{}_\mu(s)\tilde{\phi}^{}_\nu(t)\, \rangle_{\mathcal Z_{M}|\mathcal K_{global}} \bar{B}^{2,\mathrm{in}}_{\mu\nu}(s,t)
    \end{aligned}
\end{equation}

All non-physical conjugate kernels identically vanish at the saddle point: $\hat{\Phi}^{3} = \hat\Phi^1 = \hat{G}^{1} = \hat{\bar\Phi}^{2,\i} = \dots = 0$.

\subsection{Hubbard-Stratonovich transformation}
The Hubbard-Stratonovich trick allows us to rewrite the quadratic terms over conjugate fields as Gaussian integrals over linear conjugate fields:
\begin{equation}
\label{hubbard_trick_r2}
\adjustbox{max width=\textwidth}{$
    \exp\!\left(-\frac{1}{2}\,\mathbf{\hat x}^{\top} A\,\mathbf{\hat x}\right)
= \int_{\mathbb{R}^{d}}
\frac{d\mathbf{u}}{(2\pi)^{d/2} \sqrt{\det A}}\,
\exp\!\left(-\frac{1}{2}\,\mathbf{u}^{\top}A^{-1}\mathbf{u}
- i\,\mathbf{u}\!\cdot\!\mathbf{\hat x}\right)
= \big\langle \exp(-i\,\mathbf{u}\!\cdot\!\mathbf{\hat x}) \big\rangle_{\mathbf{u} \sim \mathcal{N}(0,A)}.
$}
\end{equation}
where $\mathbf{\hat x}$ is a generic conjugate fields in the partition function above.
Using Stein's lemma, (integration by parts on a Gaussian variable), we can reformulate the definitions of each of the kernels A and B as response functions: 
\begin{equation}
\label{Aellli_avg_r2}
    \begin{aligned}
A^{\,\ell}_{\mu\nu,i}(t,s)
&= -\,i\,
\big\langle
\hat{\xi}^{\,\ell}_{\nu,i}(t)\,
\sigma\!\big(h^{\,\ell-1}_{\mu,i}(s)\big)
\big\rangle_{\beta_{\nu,i}^{\ell}(s)}
\\
&=
\big[G^{\,\ell}_{i}\big]^{-1}
\Big\langle
\big(
\xi^{\,\ell}_{i}
-
B^{\,\ell}_{i}\,
\sigma\!\big(h^{\,\ell-1}_{i}\big)
\big)\,
\sigma\!\big(h^{\,\ell-1}_{i}\big)
\Big\rangle_{\beta_{\nu,i}^{\ell}(s)}
\\
&=
\left\langle\frac{\partial\sigma(h_{\mu,i}^{\ell-1}(t))}{\partial\beta_{\nu,i}^{\ell\top}(s)}\right\rangle_{\beta_{\nu,i}^{\ell}(s)}
\end{aligned}
\end{equation}
It easier now integrate over all the $\mathbf{\hat x}$'s, since the argument of the exponential in $\mathcal Z$ has been linearised with respect to them all. Doing so yields delta functions that give us the final DMFT dynamics.

\subsection{DMFT Dynamics}
Piecing together the self-consistent order parameters and the exact integrations over the Hubbard-Stratonovich fields, the fully rigorous DMFT description of the infinite-width MoE is summarized below.
\begin{equation}
\label{dmft_gaussians}
\adjustbox{max width=\textwidth}{$
\begin{aligned}
\alpha^1(t)&\sim\mathcal{N}(0,\Phi^0)\quad \quad\alpha^{\phi}(t)\sim\mathcal{GP}(0,\Phi^1) \quad\left[\alpha^{2,\i}\right]_j\sim\mathcal{GP}(0,\Phi^1) \\[4pt]
\bar\alpha^3(t)&\sim\mathcal{GP}(0,\bar\Phi^{2,\i}) \quad \beta^{g^{1,\phi}}\sim\mathcal{GP}(0,\kappa \Phi^{2,\i}\tilde\phi\tilde\phi G^3) \\[4pt]
\bar\beta^{1}(t)&\sim\mathcal{GP}(0,\frac{N_e}{\kappa}\bar G^{2,\i}) \quad \left[\beta^{2,\i}\right]_j\sim\mathcal{GP}(0,\frac\kappa{N_e}G^3\tilde\phi\tilde\phi) \\[4pt]
\beta^{3}(t)&\sim\mathcal{N}(0,\mathbb I) \quad \tilde\beta^{1,\phi}(t)\sim\mathcal{GP}(0,\frac{1}{\kappa}G^{1,\phi})
\end{aligned}
$}
\end{equation}
\begin{equation}
\adjustbox{max width=\textwidth}{$
\begin{aligned}
g^{1,\phi}_\mu(t) &= \beta^{g^{1,\phi}}_\mu(t)\, + \, \frac{\eta_0\gamma_0}{\kappa P}\int_0^tds \sum_\nu\big[] \Delta_\nu(s) G^3_{\mu\nu}(s,t) \left[\Phi^{2,in}_{\mu\nu}(s,t) \tilde\phi_\nu(t) - \bar \Phi^{2,in}_{\mu\nu}(s,t) \right]\\
&\quad + \tilde A^{\tilde\phi}\, \big] \tilde\phi_\mu(s) \\[4pt]
z^1_\mu(t) &= \bar\beta^{1}_\mu(t) + \tilde\beta^{1,\phi}_\mu(t) + \frac{\eta_0 \gamma_0}{P}\int_0^t \! ds\sum_{\nu}\big[ B^{1,\phi}_{\nu\mu}(s,t) + \bar B^1_{\mu\nu}(s,t) \\ 
&\quad+\Delta_{\nu}(s) G^{1,\phi}_{\mu\nu}(s,t) \big] \, \sigma(h^{1}_{\nu}(s)) \\[4pt]
 \left[z^{2,\i}_{\mu}(t)\right]_j &= \left[\beta^{2,\i}_\mu(t)\right]_j + \frac{\eta_0 \gamma_0}{P{ N_e}}\int_0^t \! ds\sum_{\nu}\left[\Delta_{\nu}(s) G_{\mu \nu}^{3}(s,t)\,\tilde\phi_\mu(s)\,\tilde\phi_\nu(t) + B^{2,\i}_{\mu\nu}(t,s) \right] \, \sigma\left( {\left[h^{2,\text{in}}_{\nu}(s)\right]_{j}}\right) \\[4pt]
 z_\mu^{3}(t)&=\beta_{\mu}^{3}(t)+ \frac{\eta_0 \gamma_0}{P}\int_0^t \! ds\sum_{\nu}\Delta_{\nu}(s) \,  h^{3}_{\nu}(s) \\[4pt]
h_{\mu}^1(t)&=\alpha_{\mu}^1(t)+\frac{\eta_0 \gamma_0}{P} \int_0^t \! ds \sum_{\nu} \Delta_{\nu} \,\Phi_{\mu\nu}^0\, (z^1_\mu(s)\,\dot\sigma(h^1_\mu(s)))  \\[4pt]
 {\left[h^{2,\text{in}}_{\mu}(t)\right]_{j}} &= \alpha^{2,\i}_\mu(t) + \frac{\eta_0 \gamma_0}{P} \int_0^t \! ds \sum_{\nu} \big[ \Delta_{\nu}(s)\Phi_{\mu \nu}^{1}(t,s)\\
 & \qquad\qquad\qquad\qquad\qquad\qquad+ \frac1\kappa \bar A^1_{\mu\nu}(s,t)\big]\, \left(\dot\sigma\left( \left[h^{2,\text{in}}_{\mu}(t)\right]_{j}\right)\odot  \left[z_{\nu}^{2,\i}(s)\right]_j\right) \\[4pt]
h^3_\mu(t) &= \bar\alpha^3_\mu(t) + \frac{\eta_0\gamma_0}{P}\int_0^t ds \sum_\nu \left[\bar A^{2,\i}_{\mu\nu}(t,s) + \Delta_\nu(s) \bar\Phi^{2,\i}_{\mu\nu}(s,t) \right] z^3_\nu(s)
\end{aligned}
$}
\end{equation}

\newpage

\section{DMFT Analysis for Regime II (\texorpdfstring{$\mu$}{mu}P)}\label{sec:dmft_regime2_mup}

For $\mu$P in Regime II, an analogous derivation as in the previous section, which we omit for conciseness, yields the following set of self-consistent DMFT equations:

\begin{equation*}
\adjustbox{max width=\textwidth}{$
\begin{aligned}
\alpha^1(t)&\sim\mathcal{N}(0,\Phi^0)\quad
\quad\alpha^{\phi}(t)\sim\mathcal{GP}(0,\Phi^1)
\quad\left[\alpha^{2,\i}\right]_j\sim\mathcal{GP}(0,\Phi^1)
\\[4pt]
\bar\beta^{1}(t)&\sim\mathcal{GP}(0,\frac{N_e}{\kappa}\bar G^{2,\i})\quad
\beta^{3}(t)\sim\mathcal{N}(0,\mathbb I)
\quad
\tilde\beta^{1,\phi}(t)\sim\mathcal{GP}(0,\frac{1}{\kappa}G^{1,\phi})
\\[4pt] 
\Phi_{\mu\nu}^{0} &= \frac1Dx_\mu x_\nu^\top,
\Phi_{\mu\nu}^{1}(t,s) = \langle \sigma(h_\mu^{1}(t))\sigma(h_\nu^{1}(s)) \rangle_{\mathcal Z_N^{global}}, \\
\Phi^{2,\i}_{\mu\nu}(s,t)
&=
\frac1{N_e}
\sum_{j=1}^{N_e}
\big\langle
\sigma({\left[h^{2,\text{in}}_{\mu}(t)\right]_{j}})\,\sigma({\left[h^{2,\text{in}}_{\nu}(s)\right]_{j}})
\big\rangle_{\mathcal Z_{N_e\,j}|\mathcal K_{global}} \\
\Phi^{3}_{\mu\nu}(s,t) &=\big\langle h_\mu^{3}(t) h^{3}_\nu(s)\rangle_{\mathcal Z_{N}^{global}},\quad
G_{\mu\nu}^{1}(t,s) = \langle [\dot\sigma(h_\mu^{1}(t))\odot z_\mu^{1}(t)] [\dot\sigma(h_\nu^{1}(s))\odot z_\nu^{1}(s)] \rangle_{\mathcal Z_N^{global}}\\[4pt]
\bar\Phi^{2,\i}_{\mu\nu}(s,t)
&=
\big\langle
\tilde\phi_{\mu}(s)\,\tilde\phi_\nu(t) \Phi^{2,\i}_{\mu\nu}(s,t)
\big\rangle_{\mathcal Z_{M}|\mathcal K_{global}}\qquad\bar G^{2,\i}_{\mu\nu}(s,t)
=
\big\langle
G^{2,\i}_{\mu\nu}(s,t)
\big\rangle_{\mathcal Z_{M}|\mathcal K_{global}}\\
G^{2,\i}_{\mu\nu}(s,t)
&=
\frac1{N_e}
\sum_{j=1}^{N_e}
\big\langle
\left[
\dot\sigma(\left[h^{2,\text{in}}_{\mu}(t)\right]_{j})\odot
\left[z^{2,\text{in}}_{\mu}(t)\right]_{j}\right]\,\left[
\dot\sigma(\left[h^{2,\text{in}}_{\nu}(s)\right]_{j})\odot
\left[z^{2,\text{in}}_{\nu}(s)\right]_{j}\right]
\big\rangle_{\mathcal Z_{N_e\,j}|\mathcal K_{global}}\\[4pt]
 G^{1,\phi}_{\mu\nu}(s,t)
&=
\big\langle
g_{\mu}^{1,\phi}(s)\, g_{\nu}^{1,\phi}(t)
\big\rangle_{\mathcal Z_{M}|\mathcal K_{global}},
\quad
G^{3}_{\mu\nu}(s,t)
=
\big\langle
z_{\mu }^{3}(s)\,z_{\nu }^{3}(t)
\big\rangle_{\mathcal Z^{global}_{N}}
\\[4pt]
 \bar A^1_{\mu\nu}(t,s)
 &=
 \left\langle\frac{\partial\sigma(h^1_\mu(t))}{\partial\bar\beta^{1\top}_\nu(s)}\right\rangle_{\mathcal Z^{global}_N}
 ,\quad
 \bar B^1_{\mu\nu}(t,s)
 =
 \sum_{j=1}^{N_e}\left\langle\frac{\partial \left[g^{2,\i}_\mu(t))\right]_j}{\partial\left[\alpha^{2,\i\top}_\nu(s)\right]_j}\right\rangle_{\mathcal Z_{N_e\,j}|\mathcal K_{global}}
 \\[4pt]
A^{1,\phi}_{\mu\nu}(t,s)& =\left\langle\frac{\partial\sigma(h_{\mu}^{1}(t))}{\partial\tilde\beta_{\nu}^{1,\phi\top}(s)}\right\rangle_{\mathcal Z_N^{global}},
\quad
B^{1,\phi}_{\mu\nu}(t,s)=\left\langle\frac{\partial  g_{\mu}^{1,\phi}(t))}{\partial\alpha_{\nu}^{\phi\top}(s)}\right\rangle_{\mathcal Z_{M}|\mathcal K_{global}}
\\[4pt]
g^{1,\phi}_\mu(t)
&=
\frac{\eta_0\gamma_0}{\kappa P}\int_0^tds \sum_\nu \Delta_\nu(s) G^3_{\mu\nu}(s,t) \left[\Phi^{2,in}_{\mu\nu}(s,t) \tilde\phi_\nu(t)
-
\bar \Phi^{2,in}_{\mu\nu}(s,t) \right]
\tilde\phi_\mu(s)
\\[4pt]
z^1_\mu(t) 
&= 
\bar\beta^{1}_\mu(t) + \tilde\beta^{1,\phi}_\mu(t)
+
\frac{\eta_0 \gamma_0}{P}\int_0^t \! ds\sum_{\nu}\left\{
B^{1,\phi}_{\nu\mu}(s,t)
+
\bar B^1_{\mu\nu}(s,t)+
\Delta_{\nu}(s) 
\left[
\bar G_{\mu \nu}^{2,\i}(s,t)
+
G^{1,\phi}_{\mu\nu}(s,t)
\right]
\right\}
\, \sigma(h^{1}_{\nu}(s))
\\[4pt]
 \left[z^{2,\i}_{\mu}(t)\right]_j &=\frac{\eta_0 \gamma_0}{P\sqrt{ N_e}}\int_0^t \! ds\sum_{\nu}\Delta_{\nu}(s) G_{\mu \nu}^{3}(s,t)\,\tilde\phi_\mu(s)\,\tilde\phi_\nu(t) \, \sigma\left( {\left[h^{2,\text{in}}_{\nu}(s)\right]_{j}}\right), \quad j\in\{1,2,...,N_e\}
\\[4pt]
 z_\mu^{3}(t)&=\beta_{\mu}^{3}(t)+ \frac{\eta_0 \gamma_0}{P}\int_0^t \! ds\sum_{\nu}\Delta_{\nu}(s) \,  h^{3}_{\nu}(s)
 \\[4pt]
h_{\mu}^1(t)&=\alpha_{\mu}^1(t)+\frac{\eta_0 \gamma_0}{P}
    \int_0^t \! ds
    \sum_{\nu} \Delta_{\nu} \,\Phi_{\mu\nu}^0\, (z^1_\mu(s)\,\dot\sigma(h^1_\mu(s)))  \\[4pt]
 {\left[h^{2,\text{in}}_{\mu}(t)\right]_{j}}
 &=
 \alpha^{2,\i}_\mu(t)
 +
 \frac{\eta_0 \gamma_0}{P}
    \int_0^t \! ds
    \sum_{\nu}
    \left[
    \Delta_{\nu}(s)\Phi_{\mu \nu}^{1}(t,s)
    +
    \frac1\kappa
    \bar A^1_{\mu\nu}(s,t)\right]\, 
    \left(\dot\sigma\left( \left[h^{2,\text{in}}_{\mu}(t)\right]_{j}\right)\odot  \left[z_{\nu}^{2,\i}(s)\right]_j\right) ,\quad \forall j \\[4pt]
h^3_\mu(t) &=\frac{\eta_0\gamma_0}{\kappa P}\int_0^t ds \sum_\nu 
\Delta_\nu(s) 
    \bar\Phi^{2,\i}_{\mu\nu}(s,t)
    z^3_\nu(s)
    \\[4pt]
\tilde\phi_\mu(t) &= \frac{e^{\hat\psi_\mu(t)}}{\mathcal S_\mu(t)}
,\qquad
{\mathcal S}_{\mu}(t)
=
\big\langle
e^{\hat\psi_\mu(t)}
\big\rangle_{\mathcal Z_{M}|\mathcal K_{global}}
,\qquad
\hat\psi_\mu(t) = m_\mu(t) \psi_\mu(t)
,\qquad
\\
m_\mu(t) &= \mathbf{1}(\psi_\mu(t) -\tau_\mu(t)) \qquad
\text{where } \tau_\mu(t) \text{ is chosen such that } \rho = \left\langle m_\mu(t) \right\rangle_{\mathcal Z_{M}|\mathcal K_{global}}
\\[4pt]
\frac{d f_{\mu}(t)}{dt}
&=\frac{\eta_0}{P}\sum_\nu\Delta_\nu(t)
\Bigg[
G^{1}_{\mu\nu}(t,t)\,\Phi^{0}_{\mu\nu}(t,t)
+
\left[\frac{N_{e}}\kappa \bar G_{\mu\nu}^{2,\i}(t,t)
+G^{1,\phi}_{\mu\nu}(t,t)
\right]\Phi^1_{\mu\nu}(t,t)\\
&
+
\kappa G^{3}_{\mu\nu}(t,t) \bar\Phi^{2,\i}_{\mu\nu}(t,t)
+
\Phi^{3}_{\mu\nu}(t,t)
\Bigg]
\end{aligned}
$}
\end{equation*}

\newpage

\maketitle

\section{DMFT Analysis for Regime III (MSSP)}\label{sec:dmft_regime3}

\subsection{Architectural Definitions}
In this section we are considering the case in which $M, N, N_e \to \infty$ at a fixed ratio. Specifically, we define $\kappa = \frac{M}{N}, \iota = \frac{N_e}{N}$, where $\kappa$,$\iota$ are order one in $N$. The forward pass, following \textit{MSSP}, is defined as:

\begin{equation}
\label{infinite_experts_architecture_r3}
    \begin{aligned}
\mathbb R^N \ni h_\mu^1 
&= \frac{1}{\sqrt D} W^1 x_\mu 
&\qquad
\mathbb R^M \ni \psi_\mu 
&= \frac{1}{\sqrt N} Q \sigma(h^1_\mu) \\
\mathbb R^{N_e} \ni h^{2,\text{in}}_{\mu,i} 
&= \frac{1}{\sqrt N} W^{2,\text{i}}_i \sigma(h_\mu^1) 
&\qquad
\mathbb R^M \ni \phi_\mu 
&= \text{softmax}(\psi_\mu) \\
\mathbb R^N \ni h^{2,\text{out}}_{\mu,i} 
&= \frac{1}{\sqrt N} W^{2,\text{out}}_i \sigma(h^{2,\text{in}}_{\mu,i}) 
&\qquad
\mathbb R^N \ni h^3_\mu 
&= \sum_{i=1}^M \phi_\mu^i \, h^{2,\text{o}}_{\mu,i} \\
\mathbb R \ni h^4_\mu 
&= \frac{1}{\sqrt N} w^{4\mathsf{T}} h^3_\mu 
&\qquad
f_\mu 
&= \frac{1}{\gamma} h^4_\mu
\end{aligned}
\end{equation}

\subsection{Learning Rates and Initialization}
To ensure that the network evolves dynamically in the infinite-width limit without gradients vanishing or exploding, we apply specific learning rate scalings:
\begin{equation}
    \begin{aligned}
        \eta&=\eta_0\gamma^2 \qquad
        \eta_Q = \eta_0\gamma^2 \kappa \qquad
        \eta_E = \eta_0\gamma^2 N \qquad
        \gamma =\gamma_0\sqrt N \qquad
        \eta_0,\gamma_0\sim O(1)\\
    \end{aligned}
\end{equation}

The network weights are initialized from standard Gaussian distributions:
\begin{equation}
    w_\alpha^{4}(0),\left[W_{i}^{2,\o}(0)\right]_{\alpha\beta},\left[W_{i}^{2,\i}(0)\right]_{\alpha\beta},W_{\alpha\beta}^{1}(0),Q_{\alpha\beta}(0)\sim\mathcal N(0,1)
\end{equation}
but noticing that for each expert's matrices $W_{i}^{2,\i},W_{i}^{2,\o}$ we share the same initial weights $W_{i}^{2,\i}(0)=W^{2,\i}(0), W_{i}^{2,\o}(0)=W^{2,\o}(0)$  
For brevity in our derivations, we denote the collection of all network parameters as $\bm\theta = \text{Vec}\{W^1,W_i^{2,\i},W_i^{2,\o},w^4,Q\}$. 

Due to the normalization constraint of the softmax operator over $M$ experts, the routing probabilities scale as $\phi^i\sim O(\frac1{\kappa N})$. It is mathematically tidier to formulate the DMFT using variables which remain $O(1)$ in the limit. We therefore define and track the rescaled routing variables:
\begin{equation}
    \tilde\phi^i_\mu := \kappa N \phi^i_\mu
\end{equation}

\subsection{Gradient Definitions}
We mathematically \textit{define} the pre-activation gradients, ensuring they contain the correct scaling factors to remain finite in the $N \to \infty$ limit:

\begin{equation}
\begin{aligned}
g^1_\mu 
&:= \sqrt N \frac{\partial h^4_\mu}{\partial h^1_\mu}
&\qquad
g^{2,\text{in}}_{\mu,i} 
&:= \kappa N^\frac32 \frac{\partial h^4_\mu}{\partial h^{2,\text{in}}_{\mu,i}} \\
g^{2,\text{out}}_{\mu,i} 
&:= \kappa N^{\frac32}\frac{\partial h^4_\mu}{\partial h^{2,\text{o}}_{\mu,i}}
&\qquad
g^3_\mu 
&:= \sqrt N \frac{\partial h^4_\mu}{\partial h^3_\mu} = w^4 = z^3 \\
g^\phi_\mu 
&:= \frac{1}{\sqrt N}\frac{\partial h^4_\mu}{\partial \phi_\mu}
&\qquad
z^{2,\text{out}}_{\mu,i} 
&:= \kappa N \phi^i_\mu g^3_\mu = \tilde\phi^i_\mu g^3_\mu \\
z^{2,\text{in}}_{\mu,i} 
&:= \frac{1}{\sqrt{N}} \, W^{2,\text{out}\,\top}_i g^{2,\text{out}}_{\mu,i}
&\qquad
\tilde z^\phi_\mu 
&:= \frac{1}{\kappa \sqrt N} Q^\top g^{1,\phi} \\
\tilde z^1_{\mu,i} 
&:= \frac{1}{\kappa \sqrt N} W^{2,\text{in}\,\top}_i g^{2,\text{in}}_{\mu,i}
&\qquad
z^1 
&:= \frac{1}{\kappa \sqrt N} Q^\top g^{1,\phi}
   + \frac{1}{\kappa \sqrt N} \sum_{i=1}^M W^{2,\text{in}\,\top}_i g^{2,\text{in}}_{\mu,i} \\
&&&
= \frac{1}{\kappa \sqrt N} \sum_{i=1}^M \tilde z^1_{\mu,i} + \tilde z^{1,\phi}_\mu
\end{aligned}
\end{equation}

Note that the components of the router gradient $g^\phi_\mu$ naturally scale as $O(1)$:
\begin{equation}
    (g^\phi_\mu)_i =\frac1{\sqrt N}\frac{\partial h_\mu^{4}}{\partial\phi_\mu^i} = \frac1{\sqrt N}\frac{\partial h_\mu^{4}}{\partial \bm h^3_\mu}\cdot\frac{\partial \bm h_\mu^{3}}{\partial\phi_\mu^i} = \frac{1}{ N}\bm g^3\cdot\bm h_{\mu,i}^{2,\o} = O(1).
\end{equation}

The backward pass through the router dictates that the gradient at layer $1$ involves the quantity $g^{1,\phi}$, defined component-wise to handle the Jacobian of the softmax:
\begin{equation}
    g^{1,\phi}_k := \kappa N \sum_{i=1}^{\kappa N}g_i^\phi\phi^i(\delta_{ik}-\phi^k)
    = \sum_{i=1}^{\kappa N}g_i^\phi\tilde\phi^i(\delta_{ik}-\frac{\tilde\phi^k}{\kappa N})
\end{equation}

\subsection{DMFT Kernels and Order Parameters}
We define the following macroscopic \textit{kernels}, all of which are $O_N(1)$ and will serve as the fundamental order parameters in the dynamics:

\begin{equation}
\label{inf_exp_kernels_r3}
\begin{aligned}
\begin{aligned}
\Phi^{0}_{\mu\nu}
&= \frac{1}{D}\, x_\mu \cdot x_\nu
&\qquad
\Phi^{1}_{\mu\nu}(s,t)
&= \frac{1}{N}\, \sigma\!\big(h^{1}_{\mu}(s)\big)\cdot \sigma\!\big(h^{1}_{\nu}(t)\big) \\
\Phi^{2,\text{in}}_{\mu\nu,i}(s,t)
&= \frac{1}{\iota N}\, \sigma\!\big(h^{2,\text{in}}_{\mu,i}(s)\big)\cdot \sigma\!\big(h^{2,\text{in}}_{\nu,i}(t)\big)
&\qquad
\Phi^{2,\text{o}}_{\mu\nu,i}(s,t)
&= \frac{1}{N}\, \sigma\!\big(h^{2,\text{out}}_{\mu,i}(s)\big)\cdot \sigma\!\big(h^{2,\text{out}}_{\nu,i}(t)\big) \\
\Phi^{3}_{\mu\nu}(s,t)
&= \frac{1}{N}\, h^{3}_{\mu}(s)\cdot h^{3}_{\nu}(t)
&\qquad
G^{1}_{\mu\nu}(s,t)
&= \frac{1}{N}\, g^{1}_{\mu}(s)\cdot g^{1}_{\nu}(t) \\
G^{2,\text{in}}_{\mu\nu,i}(s,t)
&= \frac{1}{\iota N}\, g^{2,\text{in}}_{\mu,i}(s)\cdot g^{2,\text{in}}_{\nu,i}(t)
&\qquad
G^{2,\text{out}}_{\mu\nu,i}(s,t)
&= \frac{1}{N}\, g^{2,\text{out}}_{\mu,i}(s)\cdot g^{2,\text{out}}_{\nu,i}(t) \\
G^{1,\phi}_{\mu\nu}(s,t)
&= \frac{1}{\kappa N}\, g^{1,\phi}_{\mu}(s)\cdot g^{1,\phi}_{\nu}(t)
&\qquad
G^{3}_{\mu\nu}(s,t)
&= \frac{1}{N}\, g^{3}_{\mu}(s)\cdot g^{3}_{\nu}(t)
\end{aligned}
\end{aligned}
\end{equation}

With the exception of the input data Gram matrix $\Phi^0$, all of these kernels are dynamically evolving macroscopic variables.

\subsection{Learning Dynamics and the Neural Tangent Kernel}

We train the network using continuous-time gradient flow. The variable learning rate scheme defined previously is necessary so that the activation updates within the experts and for the router scores do not vanish or explode as $N, M \to \infty$. 

Taking a standard empirical risk minimization loss of the form:
\begin{equation}
    \label{deep_loss_r3}
    \mathcal L=\frac 1 P \sum_{\mu=1}^P\ell(f_\mu,y_\mu)
\end{equation}

The network parameters evolve according to:
\begin{equation}
    \label{deep_learning_dynamics_r3}
    \begin{aligned}
    &\frac{d\theta}{dt}
= \frac{\eta}{P\gamma}
  \sum_{\mu} \Delta_{\mu}
  \frac{\partial h_{\mu}^{4}}{\partial \theta}\\
  &\Delta_\mu=-\frac{\partial\mathcal L}{\partial f_\mu}
    \end{aligned}
\end{equation}

For a Mean Squared Error (MSE) loss where $\mathcal L=\frac{1}{P}\sum_{\nu}(y_\nu-f_\nu)^2$, the error signal (or residual) simplifies to $\Delta_\nu=2(y_\nu-f_\nu)$.

The logits update dynamically via the chain rule, driven by the Neural Tangent Kernel (NTK):
\begin{equation}
\label{logit_updates_r3}
\frac{d f_{\mu}(t)}{dt}
= \frac{\partial f_{\mu}(t)}{\partial \theta}\cdot \frac{d\theta}{dt}
= \frac{\partial f_{\mu}(t)}{\partial \theta}\cdot \frac{\eta_\theta}{P}\sum_\alpha^P\Delta_\alpha\frac{\partial f_\alpha(t)}{\partial \theta}
= \frac{\eta}{P}
  \sum_{\alpha} \Delta_{\alpha} \,
  K^{\mathrm{NTK}}_{\mu \alpha}(t,t)
\end{equation}

Where the infinite-width NTK is defined as:
\begin{equation}
\label{ntk_kernel_r3}
K^{\mathrm{NTK}}_{\mu \alpha}(t,s)
\equiv
\frac{\partial f_{\mu}(t)}{\partial \theta}
\cdot
\frac{\partial f_{\alpha}(s)}{\partial \theta}
\end{equation}

\subsection{Decomposition of the MoE NTK}
To explicitly compute the NTK, we require that the magnitude $\eta_\theta \frac{\partial f_{\mu}(t)}{\partial \theta_i} \cdot \frac{\partial f_{\alpha}(s)}{\partial \theta_i}$ remains order 1 for each parameter block $\theta_i$. We evaluate this constraint block by block.

For the expert input layer $W^{2,\i}$:
\begin{equation}
    \begin{aligned}
        \eta_0 &\sum_{i=1}^{\kappa N}\sum_{l,m,j=1}^{N_e}\sum_{k=1}^N
        \frac{\partial h^4}{\partial \left[{h^{2,\text{in}}_i}\right]_l}
        \frac{\partial \left[{h^{2,\text{in}}_i}\right]_l}{\partial W^{2}_{jk,i}}
        \frac{\partial h^4}{\partial \left[{h^{2,\text{in}}_i}\right]_m}
        \frac{\partial \left[{h^{2,\text{in}}_i}\right]_m}{\partial W^{2}_{jk,i}}=\\
        &=
        \eta_0 N \sum_{i=1}^{\kappa N}\sum_{l,m,j=1}^{N_e}\sum_{k=1}^N
        \frac{\left[g^{2,\i}_i\right]_l}{\kappa N^{3/2}}\frac{\left[g^{2,\i}_i\right]_m}{\kappa N^{3/2}}\frac{\sigma(h^1_k)}{\sqrt N}\frac{\sigma(h^1_k)}{\sqrt N}\delta_{mj}\delta_{lj}\\
        &=\eta_0 \frac{\iota}\kappa\frac{1}{\kappa N}\sum_{i=1}^{\kappa N}\left[\frac1{N_e}\sum_{j=1}^{N_e} \left[g^{2,\i}_i\right]_j\left[g^{2,\i}_i\right]_j\right]\left[\frac1N\sum_{k=1}^N\sigma(h^1_k)^2\right]\\
        &=\eta_0 \frac{\iota}\kappa\frac{1}{\kappa N}\sum_{i=1}^{\kappa N}G^{2,\i}_i\Phi^1 = \eta_0\frac{\iota}\kappa\bar G^2\Phi^1
    \end{aligned}
\end{equation}

For the expert output layer $W^{2,\o}$:
\begin{equation}
    \begin{aligned}
        \eta_0\sum_{i=1}^{\kappa N}\sum_{l,m,k=1}^{N_e}\sum_{j=1}^N&
        \frac{\partial h^4}{\partial \left[h^{2,\o}_i\right]_l}
        \frac{\partial \left[h^{2,\o}_i\right]_l}{\partial W^{2,\o}_{jk}}
        \frac{\partial h^4}{\partial \left[h^{2,\o}_i\right]_m}
        \frac{\partial \left[h^{2,\o}_i\right]_m}{\partial W^{2,\o}_{jk}}\\
        &=
        \eta_0  \sum_{i=1}^{\kappa N}\sum_{l,m,k=1}^{N_e}\sum_{j=1}^N
        \frac{\left[g^{2,\o}_i\right]_l}{\kappa N^{3/2}}\frac{\left[g^{2,\o}_i\right]_m}{\kappa N^{3/2}}\frac{\sqrt{\kappa N}}{\sqrt{N_e}}\sigma(\left[{h^{2,\text{in}}_i}\right]_k)\frac{\sqrt{\kappa N}}{\sqrt{N_e}}\sigma(\left[{h^{2,\text{in}}_i}\right]_k)\delta_{mj}\delta_{lj}\\
        &=\eta_0 \frac{1}{\kappa N}\sum_{i=1}^{\kappa N}G^{2,\o}_i\Phi^{2,\i}_i        
    \end{aligned}
\end{equation}

For the shared base layer $W^1$:
\begin{equation}
\begin{aligned}
\eta_0
\sum_{k=1}^{D}
\sum_{j,m=1}^{N}
\frac{\partial h^{4}}{\partial h^{1}_{n}}
\frac{\partial h^{1}_{n}}{\partial W^{1}_{jk}}
\frac{\partial h^{4}}{\partial h^{1}_{m}}
\frac{\partial h^{1}_m}{\partial W^{1}_{jk}}
&=
\eta_0
\sum_{k=1}^{D}
\sum_{j,m,n=1}^{N}
\frac{g^{1}_{n}}{\sqrt{N}}
\frac{g^{1}_{m}}{\sqrt{N}}
\,\delta_{nj}\,\delta_{mj}\,
\frac{x_k}{\sqrt{D}}
\frac{x_k}{\sqrt{D}}
=
\eta_0\,G^{1}\,\Phi^{0}
\end{aligned}
\end{equation}

The router weight matrix $Q$ involves the complex Jacobian of the softmax operator. Integrating this out, we find:
\begin{equation}
\begin{aligned}
\eta_0\kappa
\sum_{j=1}^{\kappa N}
\sum_{k=1}^{N}
\frac{\partial h^{4}}{\partial Q_{jk}}
\frac{\partial h^{4}}{\partial Q_{jk}}
&=
\eta_0\kappa N
\sum_{j=1}^{\kappa N}
\sum_{i,i'=1}^{\kappa N}
\Phi^{1}
\,g^{\phi}_{i}\,g^{\phi}_{i'}
\,\phi^{i}\,\phi^{i'}
(\delta_{ij}-\phi^{j})
(\delta_{i'j}-\phi^{j})
=
{\eta_0}\Phi^1G^{1,\phi}
\end{aligned}
\end{equation}

Aggregating these contributions yields the full macroscopic evolution equation:
\begin{equation}
\begin{aligned}
    \frac{d f_{\mu}(t)}{dt}
&= \frac{\eta_0}{P}\sum_\nu\Delta_\nu
\Bigg[
G^{1}_{\mu\nu}(t,t)\,\Phi^{0}_{\mu\nu}(t,t)
+
\frac{\iota}\kappa \Big[G_{\mu\nu,i}^{2,\i} + G^{1,\phi}_{\mu\nu}(t,t)\Big](t,t)\Phi^1_{\mu\nu}(t,t)\\
&
+
\frac{\iota}\kappa G^3_{\mu\nu}(t,t)\bar\Phi^{2,\i}_{\mu\nu}(t,t)+
\Phi^{3}_{\mu\nu}(t,t)
\Bigg]
\end{aligned}
\end{equation}

\subsection{Evolution of weights, preactivations, and pregradients}
The updates to the router weights are:
\begin{equation}
\label{appendixDdQdt_r3}
\begin{aligned}
\frac{d Q_{jk}}{dt}
&=\frac{\eta}{P}\sum_\mu\Delta_\mu \frac{\partial f_\mu}{\partial Q_{jk}}
=\frac{\eta_0\gamma\kappa\sqrt N}{P}\sum_\mu\Delta_\mu \frac{\partial h^4_\mu}{\partial Q_{jk}}
= \frac{\eta_0\gamma_0\kappa N}{P}\sum_\mu\Delta_\mu \frac{\partial h^4_\mu}{\partial Q_{jk}} \\
&=
\frac{\eta_0 \gamma_0 }{P \sqrt N}
\sum_{\mu} \Delta_\mu
\left[\kappa N
\sum_{a=1}^{M}
g_a^{\phi}\,
\phi_a
(\delta_{aj}-\phi_j)\right]
\, \sigma(h_k^{1}) \\[6pt]
&= 
\frac{\eta_0 \gamma_0 }{P \sqrt N}
\sum_{\mu} \Delta_\mu
g_j^{1,\phi}
\sigma(h_k^{1}) \\[6pt]
\end{aligned}
\end{equation}
We obtain thus:
\begin{equation}
\begin{aligned}
\psi_\mu(t)
&= \frac{1}{\sqrt N}\, Q(t)\, \sigma\!\big(h_{\mu}^{1}(t)\big) \\
&= \chi^{\phi}_\mu(t)
+ \frac{\eta_0 \gamma_0}{P}\,
\int_{0}^{t} ds
\sum_{\nu} \Delta_{\nu}(s)\,
\Phi^{1}_{\mu\nu}(s,t)
g^{1,\phi}_\nu(t) \\
\tilde z^{1,\phi}_\mu &= \tilde\xi^{1,\phi}_\mu +\frac{\eta_0\gamma_0}{P}\int_0^\infty dt \sum_\nu \Delta_\nu G^{1,\phi}_{\nu\mu}\,\sigma(h^1_{\nu j})
\end{aligned}
\end{equation}
The other preactivation and pregradient updates follow similarly:

\begin{equation}
\begin{aligned}
\frac{d W^{1}(t)}{dt}
&=
\frac{\eta_0 \gamma_0 \sqrt{N}}{P}
\sum_{\mu}
\Delta_{\mu}(t)\,
\frac{g_{\mu}^{1}(t)}{\sqrt{N}}\,
\frac{1}{\sqrt{D}}\,
x_{\mu}
\\[3pt]
\Rightarrow\quad
h^{1}_{\mu}(t)
&=
\frac{1}{\sqrt{D}}\,W^{1}(0)\,x_{\mu}
+
\int_{0}^{t} ds\,
\frac{\eta_0 \gamma_0}{P}
\sum_{\nu}
\Delta_{\nu}(s)\,
g_{\nu}^{1}(s)\,
\Phi^{0}_{\mu\nu}(t,s)
\\[6pt]
\frac{d W^{2,\i}_i(t)}{dt}
&=
\frac{\eta_0 \gamma_0 \sqrt{N} N }{P}
\sum_{\mu}
\Delta_{\mu}(t)\,
\frac{g_{\mu,i}^{2,\i}(t)}{\kappa \sqrt{N}}\,
\frac{1}{N^{3/2}}\,
\sigma\!\big(h^{1}_{\mu}(t)\big)
\\[3pt]
\Rightarrow\quad
h^{2,\text{in}}_{\mu, i}(t)
&=
\frac1{\sqrt N} W^{2,\i}_i(0)\,\sigma(h^1_\mu(t))
+
\frac{\eta_0 \gamma_0}{\kappa P}
\int_{0}^{t} ds\,
\sum_{\nu}
\Delta_{\nu}(s)\,
g_{\nu_i}^{2,\i}(s)\,
\Phi^{1}_{\mu\nu}(t,s)
\\[6pt]
\frac{d W^{2,\o}_i(t)}{dt}
&=
\frac{\eta_0 \gamma_0 \sqrt{N} N }{P}
\sum_{\mu}
\Delta_{\mu}(t)\,
\frac{g_{\mu,i}^{2,\o}(t)}{N^{3/2}\kappa}\,
\frac{1}{\sqrt{N}}\,
\sigma\!\big(h^{2,\text{in}}_{\mu, i}(t)\big)
\\
\Rightarrow\quad
h^{2,\o}_{\mu,i}(t)
&=
\frac{1}{\sqrt N}\,
W^{2,\o}_i(0)\,
\sigma\!\big(h^{2,\text{in}}_{\mu, i}(t)\big)
+
\frac{\eta_0 \gamma_0 \iota}{P\kappa}
\int_{0}^{t} ds\,
\sum_{\nu}
\Delta_{\nu}(s)\,
g_{\nu,i}^{2,\o}(s)\,
\Phi^{2,\i}_{\mu\nu,i}(t,s)
\\[6pt]
\frac{d w^{4}(t)}{dt}
&=
\frac{\eta_0 \gamma_0 \sqrt{N}}{P}
\sum_{\mu}
\Delta_{\mu}(t)\,
\frac{1}{\sqrt{N}}\,
h^{3}_{\mu}(t)
=
\frac{\eta_0 \gamma_0}{P}
\sum_{\mu}
\Delta_{\mu}(t)\,
h^{3}_{\mu}(t)
\end{aligned}
\end{equation}

Using the same expressions for the evolution of the weights to derive expressions for the pregradients:

\begin{equation}
\begin{aligned}
z^{3}_\mu(t)
&=
g^{3}_\mu(t)
=
w^{4}_\mu(t)
=
\xi^{3}_\mu(0)
+
\frac{\eta_0 \gamma_0}{P}
\int_{0}^{t} ds
\sum_{\nu}
\Delta_{\nu}(s)\,
h^{3}_{\nu}(s)
\\[6pt]
z^{2,\o}_{\mu,i}(t)
&= \kappa g^{2,\o}_{\mu,i}(t)
=
\tilde\phi^{i}_{\mu}(t)\,g^{3}_{\nu}(t)
=
\frac{\eta_0 \gamma_0 }{P}
\tilde\phi^i_\mu(t)
\int_{0}^{t} ds
\sum_{\nu}
\Delta_{\nu}(s)\,
h^{3}_{\nu}(s)
\\[6pt]
z^{2,\i}_{\mu,i}(t)
&=
\frac{1}{\sqrt N}\,W^{2,\o\,\top}_i(t)\,g^{2,\o}_{\mu,i}(t)
\\
&=
\xi^{2,\i}_{\mu,i}(t)
+
\frac{\eta_0 \gamma_0}{P \kappa}
\int_{0}^{t} ds
\sum_{\nu}
\Delta_{\nu}(s)\,
G^{2,\o}_{\nu\mu,i}(s,t)\,
\sigma\!\big(h^{2,\i}_{\nu,i}(s)\big)
\\[6pt]
\tilde z^{1}_{\mu,i}(t)
&=
W^{2,\i\,\top}_i(t)\,g^{2,\i}_i(t)
\\
&=
\tilde\xi^{1}_{\mu,i}(t)
+
\frac{\eta_0 \gamma_0 \iota}{P\kappa}
\int_{0}^{t} ds
\sum_{\nu}
\Delta_{\nu}(s)\, G^{2,\i}_{\nu\mu,i}(s,t)\,
\sigma\!\big(h^{1}_{\nu}(s)\big)
\end{aligned}
\end{equation}
Finally for the router gradient:
\begin{equation}
    \begin{aligned}
        g^{1,\phi}_i
        &=
        \frac1{\kappa N}
        \sum_{m=1}^{\kappa N}g_m^\phi\tilde\phi^m(\kappa N \delta_{mi}-\tilde\phi^i)\\
        &=
        \xi^{g^{1,\phi}}_{i}
        +
        \frac{\eta_0\gamma_0 \iota}{P\kappa}\int_0^tds \sum_\nu \Delta_\nu G^3 \Phi^{2,in}_i \tilde\phi^i \tilde\phi^i
        -
        \frac{\eta_0\gamma_0 \iota}{P\kappa}\int_0^tds \sum_\nu \Delta_\nu G^3 \bar \Phi^{2,in} \tilde\phi^i
        .
    \end{aligned}
\end{equation}

\subsection{Stochastic Initial Fields}
To isolate the purely deterministic part of the trajectory from the random initialization, we define a set of initial stochastic fields $\mathcal{F}$. These fields encapsulate the randomness of the initial weights:

\begin{equation}
\mathcal{F} =\{\chi_\mu^1(t),\chi_{\mu,i}^{2,\i}(t),\chi_{\mu,i}^{2,\o}(t),\chi^\phi_\mu(t),\tilde\xi_{\mu,i}^{1}(t),\xi_{\mu,i}^{2,\i}(t),\xi_\mu^{3}(t),\xi^\phi_\mu(t),\xi^{g^\phi}_{\mu i}(t) \}_{i \in\{1,...,M\}, \mu\in\{1,...,P\}}
\end{equation}

\begin{equation}
\begin{aligned}
\chi^{1}_{\mu} &= \frac{1}{\sqrt{D}}\, W^{1}(0)\, x_{\mu} \\
\chi^{2,\i}_{\mu,i}(t) &= \frac{1}{\sqrt N}\, W^{2,\i}_i(0)\, \sigma\!\big(h^{1}_{\mu}(t)\big) \\
\chi^{2,\o}_{\mu,i}(t) &= \frac{1}{\sqrt N}W^{2,\o}_i(0)\, \sigma\!\big(h^{2,\text{in}}_{\mu, i}(t)\big) \\
\chi^{\phi}_{\mu}(t) &= \frac{1}{\sqrt N}\, Q(0)\, \sigma(h^1_{\mu}(t)) \\
\tilde\xi^{1}_{\mu,i}(t) &= \frac{1}{\sqrt N}\big(W^{2,\i}_i(0)\big)^{\top}\, g^{2,\i}_{\mu,i}(t) \\
\xi^{2,\i}_{\mu,i}(t) &= \frac{1}{\sqrt N}\, \big(W^{2,\o}_i(0)\big)^{\top}\, g^{2,\o}_{\mu,i}(t) \\
{\xi^{1,\phi}_\mu(t)} &= \frac{1}{\kappa \sqrt N} Q(0)^\top {g^{1,\phi}_\mu(t)} \\
\xi^{g^\phi}_{\mu i}(t) &= \frac{1}{N^{3/2}} g^{3\top}_\mu(t)\,W^{2,\o}_i(0)\,\sigma({h^{2,\text{in}}_{\mu, i}}(t))
\end{aligned}
\end{equation}
A critical step in rendering the MoE partition function computationally tractable is distinguishing between expert-local fields (which are specific to an expert $i$) and global fields (which impact the shared base layer or the final aggregated output). This distinction allows the massive partition function to factorize over the experts.

We define the router-averaged stochastic fields to capture the macroscopic effect of the local processes:
\begin{equation}
    \bar\chi^3_\mu(t) = \frac{1}{\kappa N}\sum_{i=1}^{\kappa N}\tilde\phi^i_\mu(t)\chi^{2,\o}_{\mu,i}(t)
\end{equation}
\begin{equation}
    \bar\xi^1_\mu(t) = \frac1{\kappa\sqrt N}\sum_{i=1}^{\kappa N}\tilde\xi^{1}_{\mu,i}(t)
\end{equation}

Substituting these, the global aggregated output $h^3_\mu$ and the gradient arriving at the base layer $z^1_\mu$ can be expressed entirely in terms of order parameters that are smooth over the ensemble of experts:
\begin{equation}
    h^3_\mu(t) = \bar\chi^3_\mu(t) + \frac{\eta_0\gamma_0\iota}{\kappa P}\int_0^t ds \sum_\nu \Delta_\nu(s) \bar\Phi^{2,\i}_{\mu\nu}(s,t) z^3_\nu(s)
\end{equation}

\begin{equation}
z^1_\mu(t) = \bar\xi^{1}_\mu(t) + \xi^{1,\phi}_\mu(t) + \frac{\eta_0 \gamma_0 \iota}{\kappa P}\int_0^t \! ds\sum_{\nu}\Delta_{\nu}(s) \left[\frac1{\sqrt N} \bar G_{\mu \nu}^{2,\i}(s,t) + G^{1,\phi}_{\mu\nu}(s,t) \right] \sigma(h^{1}_{\nu}(s))
\end{equation}

\subsection{Deriving the DMFT Action}
We formulate the DMFT by writing the moment-generating function for the system's trajectories and performing a disorder average over the initial weights $\theta_0$.
\begin{equation}
\adjustbox{max width=\textwidth}{$
\begin{aligned}
&Z\propto\Bigg\langle
\int d\mathcal{F}
\exp\Bigg(
\sum_{\mu}
    \int_{0}^{\infty} \! dt \,i\,\Bigg[
\hat{\chi}_\mu^{1} \!\cdot\! 
\big( \chi_\mu^{1} - \frac{1}{\sqrt{D}} W^{1}(0) x_\mu \big)
+ \hat{\chi}_{\mu}^{2,\i} \!\cdot\! 
\Big( \chi_{\mu}^{2,\i} - 
\frac1{\sqrt N}W^{2,\i}(0) \sigma(h_\mu^{1}(t)) \Big) \\
&
+ \hat{\bar\chi}_\mu^{3} \!\cdot\! 
\Big( \bar\chi_\mu^{3} - \frac{1}{\kappa N}\sum_{i=1}^{\kappa N}\tilde\phi^i_\mu(t)\frac1{\sqrt {N}}W^{2,\o}(0) 
\sigma({h^{2,\text{in}}_{\mu, i}}(t)) \Big)
+ \hat{\bar\xi}_\mu^{1} \!\cdot\! 
\Big( \bar\xi_\mu^{1} - \frac{1}{\kappa N}\sum_{i=1}^{\kappa N} 
\frac1{\sqrt N}W^{2,\i}(0)^{\!\top} g_{\mu,i}^{2,\i}(t) \Big) \\
&
+ \frac{1}{\kappa N}\sum_{i=1}^{\kappa N}\hat{\xi}_{\mu,i}^{2,\i} \!\cdot\! 
\Big( \xi_{\mu,i}^{2,\i} - \frac{1}{\sqrt N} 
W^{2,\o}(0)^{\!\top} g_{\mu,i}^{2,\o}(t) \Big)
+ \hat{\xi}_\mu^{3} \!\cdot\! 
\Big(\xi_\mu^{3}-w^{4}(0)^{\!\top}\Big)
\\
&
+\hat\chi^\phi_\mu\cdot\bigg(\chi^\phi_\mu-\frac{1}{\sqrt N}Q(0)\sigma(h^1_\mu(t))\bigg)
+
\hat\xi^{1,\phi}_\mu\cdot\bigg(\xi^{1,\phi}_\mu-\frac{1}{\kappa \sqrt N}Q(0)^\top g^{1,\phi}_\mu(t)
\bigg)\\
&
+ \sum_{m=1}^{\kappa N}
\hat{\xi}^{g^{1,\phi}}_{\mu, m}(t)
\Big(
\xi^{g^{1,\phi}}_{\mu, m}(t)
-
\frac{1}{\kappa N}
\sum_{i=1}^{\kappa N}
\frac{1}{N^{3/2}}
\,
g^{3\,T}_\mu(t)
W^{2,\text{out}}
\sigma\!\big(h^{2,\text{in}}_{\mu,i}(t)\big)
\tilde{\phi}_\mu^{\,i}(t)
\left(
\kappa N\,\delta_{im}
-
\tilde{\phi}^{\,m}_\mu(t)
\right)
\Big)
\Bigg]\Bigg)\\
&\times\exp\!\Bigg(
    \sum_{\mu}
    \int_{0}^{\infty} \! dt \,
    \Bigg[
    j_\mu^1(t)\cdot\chi_\mu^1(t)+j_{\mu}^{2,\i}(t)\cdot \chi_{\mu}^{2,\i}(t)
    + \bar j_{\mu}^{3}(t)\cdot \bar\chi_{\mu}^{3}(t)
    + \frac{1}{\kappa N}\sum_{i=1}^Mv_{\mu,i}^{2,\i}(t)\cdot\xi_{\mu,i}^{2,\i}(t)\\&
    +\bar v_{\mu}^{1}(t)\cdot\bar\xi_{\mu}^{1}(t)+v_\mu^{3}\cdot\xi^{3}_\mu(t)
    +j^\phi_\mu(t)\cdot\chi^\phi_\mu(t) + 
    v_\mu^{1,\phi}(t)\cdot\xi_\mu^{1,\phi}(t)
    + 
    v_\mu^{g^{1,\phi}}(t)\cdot\xi_\mu^{g^{1,\phi}}(t)
    \Bigg]
  \Bigg)\Bigg\rangle_{\theta_0}.
\end{aligned}
$}
\end{equation}

Given the shared initialisation across experts we now have terms coupling different experts (e.g. $\sigma(h_\mu^{2,\i})_i\,\sigma(h_\nu^{2,\i})_j\,\tilde{\phi}_\mu^i\tilde{\phi}_\nu^j$). In order to obtain a stable set of order parameters that scale coherently with $N$ we define the following set of "averaged fields":
\begin{equation}
\begin{aligned}
     1 & =\int \frac{d\hat{\bar{g}}^{2,\i}_{\mu}(t)}{2\pi}\; \exp\left[\,\hat{\bar{g}}^{2,\i}_{\mu}(t)\left(\kappa N\bar{g}_{\mu}^{2,\i}(t) - \sum_{m=1}^{\kappa N} g_{\mu,m}^{2,\mathrm{in}}(t)\right)\right] \\
    1 &= \int \frac{d\hat{\bar{\sigma}}{(h_{\mu}}^{2,\i}(t))}{2\pi}\; \exp\left[\,\hat{\bar{\sigma}}{(h_{\mu}}^{2,\i}(t))\left(\kappa N\bar{\sigma}{(h_{\mu}}^{2,\i}(t)) - \sum_{i=1}^{\kappa N} \sigma(h_{\mu,i}^{2,\mathrm{in}}(t))\tilde{\phi}^i(t)\right)\right] \\
    1 &= \int \frac{d\hat{\bar{\sigma}}^\phi{(h_{\mu}}^{2,\i}(t))}{2\pi}\; \exp\left[\,\hat{\bar{\sigma}}^{\phi}{(h_{\mu}}^{2,\i}(t))\left(\kappa N\bar{\sigma}^\phi{(h_{\mu}}^{2,\i}(t)) - \sum_{i=1}^{\kappa N} \sigma(h_{\mu,i}^{2,\mathrm{in}}(t))\tilde{\phi}^i(t)\hat{\xi}^{g^{1,\phi}}_{i}(t)\right)\right] \\
    1 & = \int \frac{d\hat{\hat{\bar{\xi}}}_{\mu}^{2,\mathrm{in}}(t)}{2\pi}\; \exp\left[\,\hat{\hat{\bar{\xi}}}_{\mu}^{2,\mathrm{in}}(t)\left(\kappa N\hat{\bar{\xi}}_{\mu}^{2,\mathrm{in}}(t) - \sum_{i=1}^{\kappa N} \hat\xi_{\mu,i}^{2,\mathrm{in}}(t)\tilde \phi^i (t)\right)\right] 
\end{aligned}
\end{equation}
We must constrain the solutions to be physical, so for each $i,\mu,\nu,s,t$ we multiply in the following resolutions of the identity:
\begin{equation}
    1 = \int\frac{d \Phi^1_{\mu\nu}(s,t)d\hat\Phi^1_{\mu\nu}(s,t)}{2\pi i N^{-1}}\exp\left[\hat\Phi^1_{\mu\nu}(s,t)\Big(N\Phi^1_{\mu\nu}(s,t)-\sigma(h_\mu^1(s))\cdot \sigma(h^1_\nu(t))\Big)\right]
\end{equation}

\begin{equation}
    1 = \int\frac{d G^1_{\mu\nu}(s,t)d\hat G^1_{\mu\nu}(s,t)}{2\pi i N^{-1}}\exp\left[\hat G^1_{\mu\nu}(s,t)\Big(NG^1_{\mu\nu}(s,t)-g_\mu^1(s)\cdot g^1_\nu(t)\Big)\right]
\end{equation}

\begin{equation}
    1 = \int\frac{d G^3_{\mu\nu}(s,t)d\hat G^3_{\mu\nu}(s,t)}{2\pi i N^{-1}}\exp\left[\hat G^3_{\mu\nu}(s,t)\Big(NG^3_{\mu\nu}(s,t)-g_\mu^3(s)\cdot g^3_\nu(t)\Big)\right]
\end{equation}

\begin{equation}
    1 = \int\frac{d G^{1,\phi}_{\mu\nu}(s,t)d\hat G^{1,\phi}_{\mu\nu}(s,t)}{2\pi i {\kappa N}^{-1}}\exp\left[\hat G^{1,\phi}_{\mu\nu}(s,t)\Big(\kappa N G^{1,\phi}_{\mu\nu}(s,t)-g_\mu^{1,\phi}(s)\cdot g^{1,\phi}_\nu(t)\Big)\right]
\end{equation}
\begin{equation}
\label{Phi2i_r3}
    1 = \int\frac{d \Phi^{2,\i}_{\mu\nu,i}(s,t)d\hat\Phi^{2,\i}_{\mu\nu,i}(s,t)}{2\pi i (\iota N)^{-1}}\exp\left[\hat\Phi^{2,\i}_{\mu\nu,i}(s,t)\Big({\iota N}\Phi^{2,\i}_{\mu\nu,i}(s,t)-\sigma({h^{2,\text{in}}_{\mu, i}}(s))\cdot \sigma(h^{2,\text{in}}_{\nu,i}(t))\Big)\right]
\end{equation}
\begin{equation}
\label{G2i_r3}
    1 = \int\frac{dG^{2,\i}_{\mu\nu,i}(s,t)d\hat G^{2,\i}_{\mu\nu,i}(s,t)}{2\pi i (\iota N)^{-1}}\exp\left[\hat G^{2,\i}_{\mu\nu,i}(s,t)\Big(\iota N G^{2,\i}_{\mu\nu,i}(s,t)-g_{\mu,i}^{2}(s)\cdot g_{\nu,i}^{2}(t)\Big)\right]
\end{equation}
\begin{equation}
    1 = \int\frac{d \bar A^{1}_{\mu\nu}(s,t)d\bar B^{1}_{\mu\nu}(s,t)}{2\pi i N^{-1}}\exp\left[-\bar B^{1}_{\mu\nu}(s,t)\Big(N\bar A^{1}_{\mu\nu}(s,t)+i\hat{\bar\xi}_\mu^{1}(s)\cdot \sigma(h^1_\nu(t))\Big)\right]
\end{equation}

\begin{equation}
    1 = \int\frac{d A^{1,\phi}_{\mu\nu}(s,t)d B^{1,\phi}_{\mu\nu}(s,t)}{2\pi i N^{-1}}\exp\left[-B^{1,\phi}_{\mu\nu}(s,t)\Big(NA^{1,\phi}_{\mu\nu}(s,t)+i\hat\xi_\mu^{1,\phi}(s)\cdot \sigma(h^1_\nu(t))\Big)\right]
\end{equation}

\begin{equation}
    1 = \int\frac{dA^{g^{1,\phi}}_{\mu\nu}(s,t)d B^{g^{1,\phi}}_{\mu\nu}(s,t)}{2\pi i (\kappa N)^{-1}}\exp\left[-B^{g^{1,\phi}}_{\mu\nu}(s,t)\Big(\kappa N A^{g^{1,\phi}}_{\mu\nu}(s,t)+i\hat\xi_\mu^{g^{1,\phi}}(s)\cdot \tilde\phi_\nu(t)\Big)\right]
\end{equation}

\begin{equation}
1=\int\frac{d\Phi_{\mu\nu}^{3}(t,s)\,d\hat{\Phi}_{\mu\nu}^{3}(t,s)}{2\pi i\, N^{-1}}\exp\!\left[\hat{\Phi}_{\mu\nu}^{3}(t,s)\left(N\Phi_{\mu\nu}^{3}(t,s)-h_{\mu}^{3}(t)\cdot h_{\nu}^{3}(s)\right)\right]
\end{equation}

\begin{equation}
1=\int\frac{d\bar\Phi_{\mu\nu}^{2,\i}(t,s)\,d\hat{\bar\Phi}_{\mu\nu}^{2,\i}(t,s)}{2\pi i\, }\exp\!\left[\hat{\bar\Phi}_{\mu\nu}^{2,\i}(t,s)\left(\kappa N\bar\Phi_{\mu\nu}^{2,\i}(t,s)-\sum_{i=1}^{\kappa N}\tilde\phi^i_\mu(s)\Phi^{2,\i}_{\mu\nu,i}(s,t)\right)\right]
\end{equation}

\begin{equation}
1=\int\frac{d\bar G_{\mu\nu}^{2,\i}(t,s)\,d\hat{\bar G}_{\mu\nu}^{2,\i}(t,s)}{2\pi i\, (\kappa N)^{-1}}\exp\!\left[\hat{\bar G}_{\mu\nu}^{2,\i}(t,s)\left(\kappa N\bar G_{\mu\nu}^{2,\i}(t,s)-\sum_{i=1}^{\kappa N}G^{2,\i}_{\mu\nu,i}(s,t)\right)\right]
\end{equation}

\begin{equation}
    1 = \int \frac{d\tilde{G}_{\mu\nu}^{2,\i}(s,t)\, d\hat{\tilde{G}}^{2,\i}_{\mu\nu}(s,t)}{2\pi i\, (\iota N)^{-1}} \exp\left[\hat{\tilde{G}}^{2,\i}_{\mu\nu}(s,t)\left(\iota N\, \tilde{G}^{2,\i}_{\mu\nu}(s,t) - \bar{g}^{2,\i}_\mu(s) \cdot \bar{g}^{2,\i}_\nu(t)\right)\right]
\end{equation}
\begin{equation}
    1 = \int \frac{d\tilde{\Phi}^{2,\i}_{\mu\nu}(s,t)\, d\hat{\tilde{\Phi}}^{2,\i}_{\mu\nu}(s,t)}{2\pi i\, (\iota N)^{-1}} \exp\left[\hat{\tilde{\Phi}}^{2,\i}_{\mu\nu}(s,t)\left(\iota N\, \tilde{\Phi}^{2,\i}_{\mu\nu}(s,t) - \bar{\sigma}^{2,\i}_\mu(s) \cdot \bar{\sigma}^{2,\i}_\nu(t)\right)\right]
\end{equation}
\begin{equation}
    1 = \int \frac{d\tilde{\Phi}^{2,\i,\phi}_{\mu\nu}(s,t)\, d\hat{\tilde{\Phi}}^{2,\i,\phi}_{\mu\nu}(s,t)}{2\pi i\, (\iota N)^{-1}} \exp\left[\hat{\tilde{\Phi}}^{2,\i,\phi}_{\mu\nu}(s,t)\left(\iota N\, \tilde{\Phi}^{2,\i,\phi}_{\mu\nu}(s,t) - \bar{\sigma}^{2,\i,\phi}_\mu(s) \cdot \bar{\sigma}^{2,\i}_\nu(t)\right)\right]
\end{equation}
\begin{equation}
    1 = \int \frac{d\hat{\bar{\Xi}}^{2,\i}_{\mu\nu}(s,t)\, d\hat{\hat{\bar{\Xi}}}^{2,\i}_{\mu\nu}(s,t)}{2\pi i\, (\iota N)^{-1}} \exp\left[\hat{\hat{\bar{\Xi}}}^{2,\i}_{\mu\nu}(s,t)\left(\iota N\, \hat{\bar{\Xi}}^{2,\i}_{\mu\nu}(s,t) - \hat{\bar{\xi}}^{2,\i}_\mu(s) \cdot \hat{\bar{\xi}}^{2,\i}_\nu(t)\right)\right]
\end{equation}
\begin{equation}
    1 = \int \frac{d{\Phi}^{\phi\phi,2,\i}_{\mu\nu}(s,t)\, d\hat{{\Phi}}^{\phi\phi,2,\i}_{\mu\nu}(s,t)}{2\pi i\, (\iota N)^{-1}} \exp\left[\hat{{\Phi}}^{\phi\phi,2,\i}_{\mu\nu}(s,t)\left(\iota N\, {\Phi}^{\phi\phi,2,\i}_{\mu\nu}(s,t) - \bar{\sigma}^{\phi,2,\i}_\mu(s) \cdot \bar{\sigma}^{\phi,2,\i}_\nu(t)\right)\right]
\end{equation}

\begin{equation}
    1 = \int \frac{d{\tilde A}^{2,\i}_{\mu\nu}(s,t)\, d{{\tilde B}}^{2,\i}_{\mu\nu}(s,t)}{2\pi i\, (\iota N)^{-1}} \exp\left[-\tilde B^{2,\i}_{\mu\nu}(s,t)\left(\iota N\, {\tilde A}^{,2,\i}_{\mu\nu}(s,t) - \bar{\sigma}^{2,\i}_\mu(s) \cdot \bar{\xi}^{2,\i}_\nu(t)\right)\right]
\end{equation}

\begin{equation}
    1 = \int \frac{d{\tilde A}^{\phi,2,\i}_{\mu\nu}(s,t)\, d{{\tilde B}}^{\phi,2,\i}_{\mu\nu}(s,t)}{2\pi i\, (\iota N)^{-1}} \exp\left[{{-\tilde B}}^{2,\i}_{\mu\nu}(s,t)\left(\iota N\, {\tilde A}^{2,\i}_{\mu\nu}(s,t) - \bar{\sigma}^{\phi,2,\i}_\mu(s) \cdot \bar{\xi}^{2,\i}_\nu(t)\right)\right]
\end{equation}

\subsection{Softmax}
As for the previous regimes we enforce the softmax layer order parameter via the Fourier-transformed delta function

\begin{equation}
1=\int\frac{d \mathcal S_{\mu}(t)\,d\hat{ \mathcal S}_{\mu}(t)}{2\pi i\, (\kappa N)^{-1}}\exp\!\left[\hat{\mathcal S}_{\mu}(t)\left(\kappa N \mathcal S_{\mu}^{}(t)-\sum_{i=1}^{\kappa N} e^{\psi^i_\mu(t)}\right)\right]
\end{equation}

\subsection{Partition function}
Define the set of all kernels and conjugates (indexed for time and feature, although this is omitted for brevity in \ref{kernels_set_r3}):

\begin{equation}
\label{kernels_set_r3}
\adjustbox{max width=\textwidth}{$
\begin{aligned}
    \mathcal K = \{&\Phi^1,\hat\Phi^1,,\bar\Phi^{2,\i},\hat{\bar\Phi}^{2,\i},\Phi^3,\hat\Phi^3,G^1,
    \hat G^1,\bar G^{2,\i},\hat{\bar G}^{2,\i},\Phi^{2,\i}_{i},\hat{\Phi}^{2,\i}_{i},G^{2,\i}_{i},\hat{G}^{2,\i}_{i},
    \\&G^{3},\hat G^{3},G^{1,\phi},\hat G^{1,\phi},\bar A^{1},\bar B^{1},\bar B^{2,\i},A^{1,\phi},B^{1,\phi}
    ,A^{g^{1,\phi}},B^{g^{1,\phi}}
    ,B^3, A^3,\\
    &
    \tilde{\bar G}^{2,\i},\tilde G^{2,\i}
    \hat{\tilde{\Phi}}^{2,\i}, \tilde{\Phi}^{2,\i},\hat{\tilde{\Phi}}^{2,\i,\phi}, \tilde{\Phi}^{2,\i,\phi},{\tilde{B}}^{2,\i},
    \tilde{A}^{2,\i},{\tilde{B}}^{2,\i,\phi}, \tilde{A}^{2,\i,\phi},\hat{\hat{\bar{\Xi}}}^{2,\i},\hat{\bar{\Xi}}^{2,\i} , \hat{{\Phi}}^{\phi\phi,2,\i}, {\Phi}^{\phi\phi,2,\i}\}
\end{aligned}
$}
\end{equation}

We can partition these order parameters into the set $\mathcal K_{global}$ of global order parameters, the set $\mathcal K_{exp-loc}$ of expert-local ones, together with the shared fields $\mathcal F_j^{2,\mathrm{in},\mathrm{sh}}$, the router fields $\mathcal F_i^\phi$, the experts fields $\mathcal F_{i,j}^{2,\mathrm{in}}$,$\mathcal F_n^{\mathrm{global\text{-}site}}$:
\begin{equation}
\adjustbox{max width=\textwidth}{$
\begin{aligned}
\small
\mathcal K_{\mathrm{global}}=\{&
\Phi^1,\hat\Phi^1,\bar\Phi^{2,\mathrm{in}},\hat{\bar\Phi}^{2,\mathrm{in}},
\Phi^3,\hat\Phi^3,G^1,\hat G^1,\bar G^{2,\mathrm{in}},\hat{\bar G}^{2,\mathrm{in}},
G^{1,\phi},\hat G^{1,\phi},G^3,\hat G^3,\bar A^1,\bar B^1,\bar A^{2,\mathrm{in}},
A^{1,\phi}, \\
&B^{1,\phi},A^{g1,\phi},B^{g1,\phi},
\widetilde G^{2,\mathrm{in}},\hat{\widetilde G}^{2,\mathrm{in}},
\widetilde\Phi^{2,\mathrm{in}},\hat{\widetilde\Phi}^{2,\mathrm{in}},
\widetilde\Phi^{2,\mathrm{in},\phi},\hat{\widetilde\Phi}^{2,\mathrm{in},\phi},
\widetilde B^{2,\mathrm{in}}, \\
&\widetilde A^{2,\mathrm{in}},
\widetilde B^{2,\mathrm{in},\phi},\widetilde A^{2,\mathrm{in},\phi},
\hat{\bar\Xi}^{2,\mathrm{in}},\hat{\hat{\bar\Xi}}^{2,\mathrm{in}},
\Phi^{\phi\phi,2,\mathrm{in}},\hat\Phi^{\phi\phi,2,\mathrm{in}},S,\hat S
\}.
\end{aligned}
$}
\end{equation}

\begin{equation}
\mathcal K_{\mathrm{exp-loc},i}
=
\{
\Phi_i^{2,\mathrm{in}},
\hat\Phi_i^{2,\mathrm{in}},
G_i^{2,\mathrm{in}},
\hat G_i^{2,\mathrm{in}}
\}.
\label{eq:K_exp_loc_i}
\end{equation}

\begin{equation}
\mathcal F_j^{2,\mathrm{in},\mathrm{sh}}
=
\{
\hat\chi_{\mu,j}^{2,\mathrm{in}}(t),
\bar g_{\mu,j}^{2,\mathrm{in}}(t),
\hat{\bar g}_{\mu,j}^{2,\mathrm{in}}(t),
\nonumber\
\bar\sigma_{\mu,j}^{2,\mathrm{in}}(t),
\hat{\bar\sigma}_{\mu,j}^{2,\mathrm{in}}(t),
\bar\sigma_{\mu,j}^{\phi,2,\mathrm{in}}(t),
\hat{\bar\sigma}_{\mu,j}^{\phi,2,\mathrm{in}}(t),
\nonumber
\bar\xi_{\mu,j}^{2,\mathrm{in}}(t),
\hat{\bar\xi}_{\mu,j}^{2,\mathrm{in}}(t)
\}_{\mu,t}.
\label{eq:K_j_shared_2in}
\end{equation}

\begin{equation}
\adjustbox{max width=\textwidth}{$
d\mathcal F_j^{2,\mathrm{in},\mathrm{sh}}
:=
\prod_{\mu,t}
d\hat\chi_{\mu,j}^{2,\mathrm{in}}(t)
d\bar g_{\mu,j}^{2,\mathrm{in}}(t)
d\hat{\bar g}_{\mu,j}^{2,\mathrm{in}}(t)
\times
d\bar\sigma_{\mu,j}^{2,\mathrm{in}}(t)
d\hat{\bar\sigma}_{\mu,j}^{2,\mathrm{in}}(t)
d\bar\sigma_{\mu,j}^{\phi,2,\mathrm{in}}(t)
d\hat{\bar\sigma}_{\mu,j}^{\phi,2,\mathrm{in}}(t)
\times
d\bar\xi_{\mu,j}^{2,\mathrm{in}}(t)
d\hat{\bar\xi}_{\mu,j}^{2,\mathrm{in}}(t).
$}
\label{eq:dK_j_shared_2in}
\end{equation}

\begin{equation}
\mathcal F_i^\phi
=
\{
\chi_{\mu i}^{\phi}(t),
\hat\chi_{\mu i}^{\phi}(t),
g_{\mu i}^{1,\phi}(t),
\xi_{\mu i}^{g1,\phi}(t),
\hat\xi_{\mu i}^{g1,\phi}(t),
\tilde\phi_\mu^i(t),
\psi_\mu^i(t)
\}_{\mu,t}.
\label{eq:K_i_phi}
\end{equation}

\begin{align}
\mathcal F_{i,j}^{2,\mathrm{in}}
=
\{&
\chi_{\mu,i,j}^{2,\mathrm{in}}(t),
h_{\mu,i,j}^{2,\mathrm{in}}(t),
g_{\mu,i,j}^{2,\mathrm{in}}(t),
\xi_{\mu,i,j}^{2,\mathrm{in}}(t),
\hat\xi_{\mu,i,j}^{2,\mathrm{in}}(t)
\}_{\mu,t}.
\label{eq:K_ij_2in}
\end{align}

\begin{align}
\mathcal F_n^{\mathrm{global\text{-}site}}
=
\{&
h_{\mu,n}^{1}(t),
g_{\mu,n}^{1}(t),
\chi_{\mu,n}^{1}(t),
\hat\chi_{\mu,n}^{1}(t),
\nonumber
h_{\mu,n}^{3}(t),
g_{\mu,n}^{3}(t),
\xi_{\mu,n}^{3}(t),
\hat\xi_{\mu,n}^{3}(t),
\nonumber\\
&
\xi_{\mu,n}^{1,\phi}(t),
\hat\xi_{\mu,n}^{1,\phi}(t),
\bar\chi_{\mu,n}^{3}(t),
\hat{\bar\chi}_{\mu,n}^{3}(t),
\nonumber
\bar\xi_{\mu,n}^{1}(t),
\hat{\bar\xi}_{\mu,n}^{1}(t)
\}_{\mu,t}.
\label{eq:K_n_global_site_iii}
\end{align}

Then the partition function can be written as:
\begin{equation}
\begin{aligned}
Z
\propto
&\int
\left(
\prod_{\mu,\nu}
\prod_{t,s}
d\mathcal K_{\mathrm{global}}
\right)
\exp\left\{
N\mathcal S_{\mathrm{global}}
[
\mathcal K_{\mathrm{global}}
]
\right\}
\nonumber\times
\prod_{n=1}^{N}
\mathcal Z_N^{\mathrm{global}}
[
j_n^1,
\bar j_n^3,
\bar v_n^1,
v_n^{1,\phi},
v_n^3
]
\nonumber\\
&\times
\int
\left[
\prod_{j=1}^{N_e}
d\mathcal F_j^{2,\mathrm{in},\mathrm{sh}}
\right]
\nonumber \times
\Bigg[
\exp\left\{
\sum_{j=1}^{N_e}
\mathcal S_j^{2,\mathrm{in},\mathrm{sh}}
\right\}
\nonumber\\
&\times
\prod_{i=1}^{\kappa N}
\mathcal Z_i^{\mathrm{local}}
\left[
j_i^{2,\mathrm{in}},
v_i^{2,\mathrm{in}},
j_i^{3,\mathrm{out}},
j_i^\phi,
v_i^{g1,\phi}
\ ;\
\{\mathcal F_j^{2,\mathrm{in},\mathrm{sh}}\}_{j=1}^{N_e}
\right]
\Bigg].
\label{eq:Z_full_corrected_visible_shared_iii}
\end{aligned}
\end{equation}

with the global action defined as:

\begin{equation}
\begin{aligned}
\mathcal S_{\mathrm{global}}
=&
\sum_{\mu,\nu}
\int dt
\int ds
\Big[
\hat\Phi_{\mu\nu}^{1}(t,s)
\Phi_{\mu\nu}^{1}(t,s)
+
\hat G_{\mu\nu}^{1}(t,s)
G_{\mu\nu}^{1}(t,s)
+
\kappa
\hat G_{\mu\nu}^{1,\phi}(t,s)
G_{\mu\nu}^{1,\phi}(t,s)
\nonumber\\
&\quad
-
\bar B_{\mu\nu}^{1}(t,s)
\bar A_{\mu\nu}^{1}(t,s)
-
B_{\mu\nu}^{1,\phi}(t,s)
A_{\mu\nu}^{1,\phi}(t,s)
+
\hat\Phi_{\mu\nu}^{3}(t,s)
\Phi_{\mu\nu}^{3}(t,s)
\nonumber\\
&\quad
+
\kappa
\hat{\bar\Phi}_{\mu\nu}^{2,\mathrm{in}}(t,s)
\bar\Phi_{\mu\nu}^{2,\mathrm{in}}(t,s)
\nonumber\\
&\quad
+
\kappa
\hat{\bar G}_{\mu\nu}^{2,\mathrm{in}}(t,s)
\bar G_{\mu\nu}^{2,\mathrm{in}}(t,s)
+
\kappa
\hat S_\mu(t)
S_\nu(s)
+
\hat G_{\mu\nu}^{3}(t,s)
G_{\mu\nu}^{3}(t,s)
\nonumber\\
&\quad
-
\kappa
B_{\mu\nu}^{g^{1,\phi}}(t,s)
A_{\mu\nu}^{g^{1,\phi}}(t,s)
+
\iota
\hat{\widetilde G}_{\mu\nu}^{2,\mathrm{in}}(t,s)
\widetilde G_{\mu\nu}^{2,\mathrm{in}}(t,s)
\nonumber\\
&\quad
+
\iota
\hat{\widetilde\Phi}_{\mu\nu}^{2,\mathrm{in}}(t,s)
\widetilde\Phi_{\mu\nu}^{2,\mathrm{in}}(t,s)
+
\iota
\hat{\widetilde\Phi}_{\mu\nu}^{2,\mathrm{in},\phi}(t,s)
\widetilde\Phi_{\mu\nu}^{2,\mathrm{in},\phi}(t,s)
\nonumber\\
&\quad
-
\iota
\widetilde B_{\mu\nu}^{2,\mathrm{in}}(t,s)
\widetilde A_{\mu\nu}^{2,\mathrm{in}}(t,s)
-
\iota
\widetilde B_{\mu\nu}^{2,\mathrm{in},\phi}(t,s)
\widetilde A_{\mu\nu}^{2,\mathrm{in},\phi}(t,s)
\nonumber\\
&\quad
+
\iota
\hat{\hat{\bar\Xi}}_{\mu\nu}^{2,\mathrm{in}}(t,s)
\hat{\bar\Xi}_{\mu\nu}^{2,\mathrm{in}}(t,s)
+
\iota
\hat\Phi_{\mu\nu}^{\phi\phi,2,\mathrm{in}}(t,s)
\Phi_{\mu\nu}^{\phi\phi,2,\mathrm{in}}(t,s)
\nonumber\\
&\quad
+
i\kappa
\bar A_{\mu\nu}^{2,\mathrm{in}}(t,s)
G_{\mu\nu}^{3}(t,s)
A_{\mu\nu}^{g^{1,\phi}}(t,s) - \iota \frac12 \hat{\bar{\Xi}}^{2,\i}_{\mu\nu} G^3
\nonumber\\
&\quad
+
\kappa
A_{\mu\nu}^{g^{1,\phi}}(t,s)
A_{\mu\nu}^{g^{1,\phi}}(t,s)
G_{\mu\nu}^{3}(t,s)
\widetilde\Phi_{\mu\nu}^{2,\mathrm{in}}(t,s)
\nonumber\\
&\quad
+
i\kappa\iota
\widetilde\Phi_{\mu\nu}^{2,\mathrm{in},\phi}(t,s)
G_{\mu\nu}^{3}(t,s)
A_{\mu\nu}^{g1,\phi}(t,s)
-
\frac{\iota}{2}
\Phi_{\mu\nu}^{\phi\phi,2,\mathrm{in}}(t,s)
G_{\mu\nu}^{3}(t,s)
\nonumber\\
&\quad
-
\iota\kappa
\widetilde A_{\mu\nu}^{\phi,2,\mathrm{in}}(t,s)
G_{\mu\nu}^{3}(t,s)
+
i\kappa
\widetilde A_{\mu\nu}^{2,\mathrm{in}}(t,s)
G_{\mu\nu}^{3}(t,s)
A_{\mu\nu}^{g1,\phi}(t,s)
\nonumber
\Big].
\label{eq:S_global_corrected_visible_shared_iii}
\end{aligned}
\end{equation}

and the single-site global action as:

\begin{equation}
\begin{aligned}
&\mathcal Z_N^{\mathrm{global}}
[
j_n^1,
\bar j_n^3,
\bar v_n^1,
v_n^{1,\phi},
v_n^3
]
\nonumber:=
\int
d\mathcal K_n^{\mathrm{global\text{-}site}}
\;
\exp\left\{
\mathcal S_n^{\mathrm{global\text{-}site}}
\right\}.
\label{eq:ZN_global_def}
\end{aligned}
\end{equation}

\begin{equation}
\begin{aligned}
\mathcal S_n^{\mathrm{global\text{-}site}}
=&
-\frac12
\sum_{\mu,\nu}
\int dt
\int ds
\Big[
\hat\chi_{\mu,n}^{1}(t)
\hat\chi_{\nu,n}^{1}(s)
\Phi_{\mu\nu}^{0}(t,s)
+
\hat\xi_{\mu,n}^{3}(t)
\hat\xi_{\nu,n}^{3}(s)
\nonumber\\
&\qquad
+
\hat\xi_{\mu,n}^{1,\phi}(t)
\hat\xi_{\nu,n}^{1,\phi}(s)
G_{\mu\nu}^{1,\phi}(t,s)
\Big]
\nonumber\\
&-
\sum_{\mu,\nu}
\int dt
\int ds
\Big[
\hat\Phi_{\mu\nu}^{1}(t,s)
\sigma(h_{\mu,n}^{1}(t))
\sigma(h_{\nu,n}^{1}(s))
+
\hat G_{\mu\nu}^{1}(t,s)
g_{\mu,n}^{1}(t)
g_{\nu,n}^{1}(s)
\nonumber\\
&\qquad
+
\hat\Phi_{\mu\nu}^{3}(t,s)
h_{\mu,n}^{3}(t)
h_{\nu,n}^{3}(s)
+
\hat G_{\mu\nu}^{3}(t,s)
g_{\mu,n}^{3}(t)
g_{\nu,n}^{3}(s)
\nonumber\\
&\qquad
-
\frac12
\widetilde\Phi_{\mu\nu}^{2,\mathrm{in}}(t,s)
\hat{\bar\chi}_{\mu,n}^{3}(t)
\hat{\bar\chi}_{\nu,n}^{3}(s)
+
\frac12
\widetilde G_{\mu\nu}^{2,\mathrm{in}}(t,s)
\hat{\bar\xi}_{\mu,n}^{1}(t)
\hat{\bar\xi}_{\nu,n}^{1}(s) \\
&\qquad
-\iota\kappa\tilde{\Phi}^{2,\i}_{\mu\nu}\hat{\bar\chi}^3g^3
A^{g^{1,\phi}}+ \iota \kappa \bar{A}^{2,\i}_{\mu\nu} \hat{\bar\chi}^3 g^3
\Big]
\nonumber\\
&-
i
\sum_{\mu,\nu}
\int dt
\int ds
\Big[
B_{\mu\nu}^{1,\phi}(t,s)
\hat\xi_{\mu,n}^{1,\phi}(t)
\sigma(h_{\nu,n}^{1}(s))
+
\bar B_{\mu\nu}^{1}(t,s)
\hat{\bar\xi}_{\mu,n}^{1}(t)
\sigma(h_{\nu,n}^{1}(s))\\
&
+\iota\kappa\tilde{\Phi}^{2,\i, \phi}_{\mu\nu} \hat{\bar\chi}^3 g^3 
\Big]
\nonumber\\
&+
\sum_{\mu}
\int dt
\Big[
\big(
v_{\mu,n}^{3}(t)
+
i\hat\xi_{\mu,n}^{3}(t)
\big)
\xi_{\mu,n}^{3}(t)
+
\big(
v_{\mu,n}^{1,\phi}(t)
+
i\hat\xi_{\mu,n}^{1,\phi}(t)
\big)
\xi_{\mu,n}^{1,\phi}(t)
\nonumber\\
&\qquad
+
\big(
j_{\mu,n}^{1}(t)
+
i\hat\chi_{\mu,n}^{1}(t)
\big)
\chi_{\mu,n}^{1}(t)
+
\big(
\bar j_{\mu,n}^{3}(t)
+
i\hat{\bar\chi}_{\mu,n}^{3}(t)
\big)
\bar\chi_{\mu,n}^{3}(t)
\nonumber\\
&\qquad
+
\big(
\bar v_{\mu,n}^{1}(t)
+
i\hat{\bar\xi}_{\mu,n}^{1}(t)
\big)
\bar\xi_{\mu,n}^{1}(t)
\Big].
\label{eq:S_n_global_site_iii}
\end{aligned}
\end{equation}

\begin{equation}
\begin{aligned}
\mathcal S_j^{2,\mathrm{in},\mathrm{sh}}
=&
-\frac12
\sum_{\mu,\nu}
\int dt
\int ds
\;
N
\Big[
\hat\chi_{\mu,j}^{2,\mathrm{in}}(t)
\hat\chi_{\nu,j}^{2,\mathrm{in}}(s)
\Phi_{\mu\nu}^{1}(t,s)
\nonumber
-
\hat{\widetilde G}_{\mu\nu}^{2,\mathrm{in}}(t,s)
\bar g_{\mu,j}^{2,\mathrm{in}}(t)
\bar g_{\nu,j}^{2,\mathrm{in}}(s)
\nonumber\ \\
&
-
\hat{\widetilde\Phi}_{\mu\nu}^{2,\mathrm{in}}(t,s)
\bar\sigma_{\mu,j}^{2,\mathrm{in}}(t)
\bar\sigma_{\nu,j}^{2,\mathrm{in}}(s)
\nonumber
-
\hat{\widetilde\Phi}_{\mu\nu}^{2,\mathrm{in},\phi}(t,s)
\bar\sigma_{\mu,j}^{\phi,2,\mathrm{in}}(t)
\bar\sigma_{\nu,j}^{2,\mathrm{in}}(s)
\nonumber\\
&
-
\hat{\hat{\bar\Xi}}_{\mu\nu}^{2,\mathrm{in}}(t,s)
\hat{\bar\xi}_{\mu,j}^{2,\mathrm{in}}(t)
\hat{\bar\xi}_{\nu,j}^{2,\mathrm{in}}(s)
\nonumber \\
&
-
\hat\Phi_{\mu\nu}^{\phi\phi,2,\mathrm{in}}(t,s)
\bar\sigma_{\mu,j}^{\phi,2,\mathrm{in}}(t)
\bar\sigma_{\nu,j}^{\phi,2,\mathrm{in}}(s)
\nonumber
-
\widetilde B_{\mu\nu}^{2,\mathrm{in}}(t,s)
\bar\sigma_{\mu,j}^{2,\mathrm{in}}(t)
\hat{\bar\xi}_{\nu,j}^{2,\mathrm{in}}(s)
\nonumber \\
&
-
\widetilde B_{\mu\nu}^{\phi,2,\mathrm{in}}(t,s)
\bar\sigma_{\mu,j}^{\phi,2,\mathrm{in}}(t)
\hat{\bar\xi}_{\nu,j}^{2,\mathrm{in}}(s)
\Big] \\
&
\nonumber -
i
\sum_{\mu}
\int dt
\Big[
\hat{\bar g}_{\mu,j}^{2,\mathrm{in}}(t)
\bar g_{\mu,j}^{2,\mathrm{in}}(t)
+
\hat{\bar\sigma}_{\mu,j}^{2,\mathrm{in}}(t)
\bar\sigma_{\mu,j}^{2,\mathrm{in}}(t)
\nonumber \\
&
+
\hat{\bar\sigma}_{\mu,j}^{\phi,2,\mathrm{in}}(t)
\bar\sigma_{\mu,j}^{\phi,2,\mathrm{in}}(t)
+
\hat{\hat{\bar\xi}}_{\mu,j}^{2,\mathrm{in}}(t)
\hat{\bar\xi}_{\mu,j}^{2,\mathrm{in}}(t)
\Big]
\nonumber
\\
&
-
i
\sum_{\mu,\nu}
\int dt
\int ds
\Big[
-\frac12
\Phi_{\mu\nu}^{1}(t,s)
\hat\chi_{\mu,j}^{2,\mathrm{in}}(t)
\hat\chi_{\nu,j}^{2,\mathrm{in}}(s)
\Big].
\label{eq:S_j_shared_2in_iii}
\end{aligned}
\end{equation}

\begin{equation}
\begin{aligned}
\mathcal Z_i^{\mathrm{local}}
&\left[
j_i^{2,\mathrm{in}},
v_i^{2,\mathrm{in}},
j_i^{3,\mathrm{out}},
j_i^\phi,
v_i^{g1,\phi}
\ ;\
\{\mathcal K_j^{2,\mathrm{in},\mathrm{sh}}\}_{j=1}^{N_e}
\right]
 :=
\int
d\mathcal K_{\mathrm{exp-loc},i}\\&
\int
d\mathcal K_i^\phi
\nonumber\times
\exp\left\{
\mathcal S_i^{\mathrm{exp-loc}}
+
\mathcal S_i^\phi
\right\}
\nonumber
\times
\frac{\iota}{\iota N}
\prod_{j=1}^{N_e}
\left[
\int
d\mathcal K_{i,j}^{2,\mathrm{in}}
\;
\exp\left\{
\mathcal S_{i,j}^{2,\mathrm{in}}
\right\}
\right].
\label{eq:Zi_local_conditional_visible_shared_iii}
\end{aligned}
\end{equation}

\begin{equation}
\begin{aligned}
\mathcal S_i^{\mathrm{exp-loc}}
=
\sum_{\mu,\nu}
\int_0^\infty dt
\int_0^\infty ds
\Big[
\iota
\hat\Phi_{\mu\nu,i}^{2,\mathrm{in}}(t,s)
\Phi_{\mu\nu,i}^{2,\mathrm{in}}(t,s)
+
\iota
\hat G_{\mu\nu,i}^{2,\mathrm{in}}(t,s)
G_{\mu\nu,i}^{2,\mathrm{in}}(t,s)
\Big].
\label{eq:S_i_exp_loc}
\end{aligned}
\end{equation}

\begin{equation}
\adjustbox{max width=\textwidth}{$
\begin{aligned}
\mathcal S_i^\phi
=&
\sum_{\mu}
\int_0^\infty dt
\;
g_{\mu i}^{1,\phi}(t)
g_{\mu i}^{1,\phi}(t)\hat G^{1,\phi}_{\mu\nu}
\nonumber -
\frac12
\sum_{\mu,\nu}
\int dt
\int ds
\Big[
\hat\chi_{\mu i}^{\phi}(t)
\hat\chi_{\nu i}^{\phi}(s)
\Phi_{\mu\nu}^{1}(t,s)
\Big]
\nonumber\\
&-
i
\sum_{\mu,\nu}
\int dt
\int ds
\Big[
\hat\chi_{\mu i}^{\phi}(t)
g_{\nu i}^{1,\phi}(s)
A_{\mu\nu}^{1,\phi}(t,s)
-
B_{\mu\nu}^{g1,\phi}(t,s)
\hat\xi_{\mu i}^{g1,\phi}(t)
\tilde\phi_{\nu}^{i}(s)
\Big]
\nonumber\\
&-
\sum_{\mu,\nu}
\int dt
\int ds
\Big[
\hat G_{\mu\nu}^{1,\phi}(t,s)
g_{\mu i}^{1,\phi}(t)
g_{\nu i}^{1,\phi}(s)
\nonumber
+
\hat{\bar\Phi}_{\mu\nu}^{2,\mathrm{in}}(t,s)
\tilde\phi_{\mu}^{i}(t)
\tilde\phi_{\nu}^{i}(s)
\Phi_{\mu\nu,i}^{2,\mathrm{in}}(t,s)
\\
&
+
\hat{\bar G}_{\mu\nu}^{2,\mathrm{in}}(t,s)
G_{\mu\nu,i}^{2,\mathrm{in}}(t,s)
\Big]
\nonumber\\
&-
\sum_{\mu}
\int dt
\;
\hat S_{\mu}(t)
e^{\psi_{\mu}^{i}(t)}
\nonumber +
\sum_{\mu}
\int dt
\Big[
\big(
j_{\mu i}^{\phi}(t)
+
i\hat\chi_{\mu i}^{\phi}(t)
\big)
\chi_{\mu i}^{\phi}(t)
+
\big(
v_{\mu i}^{g1,\phi}(t)
+
i\hat\xi_{\mu i}^{g1,\phi}(t)
\big)
\xi_{\mu i}^{g1,\phi}(t)
\Big].
\label{eq:S_i_phi_iii}
\end{aligned}
$}
\end{equation}

\begin{equation}
\begin{aligned}
\mathcal S_{i,j}^{2,\mathrm{in}}
=&
-
i
\sum_{\mu,\nu}
\int dt
\int ds
\Big[
\frac1\kappa
\hat\chi_{\mu,j}^{2,\mathrm{in}}(t)
g_{\nu,i,j}^{2,\mathrm{in}}(s)
\bar A_{\mu\nu}^{1}(t,s)
\Big]
\nonumber
\\
&
-
i
\sum_{\mu}
\int dt
\Big[
\hat{\bar g}_{\mu,j}^{2,\mathrm{in}}(t)
g_{\mu,i,j}^{2,\mathrm{in}}(t)
\nonumber\
\hat{\bar\sigma}_{\mu,j}^{2,\mathrm{in}}(t)
\sigma\!\left(
h_{\mu,i,j}^{2,\mathrm{in}}(t)
\right)
\tilde\phi_\mu^{i}(t)
\nonumber\\
&\qquad
-\hat{\bar\sigma}_{\mu,j}^{\phi,2,\mathrm{in}}(t)
\sigma\!\left(
h_{\mu,i,j}^{2,\mathrm{in}}(t)
\right)
\tilde\phi_\mu^{i}(t)
\hat\xi_{\mu i}^{g1,\phi}(t)
\nonumber
-
\hat{\hat{\bar\xi}}_{\mu,j}^{2,\mathrm{in}}(t)
\hat{\xi}_{\mu,i,j}^{2,\mathrm{in}}(t)
\tilde\phi_\mu^{i}(t)
\Big]
\nonumber \\
&-
\sum_{\mu,\nu}
\int dt
\int ds
\Big[
\hat\Phi_{\mu\nu,i}^{2,\mathrm{in}}(t,s)
\sigma\!\left(
h_{\mu,i,j}^{2,\mathrm{in}}(t)
\right)
\sigma\!\left(
h_{\nu,i,j}^{2,\mathrm{in}}(s)
\right)
\nonumber
+
\hat G_{\mu\nu,i}^{2,\mathrm{in}}(t,s)
g_{\mu,i,j}^{2,\mathrm{in}}(t)
g_{\nu,i,j}^{2,\mathrm{in}}(s)
\Big]
\nonumber \\
&+
\sum_{\mu}
\int dt
\;
N
\Big[
\big(
j_{\mu i}^{2,\mathrm{in}}(t)
+
i\hat\chi_{\mu,j}^{2,\mathrm{in}}(t)
\big)
\chi_{\mu,i,j}^{2,\mathrm{in}}(t)
\nonumber
+
\big(
\frac{1}{\kappa N}
v_{\mu,i}^{2,\mathrm{in}}(t)
+
i\hat\xi_{\mu,i,j}^{2,\mathrm{in}}(t)
\big)
\xi_{\mu,i,j}^{2,\mathrm{in}}(t)
\Big].
\label{eq:S_ij_2in_iii}
\end{aligned}
\end{equation}

One way to look at it is the following: there are four conditional factorization levels in the full partition function.
The first is the standard global \(N\)-site DMFT factorization over \(n\).
The remaining three form a nested local/shared sector: a shared \(N_e\)-component factorization over \(j\), an expert factorization over \(i\) conditional on the shared \(j\)-fields, and an intra-expert \(N_e\)-component factorization over
\((i,j)\)-local fields. The important point is that these are \emph{conditional} factorizations. 

\subsection{Saddle Point Approximation}
To write down the saddle point equations, we first define the single-site distributions, where $\mathcal H$ in each case is the logarithm of the integrand of the corresponding $\mathcal Z$. We can then define for each $\mathcal Z\in\{\mathcal Z_{N}^{global}, Z_{\text{sh}}\,,\mathcal Z_{{N_e}\,j },\mathcal Z_{M}\}$ and the corresponding $\mathcal H$ the average

\begin{equation}
\label{defn_average_single_site_r3}
    \langle\mathcal O(\{\chi,\xi\})\rangle_{\mathcal Z} 
    = \frac{1}{\mathcal Z}\int\prod_\mu d\mathcal F \exp(-\mathcal H[\{\chi,\xi\},\{j,v\}]) \mathcal O(\{\chi,\xi\})
\end{equation}

With this apparatus in place, we can take saddle equations. We treat expert-local kernels as microvariables which are implicitly defined in terms of $\chi,\xi$, and so do not take saddle equations of them at this point.

\subsection{Saddle-Point Equations}
As in Regime II, we can use the saddle point approximation, which consist in finding the set of equations that lead to a stationary action $S[\mathcal{K}].$

We note the following:
\begin{itemize}
    \item
At zero source, all single site averages $\langle\rangle_{\mathcal Z_{N}^{global}} ,\langle\rangle_{\mathcal Z_{sh}}$ are equivalent.
\item
\textit{Conditional on the set $\mathcal K_{global},\mathcal F_{sh}$} of global and shared kernels, $\mathcal Z^{local}$ factorises over experts, and so we can write $\langle\rangle_{\mathcal Z_{N_e j}}$ for the conditional average over the distribution over the experts neurons and $\langle\rangle_{\mathcal Z_{M}| \mathcal K_{global}}$ for the conditional average over the distribution of the router's neurons.
\item 
Expert-local variables follow single-site processes, so we can drop expert indices
\item
By a derivation similar to that in \ref{sec:dmft_regime1}, we can prove that conjugate kernels defined as covariances between conjugate fields vanish, since they have no physical meaning and can not influence the dynamics in addition to imposing constraint when introducing kernel definitions.
\end{itemize}

The global kernels are given by their expectation values over the single-site global distribution $\mathcal Z^{\text{global}}_{N}$:

\begin{equation}
\label{global_saddle_v2_r3}
\begin{aligned}
\Phi^{1}_{\mu\nu}(s,t) &= \big\langle \sigma(h_{\mu}^{1}(s))\,\sigma(h_{\nu}^{1}(t)) \big\rangle_{\mathcal Z^{\text{global}}_{N}|\mathcal{K}_{global}} \\[3pt]
\Phi^{3}_{\mu\nu}(s,t) &= \big\langle h_{\mu }^{3}(s)\,h_{\nu }^{3}(t) \big\rangle_{\mathcal Z^{\text{global}}_{N}|\mathcal{K}_{global}} \\[3pt]
G^{1}_{\mu\nu}(s,t) &= \big\langle g_{\mu }^{1}(s)\,g_{\nu }^{1}(t) \big\rangle_{\mathcal Z^{\text{global}}_{N}|\mathcal{K}_{global}} \\[3pt]
G^{3}_{\mu\nu}(s,t) &= \big\langle g_{\mu }^{3}(s)\,g_{\nu }^{3}(t) \big\rangle_{\mathcal Z^{global}_{N}|\mathcal{K}_{global}} \\[3pt]
A^{1,\phi}_{\mu\nu}(s,t) &= -i\big\langle \hat{\xi}_{\mu}^{1,\phi}(s)\, \sigma(h^1_\nu(t))(t) \big\rangle_{\mathcal Z_{N}^{global}|\mathcal{K}_{global}} \\[3pt]
\bar A^{1}_{\mu\nu}(s,t) &= - i\, \big\langle \hat{\bar\xi}^1_{\mu }(t)\sigma(h^1_{\nu }(s)) \big\rangle_{\mathcal Z_{N}^{global}|\mathcal{K}_{global}} \\[3pt]
\bar B^{2,\i}_{\mu\nu}(s,t) &= - i\, \big\langle \hat{\bar\chi}^3_{\mu }(t)g^3_{\nu }(s) \big\rangle_{\mathcal Z_{N}^{global}|\mathcal{K}_{global}}
\end{aligned}
\end{equation}

The shared kernels are given by their expectation values over the single-site shared distribution $\mathcal Z_{sh}$:
\begin{equation}
\widetilde G^{2,\mathrm{in}}_{\mu\nu}(s,t)
=
\Big\langle
\bar g^{2,\mathrm{in}}_{\mu,j}(s)
\bar g^{2,\mathrm{in}}_{\nu,j}(t)
\Big\rangle_{\mathcal{Z}_{sh}|\mathcal{K_{\text{global}}}},
\end{equation}

\begin{equation}
\widetilde \Phi^{2,\mathrm{in}}_{\mu\nu}(s,t)
=
\Big\langle
\bar \sigma^{2,\mathrm{in}}_{\mu,j}(s)
\bar \sigma^{2,\mathrm{in}}_{\nu,j}(t)
\Big\rangle_{\mathcal{Z}_{sh}|\mathcal{K_{\text{global}}}},
\end{equation}

\begin{equation}
\widetilde \Phi^{2,\mathrm{in},\phi}_{\mu\nu}(s,t)
=
\Big\langle
\bar \sigma^{\phi,2,\mathrm{in}}_{\mu,j}(s)
\bar \sigma^{2,\mathrm{in}}_{\nu,j}(t)
\Big\rangle_{\mathcal{Z}_{sh}|\mathcal{K_{\text{global}}}},
\end{equation}

\begin{equation}
\hat{\bar \Xi}^{2,\mathrm{in}}_{\mu\nu}(s,t)
=
\Big\langle
\hat{\bar \xi}^{2,\mathrm{in}}_{\mu,j}(s)
\hat{\bar \xi}^{2,\mathrm{in}}_{\nu,j}(t)
\Big\rangle_{\mathcal{Z}_{sh}|\mathcal{K_{\text{global}}}}
\end{equation}

\begin{equation}
\tilde\Phi^{\phi\phi,2,\mathrm{in}}_{\mu\nu}(s,t)
=
\Big\langle
\bar \sigma^{\phi,2,\mathrm{in}}_{\mu,j}(s)
\bar \sigma^{\phi,2,\mathrm{in}}_{\nu,j}(t)
\Big\rangle_{\mathcal{Z}_{sh}|\mathcal{K_{\text{global}}}}
\end{equation}

\begin{equation}
\widetilde A^{2,\mathrm{in}}_{\mu\nu}(s,t)
=
\Big\langle
\bar \sigma^{2,\mathrm{in}}_{\mu,j}(s)
\bar \xi^{2,\mathrm{in}}_{\nu,j}(t)
\Big\rangle_{\mathcal{Z}_{sh}|\mathcal{K_{\text{global}}}}
\end{equation}

\begin{equation}
\widetilde A^{\phi,2,\mathrm{in}}_{\mu\nu}(s,t)
=
\Big\langle
\bar \sigma^{\phi,2,\mathrm{in}}_{\mu,j}(s)
\bar \xi^{2,\mathrm{in}}_{\nu,j}(t)
\Big\rangle_{\mathcal{Z}_{sh}|\mathcal{K_{\text{global}}}}
\end{equation}

\begin{equation}
\widetilde B^{2,\mathrm{in}}_{\mu\nu}(s,t)
=
i\kappa
G^3_{\mu\nu}(s,t)
A^{g^{1,\phi}}_{\mu\nu}(s,t)
\end{equation}

\begin{equation}
\widetilde B^{\phi,2,\mathrm{in}}_{\mu\nu}(s,t)
=
-i\kappa
G^3_{\mu\nu}(s,t)
\end{equation}

The router kernels require expectations over the expert ensemble distribution $\mathcal Z_{M}$, conditioned on the global kernels and shared fields:
\begin{equation}
\begin{aligned}
\bar\Phi^{2,\i}_{\mu\nu}(s,t) &= \big\langle \tilde\phi_{\mu}(s)\,\tilde\phi_\nu(t) \Phi^{2,\i}_{\mu\nu}(s,t) \big\rangle_{\mathcal Z_{M}|\mathcal{K_{\text{global}}},\mathcal{F_{\text{shared}}}} \\[3pt]
\bar G^{2,\i}_{\mu\nu}(s,t) &= \big\langle G^{2,\i}_{\mu\nu,}(s,t) \big\rangle_{\mathcal Z_{M}|\mathcal{K_{\text{global}}},\mathcal{F_{\text{shared}}}} \\[3pt]
G^{1,\phi}_{\mu\nu}(s,t) &= \big\langle g_{\mu}^{1,\phi}(s)\, g_{\nu}^{1,\phi}(t) \big\rangle_{\mathcal Z_{M}|\mathcal{K_{\text{global}}},\mathcal{F_{\text{shared}}}} \\[3pt]
{\mathcal S}_{\mu}(t) &= \big\langle e^{\psi_\mu(t)} \big\rangle_{\mathcal Z_{M}|\mathcal{K_{\text{global}}},\mathcal{F_{\text{shared}}}} \\
A^{g^{1,\phi}}_{\mu\nu}(s,t)
&=
-i
\Big\langle
\hat{\xi}^{g^{1,\phi}}_{\mu}(s)\tilde \phi_{\nu}(t)
\Big\rangle_{\mathcal Z_{M}|\mathcal{K_{\text{global}}},\mathcal{F_{\text{shared}}}} \\
B^{1,\phi}_{\mu\nu}(s,t)
&=
-i\kappa
\left\langle
\hat \chi^\phi_\mu(s)
g^{1,\phi}_\nu(t)
\right\rangle_{\mathcal Z_{M}|\mathcal{K_{\text{global}}},\mathcal{F_{\text{shared}}}} \\
B^{g^{1,\phi}}_{\mu\nu}(s,t)
=&+
i
\bar A_{\mu\nu}^{2,\mathrm{in}}(t,s)
G_{\mu\nu}^{3}(t,s)+
A_{\mu\nu}^{g^{1,\phi}}(t,s)
G_{\mu\nu}^{3}(t,s)
\widetilde\Phi_{\mu\nu}^{2,\mathrm{in}}(t,s) \\
&+i\iota 
\widetilde\Phi_{\mu\nu}^{2,\mathrm{in},\phi}(t,s)
G_{\mu\nu}^{3}(t,s)+i
\widetilde A_{\mu\nu}^{2,\mathrm{in}}(t,s)
G_{\mu\nu}^{3}(t,s) -\iota\tilde{\Phi}^{2,\i}_{\mu\nu}\langle\hat{\bar\chi}^3g^3 \rangle_{\mathcal Z_{M}|\mathcal{K_{\text{global}}},\mathcal{F_{\text{shared}}}}
\end{aligned}
\end{equation}

\subsection{Expert-Local Order Parameters}
The local kernels are determined by independent processes within each expert's internal dimensions:
\begin{equation}
\label{local_saddle_equations_v1_r3}
    \begin{aligned}
\Phi^{2,\i}_{\mu\nu}(s,t) &=  \big\langle \sigma({\left[h^{2,\text{in}}_{\mu}(t)\right]_{j}})\,\sigma({\left[h^{2,\text{in}}_{\nu}(s)\right]_{j}}) \big\rangle_{\mathcal Z_{N_e\,j}|\mathcal{K_{\text{global}}},\mathcal{F_{\text{shared}}}} \\[3pt]
G^{2,\i}_{\mu\nu}(s,t) &= \big\langle {\left[g^{2,\text{in}}_{\mu}(t)\right]_{j}}\,{\left[g^{2,\text{in}}_{\nu}(s)\right]_{j}} \big\rangle_{\mathcal Z_{N_e\,j}|\mathcal{K_{\text{global}}},\mathcal{F_{\text{shared}}}} \\[3pt] 
A^{2,\i}_{\mu\nu}(s,t) &=  \big\langle \left[\hat\xi^{2,\i}_{\mu}(s)\right]_j\sigma(\left[h_{\nu}^{2,\i}\right]_j) \big\rangle_{\mathcal Z_{N_e\,j}|\mathcal{K_{\text{global}}},\mathcal{F_{\text{shared}}}} \\[3pt]
\bar B^1_{\mu\nu}(s,t)
&=
-\frac{i}{\kappa}
\left\langle
\hat \chi^{2,\mathrm{in}}_\mu(s)
g^{2,\mathrm{in}}_\nu(t)
\right\rangle_{\mathcal Z_{N_e\,j}|\mathcal{K_{\text{global}}},\mathcal{F_{\text{shared}}}}
\end{aligned} 
\end{equation}

All non-physical conjugate kernels identically vanish at the saddle point: $\hat{\Phi}^{3} = \hat\Phi^1 = \hat{G}^{1} = \hat{\bar\Phi}^{2,\i} = \dots = 0$.

\subsection{Hubbard-Stratonovich transformation}
The Hubbard-Stratonovich trick allows us to rewrite the quadratic terms over conjugate fields as Gaussian integrals over linear conjugate fields:
\begin{equation}
\label{hubbard_trick_r3}
\adjustbox{max width=\textwidth}{$
    \exp\!\left(-\frac{1}{2}\,\mathbf{\hat x}^{\top} A\,\mathbf{\hat x}\right)
= \int_{\mathbb{R}^{d}}
\frac{d\mathbf{u}}{(2\pi)^{d/2} \sqrt{\det A}}\,
\exp\!\left(-\frac{1}{2}\,\mathbf{u}^{\top}A^{-1}\mathbf{u}
- i\,\mathbf{u}\!\cdot\!\mathbf{\hat x}\right)
= \big\langle \exp(-i\,\mathbf{u}\!\cdot\!\mathbf{\hat x}) \big\rangle_{\mathbf{u} \sim \mathcal{N}(0,A)}.
$}
\end{equation}
where $\mathbf{\hat x}$ is a generic conjugate fields in the partition function above.
Using Stein's lemma, we can reformulate the definitions of each of the kernels A and B as response functions: 
\begin{equation}
\label{Aellli_avg_r3}
    \begin{aligned}
A^{\,\ell}_{\mu\nu,i}(t,s)
&= -\,i\,
\big\langle
\hat{\xi}^{\,\ell}_{\nu,i}(t)\,
\sigma\!\big(h^{\,\ell-1}_{\mu,i}(s)\big)
\big\rangle_{\beta_{\nu,i}^{\ell}(s)}
\\
&=
\big[G^{\,\ell}_{i}\big]^{-1}
\Big\langle
\big(
\xi^{\,\ell}_{i}
-
B^{\,\ell}_{i}\,
\sigma\!\big(h^{\,\ell-1}_{i}\big)
\big)\,
\sigma\!\big(h^{\,\ell-1}_{i}\big)
\Big\rangle_{\beta_{\nu,i}^{\ell}(s)}
\\
&=
\left\langle\frac{\partial\sigma(h_{\mu,i}^{\ell-1}(t))}{\partial\beta_{\nu,i}^{\ell\top}(s)}\right\rangle_{\beta_{\nu,i}^{\ell}(s)}
\end{aligned}
\end{equation}
It easier now integrate over all the $\mathbf{\hat x}$'s, since the argument of the exponential in $\mathcal Z$ has been linearised with respect to them all. Doing so yields delta functions that give us the final DMFT dynamics.

\subsection{DMFT Dynamics}
Piecing together the self-consistent order parameters and the exact integrations over the Hubbard-Stratonovich fields, the DMFT description of the infinite-width MoE is summarized below.

\begin{equation}
\label{preactivations_DMFT}
\begin{aligned}
h_{\mu}^1(t) 
&= \alpha_\mu^1
  + \frac{\eta_0 \gamma_0}{P}
    \int_0^t \! ds
    \sum_{\nu} \Delta_{\nu} \, g_{\nu}^1(s) \, \Phi_{\mu\nu}^0 \qquad \alpha^1\sim\mathcal{N}(0,K^x)\\
{h^{2,\i}_{\mu,i}}(t)
&= \alpha^{2,\i}_\mu(t)
  + \frac{\eta_0 \gamma_0}{\kappa P}
    \int_0^t \! ds
    \sum_{\nu} \Big[\Delta_{\nu}\Phi_{\mu \nu}^{1}(s,t) + \bar A^1_{\nu\mu}(s,t) \Big] \, g_{\nu,i}^{2,\i}(s)
    \, \qquad \alpha^{2,\i}\sim\mathcal{N}(0,\Phi^1(t))\\    
h_{\mu}^{3}(t)&= 
\bar\alpha^{3}_\mu(t)
  + \frac{\eta_0 \gamma_0}{P}
    \int_0^t \! ds
    \sum_{\nu} \Big[\Delta_{\nu}\bar\Phi_{\mu \nu}^{2,\i}(s,t) +\mathcal{R}^{3}_{\nu\mu}(s,t)\Big] \, g_{\nu}^{3}(s)
    \, \qquad \bar\alpha^3\sim\mathcal{N}(0,\tilde\Phi^{2,\i}(t)) \\
&\text{with  } \mathcal{R}^{3}_{\nu\mu}(s,t)
=
\kappa \bar A^{2,\mathrm{in}}_{\nu\mu}(s,t)
+
\kappa \widetilde\Phi^{2,\mathrm{in},\phi}_{\nu\mu}(s,t)
-
\kappa \widetilde\Phi^{2,\mathrm{in}}_{\nu\mu}(s,t)
A^{g^{1,\phi}}_{\nu\mu}(s,t) \\
\psi_\mu(t)
&= \alpha^\phi_\mu
   + \frac{\eta_0 \gamma_0}{P}
     \int_{0}^{t} \! ds \,
     \sum_{\nu=1}^{P}
     \Big[\Delta_\nu(s)\Phi^1_{\mu\nu}(s,t) + A_{\mu\nu}^{1,\phi}\Big]\,
     g^{1,\phi}_\nu(s)\,
     \qquad \alpha^\phi\sim\mathcal{N}(0,\frac1\kappa\Phi^1(t)) \\
\tilde \phi_\mu(t) &= \frac{e^{\psi_\mu(t)}}{\mathcal{S}_\mu(t)},\qquad
\hat\psi_\mu(t) = m_\mu(t) \psi_\mu(t)
,\qquad m_\mu(t) = \1(\psi_\mu(t) -\tau_\mu(t))\\
\qquad &\text{where } \tau_\mu(t) \text{ is chosen such that } \rho = \left\langle m_\mu(t) \right\rangle_{\mathcal Z_{M}|\mathcal K_{global}}
\end{aligned}
\end{equation}

\begin{equation}
\label{gradients_DMFT}
    \begin{aligned}
        g_{\mu}^\phi(t) &= \frac{\eta_0\gamma_0\iota}{P\kappa}\int_0^tds\sum_{\nu}G^3_{\mu\nu}(t,s)\tilde\phi_\mu\Big[ \Phi^{2,\i}_{\mu\nu,i}(t,s)\tilde\phi_\nu - \bar\Phi^{2}_{\mu\nu}(t,s) \Big] \Delta_{\nu} \\ 
        z_{\mu}^{1}(t)&= \bar\beta_{\mu}^{1}(t)+\beta_{\mu}^{1,\phi}(t)+ \frac{\eta_0 \gamma_0}{P}\int_0^t \! ds\sum_{\nu}\Big[ \bar B^1_{\mu\nu} + B_{\mu\nu}^{1,\phi}(t,s)+\Big[\Delta_{\nu}\,\frac\iota\kappa\bar G_{\mu \nu}^{2}(s,t) \\
        & \quad + \Delta_{\nu}G^{1,\phi}_{\mu\nu}(s,t) \Big]\Big] \, \sigma(h^1_{\nu}(s))\\
        & \bar\beta_{\mu}^{1}(t) \sim \mathcal{N}(0,\tilde G^{2,\i}(t,t)) \qquad \beta_{\mu}^{1,\phi}(t) \sim \mathcal{N}(0,G^{1,\phi}(t,t))\\
        z_{\mu}^{2,\i}(t)&=\beta_{\mu}^{2,\i}(t)\tilde \phi_\mu^i+ \frac{\eta_0 \gamma_0}{\kappa P}\int_0^t \! ds\sum_{\nu}\Delta_{\nu} \, \sigma(h^{2,\i}_{\nu,i}(s)) \,G_{\mu \nu}^{3}(s,t)\tilde\phi_\mu\tilde\phi_\nu \qquad \beta^{2,\i}(t) \sim \mathcal{N}(0,G^3(t,t))\\
        z_\mu^{3}(t)&=\beta^{3}(t)+\frac{\gamma_0\eta_0}{P}\int_0^tds\sum_\nu\Delta_\nu h_\nu^3(s) \qquad \beta^3\sim\mathcal{N}(0,1),
    \end{aligned}
\end{equation}

\newpage
\section{DMFT Analysis for Regime III (\texorpdfstring{$\mu$}{mu}P)}
\label{sec:dmft-regime-iii-mup}
In this section we are considering the case in which $M, N, N_e \to \infty$ at fixed ratios. Specifically, we define $\kappa = \frac{M}{N}, \iota = \frac{N_e}{N}$, where $\kappa$,$\iota$ are order one in $N$. The forward pass, following \textit{$\mu$P} is defined as:

\begin{equation}
\label{infinite_experts_architecture_r3_mup}
    \begin{aligned}
        \mathbb R^ N\ni h_\mu^1 &=\frac{1}{\sqrt D}W^1x_\mu \\
        \mathbb R^M\ni\psi_\mu &=\frac{1}{\sqrt N} Q\sigma(h^1_\mu) \\
        \mathbb R^M\ni\phi_\mu&=\text{softmax}(\psi_\mu)\\
        \mathbb R^{N_{e}} \ni {h^{2,\text{in}}_{\mu, i}}&=\frac{1}{\sqrt N}W^{2,\i}_i\sigma(h_\mu^{1}),\\
        \mathbb R^ N\ni h^{2,\o}_{\mu,i}&=\frac1{\sqrt N}W^{2,\o}_i\sigma({h^{2,\text{in}}_{\mu, i}}),\\
        \mathbb R^ N\ni h^3_\mu&=\sum_{i=1}^M\phi_\mu^i h_{\mu,i}^{2,\o}\\
        \mathbb R\ni h^{4}_\mu&=\frac{1}{\sqrt{N}}w^{4\mathsf{T}}h^3_\mu\\
        f_\mu&=\frac{1}{\gamma}h_\mu^{4}\\
        \eta&=\eta_0\gamma^2\\
        \eta_E &= \eta_0\gamma^2 N\\
        \eta_Q &= \eta_0\gamma^2 \kappa \\
        \gamma&=\gamma_0\sqrt N\\
        \eta_0,\gamma_0&\sim O(1)\\
        w_\alpha^{4}(0),\left[W_{i}^{2,\o}(0)\right]_{\alpha\beta},\left[W_{i}^{2,\i}(0)\right]_{\alpha\beta}&,W_{\alpha\beta}^{1}(0),Q_{\alpha\beta}(0)\sim\mathcal N(0,1)
    \end{aligned}
\end{equation}

Again, we let $\bm\theta = \text{Vec}\{W^1,W_i^{2,\i},W_i^{2,\o},w^4,Q\}$, and define

\begin{equation}
    \tilde\phi^i_\mu := \kappa N \phi^i_\mu
\end{equation}

Following an analogous derivation as in \Cref{sec:dmft_regime2}, which we omit for conciseness, we arrive at the following self-consistent set of DMFT equations.

\newpage
\begin{equation*}
\adjustbox{max width=\textwidth}{$
\begin{aligned}
\alpha^1(t)&\sim\mathcal{N}(0,\Phi^0)\qquad
\quad\beta^{3}(t)\sim\mathcal{N}(0,\mathbb I)\qquad
\beta^{1,\phi}(t)\sim\mathcal{GP}(0,G^{1,\phi})\\
\quad\alpha^{\phi}(t)&\sim\mathcal{GP}(0,\Phi^1)\qquad
\quad
\alpha^{2,\i}(t) \sim \mathcal{GP}(0,\Phi^1)\qquad
\quad
\beta^{2,\i}(t) \sim \mathcal{GP}(0,\tilde\phi\,\tilde\phi \,G^3)
\\[4pt] 
\Phi_{\mu\nu}^{0} &= \frac1Dx_\mu x_\nu^\top,
\quad
\Phi_{\mu\nu}^{1}(t,s) = \langle \sigma(h_\mu^{1}(t))\sigma(h_\nu^{1}(s)) \rangle_{\mathcal Z_N^{global}},
\quad
\Phi^{2,\i}_{\mu\nu}(s,t)
=
\big\langle
\sigma({h^{2,\text{in}}_{\mu}(t)})\,\sigma({h^{2,\text{in}}_{\nu}(s)})
\big\rangle_{\mathcal Z_{N_e}|\mathcal K_{global}} \\[4pt]
\bar\Phi^{2,\i}_{\mu\nu}(s,t)
&=
\big\langle
\tilde\phi_{\mu}(s)\,\tilde\phi_\nu(t) \Phi^{2,\i}_{\mu\nu}(s,t)
\big\rangle_{\mathcal Z_{M}|\mathcal K_{global}}\qquad\bar G^{2,\i}_{\mu\nu}(s,t)
=
\big\langle
G^{2,\i}_{\mu\nu}(s,t)
\big\rangle_{\mathcal Z_{M}|\mathcal K_{global}}\\
\Phi^{3}_{\mu\nu}(s,t) &=\big\langle h_\mu^{3}(t) h^{3}_\nu(s)\rangle_{\mathcal Z_{N}^{global}},\quad
G_{\mu\nu}^{1}(t,s) = \langle [\dot\sigma(h_\mu^{1}(t))\odot z_\mu^{1}(t)] [\dot\sigma(h_\nu^{1}(s))\odot z_\nu^{1}(s)] \rangle_{\mathcal Z_N^{global}}\\[4pt]
G^{2,\i}_{\mu\nu}(s,t)
&=
\big\langle
\left[
\dot\sigma(h^{2,\text{in}}_{\mu}(t))\odot
z^{2,\text{in}}_{\mu}(t)\right]\,\left[
\dot\sigma(h^{2,\text{in}}_{\nu}(s))\odot
z^{2,\text{in}}_{\nu}(s)\right]
\big\rangle_{\mathcal Z_{N_e}|\mathcal K_{global}}\\[4pt]
 G^{1,\phi}_{\mu\nu}(s,t)
&=
\big\langle
g_{\mu}^{1,\phi}(s)\, g_{\nu}^{1,\phi}(t)
\big\rangle_{\mathcal Z_{M}|\mathcal K_{global}},
\quad
G^{3}_{\mu\nu}(s,t)
=
\big\langle
z_{\mu }^{3}(s)\,z_{\nu }^{3}(t)
\big\rangle_{\mathcal Z^{global}_{N}}
\\[4pt]
 A^{2,\i}_{\mu\nu}(t,s)
 &=
 \left\langle\frac{\partial\sigma(h^{2,\i}_\mu(t))}{\partial\beta^{2,\i\top}_\nu(s)}\right\rangle_{\mathcal Z_{N_e}|\mathcal K_{global}},
 \qquad
 \bar A^{2,\i}_{\mu\nu}(t,s)=\frac\kappa\iota\left\langle 
 \tilde\phi_\mu(t)\tilde\phi_\nu(s)A^{2,\i}_{\mu\nu}(t,s)
 \right\rangle_{\mathcal Z_{M}|\mathcal K_{global}}
 \\[4pt]
A^{1,\phi}_{\mu\nu}(t,s)&=\left\langle\frac{\partial\sigma(h_{\mu}^{1}(t))}{\partial\beta_{\nu}^{1,\phi\top}(s)}\right\rangle_{\mathcal Z_N^{global}},
B^{1,\phi}_{\mu\nu}(t,s)=\left\langle\frac{\partial  g_{\mu }^{1,\phi}(t))}{\partial\alpha_{\nu }^{\phi\top}(s)}\right\rangle_{\mathcal Z_{M}|\mathcal K_{global}}.
\bar B^1_{\mu\nu}(s,t) 
=
\left\langle
\frac{\partial  g_{\mu }^{2,\i}(t))}{\partial\alpha_{\nu }^{2,\i\top}(s)}
\right\rangle
_{\mathcal Z_{N_e}|\mathcal K_{global}}
\\[4pt]
g^{1,\phi}_\mu(t)
&=
\frac{\iota\eta_0\gamma_0}{\kappa P}\int_0^tds \sum_\nu \Delta_\nu(s) G^3_{\mu\nu}(s,t) \left[\Phi^{2,in}_{\mu\nu}(s,t) \tilde\phi_\nu(t)
-
\bar \Phi^{2,in}_{\mu\nu}(s,t) \right]
\tilde\phi_\mu(s)
\\[4pt]
z^1_\mu(t) 
&= 
\beta^{1,\phi}_\mu(t)
+
\frac{\eta_0 \gamma_0}{P}\int_0^t \! ds\sum_{\nu}\left\{
\bar B^{1}_{\nu\mu}(s,t)
+
B^{1,\phi}_{\nu\mu}(s,t)
+
\Delta_{\nu}(s) 
\left[
\frac\iota\kappa\bar G_{\mu \nu}^{2,\i}(s,t)
+ G^{1,\phi}_{\mu\nu}(s,t)
\right]
\right\}
\, \sigma(h^{1}_{\nu}(s))
\\[4pt]
z^{2,\i}_{\mu}(t) &=
\beta^{2,\i}_{\mu}(t)
+
\frac{\eta_0 \gamma_0}{P\kappa}\int_0^t \! ds\sum_{\nu}\Delta_{\nu}(s) G_{\mu \nu}^{3}(s,t)\,\tilde\phi_\mu(s)\,\tilde\phi_\nu(t) \, \sigma\left( {h^{2,\text{in}}_{\nu}(s)}\right), \quad j\in\{1,2,...,N_e\}
\\[4pt]
 z_\mu^{3}(t)&=\beta_{\mu}^{3}(t)+ \frac{\eta_0 \gamma_0}{P}\int_0^t \! ds\sum_{\nu}\Delta_{\nu}(s) \,  h^{3}_{\nu}(s),
 \\[4pt]
h_{\mu}^1(t)&=\alpha_{\mu}^1(t)+\frac{\eta_0 \gamma_0}{ P}
    \int_0^t \! ds
    \sum_{\nu} \Delta_{\nu} \,\Phi_{\mu\nu}^0\, (z^1_\mu(s)\cdot\dot\sigma(h^1_\mu(s)))  \\[4pt]
 {h^{2,\text{in}}_{\mu}(t)}&=
 {\alpha^{2,\text{in}}_{\mu}(t)}
 +
 \frac{\iota\eta_0 \gamma_0}{\kappa P}
    \int_0^t \! ds
    \sum_{\nu} \Delta_{\nu}(s)\Phi_{\mu \nu}^{1}(t,s)\,\left(\dot\sigma\left(h^{2,\text{in}}_{\mu}(t)\right)\odot  z_{\nu}^{2,\i}(s)\right) ,\quad j\in\{1,2,...,N_e\} \\[4pt]
h^3_\mu(t) &=\frac{\iota\eta_0\gamma_0}{\kappa P}\int_0^t ds \sum_\nu \left[ \bar A^{2,\i}_{\mu\nu}(t,s)
+
\Delta_\nu(s) 
    \bar\Phi^{2,\i}_{\mu\nu}(s,t)\right]
    z^3_\nu(s)
    \\[4pt]
\psi_{\mu}(t)&=
\alpha^\phi_{\mu}(t)
+
\frac{\eta_0 \gamma_0}{P}\,
\int_{0}^{t} ds
\sum_{\nu}\left[
A^{1,\phi}_{\mu\nu}(s,t)+\Delta_{\nu}(s)\,
\Phi^{1}_{\mu\nu}(s,t)\right]g^{1,\phi}_{\nu }(t)
\\[4pt]
\tilde\phi_\mu(t) &= \frac{e^{\hat\psi_\mu(t)}}{\mathcal S_\mu(t)}
,\qquad
\hat\psi_\mu(t) = m_\mu(t) \psi_\mu(t)
,\qquad
m_\mu(t) = \1(\psi_\mu(t) -\tau_\mu(t))
\\[4pt]
\text{where }& \tau_\mu(t) \text{ is chosen such that } \rho = \left\langle m_\mu(t) \right\rangle_{\mathcal Z_{M}|\mathcal K_{global}}
\\[4pt]
    \frac{d f_{\mu}(t)}{dt}
&=\frac{\eta_0}{P}\sum_\nu\Delta_\nu
\Bigg[
G^{1}_{\mu\nu}(t,t)\Phi^{0}_{\mu\nu}(t,t)
+
\left[\frac\iota\kappa \bar G_{\mu\nu}^{2,\i}(t,t)  + G^{1,\phi}_{\mu\nu}(t,t)\right] \Phi^1_{\mu\nu}(t,t) \\
&
\quad +
\frac\iota\kappa G^{3}_{\mu\nu}(t,t)\bar \Phi^{2,\i}_{\mu\nu}(t,t)
+
\Phi^{3}_{\mu\nu}(t,t)
\Bigg]
\end{aligned}
$}
\end{equation*}

\emph{Comparison to \(\mu\)P.} 
Limiting dynamics under \(\mu\)P admit a simpler mean-field structure than under MSSP.
Under \(\mu\)P, the expert weights are initialized independently across experts. 
Consequently, there is no expert-hidden disorder shared by the expert population, and the limiting theory does not require a separate shared expert-hidden single-site process. 
The expert averages self-average directly over independently initialized experts. The corresponding DMFT is provided in App.\ref{sec:dmft-regime-iii-mup}.

\newpage
\part{Experiments}

This part covers the experimental setup for our MLP MoE as well as Transformer MoE experiments in \Cref{sec:exp_setup}, provides more details concerning the main figures in \Cref{sec:fig_details}, followed by detailed empirical evidence for our claims about learning rate transfer and scaling properties of $\mu$P and MSSP for both SGD and Adam in all 3 scaling regimes in \Cref{sec:add_exper}. We close with further empirical scaling insights that practitioners should be aware of in Sections~\ref{sec:exp_softmax_collapse} to \ref{sec:global_eps}.

\section{Experimental Setup}\label{sec:exp_setup}

Open source code to fully reproduce our experiments can be found at:
\begin{center}
    \url{https://github.com/vankadara-lab/mssp-moe}
\end{center}

\subsection{MLP MoE experiments on TinyImagenet}

\textbf{MoE Architecture.} We train the same embedding-MoE-readout architecture with 2-layer fully connected experts that we analyse theoretically (without scale-dependent weight multipliers) using PyTorch \citep{pytorch}. It consists of a linear input layer with GeLU activation function, a linear router layer, 2-layer expert MLPs with GeLU activation function, followed by $M^{-1}\cdot$ sigmoid aggregation, followed by a linear output layer. In this architecture without normalization layers and residual connections, instabilities in the scaling procedure become apparent at more moderate scale.

\textbf{Data.} We train single pass over all $50000$ available images from $100$ classes of TinyImageNet \citep{le2015tiny} for $1000$ update steps with batch size $50$. Our initial experiments used CIFAR-10 \citep{cifar10}, but robustly benefitting from expert specialization in MoEs requires more diverse data.

We vary the optimizer (SGD and Adam), the scaling regime (fixed $M$, bottleneck and all-scaling), the routing mechanism (soft routing and top-$k$ routing) and the parameterization ($\mu$P and MSSP).

\subsubsection{Scaling configurations}

Unless otherwise specified, we initialize the last layer to zero.

\textbf{MSSP.} Depending on the optimizer and scaling regime, \Cref{tab:moe_mixtral_summary_template_main} summarizes our proposed parameterization MSSP as a function of width $N$, expert width $N_e$ and number of experts $M$. Scaling of non-MoE trainable weights remains in $\mu$P.

\textbf{$\mu$P.} \Cref{tab:moe_mixtral_summary_template_main} specifies $\mu$P for SGD and Adam in each scaling regime. The provided HP scaling rules 
achieve stability as well as scale-independent effective and propagating updates after sufficiently many update steps, as verified in \Cref{sec:rcc_mlp}. However their initial vanishing signal propagation induces more scale dependence in the dynamics, delayed learning and/or no monotonic improvement with scale.

For Regime I, we compare the baseline router init std $1/N$ (standard muP) versus zero router initialization for improved scale independence. Both variants barely differ. 

\subsubsection{Multiplier tuning}
\label{mlp:multiplier_tuning}

For MLPs, we use base width $1$, which uncovers scaling properties of each parameterization at more moderate scale. Before running extended evaluations or learning rate sweeps, we tune multipliers at small model scale $N=128$. The network at size $N=128$ has dimensions ($N=N_e=128, \; M=8$) in Regime I, ($N=128,\; M=N/16,\; N_e=16$) in Regime II and ($N=N_e=128, \; M=N/16$) in Regime III, from which we scale up proportionally $N\to\infty$.

We observe that the optimal multipliers can be as far as $10^6$ from all ones (since the gradient RMS norm of the expert output layer decays as fast as $N^{-2}$ in Regimes II and III (see \Cref{sec:rcc_regime2,sec:rcc_regime3})), and some update terms can be negligibly small without tuning, resulting in non-balanced learning (\Cref{sec:exp_mult_tuning}).

For each configuration of optimizer, parameterization, routing mechanism and scaling regime, we tune the following set of multipliers:
\begin{itemize}
    \item global initialization variance,
    \item global learning rate,
    \item layerwise learning rate multiplier (input, router, expert input, expert output, output layer)
\end{itemize}

Our tuning algorithm proceeds as follows. We first tune the learning rate at width $N=128$, then run a broad 6D grid of the above init and layerwise learning rate multipliers, then run a full 6D grid with 5 grid points per dimension centered at the optimum from the previous stage at multiplicative resolution 4. This is possible since several small model training runs in parallel take less than 40 seconds on one A100 GPU.

\begin{figure}[t]
    \centering
    \begin{subfigure}[t]{0.49\textwidth}
        \centering
        \includegraphics[width=\textwidth]{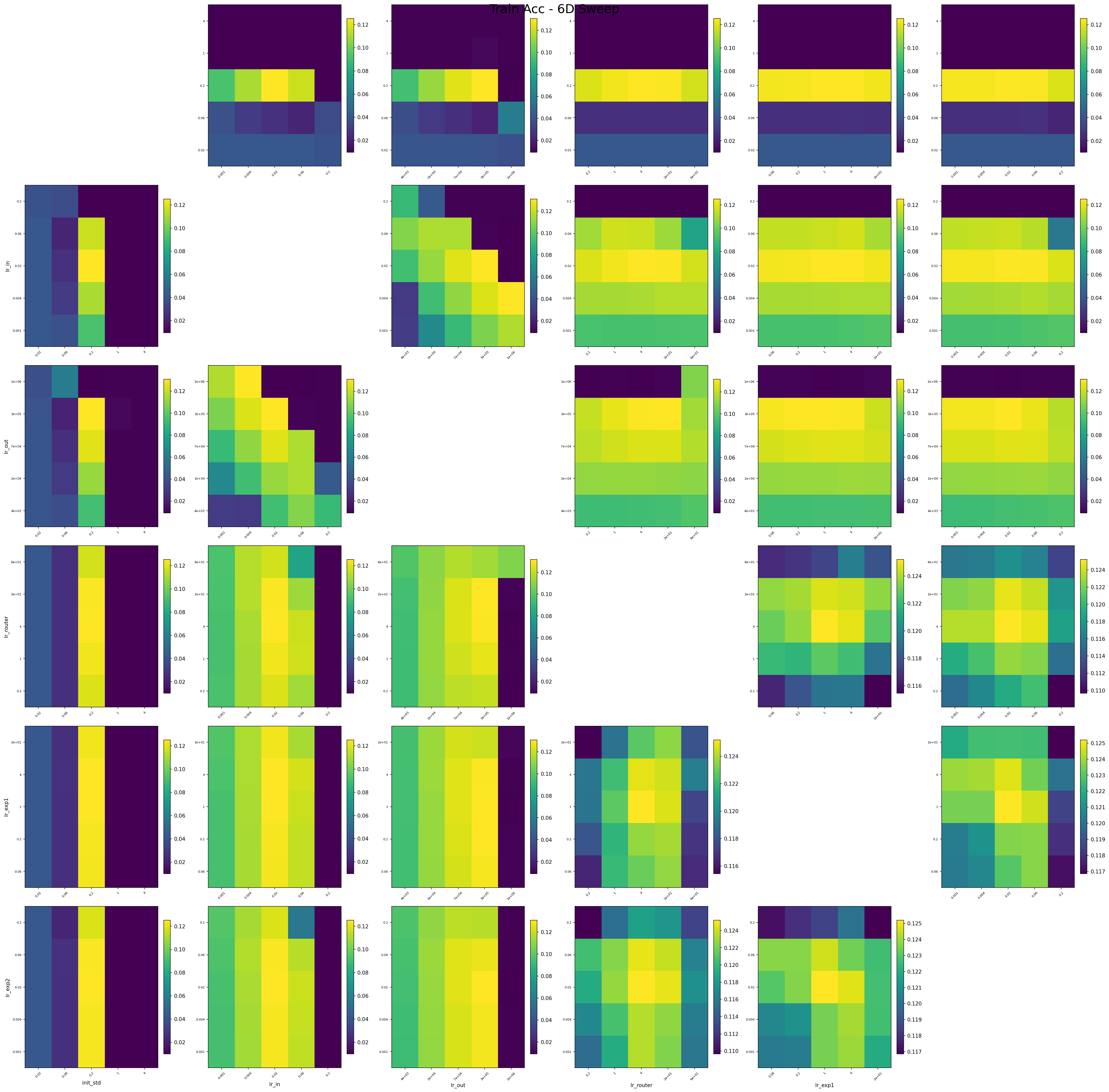}
    \end{subfigure}
    \hfill
    \begin{subfigure}[t]{0.49\textwidth}
        \centering
        \includegraphics[width=\textwidth]{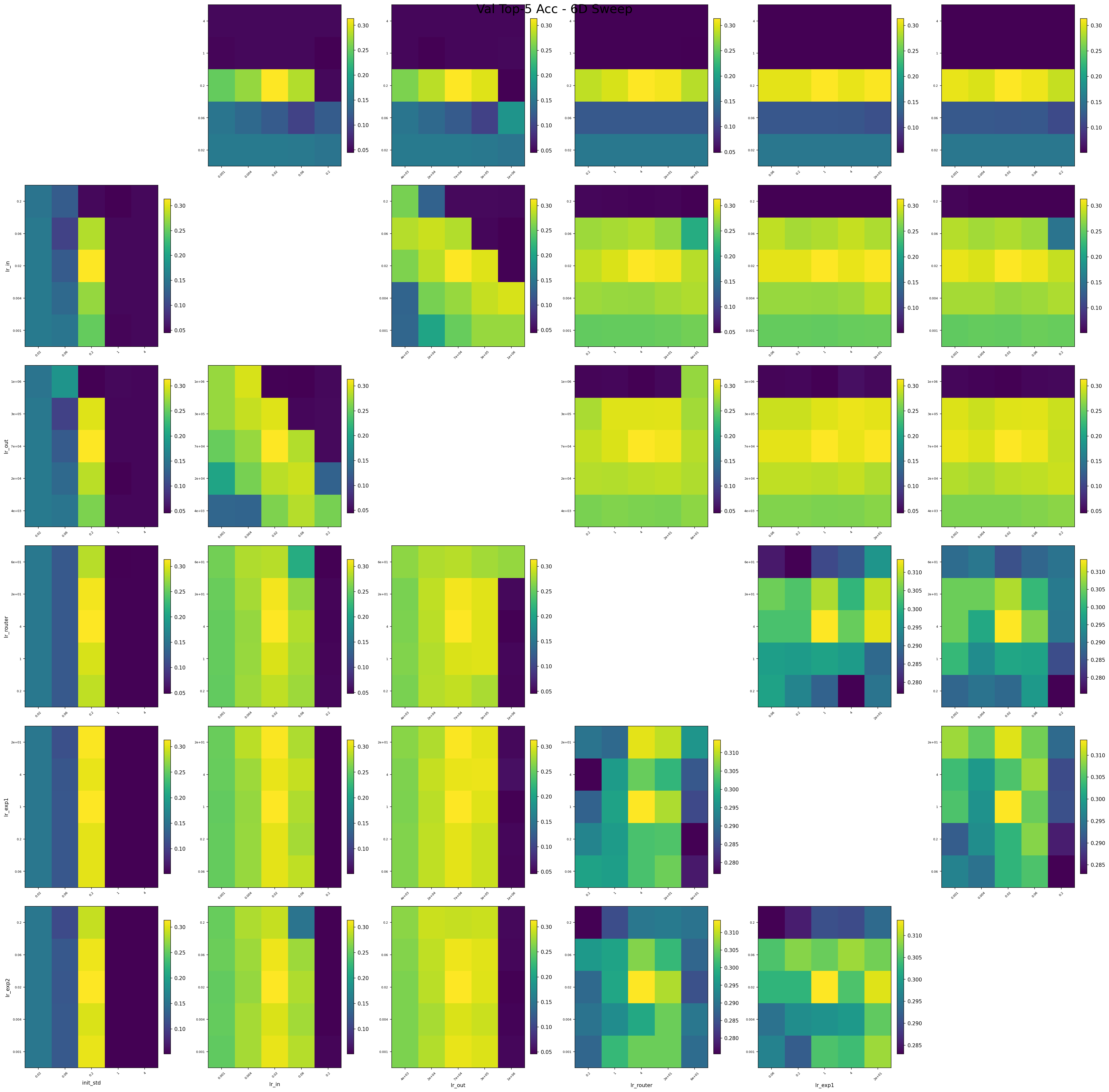}
    \end{subfigure}

    \caption{\textbf{6D multiplier sweeps at small scale $N=128$ (MSSP, SGD, bottleneck).} 2D heatmaps showing training accuracy (left) and top-5 validation accuracy (right) of all HP pairs, while fixing all remaining HPs at the optimum. The optimum in each multiplier remains consistent across HP pairs. Train and validation optimum can slightly differ.}
    \label{fig:6d_sgd_bottleneck}
\end{figure}

\Cref{fig:6d_sgd_bottleneck} shows the example of the last step for the case of SGD MSSP in Regime II. Top-5 validation accuracy optima tend to me more localized than training accuracy optima. Hence we choose the optimum based on top-5 validation accuracy. The optimal initialization variance multiplier tends to be very clearly localized and essential in all combinations of optimizer, parameterization and scaling regime.

\Cref{sec:2d_tuning} shows that a cheaper multiplier tuning procedure, which starts with a random search over the grid, followed by extensive 2D sweeps over all combinations of multipliers does not suffice to reliably provide a near-optimal combination of multipliers.

\subsubsection{Coordinate checks and learning rate sweeps}

For each scaling configuration, we repeat the same experiment for 4 independent random seeds, affecting the random weight initialization as well as data shuffling. Uncertainty bands denote $2\sigma$-confidence bands. If not mentioned otherwise, MLP experiments use soft routing, and the same compute budget for multiplier tuning was invested for $\mu$P and MSSP.

\subsection{Transformer MoE experiments}\label{sec:transformer_setup}

We adapt the \href{https://github.com/karpathy/nanoGPT}{nanoGPT} and \href{https://github.com/wolfecameron/nanoMoE}{nanoMoE} codebases to train transformer models with a simple but modern architecture. 

{\bf General Architecture and Training Details.} By default, models have 8 blocks with mixture-of-expert layers (no dense layers).  
We scale the number of attention heads proportionally with width, while the head dimension remains fixed at $\texttt{d\_head}=64$. We use standard $\texttt{d\_head}^{-1/2}$ attention scaling, which can be viewed as a tunable multiplier since $\texttt{d\_head}$ remains fixed. We use pre-attention and qk-RMSNorm \citep{wortsman2023small}. 
We train for 4768 steps using AdamW with a single, tuned maximal learning rate, $(\beta_1,\beta_2)=(0.9, 0.95)$, $\epsilon=10^{-8}$, sequence length $1024$, batch size $512$, $700$ steps of warmup followed by cosine learning rate decay to $10\%$ of the maximal learning rate, weight decay $0.1$, and gradient clipping. 

{\bf MoE layers.} Mixture-of-expert layers are scaled according to the respective scaling regime, using a sigmoid activation function and an auxiliary loss with weight 0.01 for load balancing \citep{shazeer2017}. We use token-choice routing and the number of active experts per token is half the total number of experts. We do not drop tokens. 

{\bf Architecture for Regime III.} In all all-scaling Regime III experiments, we use the following architecture. At base width $N=256$ each of the 8 MoE layers has $n_{\text{exp}}=8$ experts of hidden width $N_e=N/2=128$; at $N=2048$ the same layer has $n_{\text{exp}}=64$ experts of hidden width $N_e=1024$.      

{\bf Architecture for Coordinate Checks in Regime II.} For the coordinate checks in bottleneck Regime II, we use 8 MoE layers with a fixed expert hidden width of $N_e=16$. For keeping a total expansion ratio of $4$, the expert count is $M=64$ at $N=256$ and grows to $M=512$ at $N=2048$.

{\bf Architecture for Learning Rate Sweeps for Regime II.} For the learning rate sweeps in bottleneck Regime II, we use a more computationally efficient bottleneck architecture due to compute constraints. We use $4$ blocks alternating between dense and MoE layers (Dense, MoE, Dense, MoE). At width $N=256$, each MoE block holds $M=32$ experts of fixed hidden width $N_e=32$, scaling to $M=256$ experts at $N=2048$. This amounts to 408M parameters at $N=2048$. 

{\bf Data.} Models are trained on 2.5B tokens of \href{https://huggingface.co/datasets/allenai/dolma3_mix-150B-1025}{dolma3\_mix-150B-1025} \citep{olmo2025olmo3}. 

\subsubsection{Scaling configurations}

As in the MLP experiments, models are trained with the respective scaling configuration. We initialize the last layer to zero. 

\subsubsection{Multiplier tuning}

For every layer type, we tune initialization and learning rate multipliers. We also tune the global learning rate. Multipliers are tuned at width 256 by training on 1B tokens. We experiment both with random search and with round-robin algorithms that tune multipliers individually and across 2D grids. We evaluate multipliers across four different seeds, and find that both approaches obtain similar. An exhaustive grid search for multiplier tuning, as described for MLPs in Section \ref{mlp:multiplier_tuning}, was not computationally feasible for transformers because of the significantly longer training time. 

\subsubsection{Coordinate checks and learning rate sweeps}

We use \href{https://github.com/tml-tuebingen/torch-module-monitor}{torch-module-monitor} to monitor the training dynamics and perform coordinate checks. We use soft routing for the coordinate checks.

\subsection{Figure details}\label{sec:fig_details}

\textbf{\Cref{fig:performance_transfer_main1}}: \textit{Right:} Subset of \Cref{fig:llm_lr_sweep_main}, with description below. 
\textit{Left:} We compare the top-5 training accuracy averaged over the last 50 steps of $\mu$P (dashed lines) and MSSP (solid lines) for the soft routing MLP MoE Adam scaling runs from each scaling regime in App.~\ref{sec:rcc_regime1} for Regime I, \ref{sec:rcc_regime2} for Regime II and \ref{sec:rcc_regime3} for Regime III. Where both variants are available, we use the $0$ last-layer initialization variant for the $\mu$P baseline for direct comparability with MSSP. These experiments use the optimal multipliers and learning rate from $N=128$ and transfer them to large model scales, as is common for $\mu$P in dense networks. Both $\mu$P and MSSP use the same compute budget for multiplier tuning at small size $N=128$. In Regime I, $\mu$P and MSSP barely differ, so that the lines overlap heavily for both SGD and Adam.

\textbf{\Cref{fig:bottleneck_sgd_mup}}: $\mu$P uses zero last-layer initialization. Under maximal stable last-layer initialization $\sigma\asymp 1/N$, exponents are even more unstable and learning even more delayed. More detailed evaluations are provided in \Cref{sec:rcc_regime2}. All width-scaling exponents Expon$(v_N)$ are computed as OLS linear regression in log-log-space, hence fitting $\alpha\in\mathbb{R}$ in a model $v_N = C \cdot N^\alpha$ based on all available widths. 

The right subplots show, at each time step the width-scaling exponents of the individual terms of \Cref{eq:agg_decomp} and the overall updates to the post-aggregation activations $\|\Delta h^l_t\|_{RMS}$, which are the sum of effective and propagating updates, where the $M^{-1}$ aggregation scaling is included in all terms.

\textbf{\Cref{fig:exponents_over_time}} shows width-scaling exponents of the propagating and effective updates of each linear weight matrix for the same setting as in \Cref{fig:bottleneck_sgd_mup}. The expert output weights are also evaluated pre-aggregation. Missing lines indicate that the respective term is exactly $0$ (such as propagating updates in the input layer). \Cref{sec:rcc_regime2} shows that for Adam in Regime II, while less severe, analogous width dependence in $\mu$P is resolved by MSSP, and monotonic improvement with scale is recovered.

\textbf{\Cref{fig:mlp_transfer_sgd}}: Learning rate sweeps after multiplier tuning as reported in App.~\ref{sec:lr_regime2} for Regime II and App.~\ref{sec:lr_regime3} for Regime III. The learning rate curves for Adam in $\mu$P-Regime II are also strongly shifting (\Cref{fig:lr_bottleneck_adam_soft_std}). For Adam in Regime III, the shift is much weaker, but performance saturates and does not monotonically improve with scale (\Cref{fig:lr_allscale_adam_soft_nsh_mup}). MSSP recovers approximate learning rate transfer and monotonic improvement with scale in all cases.

\textbf{\Cref{fig:llm_lr_sweep_main}}: Learning rate sweeps for GPT MoEs trained with Adam in Regime II and III, as described in \Cref{sec:transformer_setup}. For Regime II, the figure shows the mean over 4 seeds. \Cref{sec:lr_sweeps_llm} shows the uncertainty bands as well as the corresponding training loss.

\section{Additional Experiments}\label{sec:add_exper}

\subsection{Learning rate sweeps for Transformer MoEs}\label{sec:lr_sweeps_llm}

\begin{figure}[H]
    \centering

    \begin{subfigure}[b]{0.99\textwidth}
    \centering
    \includegraphics[width=\textwidth]{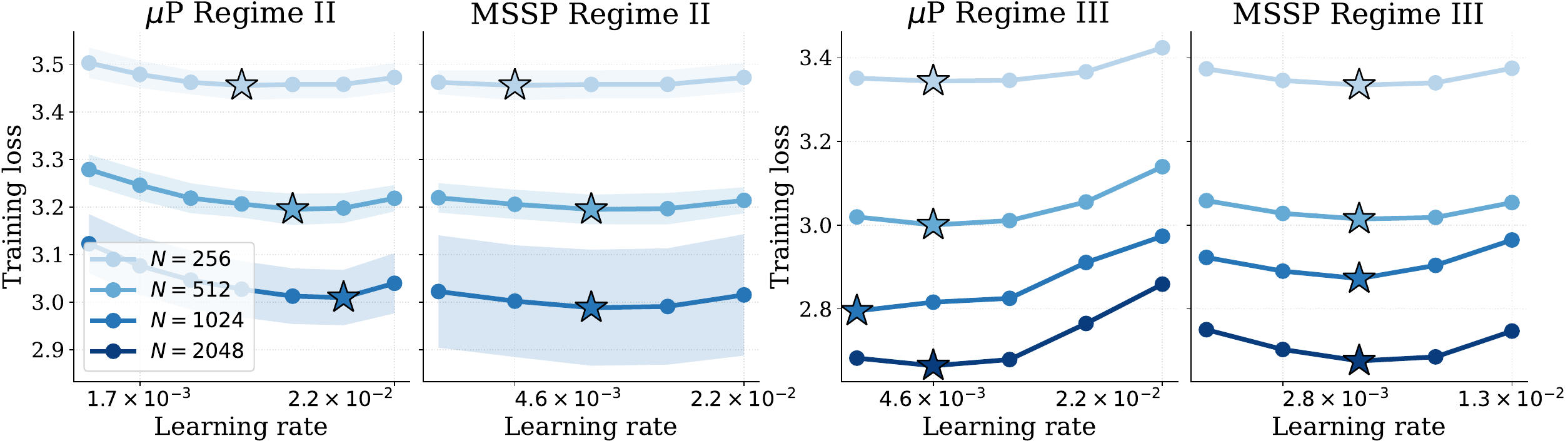}
    \end{subfigure}
    \hfill
    \begin{subfigure}[b]{0.99\textwidth}
    \centering
    \includegraphics[width=\textwidth]{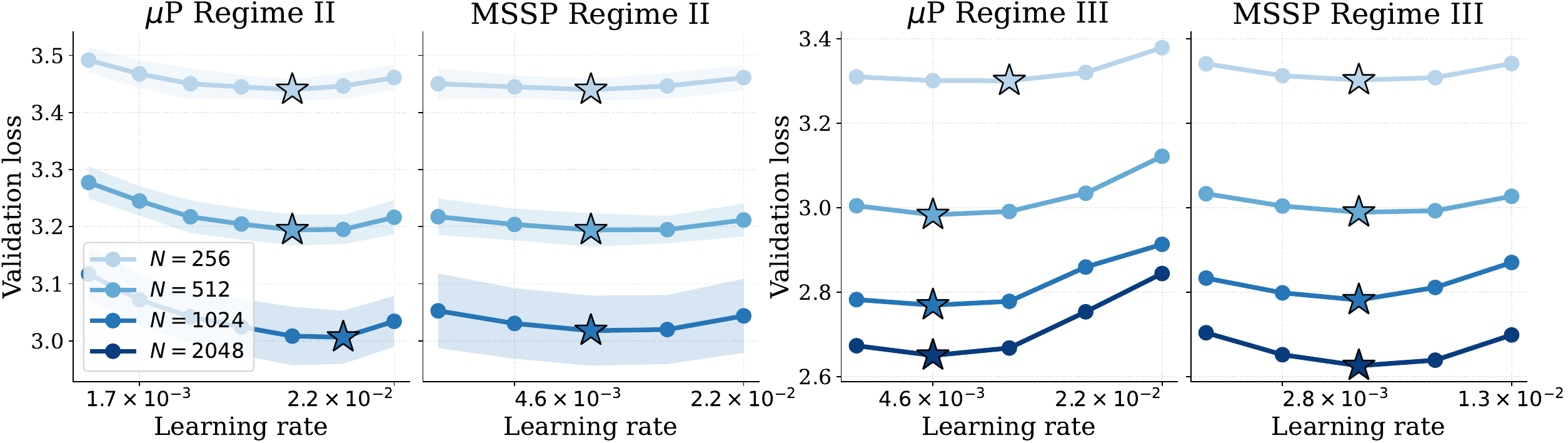}
    \end{subfigure}

    \caption{\textbf{Learning rate sweeps for $\mu$P and MSSP (Adam).} Same as \Cref{fig:llm_lr_sweep_main} but for training loss (top) and validation loss (bottom), and with $\sigma$-confidence bands across 4 seeds for bottleneck Regime II.} 
    \label{fig:llm_lr_sweep_app}
\end{figure}

\Cref{fig:llm_lr_sweep_app} shows that learning rate transfer is achieved more cleanly in MSSP than in $\mu$P. The optimal learning rate in $\mu$P-Regime-II tends to grow (due to vanishing terms in the expert aggregation operations). Since at base width $N=256$ both parameterizations are equivalent, scaling differences only become apparent at large scale. Both parameterizations perform similarly well with MSSP having slightly more variance in Regime II.

\subsection{Refined coordinate checks for Transformer MoEs}\label{sec:rcc_llm}

We use \href{https://github.com/tml-tuebingen/torch-module-monitor}{\texttt{torch-module-monitor}} to measure refined coordinate checks as a sanity check for correct implementation of the MSSP scaling rules in each scaling regime. Propagating update exponents $0$ verify correct initialization variance of the respective layer, given correctly scaled inputs. Effective update exponents $0$ verify correct learning rate scaling of the respective layer, given correctly scaled inputs. One could isolate the effect of the layer's learning rate scaling by normalizing the input activations $x_{\text{in}}/\|x_{\text{in}}\|_{RMS}$. We instead verify that $\|x_{\text{in}}\|_{RMS}=\Theta(1)$ approximately holds in all layers (not shown).

\begin{figure}[H]
    \centering

    \begin{subfigure}[b]{0.99\textwidth}
    \centering
    \includegraphics[width=\textwidth]{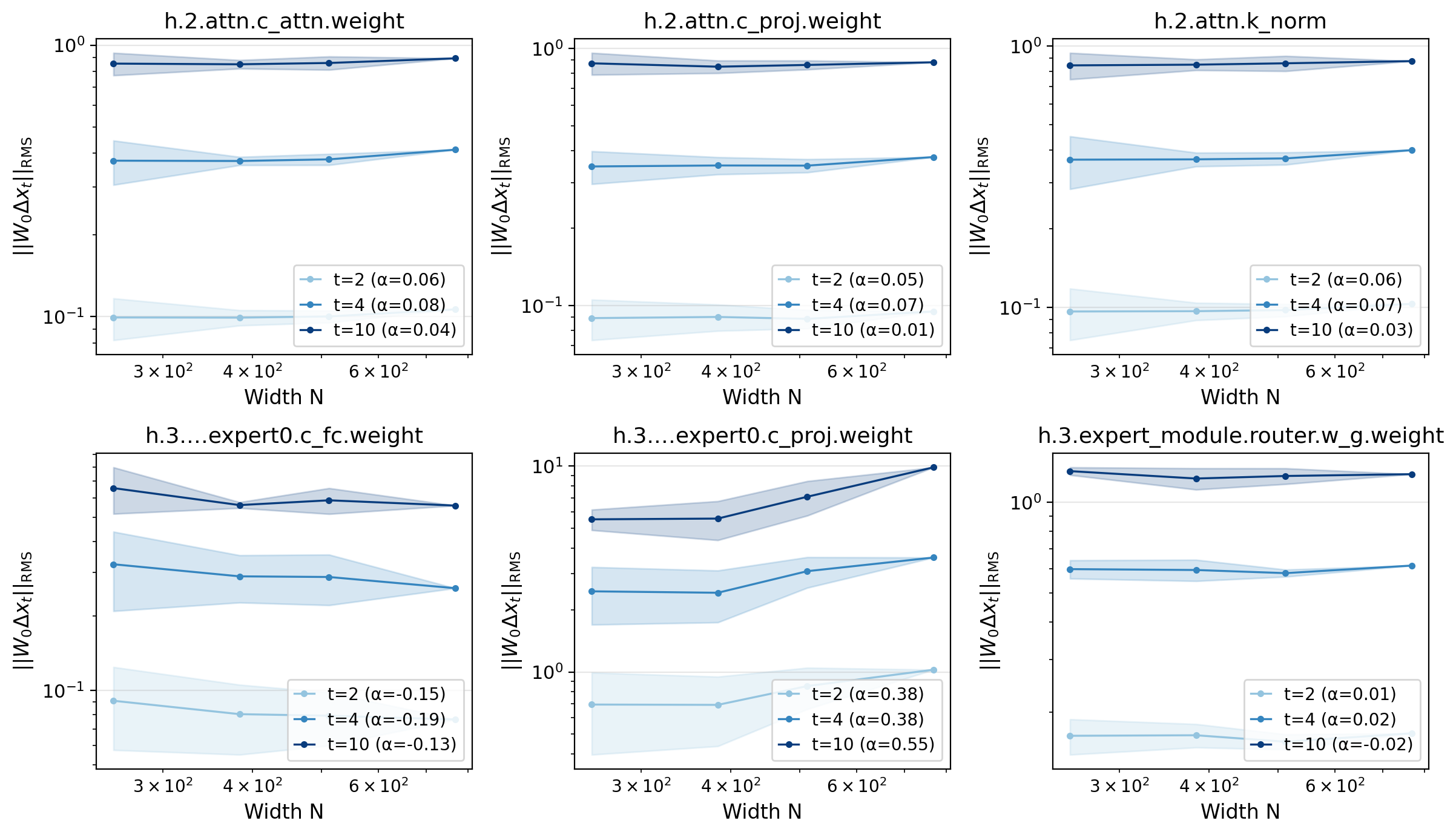}
    \end{subfigure}

    \caption{\textbf{Propagating updates in MSSP (Adam, Regime II).} Propagating updates of some example layers approximately approximately follow the desired scaling exponents. First and second expert layer stem from a single expert. Measuring such small experts is particularly noisy, depending heavily on how many tokens are routed to them in the respective step. Recall that the desired propagating update exponent in the expert output layer $\texttt{expert0.c\_proj}$ is $0.5$ in Regime II.}
    \label{fig:prop_rcc_llm_regime2}
\end{figure}

\begin{figure}[H]
    \centering

    \begin{subfigure}[b]{0.99\textwidth}
    \centering
    \includegraphics[width=\textwidth]{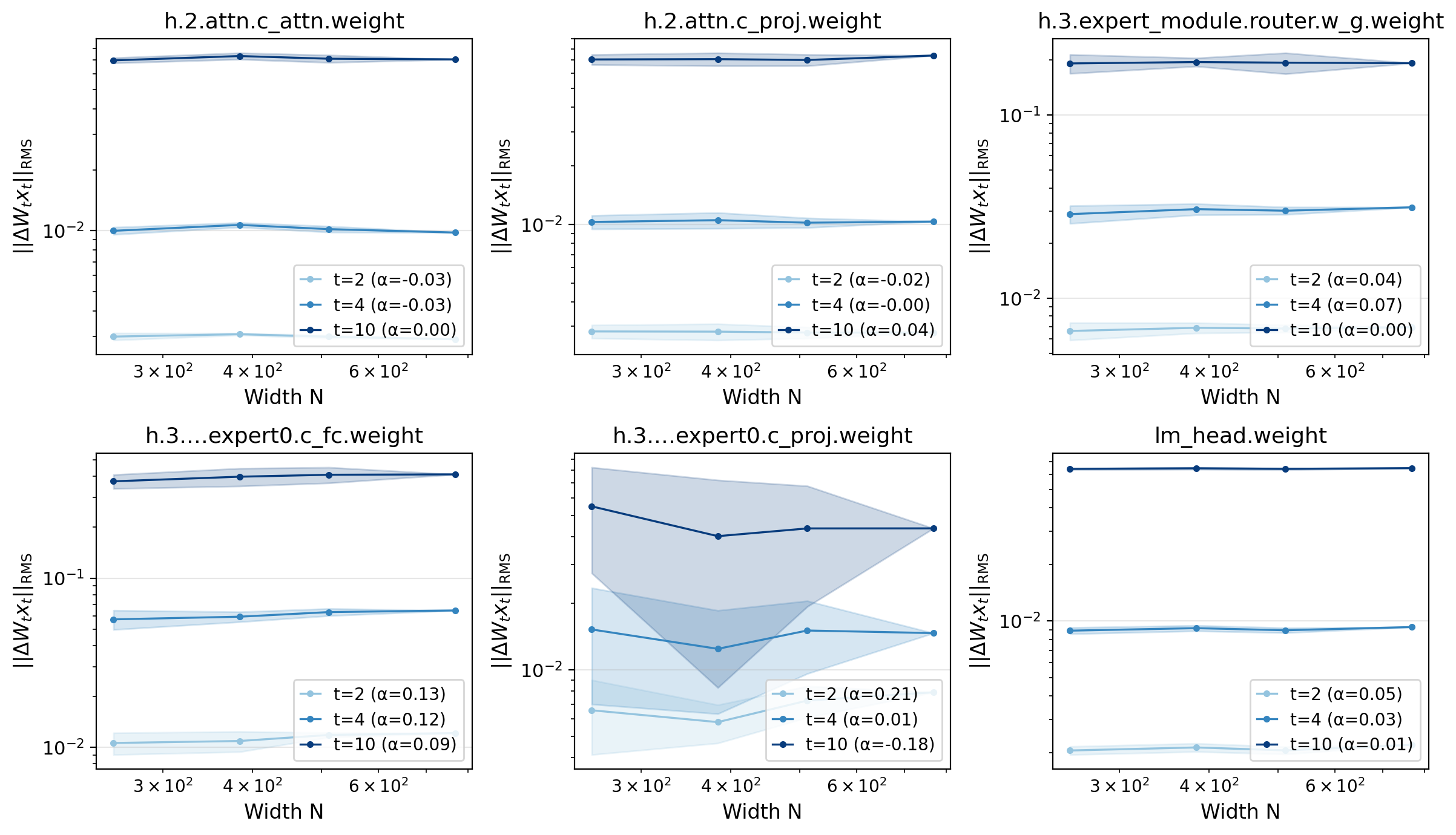}
    \end{subfigure}

    \caption{\textbf{Effective updates in MSSP (Adam, Regime II).} Effective updates of some example layers are approximately scale-preserving at all times. First and second expert layer stem from a single expert. Measuring such small experts is particularly noisy, depending heavily on how many tokens are routed to them in the respective step.}
    \label{fig:eff_rcc_llm_regime2}
\end{figure}

\begin{figure}[H]
    \centering

    \begin{subfigure}[b]{0.99\textwidth}
    \centering
    \includegraphics[width=\textwidth]{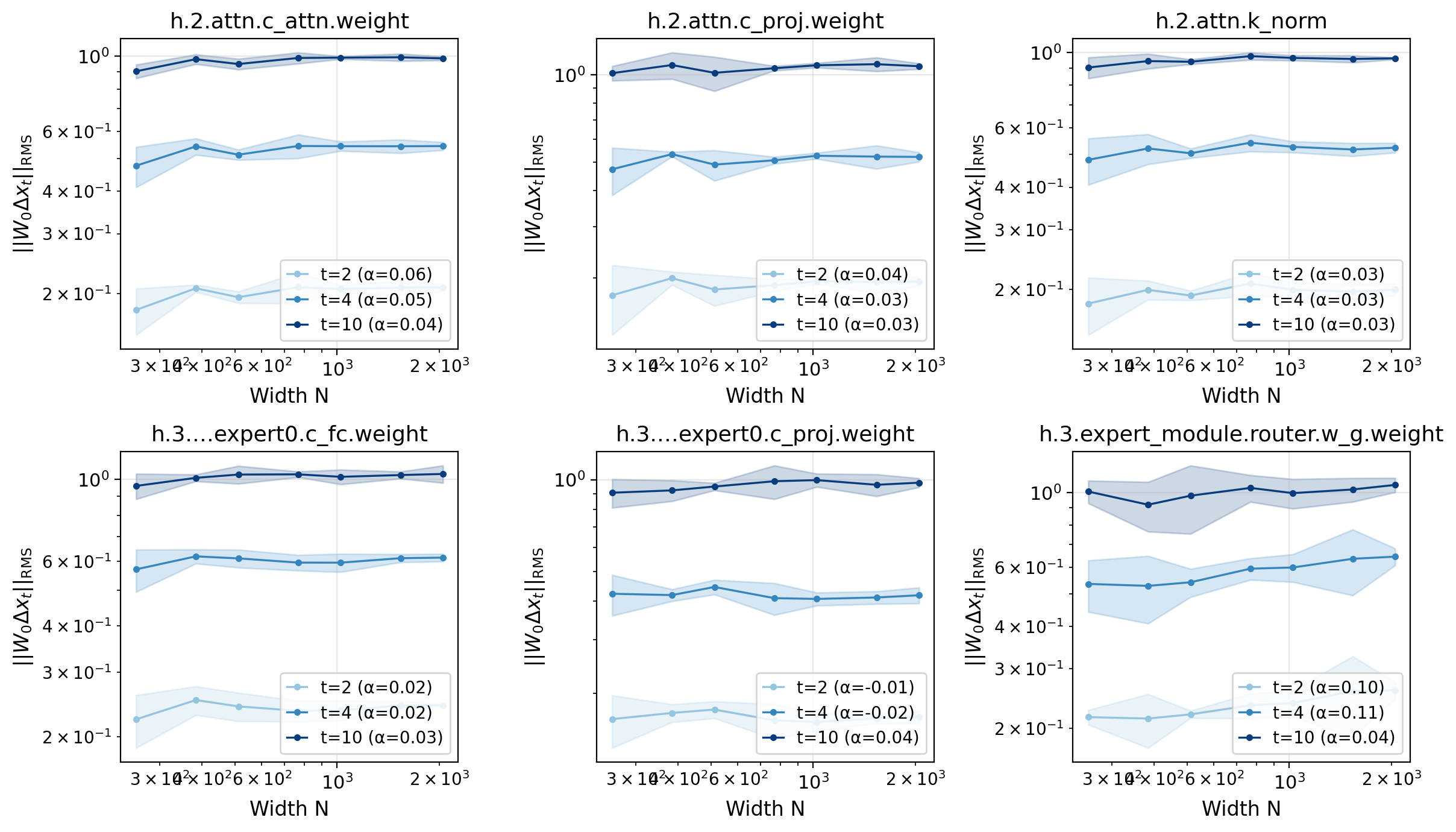}
    \end{subfigure}

    \caption{\textbf{Propagating updates in MSSP (Adam, Regime III).} Propagating updates of some example layers are approximately scale-preserving at all times. First and second expert layer stem from a single expert.}
    \label{fig:prop_rcc_llm_regime3}
\end{figure}

\begin{figure}[H]
    \centering

    \begin{subfigure}[b]{0.99\textwidth}
    \centering
    \includegraphics[width=\textwidth]{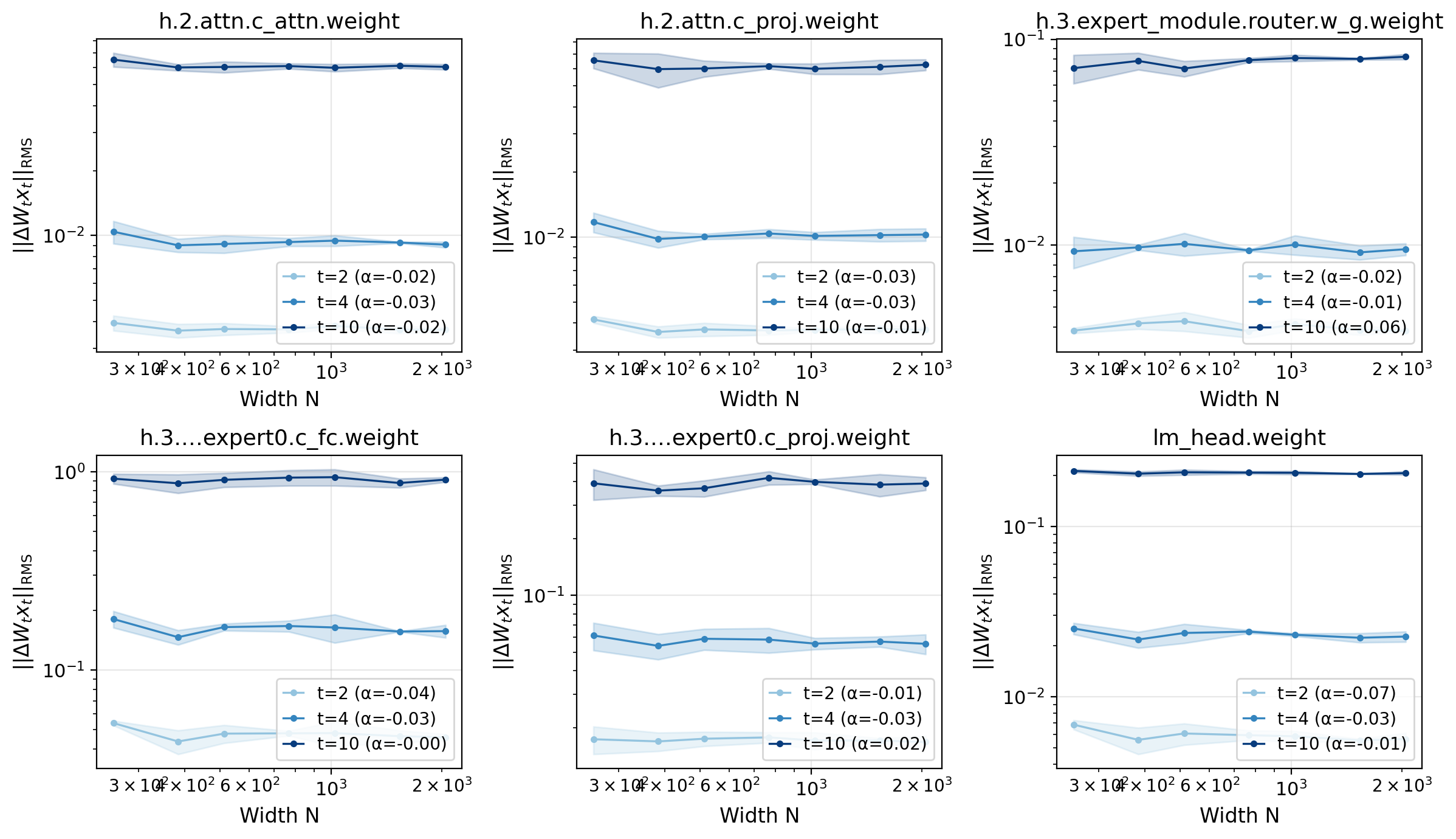}
    \end{subfigure}

    \caption{\textbf{Effective updates in MSSP (Adam, Regime III).} Effective updates of some example layers are approximately scale-preserving at all times. First and second expert layer stem from a single expert.}
    \label{fig:eff_rcc_llm_regime3}
\end{figure}

\subsection{Learning rate sweeps for MLP MoEs}\label{sec:lr_sweeps_mlps}

Generally observe cleaner learning rate transfer across model sizes in MSSP than in $\mu$P. Also observe more robust monotonic improvement with scale in MSSP across optimizers and scaling regimes.

\subsubsection{Regime I: Fixed number of experts}\label{sec:lr_regime1}

While the optimal learning rate transfers and performance monotonically improves with scale, the performance saturates at large scales. The differing router initialization has negligible impact. 

\begin{figure}[H]
\centering
\includegraphics[width=0.48\textwidth]{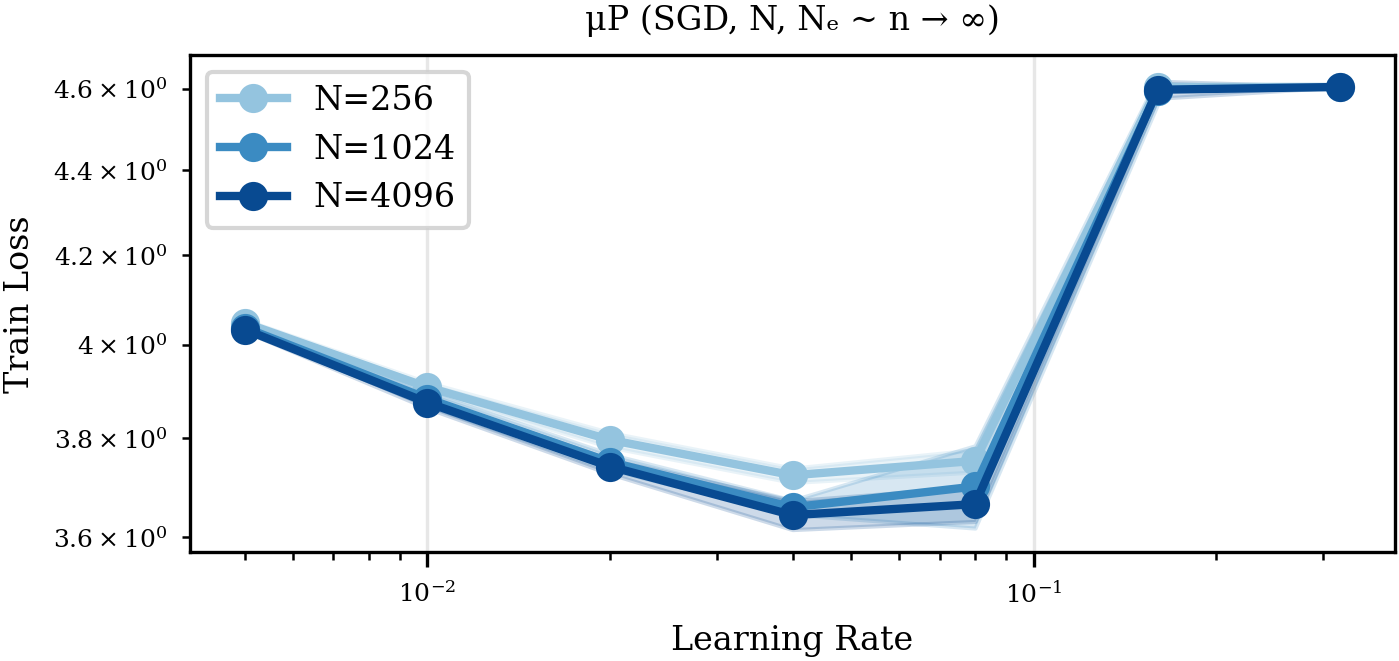}
\hfill
\includegraphics[width=0.48\textwidth]{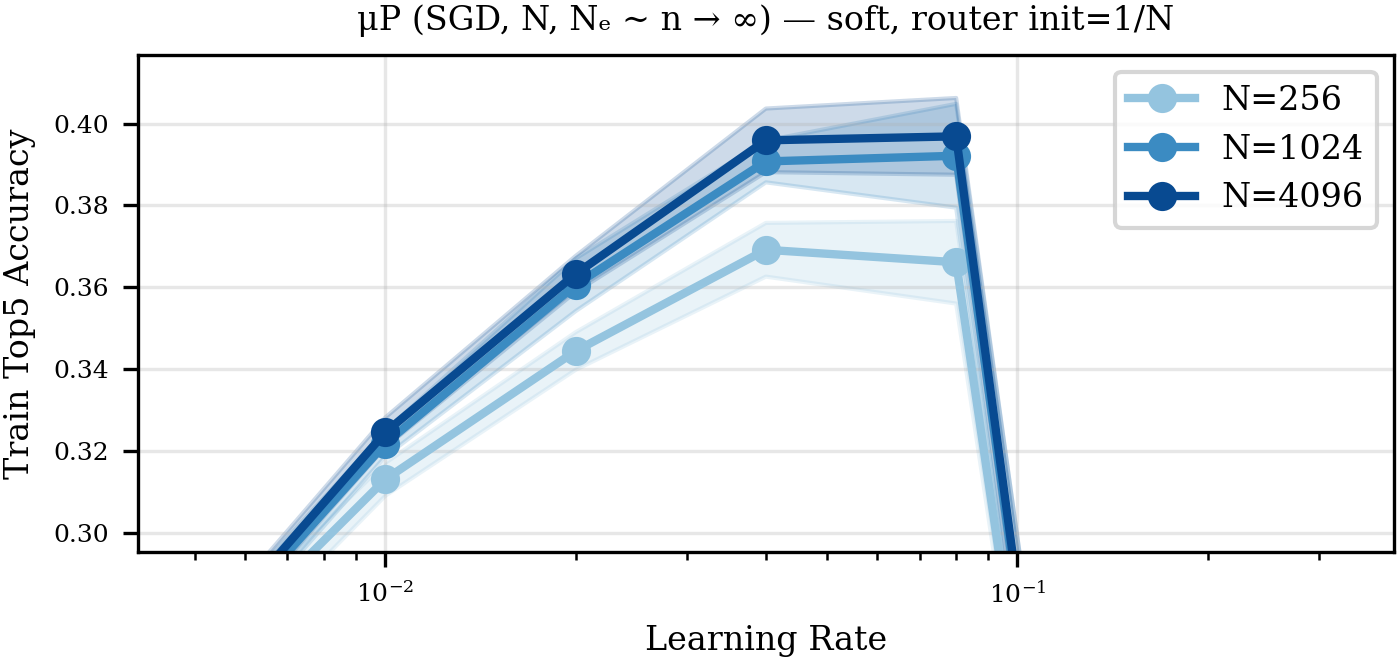}
\caption{\textbf{LR sweep, $\mu$P with $1/N$ router init (SGD, soft, Regime I).}}
\label{fig:lr_fixedE_sgd_soft_rmup}
\end{figure}

\begin{figure}[H]
\centering
\includegraphics[width=0.48\textwidth]{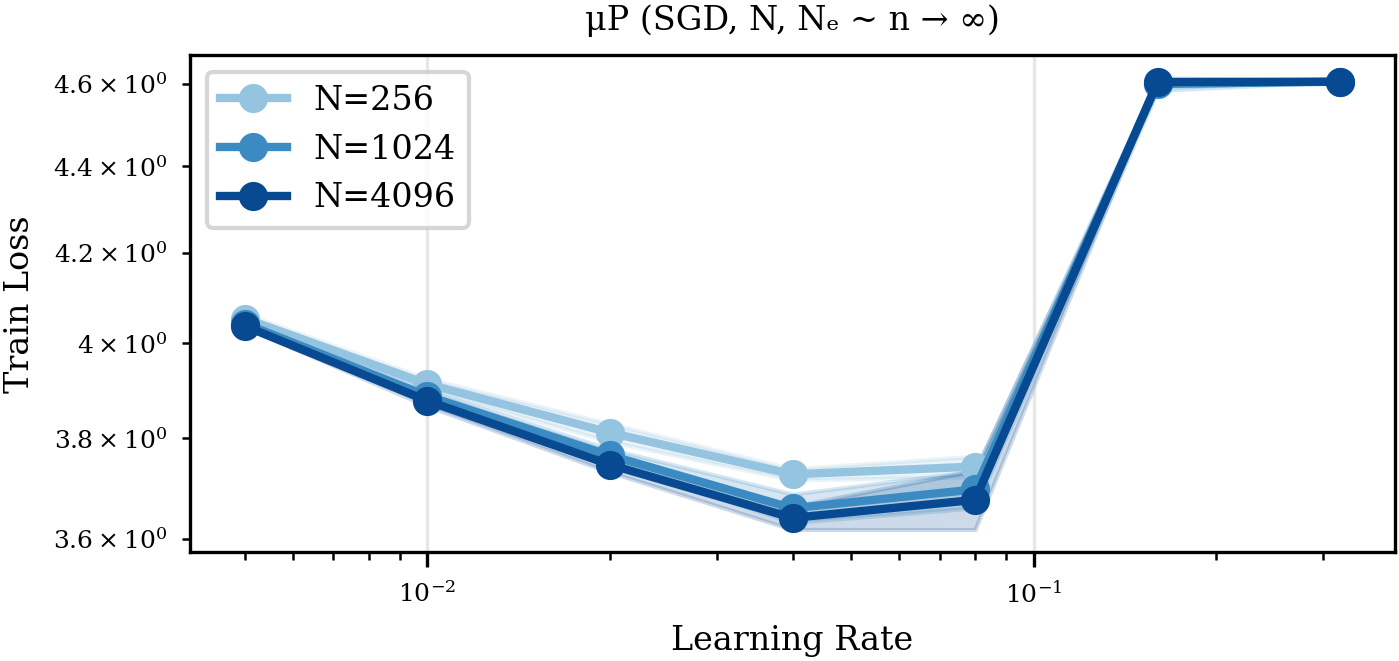}
\hfill
\includegraphics[width=0.48\textwidth]{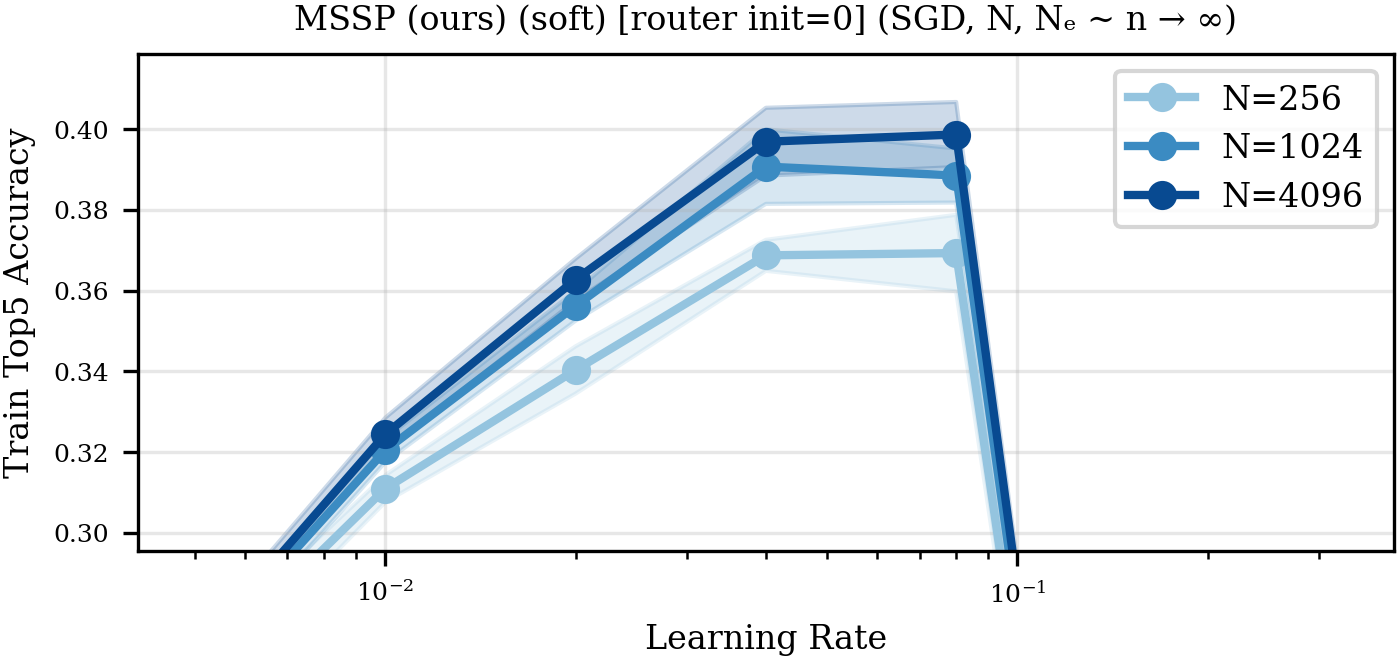}
\caption{\textbf{LR sweep, MSSP with zero router init (SGD, soft, Regime I).}}
\label{fig:lr_fixedE_sgd_soft_r0}
\end{figure}

\begin{figure}[H]
\centering
\includegraphics[width=0.99\textwidth]{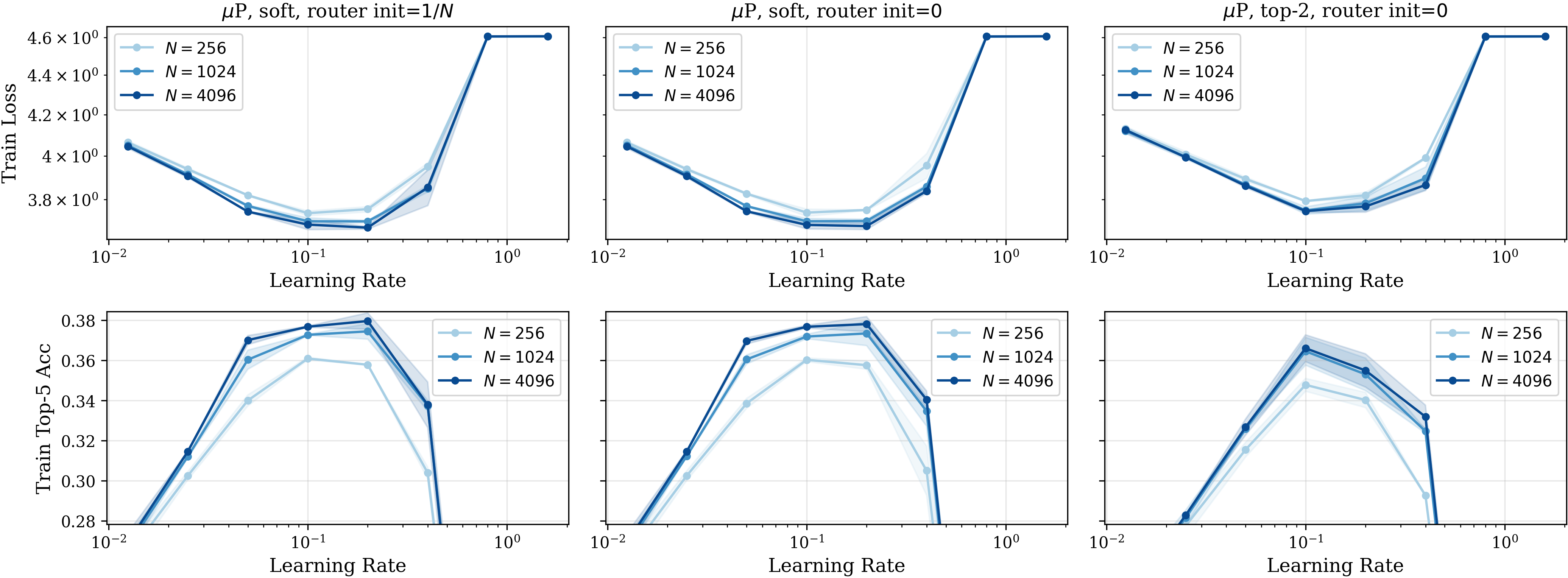}
\caption{\textbf{LR sweep comparison (Adam, Regime I).}}
\label{fig:lr_fixedE_adam_comp}
\end{figure}

\subsubsection{Regime II: Fixed expert width}\label{sec:lr_regime2}

Observe greatly improved learning rate transfer of MSSP over $\mu$P in both SGD and Adam in Regime II. Larger performance gains from small to large model width result in improved absolute performance at scale across optimizers.

The optimal learning rate in $\mu$P tends to grow toward the instability threshold since subcomponents of the expert aggregation dynamics are vanishing at fixed learning rate. Since the correctly scaled subcomponents of the expert aggregation would diverge when increasing the learning rate, the stability threshold does not grow with width. These conflicting objectives of stability of some terms versus effective learning in others results in worse performance at scale. By balancing all subcomponents, MSSP recovers width independence of both the maximal stable and most effective learning rate scaling for all terms in the training dynamics, so that monotonic improvement with scale is preserved. Slight saturating growth of the optimal learning rate can occur in MSSP. Analogous slight shifts have often been observed in dense architectures in $\mu$P \citep{tp5_2022,blake2024u,ghosh2025understanding}.

\begin{figure}[H]
\centering
\includegraphics[width=0.48\textwidth]{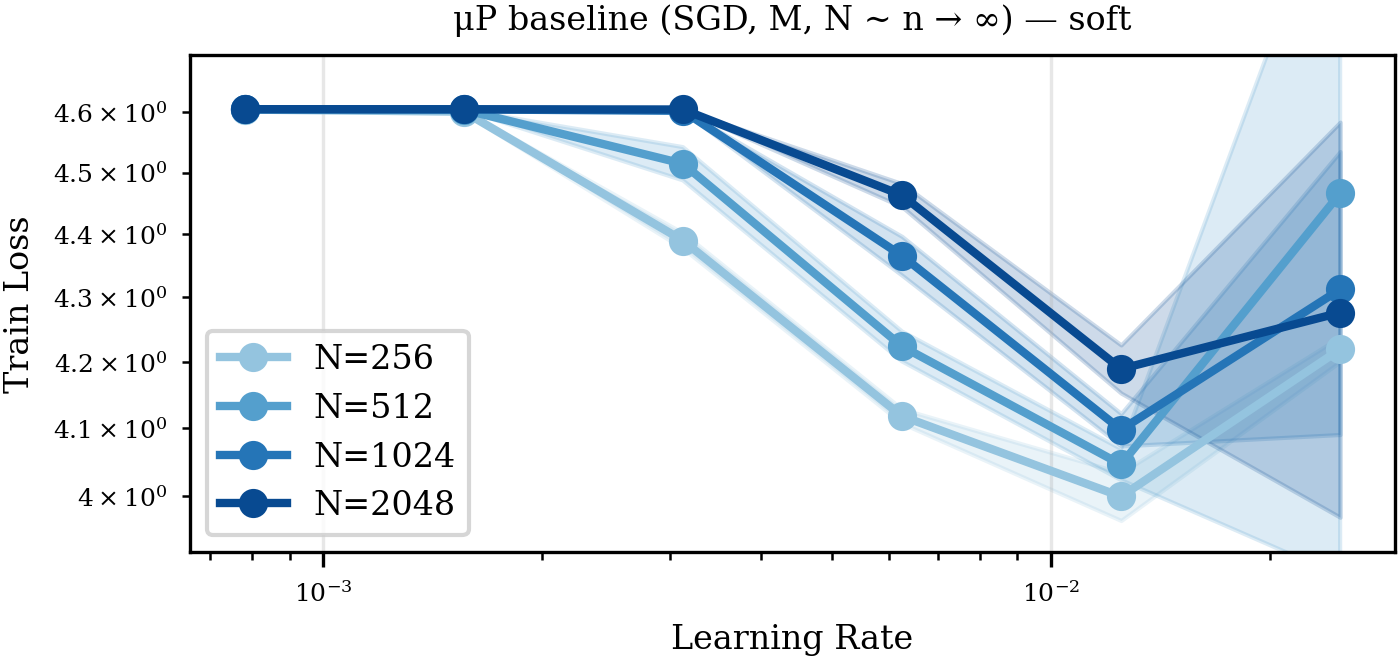}
\hfill
\includegraphics[width=0.48\textwidth]{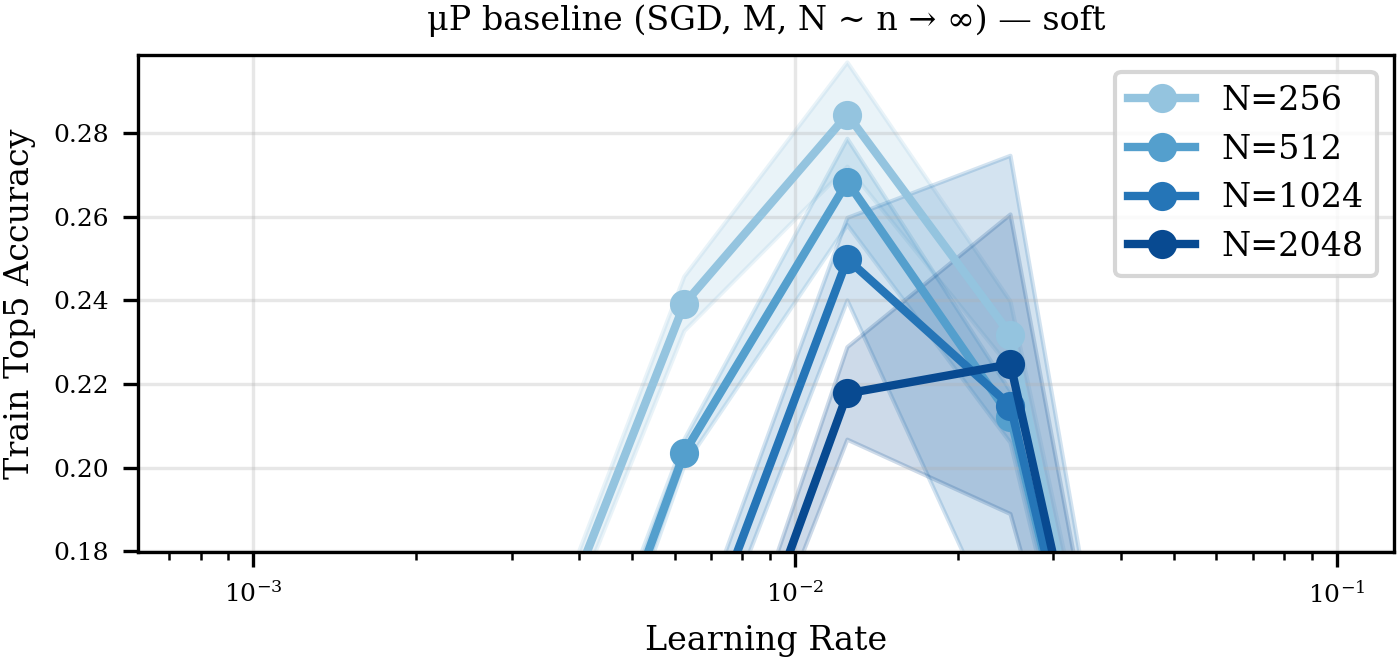}
\caption{\textbf{LR sweep, $\mu$P baseline (SGD, soft, Regime II).} Ending lines denote divergence of the training run. The optimal learning rate drifts toward the maximal stable threshold, which does not increase with width. Observe delayed learning in the corresponding coordinate checks, which results in performance getting worse with model scale.}
\label{fig:lr_bottleneck_sgd_soft_std}
\end{figure}

\begin{figure}[H]
\centering
\includegraphics[width=0.48\textwidth]{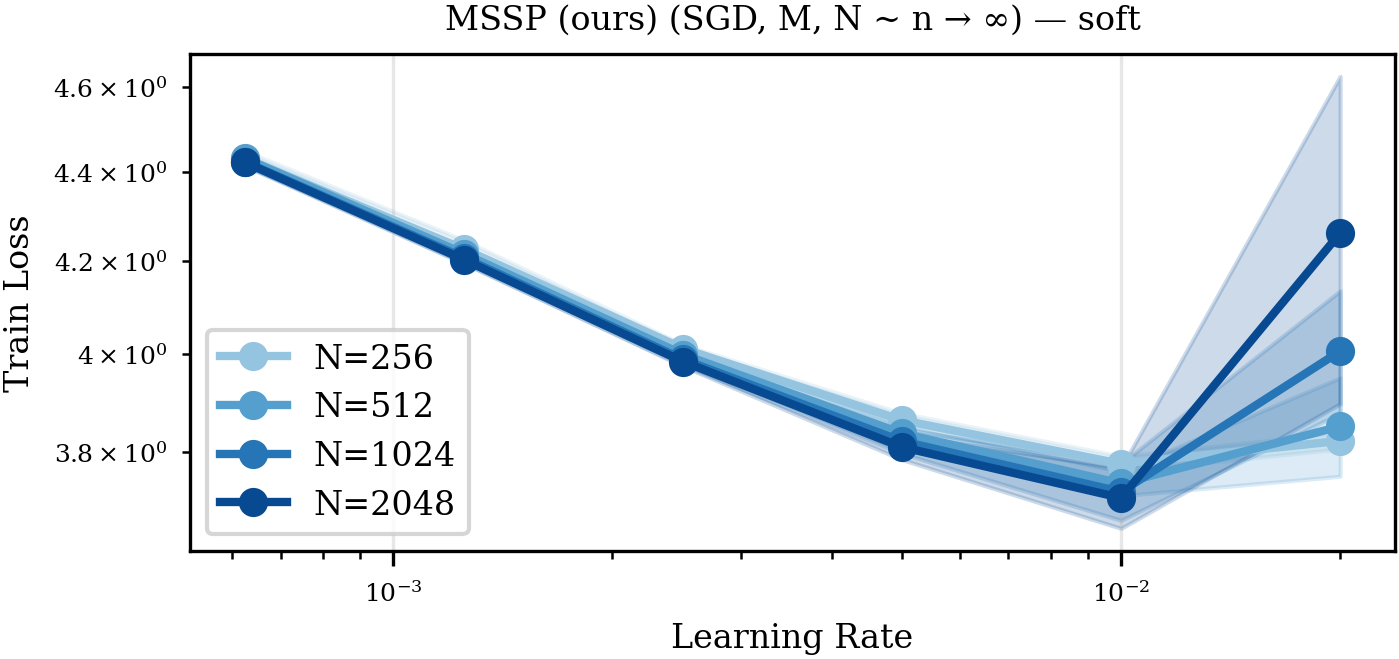}
\hfill
\includegraphics[width=0.48\textwidth]{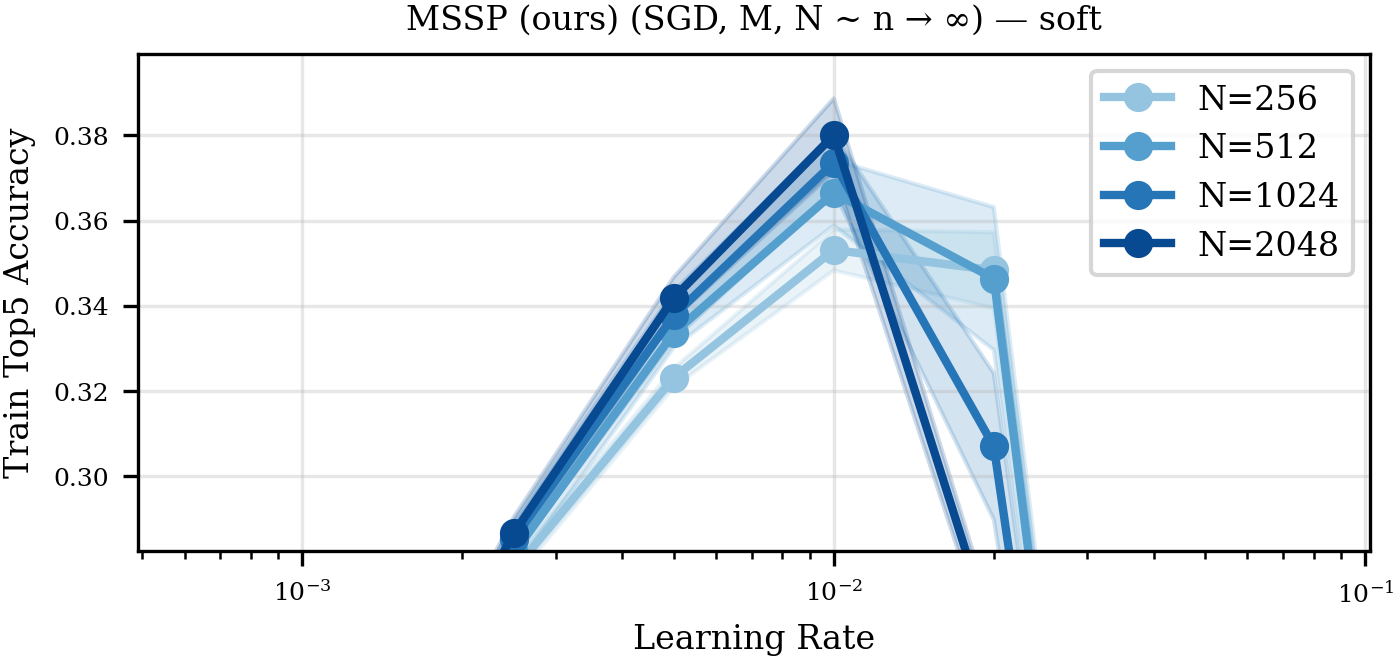}
\caption{\textbf{LR sweep, MSSP (SGD, soft, Regime II).}}
\label{fig:lr_bottleneck_sgd_soft_ours}
\end{figure}

\begin{figure}[H]
\centering
\includegraphics[width=0.48\textwidth]{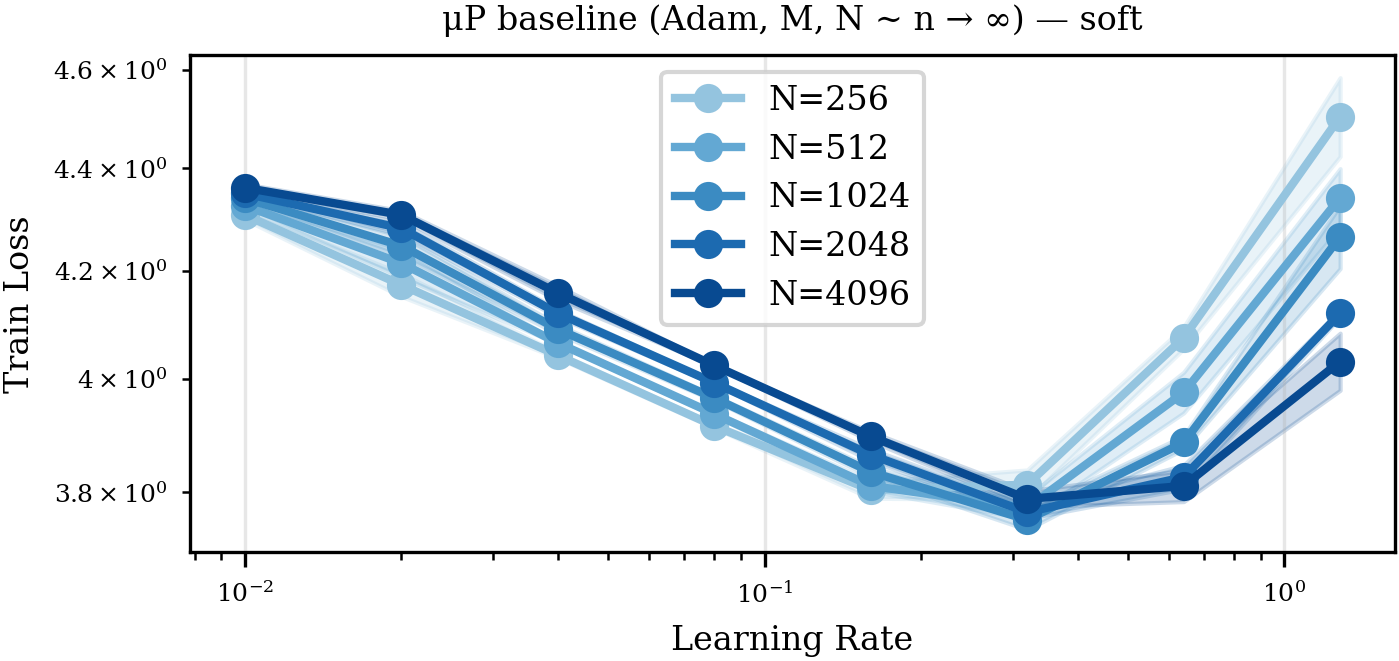}
\hfill
\includegraphics[width=0.48\textwidth]{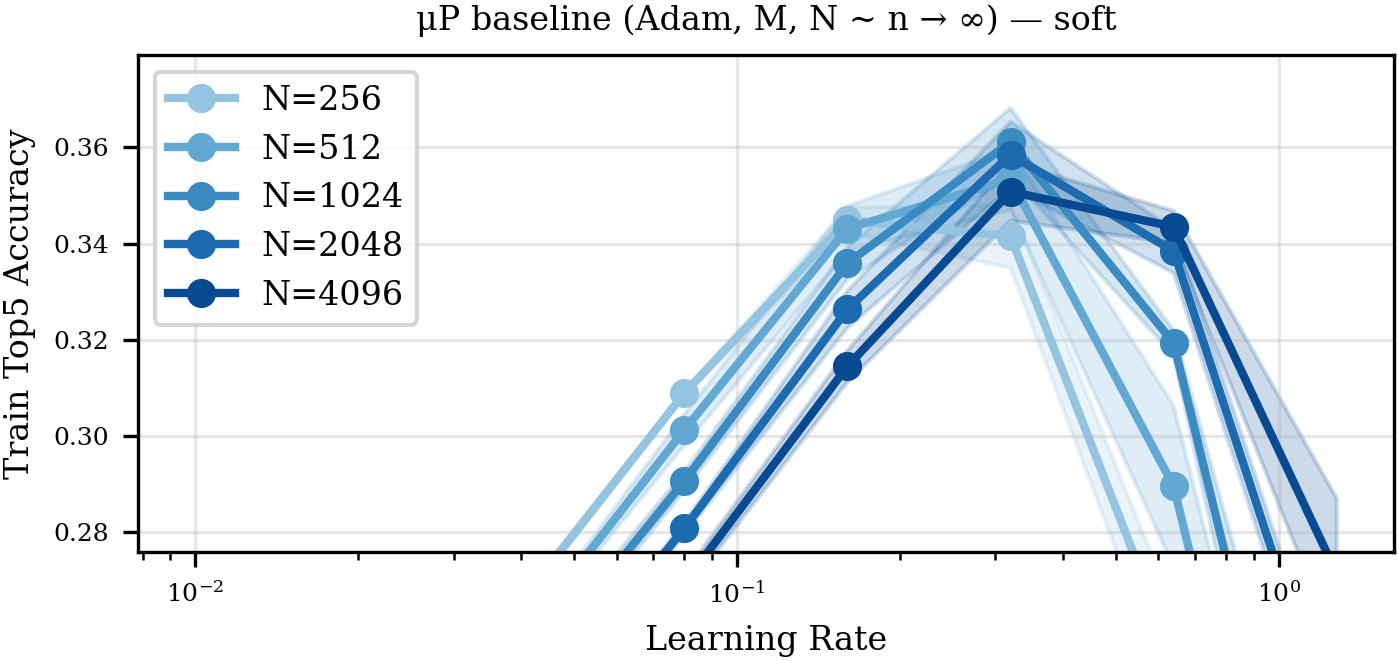}
\caption{\textbf{LR sweep, $\mu$P baseline (Adam, soft, Regime II).}}
\label{fig:lr_bottleneck_adam_soft_std}
\end{figure}

\begin{figure}[H]
\centering
\includegraphics[width=0.48\textwidth]{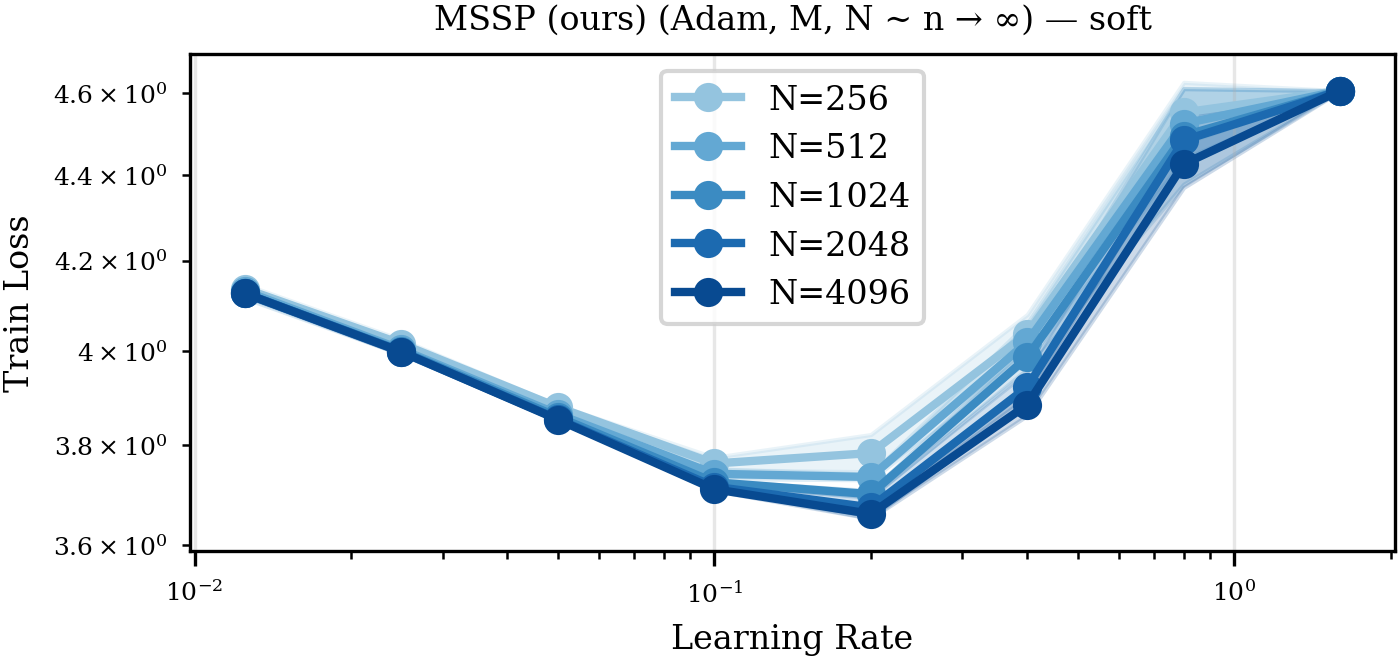}
\hfill
\includegraphics[width=0.48\textwidth]{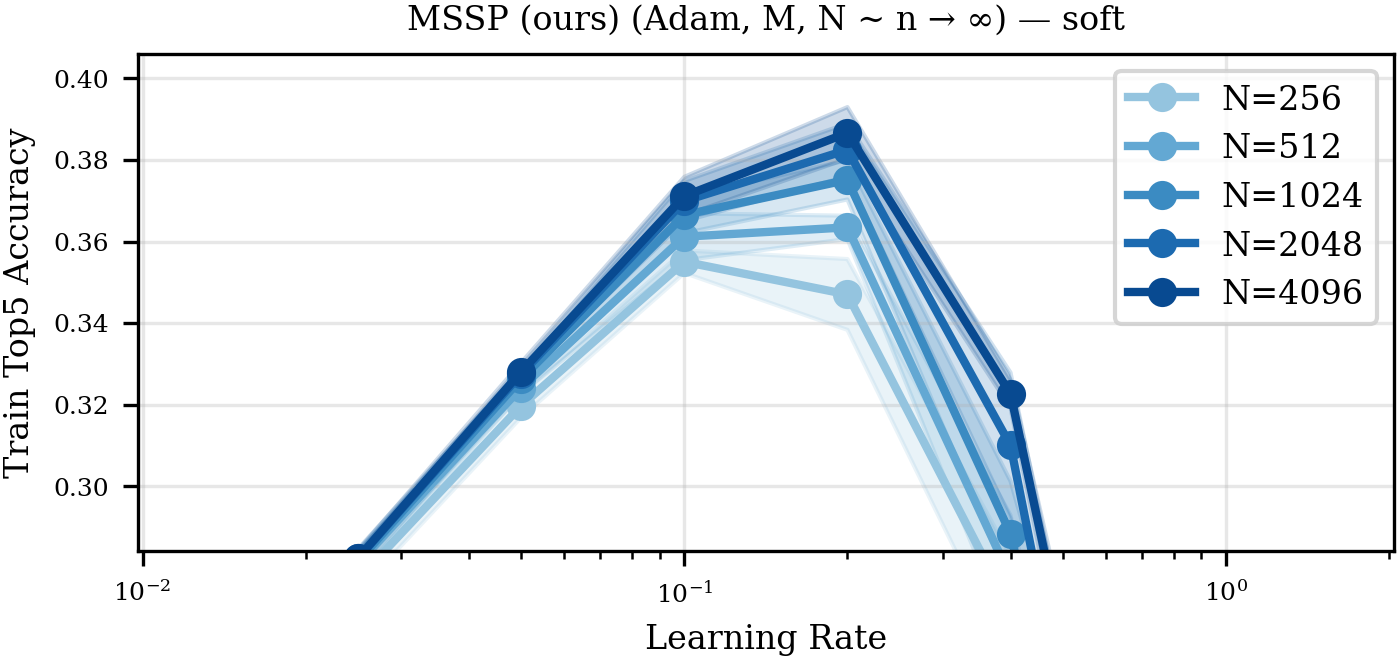}
\caption{\textbf{LR sweep, MSSP (Adam, soft, Regime II).}}
\label{fig:lr_bottleneck_adam_soft_ours}
\end{figure}

\subsubsection{Regime III: Joint proportional scaling}\label{sec:lr_regime3}

Similar to Regime II, the optimal learning rate in $\mu$P saturates at the maximal stable learning rate. The optimal learning rate in MSSP approximately transfers from small to large scale, and achieves better performance than $\mu$P at large scale by consistently recovering monotonic improvement with scale.

\begin{figure}[H]
\centering
\includegraphics[width=0.48\textwidth]{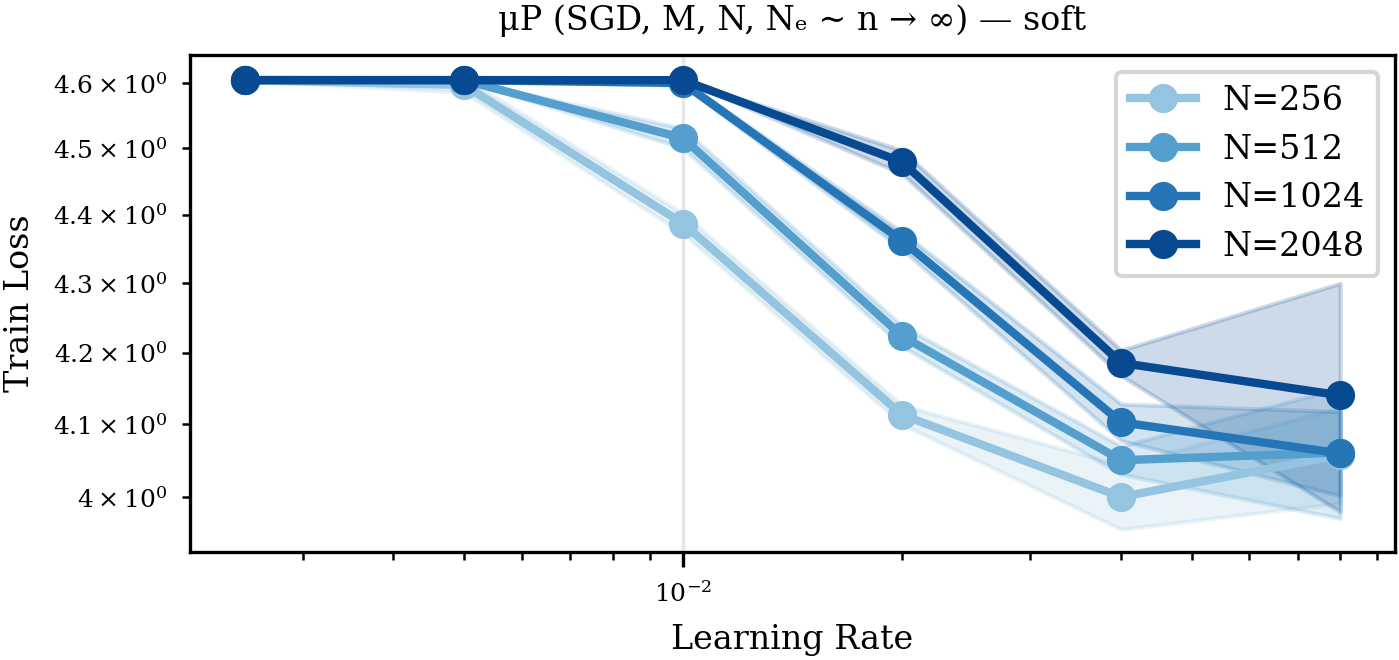}
\hfill
\includegraphics[width=0.48\textwidth]{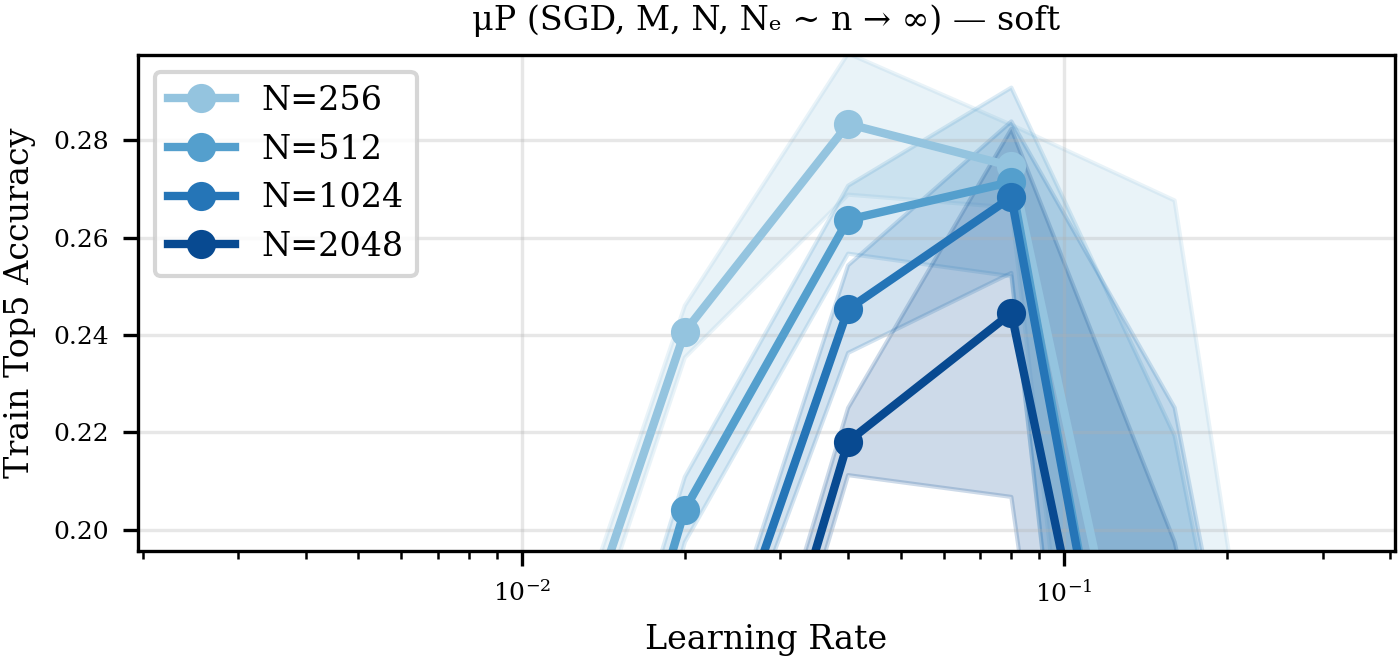}
\caption{\textbf{LR sweep, $\mu$P without shared experts (SGD, soft, Regime III).}}
\label{fig:lr_allscale_sgd_soft_nsh}
\end{figure}

\begin{figure}[H]
\centering
\includegraphics[width=0.48\textwidth]{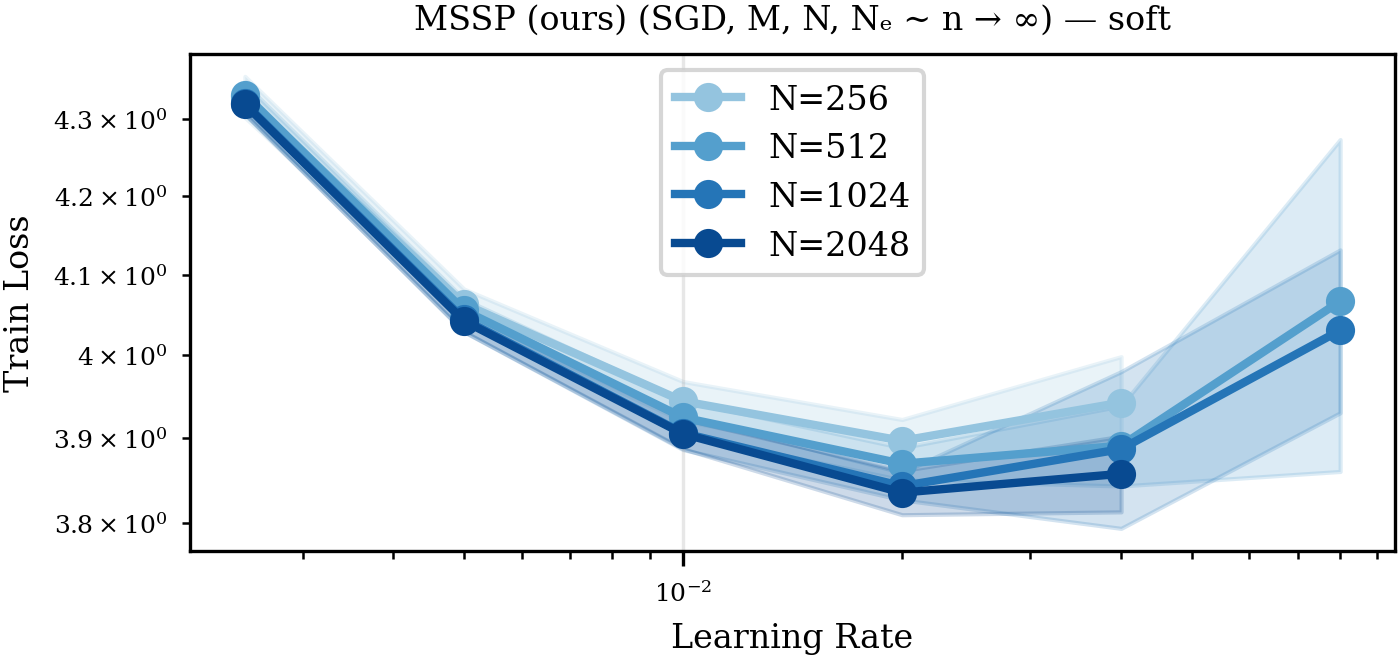}
\hfill
\includegraphics[width=0.48\textwidth]{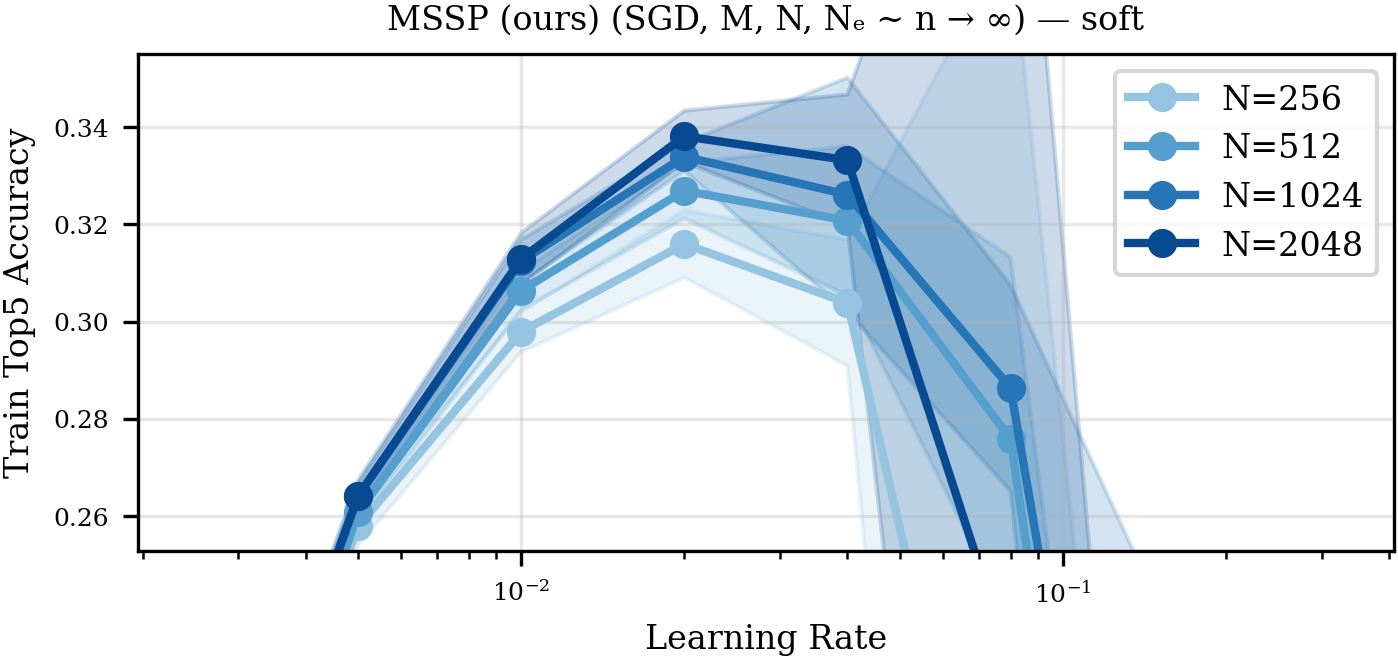}
\caption{\textbf{LR sweep, MSSP with shared experts (SGD, soft, Regime III).}}
\label{fig:lr_allscale_sgd_soft_sh}
\end{figure}

\begin{figure}[H]
\centering
\includegraphics[width=0.48\textwidth]{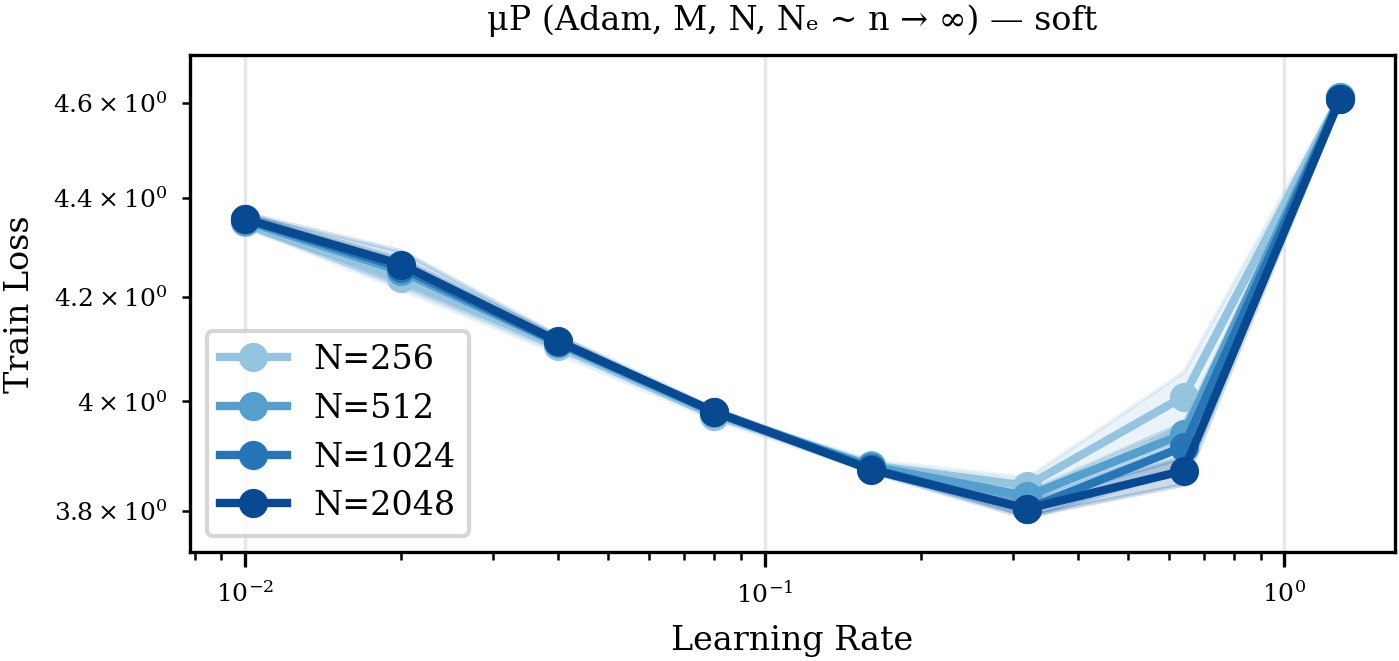}
\hfill
\includegraphics[width=0.48\textwidth]{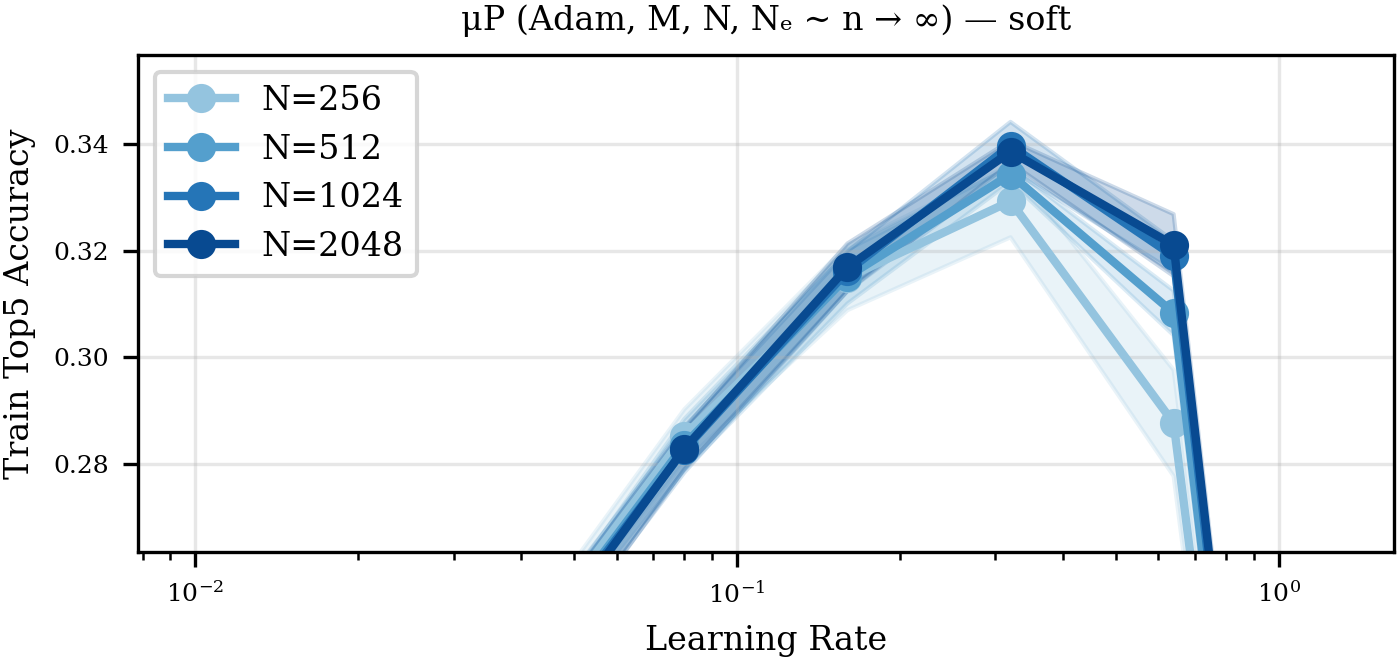}
\caption{\textbf{LR sweep, $\mu$P without shared experts (Adam, soft, Regime III).}}
\label{fig:lr_allscale_adam_soft_nsh_mup}
\end{figure}

\begin{figure}[H]
\centering
\includegraphics[width=0.48\textwidth]{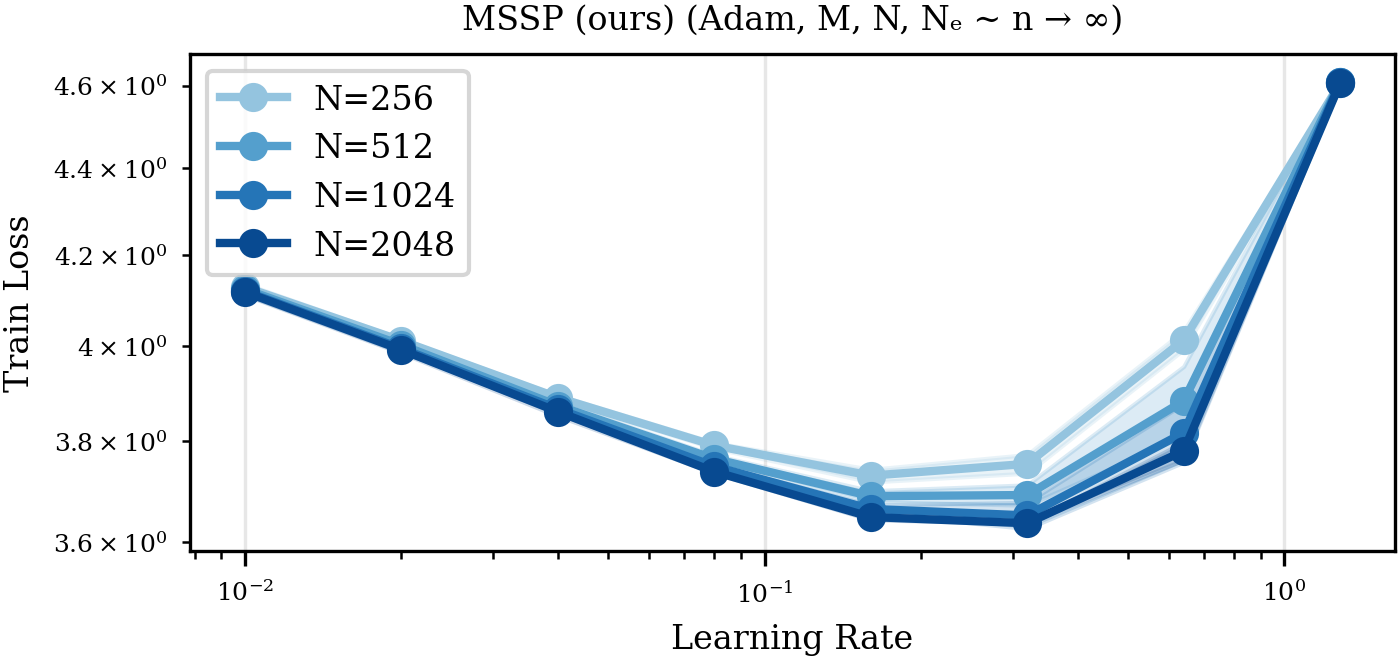}
\hfill
\includegraphics[width=0.48\textwidth]{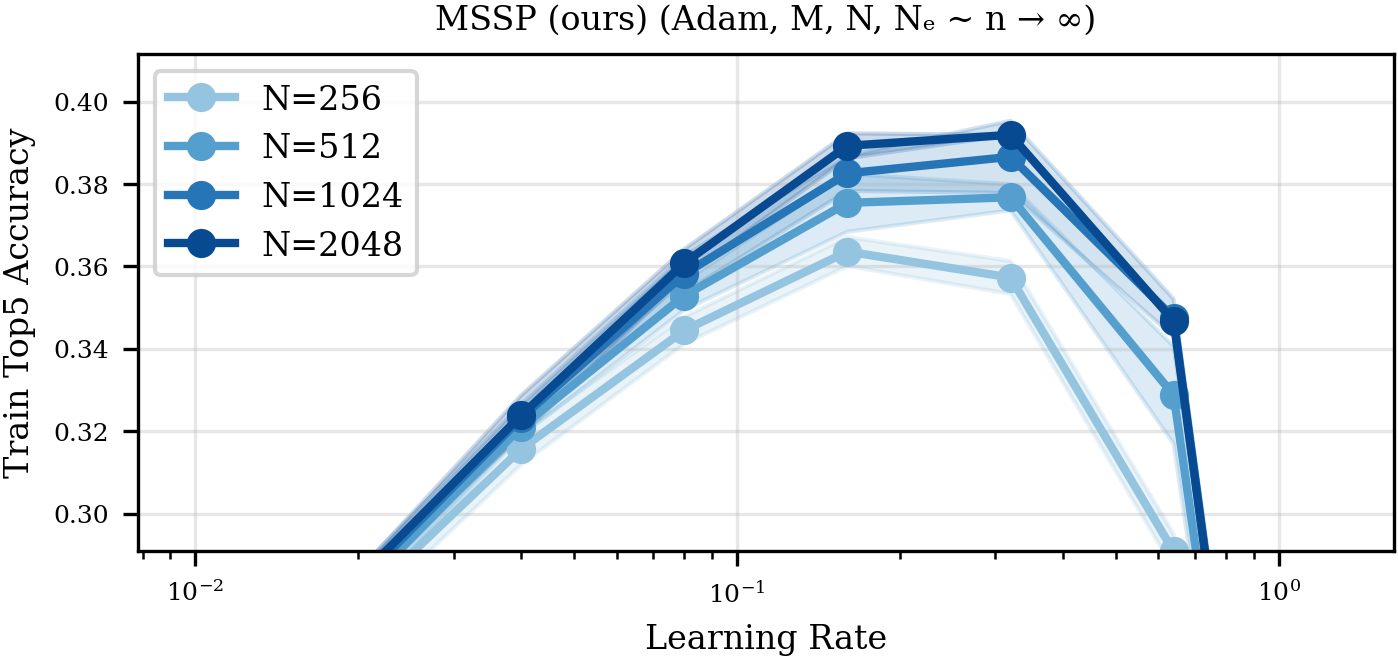}
\caption{\textbf{LR sweep, MSSP with shared experts (Adam, soft, Regime III).}}
\label{fig:lr_allscale_adam_soft_sh_0315}
\end{figure}

\begin{figure}[H]
\centering
\includegraphics[width=0.48\textwidth]{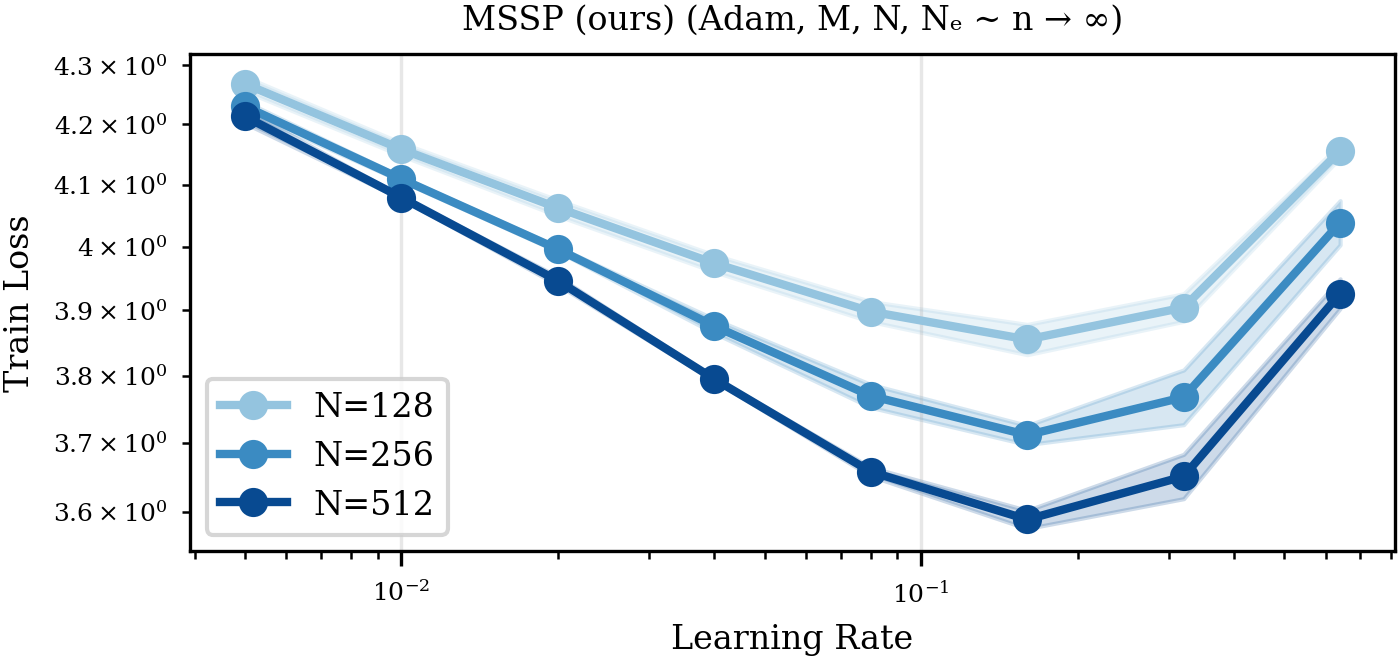}
\hfill
\includegraphics[width=0.48\textwidth]{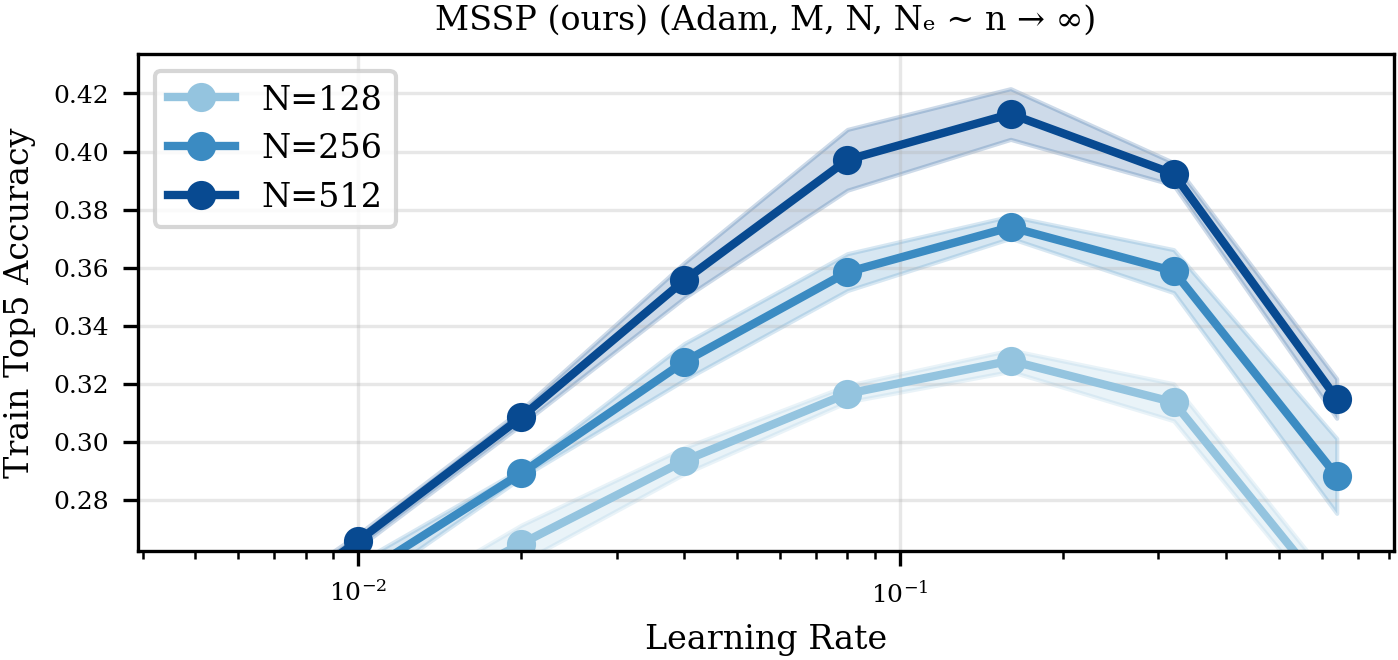}
\caption{\textbf{LR sweep, MSSP with shared experts (Adam, top-$k$, Regime III).}}
\label{fig:lr_allscale_adam_topk_sh_0311}
\end{figure}

\subsection{Fine-grained scaling evaluations in MLP MoEs}\label{sec:rcc_mlp}

Here we show standardized evaluation figures for training the optimal configuration of many combination of optimizer, parameterization, routing mechanism and scaling regime. Across the board, we observe more desirable scaling properties in MSSP than in $\mu$P.

The first row shows training loss, training accuracy, the cumulative feature learning of the entire MoE $\|\Delta h^L_t\|_{RMS}$, as well as the routing logit norm $\|\psi_t\|_{RMS}$ before the sigmoid. These allow to verify delayed learning versus monotonic improvement with scale, as well as the scale dependence in feature and router learning.

The second and third rows show refined coordinate check exponents as a function of width $N$. The general desired exponent for propagating and effective updates is zero, signaling scale independence of the respective component of the training dynamics (with the exception of expert output propagating update exponent $0.5$ in MSSP in Regime II). The expected layerwise gradient scaling exponents vary between $-2$ and $0$, depending on the scaling regime. Dashed horizontal lines indicate where expected exponents are non-zero. Decomposition exponents (third row first and second subplots) should generally be balanced, otherwise subcomponents of the dynamics strictly dominate others, which causes strong finite-width effects. Exponents should ideally also be independent of the step, otherwise non-trivial initial dynamical effects such as delayed learning are to be expected.

Missing lines in the propagating update and decomposition plots denote $0$ values due to zero initialization of the relevant layer. For example under zero last-layer initialization, the initial gradient is zero except in the output layer. This helps to reduce width dependence from vanishing initial terms.

We show both raw effective updates (second row, center) and a variant with normalized incoming activation norm (second row, right) to distinguish width dependence from the combination of the current and previous layers versus isolated width dependence in the current layer.

In all regimes, the predicted exponents hold surprisingly well across training, suggesting that our theory is predictive of practical training dynamics far beyond the first few iterations. The router gradient exponents are most noisy, but the closest clean exponent of all layers still follows our prediction in MSSP remarkably well throughout training.
 
In $\mu$P, width dependence in individual subterms of the training dynamics cascades into the entire dynamics such that exponents become much more width and time dependent across layers, and often converge to intermediate values between the clean exponents $\{-0.5, 0, 0.5\}$.

Extensive layerwise multiplier tuning for each scaling config, optimizer and routing type is paramount for achieving stable training at all.

\subsubsection{Regime I: Fixed number of experts}\label{sec:rcc_regime1}

Under maximal stable router initialization $\sigma=1/N$, the propagating updates in the router are too small. Setting the router initialization to 0 removes this source of width dependence and results in cleaner scaling exponents. The effect of this intervention on the final performance is negligible.

The top-$k$ selection mechanism with $k\asymp M$ does not change any expected scaling exponents in this paper. Indeed, \Cref{fig:fixedE_base_sgd_topk2_ll0_0421,fig:fixedE_base_sgd_topk2_r0_ll0_0421,fig:fixedE_base_adam_topk2_ll0_0417,fig:fixedE_base_adam_topk2_r0_ll0_0417} verify that all exponents remain unaltered, albeit slightly more noisy.

\begin{figure}[H]
\centering
\includegraphics[width=\textwidth]{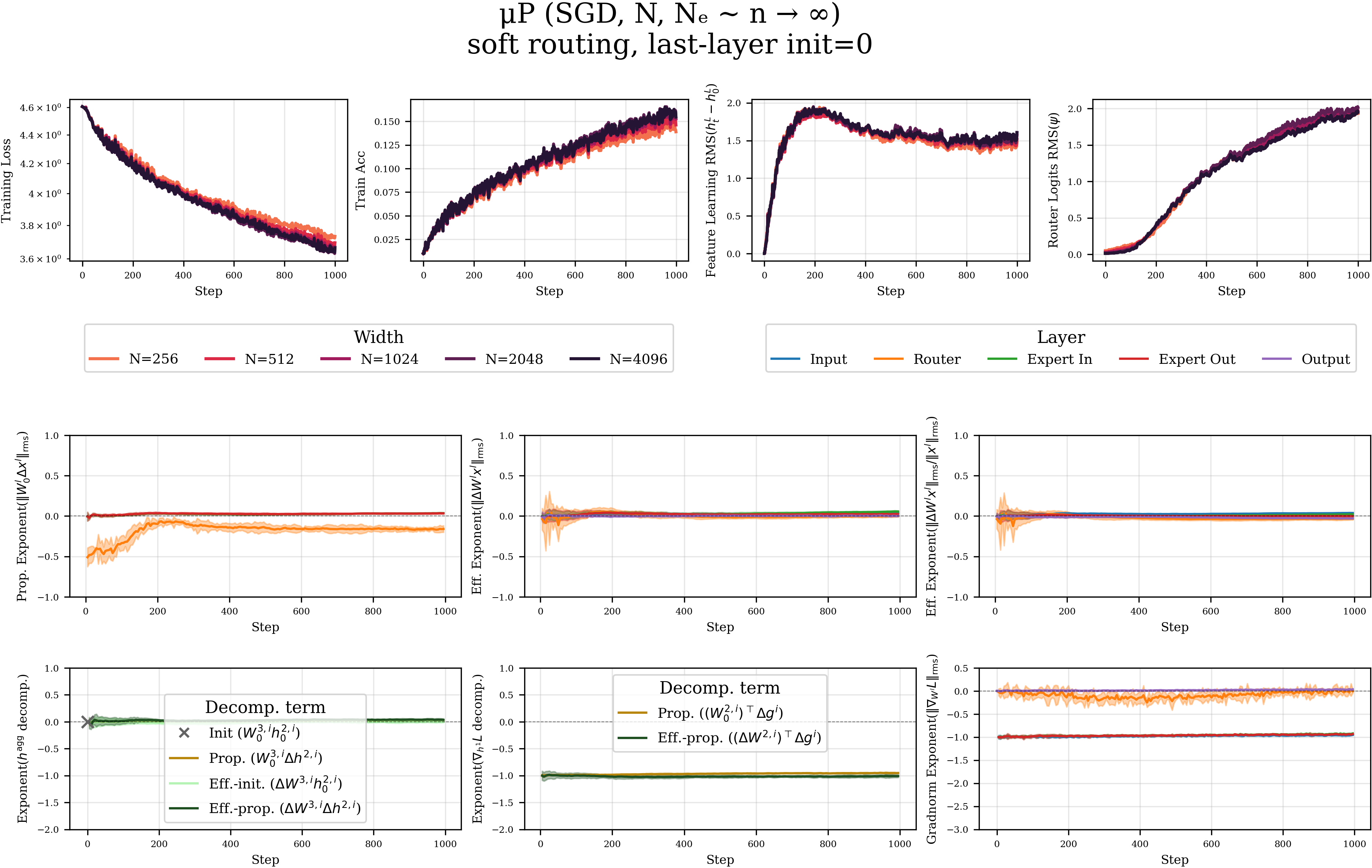}
\caption{\textbf{$\mu$P with 1/N router init (SGD, Regime I).}}
\label{fig:fixedE_base_sgd_soft_ll0_0421}
\end{figure}

\begin{figure}[H]
\centering
\includegraphics[width=\textwidth]{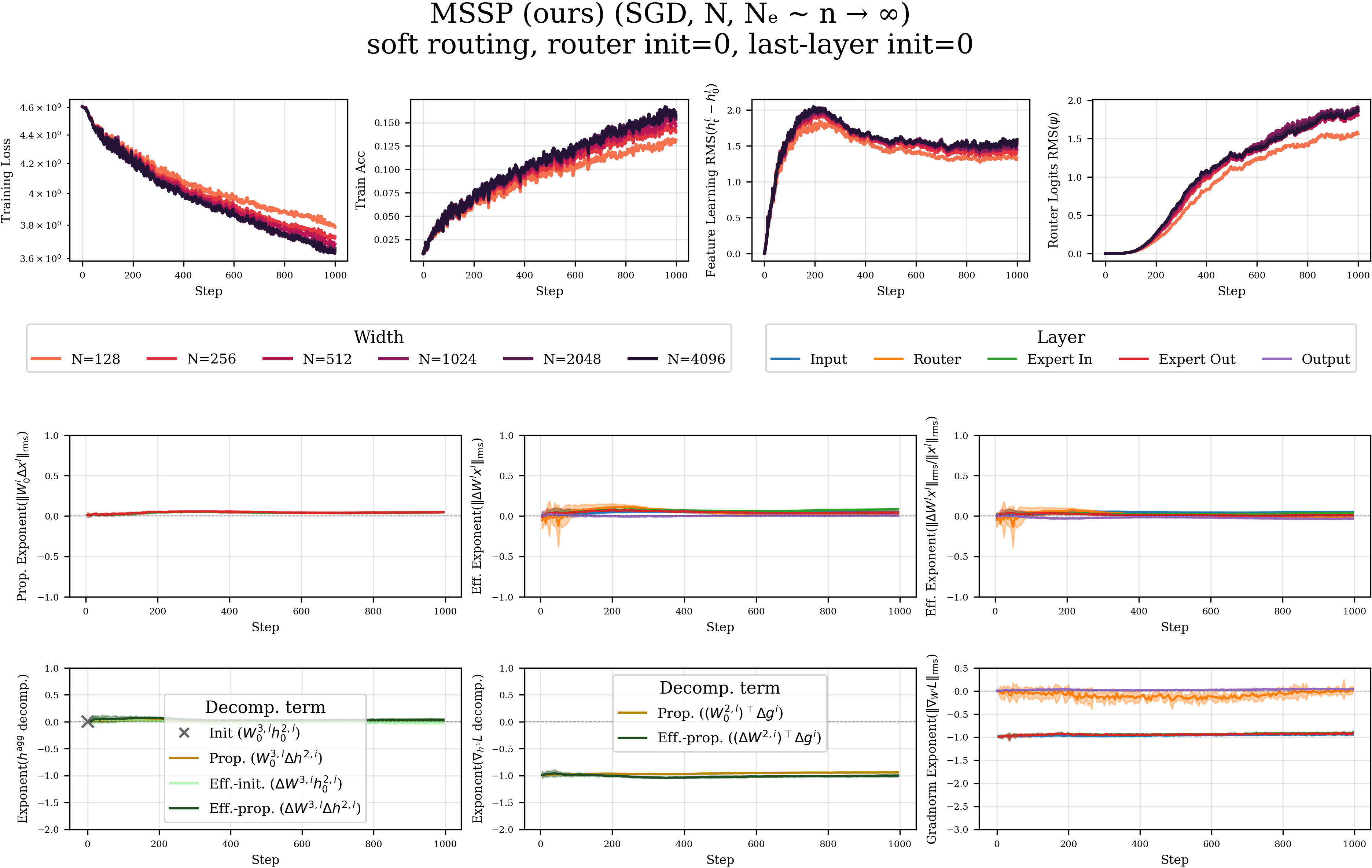}
\caption{\textbf{$\mu$P with zero router init (SGD, Regime I).}}
\label{fig:fixedE_base_sgd_soft_r0_ll0_0420}
\end{figure}

\begin{figure}[H]
\centering
\includegraphics[width=\textwidth]{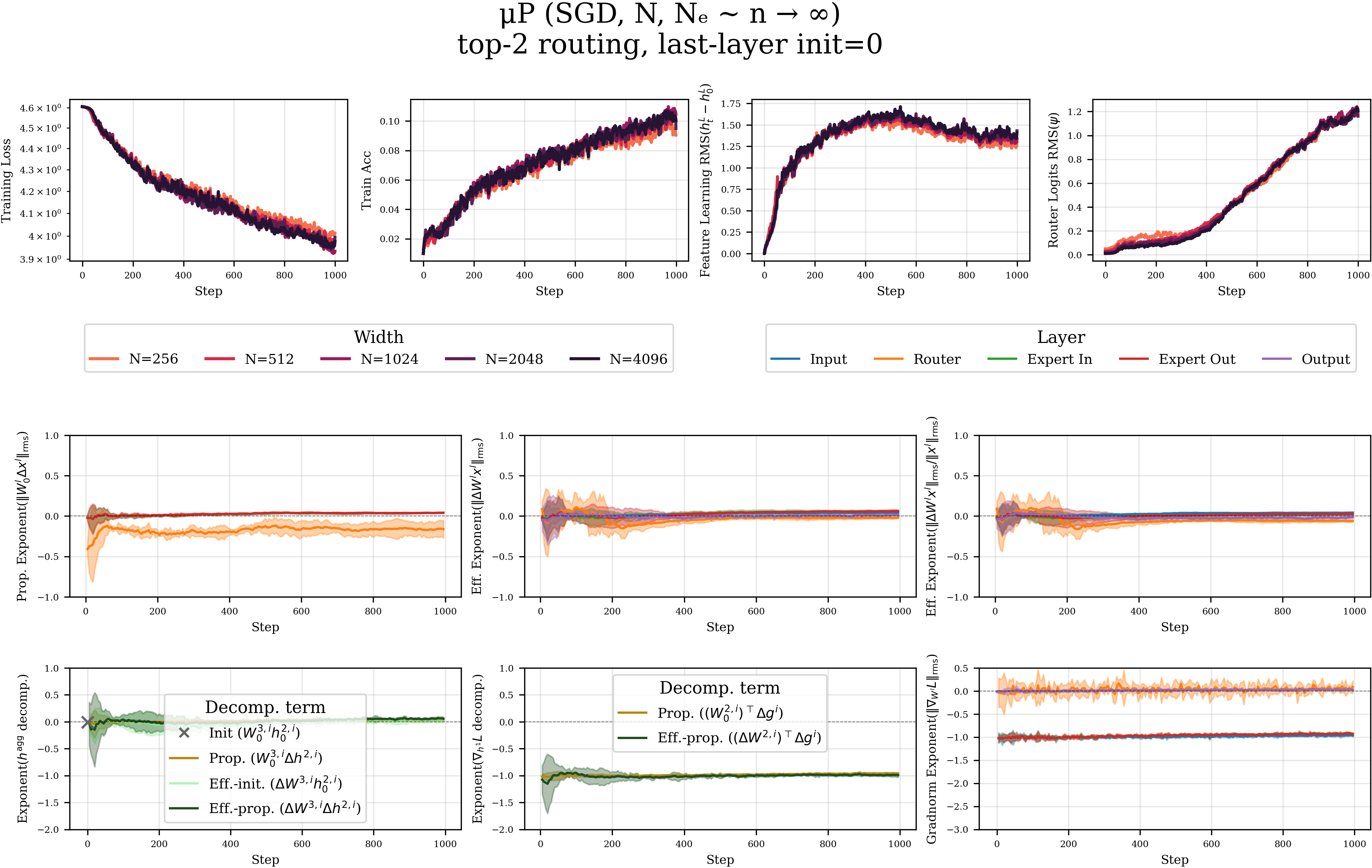}
\caption{\textbf{$\mu$P with 1/N router init (SGD, Regime I, top-$k$).}}
\label{fig:fixedE_base_sgd_topk2_ll0_0421}
\end{figure}

\begin{figure}[H]
\centering
\includegraphics[width=\textwidth]{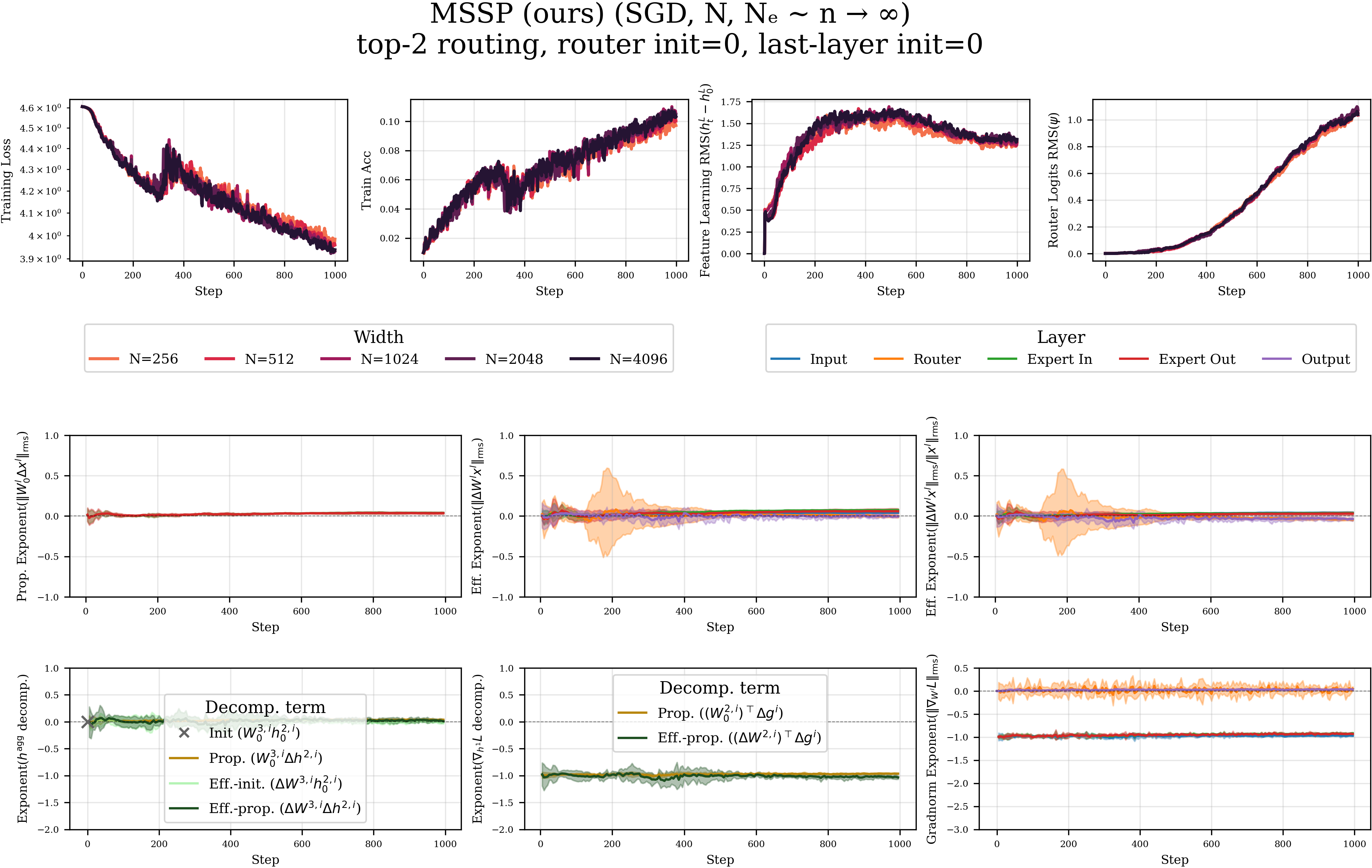}
\caption{\textbf{MSSP with zero router init (SGD, Regime I, top-$k$).}}
\label{fig:fixedE_base_sgd_topk2_r0_ll0_0421}
\end{figure}

\begin{figure}[H]
\centering
\includegraphics[width=\textwidth]{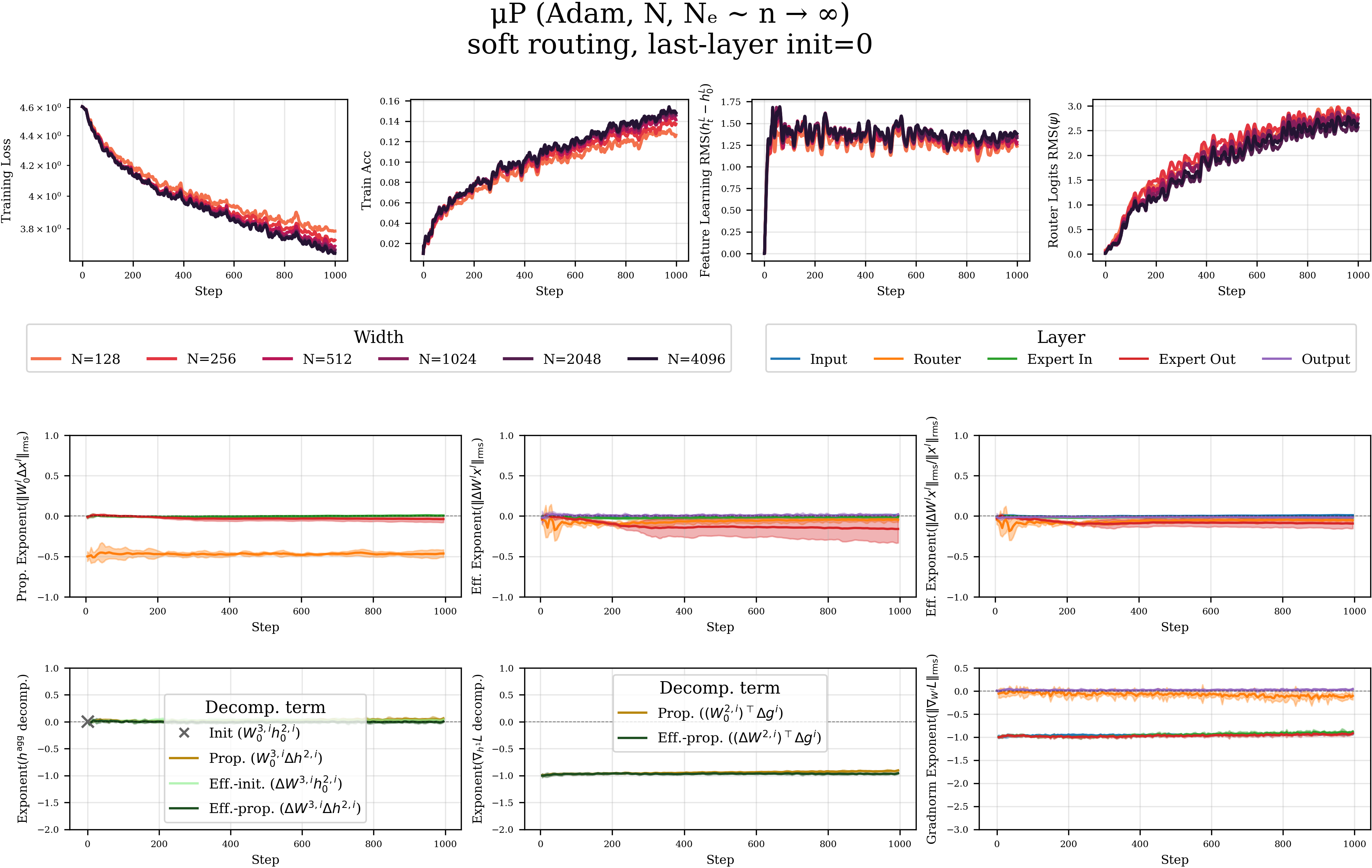}
\caption{\textbf{$\mu$P with 1/N router init (Adam, Regime I).}}
\label{fig:fixedE_base_adam_soft_ll0_0416}
\end{figure}

\begin{figure}[H]
\centering
\includegraphics[width=\textwidth]{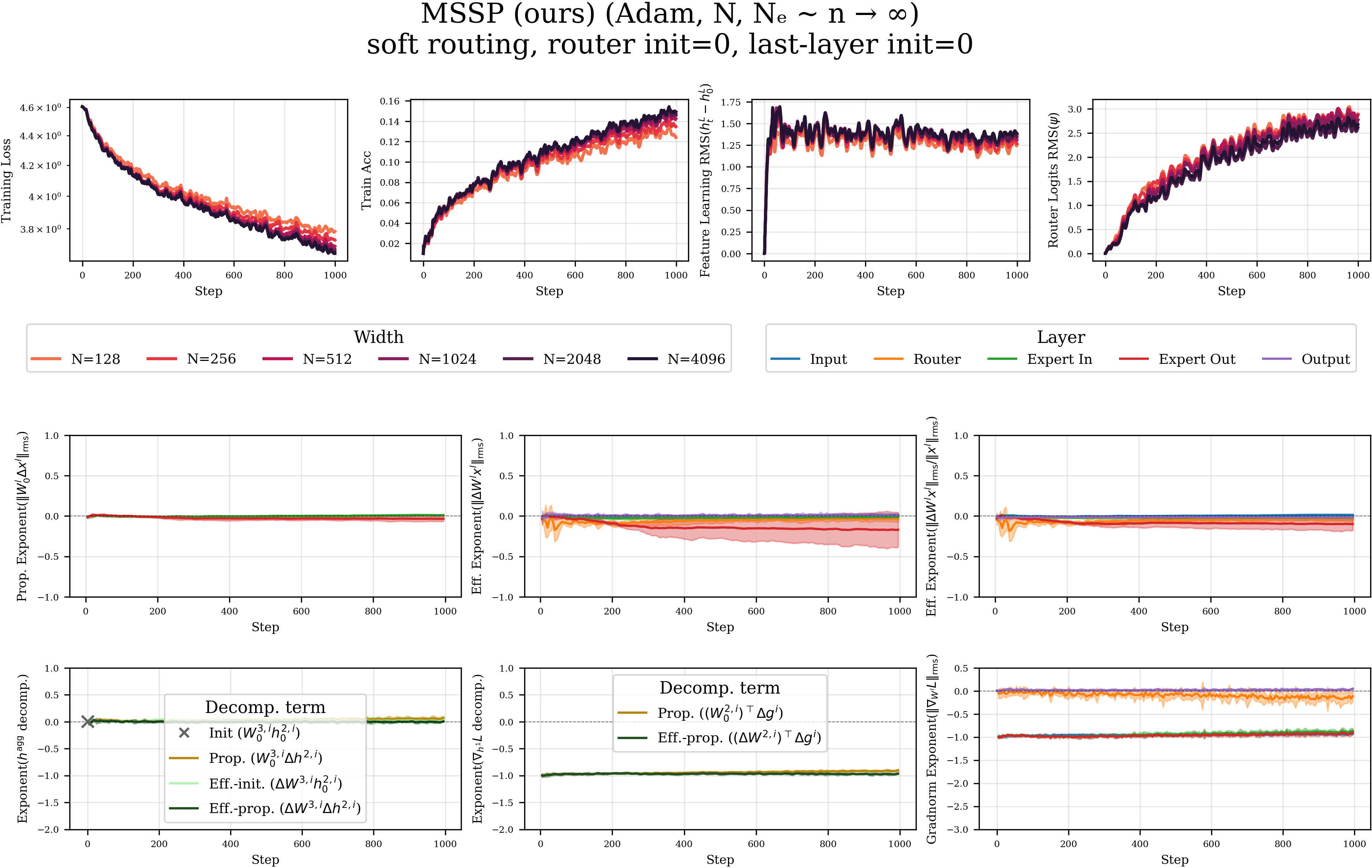}
\caption{\textbf{MSSP with zero router init (Adam, Regime I).}}
\label{fig:fixedE_base_adam_soft_r0_ll0_0417}
\end{figure}

\begin{figure}[H]
\centering
\includegraphics[width=\textwidth]{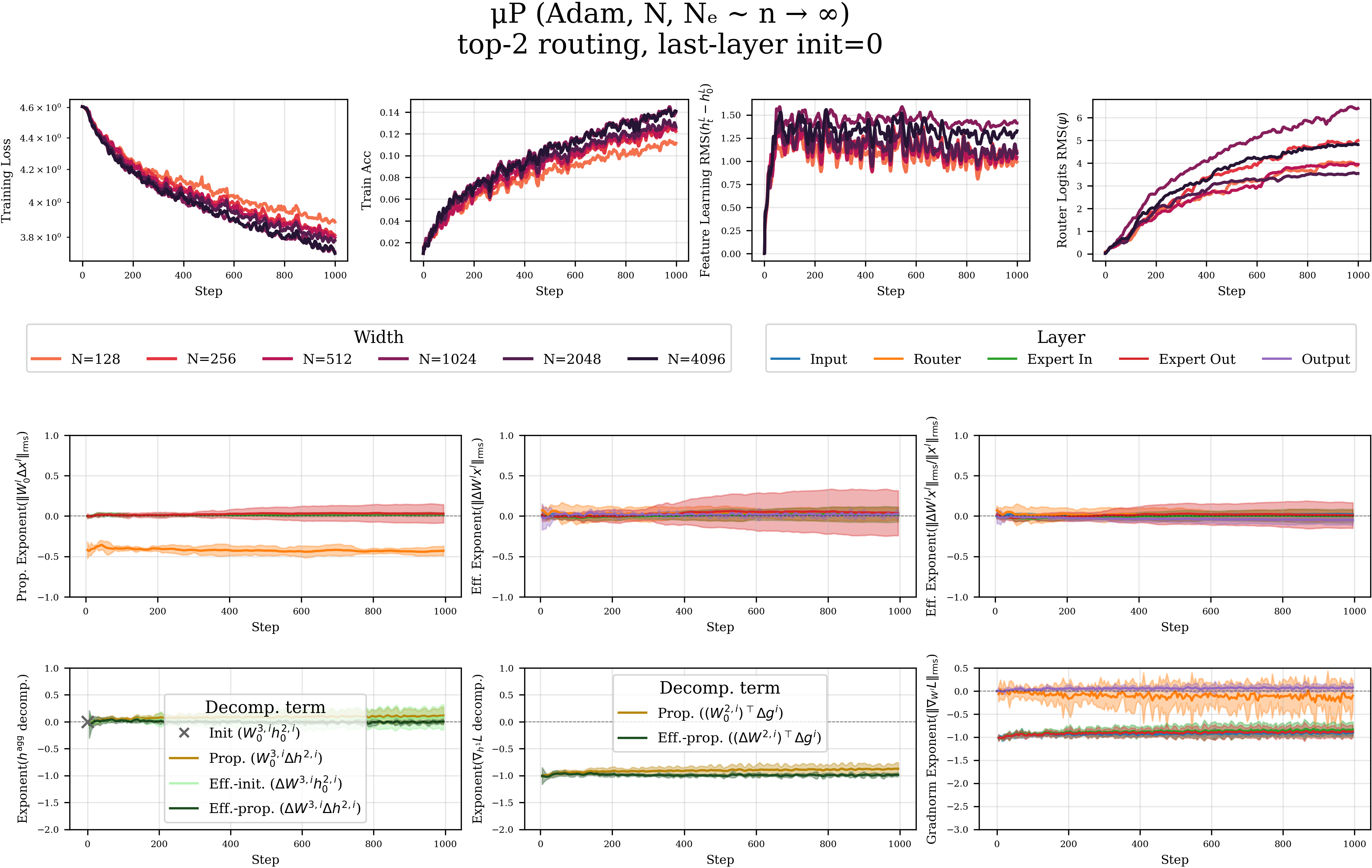}
\caption{\textbf{$\mu$P with 1/N router init (Adam, Regime I, top-$k$).}}
\label{fig:fixedE_base_adam_topk2_ll0_0417}
\end{figure}

\begin{figure}[H]
\centering
\includegraphics[width=\textwidth]{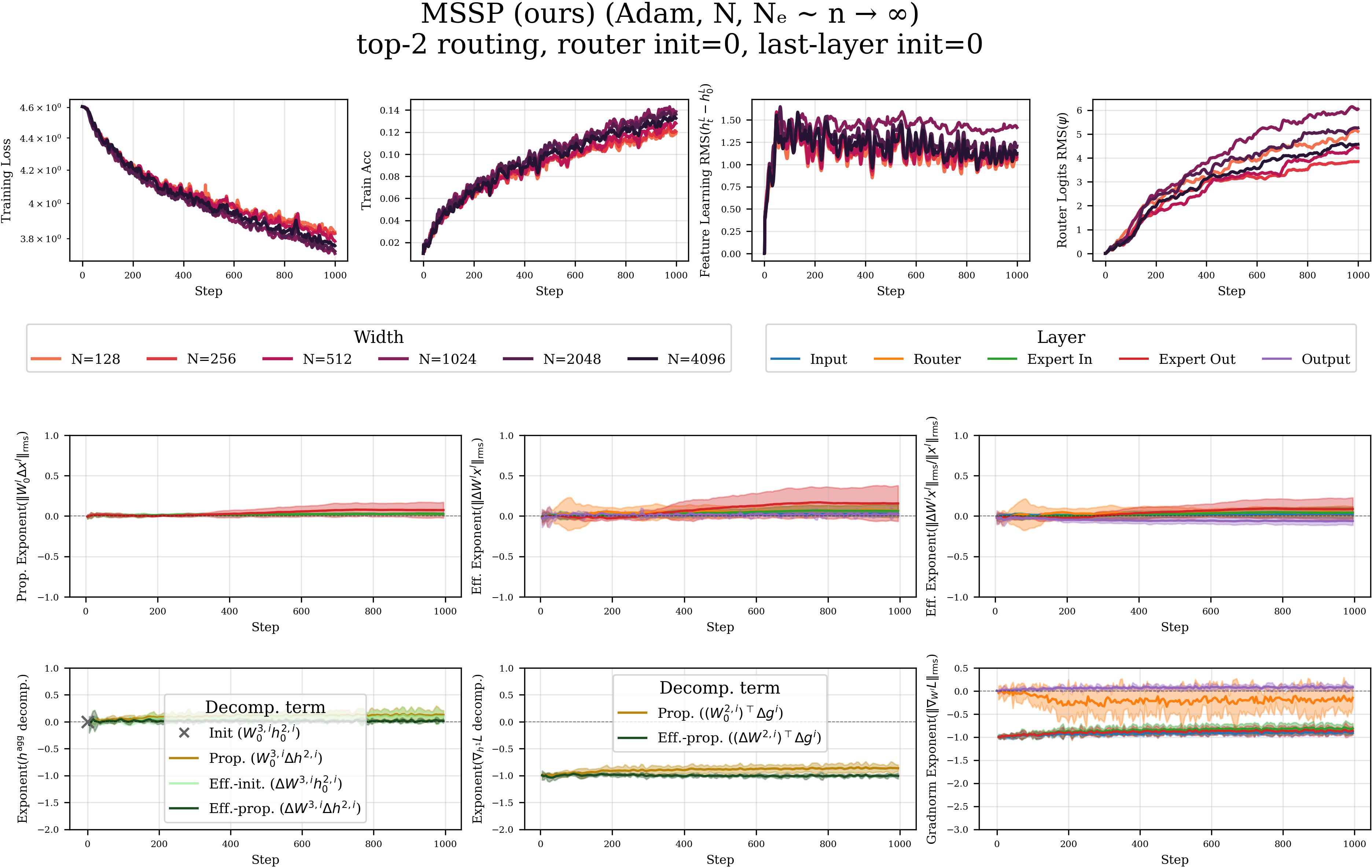}
\caption{\textbf{MSSP with zero router init (Adam, Regime I, top-$k$).}}
\label{fig:fixedE_base_adam_topk2_r0_ll0_0417}
\end{figure}

\subsubsection{Regime II: Fixed expert width}\label{sec:rcc_regime2}

To ensure a stable baseline, we provide $\mu$P both with maximal stable and with $0$ last-layer initialization. Both suffer from the same scaling degeneracies for SGD and Adam.

Starting with SGD, we observe strong delayed learning under maximal stable last-layer initialization $\sigma=1/N$. This delay is visible not only in training loss, but also feature and router logit learning. Initial vanishing terms in the expert aggregation cascade even stronger into all layers than under last-layer 0 initialization presented in the main paper, resulting in final exponents that do not follow any clean scaling exponent from $\{-0.5, 0, 0.5\}$. $\mu$P with last-layer zero initialization starts out too small but partially self-corrects, which verifies that this parameterization indeed satisfies the $\mu$P desiderata, as any larger initialization or learning rate scaling would induce divergence in at least one layer after sufficiently many steps.

Adam in $\mu$P stabilizes much faster than SGD, and, while exponents are still not clean, they differ less from $0$. Still, performance does not monotonically improve with scale and reduced feature learning at large scales is visible. Again, MSSP resolves these issues and shows monotonic improvement with scale as well as clean and balanced scaling exponents.

Propagating update scaling in $\mu$P remains vanishing throughout training for both SGD and Adam, as our theory predicts.

Note that, as predicted, expert output layer gradient entries decay extremely fast as $\Theta(N^{-2})$ in both SGD and Adam. If ignored, this can cause numerical precision issues in practice at moderate model sizes. We recommend layerwise gradient scaling or equivalent Adam moment scaling to prevent numerical underflows.

\begin{figure}[H]
\centering
\includegraphics[width=\textwidth]{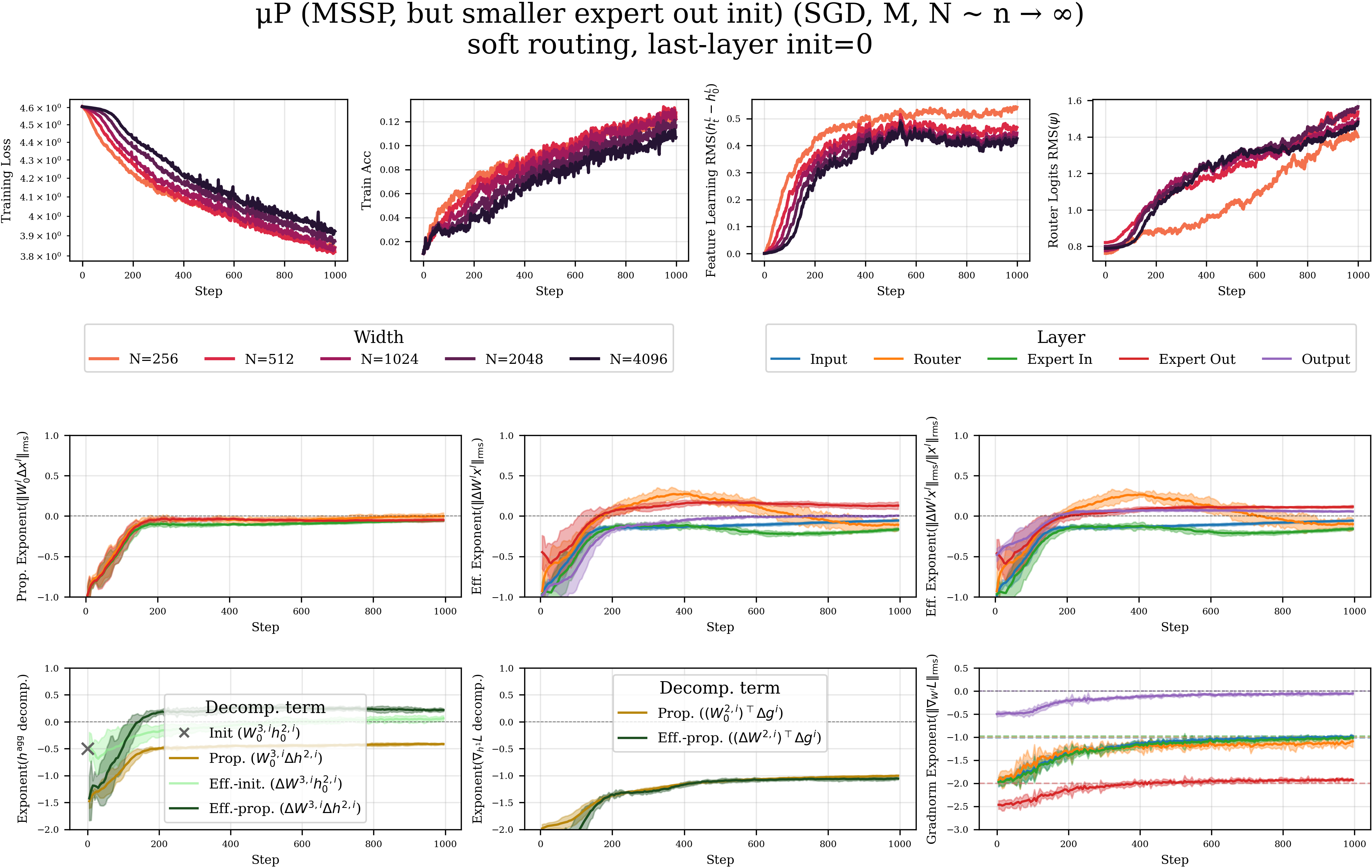}
\caption{\textbf{$\mu$P baseline (SGD, Regime II).}}
\label{fig:bottleneck_std_sgd_soft_ll0_0416}
\end{figure}

\begin{figure}[H]
\centering
\includegraphics[width=\textwidth]{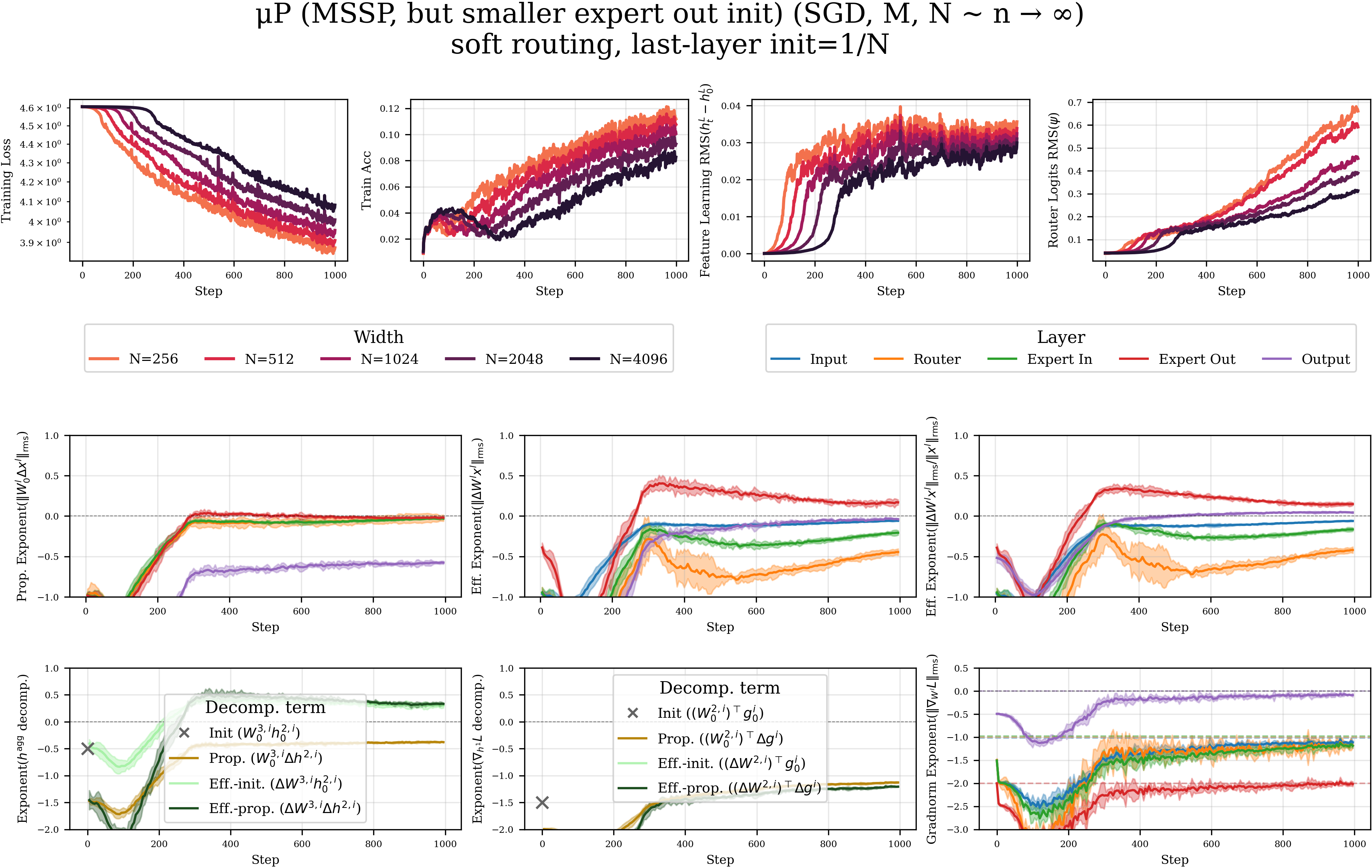}
\caption{\textbf{$\mu$P baseline with 1/N last-layer init (SGD, Regime II).} Interesting initial growth in training accuracy even though training loss and features are barely moving. Learning is still delayed with increasing width.}
\label{fig:bottleneck_std_sgd_soft_ll1n_0429}
\end{figure}

\begin{figure}[H]
\centering
\includegraphics[width=\textwidth]{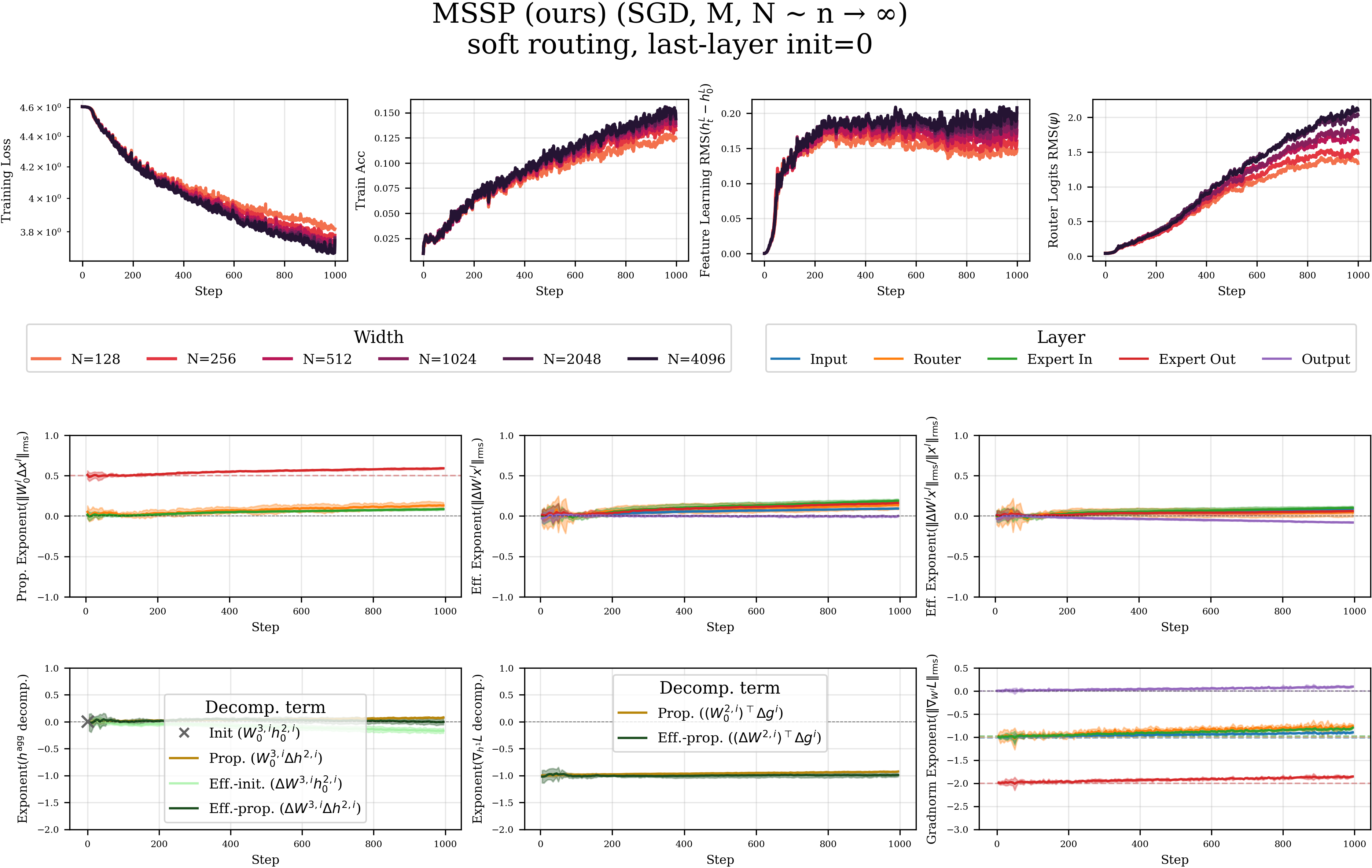}
\caption{\textbf{MSSP (SGD, Regime II).} Note that the propagating update exponent $0.5$ of the expert output layer is desired here, since it is cancelled out by a clean CLT effect in the expert aggregation. The width independence of all other quantities throughout training verifies that the divergence of the individual expert outputs $h^{2,\text{out}}$ through this propagating update term results in approximately width-independent training dynamics.}
\label{fig:bottleneck_ours_sgd_soft_ll0_0420}
\end{figure}

\begin{figure}[H]
\centering
\includegraphics[width=\textwidth]{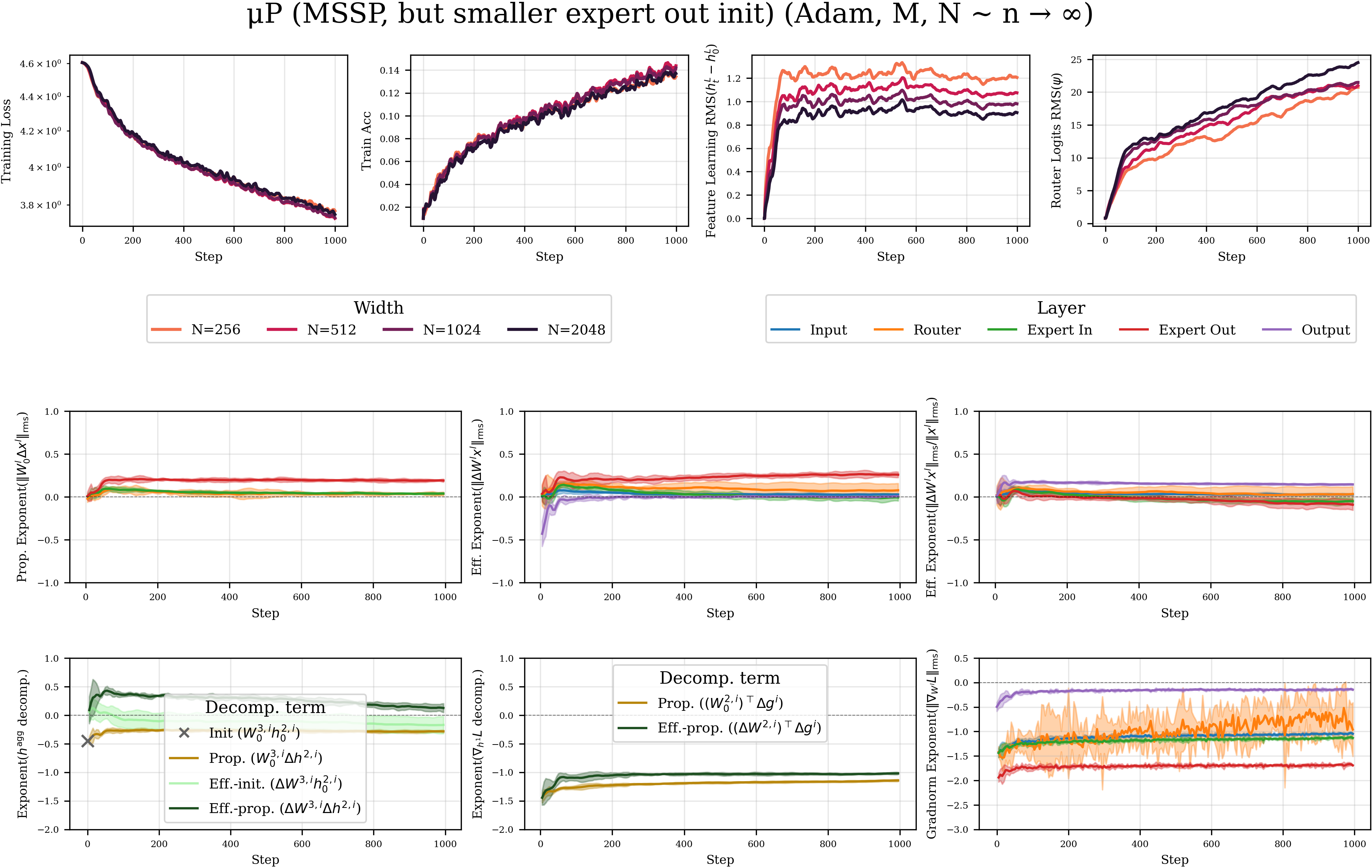}
\caption{\textbf{$\mu$P baseline (Adam, Regime II).}}
\label{fig:bottleneck_std_adam_soft_ll0_0414}
\end{figure}

\begin{figure}[H]
\centering
\includegraphics[width=\textwidth]{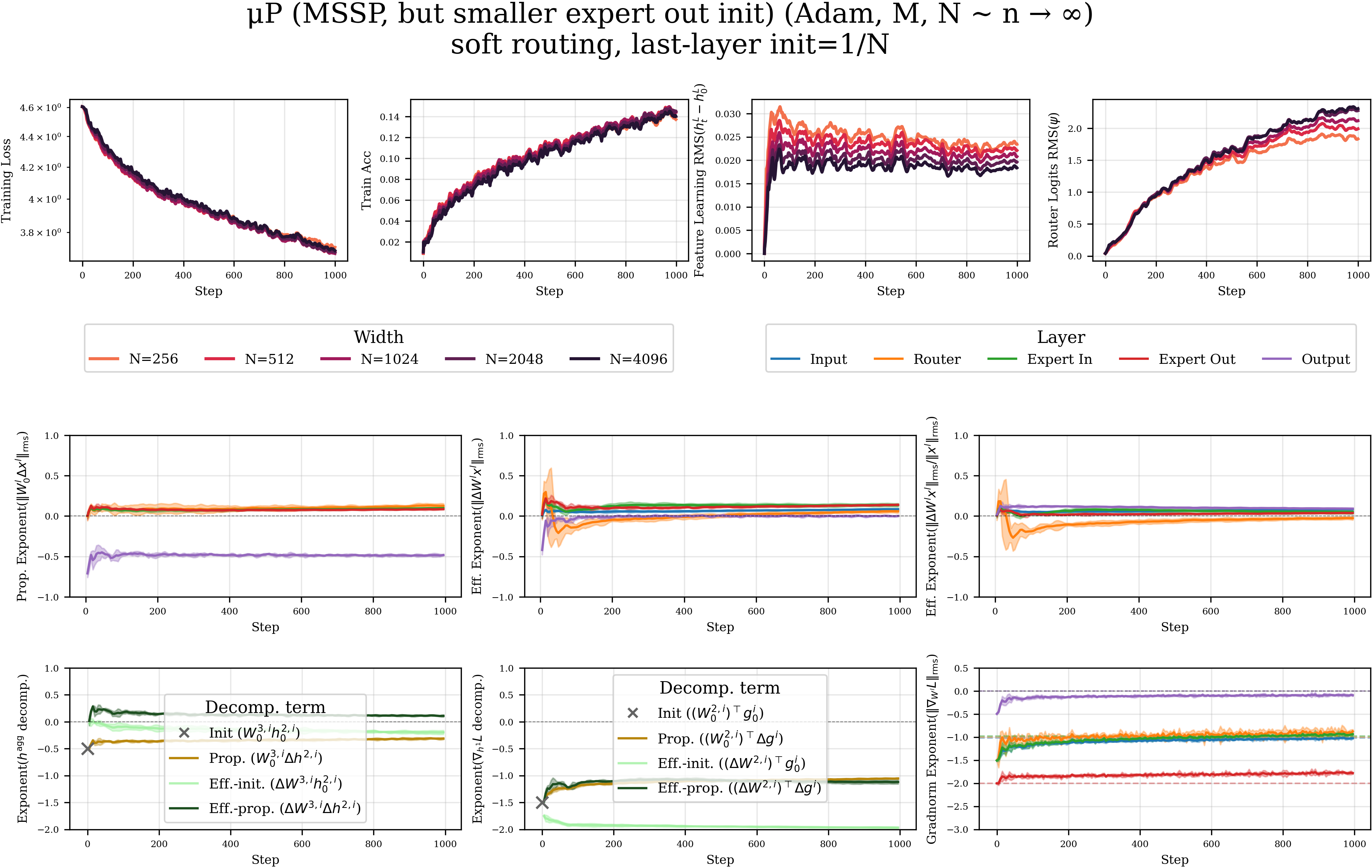}
\caption{\textbf{$\mu$P baseline with 1/N last-layer init (Adam, Regime II).}}
\label{fig:bottleneck_std_adam_soft_ll1n_0429}
\end{figure}

\begin{figure}[H]
\centering
\includegraphics[width=\textwidth]{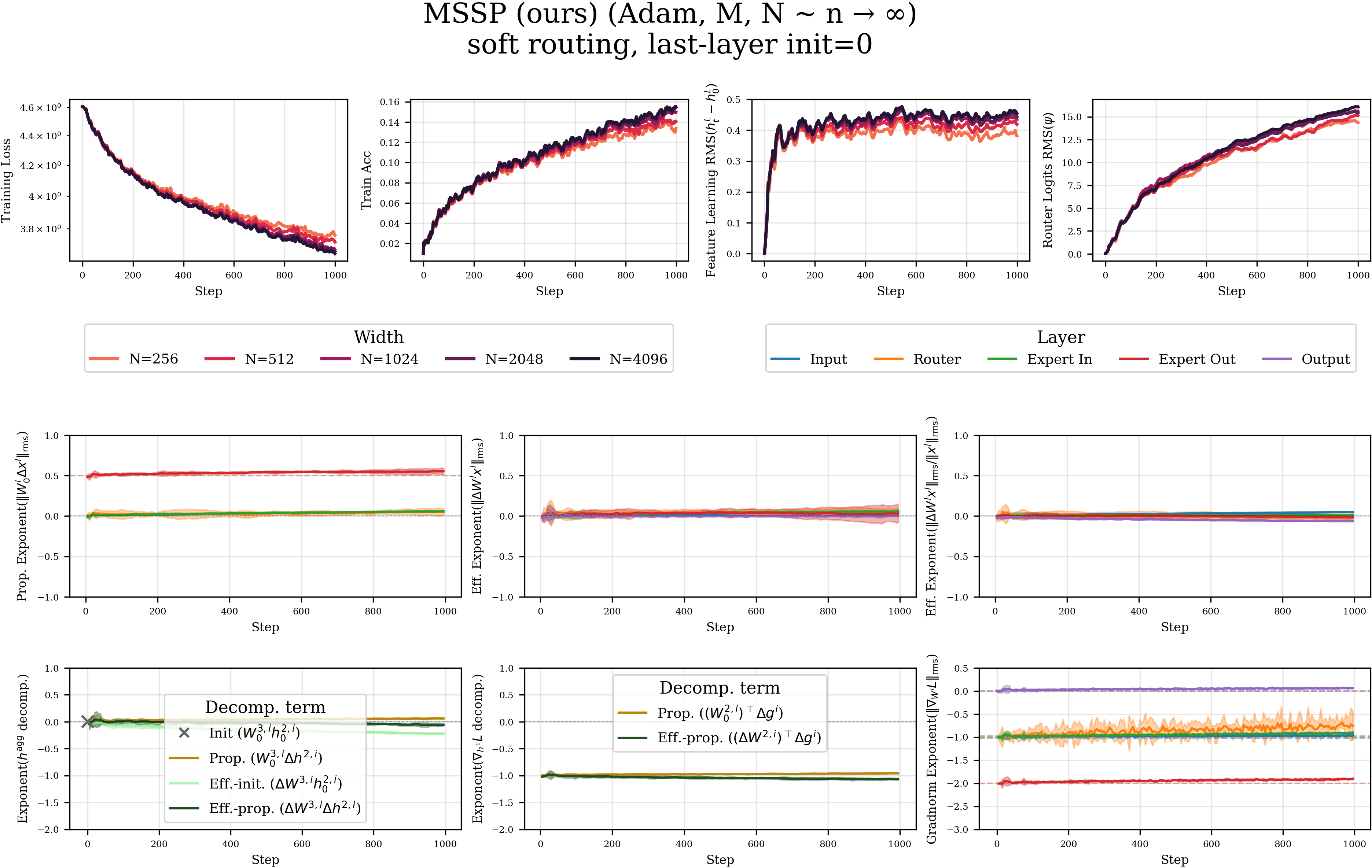}
\caption{\textbf{MSSP (Adam, Regime II).} Note that the propagating update exponent $0.5$ of the expert output layer is desired here, since it is cancelled out by a clean CLT effect in the expert aggregation. The width independence of all other quantities throughout training verifies that the divergence of the individual expert outputs $h^{2,\text{out}}$ through this propagating update term results in approximately width-independent training dynamics.}
\label{fig:bottleneck_ours_adam_soft_ll0_0414}
\end{figure}

\begin{figure}[H]
    \centering

    \begin{subfigure}[b]{0.99\textwidth}
    \centering
    \includegraphics[width=\textwidth]{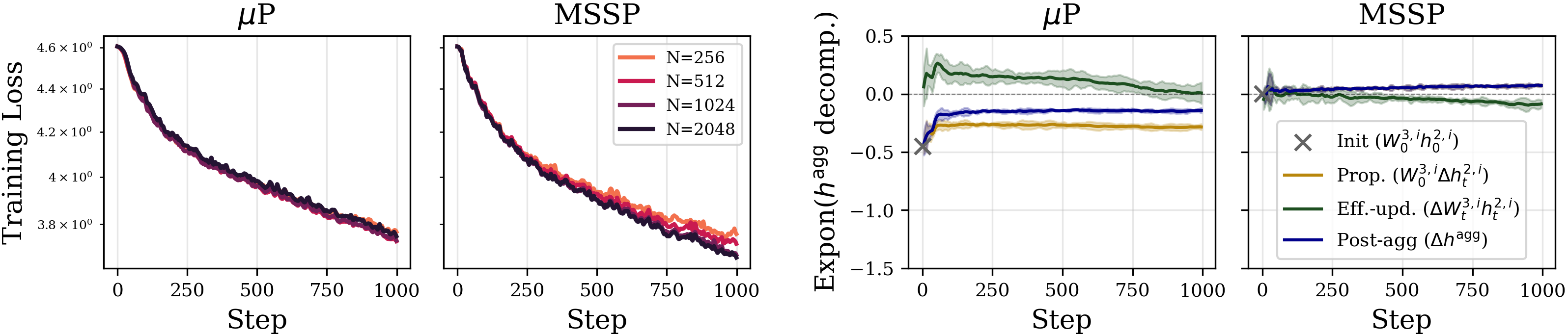}
    \end{subfigure}

    \begin{subfigure}[b]{0.99\textwidth}
    \centering
    \includegraphics[width=\textwidth]{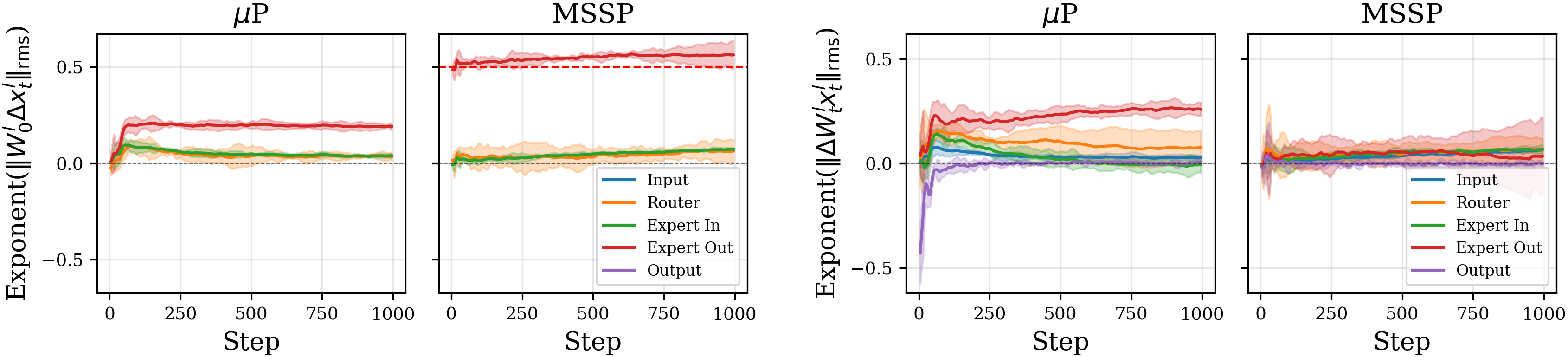}
    \end{subfigure}

    \caption{\textbf{Monotonic improvement with scale only in MSSP, not in $\mu$P (Adam, Regime II).} Same as \Cref{fig:bottleneck_sgd_mup}, but for Adam. While Adam's stability is less severely impacted at moderate model scale, expert output exponents are not clean and performance does not improve with model scale in $\mu$P.} 
    \label{fig:adam_mainfigs_bottleneck}
\end{figure}

\subsubsection{Regime III: Joint proportional scaling}\label{sec:rcc_regime3}

For SGD, observe delayed learning in $\mu$P similar to Regime II. In both $\mu$P and MSSP, the router gradient norm is very noisy, but this does not affect the width-independent signal propagation in MSSP.

Adam in $\mu$P is surprisingly stable in the all-scaling Regime III. Still, Adam's performance only improves robustly and significantly with scale in MSSP, both under soft and top-k routing.

top-k selection does not affect qualitative scaling properties as theoretically predicted.

As in Regime II, note that expert input and output layer gradients are decaying extremely fast with exponent $-2$ as predicted in both SGD and Adam, requiring layerwise gradient or Adam moment scaling to prevent numerical underflows at moderate scale.

\begin{figure}[t]
    \centering

    \begin{subfigure}[b]{0.49\textwidth}
    \centering
    \includegraphics[width=\textwidth]{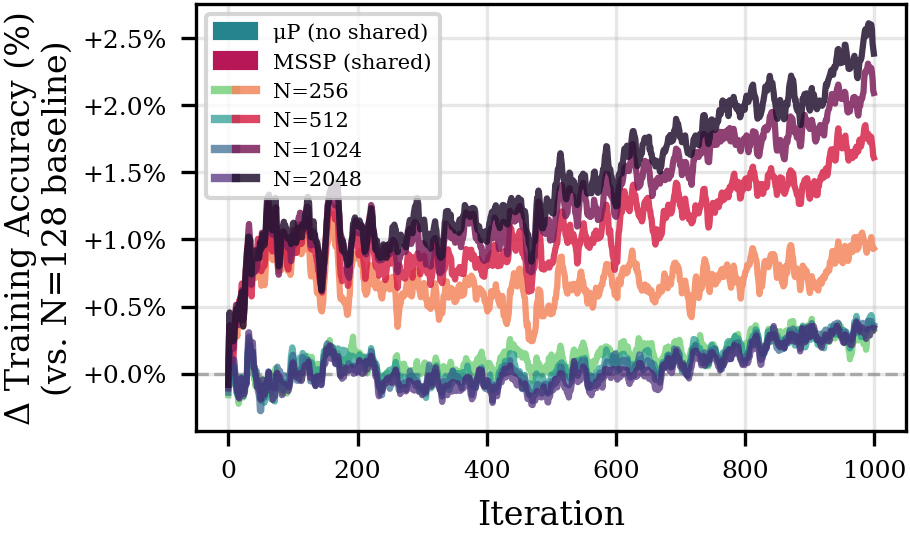}
    \end{subfigure}
    \begin{subfigure}[b]{0.49\textwidth}
    \centering
    \includegraphics[width=\textwidth]{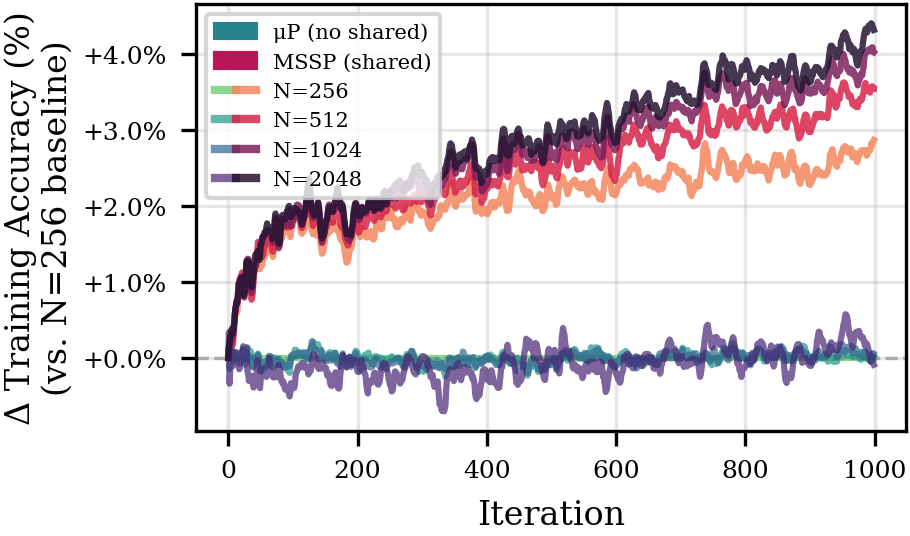}
    \end{subfigure}

    \caption{\textbf{Monotonic improvement with scale only in MSSP, not in $\mu$P (Adam, Regime III).} Training accuracy difference compared to $\mu$P with $1/N$ (left) and $0$ (right) last-layer initialization at $N=128$, with separately tuned multipliers. MSSP outperforms both versions of $\mu$P.}    
    \label{fig:loss_adam_allscaling}
\end{figure}

\begin{figure}[H]
\centering
\includegraphics[width=\textwidth]{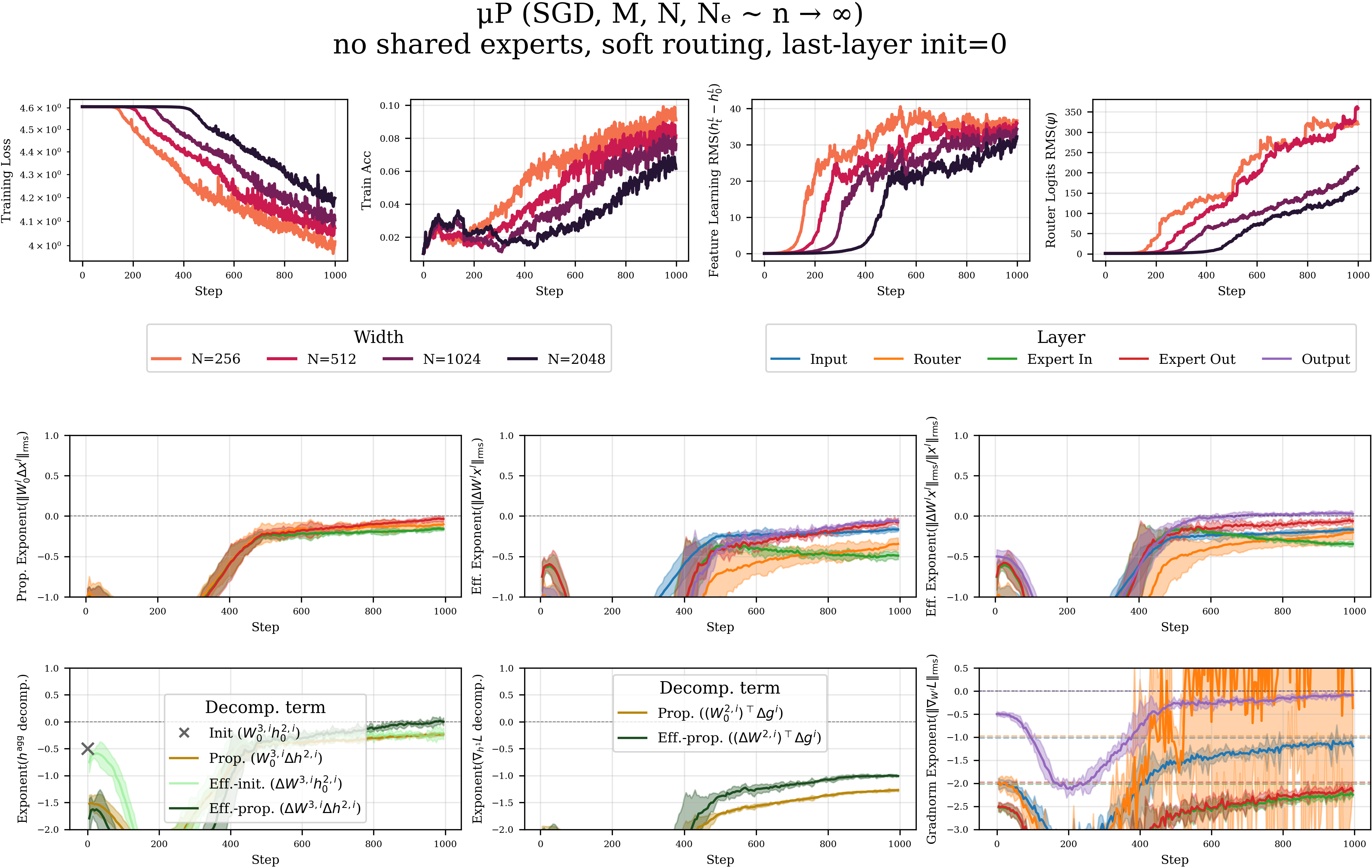}
\caption{\textbf{$\mu$P without shared experts (SGD, Regime III).}}
\label{fig:allscale_std_sgd_soft_nsh_ll0_0415}
\end{figure}

\begin{figure}[H]
\centering
\includegraphics[width=\textwidth]{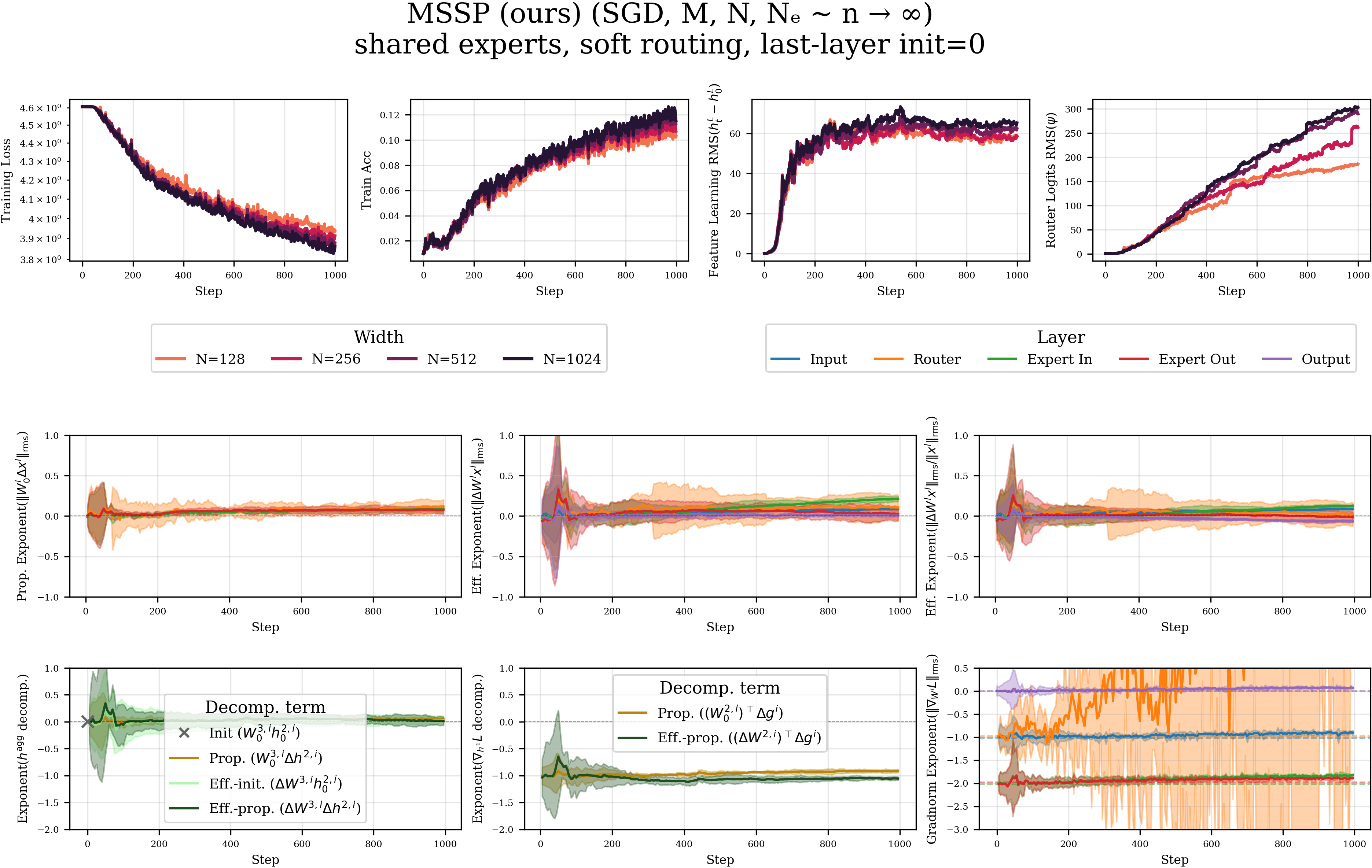}
\caption{\textbf{MSSP with shared experts (SGD, Regime III).}}
\label{fig:allscale_std_sgd_soft_sh_ll0_0419}
\end{figure}

\begin{figure}[H]
\centering
\includegraphics[width=\textwidth]{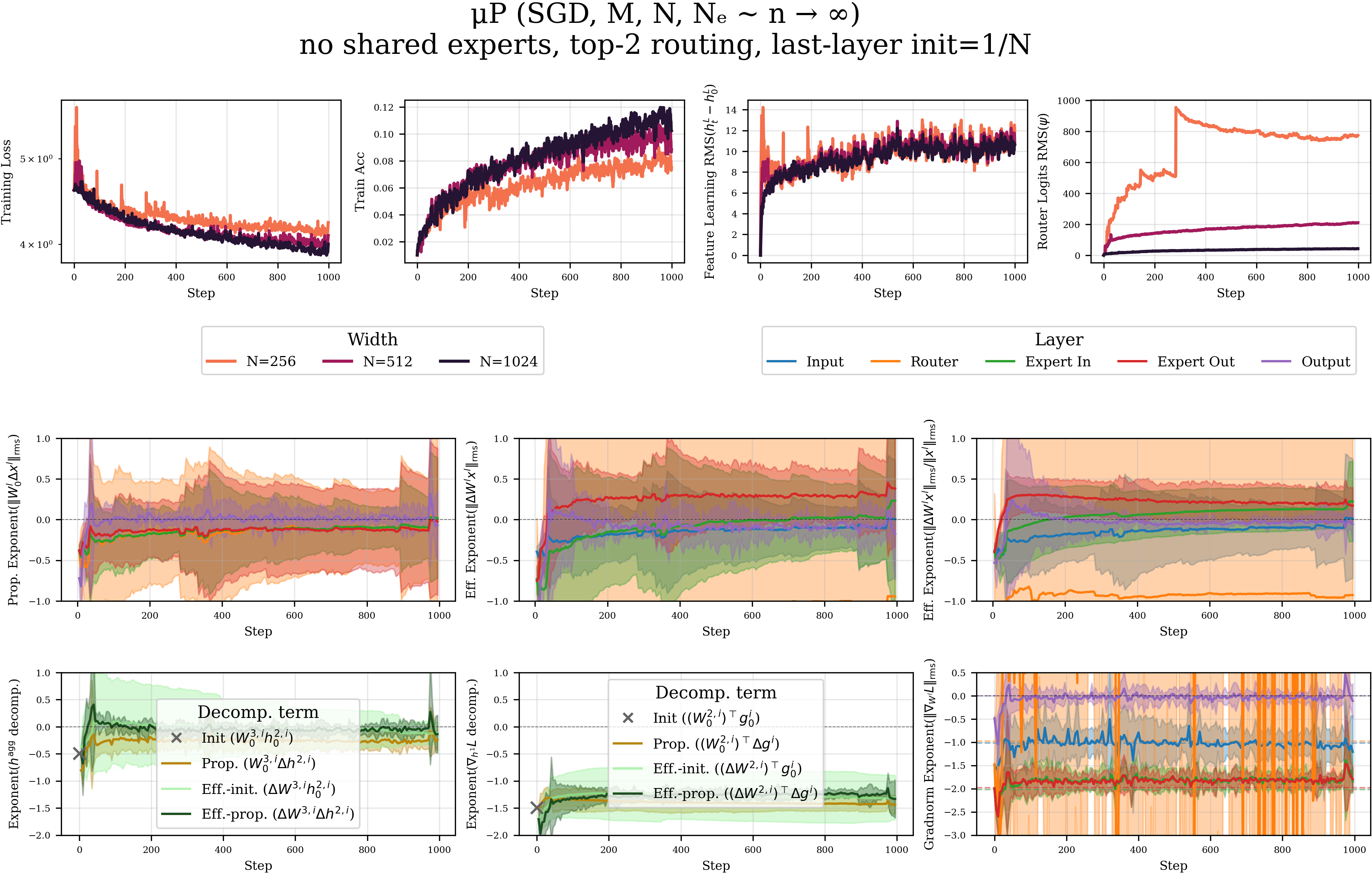}
\caption{\textbf{$\mu$P without shared experts with 1/N last-layer init (SGD, Regime III, top-$k$).}}
\label{fig:allscale_std_sgd_topk2_nsh_ll1n_0420}
\end{figure}

\begin{figure}[H]
\centering
\includegraphics[width=\textwidth]{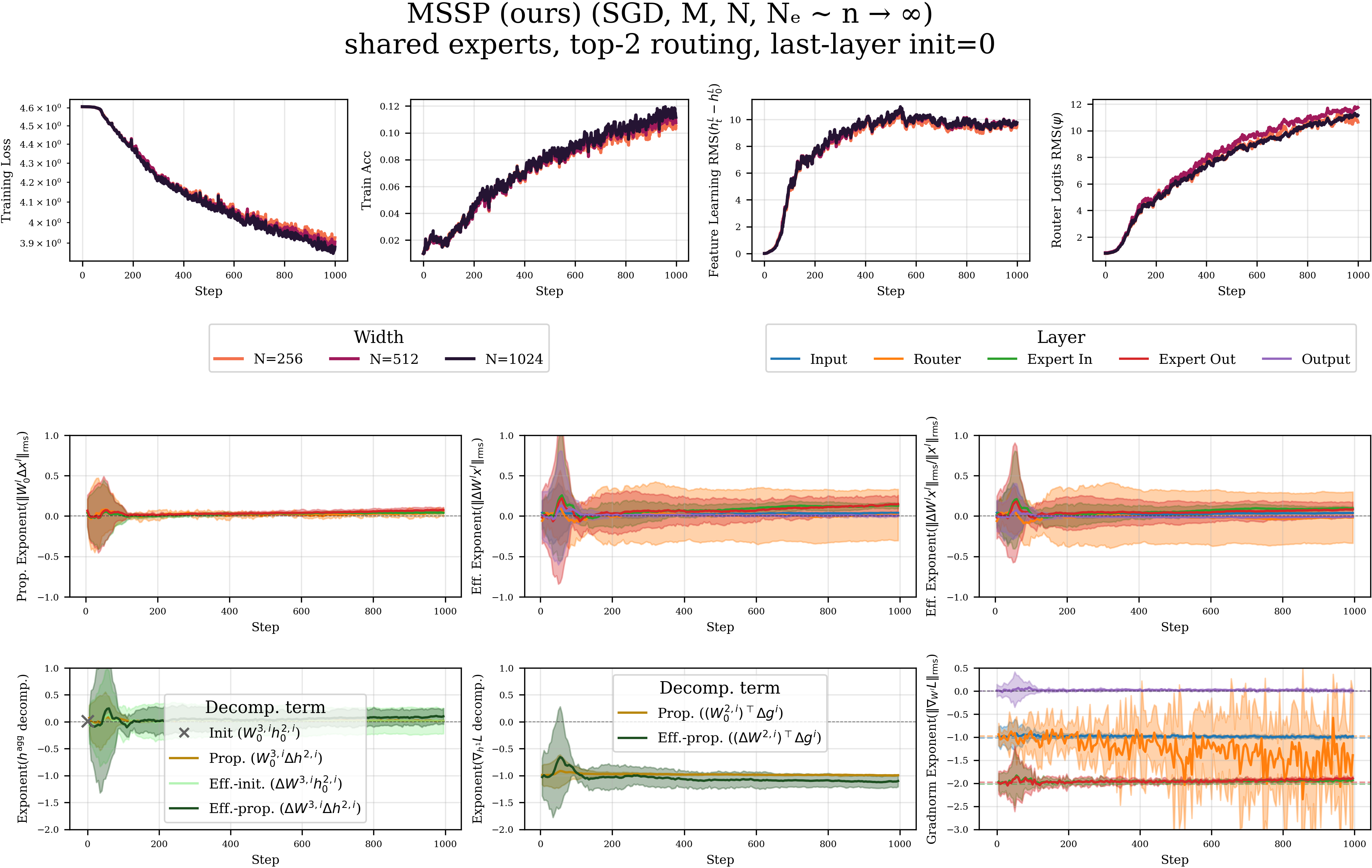}
\caption{\textbf{MSSP with shared experts (SGD, Regime III, top-$k$).}}
\label{fig:allscale_std_sgd_topk2_sh_ll0_0317}
\end{figure}

\begin{figure}[H]
\centering
\includegraphics[width=\textwidth]{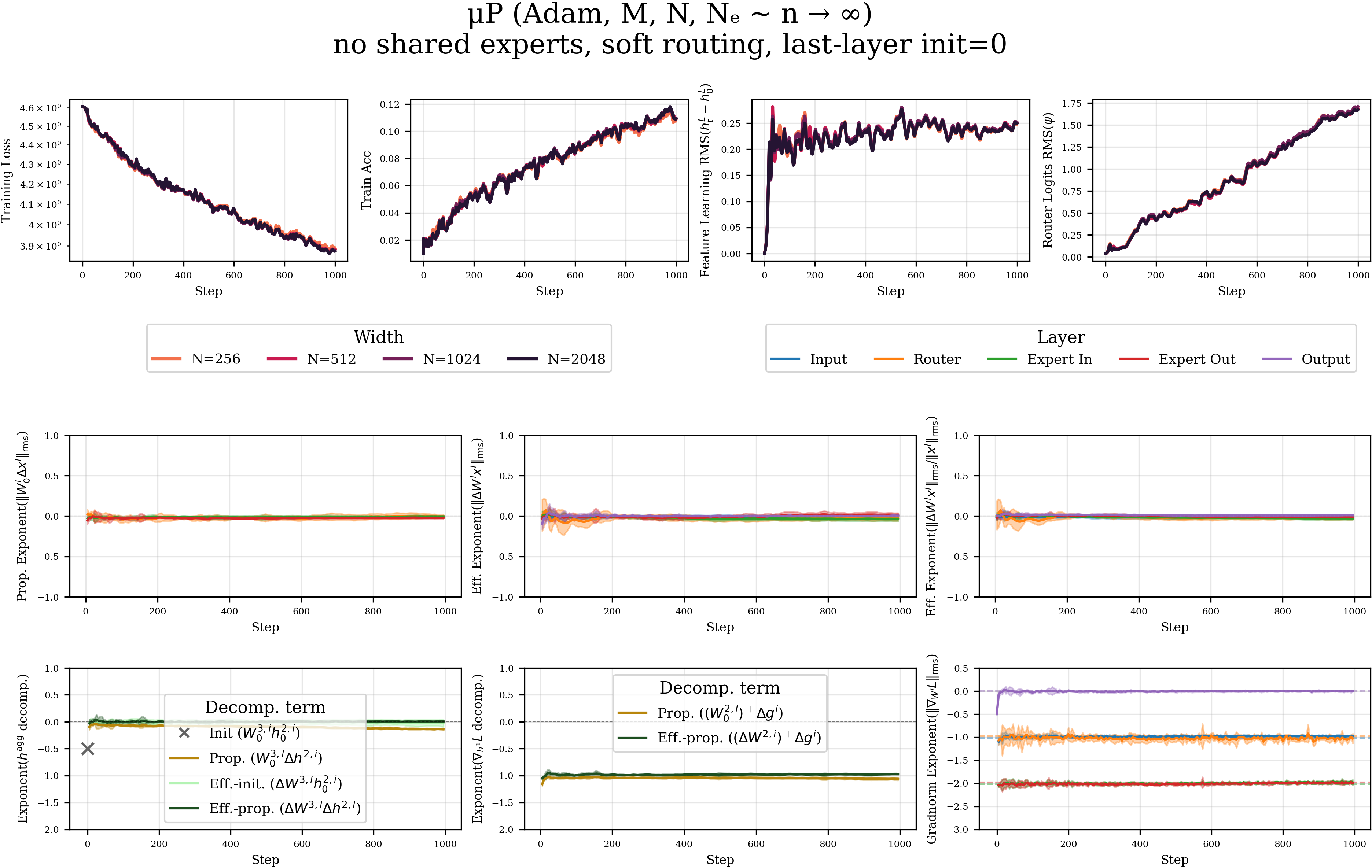}
\caption{\textbf{$\mu$P without shared experts (Adam, Regime III).}}
\label{fig:allscale_ours_adam_soft_nsh_ll0_0415}
\end{figure}

\begin{figure}[H]
\centering
\includegraphics[width=\textwidth]{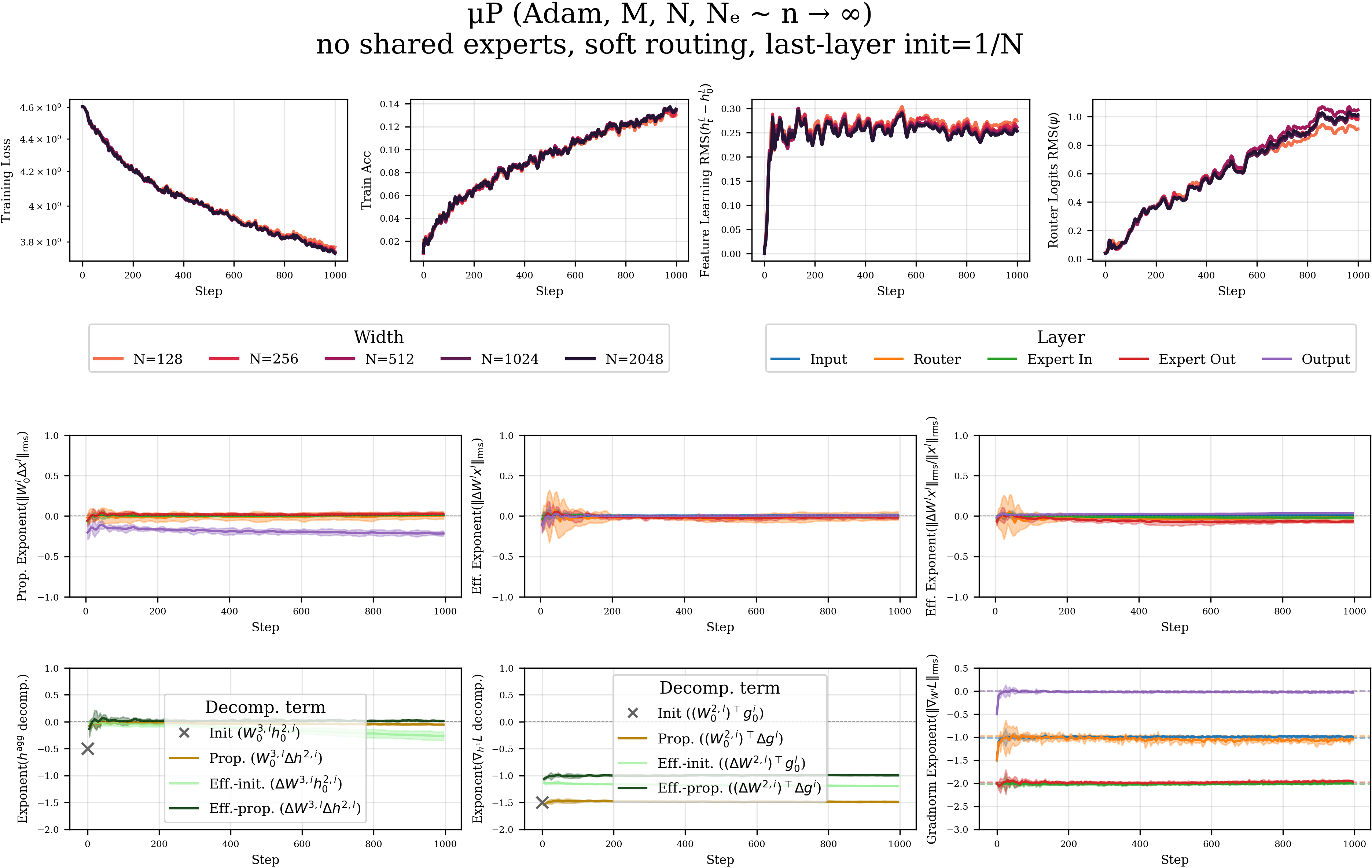}
\caption{\textbf{$\mu$P with 1/N last-layer init, no shared experts (Adam, Regime III).}}
\label{fig:allscale_ours_adam_soft_nsh_ll1n_0427}
\end{figure}

\begin{figure}[H]
\centering
\includegraphics[width=\textwidth]{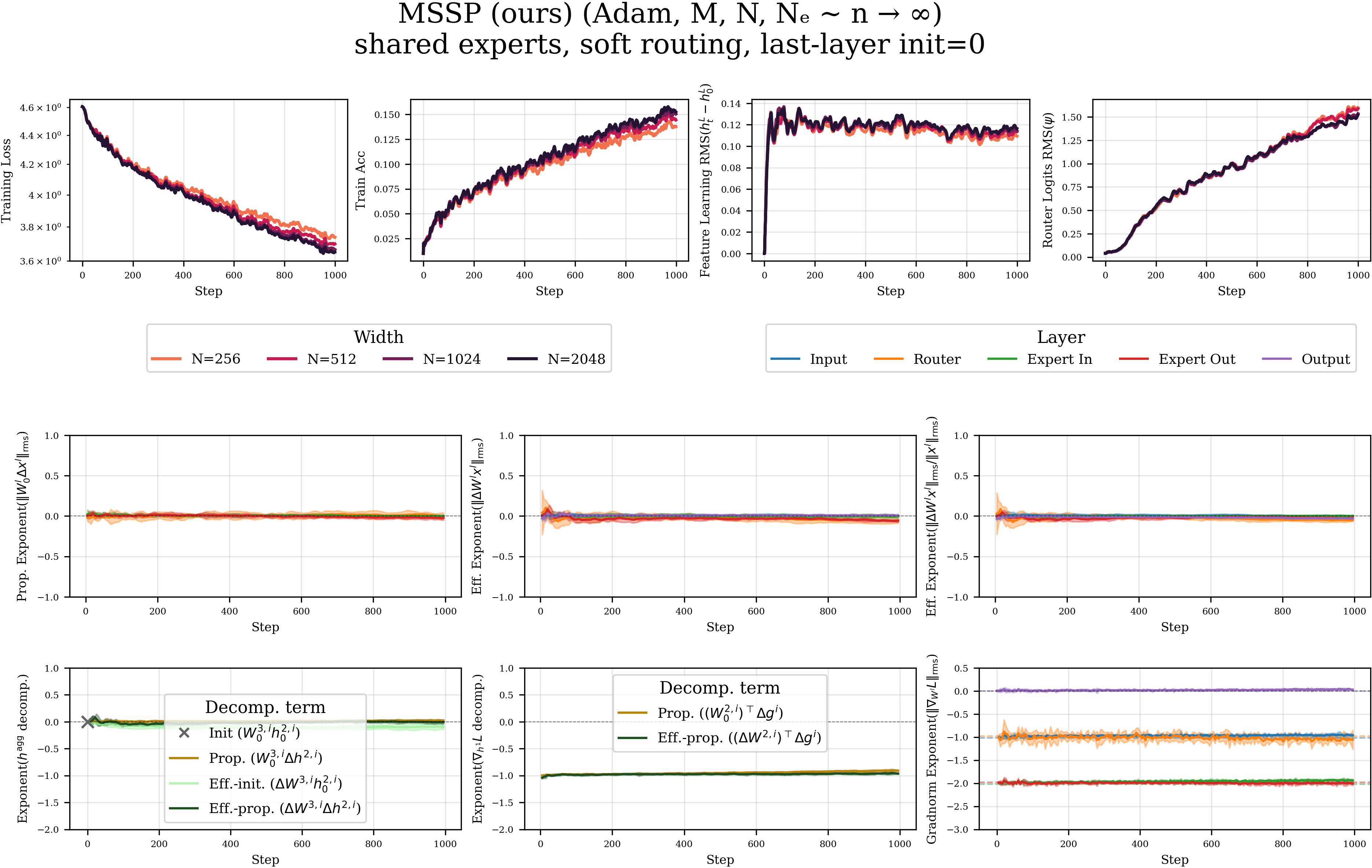}
\caption{\textbf{MSSP with shared experts (Adam, Regime III).}}
\label{fig:allscale_ours_adam_soft_sh_ll0_0418}
\end{figure}

\begin{figure}[H]
\centering
\includegraphics[width=\textwidth]{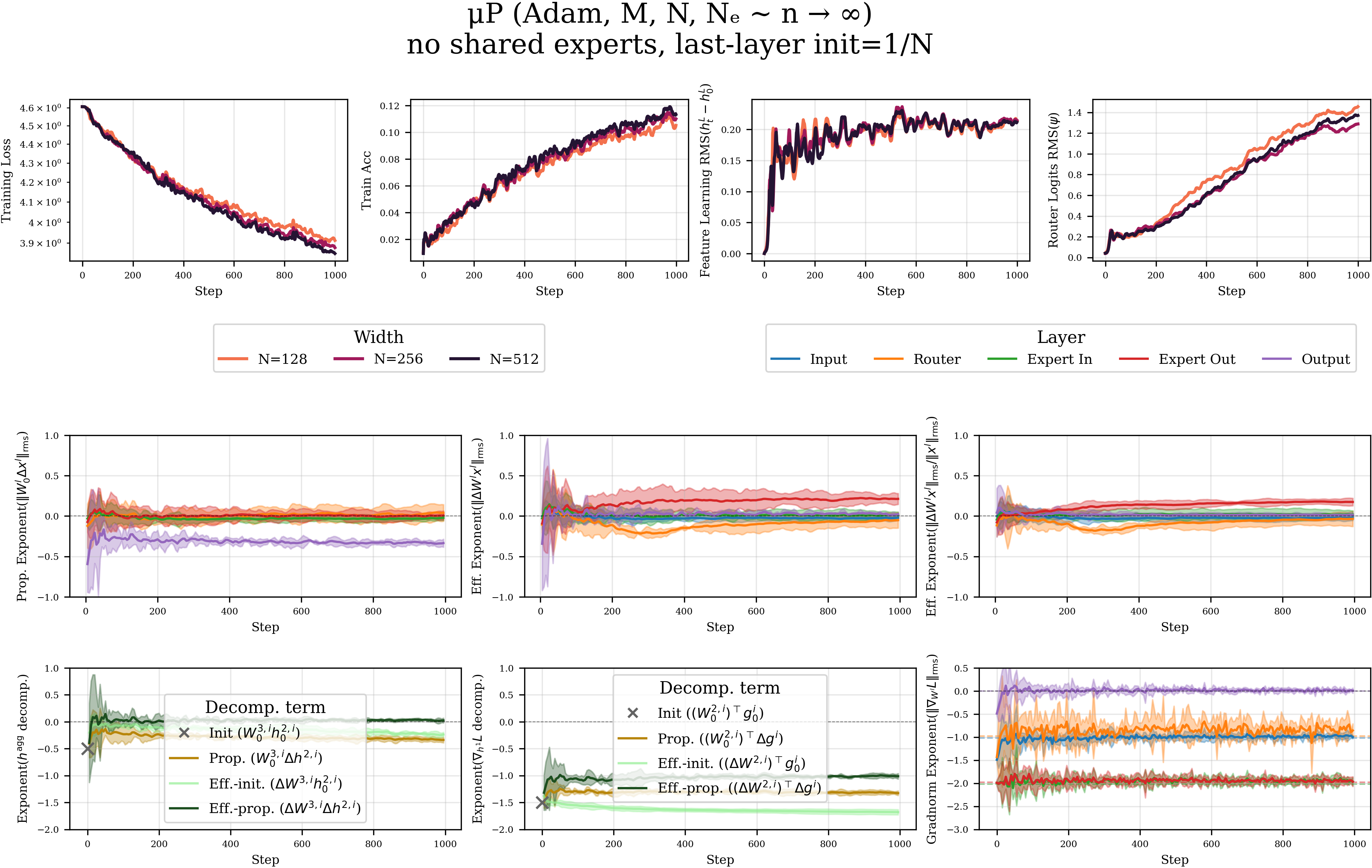}
\caption{\textbf{$\mu$P without shared experts with 1/N last-layer init (Adam, Regime III, top-$k$).}}
\label{fig:allscale_ours_adam_topk_nsh_ll1n_0427}
\end{figure}

\begin{figure}[H]
\centering
\includegraphics[width=\textwidth]{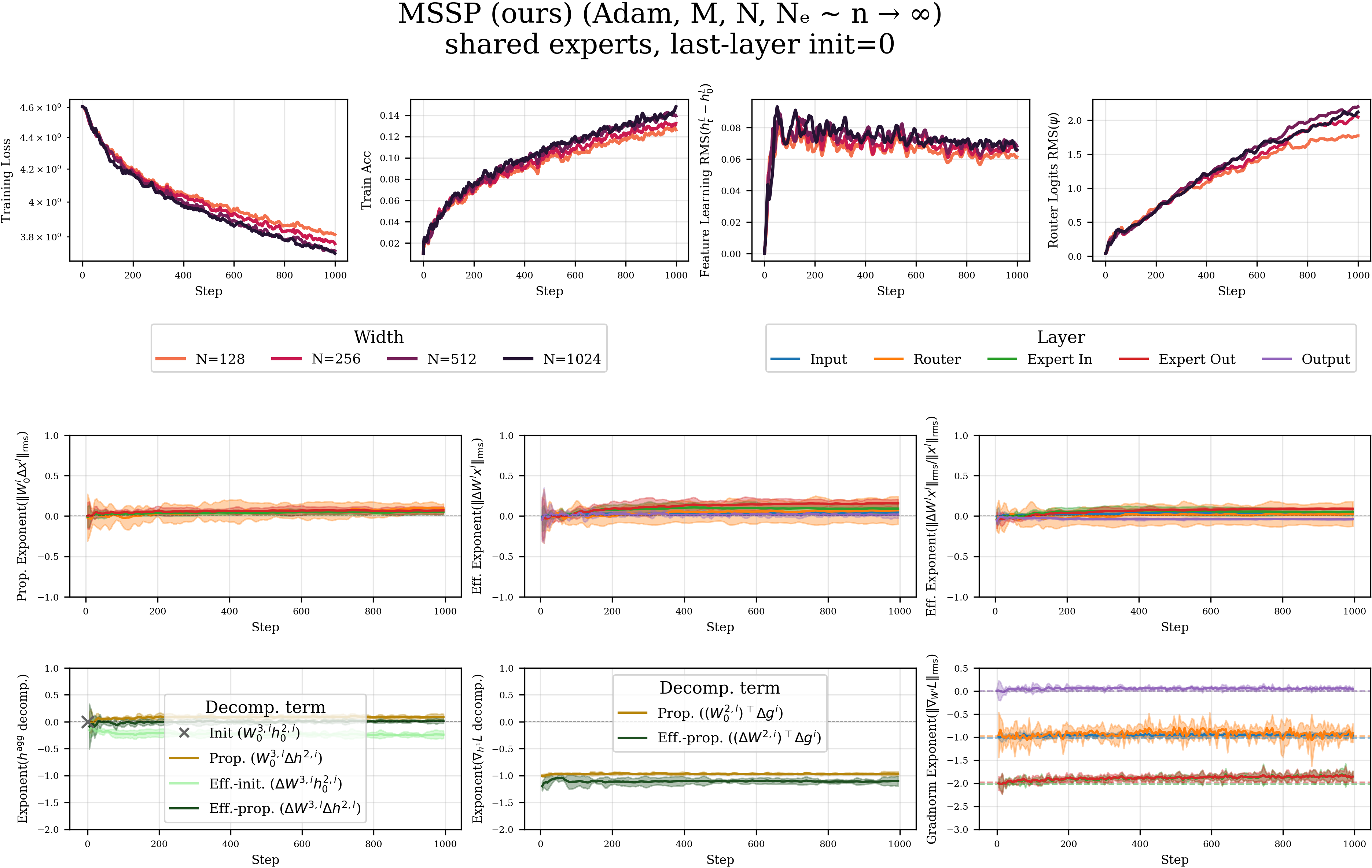}
\caption{\textbf{MSSP with shared experts (Adam, Regime III, top-$k$).}}
\label{fig:allscale_ours_adam_topk_sh_ll0_0424}
\end{figure}

\subsection{Soft softmax routing collapses to uniform for \texorpdfstring{$\mu$}{mu}P in Regime I}\label{sec:exp_softmax_collapse}

Here we train with single pass SGD with the optimal learning rate under MSE loss using soft softmax routing over binary classification from CIFAR-10 using all data points from class 'airplane' $+1$ and class 'automobile' $-1$ with batch size 64. 

\begin{figure}[H]
    \centering
    \includegraphics[width=0.66\textwidth]{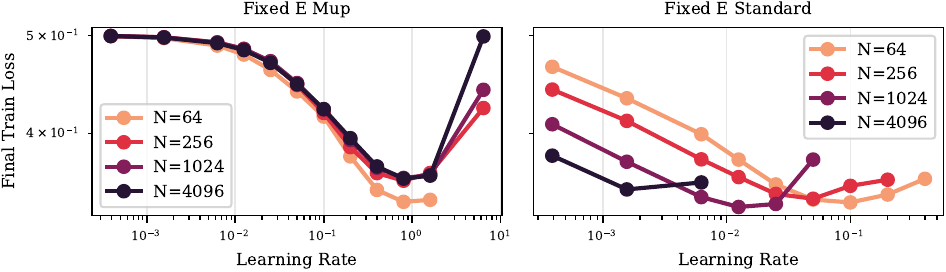}
    
    \vspace{0.5em}
    
    \includegraphics[width=0.66\textwidth]{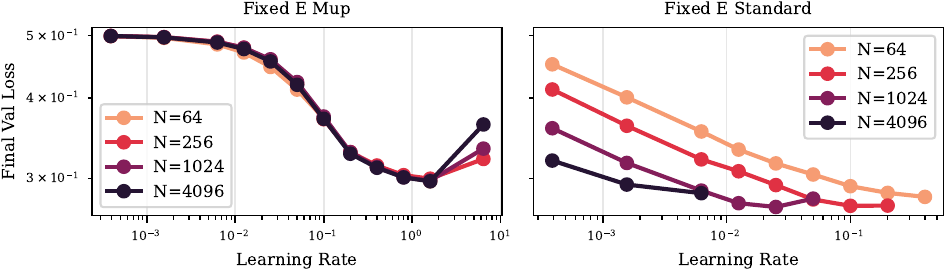}
    
    \caption{\textbf{Learning rate transfer in Regime I (soft softmax routing, MSE loss, CIFAR-10).} Top: training loss. Bottom: validation loss. Left: $\mu$P. Right: SP. Ending lines denote divergence. Observe the maximal stable and optimal learning rate shrinking in SP, but staying consistent in $\mu$P. Performance does not monotonically improve due to router and expert collapse.}
    \label{fig:lr_sweep_joint_fixed_E}
\end{figure}

\Cref{fig:lr_sweep_joint_fixed_E} shows learning rate sweeps scaling with $\mu$P versus standard parameterization (SP) under MSE loss. Lines ending on the right indicates divergence with NaNs under larger learning rates. Validation performance gets worse with scale in SP. Due to router collapse, performance does not improve with scale under soft routing in $\mu$P either. The optimal and maximal stable learning rate shrinks with width in SP, but transfers in $\mu$P. 

\begin{figure}[H]
    \centering

    \begin{subfigure}[b]{0.22\textwidth}
        \includegraphics[width=\textwidth]{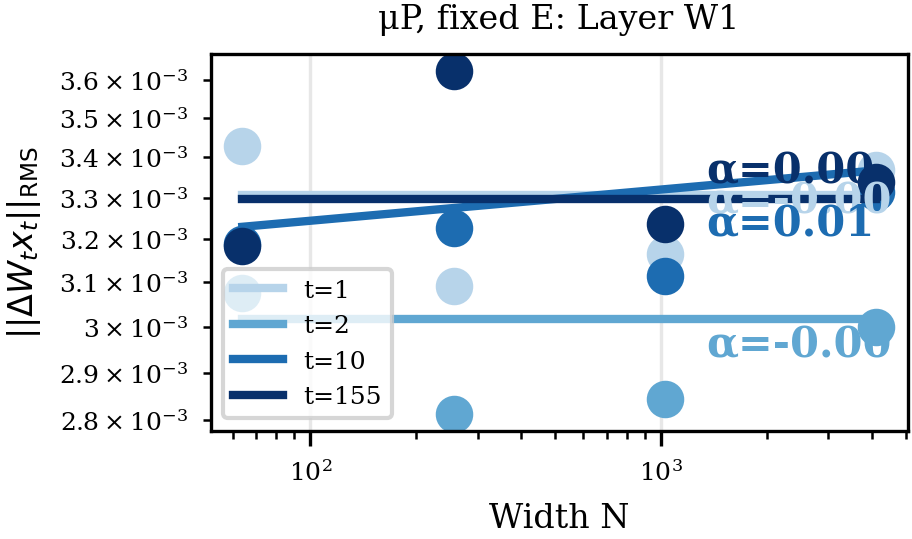}
    \end{subfigure}
    \begin{subfigure}[b]{0.22\textwidth}
        \includegraphics[width=\textwidth]{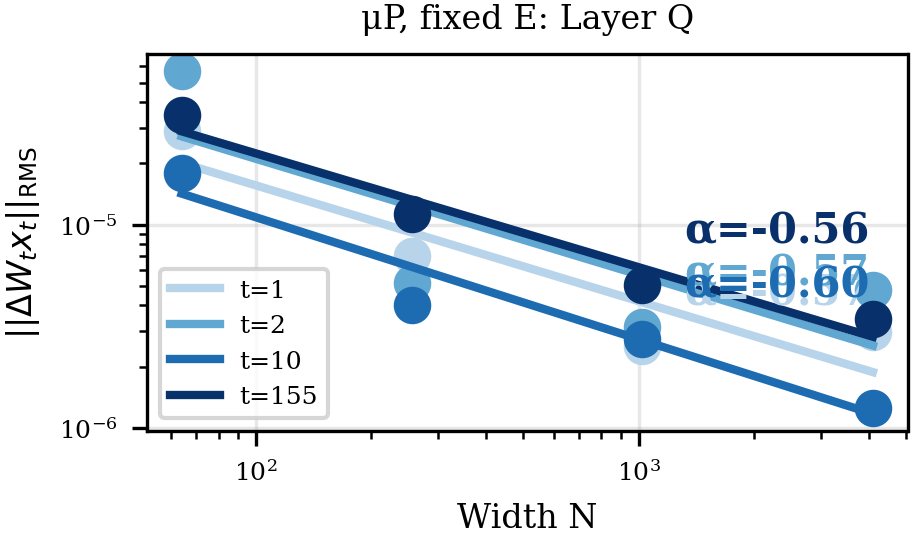}
    \end{subfigure}
    \begin{subfigure}[b]{0.22\textwidth}
        \includegraphics[width=\textwidth]{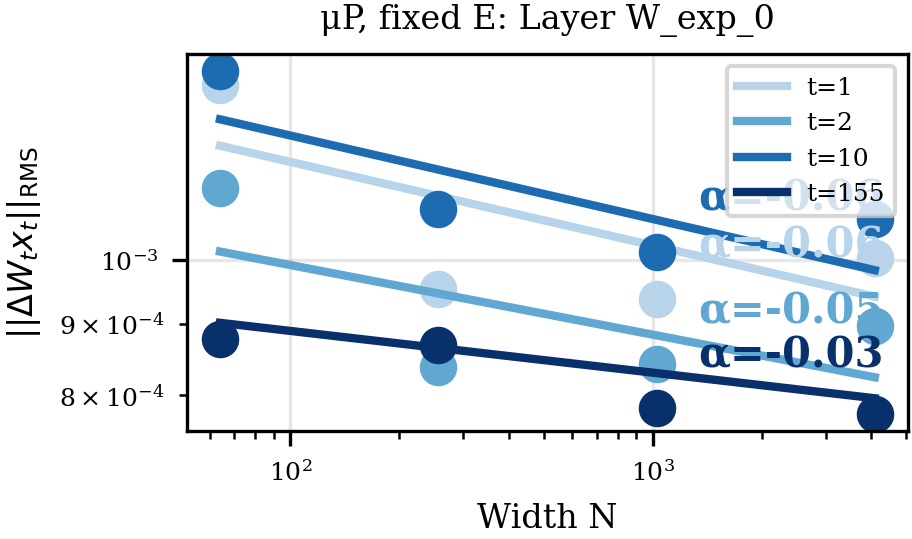}
    \end{subfigure}
    \begin{subfigure}[b]{0.22\textwidth}
        \includegraphics[width=\textwidth]{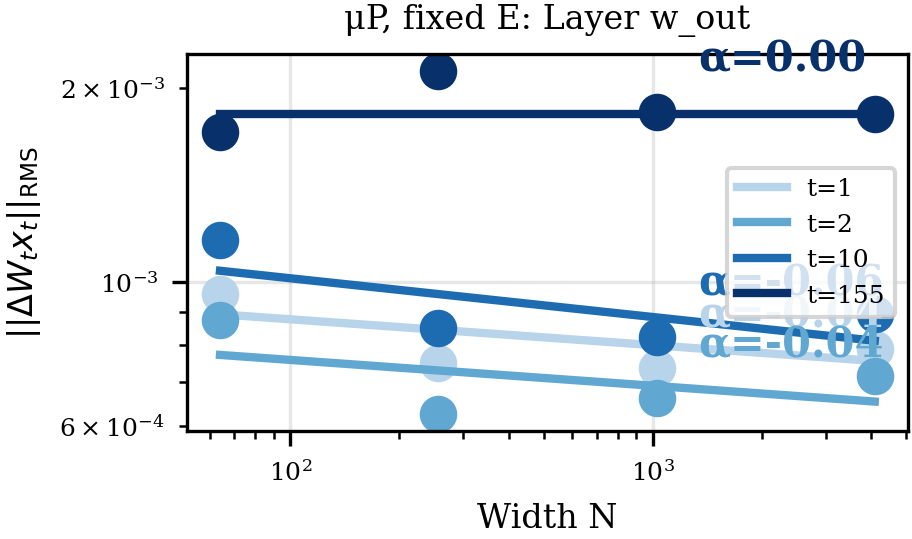}
    \end{subfigure}

    \vspace{0.3em}

    \caption{\textbf{Effective updates across layers ($\mu$P, soft, Regime I).} 
    Columns (left to right): input layer (W1), router layer (Q), first expert layer (W\_exp\_0), output layer (w\_out). Approximate width-independence in $\mu$P, except in the router, which collapses under soft softmax routing.}
    
    \label{fig:coordinate_check_fixed_E_soft}
\end{figure}

\begin{figure}[H]
    \centering

    \begin{subfigure}[b]{0.30\textwidth}
        \includegraphics[width=\textwidth]{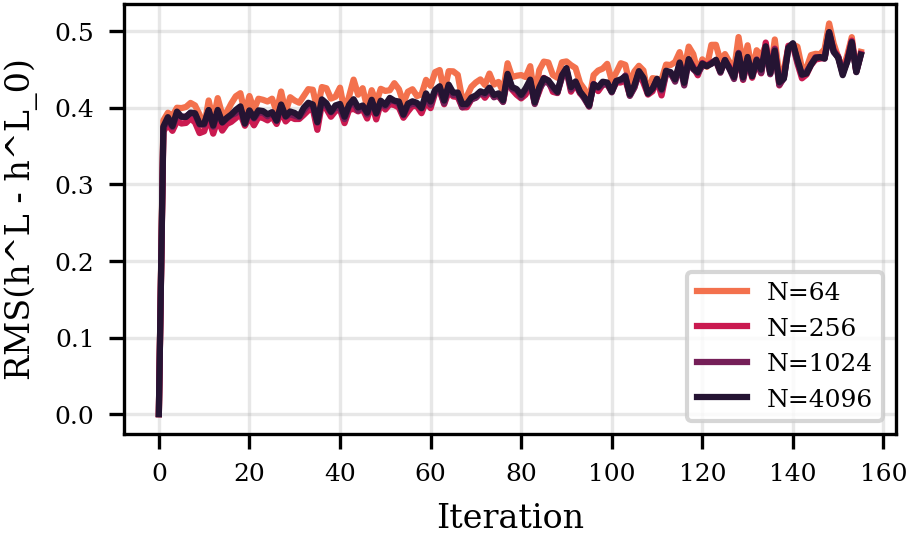}
    \end{subfigure}
    \begin{subfigure}[b]{0.30\textwidth}
        \includegraphics[width=\textwidth]{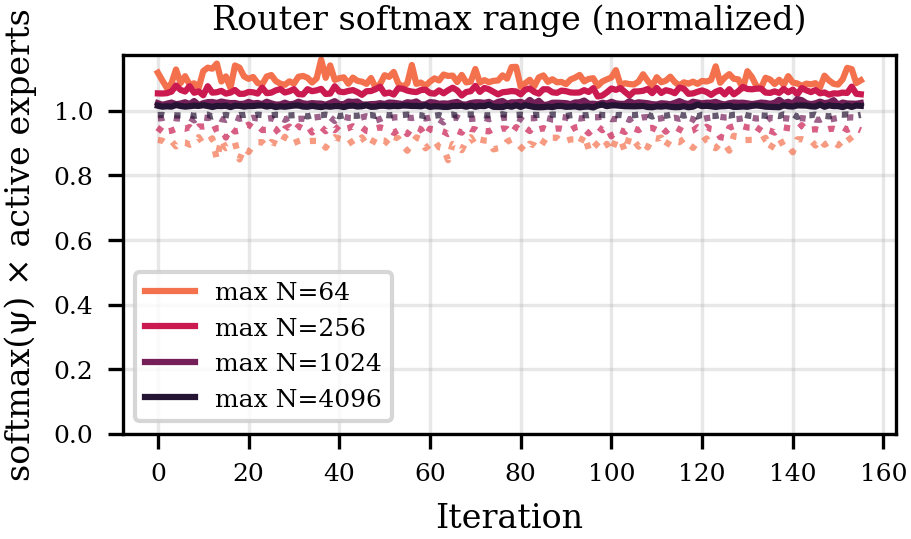}
    \end{subfigure}
    \begin{subfigure}[b]{0.30\textwidth}
        \includegraphics[width=\textwidth]{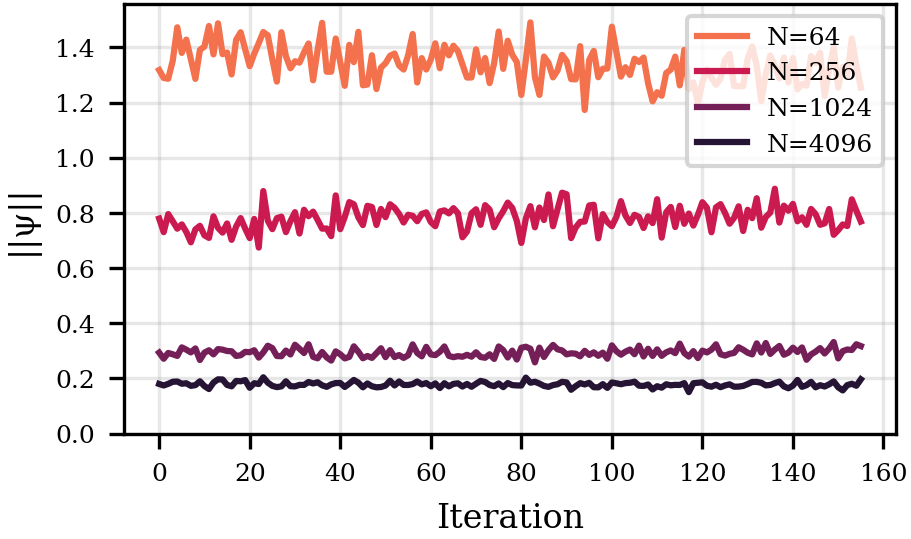}
    \end{subfigure}

    \caption{\textbf{Soft softmax routing collapses to uniform in $\mu$P in Regime I.} Here, we train single pass SGD in $\mu$P over the CIFAR-10 subset described in this section. Columns (left to right): feature learning $\|\Delta h^L\|_{RMS}$, normalized post-softmax routing weights, router logit norm $\|\psi\|_{RMS}$. 
    Routing collapses as routing logits converge to $0$ with scale over the entire course of training. Feature learning is still approximately width-independent.}

    \label{fig:training_dynamics_fixed_E_soft}
\end{figure}

As theoretically predicted, the router gradient vanishes with increased width (\Cref{fig:coordinate_check_fixed_E_soft}), which results in vanishing router updates inducing vanishing router logits inducing uniform routing. \Cref{fig:training_dynamics_fixed_E_soft} shows that this collapse prevails over the entire course of training.

Across several settings, we found that expert specialization does not necessarily improve performance on CIFAR-10, as the dataset is too small and not diverse enough. Hence we use TinyImageNet for all other MLP experiments.

\subsection{MoEs require layerwise learning rate tuning}\label{sec:exp_mult_tuning}

\begin{figure}[H]
\centering
\includegraphics[width=\textwidth]{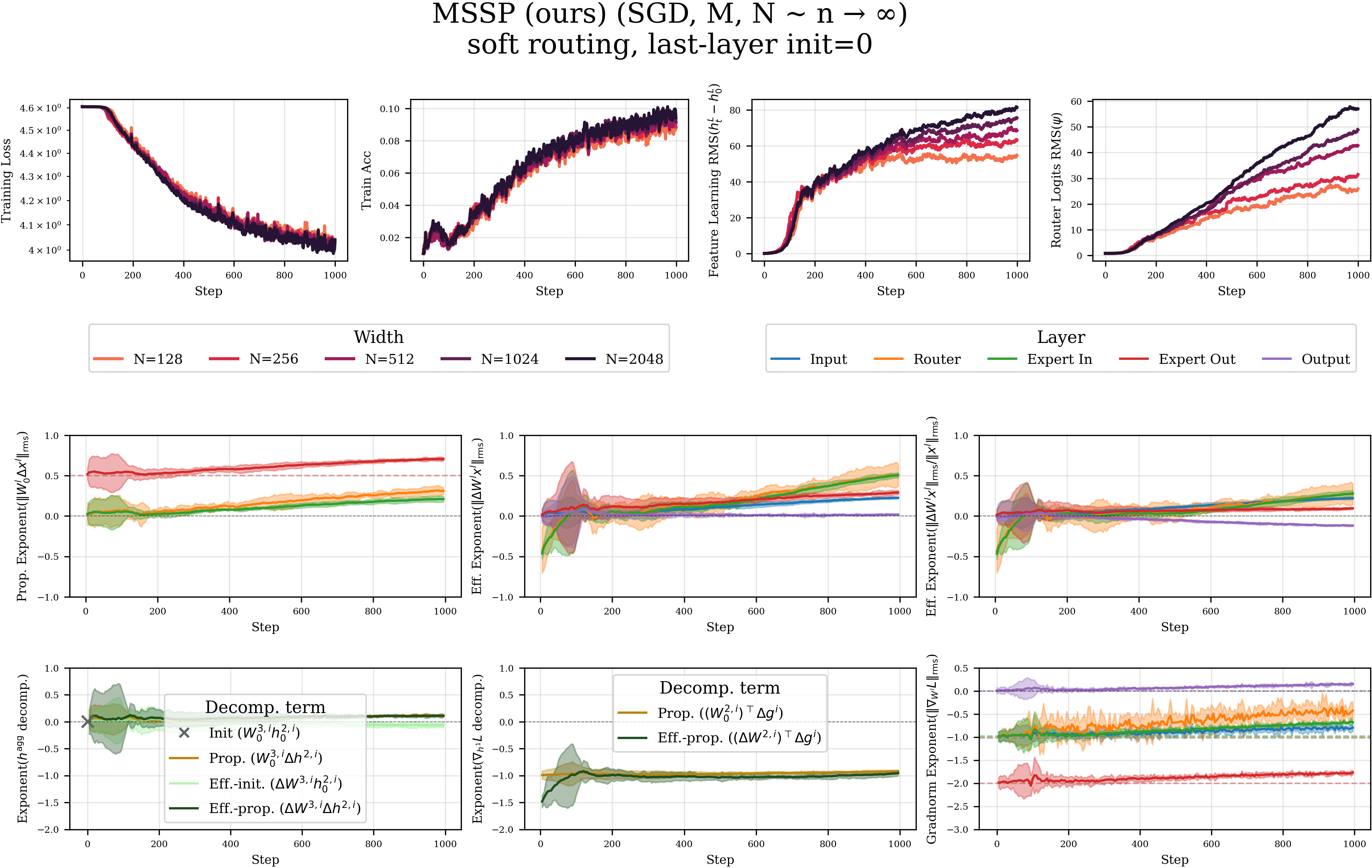}
\caption{\textbf{MSSP without tuned multipliers (SGD, Regime II).} Here we set all initialization variance and layerwise learning rate multipliers to $1.0$ and only tune the global learning rate.}
\label{fig:multtuning_rcc}
\end{figure}

Without layerwise learning rate multiplier tuning, effective updates in router and expert input layer can initially vanish even in MSSP (\Cref{fig:multtuning_rcc}).

\begin{figure}[H]
    \centering

    \begin{subfigure}[b]{0.24\textwidth}
        \includegraphics[width=\textwidth]{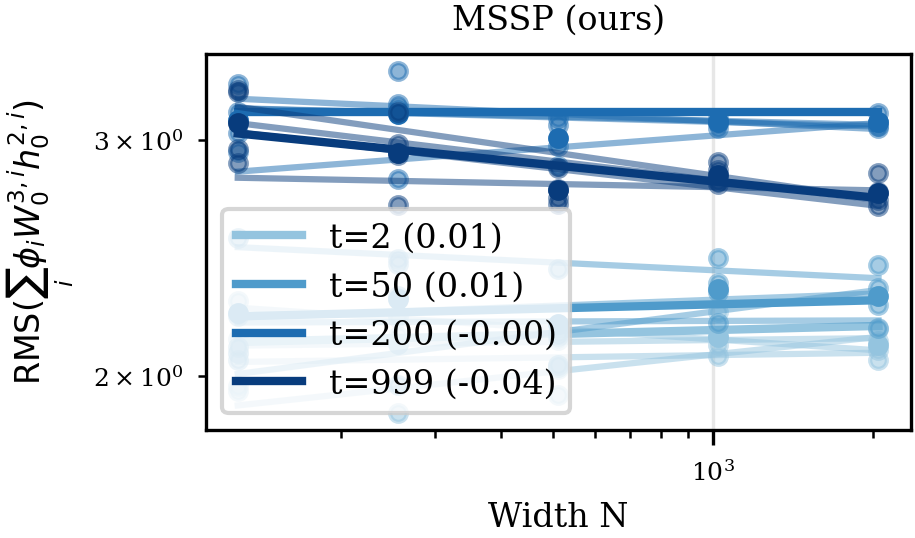}
    \end{subfigure}
    \begin{subfigure}[b]{0.24\textwidth}
        \includegraphics[width=\textwidth]{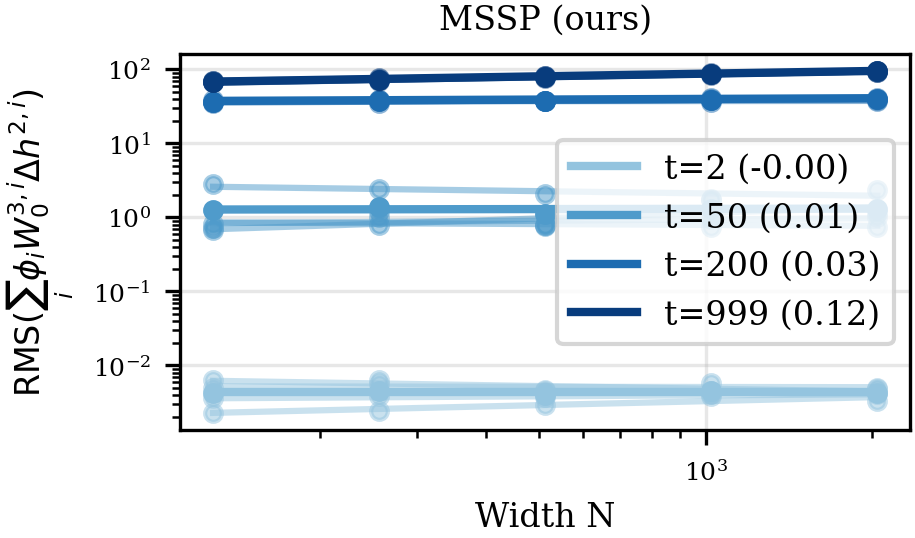}
    \end{subfigure}
    \begin{subfigure}[b]{0.24\textwidth}
        \includegraphics[width=\textwidth]{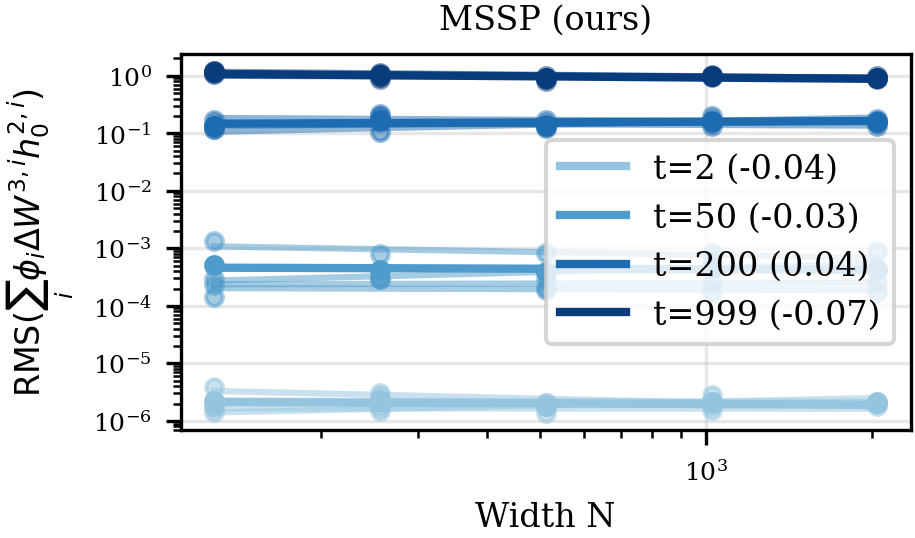}
    \end{subfigure}
    \begin{subfigure}[b]{0.24\textwidth}
        \includegraphics[width=\textwidth]{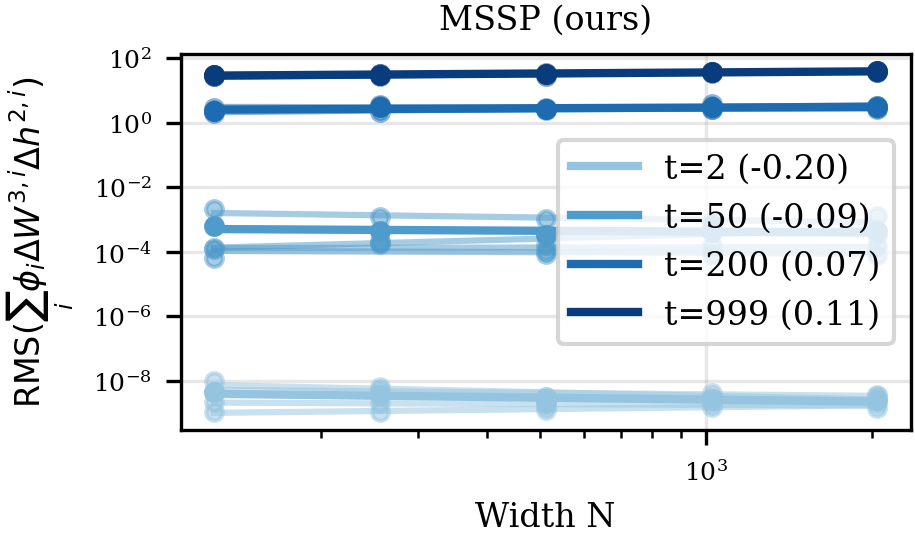}
    \end{subfigure}

    \begin{subfigure}[b]{0.49\textwidth}
        \includegraphics[width=\textwidth]{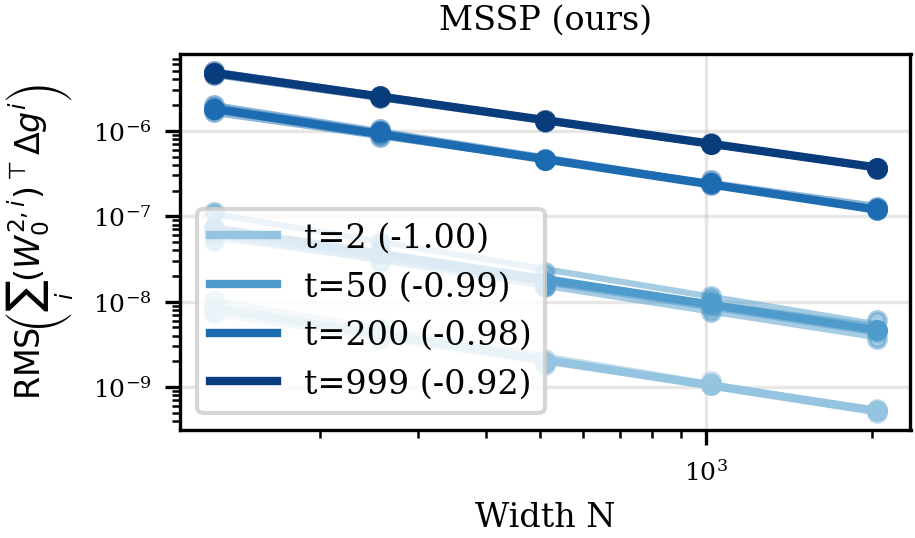}
    \end{subfigure}
    \begin{subfigure}[b]{0.49\textwidth}
        \includegraphics[width=\textwidth]{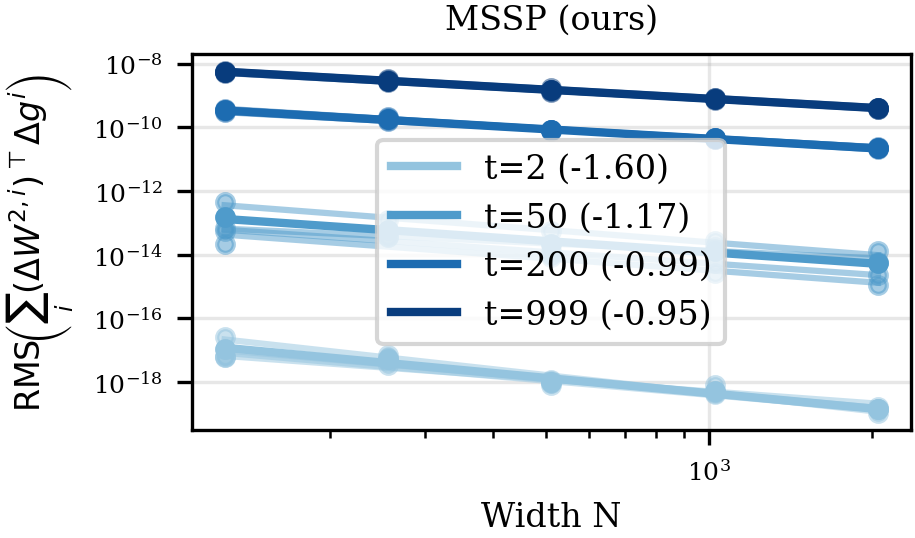}
    \end{subfigure}

    \caption{\textbf{Without tuned multipliers, sub-term contributions differ by orders of magnitude (MSSP, SGD, Regime II).} The forward pass expert aggregation components (top) and the backward pass components to $\nabla_{h^1}\mathcal{L}_t$. Propagating updates dominate the effective updates in absolute scale for several hundred steps. In the forward pass, propagating updates start out at around $10^0$ and effective updates around $10^{-6}$. In the backward pass, the propagating update term starts around $10^{-9}$ at $N=1024$ and $t=2$, whereas the effective update term starts around much smaller $10^{-18}$. Hence the weight updates contribute vanishingly to the overall first-layer gradient $\nabla_{h^1}\mathcal{L}_t$.}

    \label{fig:tuning_constants}
\end{figure}

\Cref{fig:tuning_constants} shows that, while all exponents are approximately $0$ as theoretically predicted, the propagating updates dominate the effective updates in absolute scale in the first steps by a factor of more than $10^6$ in both the forward and backward pass aggregation operations. 

Hence, in parameterizations where correctly scaled effective updates should dominate vanishing propagating updates, the propagating updates can still dominate in absolute value for several hundred steps at realistic scales. Hence it can seem that the empirical exponents do not follow the predicted ones, when not measured in sufficient granularity.

After layerwise learning rate tuning, which can consequently require grids with ranges beyond $10^6$, we observe that propagating and effective updates generally end up having a similar order of magnitude in the first 1000 steps, and hence we accurately measure the predicted limiting exponents at moderate model scales.

Overall this highlights that extensively tuning layerwise learning rate multipliers is even more essential in MoEs than in dense models.

\subsection{Random search and 2D multiplier tuning do not suffice}\label{sec:2d_tuning}

Instead of full 6D sweeps, one could also tune multipliers more efficiently by running a random search, and, starting from the HPs found in the first stage, doing 2D sweeps of all HP pairs around the optimum. This approach finds near-optimal HPs if the suboptimality of the random stage is predominantly restricted to a subspace allowing 2 interacting multipliers.

\begin{figure}[H]
    \centering
    \begin{subfigure}[t]{0.32\textwidth}
        \centering
        \includegraphics[width=\textwidth]{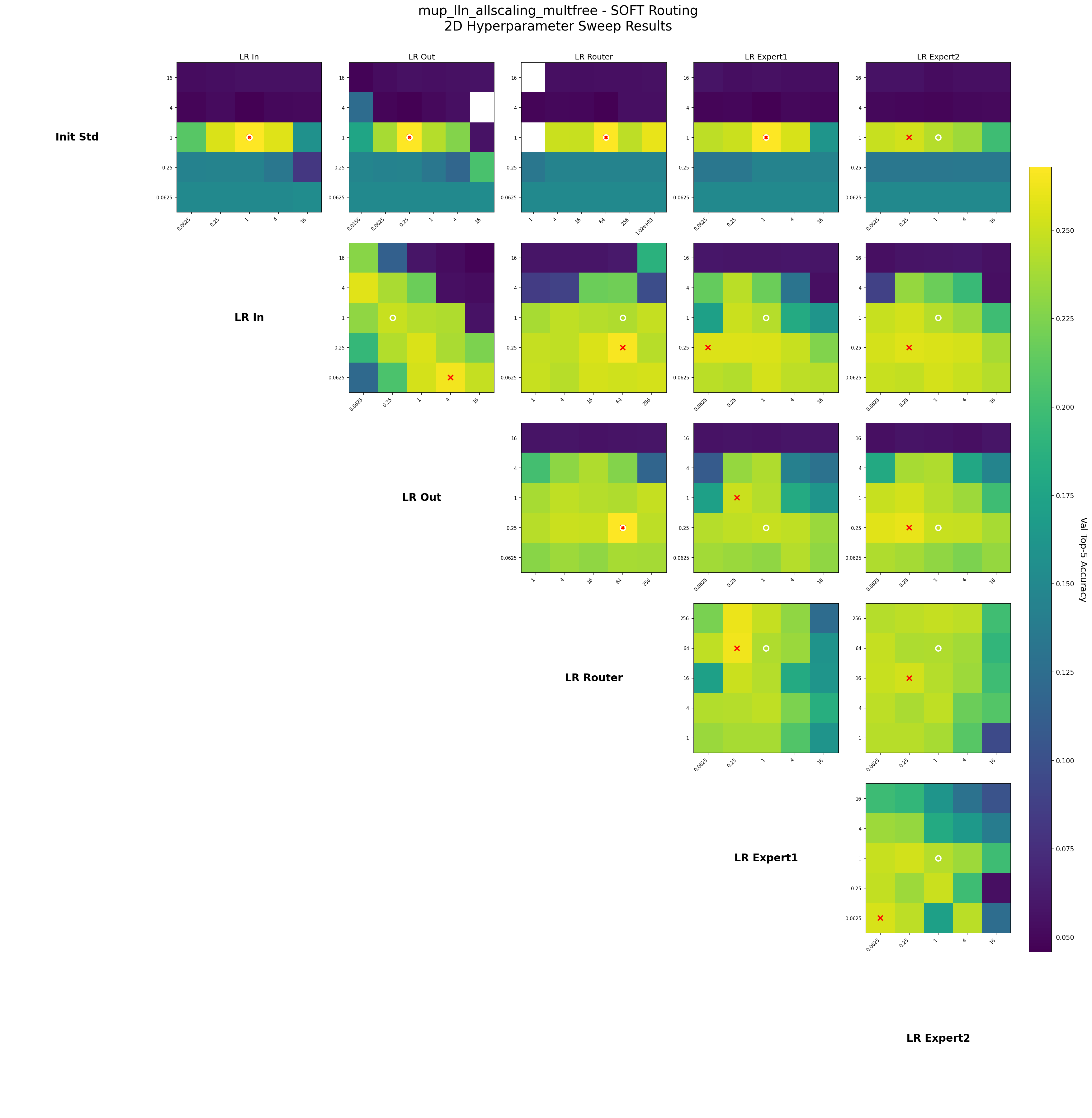}
    \end{subfigure}
    \hfill
    \begin{subfigure}[t]{0.32\textwidth}
        \centering
        \includegraphics[width=\textwidth]{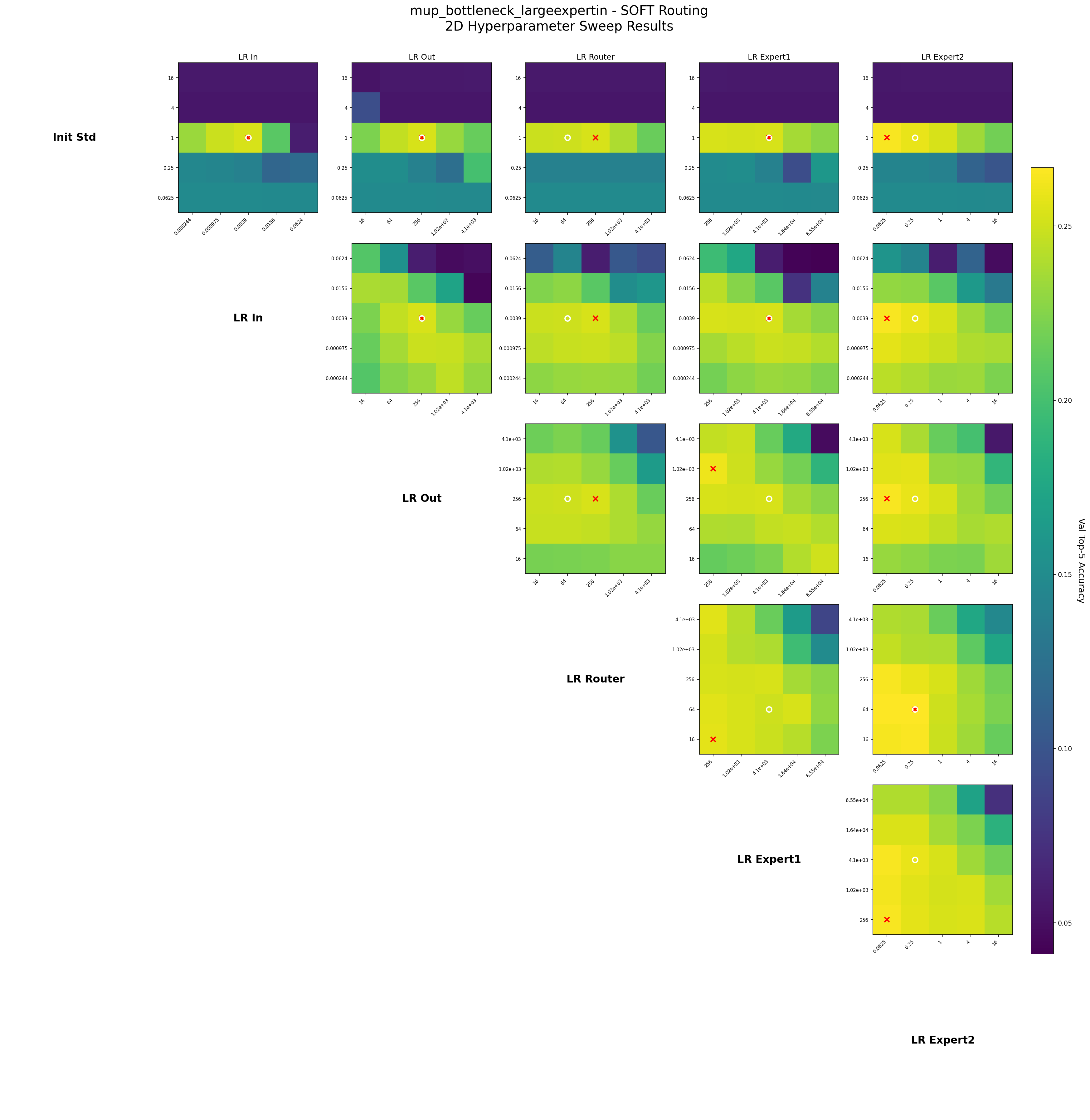}
    \end{subfigure}
    \hfill
    \begin{subfigure}[t]{0.32\textwidth}
        \centering
        \includegraphics[width=\textwidth]{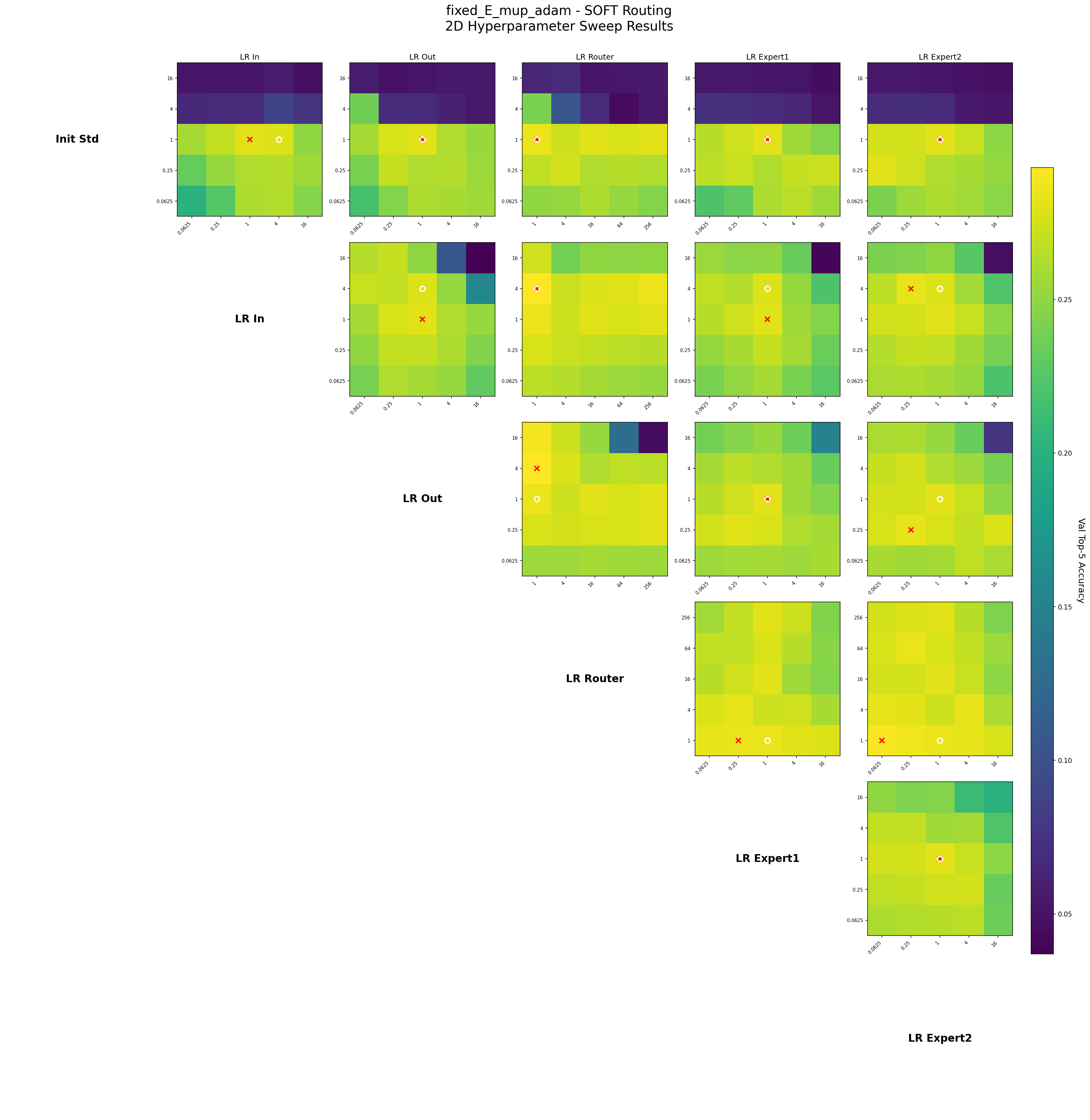}
    \end{subfigure}

    \caption{\textbf{2D multiplier sweeps at small scale $N=128$ from random search stage.} 2D heatmaps showing top-5 validation accuracy of all HP pairs, while fixing all other multipliers at the optimum found from a random sweep over at least $1024$ runs. From left to right: SGD $\mu$P Regime III, SGD $\mu$P Regime II, Adam MSSP Regime I. White circles denote the optimum after this 2D stage, red 'x' denote the optimum within the respective 2D sweep. Often these do not align, suggesting that higher dimensional interactions make this hyperparameter tuning strategy insufficient.}
    \label{fig:2dsweeps_sgd}
\end{figure}

However, \Cref{fig:2dsweeps_sgd} shows that 2D sweeps after a random sweep do not suffice. For example in SGD $\mu$P Regime III (left), all 2D sweeps containing the expert input or output layer lr multiplier suggest that smaller values would perform better,  but the joint change of (lr router, lr out) dominates and does not allow a third change. Similarly in SGD $\mu$P Regime II (center) the pair (lr router, lr expert2) marginally dominates other improvements such as even smaller expert output learning rate. Hence, after such a 2D sweep, it remains unclear whether the new expert output learning rate, and all other HPs, are indeed robustly near-optimal, or whether higher-order interactions continue to induce an indefinite 2D update without ever converging to a local optimum. In Adam MSSP Regime I (right), (lr router, lr in) are updated and it remains unclear whether the global optimum would also include a reduced expert output lr multiplier.

\subsection{Global Adam \texorpdfstring{$\eps$}{epsilon} induces width dependence at sufficient scale}\label{sec:global_eps}

\begin{figure}[H]                            
        \centering
        \begin{subfigure}[b]{0.24\textwidth}
        \centering
        \includegraphics[width=\textwidth]{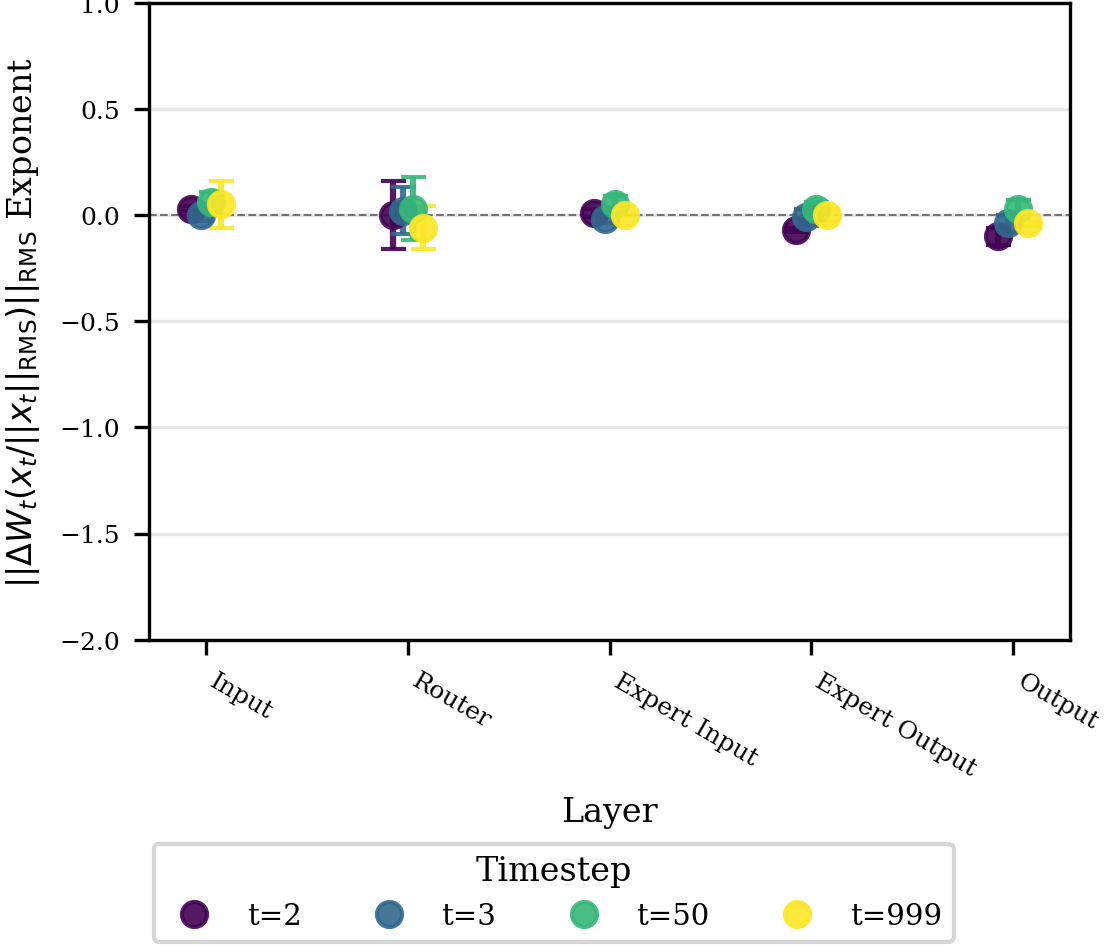}
        \end{subfigure}
        \hfill
        \begin{subfigure}[b]{0.24\textwidth}
        \centering
        \includegraphics[width=\textwidth]{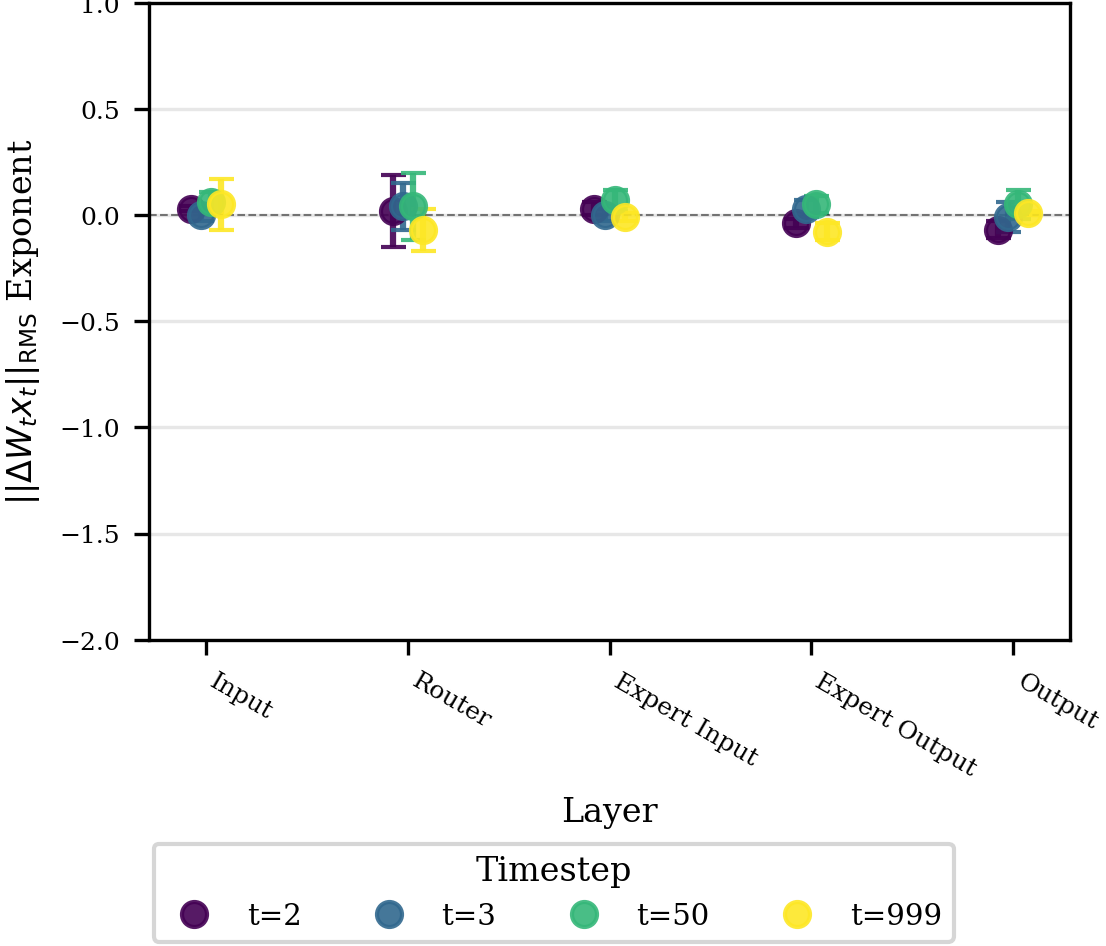}
        \end{subfigure}
        \hfill
        \begin{subfigure}[b]{0.24\textwidth}
        \centering
        \includegraphics[width=\textwidth]{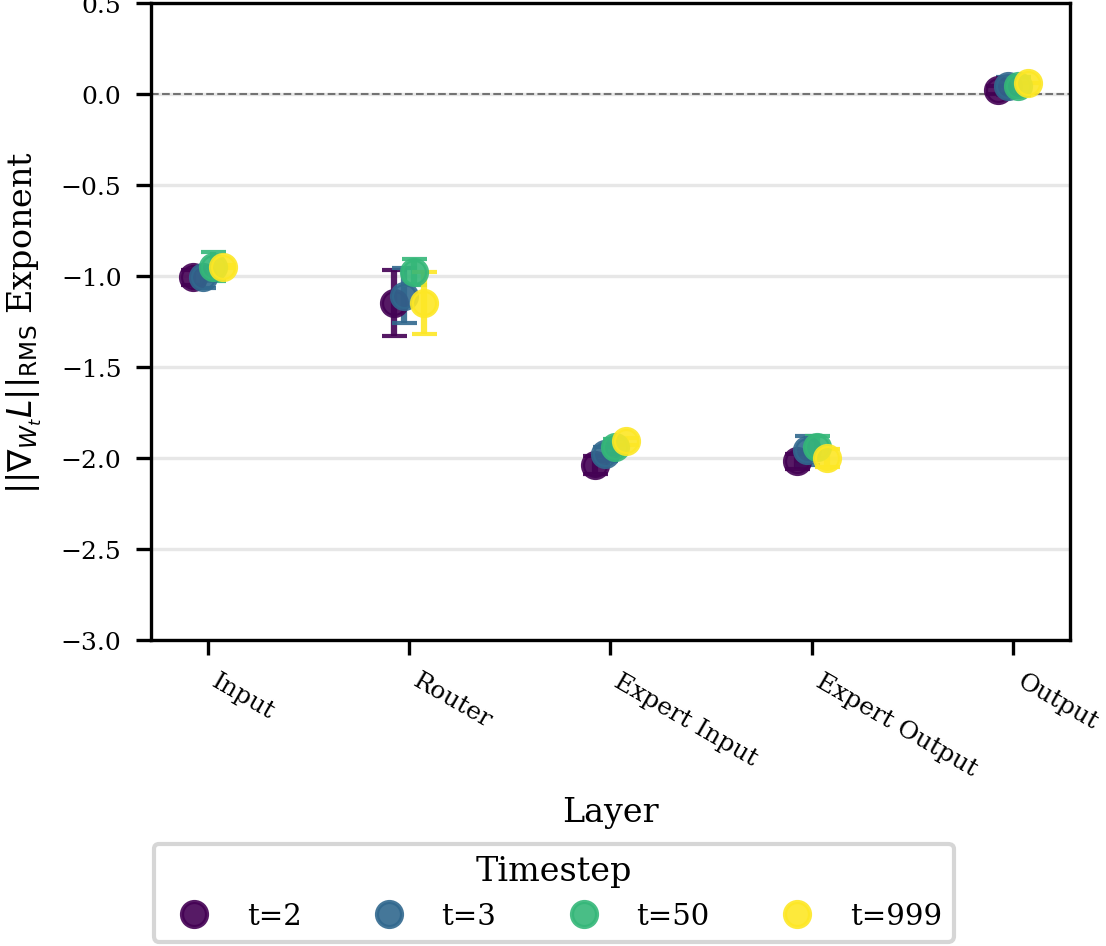}
        \end{subfigure}
        \hfill
        \begin{subfigure}[b]{0.24\textwidth}
        \centering
        \includegraphics[width=\textwidth]{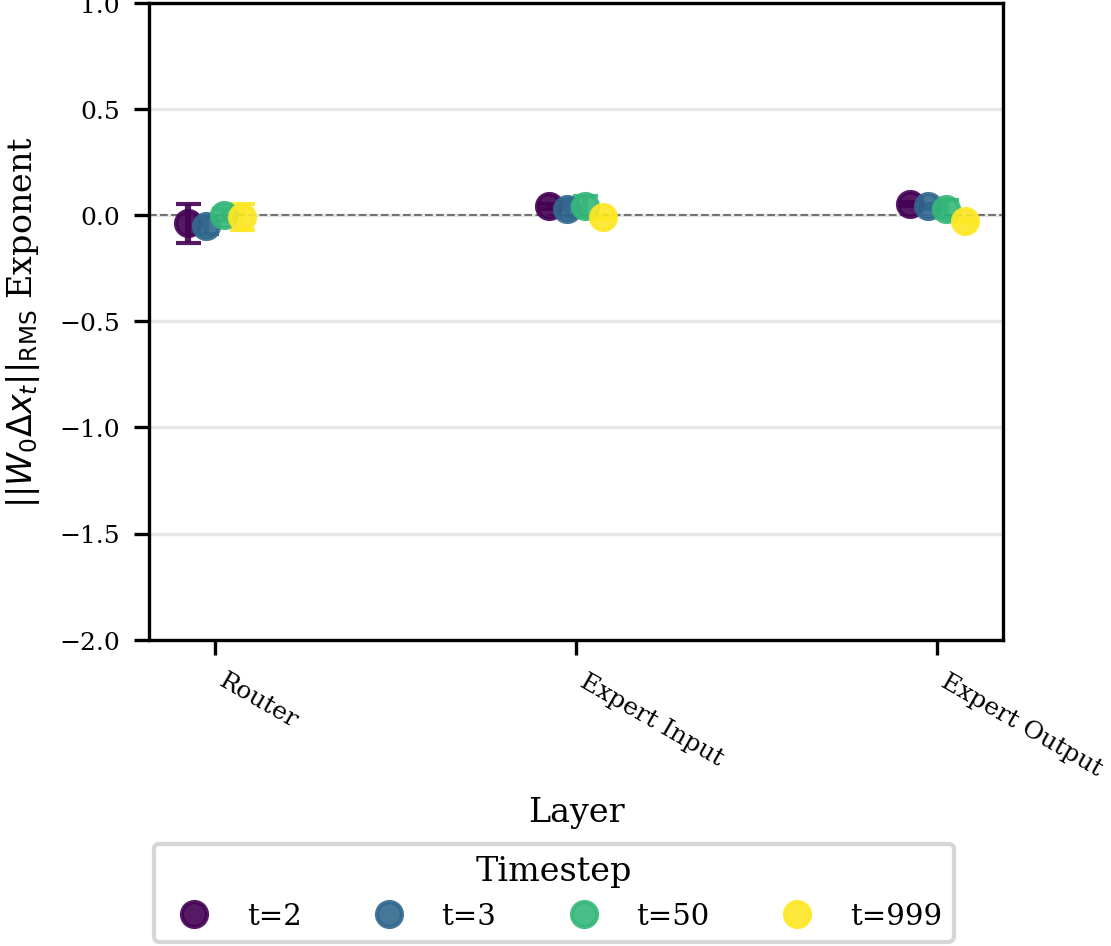}
        \end{subfigure}

        \vspace{0.3em}

        \begin{subfigure}[b]{0.24\textwidth}
        \centering
        \includegraphics[width=\textwidth]{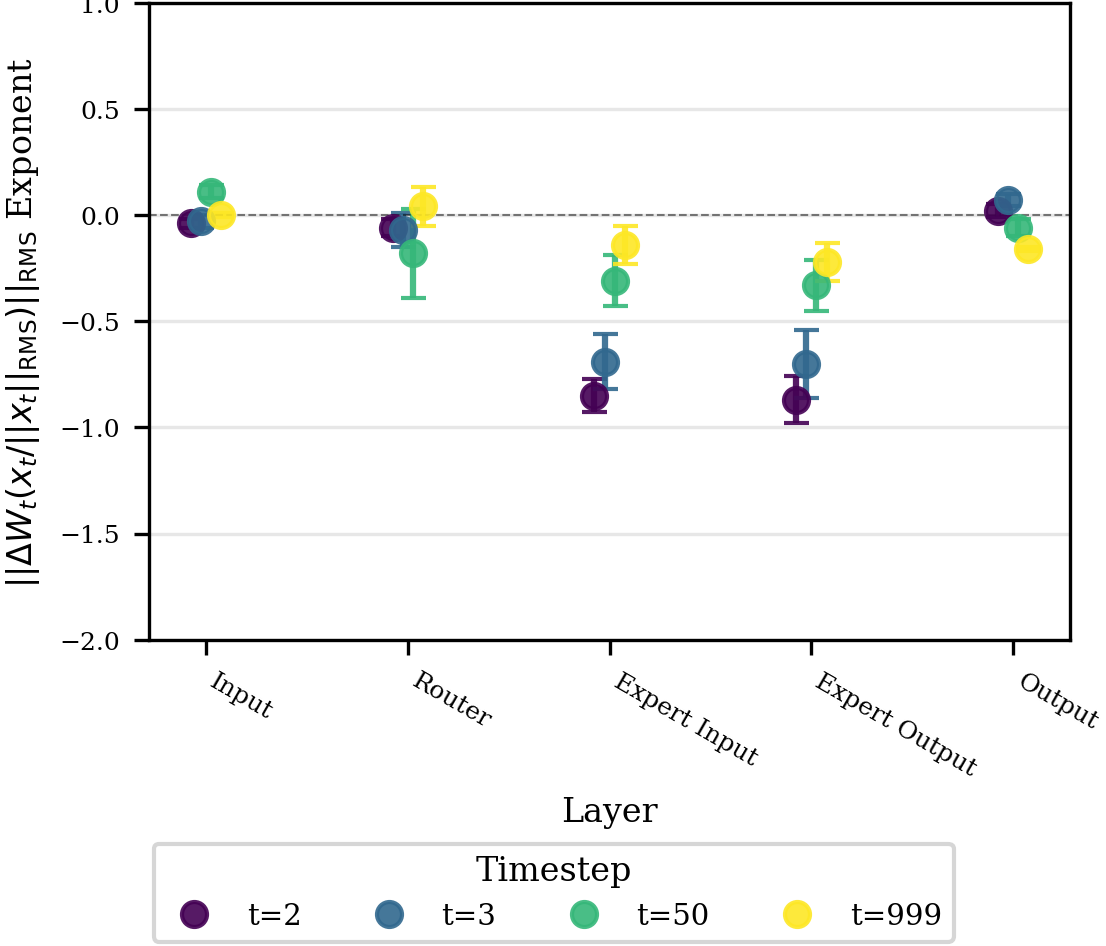}
        \end{subfigure}
        \hfill
        \begin{subfigure}[b]{0.24\textwidth}
        \centering
        \includegraphics[width=\textwidth]{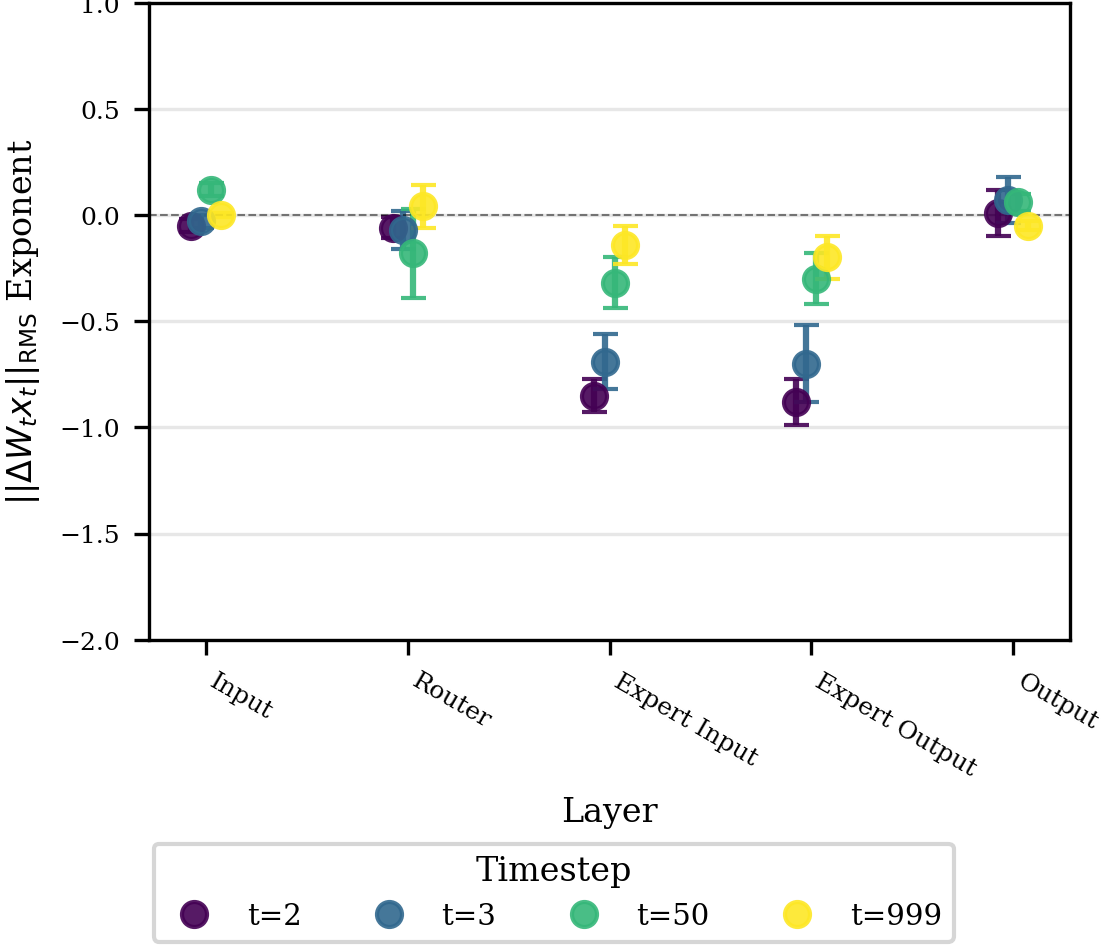}
        \end{subfigure}
        \hfill
        \begin{subfigure}[b]{0.24\textwidth}
        \centering
        \includegraphics[width=\textwidth]{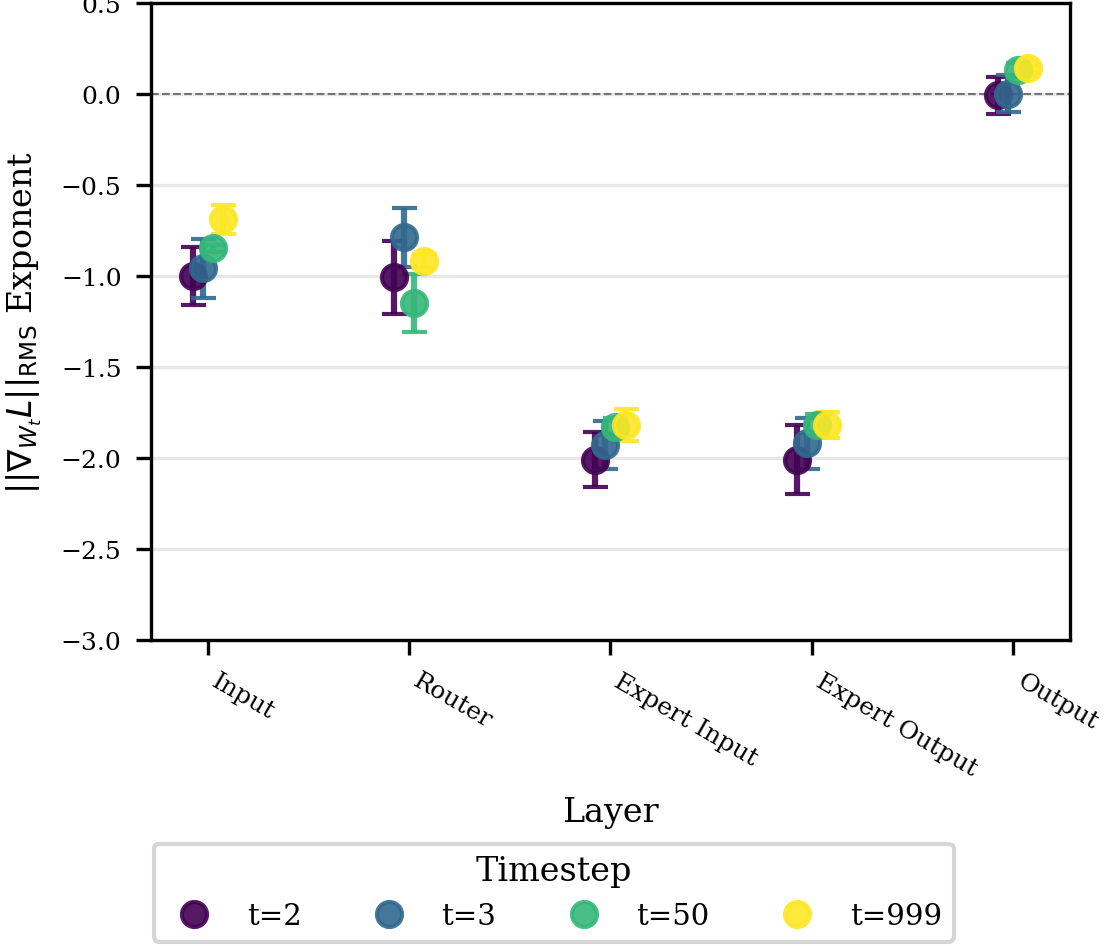}
        \end{subfigure}
        \hfill
        \begin{subfigure}[b]{0.24\textwidth}
        \centering
        \includegraphics[width=\textwidth]{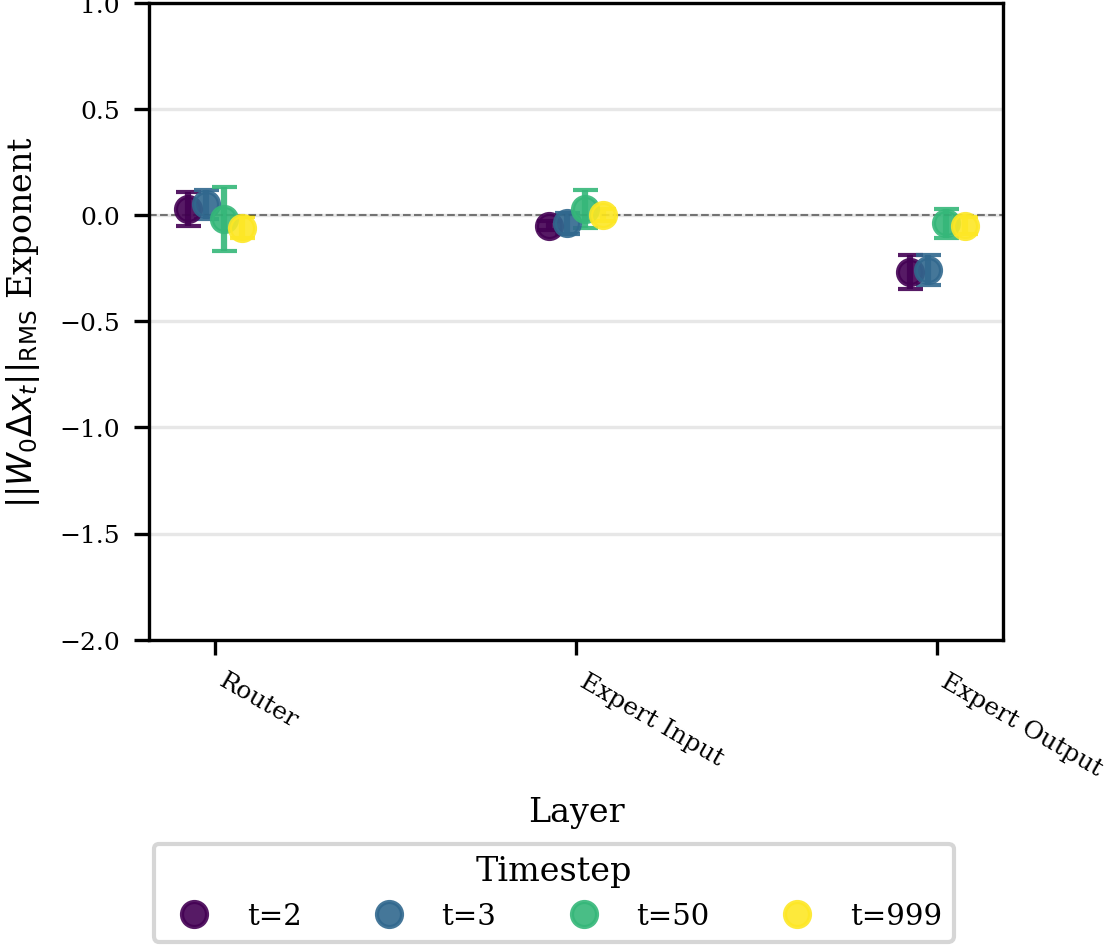}
        \end{subfigure}    

    \caption{\textbf{Global Adam $\eps$ induces entangled exponents.} RCC exponents of Adam MSSP in Regime III with layerwise $\eps$ scaling (Ours, top), the same scaling rule with constant global $\eps$ (bottom). Columns (left to right): Effective Updates (Normalized), Effective Updates (Raw), Gradient Norms (Raw), Propagating Updates (Raw).}
        \label{fig:adam_eps1e-5}
    \end{figure}

\begin{figure}[H]
    \centering
    \begin{subfigure}[b]{0.19\textwidth}
    \centering
    \includegraphics[width=\textwidth]{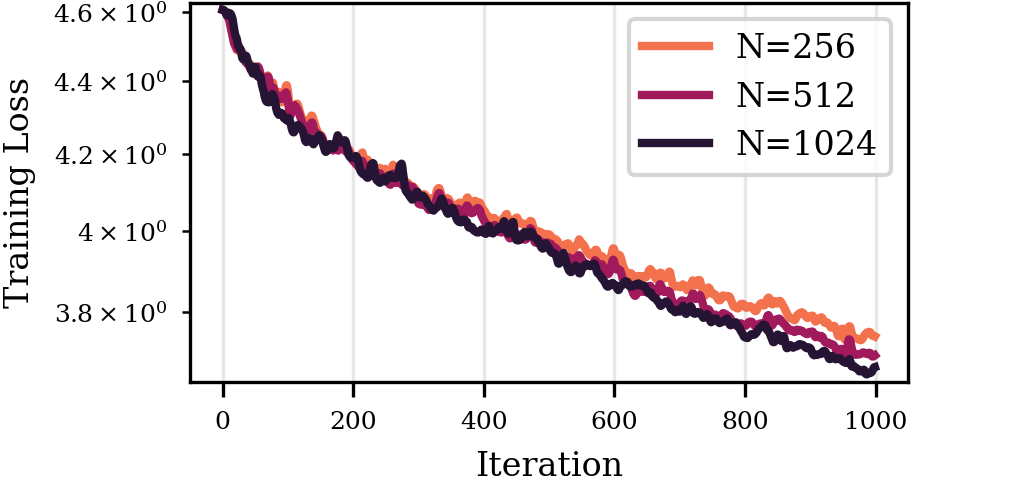}
    \end{subfigure}
    \hfill
    \begin{subfigure}[b]{0.19\textwidth}
    \centering
    \includegraphics[width=\textwidth]{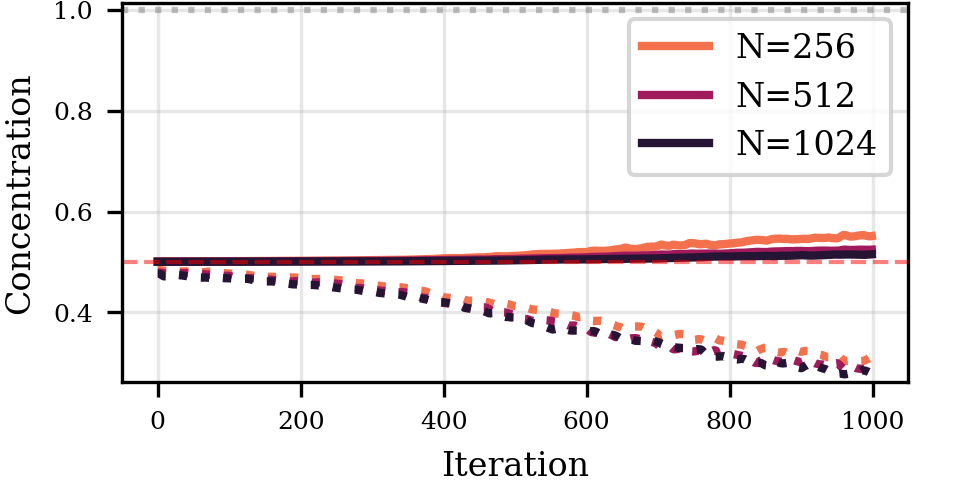}
    \end{subfigure}
    \hfill
    \begin{subfigure}[b]{0.19\textwidth}
    \centering
    \includegraphics[width=\textwidth]{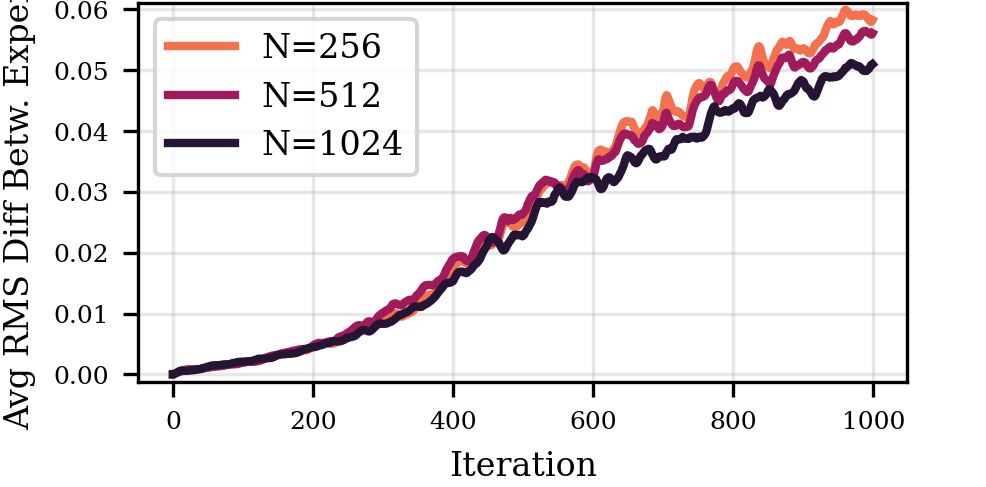}
    \end{subfigure}
    \hfill
    \begin{subfigure}[b]{0.19\textwidth}
    \centering
    \includegraphics[width=\textwidth]{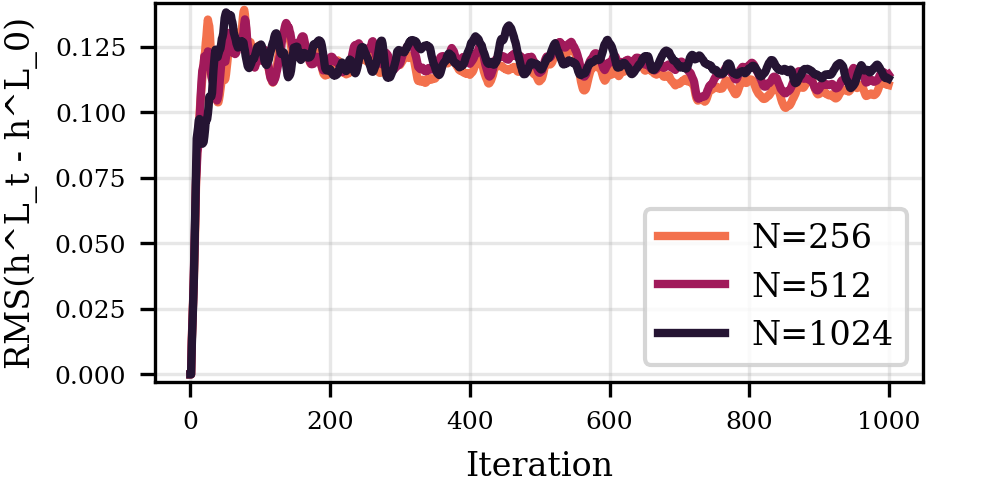}
    \end{subfigure}
    \hfill
    \begin{subfigure}[b]{0.19\textwidth}
    \centering
    \includegraphics[width=\textwidth]{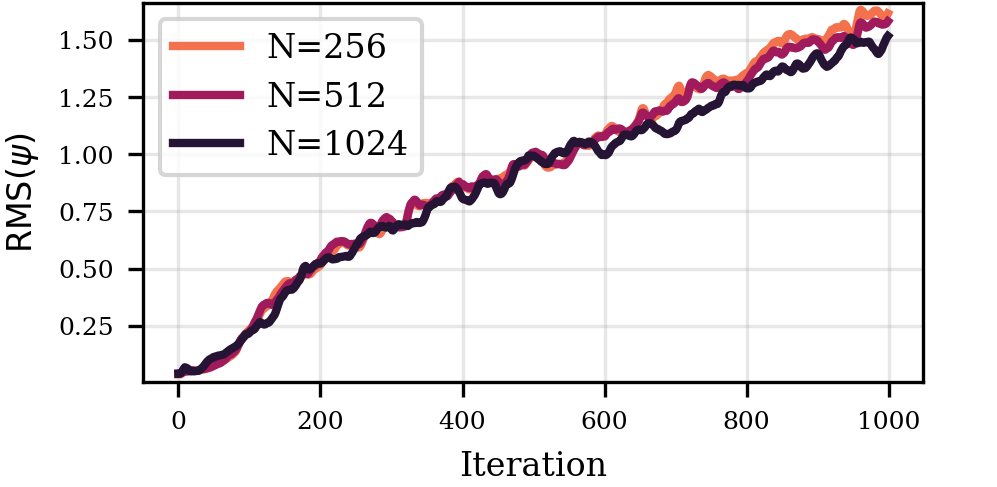}
    \end{subfigure}

    \vspace{0.3em}

    \begin{subfigure}[b]{0.19\textwidth}
    \centering
    \includegraphics[width=\textwidth]{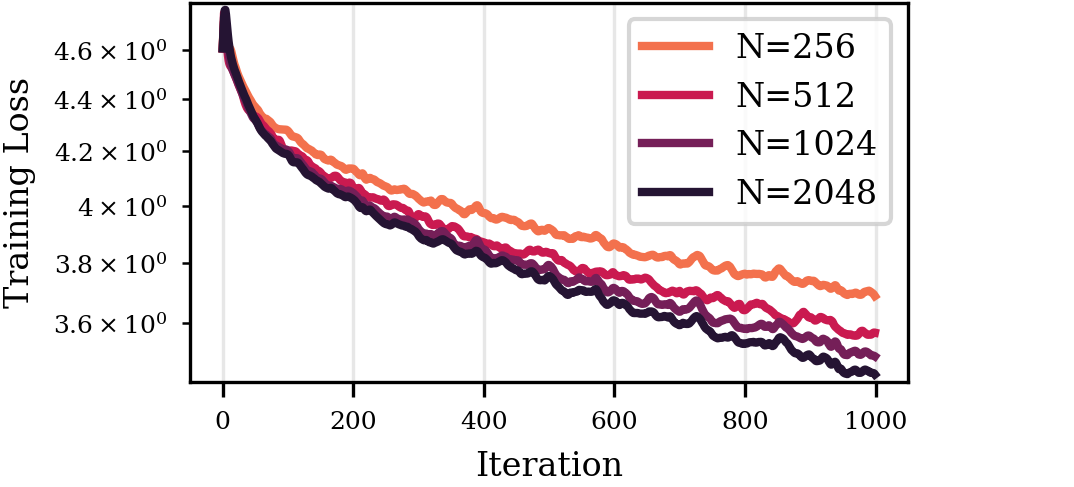}
    \end{subfigure}
    \hfill
    \begin{subfigure}[b]{0.19\textwidth}
    \centering
    \includegraphics[width=\textwidth]{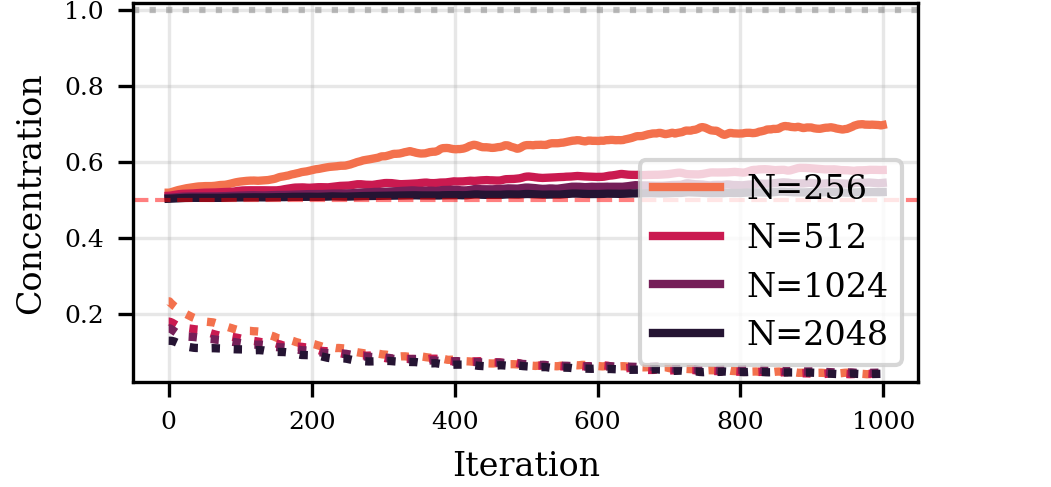}
    \end{subfigure}
    \hfill
    \begin{subfigure}[b]{0.19\textwidth}
    \centering
    \includegraphics[width=\textwidth]{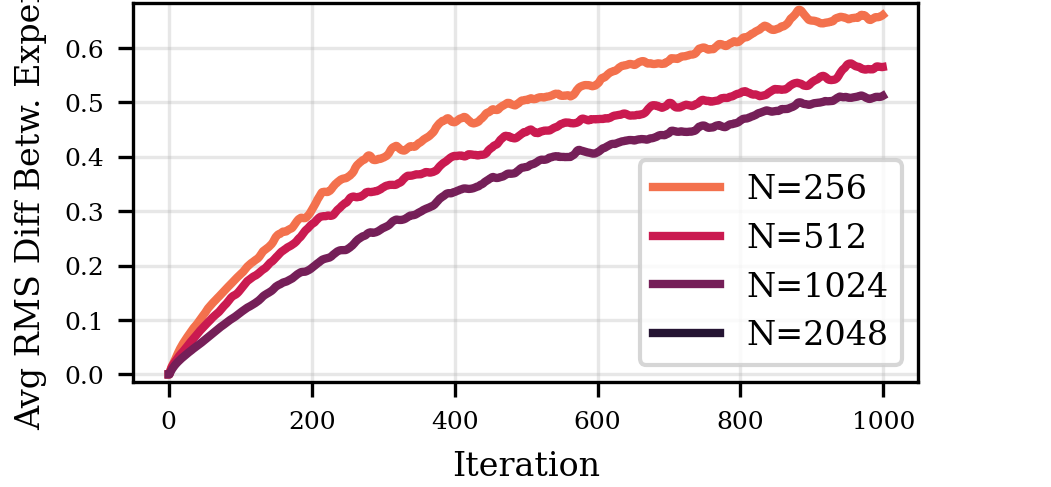}
    \end{subfigure}
    \hfill
    \begin{subfigure}[b]{0.19\textwidth}
    \centering
    \includegraphics[width=\textwidth]{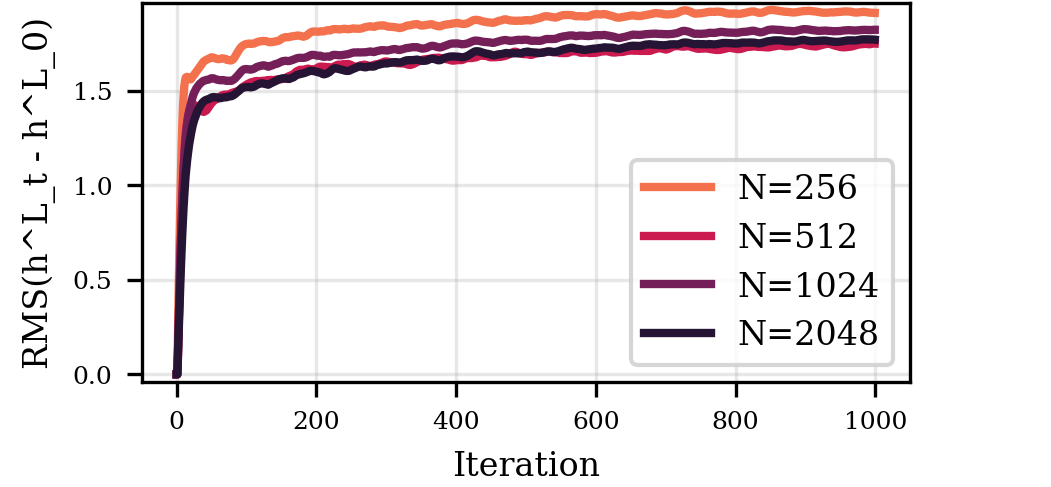}
    \end{subfigure}
    \hfill
    \begin{subfigure}[b]{0.19\textwidth}
    \centering
    \includegraphics[width=\textwidth]{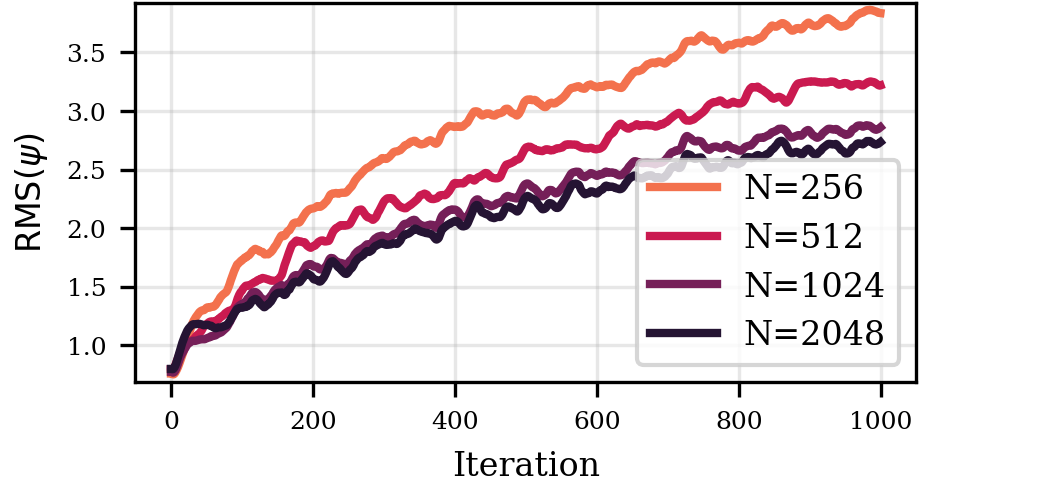}
    \end{subfigure}

\caption{\textbf{Reduced feature learning and router learning under global Adam $\eps$.} Adam MSSP in Regime III with layerwise $\eps$ scaling (top) versus the same scaling rule with constant global Adam $\eps$ (bottom). Columns (left to right): Training loss, router concentration (0.5 indicates uniform routing, solid lines denote the maximal and dotted lines the minimal routing weight in relation to uniform routing), average RMS norm difference between experts, accumulated feature learning ($\|\Delta h^L\|_{RMS})$, router logit norm $\|\psi\|_{RMS}$.}
    \label{fig:adam_eps1e-5_stats}
\end{figure}

Here, we mimick the effects of further scaling -- which we cannot run due to compute constraints -- by increasing Adam $\eps$, since gradient RMS norms decay with model scale. We compare naive constant Adam $\eps$ versus our layerwise Adam $\eps$ scaling under allscaling and layerwise LR multipliers tuned at width 128. \Cref{fig:adam_eps1e-5} shows more time-dependent and less clean effective and propagating update exponents under global $\eps$, especially of expert effective updates. \Cref{fig:adam_eps1e-5_stats} shows feature learning and router learning is reduced with scale under global epsilon, as updates are beginning to vanish.

\end{appendices}

\end{document}